\documentclass[12pt,oneside]{book}
\usepackage[utf8]{inputenc}

\usepackage[top=1.5in, bottom=1.25in, hmargin=1in]{geometry}

\usepackage[hidelinks]{hyperref}
\usepackage{bm}
\usepackage{graphicx}
\usepackage{float}
\usepackage[round]{natbib}
\usepackage{tcolorbox}
\usepackage{multicol}
\usepackage{bookmark}

\usepackage{booktabs}
\usepackage{amsmath}
\usepackage{amssymb}
\usepackage{multirow}
\usepackage{multicol}

\usepackage{tikz}
\usepackage{adjustbox}
\usepackage{textcomp}
\usepackage{xcolor}

\usepackage{booktabs}
\usepackage{mathtools}
\usepackage{setspace}
\usepackage{bbm}
\usepackage{placeins}
\usepackage{enumitem}

\usepackage{caption}
\usepackage{subcaption}
\usepackage{wrapfig}
\usepackage[ruled]{algorithm2e}
\usepackage{setspace}
\usepackage{cleveref}
\usepackage{nomencl}

\usepackage{fancyhdr}

\usepackage{titletoc}
\usepackage{tocloft,calc}

\SetCommentSty{mycommfont}
\newcommand{\bb}{\mathbf}  
\SetKwFor{For}{for (}{) $\lbrace$}{$\rbrace$}

\AtBeginDocument{\addtolength\cftchapnumwidth{\widthof{\bfseries Chapter }}}

\newtheorem{definition}{\textbf{Definition}}
\newtheorem{theorem}{\textbf{Theorem}}

\SetKwInput{KwInit}{Initialise}
\SetKwInput{KwOffline}{Offline}

\makenomenclature
\usepackage{etoolbox}
\renewcommand\nomgroup[1]{%
  \item[\bfseries
  \ifstrequal{#1}{A}{General notation }{%
  \ifstrequal{#1}{B}{Linear algebra }{%
  \ifstrequal{#1}{C}{Probability theory }{%
  \ifstrequal{#1}{D}{Machine learning }{%
  \ifstrequal{#1}{E}{Motion trajectories and control }{}}}}}%
]}

\newcommand\blfootnote[1]{%
  \begingroup
  \renewcommand\thefootnote{}\footnote{#1}%
  \addtocounter{footnote}{-1}%
  \endgroup
}

\begin{document}
\begin{titlepage}
	\centering
	\vspace*{3cm}
	\hrulefill\\
	\vspace{0.5cm}
	{\scshape\LARGE Learning to represent surroundings, anticipate motion and take informed actions in unstructured environments \par}
	\vspace{0.5cm}
	\hrulefill\\
	\vspace{1.2cm}
	{\Large Weiming Zhi\par}
	\vspace{1.5cm}
	{\normalsize  A thesis submitted in fulfilment of the requirements for the degree of\par}
	
	\vspace{0.3cm}
	
	{\normalsize Doctor of Philosophy}

	\vspace{1.2cm}
	\vspace{0.2cm}
	{\normalsize Faculty of Engineering \\ \vspace{4pt}
	University of Sydney \\ \vspace{4pt}
	Australia}
	
	\vspace{1.3cm}
	{\normalsize 2023\par}

\end{titlepage}

\frontmatter
\begin{centering}
\chapter{Declaration}
\end{centering}
I hereby declare that this submission is my own work and that, to the best of my
knowledge and belief, it contains no material previously published or written by
another person nor material which to a substantial extent has been accepted for
the award of any other degree or diploma of the University or other institute of
higher learning, except where due acknowledgement has been made in the text. 

\begin{flushright}
\vspace{50pt}
\textbf{Weiming Zhi}\\
 January 2023
\end{flushright}

\pagebreak

\begin{centering}
\chapter{Author Attribution and Publications}
\end{centering}
Elements of this thesis have appeared in peer-reviewed conferences and journal publications. I am the first author of each of these papers. My contributions include, but are not limited to, conceptualising the method, developing the theory, designing and conducting the experiments, and writing the paper. The specific papers are listed here:
\begin{enumerate}
    \item \Cref{chap3} has appeared as: \citep{HM} \textbf{W. Zhi}, L. Ott, R. Senanayake, F. Ramos. Continuous Occupancy Map Fusion with Fast Bayesian Hilbert Maps. \textit{International Conference on Robotics and Automation (ICRA)}, 2019. 
    \item \Cref{chap4} has appeared as: \citep{sptemp} \textbf{W. Zhi}, R. Senanayake, L. Ott, F. Ramos. Spatiotemporal Learning of Directional Uncertainty in Urban Environments. \textit{IEEE Robotics and Automation Letters (RA-L)}, 2019. 
    \item \Cref{chap5} has appeared as: \citep{SPAN_nav} \textbf{W. Zhi}, T. Lai, L. Ott, F. Ramos. Anticipatory Navigation in Crowds by Probabilistic Prediction of Pedestrian Future Movements.  \textit{International Conference on Robotics and Automation (ICRA)}, 2021. 
    \item \Cref{chap5b} has appeared as: \citep{OTNet} \textbf{W. Zhi}, Tin Lai, L. Ott, F. Ramos. Trajectory Generation in New Environments from Past Experiences. \textit{IEEE/RSJ International Conference on Intelligent Robots and Systems (IROS)}, 2021.
    \item \Cref{chap6} has appeared as: \citep{Prob_struct_const} \textbf{W. Zhi}, L. Ott, F. Ramos. Probabilistic Trajectory Prediction with Structural Constraints. \textit{IEEE/RSJ International Conference on Intelligent Robots and Systems (IROS)}, 2021. 
    \item \Cref{chap7} has appeared as: \citep{Diff_templates} \textbf{W. Zhi},  T. Lai, L. Ott, F. Ramos. Diffeomorphic Transforms for Generalised Imitation Learning. \textit{Annual Learning for Dynamics and Control Conference (L4DC)}, 2022. {\color{red} \textbf{This work received the best paper award at L4DC}}.
    \item \Cref{chap8} has been accepted, and will appear as: \citep{GeoFab_gloabL_opt} \textbf{W. Zhi}, K. Van Wyk, I. Akinola, N. Ratliff, F. Ramos. Global and Reactive Motion Generation with Geometric Fabric Command Sequences. \textit{IEEE International Conference on Robotics and Automation (ICRA)}, 2023. To appear
\end{enumerate}

Additionally, the following peer-reviewed papers were produced during the author's PhD, but are not components of this thesis.
\begin{enumerate}[resume]
    \item \citep{Fast_diff_int} \textbf{W. Zhi}, T. Lai, L. Ott, E. V. Bonilla, F. Ramos. Learning Efficient and Robust Ordinary Differential Equations via Invertible Neural Networks. \textit{International Conference on Machine Learning (ICML)}, 2022.
    \item \citep{KTM} \textbf{W. Zhi}, L. Ott, F. Ramos. Kernel Trajectory Maps for Multi-Modal Probabilistic Motion Prediction. \textit{Conference on Robot Learning (CoRL)}, 2019.  
    \item \citep{PDMP} T. Lai, \textbf{W. Zhi}, T. Hermans, F. Ramos. Parallelised Diffeomorphic Sampling-based Motion Planning. \textit{Conference on Robot Learning (CoRL)}, 2021.
\end{enumerate}

In addition to the statements above, in cases where I am not the corresponding author of a published item, permission to include the published material has been granted by the corresponding author.
\begin{flushright}
\vspace{20pt}
\textbf{Weiming Zhi}\\
January 2023
\end{flushright}


\pagebreak

\begin{centering}
\chapter{Abstract}
\end{centering}


Contemporary robots have become exceptionally skilled at achieving specific tasks in structured environments. However, they often fail when faced with the limitless permutations of real-world unstructured environments. This motivates robotics methods which \emph{learn} from experience, rather than follow a pre-defined set of rules. In this thesis, we present a range of learning-based methods aimed at enabling robots, operating in dynamic and unstructured environments, to better understand their surroundings, anticipate the actions of others, and take informed actions accordingly. 

In the first part of the thesis, we investigate methods which leverage learning to represent the structure and motion in a robot's operating environment, in a \emph{continuous} manner. These methods do not impose a fixed-size grid on the environment, and can be queried at arbitrary resolutions. We present Fast Bayesian Hilbert Maps (Fast-BHMs), an efficient continuous representation of environment occupancy, and contribute a fusion algorithm to merge them in a multi-agent setting. We show that Fast-BHMs can be represented more succinctly than discretised grid-cells, motivating its use in multi-agent map-building. Then, we present a continuous and probabilistic spatiotemporal model of motion distributions in the environment. This allows a robot to understand long-term motion patterns in its surroundings, and reason about how things in this environment move. 

Next, in the second part of this thesis, we develop methods for \emph{anticipatory navigation}, where we aim to endow robots with the ability to predict the motion of dynamic obstacles in the vicinity, and to take decisions which account for these predictions. We contribute \emph{Stochastic Process Anticipatory Navigation (SPAN)}, a framework that enables a robot to smoothly navigate through crowds. SPAN learns the future motion of nearby moving entities and represents the motions as stochastic processes, and integrates them into a control problem, which is then continuously solved. Then, we investigate the problem of whether we can transfer motion trajectories observed in past environments to novel environments. Humans are able to infer likely motion patterns in an environment, by simply reasoning about the floor plan of the environment. We wish to endow robots with the same ability, and introduce \emph{Occupancy-Conditional Trajectory Network (OTNet)}. OTNet embeds occupancy representations as a vector of similarities with previously observed maps to predict a multi-modal distribution over likely trajectories in the environment. This allows us to generate motion trajectories which match motion patterns observed in past environments. Finally, we contribute a combined learning and optimisation framework for probabilistic motion prediction. Motion prediction methods often lack mechanisms to explicitly account for the environment structure as constraints. Our framework enables constraints to be applied to the predictions, such that they remained compliant with the structure of the environment. 

Finally, in the third part of this thesis, we examine methods to generate motions for robot manipulators in unstructured environments. We present the \emph{Diffeomorphic Template (DT)} framework for generalised imitation learning. Generalised imitation learning seeks to teach a robot, by providing a small set of expert demonstrations, to acquire novel skills, and then generalise the skills when circumstances change. Individual DTs modularise the robot's motion, each imbuing a different behaviour, such as imitating demonstrations or avoiding new obstacles. Multiple DTs can then be combined to generate generalised behaviour. Next, we contribute the Geometric Fabrics Command Sequence (GFCS) method to generate motion trajectories which are both reactive and global. GFCS builds upon Geometric Fabrics, a reactive but local method. The local nature often results in the robot becoming stuck in local minima, unable to reach the designated goals. GFCS formulates an optimisation problem, over parameters of multiple Geometric Fabrics, which is subsequently solved via global optimisation. To speed up the optimisation, we learn a model that conditions on the problem to generate solutions to warm-start the optimiser. 

\thispagestyle{empty}

\pagebreak

\begin{centering}
\chapter{Acknowledgements} 
\end{centering}
This thesis was made possible by the kind support of many others, throughout this adventure.

Foremost, my gratitude goes to my supervisor Fabio Ramos. I would like to thank Fabio for not only providing me with valuable research guidance and unwavering support, but also imparting me with the knowledge of how to become a well-rounded researcher. I consider myself extremely fortunate to pursue my PhD in Fabio's lab. I am also immensely grateful to Lionel Ott, who though not listed on paper as a supervisor, acted very much as my secondary supervisor. He enlightened me with his depth of technical expertise and helped me develop foundational research skills. I would also like to express my gratitude to Edwin Bonilla, who also mentored me closely, tremendously honing my research abilities, and shaping my perspective on machine learning research.  

My warmest thanks go to the many lab mates I met throughout my PhD. Many ideas came to fruition alongside coffee chats next to our beloved espresso machine. In particular, I would like to acknowledge the oversized impacts Ransalu Senanayake and Tin lai have played in my PhD journey. As a senior PhD student, Ransalu took me under his wing from the first day of my PhD, and walked me through the intricacies of robotics research. Tin is my serial co-author and force multiplier, who never fails to bring fresh perspectives and spark new research. I deeply enjoyed my time remotely interning with NVIDIA's robotics research lab, led by Dieter Fox. Along with Fabio, I would like to thank my mentors at NVIDIA --- Karl Van Wyk, Iretiayo Akinola and Nathan Ratliff. I was also greatly fortunate to collaborate with Tucker Hermans on sampling-based motion planning research.

I am also grateful to Nikita for her indispensable support and patience, and also for keeping me sane during Covid-19 lockdowns. Last but not least, none of this would be possible without the support of my parents and siblings, who have made sacrifices for me to receive the highest quality education, whenever possible --- they have my sincerest gratitude.

\phantomsection
\addcontentsline{toc}{chapter}{Nomenclature}

\nomenclature[A,011]{$f(\cdot)$, $g(\cdot)$, $h(\cdot)$}{Functions}
\nomenclature[A,012]{$i$, $j$, $k$}{Indices}
\nomenclature[A,013]{$\approx$}{Approximately equal to}
\nomenclature[A,014]{$>>$}{Much greater than}
\nomenclature[A,02]{$\lvert\lvert \cdot \lvert\lvert_{p}$}{p-norm}
\nomenclature[A,03]{$\lvert\cdot \lvert$}{Absolute value of a scalar or cardinality of a set}
\nomenclature[A,04]{$\{\cdot\}$}{A set}
\nomenclature[A,05]{$\min_{x}(\cdot), \max_{x}(\cdot)$}{Minimise, maximise with respect to $x$}
\nomenclature[A,06]{$\arg\min_{x}(\cdot), \arg\max_{x}(\cdot)$}{Minimise, maximise with respect to $x$ and return the argument}
\nomenclature[A,07]{$\mathbbm{1}$}{Indicator function}

\nomenclature[A,31]{$\mathbb{R}$}{Set of real numbers}
\nomenclature[A,33]{$\mathbb{R}^{m \times n}$}{Set of $m$ by $n$ matrices over the reals}
\nomenclature[A,34]{$\mathbb{R}^{m \times m}_{++}$}{Set of $m$ by $m$ positive definite matrices over the reals}
\nomenclature[A,35]{$\mathbb{N}$}{Set of natural numbers}

\nomenclature[B,01]{$x$}{Scalar}
\nomenclature[B,02]{$\mathbf{x}$}{Vector}
\nomenclature[B,03]{$\mathbf{x}^{\top}$}{Transpose of $\mathbf{x}$}
\nomenclature[B,05]{$\mathbf{X}$}{Matrix}
\nomenclature[B,06]{$\mathbf{X}^{-1}$}{Inverse of matrix $\mathbf{X}$}
\nomenclature[B,07]{$\lvert\mathbf{X}\lvert$}{Determinant of matrix $\mathbf{X}$}
\nomenclature[B,08]{$\mathbf{X}^{\dagger}$}{Generalised inverse matrix $\mathbf{X}$}
\nomenclature[B,09]{$\mathbf{I}$}{Identity matrix}
\nomenclature[B,10]{$\mathrm{vec}(\mathbf{X})$}{Vectorise matrix $\mathbf{X}$}
\nomenclature[B,11]{$\mathrm{tr}(\mathbf{X})$}{Trace of matrix $\mathbf{X}$}
\nomenclature[B,12]{$\mathrm{diag}(\mathbf{X})$}{Diagonal of matrix $\mathbf{X}$}
\nomenclature[B,13]{$\mathrm{LowerTrig}(\mathbf{X})$}{Lower triangular matrix of matrix $\mathbf{X}$}

\nomenclature[C,01]{$p(A)$}{Probability of event $A$}
\nomenclature[C,02]{$p(A|B)$}{Probability of event $A$ conditional on $B$}
\nomenclature[C,03]{$p(A,B)$, $p(A \bigcap B)$}{Probability of events $A$ and $B$}
\nomenclature[C,04]{$\bm{\mu}$, $\mathbf{m}$}{Mean vector}
\nomenclature[C,05]{$\bm{\Sigma}$}{Covariance matrix}
\nomenclature[C,06]{$\sim$}{Distributed as}
\nomenclature[C,07]{$\mathcal{N}$}{Normal distribution}
\nomenclature[C,071]{$\mathcal{U}$}{Uniform distribution}
\nomenclature[C,08]{$MN$}{Matrix normal distribution}
\nomenclature[C,09]{$VM$}{von-Mises distribution}
\nomenclature[C,10]{$erf$}{Error function}
\nomenclature[C,11]{i.i.d}{Independent and identically distributed}
\nomenclature[C,111]{$\&$}{Conflation}
\nomenclature[C,112]{$D_{KL}(\cdot\lvert\lvert \cdot)$}{Kullback–Leibler divergence}
\nomenclature[C,12]{$\mathbb{E}$}{Expected value}
\nomenclature[C,13]{$\varepsilon$}{Noise}

\nomenclature[D,01]{$\mathbf{x}$}{Vector of inputs}
\nomenclature[D,02]{$y$}{Outputs}
\nomenclature[D,03]{$\mathbf{x}^{*}$}{Test point of interest}
\nomenclature[D,031]{$\mathcal{X}$}{Instance space}
\nomenclature[D,032]{$\mathcal{Y}$}{Label space}
\nomenclature[D,04]{$D$}{Dataset}
\nomenclature[D,05]{$\mathbf{w}$, $\mathbf{W}$}{Weight vector, weight matrix}
\nomenclature[D,07]{$\bm{\theta}$}{Parameters}
\nomenclature[D,071]{$\mathcal{L}$}{Loss function}
\nomenclature[D,08]{$\sigma(\cdot)$}{Sigmoid function}
\nomenclature[D,09]{$\bm{\phi}$, $\bm{\Phi}$}{Feature vector, feature matrix}
\nomenclature[D,11]{$\hat{\mathbf{x}}$, $\hat{t}$, $\hat{\mathbf{s}}$}{Inducing points}
\nomenclature[D,12]{$k$}{Kernel function}
\nomenclature[D,121]{$\delta_{H}$}{Hausdorff distance}
\nomenclature[D,122]{$\mathcal{L}$}{Loss function}
\nomenclature[D,13]{$\gamma$, $\ell$}{Length-scale hyper-parameter}
\nomenclature[D,14]{$\lambda$}{Regularisation hyper-parameter}

\nomenclature[E,01]{$\mathcal{Q}$}{Configuration space}
\nomenclature[E,02]{$\mathbf{q}$}{Configuration-space coordinates}
\nomenclature[E,021]{$\dot{\mathbf{q}}$}{Configuration-space velocity}
\nomenclature[E,022]{$\ddot{\mathbf{q}}$}{Configuration-space acceleration}
\nomenclature[E,023]{$\mathbf{q}_{g}$}{Goal configuration}
\nomenclature[E,024]{$\mathbf{x}$}{Task-space positions}
\nomenclature[E,025]{$\mathbf{u}$}{Controls}
\nomenclature[E,03]{$f_{e}$}{Forward kinematics}
\nomenclature[E,031]{$\mathbf{J}_{f}$}{Jacobian of mapping $f$}
\nomenclature[E,04]{$\xi$}{Trajectories}
\nomenclature[E,05]{$\bm{\Xi}$}{Distributions of trajectories}
\nomenclature[E,06]{$\mathbf{M}$}{Manifold}
\nomenclature[E,07]{$\varphi$}{Diffeomorphism}
\nomenclature[E,08]{$\gamma$}{Flow of ODE}


\printnomenclature

\pagebreak

\addcontentsline{toc}{chapter}{\listfigurename}
\tableofcontents
\pagebreak

\phantomsection
\listoffigures

\phantomsection
\addcontentsline{toc}{chapter}{\listtablename}
\listoftables

\mainmatter
\chapter{Introduction}
\pagenumbering{arabic}
\section{Motivations}
Robots have become exceedingly adept at completing specific tasks in controlled conditions, such as in warehouses or manufacturing plants. The introduction of autonomous robots in these limited environments have reduced human fatigue, mitigated safety hazards, and improved execution precision. The increasing visibility of a diverse range of robots give the general public a perception that the ubiquitous deployment of general-purpose robots, capable of operating for long durations in the wild, is just around the corner. The reality is far more sobering -- modern robots are often delicate, vulnerable to unexpected situations, and confined to performing tasks in a narrow domain. In its essence, existing robot systems lack the remarkably human-esque ability to understand the context of the task at hand in its entirety and extrapolate from prior experience to acquire new skills. 

A colossal bounty is placed on solving the open problem of how to develop robotic systems that are more multi-purpose and can conduct a variety of tasks safely in the unstructured world. Solving this challenge would bring about profound changes to human society, paving the way for a range of emerging robotics applications in the service, transport and construction sectors. Envision a future where robots are capable of safely coexisting and collaborating with humans in everyday environments -- helping in the kitchen, walking the dog in a park, delivering parcels through crowds. These technological advancements have the potential to release a torrent of productivity, and provide a strong response to persistent macroeconomic woes brought about by aging populations. 

Operating in varied and dynamics environments, with minimal human intervention, requires robots to handle changes in its vicinity in a natural and predictable manner. At the same time, the unstructured nature of real-world conditions can present the robot with endless permutations of scenarios to handle. In many cases, it becomes infeasible to hand-specify an exhaustive set of predefined rules to capture the variety of conditions which may arise. Advances in the field of machine learning have provided roboticists with novel approaches to leverage the data for generalisation. This thesis presents contributions developed by the author to endow robots with the ability to extract patterns from data, and learn to adapt to its environment from experience. Machine learning can be applied to a diverse range of components throughout the autonomy stack, allowing the robot to learn both \emph{about} and \emph{from} its surroundings. In particular, we explore ideas and develop methods (1) to aid robots in constructing concise representations of its surroundings in multi-agent and dynamic settings; (2) to probabilistically predict the movement patterns of dynamic agents in the environment and act according to the predictions; (3) to learn to generate motion for robot manipulators in complex environments.

\begin{figure}[t]
\centering
\begin{subfigure}{0.24\textwidth}
    \frame{\includegraphics[width=\textwidth]{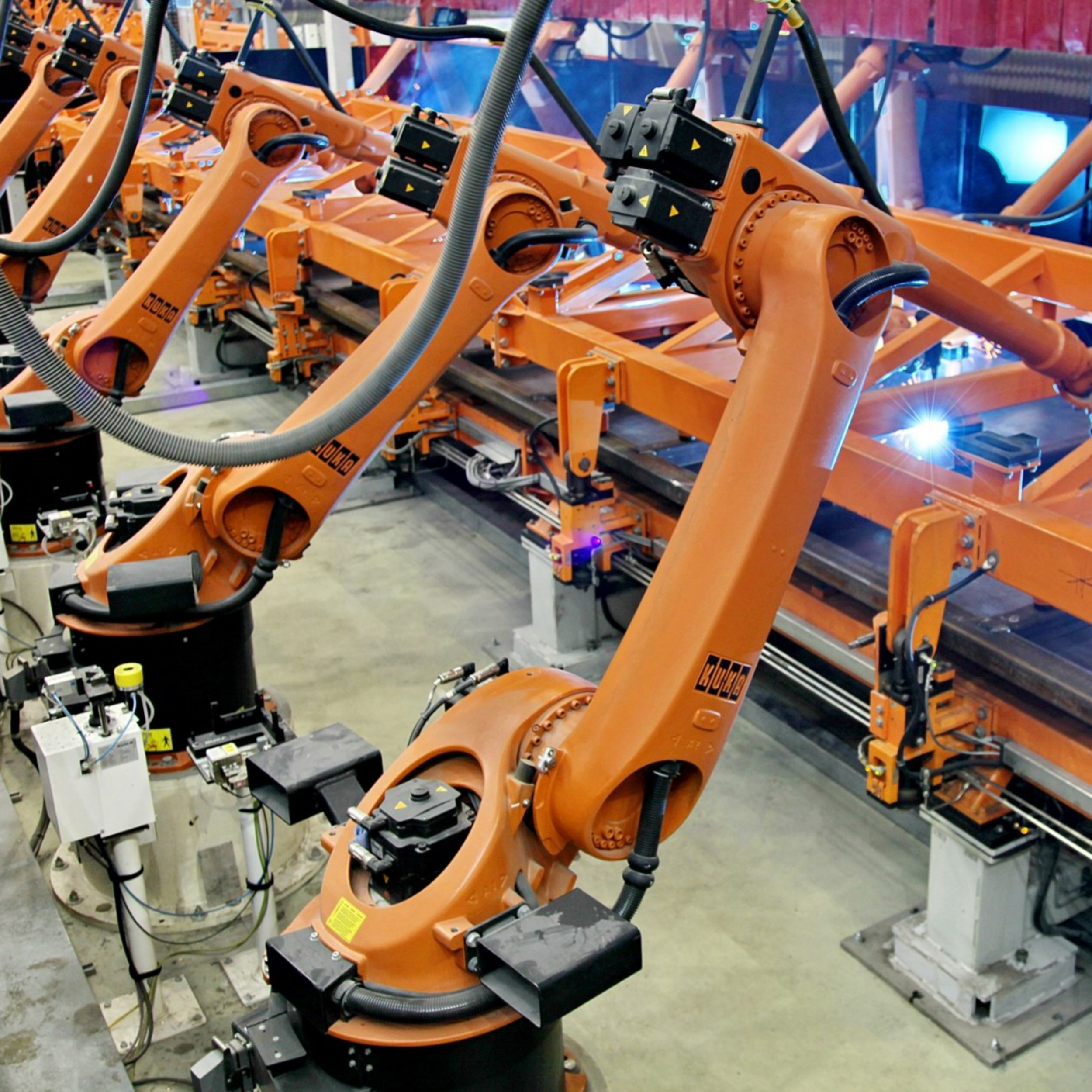}}
    \caption{Robot Manipulator}
\end{subfigure}%
\begin{subfigure}{0.24\textwidth}
    \frame{\includegraphics[width=\textwidth]{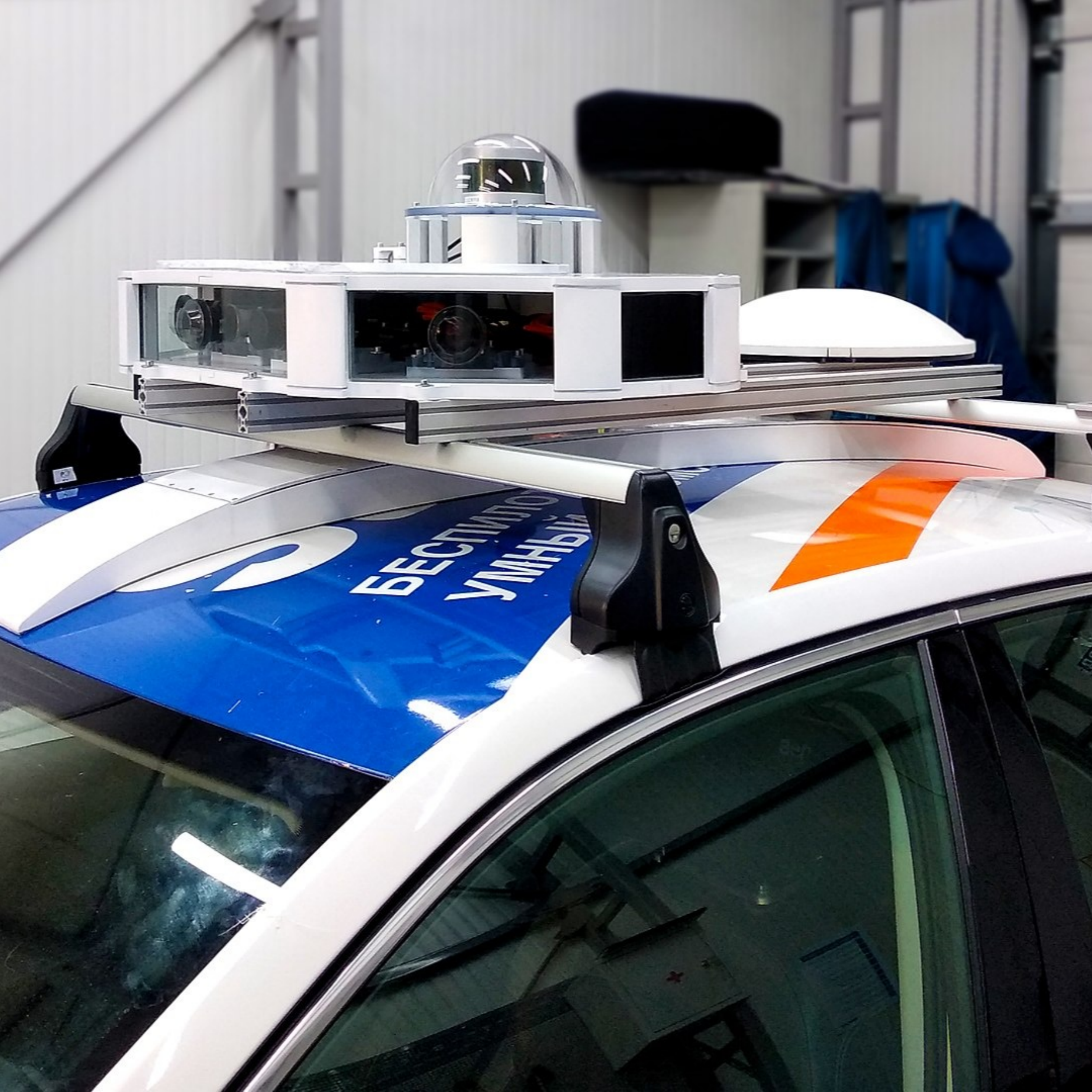}}
    \caption{Autonomous vehicle}
\end{subfigure}%
\begin{subfigure}{0.24\textwidth}
    \frame{\includegraphics[width=\textwidth]{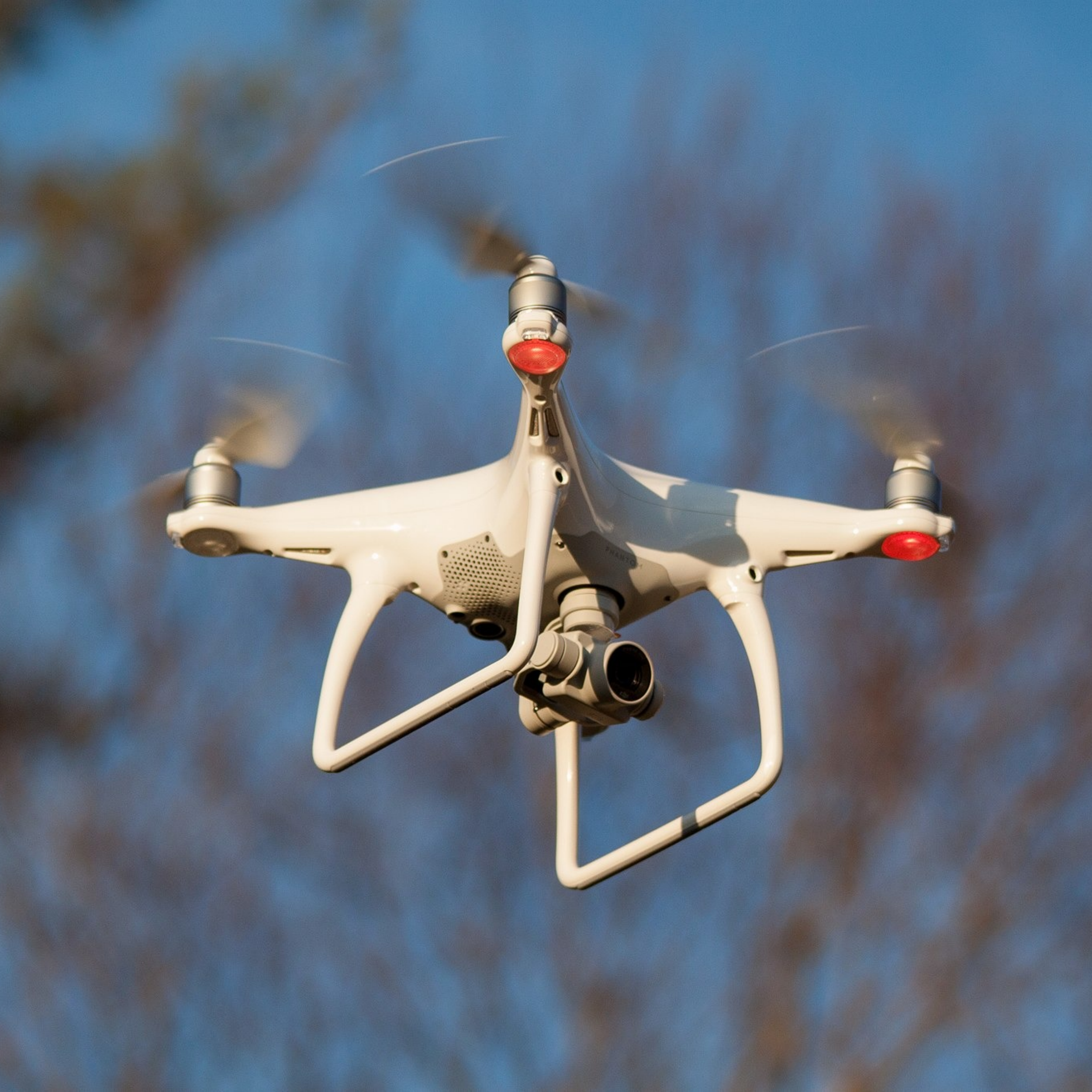}}
    \caption{Quadrotor drone}
\end{subfigure}%
\begin{subfigure}{0.24\textwidth}
    \frame{\includegraphics[width=\textwidth]{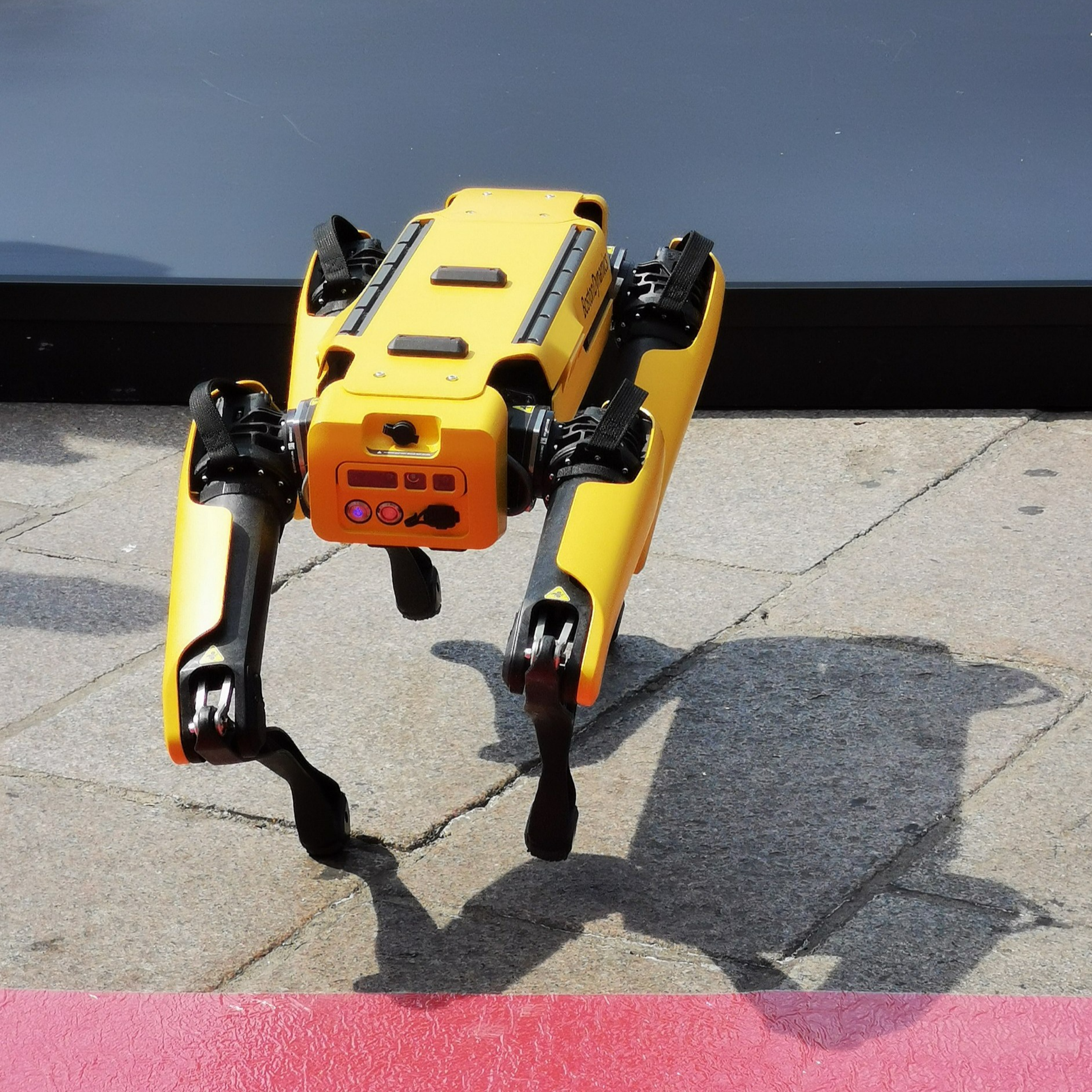}}
    \caption{Quadruped robot}
\end{subfigure}%
\caption{Contemporary robots deployed in their respective environments. Courtesy of Wikimedia Commons.}
\end{figure}

\section{Contributions}
Under the overarching goal of integrating machine learning in robots to perform more general tasks in unstructured environments, we develop robot learning methods to represent the environment, anticipate the movement of others in the vicinity, and generate informed motions. Specifically, in each of the three parts of this thesis, we tackle a distinct research theme and make several contributions within each theme.

\begin{tcolorbox}[colback=white,colframe=red!75!black,title=Highlight figures of Part I: Learning Continuous Environment Representation]
\begin{multicols}{2}
\centering

{\small Chapter 3: Fusion of Continuous Occupancy Maps}

\vspace{1em}
    \includegraphics[width=0.48\textwidth]{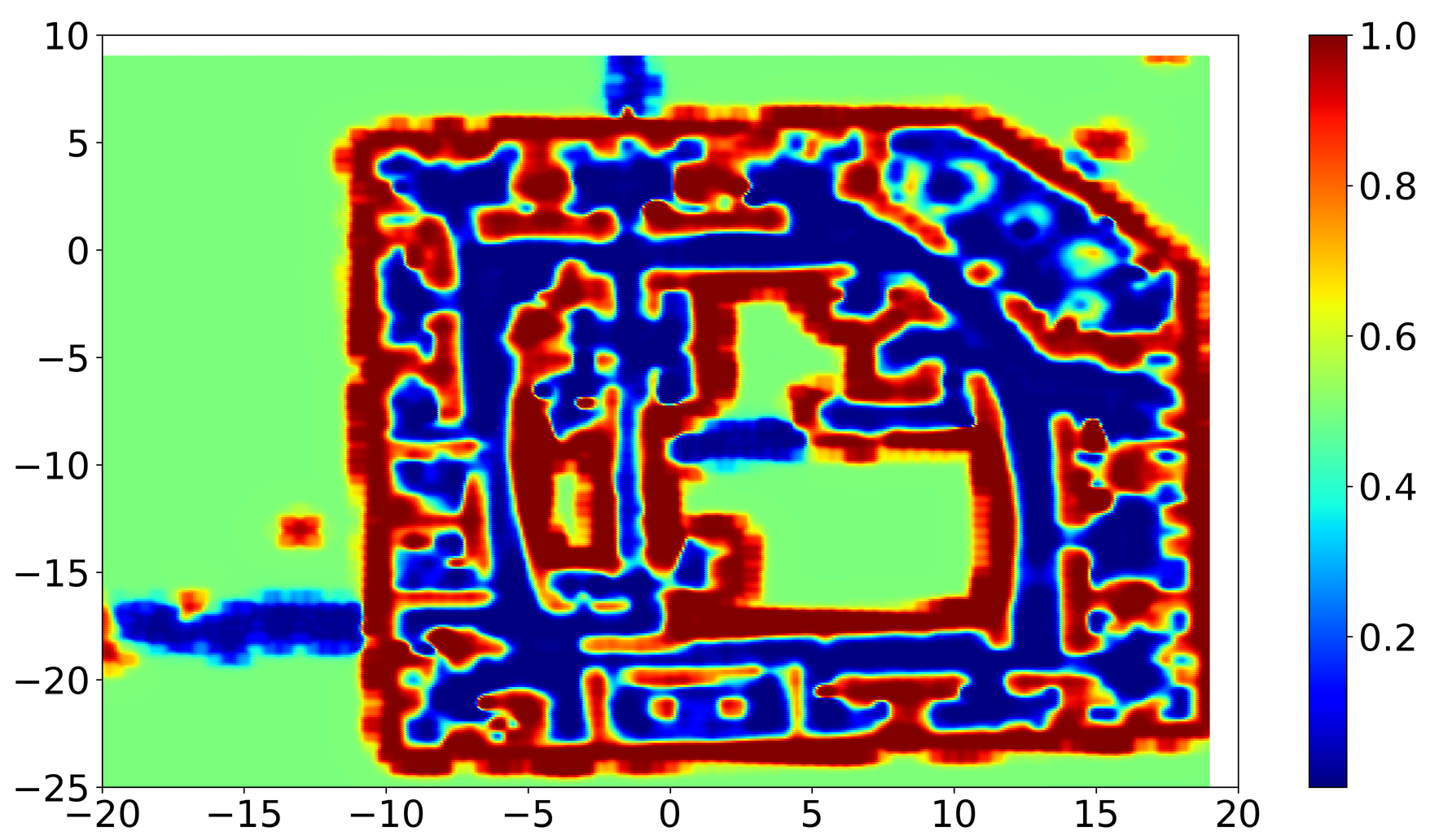}
    
    
\columnbreak

{\small Chapter 4: Continuous Spatiotemporal Maps of Motion Directions}

\centering
\vspace{1em}
\includegraphics[width=0.35\textwidth]{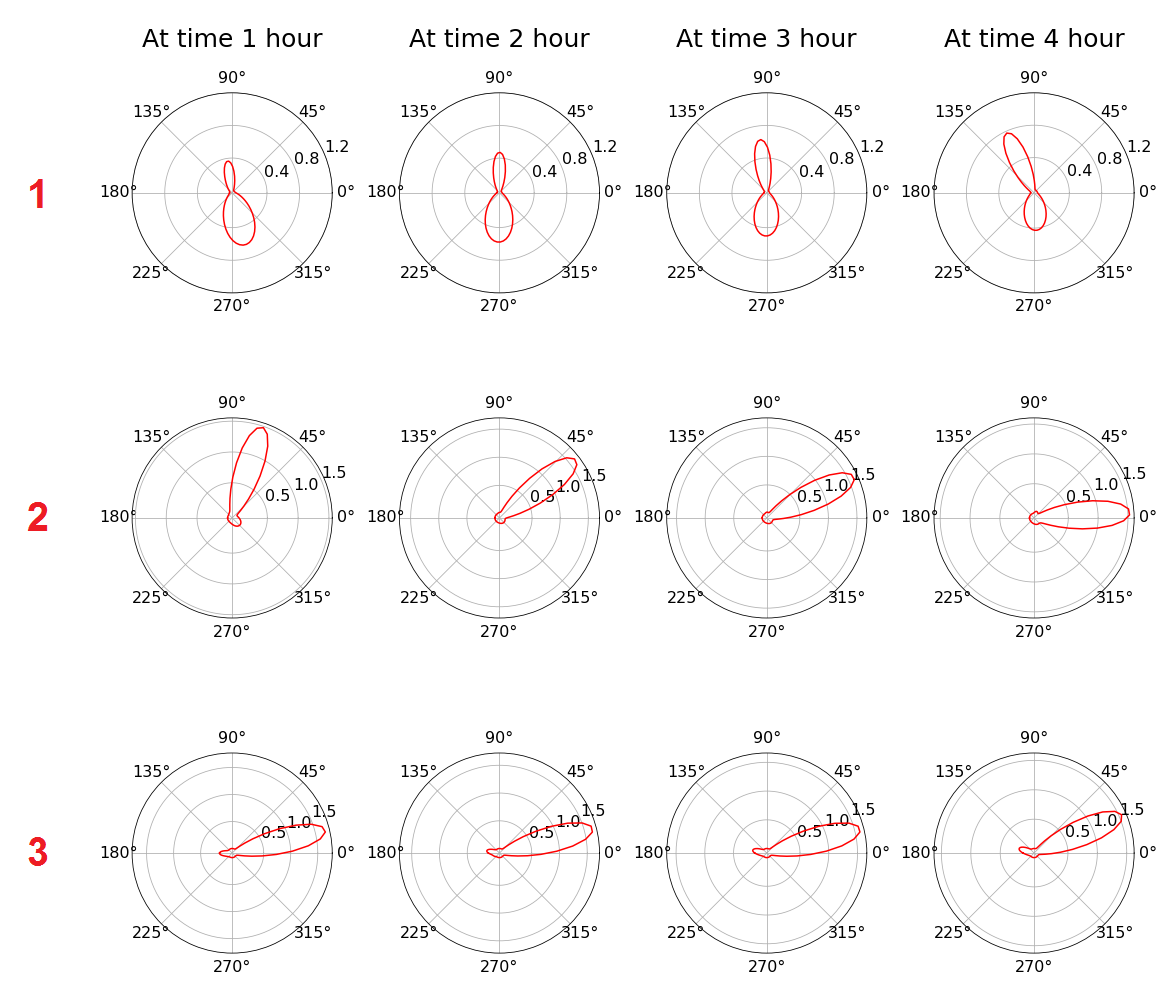}

\end{multicols}
\end{tcolorbox}

\begin{enumerate}

    \item \textbf{Part I: Learning Continuous Environment Representations:} We seek to apply learning to aid robots in representing their perceived surroundings. We develop efficient methods to learn models which represent structure and motion patterns within robots' environments in a continuous manner, particularly in a multi-agent setup or in a dynamic environment. Early methods to represent robots' operating environment sought to enforce a grid structure and compute values for each grid cell. We take an alternative approach, and work with learning-based approaches for robots to continuously represent their surroundings without discretising the inherently continuous real-world. Specifically, we are interested in answering the following question: 
    
    \emph{How do we leverage learning to produce compact representations of occupancy and movement in a multi-agent set-up or a dynamic environment?} 
    
    The specific contributions include: 
    \begin{itemize}
        \item \textbf{Fusion of continuous occupancy representations:} In \cref{chap3}, we introduce the \emph{Fast-BHM} method, an efficient variant of \emph{Bayesian Hilbert Maps} (BHMs) \citep{Senanayke:2017}, to build maps which are continuous and compact. We show that the compact nature of Fast-BHMs makes them ideal to use in bandwidth-limited multi-agent setups. To this end, we develop a fusion algorithm to merge multiple fast-BHMs models maps iteratively, enabling a group of robots to build separate maps, which are then fused in a decentralised manner into an aggregated model.
        
        \item \textbf{Continuous spatiotemporal model of direction distributions:} In \cref{chap4}, we introduce a continuous spatiotemporal model to represent the conditional probability distribution of movement directions of dynamic objects. This distribution is capable of capturing multi-modality and is correctly wrapped around a unit circle. The contributed model allows the robot to understand long-term patterns of how dynamic objects move in the environment. 
    \end{itemize}

\begin{tcolorbox}[colback=white,colframe=green!75!black,title=Highlight figures of Part II: Learning for Anticipatory Navigation]
\begin{multicols}{2}
\centering
{\small Chapter 5: Stochastic Process Anticipatory Navigation}

\vspace{1em}
    \includegraphics[width=0.48\textwidth]{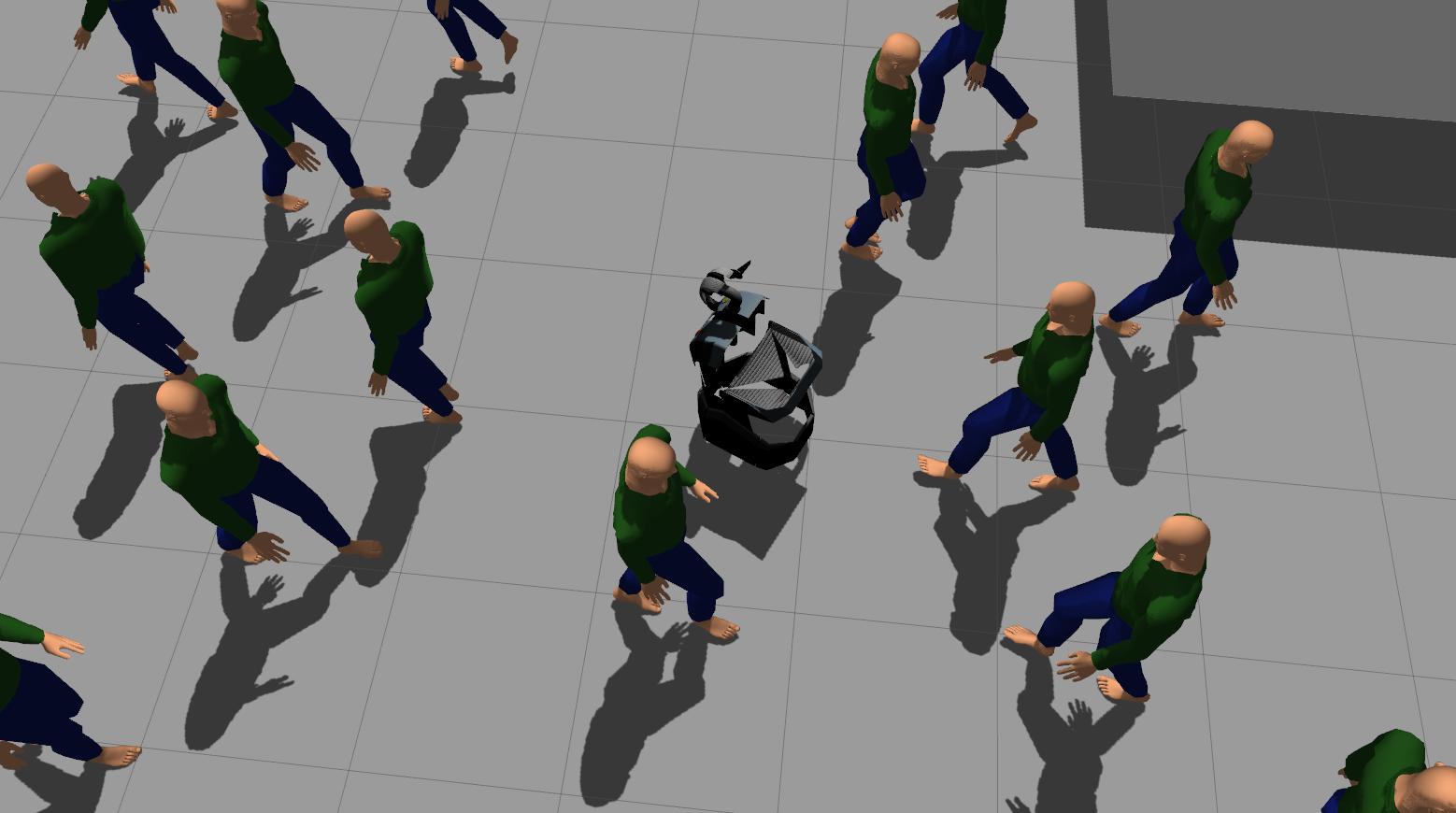}
    
    
\columnbreak

{\small Chapter 6: Trajectory Generation in New
Environments from Past Experiences}

\centering
\vspace{1em}
\includegraphics[width=0.4\textwidth]{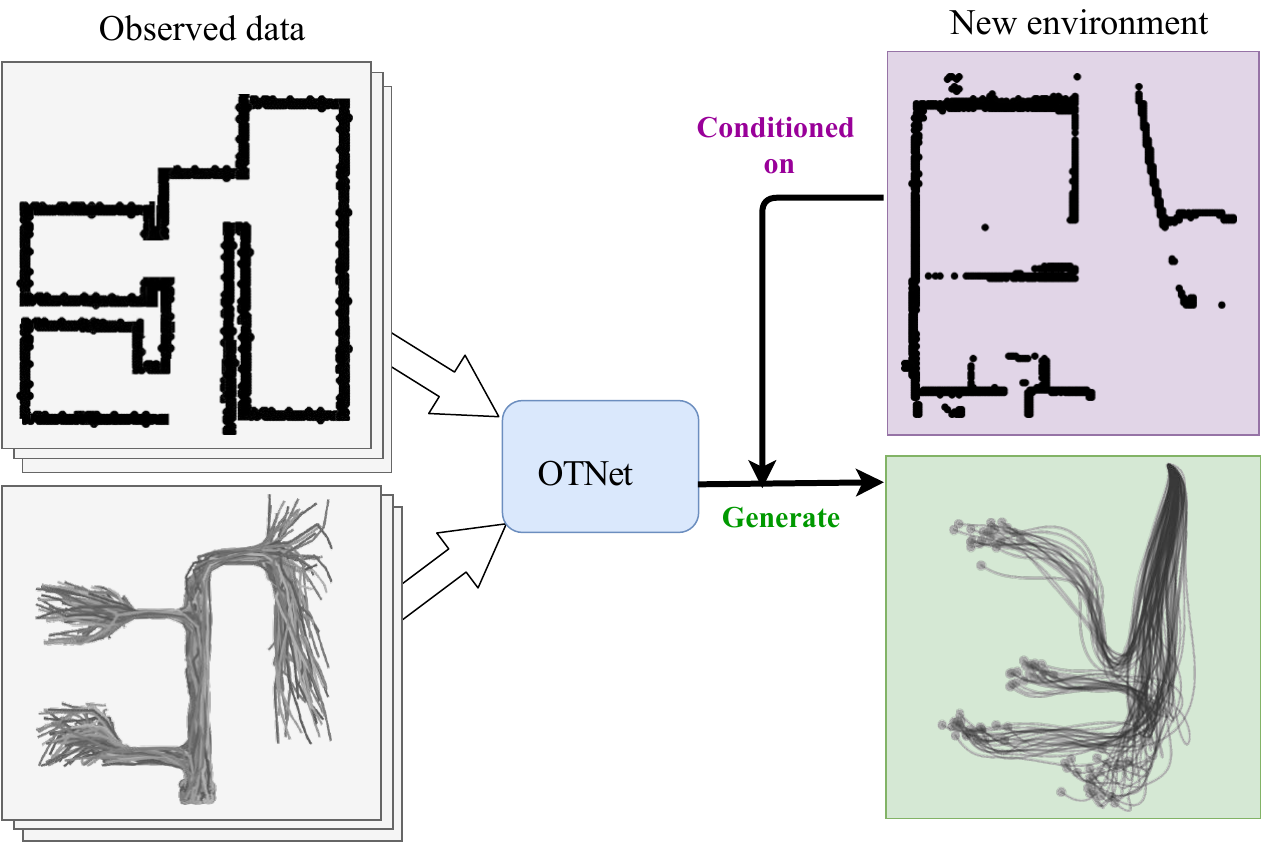}

\end{multicols}
\centering
{\small Chapter 7: Structurally Constrained Motion Prediction}
\vspace{1em}
\includegraphics[width=0.45\textwidth]{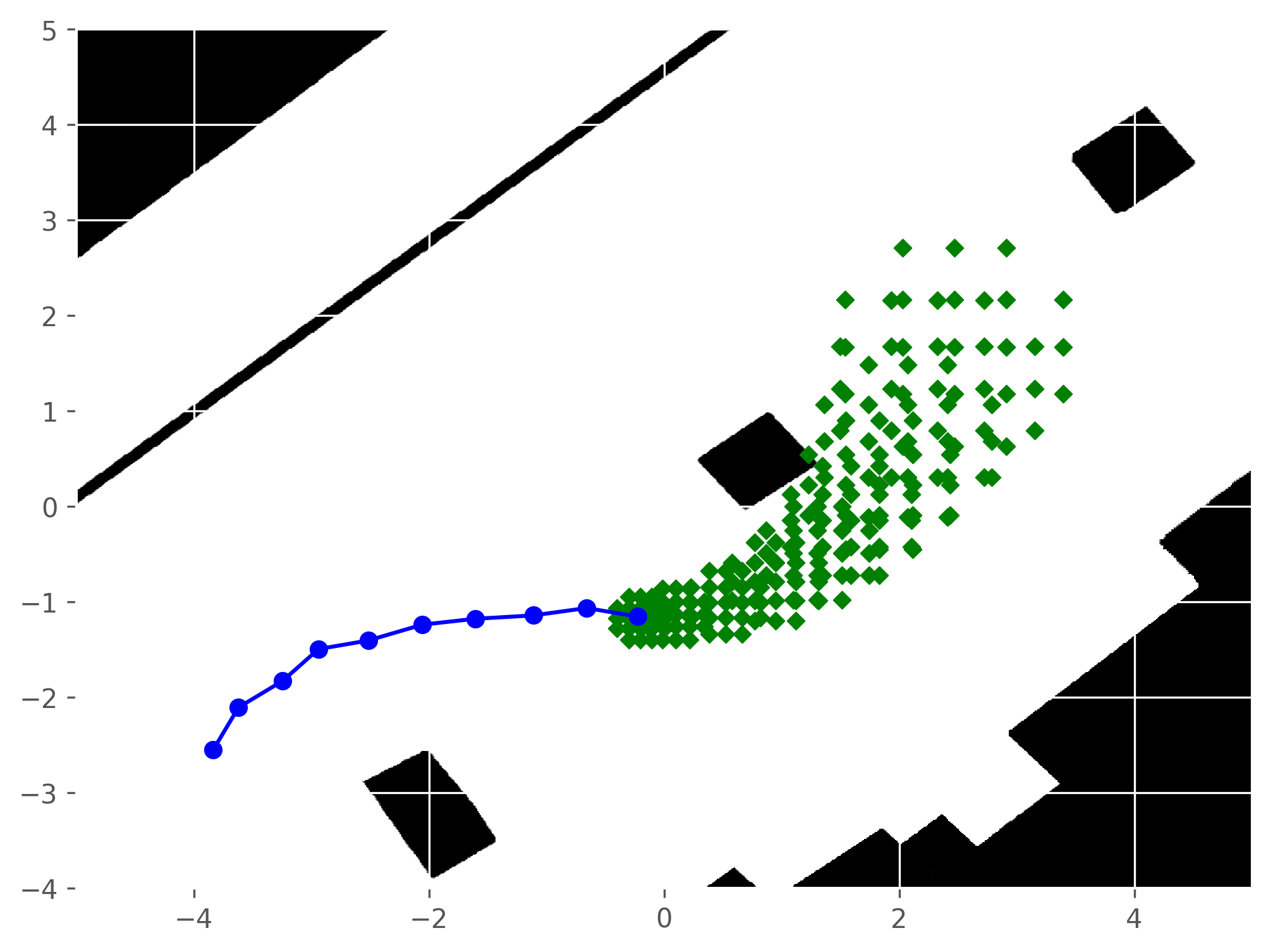}
\end{tcolorbox}
     \item \textbf{Part II: Learning for Anticipatory Navigation:} Humans are able to anticipate how others in the vicinity move, and act accordingly to negotiate a path through dynamic environments -- we seek to enable robots to do the same. We study the problem of navigating through environments with crowds while anticipating how the pedestrians in the environment move. We are interested in answering the questions: 
     
     \emph{How do we learn a probabilistic predictive model of dynamic objects, such as pedestrians, from data and past experience? Additionally, how can we take these predictions into account and anticipate the motion of others during navigation?} 
     
     The specific contributions include: 
     \begin{itemize}
         \item \textbf{Stochastic Process Anticipatory Navigation:} In \cref{chap5}, we introduce the \emph{Stochastic Process Anticipatory Navigation} (SPAN) framework. A probabilistic predictive model produces distributions of where dynamic pedestrians are expected to move. This predictive model is integrated into a time-to-collision controller to smoothly navigate through moving crowds. The robot is able to actively predict and preempt the behaviours of other agents in its proximity. We observe the emergence of behaviours, such as the controlled robot following behind others that are moving in the same direction, when squeezing through crowds. 
         
         \item \textbf{Occupancy-Conditioned Trajectory Networks:} In \cref{chap5b}, we introduce \emph{Occupancy-Conditioned Trajectory Network} (OTNet), a model to generalise motion patterns observed in past environments to new environments. Humans have the ability to intuit probable motion trajectories in an environment simply by examining the environment structure from a floor plan. OTNet aims to use machine learning to allow robots to do the same. OTNet encodes the environment occupancy as a vector of similarities between an environment of interest and previously observed environments, and trains a neural network to map these embedding vectors to multi-modal probability distributions over trajectories. The trajectory distributions can be further refined by enforcing the start points of the trajectories. We envision that the trajectory distribution over the entire environment can be used as a prior for trajectory prediction models, which can then be refined to accurately extrapolate partially observed trajectories.
         
         \item \textbf{Probabilistic motion prediction model with structural constraints:} In \cref{chap6}, we introduce a method to predict motion trajectories capable of imposing structural constraints to multi-modal distributions over trajectories. This enables the probabilistic motion prediction model to enforce known constraints in the environment. For example, we know that moving objects in the environment are highly unlikely to go through occupied coordinates, so we explicitly construct and subsequently enforce chance-constraints on the predicted distribution of future trajectories, such that it complies with the known occupancy structure of the environment.   
     \end{itemize}

\begin{tcolorbox}[colback=white,colframe=blue!75!black,title= Highlight figures of Part III: Learning for Robot Manipulator Motion Generation]
\begin{multicols}{2}
\centering
{\small Chapter 8: Diffeomorphic Transforms for Generalised Imitation Learning}

\vspace{1em}    
\includegraphics[width=0.48\textwidth]{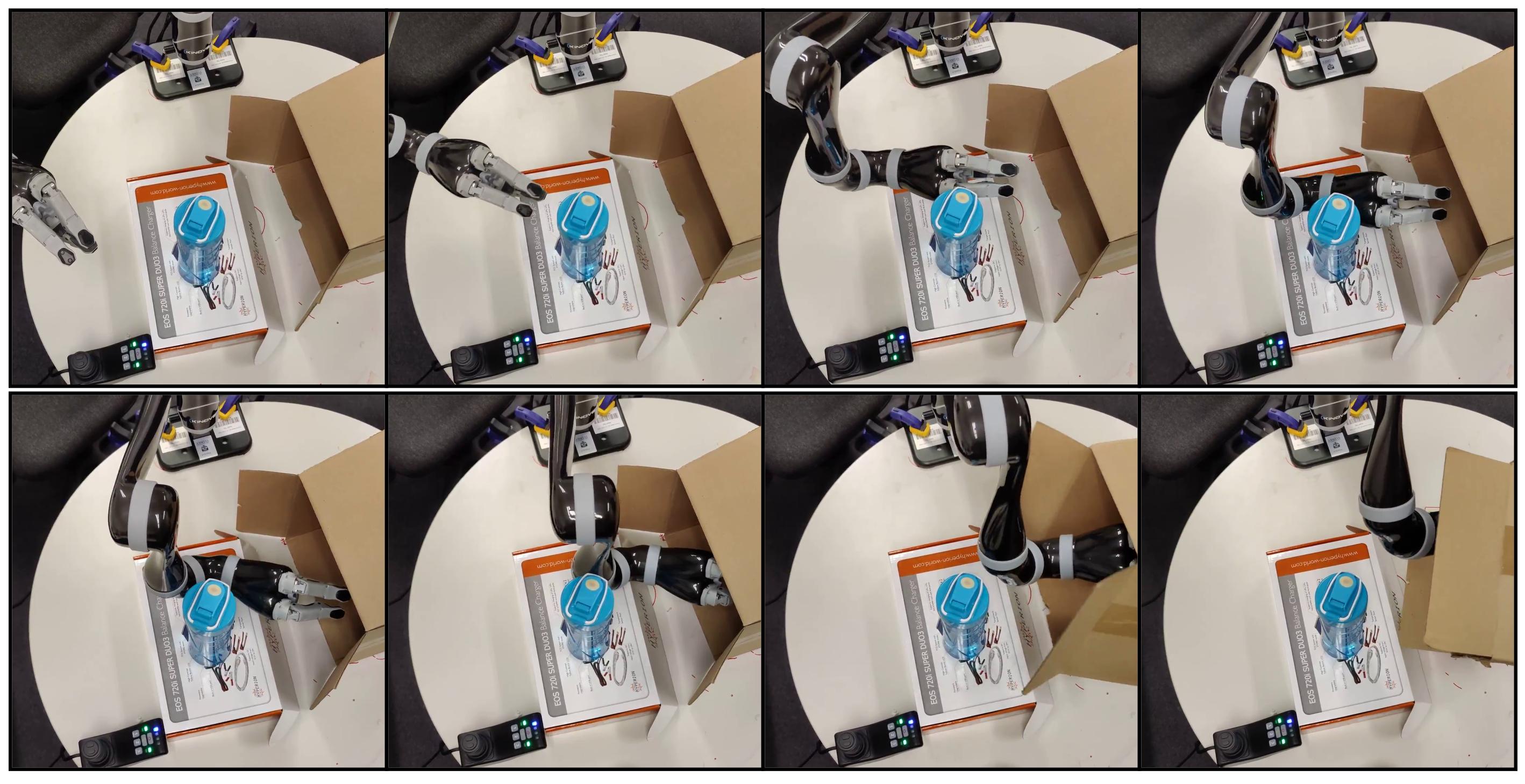}
    
    
\columnbreak

{\small Chapter 9: Motion Generation with Geometric Fabric Command Sequences}

\centering
\vspace{1em}
\includegraphics[width=0.35\textwidth]{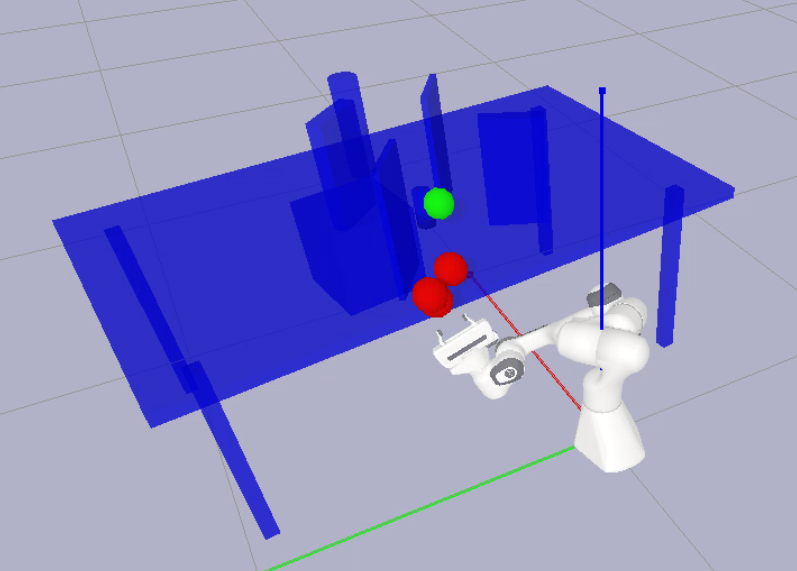}

\end{multicols}
\end{tcolorbox}

     \item \textbf{Part III: Learning for Robot Manipulator Motion Generation:} We study robot learning methods to generate motion trajectories for manipulators. Compared to ground robots, robot manipulators often have higher degrees of freedom, resulting in higher dimensional motion trajectories. However, they are typically not subject to non-holonomic constraints typical of ground robots. In \cref{part3}, we seek to enable robot manipulators to operate in dynamic environments around humans. Specifically, we seek answers to the questions: 
     
     \emph{How can we enable robots to (i) learn from demonstrations to acquire new skills and generalise these skills when the environment or problem setting changes, and (ii) efficiently produce reactive motions that are also globally optimal?}
     
     The specific contributed deliverables include: 
     \begin{itemize}
         \item \textbf{Diffeomorphic Templates for generalised imitation learning:} In \cref{chap7}, we introduce \emph{Diffeomorphic Templates} for generalised imitation learning. Oftentimes, it is difficult to specify the exact skill that the robot is expected to perform. In these setups, we aim to provide a framework for humans to describe the desired motion by providing a small number of demonstrations for the robot to learn from. Additionally, we require the robot to \emph{generalise} and account for changes in the environment or other additional user specifications, which may not be present in the demonstration data. Our Diffeomorphic Templates framework decomposes each specific behaviour of the motion, such as imitating data, or avoiding new obstacles, as individual modules. These modules can then be combined in a principled manner to produce stable dynamical systems representing the desired combined motion.
         \item \textbf{Geometric Fabric Command Sequences:} In \cref{chap8}, we introduce \emph{Geometric Fabric Command Sequences (GFCS)}, a method capable of generating collision-free and kinematically feasible robot motion from start to goal. The produced motion is both reactive and globally optimal. The proposed GFCS extends upon, and solves an important limitation of, \emph{Geometric Fabrics} \citep{geoFabs}. Geometric Fabrics can reactively generate smooth motion, but the generated motion trajectories often suffer from a tendency of becoming trapped in local non-convex regions, and are unable to reach the goals. We formulate global motion generation as a global optimisation problem over the parameters of several Geometric Fabric models, which is then solved by a black-box optimiser. We additionally develop a self-supervised learning framework to speed up the optimisation by learning a generative model to warm start the black-box optimisation.
     \end{itemize}
\end{enumerate}

\section{Thesis Structure}
We begin in \textbf{\cref{chap2}} by providing background knowledge on relevant topics, and describe some of the frequently-used methods which appear in subsequent contribution chapters. These topics span both machine learning and robotics, including supervised regression with linear regression with non-linear features and neural networks, optimisation, environment representations, robot kinematics, and dynamical systems. 

Next, in \textbf{\cref{part1}}, we detail contributions to environment representation via learning. In \textbf{\cref{chap3}}, we introduce approaches to fuse \emph{Fast Bayesian Hilbert Maps}, a continuous occupancy representation. In \textbf{\cref{chap4}}, we propose a continuous spatiotemporal model to capture the distributions of movement directions in an environment. 

Subsequently, in \textbf{\cref{part2}}, we describe contributions to learning for anticipation navigation. In \textbf{\cref{chap5}}, we present the \emph{Stochastic Process Anticipatory Navigation} (SPAN) framework, to predict the movement of other agents and use these predictions when navigating through crowds. In \textbf{\cref{chap5b}}, we introduce \emph{Occupancy-conditioned Trajectory Network} (OTNet) a model capable of conditioning on an environment's occupancy to generate the likely motion patterns in the environment; in \textbf{\cref{chap6}}, we outline a combined learning and optimisation framework, which enforce constraints on probabilistic motion prediction models. 

Then, in \textbf{\cref{part3}}, we elaborate on contributions to generating motion for robot manipulators. In \textbf{\cref{chap7}}, we propose the \emph{Diffeomorphic Templates} (DTs) framework for generalised imitation learning, which enables robots to learn from expert demonstrations and to generalise acquired behaviour when the problem setup changes. In \textbf{\cref{chap8}}, we introduce \emph{Geometric Fabric Command Sequence} (GFCS), along with a self-supervised learning framework to integrate GFCS, to produce manipulator motion which is both reactive and global. 

Finally, in \textbf{\cref{chap9}}, we provide concluding remarks for this thesis, revisit connections between the presented themes, and outline possible future research directions. 


\pagebreak

\chapter{Background}\label{chap2}
\section{Overview}
In this chapter, we shall provide background to tools used to develop the robot learning methods introduced in the later chapters. We begin by laying out the background to some of the topics relevant to machine learning. First, in \cref{reg_sec}, we describe supervised regression problems, which are some of the most commonly encountered machine learning problems. To tackle these regression problems, in \cref{linreg_background_sec}, we outline the background around fitting linear models, which can be used in conjunction with non-linear features. These models are easy to implement and interpret. We give an illustrative example of non-linear features in \cref{kernel_backg}, and show how kernel functions can be used to greatly reduce computation. However, linear models, even when applied with non-linear features, are often limited in their flexibility. Hence, in \cref{FC_networks}, we discuss fully-connected neural networks, the quintessential neural network model. These flexible models can be highly parameterised, while also trained efficiently on modern GPUs. Both linear models and neural network models have been used throughout this thesis to approximate functions of interest. Then in \cref{sec_optimise}, we sketch the background for some of the optimisation techniques used in this thesis. This includes \emph{ADAptive Moment estimation} (ADAM) to estimate parameters of neural network models, \emph{Sequential Quadratic Programming} (SQP) to solve constrained optimisation problems, where first and second derivatives are available, and \emph{Covariance Matrix Adaptation Evolution Strategy} (CMA-ES), a black-box optimiser which makes little assumptions about the function that is optimised, and does not require function derivatives to find globally optimal solutions. 

Then, we provide context to the relevant topics from robotics. We begin, in \cref{sec:env_reps}, by providing background on methods to represent a robot's environment. We discuss occupancy grid maps, an early model to discretely represent the environment of a robot, and outline how the grid values can be updated using Baye's rule. We also introduce Hilbert Maps, a learning-based approach that produces continuous representations of the environment. Next, in \cref{kinematics}, we provide a brief overview of the kinematics of robot manipulators, and explain the various concepts, such as \emph{configuration space}, \emph{task space} and \emph{forward kinematics}. We then give an example of how to derive the forward kinematics for a simple planar robot. Finally, in \cref{sec:background_dyn_sys}, we give a gentle background to the concepts around dynamical systems and ODEs which are relevant to our contributions. Robot motion is often described as a dynamical system and individual robot motion trajectories are integrals or roll-outs of the system. Additionally, we also outline the concept of asymptotic stability for dynamical systems. This is revisited in \cref{chap7}, where we tackle generalised imitation learning while maintaining stability.


\section{Supervised Regression Problems} \label{reg_sec}
Many robot learning problems that arise in this thesis fall under the general umbrella of supervised learning. In particular, the supervised regression problems that appear in this thesis include the problem of finding continuous trajectories in \cref{chap5} and \cref{chap6}, as well as that of learning from expert demonstrations in \cref{chap7}.

Consider a dataset containing $N$ instances of input-output pairs $\mathcal{D}=\{(\mathbf{x}_{i},y_{i})\}_{i=1}^{N}$. Here, we assume that the inputs $\mathbf{x}$ are $d$-dimensional vectors from a set $\mathcal{X}\subseteq \mathbb{R}^{d}$. We shall, in particular, focus on regression problems, where the outputs $y$ are real-valued scalars from a set $\mathcal{Y}\subseteq\mathbb{R}$. Though the presented methods can be easily generalised to multi-variate outputs. We assume that observed instances are independently drawn from a joint distribution $p(\mathbf{x},y)$ over $\mathcal{X}$ and $\mathcal{Y}$. Additionally, the observed $y_{i}$ are outputs of some unknown target function on the inputs, corrupted by zero-mean Gaussian noise with constant variance $\sigma^{2}$, i.e. 
\begin{align}
y_{i}=g(\mathbf{x}_{i})+\varepsilon && \text{where } \varepsilon \sim N(0,\sigma^{2}).
\end{align}
We refer to the inputs, $\mathbf{x}_{i}\in \mathcal{X}$, as \emph{features} and the outputs, $y_{i}\in\mathbb{R}$ as \emph{labels}. Our aim is to approximate the target function $g:\mathcal{X}\rightarrow \mathcal{Y}$ as accurately as possible with another function $f$. We refer to $f$ as a function approximator. 


\subsection{Linear Regression Models}\label{linreg_background_sec}

Here we study linear regression, one of the simplest parametric approaches to tackle supervised learning. We begin by making assumptions about the set of trial functions to search in, and specify a parametric form for the unknown function. Take the linear case: if we assume that $g$ is linear and restrict our approximation to be likewise linear, omitting the intercept for brevity, our approximation takes the form:
\begin{equation}
    f(\mathbf{x})=\mathbf{w}^{\top}\mathbf{x},
\end{equation}
where vector $\mathbf{w}\in\mathbb{R}^{d}$ contains the weight parameters of our model. Fitting a linear model on given inputs under the regression problem setup is referred to as \emph{linear regression}. More generally, we can introduce non-linearities to our modelling by adopting a set of $k$ non-linear basis functions (also known as feature maps), $\{\phi_{i}:\mathbb{R}^{d}\rightarrow\mathbb{R}\}_{i=1}^{k}$, and typically $k>d$. We define $\bm{\phi}(\mathbf{x})=[\phi_{1}(\mathbf{x}),\ldots, \phi_{k}(\mathbf{x})]^{\top}$ as our transformed set of features, and our model is now: 
\begin{equation}
f(\mathbf{x})=\mathbf{w}^{\top}\bm{\phi}(\mathbf{x}), \label{linbasis_mod}
\end{equation}
where the vector of parameters is now $\mathbf{w}\in\mathbb{R}^{k}$. Note that although the function $f$ is no longer linear with respect to the original inputs $\mathbf{x}$, it remains linear with respect to the transformed features $\bm{\phi}(\mathbf{x})$. Therefore, we can simply view this as applying linear regression to the new set of transformed features.

After assuming a suitable parametric form of our approximator $f$, our task is now to find the set of parameters $\mathbf{w}$ such that $f$ is as ``close'' as possible to $g$. To accomplish this, we need to define a \emph{loss function} to measure the "closeness" of our model prediction $f(\mathbf{x})$, when the label is $y$. A commonly used loss function used in regression problems is the square error loss:
\begin{equation}
    \mathcal{L}(f(\mathbf{x}),y)=(f(\mathbf{x})-y)^{2}
\end{equation}
The best-fit function approximator is then found by optimising the parameters $\mathbf{w}\in\mathbb{R}^{k}$ such that the expected value of the loss function is minimised. This expected value cannot be directly computed, as the joint distribution $p(\mathbf{x},y)$ is unknown. However, we can compute an approximation, or the \emph{empirical risk}, by taking the average loss over the training dataset. Specifically,

\begin{align}
\mathbf{w}^{*}=&\arg\min_{\mathbf{w}\in\mathbb{R}^{k}}\mathbb{E}_{p(\mathbf{x},y)}[\mathcal{L}(f(\mathbf{x}),y)]\\
\approx &\arg\min_{\mathbf{w}\in\mathbb{R}^{k}}\frac{1}{N}\sum_{i=1}^{N}(f(\mathbf{x}_{i})-y_{i})^{2},\\
\text{and }& f(\mathbf{x}_{i})=\mathbf{w}^{\top}\bm{\phi}(\mathbf{x}_{i}), \label{opt_cost}
\end{align}
where the expected value of the loss, over the joint distribution $p(\mathbf{x},y)$, is empirically approximated by the i.i.d samples contained in the dataset $\mathcal{D}$. We refer to the computed loss as the \emph{mean squared error} (MSE) loss. We can then proceed to solve \cref{opt_cost} to find the optimal parameters $\mathbf{w}^{*}$. In the case of linear models with a MSE loss, the solution admits a closed-form:
\begin{align}
    \mathbf{w}^{*}=(\bm{\Phi}^{\top}\bm{\Phi})^{-1}\bm{\Phi}^{\top}\mathbf{y}, 
    && 
    \text{where } \bm{\Phi}=\begin{bmatrix}
    \bm{\phi}(\mathbf{x}_{1})\\
    \vdots\\
    \bm{\phi}(\mathbf{x}_{N})
    \end{bmatrix}
    &&
    \text{and } \mathbf{y}=\begin{bmatrix}
    y_{1}\\
    \vdots\\
    y_{N}
    \end{bmatrix}.\label{backg_solve_ols}
\end{align}
This allows us to directly efficiently obtain the optimised weights, without resorting to numerical optimisation approaches.

\subsection{Non-linear Features and Kernel Functions: An Example}\label{kernel_backg}
As outlined in the previous subsection \cref{linreg_background_sec}, non-linear features extend linear regression models to estimate non-linear functions from inputs to targets, by first mapping the original $d$-dimensional inputs to $k$-dimensional non-linear features with a feature map, $\bm{\phi}:\mathbb{R}^{d}\rightarrow\mathbb{R}^{k}$, and then applying linear methods on the features. Intuitively, the unknown function may be linear in the generally higher dimensional \emph{feature space}, while being non-linear with respect to the original inputs. Here, we show an illustrative example of regression with polynomial features, that are explicitly computed. Then, we shall discuss the concept of ``kernelisation'', which speeds up computations by avoiding explicitly computing inner products on features. 

Here, we consider an example non-linear regression problem from \cite{kernelDMD}. Suppose we have 3-dimensional inputs $\mathbf{x}\in\mathbb{R}^{3}$ and aim to find a function approximator $f$ which maps these inputs to targets. Suppose we know \emph{a priori}, from physical intuition or empirical evidence, that the set of pair-wise products spans the space of candidate functions -- our function approximator can be expressed as a linear combination of pair-wise quadratic basis functions,
\begin{align}
\bm{\phi}(\mathbf{x})=[x_{1}x_{2},x_{1}x_{3},x_{2}x_{3},x_{1}^{2},x_{2}^{2},x_{3}^{2}]^{\top}\in\mathbb{R}^{6}, && f(\mathbf{x})=\mathbf{w}^{\top}\bm{\phi}(\mathbf{x}),
\end{align}
where $\mathbf{w}\in\mathbb{R}^{6}$ are weights which can be computed via \cref{backg_solve_ols}. We observe that solving for the weights requires products between the features. If we explicitly construct features from two inputs $\mathbf{x},\mathbf{x'}$, and compute their dot product, $\bm{\phi}(\mathbf{x})^{\top}\bm{\phi}(\mathbf{x'})$, the number of operations required is 23. We need to perform 6 products to explicitly construct each feature vector, and then 6 products and 5 sums to compute the dot product. This computation can be sped up by constructing an equivalent \emph{kernel function}. 

Without loss of generality, we consider a slightly re-scaled feature map for a simpler reduction of coefficients,
\begin{equation}
\bm{\varphi}(\mathbf{x})=[\sqrt{2}x_{1}x_{2},\sqrt{2}x_{1}x_{3},\sqrt{2}x_{2}x_{3},x_{1}^{2},x_{2}^{2},x_{3}^{2}]^{\top}\in\mathbb{R}^{6}.
\end{equation}
The dot product between two the feature vectors of two inputs $\mathbf{x},\mathbf{x'}\in\mathbb{R}^{3}$ is then,
\begin{align}
\bm{\varphi}(\mathbf{x})^{\top}\bm{\varphi}(\mathbf{x'})&=2x_{1}x_{2}x_{1}'x_{2}'+2x_{1}x_{3}x_{1}'x_{3}'+2x_{2}x_{3}x_{2}'x_{3}'+x_{1}^{2}(x_{1}')^{2}+x_{2}^{2}(x_{2}')^{2}+x_{3}^{2}(x_{3}')^{2}\\
&=(x_{1}x_{1}'+x_{2}x_{2}'+x_{3}x_{3}')^{2}\\
&=(\mathbf{x}^{\top}\mathbf{x}')^{2}:=k(\mathbf{x},\mathbf{x}').
\end{align}
Lo and behold, there is an alternative route to evaluating the dot product between the feature vectors of two inputs! We can instead evaluate the defined kernel function $k:\mathbb{R}^{3}\times\mathbb{R}^{3}\rightarrow\mathbb{R}$, which requires only 6 operations -- 3 products and 2 sums for the dot product, and 1 for the square. The computational savings of evaluating the kernel function instead of explicitly evaluating feature vectors is even more apparent when the inputs are higher dimensional or the interaction terms of higher dimensions are modelled. Kernel functions have been designed for a variety of common feature maps, and rules to combine kernel functions have also been explored. More details on kernels and kernel designing can be found in Chapter 2 of \cite{cookbook}.

\subsection{Regression with Fully-Connected Neural Networks}\label{FC_networks}

Linear models on non-linear features have been used extensively throughout this thesis, e.g. to construct continuous environment representations (\cref{chap3,chap4}), continuous trajectory representations (\cref{chap5,chap6}), and invertible functions (\cref{chap7}). They are straightforward to apply and easy to interpret, when we have \emph{a priori} knowledge of what basis functions to select. However, the true workhorses of contemporary machine learning are neural network models. Neural networks have been used in each of the chapters between \cref{chap4} to \cref{chap8}. In this section, we shall introduce fully-connected neural networks, often also referred to as ``feedforward neural networks'' or ``dense neural networks'', which are the quintessential neural network models.  

Consider the regression problem outlined in \cref{reg_sec}. Instead of assuming that the function approximator $f$ is linear, we instead assume that $f$ is a neural network given by a composition of $q$ layers, i.e.,
\begin{equation}
    f(\mathbf{x})=h^{q}(\ldots h^{2}(h^{1}(\mathbf{x}))),
\end{equation}
where $q$ is known as the \emph{depth} of network, and each $h^{i}$, $i=1,\ldots,q$ is known as a fully-connected layer. Each layer itself is a linear function, with a weight matrix $\mathbf{W}^{i}$ and bias $\mathbf{b}^{i}$, along with an activation function $\varphi^{i}$ which may be non-linear, that is:
\begin{equation}
    \mathbf{u}^{i+1}=h^{i}(\mathbf{u}^{i})=\varphi^{i}({\mathbf{W}^{i}}^{\top}\mathbf{u}^{i}+\mathbf{b}^{i}),
\end{equation}
where $\mathbf{u}^{i}$ is the input from the previous layer and $\mathbf{u}^{i+1}$ is the output of the layer. Specifically, the input of the first layer $h^{1}$ is $\mathbf{x}$, and the inputs of the remaining layers are the outputs of the previous layer. Typically, in regression problems, the activation function at the final layer, $\varphi^{q}$, is simply the identity function, while the other activation functions are non-linear. Various non-linear functions can be used as activation functions including the $ReLU$, $Tanh$, and $Softmax$ functions \citep{GoodBengCour16}.  

Unlike linear models under an MSE loss scheme, the parameters of the best fit neural network function cannot be expressed in closed form, and instead requires numerical optimisation. Let us denote the neural network function approximator as $f_{\bm{\theta}}$, where $\bm{\theta}=\{\mathbf{W}^{i},\mathbf{b}^{i}\}_{i=1}^{q}$ is the set of all parameters of the network. We apply first-order gradient descent methods, such as Stochastic Gradient Descent (SGD), where the most basic update rule for the parameters, with respect to a loss $\mathcal{L}$, is:
\begin{align}
    \bm{\theta}^{j+1}\leftarrow \bm{\theta}^{j}-\alpha \nabla_{\bm{\theta}}\frac{1}{\lvert I^{j} \lvert}\sum_{i\in I^{j}}\mathcal{L}(f_{\bm{\theta}}(\mathbf{x}_{i}),y_{i})
\end{align}
where $j$ denotes the current iterations, and $\alpha$ is the step-size. It is often prohibitively expensive to compute the derivatives of the parameters with respect to the loss over all of the examples. Therefore, we estimate this derivative by randomly selecting a subset of examples from the dataset. The set $I^{j}$ contains a batch of randomly selected indices of the dataset, drawn every iteration. The number of corresponding data points selected in each batch is in practice significantly smaller than the entire dataset, with the batch known as a \emph{mini-batch}. The parameters are updated iteratively, with the initial values of the parameters $\bm{\theta}^{0}$ drawn from a random distribution. Selecting a fit-for-purpose optimiser impacts the performance of the network. To this end, various newer and more sophisticated optimisers have been developed upon SGD, such as ADAM \citep{Kingma2015AdamAM}.

We observe that a challenge of utilising neural networks is efficiently obtaining derivatives of the loss, with respect to network parameters, over a sizeable batch of data examples. Indeed, the adoption of neural networks is driven by the availability of efficient automatic differentiation libraries such as \emph{PyTorch} and \citep{pytorch} and \emph{Tensorflow} \citep{tflow}, which can be parallelised on modern GPUs. These developments have given rise to neural network models which are increasingly deep, and allow for training on datasets which are increasingly large in size.  


\section{Optimisation}\label{sec_optimise}
Beyond using gradient descent to train neural networks, many methods developed in this thesis make use of optimisation techniques. In this section, we provide an introduction to the optimisation methods used. Optimisation seeks to optimise (typically minimise) an \emph{objective function} with respect to a set of \emph{decision variables}, while satisfying a set of \emph{constraints}. We can get a general sense of the difficulty of the optimisation problem by the global structure (convex vs non-convex) and local structure (derivatives available vs derivatives not available) of the objective and constraints. In this thesis, we use (1) ADAptive Moment estimation (ADAM) a variant of Stochastic Gradient Descent (SGD) (discussed in \cref{FC_networks}) to optimise the parameters of neural network models, (2) Sequential Least SQuares Programming (SLSQP) \cite{slsqp}, an implementation of Sequential Quadratic Programming (SQP) in \cref{chap6}, and (3) the derivative-free Covariance Matrix Adaptation Evolution Strategy (CMA-ES) \cite{CMA_tut} in \cref{chap8}. We shall provide background for these approaches.

\subsection{ADAptive Moment estimation (ADAM)}

Parameters of neural networks in this thesis are typically optimised with ADAptive Moment estimation (ADAM) \citep{Kingma2015AdamAM}, a first-order optimiser which extends Stochastic Gradient Descent (SGD). Vanilla SGD has been shown to converge much more efficiently and better escape local minima when \emph{momentum} is considered \citep{Momentum_cite}. That is, when we update the parameters in the current gradient descent iteration, the updates in the previous iterations are also taken into account. A line of work, begin from \emph{AdaGrad} \citep{Adagrad}, \emph{RMSProp} \citep{rmsprop}, to ADAM makes use of momentum. Here, we provide background on ADAM, which has become the go-to optimiser for modern neural network models.

When optimising a set of $d$-dimensional parameters, ADAM keeps track of two properties of the optimiser at its current iteration: (1) the first moment, $\mathbf{m}\in\mathbb{R}^{d}$: the exponential moving average of gradients; (2) the second moment, $\mathbf{v}\in\mathbb{R}^{d}$: the exponential moving average of the gradients square of gradients. Assuming that we are optimising a loss $\mathcal{L}$ with model parameters $\mathbf{\theta}$ and the decay rates for the exponential moving average are given hyper-parameters, $\beta_{1}, \beta_{2}\in [0,1)$, for the first and second moments respectively. Then, the update rules for the loss gradients and moments at iteration $j+1$ are:
\begin{align}
    \mathbf{g}^{j+1}&\leftarrow \nabla_{\bm{\theta}}\mathcal{L}(\bm{\theta}^{j})\\
    \mathbf{m}^{j+1}&\leftarrow \beta_{1}\mathbf{m}^{j}+(1-\beta_{1})\mathbf{g}^{j+1}\\ 
    \mathbf{v}^{j+1}&\leftarrow \beta_{2}\mathbf{v}^{j}+(1-\beta_{2})(\mathbf{g}^{j+1}\odot \mathbf{g}^{j+1}), 
\end{align}
where $\odot$ refers to the Hadamard (element-wise) product, and the initial first and second moments are set as vectors of 0. To account for the bias towards zeros of the computed moments, ADAM performs bias-corrections for the moments:
\begin{align}
\mathbf{\hat{m}}^{j+1}\leftarrow \frac{\mathbf{\hat{m}}^{j+1}}{(1-(\beta_{1})^{j+1})}, && \mathbf{\hat{v}}^{j+1}\leftarrow \frac{\mathbf{\hat{v}}^{j+1}}{(1-(\beta_{2})^{j+1})}. 
\end{align}
Here, $(\beta_{1})^{j+1}$ and $(\beta_{1})^{j+1}$ refers to the decay parameters to the power of $j+1$. We can then use the corrected moments to update our decision variables (i.e. the neural network parameters),
\begin{equation}
\bm{\theta}^{j+1}\leftarrow \bm{\theta}^{j}-\alpha \frac{\mathbf{\hat{m}}^{j+1}}{\sqrt{\mathbf{\hat{v}}^{j+1}}+\epsilon},
\end{equation}
where $\alpha$ is the learning rate, $\epsilon$ is a small value (typically $10^{-8}$) to prevent divide by zero errors, and $\sqrt{\mathbf{\hat{v}}^{j+1}}$ refers to the element square-root of $\mathbf{\hat{v}}^{j+1}$. In the next subsection, we shall provide background on optimisers which are used beyond optimising for parameters for learning models.
\subsection{Sequential Quadratic Programming}
Sequential quadratic programming (SQP) can be applied to constrained optimisation problems, where the objective and constraints are twice continuously differentiable. SQP is an iterative method which begins at a random guess as the solution, and sequentially solves \emph{quadratic programs}, one at each iteration to refine the solution. A quadratic program (QP) has a quadratic objective function with linear constraints and can be efficiently solved to optimality \citep[Chap~16]{opt_book}. Intuitively, SQP resembles applying Newton's method to the objective function, while accounting for the constraints. When SQP is applied to an unconstrained problem, SQP reduces to Newton's method, as solving an unconstrained quadratic program amounts to finding the minimum of a convex quadratic function. SQP will converge on a locally-optimal solution, and can be used in conjunction with global search meta-heuristics to escape local minima and find more global solutions. Consider a canonical constrained optimisation problem:

\begin{align}
    \min_{\mathbf{x}}&f(\mathbf{x})\\
    \text{subject to:  }  g(\mathbf{x})&=0,\\
        h(\mathbf{x})&\geq 0.
\end{align}
where $\mathbf{x}\in\mathbb{R}^{d}$ denotes the decision variables, $f:\mathbb{R}^{d}\rightarrow \mathbb{R}$ is the objective function, $g:\mathbb{R}^{d}\rightarrow \mathbb{R}^{p}$ gives $p$ equality constraints and $h:\mathbb{R}^{d}\rightarrow \mathbb{R}^{q}$ gives $q$ inequality constraints. The Lagrangian of this problem is given by:

\begin{equation}
    L(\mathbf{x},\bm{\mu},\bm{\lambda})=f(\mathbf{x})+\bm{\mu}^{\top}g(\mathbf{x})+\bm{\lambda}^{\top}h(\mathbf{x}),
\end{equation}
where $\bm{\mu},\bm{\lambda}$ are Lagrangian multipliers. We construct a quadratic program by using a quadratic approximation of the objective and linear approximations of the constraints, to give the increment step for the current iteration of the solution. For the $k^{th}$ iterate $\mathbf{x}_{k}$, we seek to find the current increment step $\mathbf{d}_{k}$ via the QP:

\begin{align}
    \min_{\mathbf{d}_{k}}~& f(\mathbf{x}_{k})+\nabla f(\mathbf{x}_{k})^{\top}\mathbf{d}_{k}+\frac{1}{2}\mathbf{d}_{k}^{\top}\nabla^{2}_{\mathbf{x}_{k}\mathbf{x}_{k}}L\mathbf{d}_{k},\\
    \text{s.t.  }~& g(\mathbf{x}_{k})+\nabla g(\mathbf{x}_{k})^{\top}\mathbf{d}_{k}=0,\\
    & h(\mathbf{x}_{k})+ \nabla h(\mathbf{x}_{k})^{\top} \mathbf{d}_{k}\geq 0.
\end{align}
The QP is then solved and the solution is updated for the next iterate, $\mathbf{x}_{k+1}\leftarrow \mathbf{x}_{k}+\mathbf{d}_{k}$, and the Lagrangian multipliers are updated by recomputing stationary points of the Lagrangian. We begin by providing a random initialisation of the solution, $\mathbf{x}_{0}$, and iteratively formulate and solve QPs to update the solution until we converge onto a solution. Convergence analysis and practical implementation tips are provided in \cite[Chap. 18]{opt_book}.

\subsection{Covariance Matrix Adaptation Evolution Strategy}
Sequential quadratic programming requires the objective function to be twice differentiable. However, in many problems, derivatives do not exist or are not easily available. Here, we provide background for the derivative-free method Covariance Matrix Adaptation Evolution Strategy (CMA-ES), which in practice is capable of robustly handling objective functions which are non-smooth, noisy, and contain multiple local optima \citep{benchmark-es}. CMA-ES is often used in scenarios where each function evaluation is slow or expensive, such as optimising model hyper-parameters or shape optimisation of airfoils. 
\begin{figure}[t]
    \centering
    \includegraphics[width=0.7\textwidth]{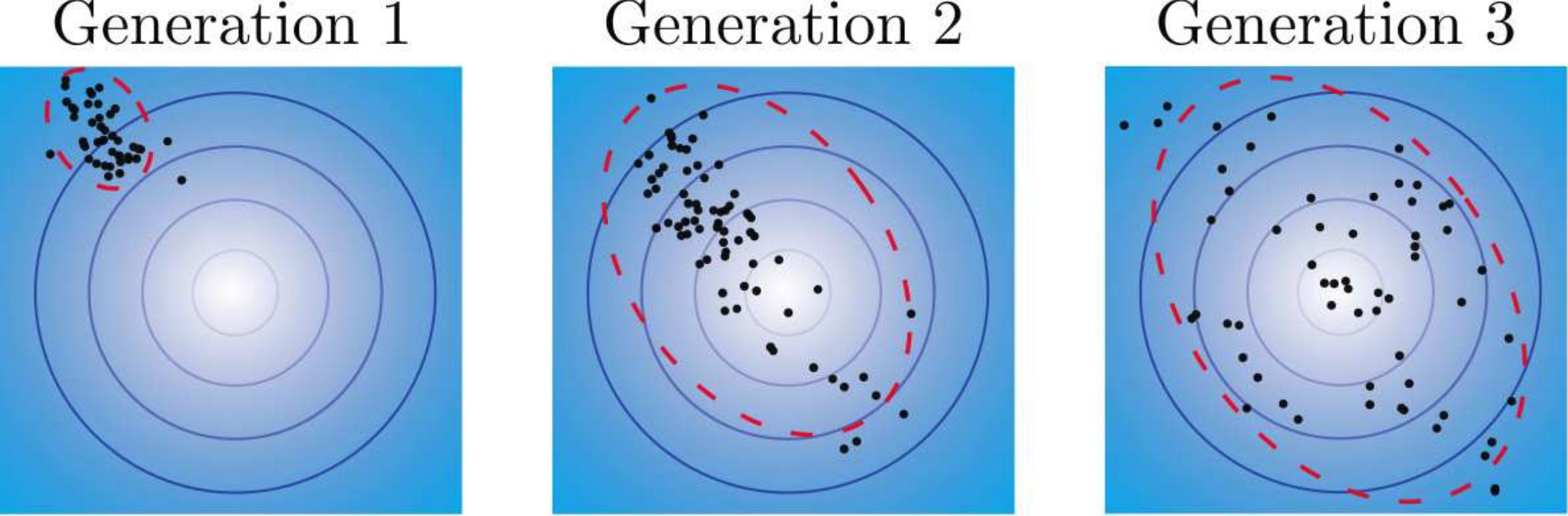}
    \vspace{0.5em}
    \includegraphics[width=0.7\textwidth]{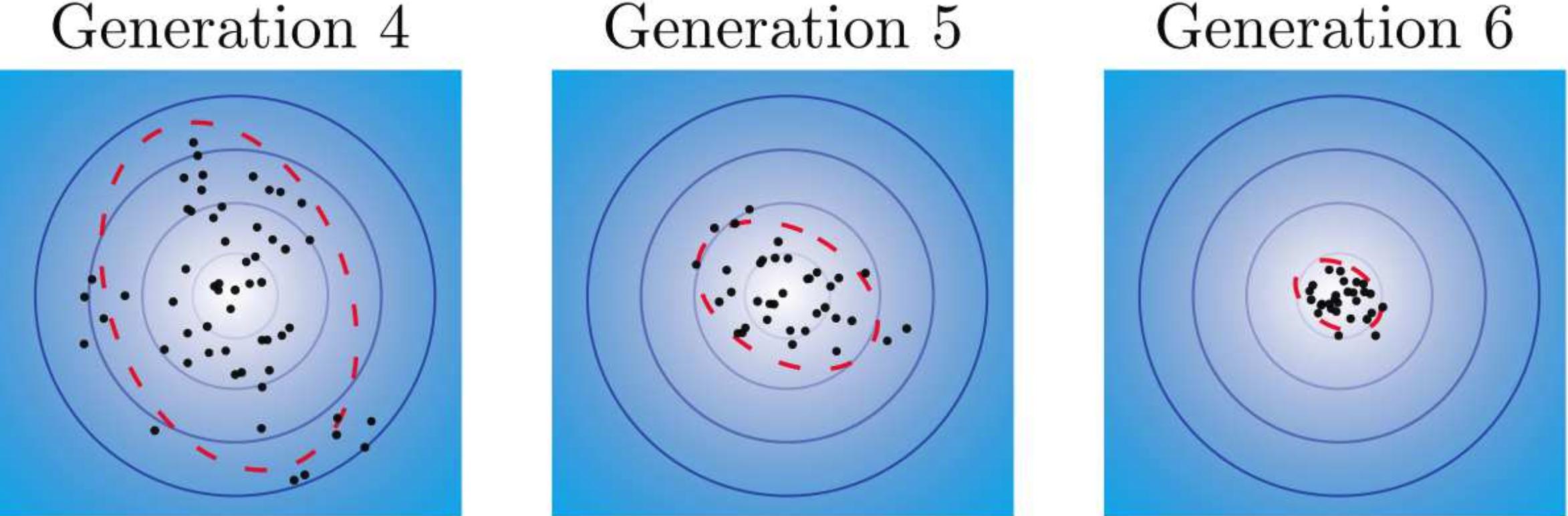}
    \caption{An example of CMA-ES optimising a cost function for 6 iterations (generations), where the contours are displayed, and lighter colours indicate a lower cost. Samples are given in black, and the ellipse iso-contour of the sampling distribution is in red. We observe that CMA-ES is able to rapidly locate the region with low cost. Figures from \cite{cma_figure_cite}.}
    \label{fig:cma-Example}
\end{figure}
We assume that we are faced with a black-box objective function $f:\mathbb{R}^{d}\rightarrow\mathbb{R}$. Here, we consider the original unconstrained formulation of CMA-ES, presented in \cite{benchmark-cma}, though constrained variants of CMA-ES have been subsequently introduced \citep{constrained_CMA}. Black-box optimisation assumes that we have no knowledge about the local or global structure of the objective, and only have the option of evaluating the objective function. Many black-box optimisation algorithms follow a generic template \citep{auger2012tutorial}, sketched in \cref{Gen_black_box}. According to this template, candidate solutions are iteratively sampled according to some distribution, and the function at the candidates is then evaluated. At each iteration, the candidates of the previous iteration, along with their function evaluations, are then used to update the sampling distribution. The essence of these black-box optimisation methods lies in how the sampling distribution $p(\mathbf{x}|\bm{\theta})$ is constructed and how the sampling distribution is updated.

\begin{algorithm}[t]
\caption{Black-box minimisation of $f:$}\label{Gen_black_box}
\textbf{Initialise:} Search distribution parameters $\bm{\theta}$, population size of each sampled batch $\lambda\in \mathbb{N}$, and solutions list $Solutions\leftarrow[]$\\
\While{not terminated}{
$\{\mathbf{x}_{1},\ldots,\mathbf{x}_{\lambda}\} \sim p(\mathbf{x}|\bm{\theta})$ \tcp*{Sample a batch of candidates}
Evaluate $f(\mathbf{x}_{1}),\ldots,f(\mathbf{x}_{\lambda})$; \\
$solutions\mathtt{.insert}(\{(\mathbf{x}_{1},f(\mathbf{x}_{1}),\ldots (\mathbf{x}_{\lambda},f(\mathbf{x}_{\lambda}))\})$ \tcp*{Keep track of candidates}
$\bm{\theta}\leftarrow F(\bm{\theta},\mathbf{x}_{1},\ldots,\mathbf{x}_{\lambda},f(\mathbf{x}_{1}),\ldots,f(\mathbf{x}_{\lambda}))$ \tcp*{Update distribution parameters}}

\textbf{Return} best $(\mathbf{x},f(\mathbf{x}))$ from $Solutions$, where $f(\mathbf{x})$ is lowest.
\end{algorithm}

In CMA-ES, the sampling distribution is assumed to be a multivariate Gaussian distribution. At a high level, in each iteration, CMA-ES draws $\lambda$ candidates to evaluate from a Gaussian sampling distribution, then evaluates the objective function and ranks the results, finally, the Gaussian sampling distribution is updated with the candidates with the best-$\mu$ function values. Specifically, at a specific iteration $k$, we have $p(\mathbf{x}_{k}|\bm{\theta}_{k})=\mathcal{N}(\mathbf{m}_{k},\sigma_{k}^{2}\mathbf{C}_{k})$, where $\mathbf{m}_{k}\in\mathbb{R}^{d}$ is a mean vector, $\sigma_{k}\in\mathbb{R}$ is a step-size parameter, and $\mathbf{C}_{k}\in \mathbb{S}^{d}_{++}$ is a positive definite covariance matrix of size $d\times d$. Additionally, we have two \emph{evolution path} parameters $p^{\sigma}_{k}\in \mathbb{R}$ and $\mathbf{p}^{\mathbf{C}}_{k}\in \mathbb{R}^{d}$ which aid convergence. CMA-ES also requires many hyper-parameters which need to be specified. These include $\mu<\lambda\in\mathbb{N}$, which is the number of samples which we consider in the update, weights $w_{1},\ldots,w_{\mu}\in \mathbb{R}$ and $\mu_{eff}=(\Sigma_{i=1}^{\mu}w_{i})^{-1}$ a variance effective selection mass factor; decay rates $c_{\sigma}, c_{\mu}, c_{\mathbf{C}}, c_{1} \in \mathbb{R}$; damping factor $d_{\sigma}$. 

At every iteration, we rank the function evaluations $f(\mathbf{x}_{1}),\ldots,f(\mathbf{x}_{\lambda})$, and select the $\mu$ samples with the best function evaluations out of the total $\lambda$ samples. We now denote the selected samples as $\{\mathbf{x}_{1},\ldots,\mathbf{x}_{\mu}\}$. Then, we can use the basic update equations for our parameters, which are designed as \citep{CMA_tut}:

\begin{align}
    \mathbf{m}_{k+1}&\leftarrow \sum_{i=1}^{\mu}w_{i}\mathbf{x}_{i},\label{update_eq_cma_1}\\
    p^{\sigma}_{k+1}&\leftarrow (1-c_{\sigma})p^{\sigma}_{k}+\sqrt{c_{\sigma}(2-c_{\sigma})\mu_{eff}} \mathbf{C}_{k}^{-\frac{1}{2}}\big(\frac{\mathbf{m}_{k+1}-\mathbf{m}_{k}}{\sigma_{k}}\big),\label{update_eq_cma_2}\\
    \sigma_{k+1}&\leftarrow\sigma_{k}\exp\big(\frac{c_{\sigma}}{d_{\sigma}}\big(\frac{\lvert\lvert p^{\sigma}_{k+1} \lvert\lvert_{2}}{\mathbb{E}\lvert\lvert N(0,I) \lvert\lvert_{2}}-1\big) \big),\label{update_eq_cma_3}\\
    \mathbf{p}^{\mathbf{C}}_{k+1}&\leftarrow (1-c_{\mathbf{C}})\mathbf{p}^{\mathbf{C}}_{k}\sqrt{c_{\mathbf{C}}(2-c_{\mathbf{C}})\mu_{eff}} \big(\frac{\mathbf{m}_{k+1}-\mathbf{m}_{k}}{\sigma_{k}}\big),\label{update_eq_cma_4}\\
    \mathbf{C}_{k+1}&\leftarrow (1-c_{1}+c_{\mu}\sum_{i=1}^{\mu}x_{i})\mathbf{C}_{k}+c_{1}\mathbf{p}^{\mathbf{C}}_{k+1}{\mathbf{p}^{\mathbf{C}}_{k+1}}^{\top}+c_{\mu}\sum_{i=1}^{\mu}w_{i}\big(\frac{\mathbf{x}_{i}-\mathbf{m}_{k+1}}{\sigma_{k+1}}\big)\big(\frac{\mathbf{x}_{i}-\mathbf{m}_{k+1}}{\sigma_{k+1}}\big)^{\top}.\label{update_eq_cma_5}
\end{align}
CMA-ES algorithm then conforms to the template in \cref{Gen_black_box}, with the samples drawn from $p(\mathbf{x}_{k}|\bm{\theta}_{k})=\mathcal{N}(\mathbf{m}_{k},\sigma_{k}^{2}\mathbf{C}_{k})$, and the update of parameters given by \cref{update_eq_cma_1,update_eq_cma_2,update_eq_cma_3,update_eq_cma_4,update_eq_cma_5}. Additional details on the selection of hyper-parameters and additional complications to the update equations can be found in \cite{CMA_tut}. An illustration of CMA-ES optimising a toy function is provided in \cref{fig:cma-Example}. We observe that the sampling distribution of candidate solutions quickly hones into the low cost region.

\section{Environment Representations in Robotics}\label{sec:env_reps}
A robot navigating in an unknown environment needs to construct a representation of what it believes to be the environment -- we call this representation a \emph{map}. The robot wishes to know whether some specified coordinates of the environment are occupied or free, or the \emph{occupancy} of the environment. In this section, we provide background on \emph{occupancy grid maps}, a discretised representation, and \emph{Hilbert Maps}, a more modern continuous representation of the environment. 
\subsection{Occupancy Grid Maps}\label{sec:background_grid_map}
 Here, we briefly outline the traditional discrete map model of \emph{occupancy grid maps} \citep{OccupancyGridMaps}, and its update via Baye's rule. Contributions in \cref{part1} of this thesis pertain to learning-based continuous mapping methods, which are contrasted and compared against discrete approaches like occupancy grid maps. 


Occupancy grid maps start by discretising the environment into a grid with some fixed resolution. We label each of the grids with an index, $i=1,\ldots,d$, and assign a binary variable $y_{i}\in\{0,1\}$, where $y_{i}=1$ indicates that the event that the $i^{th}$ cell is occupied. As the robot is moving around in the environment, it shall obtain new sensor measurements. We denote each measurement, which contains sensor data and robot position, as $s_{1},s_{2},\ldots,s_{n}$, and their combined measurements as $s_{1:n}$. For tractability, occupancy grid maps assume that each grid cell is independent, thus the joint probability of being occupied, given the sensors we have observed, is the product of its marginals:
\begin{equation}
p(\bigcap_{i=1}^{d}y_{i}=1|s_{i:n})=\prod_{i=1}^{d}p(y_{i}=1|s_{i:n}).
\end{equation}
where $\bigcap$ denotes the intersections of events. We are given a new sensor measurement $s_{n+1}$ and wish to incorporate it into our existing map model. To further simplify, we assume that each sensor measurement is independent of the others. Thus, for each grid cell,
\begin{equation}
    p(s_{1:n+1}|y_{i}=1)=p(y_{i}=1)\prod_{j=1}^{n+1} p(s_{j}|y_{i}=1),
\end{equation}
where $p(y_{i}=1)$ is the prior, which is typically set to $0.5$. Following \cite{thrun2005probabilistic}, we can develop a recursive Baye's rule update equation in a stable and efficient with a \emph{log-odds} representation. Where the odds of an event $y=1$ is given by $\log\mathrm{odds}(y=1)=\log(p(y=1)(1-p(y=1))^{-1})$, and the probability can be recovered by $p(y=1)=1-(1+\exp(\log\mathrm{odds}(y=1)))$. Incremental updates for each cell can then be expressed as:
\begin{equation}
    \log\mathrm{odds}(y_{i}=1|s_{1:n+1})=\log\mathrm{odds}(y_{i}|s_{n+1})+\log\mathrm{odds}(y_{i}|s_{1:n})-\log\mathrm{odds}(y_{i}).
\end{equation}
Here, we are assumed to have the inverse measurement model and know $p(y_{i}=1|s_{n+1})$, which depends on the sensor used. Each grid cell then simply keeps track of $\log\mathrm{odds}(y_{i}|s_{1:n})$. We note that the assumption each grid cell is independent may be an inaccurate one, as common obstacles would generally be expected to span multiple cells. This shortcoming is addressed by continuous representation models.
\subsection{Hilbert Maps}\label{subsec_background_hm}
\emph{Hilbert Maps} (HM)~\citep{HilbertMaps} are continuous representations of occupancy in environments, and have been shown to outperform occupancy grid maps significantly when there are fewer data points \citep{HilbertMaps}. This is owing to the fact that natural environments are inherently continuous, and obstacles may span multiple predefined cells. HMs utilise a logistic regression classifier with projections of occupancy data into high dimensional space to obtain non-linear features. HMs are parameterised by a vector of weights with the corresponding set of features. Stochastic Gradients Descent (SGD) is then used to learn the weights of the classifier online. 

\begin{figure}[t]
    \centering
    \includegraphics[width=0.4\textwidth]{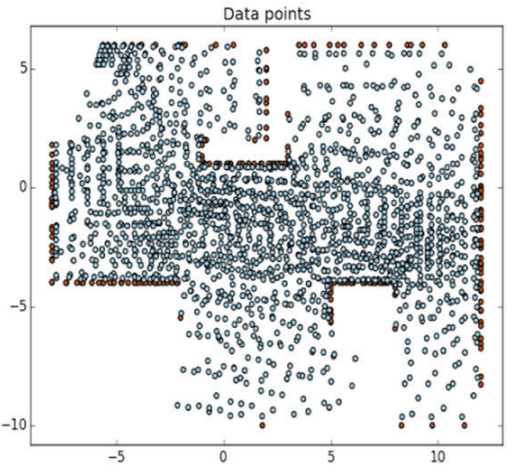}
    \includegraphics[width=0.4\textwidth]{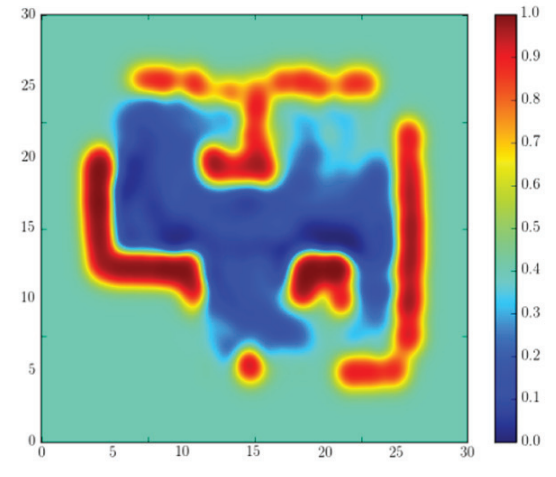}
    \caption{An illustration of an example Hilbert Map. (Left) Training data points used for training, blue and brown represent occupied and unoccupied points. (Right) The probability of being occupied is given by a constructed Hilbert Map. Figures from \cite{HilbertMaps}.}
    \label{fig:hilbert_map_example}
\end{figure}

We will base our discussions on features produced by projecting coordinates of interest to inducing points over spatial coordinates. These features have strong performance when used in HMs. They are known as ``hinged features'' as introduced in \citep{Senanayke:2017}, and similar to ``sparse features'' outlined in \citep{HilbertMaps}. We denote the $m$ inducing points $\hat{\mathbf{x}}_{1},\hat{\mathbf{x}}_{2},\ldots, \hat{\mathbf{x}}_{m}\in\mathbb{R}^{2}$ are spatially fixed inducing points. We are assumed to be given a dataset of $n$ coordinates, and labels of whether the sensor detects the coordinate to be occupied or not. That is, we have $\mathcal{D}=\{(\mathbf{x}_{i},y_i)\}_{i=1}^{n}$, where $\mathbf{x}_{i}\in\mathbb{R}^{2}$ is a coordinate in space and $y_{i}\in\{0,1\}$ is a binary variable indicating whether the coordinate is occupied or not. We begin the map-building by projecting coordinates to the inducing points, using a Gaussian radial basis function, specifically,

\begin{align}
\bm{\phi}(\mathbf{x})=[\phi(\mathbf{x},\hat{\mathbf{x}}_{1}),\phi(\mathbf{x},\hat{\mathbf{x}}_{2}),\ldots,\phi(\mathbf{x},\hat{\mathbf{x}}_{m})], && \phi(\mathbf{x},\hat{\mathbf{x}})=\exp(-\gamma\lvert\lvert\mathbf{x}-\hat{\mathbf{x}}\lvert\lvert_{2}^{2}), \label{eqn:bk_project_hm}
\end{align}
where $\gamma$ is a length-scale hyperparameter, which controls how strongly spatially-neighbouring coordinates influence one another. The probability that a coordinate $\mathbf{x}$ in the environment is unoccupied can then be expressed using a simple classifier as, 
\begin{equation}
p(y=1|\mathbf{x},\mathbf{w})=\big(1+\exp(\mathbf{w^\top \bm{\phi}(\mathbf{x})})\big)^{-1}, \label{eqn:bk_forward_hm}
\end{equation}
where $\mathbf{w}$ are weight parameters that are learnt from gathered data. The classifier can be trained by optimising a regularised binary cross-entropy loss:

\begin{align}
\mathcal{L}_{\mathbf{w}}(\mathcal{D})=\sum_{i=1}^{n}\Big[y_{i}\log(p(y_{i}=1|\mathbf{x},\mathbf{w}))+(1-y_{i})\log(1-p(y_{i}=1|\mathbf{x},\mathbf{w}))\Big]+\mathcal{R}(\mathbf{w}), 
\end{align}
where $\mathcal{R}(\mathbf{w})$ is the elastic-net regulariser, defined as,
\begin{equation}
\mathcal{R}(\mathbf{w})=\lambda_{1}\lvert\lvert\mathbf{w}\lvert\lvert_{1}+\lambda_{2}\lvert\lvert\mathbf{w}\lvert\lvert_{2}^{2},
\end{equation}
where $\lambda_1$ and $\lambda_2$ are regularisation hyperparameters, which control how strongly the loss is regularised. We can train Hilbert Maps efficiently online, with streaming data, via stochastic gradient descent. When we wish to predict the probability of coordinates being occupied, we can simply project the query coordinates to the inducing points with \cref{eqn:bk_project_hm}, and then compute the probabilities via \cref{eqn:bk_forward_hm}. An example of a built Hilbert Map along with the training data used is illustrated in \cref{fig:hilbert_map_example}.

\section{Robot Manipulator Kinematics}\label{kinematics}
This section provides a brief introduction to manipulator kinematics, which is crucial to understand many of the techniques in \cref{chap7} and \cref{chap8}. We begin by providing describing the configuration of a manipulator. Then, we introduce forward kinematics, mapping the joint configurations of the manipulator to the Cartesian coordinates of its end-effector. 

\subsection{Manipulator Configurations}
\begin{figure}[t]
    \centering
    \includegraphics[width=0.7\textwidth]{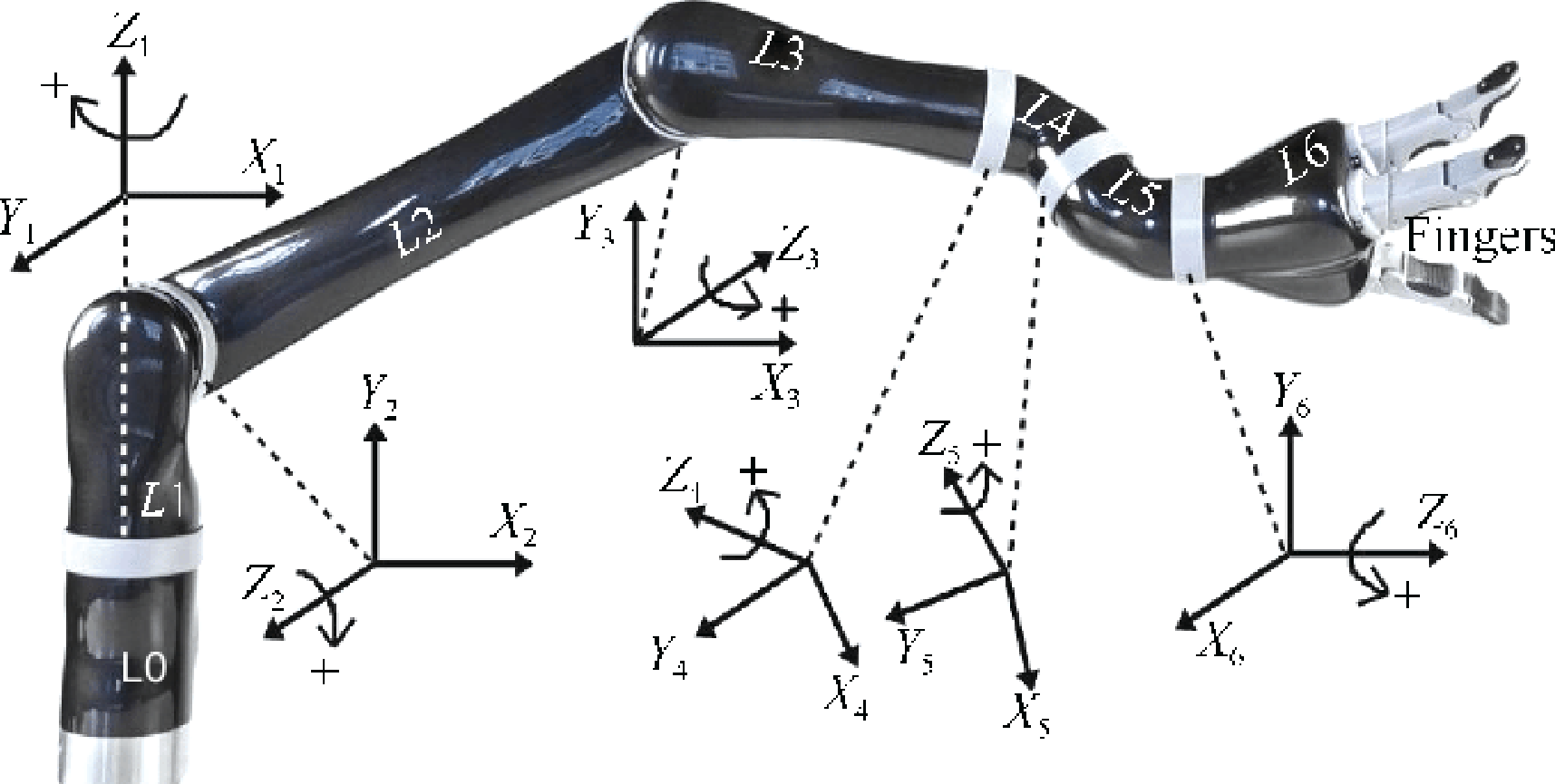}
    \caption{The JACO arm used in real-world experiments in \cref{chap7} and \cref{chap8} has 6 degrees of freedom (not including the gripper), one for each of 6 revolute joints. Figure adapted from \cite{jaco_arm_img}.}
    \label{arm_jaco}
\end{figure}
In this thesis, the manipulators explored are restricted to be fixed base, where the root link is connected to a stationary platform, and between each link, or between the base and a link, there is a single joint. These robot manipulator systems can be modelled by $n_{q}$ joints, $n_{q}$ links, along with a fixed-base. We denote the displacement for each joint as $q_{i}$, $i=1,\ldots,n_{q}$, where each joint typically has some upper and lower displacement limits. We describe the current state of such as robot by its \emph{configuration} $\mathbf{q}$, and the set of all feasible configurations as the \emph{Configuration Space (C-space)}, denoted as $\mathcal{Q}$. Specifically, 
\begin{align}
\mathbf{q}=\begin{bmatrix}
q_{1}\\
\vdots\\
q_{n_{q}}
\end{bmatrix},
&&
\text{where }\mathbf{q}\in \mathcal{Q}.
\end{align}
In this thesis, all of the manipulators contain revolute joints, and the configurations correspond to the rotation angles of the joints. The JACO manipulator used in the experiments of \cref{chap7} and \cref{chap8} has 6 degrees of freedom (not including the gripper), and is illustrated in \cref{arm_jaco}. In the absence of additional kinematic constraints, which are restrictions on the movements of components of the robot, the coordinates of the robot's C-space are \emph{minimal coordinates}, specifically, that its dimension is exactly the degrees-of-freedom of the system. 

Although we can specify the motion of a robot within its C-space coordinates, it can often be difficult to translate the geometry of the surrounding environment into C-space coordinates. Therefore, we would often reason about tasks in the robot's natural environment with \emph{task-space} coordinates, the Euclidean coordinate system with respect to a point on the robot, typically the end-effector. 

\subsection{Forward Kinematics}

The \emph{forward kinematics} then refers to the mapping between the configuration of the robot to the displacement and orientation (the pose). Note that in this thesis, we may on occasion restrict ourselves to only reasoning about the displacement of the end-effector, and discard the rotation from consideration. For brevity, we shall also loosely refer to the mapping between robot configurations and end-effector displacement only as ``forward kinematics''. The forward kinematics $f_{e}(\mathbf{q})$ can often be surjective -- there are many joint configurations that can result in the same end-effector displacement. Additionally, evaluating the forward kinematics is often efficient, and consists of a sequence of rigid body transformations, which only requires linear algebra and trigonometry. On the other hand, \emph{inverse kinematics}, mapping from end-effector coordinates to joint configurations is typically much more cumbersome and may require numerical optimisation. 
\begin{figure}[t]
    \centering
    \includegraphics[width=0.6\textwidth]{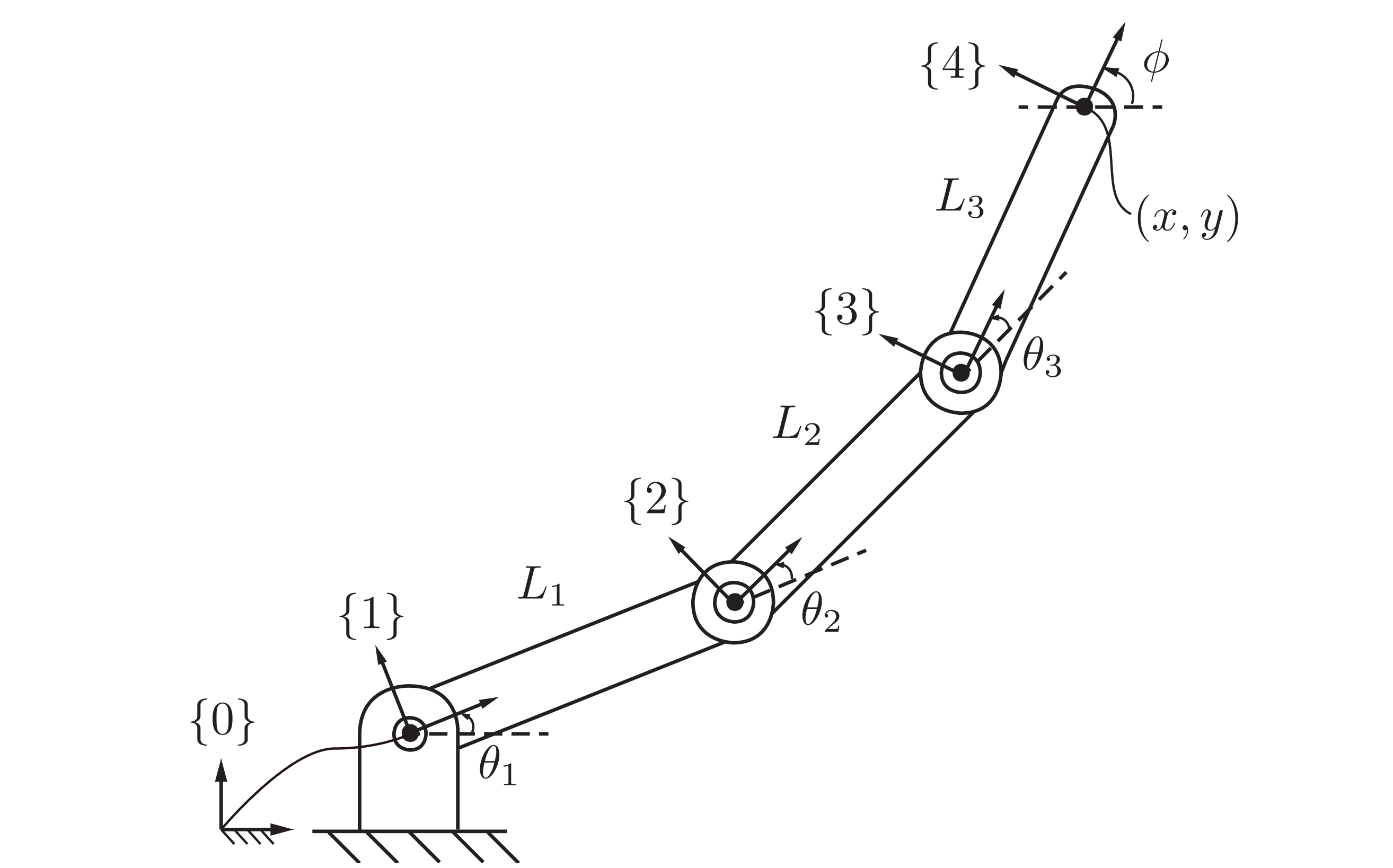}
    \caption{The 3 degrees of freedom planar manipulator used in the example. Example and figure from \cite{modern_robotics}.}
    \label{fig:example_manp_bk}
\end{figure}

Here, we give an example (from \cite{modern_robotics}) of constructing the forward kinematics of a simple planar manipulator with 3 links. Forward kinematics can be written as a sequence of transformation matrix products, starting from the manipulator base to the end-effector. Suppose the robot has 3 links of length $L_{1}$, $L_{2}$, $L_{3}$, and 3 controllable joints $\theta_{1},\theta_{2},\theta_{3}$ at the end of each link. The fixed frame reference at the origin is labelled as $\{0\}$, three link reference frames are respectively labeled $\{1\}$, $\{2\}$, $\{3\}$, and the reference frame at the end-effector as $\{4\}$. We define the robot configuration as $\mathbf{q}=(\theta_{1},\theta_{2},\theta_{3})$, and denote the end-effector orientation as $\phi$.  An illustration of the 3 degrees of freedom planar manipulator is shown in \cref{fig:example_manp_bk}. Then, we can define the transformation matrix between the reference frames as:

\begin{align}
    \mathbf{T}_{0,1}&=
    \begin{bmatrix}
\cos(\theta_{1}) & -\sin(\theta_{1}) & 0 & 0\\
\sin(\theta_{1}) & \cos(\theta_{1}) & 0 & 0\\
0 & 0 & 1 & 0\\
0 & 0 & 0 & 1\\
\end{bmatrix},
&& 
    \mathbf{T}_{1,2}&=
    \begin{bmatrix}
\cos(\theta_{2}) & -\sin(\theta_{2}) & 0 & L_{1}\\
\sin(\theta_{2}) & \cos(\theta_{2}) & 0 & 0\\
0 & 0 & 1 & 0\\
0 & 0 & 0 & 1\\
\end{bmatrix}\\
\mathbf{T}_{2,3}&=
    \begin{bmatrix}
\cos(\theta_{3}) & -\sin(\theta_{3}) & 0 & L_{2}\\
\sin(\theta_{3}) & \cos(\theta_{3}) & 0 & 0\\
0 & 0 & 1 & 0\\
0 & 0 & 0 & 1\\
\end{bmatrix},
&& 
\mathbf{T}_{3,4}&=
    \begin{bmatrix}
1 & 0 & 0 & L_{3}\\
0 & 1 & 0 & 0\\
0 & 0 & 1 & 0\\
0 & 0 & 0 & 1\\
\end{bmatrix}
\end{align}
We seek the forward kinematics, $f_{e}$, that maps our configurations $\mathbf{q}$ to positions and orientations $(x,y,\phi)$. We begin by finding the transformation matrix from reference frame $\{0\}$ to $\{4\}$, given by:
\begin{equation}
     \mathbf{T}_{0,4}=\mathbf{T}_{0,1}\mathbf{T}_{1,2}\mathbf{T}_{2,3}\mathbf{T}_{3,4}.
\end{equation}
The displacements in the $x,y$-plane of the end-effector, relative to the fixed-base origin, can be extracted from the matrix as $x=\mathbf{T}_{0,4}^{(1,4)}$ and $y=\mathbf{T}_{0,4}^{(2,4)}$, where the super-scripted tuples indicate the rows and columns of the matrix, respectively. The $2\times 2$ upper-left sub-matrix of $\mathbf{T}_{0,4}$, i.e. $\mathbf{T}_{0,4}^{(1:2,1:2)}$, is the rotation matrix of $\phi$. Hence, $\phi$ can be obtained via $\phi=\mathtt{atan2}(\mathbf{T}_{0,4}^{(2,1)},\mathbf{T}_{0,4}^{(1,1)})$, where $\mathtt{atan2}$ is the 2-argument arctan function \citep{c_standard}. Here the forward kinematics $f_{e}$, with end-effector orientation, is given by:
\begin{equation}
    \begin{bmatrix}
    x\\
    y\\
    \phi
    \end{bmatrix}=f_{e}(\mathbf{q})=
    \begin{bmatrix}
    \mathbf{T}_{0,4}^{(1,4)}\\
    \mathbf{T}_{0,4}^{(2,4)}\\
    \mathtt{atan2}(\mathbf{T}_{0,4}^{(2,1)},\mathbf{T}_{0,4}^{(1,1)})
    \end{bmatrix}
\end{equation}
We are also often interested in the velocities in both the task-space and the C-space. Let us denote the end-effector position coordinates as $\mathbf{x}_{e}\in\mathbb{R}^{3}$. The instantaneous velocities at a specific configuration $\mathbf{q}$ are linked via:
\begin{align}
    \dot{\mathbf{x}}_{e}=\mathbf{J}_{f_{e}}(\mathbf{q})\dot{\mathbf{q}} && \Longleftrightarrow && \dot{\mathbf{q}}=\mathbf{J}_{f_{e}}^{\dagger}(\mathbf{q})\dot{\mathbf{x}}_{e},
\end{align}
where $\mathbf{J}_{f_{e}}$ denotes the Jacobian of the forward kinematics. $\mathbf{J}_{f_{e}}^{\dagger}$ is a \emph{generalised inverse} of $\mathbf{J}_{f_{e}}$, as typically many configurations can result in the same end-effector displacement, i.e. $\mathbf{J}_{f_{e}}$ has more rows than columns. In particular, in this thesis, we shall use the Moore-Penrose \citep{penrose_1956}, which obtains the least squares solution to the over-determined system, where $\mathbf{J}_{f_{e}}^{\dagger}=(\mathbf{J}_{f_{e}}^{\top}\mathbf{J}_{f_{e}})^{-1}\mathbf{J}_{f_{e}}^{\top}$. 

\section{Dynamical Systems and Differential Equations}\label{sec:background_dyn_sys}
Throughout this thesis, we often model the motion of the robot as a dynamical system, described by an ordinary differential equation (ODE). A dynamical system represents how a \emph{state}, i.e. a collection of values that abstract the system at the current snapshot, evolves through time. A state can be viewed as a point in \emph{state-space}, the set of all possible configurations of a system. For example, we may wish to describe how the position of a point in the task-space of a manipulator evolves. Here, the state-space can be the 3d Cartesian coordinate system and the state a position coordinate. Beyond modelling in the task-space, we can construct dynamical systems in the configuration space of the robot manipulator, where a state of a robot is often given as its joint configurations. In particular, dynamical systems in configuration space are used to model manipulator motion in \cref{chap7,chap8}.

\subsection{Trajectories of Dynamical Systems}
\begin{figure}[t]
    \centering
    \includegraphics[width=0.5\textwidth]{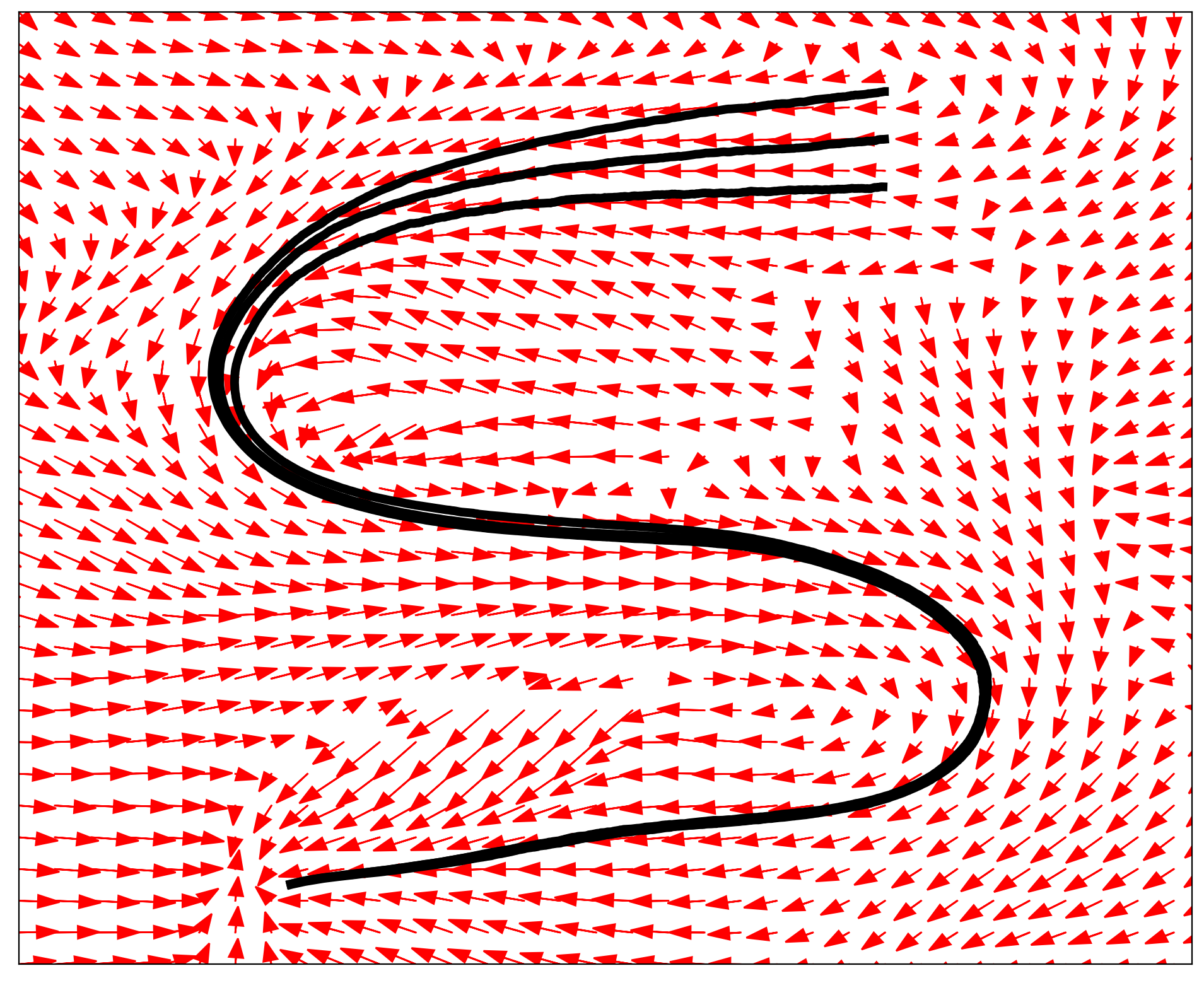}
    \caption{An example of a non-linear time-invariant dynamical system from \cref{chap7}, three trajectories are integrated are shown in black, and the dynamics are visualised as a vector field in red. This system is globally asymptotically stable, and the integrated trajectories converge.}
    \label{fig:example_nl_dynamical_system}
\end{figure}

In this thesis, we investigate \emph{continuous-time} dynamical systems, where the system does not commit to a fixed time resolution. We limit our discussion to \emph{first-order} systems, where the order of the time derivatives is at most one. Second-order dynamical systems arise in robotic systems when taking into account acceleration, such as in \cref{chap8}. However, these can be converted into a first-order system, by absorbing the state velocities and augmenting the state vector. A $d$-dimensional continuous-time first-order system with state-space $\mathcal{S}\subseteq \mathbb{R}^{d}$ and states $\mathbf{x}\in \mathcal{X}$ can be expressed as the initial value problem,

\begin{align}
    \dot{\mathbf{x}}(t)=f(\mathbf{x}(t),t), && \mathbf{x}(0)=\mathbf{x}_{0},\label{eq_back_dyn_sys}
\end{align}
where $\dot{\mathbf{x}}\in \mathcal{T}\mathcal{X}$ is the time derivative of the state, $t\in\mathbb{R}$ is a time variable, $\mathbf{x}_{0}\in \mathcal{X}$ gives the initial state of the system, and $f:\mathcal{X}\times \mathbb{R}\rightarrow \mathcal{T}\mathcal{X}$ is the system dynamics function. Here, we denote the tangent space of the state space as $\mathcal{T}\mathcal{X}$, i.e. the space of possible velocities for a particle in $\mathcal{X}$. Intuitively, we can think of $f$ as a vector field, mapping each coordinate in the state-space to a valid velocity. This vector field perspective of dynamical systems is revisited in \cref{chap7} when we introduce our Diffeomorphic Templates method. An example of a non-linear dynamical system along with three integrated trajectories is shown in \cref{fig:example_nl_dynamical_system}. The dynamics of the system can be visualised as a 2d vector field, illustrated in red.

Motion trajectories of a robot can be obtained from a dynamical system description of robot motion by ``rollouts''. Trajectories of \cref{eq_back_dyn_sys}, which we denote as $\xi(t,\mathbf{x}_{0})$, describe the state of the system at a given time after starting at a given initial condition and can be obtained via evaluating the integral:

\begin{equation}
    \xi(t,\mathbf{x}_{0})=\mathbf{x}_{0}+\int_{0}^{t}f(\mathbf{x}(s),s)\mathrm{d}s.\label{backg_ode_int}
\end{equation}

The tractability of this integral depends on the class of dynamical systems our system belongs to. If $f$ is independent of time, i.e. $f(\mathbf{x},t)=f(\mathbf{x})$ for all $t$, the system is known to be \emph{time-invariant} or \emph{autonomous}. If the dynamics are linear in $\mathbf{x}$, that is $f(\mathbf{x},t)=\mathbf{A}(t)\mathbf{x}$, then the system is known as a linear system. Linear time-invariant systems, i.e. systems of the form $\dot{\mathbf{x}}=\mathbf{A}\mathbf{x}$, admit closed-form integrals. However, linear time-invariant systems are very restricted in its ability to model real-world phenomena, and the dynamical systems discussed in this thesis are generally non-linear. The use of numerical ODE integrators are needed to evaluate \cref{backg_ode_int}.

Numerical ODE integrators discretise time, and recursively integrate the states to roll-out a trajectory of states. Integrators for first-order dynamical systems can be categorised as \emph{explicit} or \emph{implicit}: explicit methods express the state of the system in the future from the current state, while implicit methods require solving an algebraic expression which involves both the current state and the latter state. Explicit integrators are generally more efficient, but suffer from instabilities. Here, we shall outline explicit Euler's method and implicit Euler's method, which are some of the simplest examples of explicit and implicit methods. For both of these methods, we shall first specify a step-size $\Delta t$. In explicit Euler's method, we compute the states at the next step using the update: 
\begin{equation}
    \mathbf{x}(t+\Delta t)\approx\mathbf{x}(t)+hf(\mathbf{x}(t),t). \label{bg_eq_expl}
\end{equation}
The implicit Euler's method, on the other hand, considers the velocities at the next step. Specifically, the update equation is given as:
\begin{equation}
\mathbf{x}(t+\Delta t)\approx\mathbf{x}(t)+hf(\mathbf{x}(t+\Delta t),t+\Delta t). \label{bg_eq_impl}
\end{equation}
This \cref{bg_eq_impl} requires solving an algebraic equation to recover $\mathbf{x}(t+\Delta t)$. In practice, this is done with a root-finding algorithm such as Newton-Raphson's method \citep{newton_raph}. 

\subsection{Asymptotic Stability of Dynamical Systems} \label{bk_asymtotically_sec}
In the contributing \cref{chap7}, we require our method to preserve \emph{asymptotic stability}. Here, we shall give a brief definition on the asymptotic stability of systems. One often wishes to examine the long-term behaviour of dynamical systems -- some of the questions that we may seek to answer include: When we start at some initial conditions and integrate to roll-out trajectories, what happens to the states as time progresses? Do the trajectories converge to a single point, or multiple points, or fly off and diverge? How does the initial condition impact where a trajectory ends up?

\begin{figure}[t]
    \centering
    \includegraphics[width=0.49\textwidth]{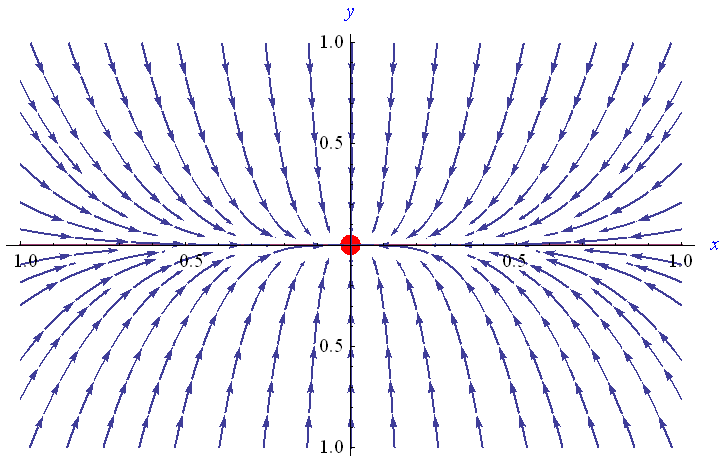}
    \includegraphics[width=0.49\textwidth]{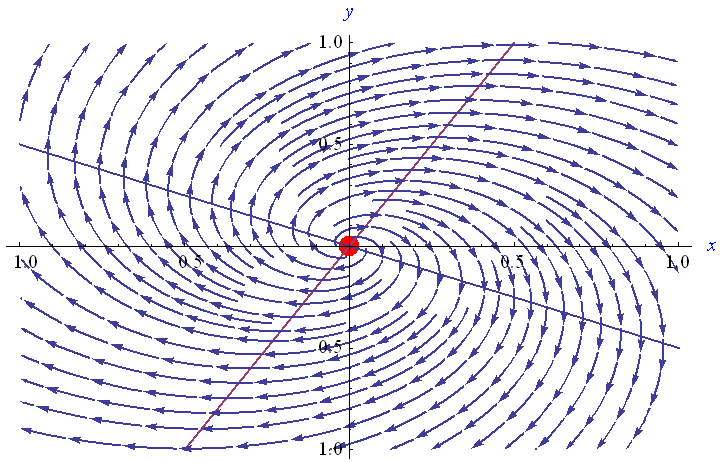}
    \caption{(Left) An example of a globally asymptotically stable equilibrium (an attractor); (Right) An unstable equilibrium (unstable spiral). The equilibrium points are shown in red, and dynamics are indicated by arrows. Figures from \cite{vector_field_spiral}.}
    \label{fig:stable_unstable_example}
\end{figure}

We shall describe the stability of a system with respect to equilibrium points. An equilibrium point, $\bm{x}^{*}$, is a point in state-space where velocity is zero, $\bm{x}^{*}\in\mathbb{R}^{d}: f(\bm{x}^{*})=0$. Additionally, a system is \emph{locally asymptotically stable} in a region $S\subset\mathbb{R}^{d}$ if trajectories starting in $S$ converge to some equilibrium $\bm{x}^{*}\in S$, 
\begin{align}
\lim_{t\rightarrow\infty}\xi(t, \bm{x}_{0})=\bm{x}^{*}, && \forall \bm{x}_{0}\in S. 
\end{align}
Furthermore, a system is \emph{globally asymptotically stable} if $S=\mathbb{R}^{d}$, and all trajectories converge to a unique equilibrium point. Alternatively, equilibrium is known to be unstable if a small perturbation to a particle at the equilibrium shall result in the particle being repelled from the equilibrium. An example
of a globally asymptotically stable equilibrium and one of an unstable equilibrium are shown in
\cref{fig:stable_unstable_example}. The system illustrated in \cref{fig:example_nl_dynamical_system} is also globally asymptotically stable.

\section{Summary}
In this chapter, we have introduced some of the basic conceptions which will resurface throughout this thesis. Machine learning methods can be generally described as approaches which leverage data to make predictions or find patterns without being explicitly given the rules of how to do so. The archetype of a machine learning problem is the regression problem, which we outline in \cref{reg_sec}, early on in this chapter. In \cref{linreg_background_sec}, we discuss linear regression, potentially used in conjunction with non-linear features, as one of the most straightforward approaches for solving regression problems. We illustrate with an example, in \cref{kernel_backg}, how dot products on non-linear features can be more efficient with the introduction of kernel functions. Next, in \cref{FC_networks}, we give an overview of fully-connected neural networks, the most fundamental of neural network models. Modern machine learning is heavily dominated by neural network models, which are highly flexible over-parameterised models, that are exceptionally parallelisable. Then in \cref{sec_optimise}, we provide background on optimisation techniques used in the thesis, These include (1) ADAptive Moment estimation (ADAM), an extension of Stochastic Gradient Descent that uses momentum, and is typically used to optimise neural network models; (2) Sequential Quadratic Programming, capable of efficiently finding local solutions for constrained optimisation problems where derivatives are available; (3) Covariance Matrix Adaptation Evolution Strategy, a black-box derivative-free optimiser. We have also lightly introduced background knowledge to related robotics topics. In \cref{sec:env_reps}, we describe methods to represent a robot's environment. These include occupancy grid maps, an early and widely-used model to represent the free and occupied space in an environment, along with Hilbert Maps, a method that avoids discretisation of the environment. Then, in \cref{kinematics}, we give some background on the essentials of robot manipulator manipulators. We elaborate on the foundational concept of a ``configuration space'', a vector space that consists of every possible geometric permutation of the robot. Then, we elaborate on \emph{forward kinematics} a mapping between the configuration space and the position of the end-effector. Finally, in \cref{sec:background_dyn_sys}, we give background on dynamical systems and differential equations, and discuss the asymptotic stability of dynamical systems. 

In the rest of the thesis, we shall develop our contributed robot learning methods, and the background presented in this chapter will be revisited throughout the contributing chapters. Coming up next is our first contributing chapter, \cref{chap3}, where we introduce our contributions to learning continuous occupancy representations and more importantly the fusion of these models.

\pagebreak

\part{Learning for Continuous Environment Representation}\label{part1}

\chapter{Fusion of Continuous Occupancy Maps}\label{chap3}\blfootnote{This chapter has been published in ICRA as \cite{HM}.}
\section{Introduction}
The deployment of multiple robots can deliver many benefits by allowing for the parallelisation of tasks. Compared with an individual robot, the deployment of multiple robots can provide speed improvements, and increase the robustness of the system, by reducing the dependency on any single robot. In particular, within a decentralised multi-robot system, there is no single central fusion center \citep{DurrantWhyte2008MultisensorDF}, hence the impact of an individual robot malfunctioning to the system is limited. In light of these advantages, this chapter examines continuous occupancy mapping in multi-robot systems. Building reliable representations of an environment is an integral part of the exploration of environments, a fundamental problem in mobile robotics. In a multi-robot scenario, the data collected by individual robots need to be integrated into a single consistent model of the environment \citep{ExplorationFox}. Real-time decentralised mapping using multiple robots~\citep{multi1} \citep{multi2} has various applications including agriculture \citep{multi_agri}, environmental monitoring \citep{multi_env1}, and disaster-relief \citep{dis_rel1} \citep{dis_rel2}. 

Historically, discrete grid maps, which assume independence between grid cells, have been used to represent the occupancy of the environment~\citep{OccupancyGridMaps}. Although the strong assumption of independence between cells allows for efficient operations on grid maps, it ignores the spatial dependency of the environment. As occupancy in the real world is continuous and not discretised into a grid with independent cells, an occupancy grid representation may not be able to adequately capture occupancy of the real-world. Gaussian process occupancy maps (GPOM)  \citep{SimonGPOM} were introduced as a method to build continuous occupancy maps, by using kernels to capture spatial dependencies between occupancy data. However, GPOMs do not scale efficiently, due to its cubic time complexity on the number of all data-points used, making it impractical to use real-time online on a robot. Another framework to continuously represent the environment, the Hilbert Maps (HM) framework~\citep{HilbertMaps}, was introduced as a much faster framework to represent the environment continuously, and does not require the retention of past data points. The Hilbert Map framework was subsequently extended to Bayesian Hilbert Maps (BHM)~\citep{Senanayke:2017} to capture the uncertainty of parameters in the model. However, training BHMs require the inversion of a covariance matrix between parameters, a computationally expensive operation of cubic time complexity. Due to the long run-times needed for this operation, multi-robot mapping with BHMs in real time is impractical. To reduce run-time, we introduce Fast Bayesian Hilbert Maps (Fast-BHM) that remove the need to invert a covariance matrix, significantly speeding up the training.

Motivated by the advantages that continuous occupancy mapping and multi-robot mapping could provide, this chapter explores decentralised multi-robot merging of continuous occupancy maps, within variations of the Hilbert Map framework. The main contributions of this chapter are:

\begin{enumerate}
\item Developing Fast-BHM, a significantly sped-up variation of the Bayesian Hilbert Map model \citep{Senanayke:2017}.
\item Developing methods to fuse Fast-BHMs models, which can be built by individual robots, to obtain a unified Fast-BHM model.
\end{enumerate}

The chapter is organised as follows. In \cref{Fast}, we introduce the Fast-BHM model, as a significantly sped-up variation of the BHM model, mitigating the impractical run-times required to train BHMs. This is followed, in \cref{MergeSec}, by the presentation of a decentralised scheme to fuse Fast-BHMs. Empirical results highlighting the effectiveness of our merging scheme, and the speed improvement of Fast-BHMs are then shown in \cref{experiments}. 

\begin{figure}[t]
    \centering
    \begin{subfigure}[h]{0.45\textwidth}
        \centering
        \includegraphics[width=0.95\textwidth]{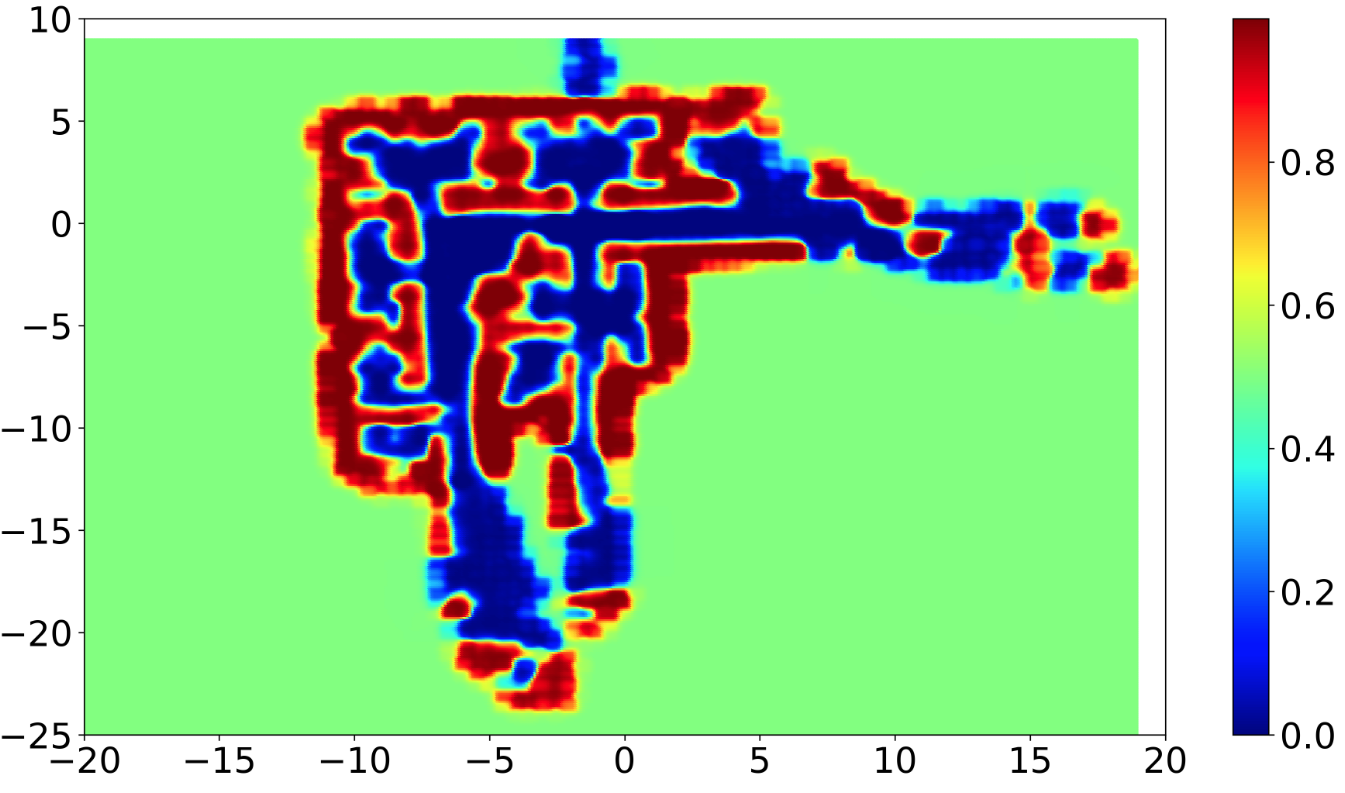}
        \caption{ Upper-Left Sub-map}
    \end{subfigure}%
    ~ 
    \begin{subfigure}[h]{0.45\textwidth}
        \centering
        \includegraphics[width=0.95\textwidth]{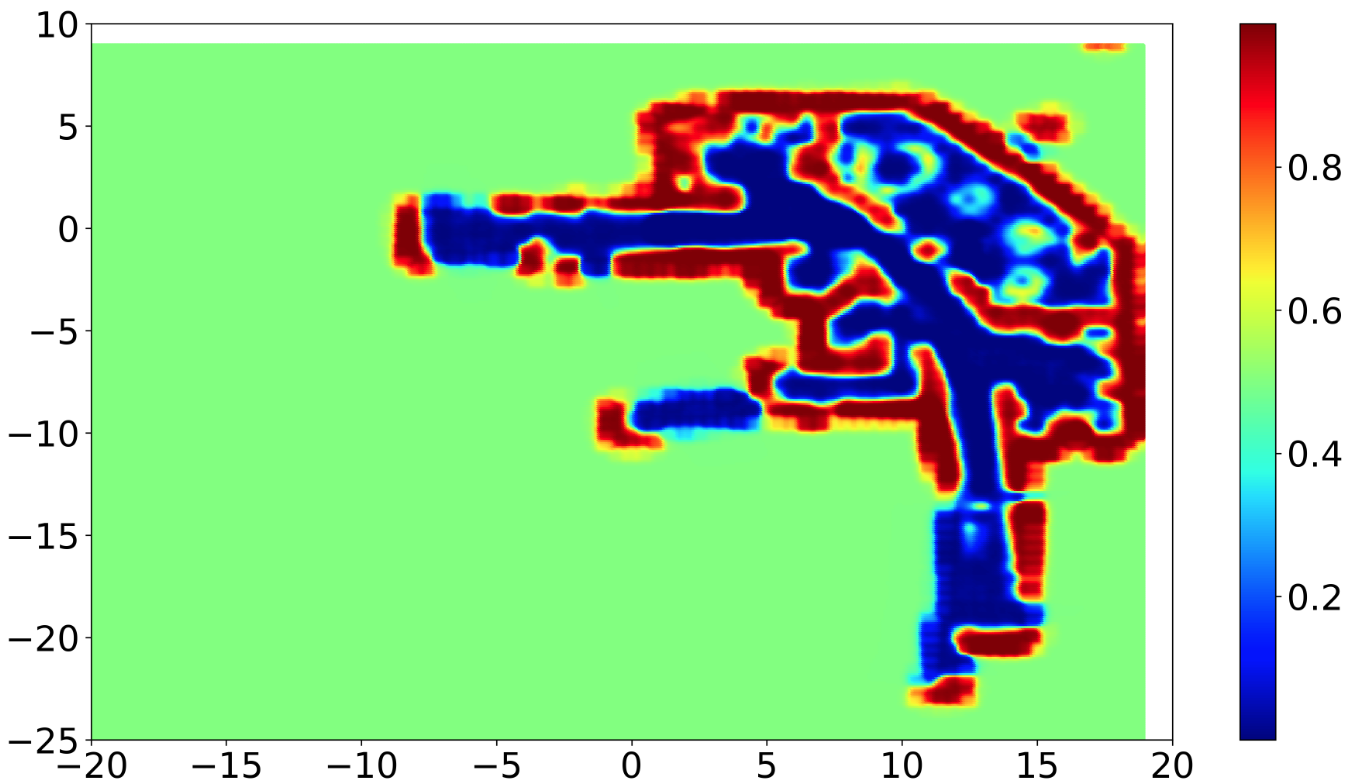}
        \caption{ Upper-Right Sub-map}
    \end{subfigure}
    
    \begin{subfigure}[h]{0.45\textwidth}
        \centering
        \includegraphics[width=0.95\textwidth]{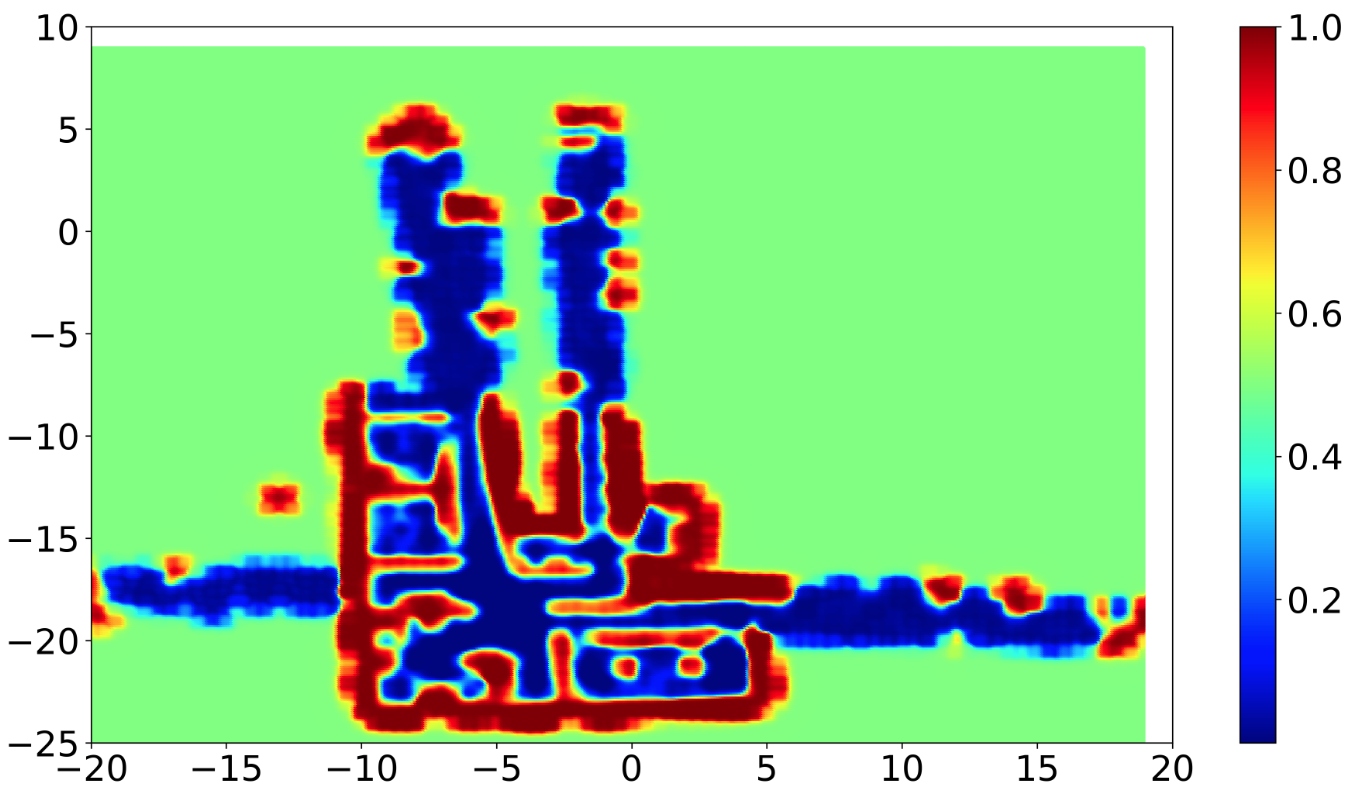}
        \caption{ Lower-Left Sub-map}
    \end{subfigure}%
    ~ 
    \begin{subfigure}[h]{0.45\textwidth}
        \centering
        \includegraphics[width=0.95\textwidth]{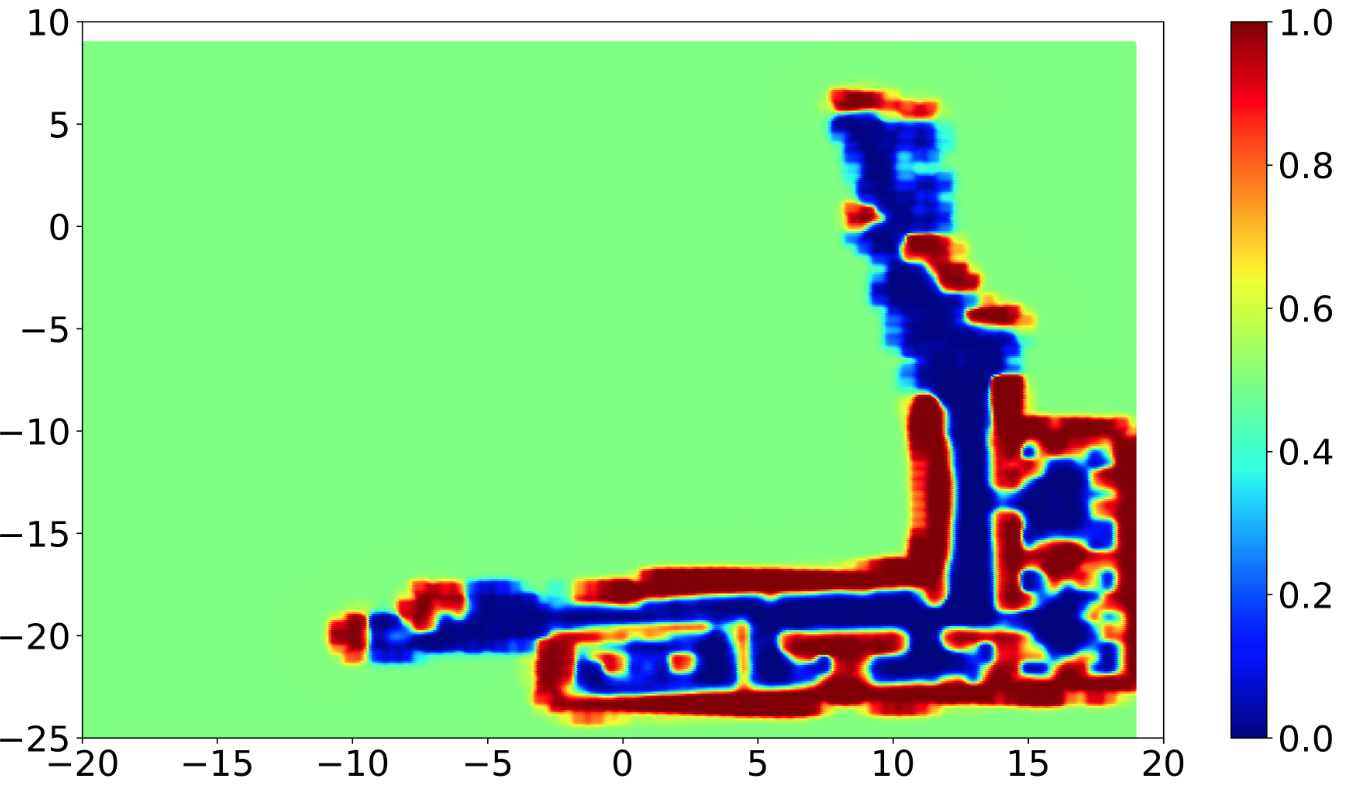}
        \caption{Lower-Right Sub-map}
    \end{subfigure}
    \caption{ Sub-maps trained on scans from different sections of the Intel \citep{Intel} dataset. In a multi-agent set-up, we require a method to fuse the sub-maps into a unified model.}
    \label{FigSubMaps}
\end{figure}

\section{Related work}


Map fusion aims to build a consistent representation of the environment based on data collected and maps built by different agents. As each individual robot moves around the environment, it builds local sub-maps based on the data it receives. These individual sub-maps can then be periodically fused to obtain a global map. Decentralised fusion methods allow each node the ability to build a global map, without relying on a centralised node \citep{DurrantWhyte2008MultisensorDF}. Each robot can obtain a local copy of the global map, potentially providing each individual robot with information about regions that it has not yet explored, and continue to refine the map with additional data. A well-known approach to building multi-robot occupancy grid maps with known poses is through concurrent exploration and updating via Bayes filters \citep{BayesFilter}. Methods have also been developed for multi-robot mapping using grid maps where the pose is unknown \citep{unknownPose} \citep{ThurnUnknownPose}. 

Attempts have also been made to fuse predictions from continuous occupancy representations, under the assumption of known pose and local measurements are mapped to a global reference frame. These include fusing predictions from Gaussian Process Maps \citep{Jadidigpfusion} \citep{Kim2014RecursiveBU} using Bayesian Committee Machines \citep{BCM} and iteratively fusing the predictions made by Hilbert Maps using an update rule \citep{3dMerge}. These methods obtain local representations of the environment, then either pre-select sample points to query from, and combine the estimates from the individual local estimations, or represent the global model as a discretised grid \citep{3dMerge}. Unlike these methods which aim to fuse predictions from individual models, our method merges the underlying local Fast-BHM models to arrive at a unified and compact global Fast-BHM model. This has the advantage of enabling the points to be queried to be decided after the fusion occurs, freeing users from having to define sample points {\em a priori} to build a grid map from queries. After this model is obtained, we can query any point in the environment.

\section{Fast Bayesian Hilbert Maps}\label{Fast}
Bayesian Hilbert Maps (BHM) have been introduced as an extension to Hilbert Maps \citep{Senanayke:2017}. We refer the reader to \cref{subsec_background_hm} for background on Hilbert Maps. Unlike Hilbert Maps, BHMs do not heavily depend on regularisation, eliminating the need to pick regularisation hyperparameters. Bayesian Hilbert Maps (BHM)~\citep{Senanayke:2017} are obtained under the assumption that weights approximately follow a multivariate normal distribution, $p(\mathbf{w})\approx Q(\mathbf{w}) = \mathcal{N}(\mathbf{w}|\bm{\mu},\bm{\Sigma})$. An Expectation-Maximisation (EM) \citep{Dempster77maximumlikelihood} -like approach is then used to iteratively learn the parameters. The update of the parameters requires the inversion of the full covariance matrix, $\bm{\Sigma} \in \mathbb{R}^{T \times T}$. This operation has a time complexity of approximately $O({T}^{3})$. The inversion of $\bm{\Sigma}$ hinders the usage of Bayesian Hilbert Maps in real-time on robots -- it is not practical to train Bayesian Hilbert Maps in real time for large open environments. Furthermore, the full covariance matrix can also be relatively large, requiring good bandwidth to conduct multi-robot map fusion. We shall now develop Fast Bayesian Hilbert Maps (Fast-BHM), which enables the maps to be trained without the need to invert the covariance matrix. 

With the aim of avoiding the usage and inversion of large matrices in mind, we propose a variation of BHMs which assumes independence between weights, Fast Bayesian Hilbert Maps (Fast-BHMs). Fast-BHMs enable us to train BHMs without constructing the full covariance matrix. As each weight in the logistic regressor, used in the original Hilbert Maps formulation \citep{HilbertMaps}, is simply a scalar value and does not depend on one another, we assume that each weight of the Bayesian Hilbert Map can be approximated as a normal distribution and independent of one another, giving a mean-field Gaussian distribution: 
\begin{equation}
p(\mathbf{w})=\prod_{t=1}^{T}p(w_t)\approx \prod_{t=1}^{T}\mathcal{N}(w_t|\mu_t,\sigma_{t}^2), 
\end{equation}
where $T$ is the number of weights present in the Bayesian Hilbert Map. Then the update equations for Fast-BHM can be derived in a manner similar to the derivation of the variational logistic regression \citep{Bishop:2006}. It is important to note that the weights are parameters of a continuous function, assuming that these weights are independent is not the same assumption as the cell independence assumption made in occupancy grid maps. The parameters are learnt in an EM approach \citep{Dempster77maximumlikelihood}. See the derivation of the variational logistic regression \citep{Bishop:2006}, \citep{vi2} for details.

We denote $\mathbf{y}$ as a vector of binary variables indicating the occupancy of coordinates. $\bm{\phi}(\mathbf{x})$ denotes a feature, obtained by applying a kernel transformation on data point $\mathbf{x}$. We take the variational logistic regression paradigm outlined in chapter 10 of \cite{Bishop:2006} and reproduce the framework here. The weights parameterising the model are denoted as $\mathbf{w}$, and a vector containing intermediate variables $\mathbf{z}\in\mathbb{R}^{N}$, where $N$ is the number of data points for the training batch. Each intermediate variable (each element in $\mathbf{z}$) is used to produce a lower bound on the sigmoid function:
\begin{align}
    \sigma(\hat{z})&\geq \sigma(z)\exp\big\{\frac{(\hat{z}-z)}{2}-\lambda(z)(\hat{z}^{2}-z^{2})\big\}, \text{ for any }\hat{z}\in\mathbb{R},\\
    &\text{where, }\lambda(z)=\frac{1}{2z}\big[\sigma(z)-\frac{1}{2}\big].
\end{align}
Under the {\em variational inference} framework \citep{vi2}, \citep{vi1}, we aim to maximise a lower bound to $\log p(\mathbf{y}\vert \bm{\phi}(\mathbf{x}))$ by learning parameters to $h(\mathbf{w},\mathbf{z})$, a lower bound of $p(\mathbf{y}|\bm{\phi}(\mathbf{x}),\mathbf{w})$,

\begin{equation}
\begin{split}
\log p(\mathbf{y}\vert \bm{\phi}(\mathbf{x}))
&=\log\int p(\mathbf{y}|\bm{\phi}(\mathbf{x}), \mathbf{w})p(\mathbf{w}) \mathrm{d}\mathbf{w}\\
&\geq \log \int h(\mathbf{w},\mathbf{z})p(\mathbf{w})\mathrm{d}\mathbf{w}
\end{split}
\end{equation}
\textbf{E-step:}
\begin{equation}
Q(\mathbf{z},\mathbf{z}^{old})=\mathbb{E}\big[\log\big(h(\mathbf{w},\mathbf{z})p(\mathbf{w})\big)\big]
\end{equation}
\textbf{M-step:}
\begin{equation}
\mathbf{z}=\arg \max_{\mathbf{z}}Q(\mathbf{z},\mathbf{z}^{old})
\end{equation}
We determine the parameters by maximising the lower bound of the marginal likelihood. We can derive the lower bound of the marginal likelihood by building on a lower bound of the sigmoid logistic, used by the authors introducing Bayesian Hilbert Maps \citep{Senanayke:2017}, and noted in \citep{Bishop:2006}. Similar to the Sequential Bayesian Hilbert Maps method \citep{Senanayke:2017}, we wish to train the model sequentially, using scans from the current time step to train weights estimates obtained from the previous step. Following the derivation of sequentially trained BHMs, we assume that the prior can be written as $p(\mathbf{w})=\prod_{t=1}^{T}\mathcal{N}(w_t|\mu_{t,k-1},\sigma_{t,k-1}^2)$, where $k$ is the current time step. 
Closely following the derivation touched on in \citep{Senanayke:2017}, and detailed in \citep{Bishop:2006}, we arrive at the update formulae for the E-step:
\begin{equation}
\mu_{t,k}=\sigma_{t,k}^2\big(\sigma_{t,k-1}^{-2}\mu_{t,k-1}+\sum_{n=1}^{N_k}(y_n-0.5)\bm{\phi}(x_{n,t})\big)\\
\end{equation}
\begin{equation}
\begin{aligned}
\sigma_{t,k}^{-2}=&\sigma_{t,k-1}^{-2}+2\sum_{n=1}^{N_k}\lambda(z_n)\bm{\phi}(x_{t,n})^2\\
\text{for all } &t\in\{1,...,T\}
\end{aligned}
\end{equation}
There are $N_{k}$ data points in the scan of time $k$, and $T$ number of weights in total.
The updating formula for the M-step is obtained as:

\begin{equation}
z_{n}^{2}=\sum_{t=1}^{T} \bm{\phi}(x_{n,k})^{2} (\sigma_{t,k}^{2}+\mu_{t,k}^{2})
\end{equation}
Where $n$ is an index in ${1,2,...,N_{k}}$. These equations allow us to train Fast-BHMs. The bottleneck for Fast-BHMs in the method is matrix multiplication, which is of sub-quadratic complexity relative to the cubic complexity of matrix inversion. In the next section, we present a method to merge Fast-BHM maps in a decentralised manner.

\section{Merging Bayesian Hilbert Maps}\label{MergeSec}
We develop methods to combine several individual Fast-BHMs trained on different, but potentially overlapping, data points into a single Fast-BHM. The merged Fast-BHM itself is a continuous representation of the environment. It can be trained further with new data and merged with other Fast-BHMs. Denoting the probability density functions of weights as $f$, the merging of several individual Fast-BHMs into a single BHM can be written as:

\begin{equation}
f(\mathbf{w}|\cup_{m=1}^{M}D_m)=\mathbb{F} [f(\mathbf{w}|D_1),...,f(\mathbf{w}|D_{M})],\\
\end{equation}
where there are $M$ individual Hilbert Maps, and there are $N_m$ data points used to train the $m^{th}$ Fast-BHM is expressed as $D_m=\{\mathbf{x}_i^m,y_i^m\}_{i=1}^{N_m}$. $\mathbb{F}$ denotes the method used to merge Fast-BHMs. 

We assume that the individual maps being merged shared the same set of features, and therefore the size of each weight vector, $f(\mathbf{w}|D_{m})$, and the merged vector, $f(\mathbf{w}|\cup_{m=1}^{M}D_m)$, will be of the same size. For the resultant merged weights to be constructed as a Bayesian Hilbert Map, we also need to maintain that the merged weights of the resultant map are normally distributed. 

We can see that the difficulty of merging the weights from individual maps lies in combining local estimates of weights when the estimates differ, as there is common information between the two local estimates of the same weight  \citep{DurrantWhyte2008MultisensorDF}. We also need to maintain that the individual weights of the merged map are Gaussian. Consider the merging of two BHMs:

 \begin{equation}
 \begin{aligned}
 f(\mathbf{w}|D_1\cup D_2)&=\frac{f(D_1\cup D_2|\mathbf{w})f(\mathbf{w})}{f(D_1\cup D_2)},\\
 \text{where } f(D_1\cup D_2|\mathbf{w})&\propto \frac{f(D_1|\mathbf{w})f(D_2|\mathbf{w})}{\underbrace{f(D_1\cap D_2|\mathbf{w})}_ \text{Common Information}}. 
 \end{aligned}
 \end{equation}

\subsection{Conflation}

We will now introduce the concept of Conflation \citep{Hill:2008} as a method to fuse distributions of random variables. The Conflation operation can be used to consolidate results from independent experiments that estimate the same quantity. Conflation is defined in Definition \ref{Def1}. The original authors prove Conflation minimises the maximum loss in Shannon Information and is the best linear unbiased estimate. Before the first fusion of any sub-maps, we assume that the valid local estimates of a particular weight are conditionally independent only on the global weight. This is based on the assumption that the data points retrieved from the sensor are assumed to be independent, and are free from any biases of the sensor. This assumption holds fairly well, as demonstrated by the results of conflation in our experiments. After the initial merger of separate Hilbert Maps, we may wish to continue training on copies of the merged map. In this scenario, we assume that the global map from the previous merge is the common information between the different updated copies in the next merge.\\

\begin{definition}[\citep{Hill:2008}]\label{Def1}
Let $X_1,X_2,...,X_n$ have PDFs $f_1,f_2,...,f_n$ satisfying $0<\int_{-\infty }^{\infty }\prod_{i=1}^{n} f_i(x) \mathrm{d}x < \infty $. Then the conflation $\&(X_1,X_2,...,X_n)$ is continuous with density:
\begin{equation}
f(x)=\frac{f_1(x)f_2(x)...f_n(x)}{\int_{-\infty }^{\infty}f_1(y)f_2(y)...f_n(y) \mathrm{d}y}
\end{equation}
\end{definition}

\bigbreak
As the product of the probability density functions (PDF) of normal random variables is the PDF of another normal distribution, $f_1(y)f_2(y)...f_n(y)$ is the PDF of a normal distribution, and the integral with respect to $y$ is equal to 1. Therefore, the conflation of $N$ normally distributed random variables is simply the product of the PDF of the variables, $f_{Merged}(x)\approx \&(f_1,...,f_N)=\prod_{n=1}^{N} f_{n}(x)$, which is also a PDF of a normal distribution. This property allows multiple cycles of training and merging to occur.

The fusion to arrive at the global estimate of a particular weight $w_{global}$ from $n$ valid local estimates, independent conditioned on the true value of the global weight, can be written as:
\begin{equation}\label{Conflate}
w_{global}\approx\&(w_1,..,w_n)\sim\mathcal{N}\Bigg(\frac{\sum_{i=1}^{n}\frac{\mu_i}{\sigma_{i}^2}}{\sum_{i=1}^{n}\sigma_{i=1}^{-2}},\frac{1}{\sum_{i=1}^{n}\sigma_{i}^{-2}}\Bigg)
\end{equation}


\begin{algorithm}[h]
\caption{Fast-BHM Sequential Fusion}\label{SeqMergeAlgo}
\textbf{Input:} \\
\text{$N$ sub-maps to fuse, $\mathbf{w}_1,...,\mathbf{w}_N$.;} \\
\text{Number of merges before stopping, $M$}
\text{Map Fusion at Robot $n$, with map $\mathbf{w}_n$}

\For{$m \in \{1,...,M\}$}{
\For{$i\in \{1,...,N\}$}{
$\mathbf{w}^{old}_i \gets \mathbf{w}_i$\;
$\mathbf{w}_i  \gets Train(\mathbf{w}_i)$\;
\If{$m = 1$ or $i=n$}{
$\Delta \mathbf{w}_i\gets\mathbf{w}_i$\;
}{
$\Delta \mathbf{w}_i\gets GetIncrement(\mathbf{w}_i, \mathbf{w}_i^{old})$\;
}
}
$\{\mathbf{w}_1,...,\mathbf{w}_N\} \gets Combine(\{\Delta \mathbf{w}_1,...,\Delta \mathbf{w}_N\})$\;
}
\textbf{Output: }$\{\mathbf{w}_1,...,\mathbf{w}_N\}$

\end{algorithm}

Conflation assumes that the estimates were computed independently, and using it directly without removing the common information may result in double counting. We assume that the global estimate from the previous merge is the only common information, and empirically demonstrate this assumption holds in \cref{experiments}. 

From the theory of conflations for normal distributions \citep{Hill:2008}, the combined estimate of the weights, $w$, given independent datasets $D_1$ and $D_2$ can be written as:
\begin{equation}
f(w|D_1\cup D_2)\approx \&[f(w|D_1),f(w|D_2)]=f(w|D_1)f(w|D_2)
\end{equation}

Suppose any resultant weight from additional training after the map merge (with PDF $f(w|D_{M}\cup D_{N})$) can be approximated as the conflation between the previous weight of the merged map (with PDF $f(w|D_{M})$), and a normally distributed increment, with PDF $f(w|D_{N})$, resulting from the new data. 

\begin{equation}\label{Merge}
\begin{aligned}
f(w|D_{M}\cup D_{N})&\approx \&[f(w|D_{M}),f(w|D_{N})]\\
&=f(w|D_{M})f(w|D_{N})
\end{aligned}
\end{equation}

From \cref{Conflate} and \cref{Merge}, we can arrive at the parameters for the weight increment estimates, given the new data as:

\begin{equation}
\Delta w \sim \mathcal{N} \Bigg(\frac{(\sigma_{M}^{2}+\sigma_{N}^{2})\mu_{M\cup N}-\sigma_{N}^{2}\mu_{M}}{\sigma_{M}^{2}},\frac{\sigma_{M\cup N}^{2}\sigma_{M}^{2}}{\sigma_{M}^{2}-\sigma_{M\cup N}^{2}}\Bigg)
\end{equation}

Where $\Delta w\sim\mathcal{N}(\mu_{N},\sigma_{N})$ is the estimate of the increment of weight from the new data, $w_{M}\sim\mathcal{N}(\mu_{M},\sigma_{M})$ is the global weight estimate from the previous merge, and $w_{M\cup N}\sim\mathcal{N}(\mu_{M\cup N},\sigma_{M\cup N})$ is the local estimate after further training on a copy of the previous global estimate. 

Unless an agent has not used its sub-map for a merge before, it stores a copy of the global weights from the previous fusion until the next fusion. At the next fusion, each local estimate is compared to the copy of the previous global estimate, and estimates of the increment of weight are found and used in the merging. Algorithm \ref{SeqMergeAlgo} outlines the repeated merging process, where \textit{GetIncrement} returns the estimated increment of weight, \textit{Train} trains the map model with additional data starting with the inputted parameters, and \textit{Combine} merges the weights assuming there is no common information.

\subsection{Filtering Out Uncertain Local Estimates of Weights}
As there are physical coordinates associated with the features, if there are no Lidar scans close to the feature coordinates used to train the Fast-BHM, the weights assigned to those features will have parameters very close to the initial values. We can filter out local estimates of weights that are very uncertain, by defining a threshold for the variance of a weight. When several sub-maps are combined together, local estimates of weights with variances below the threshold are used for the fusing to obtain the global estimates. For any particular weight in the vector of weights, if there is only one reasonably confident local estimate, the global estimate of that weight will have the same mean and variance of the local estimate with acceptable variance. If none of the local estimates for a particular weight is below the threshold, the global estimate for the weight will be the default mean and variance values.

\section{Experiments}\label{experiments} 
We conduct experiments to empirically study:
\begin{enumerate}
\item The relative performance, and training time required, for Bayesian Hilbert Maps (BHM) and Fast-BHM
\item The quality of Fast-BHMs trained on separate scans of the environment, then fused together using the scheme described in \cref{MergeSec}
\item Whether errors that occur during fusion will accumulate as local maps are repeatedly fused,
\item The performance of Fast-BHMs relative fusion methods that require discretisation of a continuous representation, when the size of information transmitted is constrained.
\end{enumerate}
Experiments are conducted on the Intel Dataset \citep{Intel}, the Freiburg Campus Dataset \citep{Radish}, and a simulated dataset of two robots in an indoor environment. Unless stated otherwise, we use 5600, 9408, and 4900 features respectively when experimenting with the aforementioned datasets. We use the Area under Receiver Operating Curve (AUC) as the performance measure \citep{AUCROC}. 
\subsection{Fast-BHM vs BHM}
Experiments were conducted to investigate the relative performances of Fast Bayesian Hilbert maps (Fast-BHM) and Bayesian Hilbert Maps. The results of the performance and run-time, of a Python implementation of both models, are tabulated in \cref{FullComp}.  

\begin{table}[h]
\centering
\caption{ {Each map representation with the performance measure, and the time needed to train on 90\% of the scans in the Intel Dataset \citep{Intel}}}
\begin{tabular}{|l|l|l|}
\hline
                       & \textbf{Fast-BHM} & \textbf{BHM} \\ \hline
\textbf{AUC}           & 0.95$\pm$0.04          & 0.96$\pm$0.03        \\ \hline
\textbf{Training Time} & 7 min    & 339 min  \\ \hline
\end{tabular}
\label{FullComp}
\end{table}

We see that the time required to train up a Fast-BHM is significantly less than that needed for a BHM, while the performance of the Fast-BHM is only marginally weaker, as measured by the area under the Receiver Operating Characteristic (ROC) curve. The training times were achieved by running python implementations on a typical desktop computer, and demonstrate the significant relative speed-up Fast-BHMs achieve. Being able to create maps quickly is crucial for autonomous robots in the field. As a covariance matrix of weight parameters is no longer kept, the size of information needed to be communicated required between each agent to define the model is also smaller for Fast-BHMs.

\subsection{Fusion of Fast-BHM}\label{Fusion}
Experiments were conducted to evaluate the performance of our fused global map. A baseline to compare the performance of our fusion scheme is a single Fast-BHM trained on all the data gathered by the individual robots. This baseline value would be the best possible result achievable, indicating that no information was loss or misinterpreted during the fusion. As assumptions and approximations were made during the fusion process, we would expect a slightly lower performance. We also wish to compare a global map obtained from multiple fusions, with one from just a single fusion, and determine whether the errors due to approximations and assumptions made during each merge would accumulate, or be corrected by subsequent training.

We split the scans in the Intel Dataset \citep{Intel} into training and test sets at a ratio of 9:1. Using the training set, the average $x$ and $y$ coordinates of each data point in the scan is calculated. Based on the average coordinates, each scan is categorised to be in one of the following quadrants: Upper-Left; Upper-Right; Lower-Left; Lower-Right. The split for the Intel dataset \citep{Intel} is illustrated in \cref{FigSubMaps}. This is equivalent to running individual robots in those four quadrants. A separate Fast-BHM for each quadrant is then trained. As the split was made considering the average of all the data-points in a scan, there will be data-points that overlap with scans belonging in other quadrants. In the repeated fusion experiment, sub-maps of each quadrant are merged together upon completing training on 25\%, 50\%, 75\% and 100\% of the training data. For obtaining a map after a single merge, the sub-maps of each quadrant are merged when 100\% of training has been used. The Freiburg Campus dataset \citep{Radish} was treated similarly, but with training and testing sets split at a ratio of roughly 4:1. The training and testing sets for the simulated dataset were split at a roughly 9:1 ratio. Scans from the simulated did not need to be divided into quadrants, as the simulated scenario was two robots collecting scans.

\Cref{PerformanceComp} compares the Area Under the Curve (AUC) metric of the different global map models obtained by querying the test set. The accuracies of the fused maps were high across all datasets examined, indicating that assumptions and approximations made during fusion were sound, and the fusion method proposed is well suited to fuse local Fast-BHM models. If we compare the performances of a global map obtained from a single merge and that from repeated merges. We see that repeated merging results in a global map that performs approximately as well as a map from only one merge. This indicates that errors from merges do not accumulate, as further training on a global map will help amend errors in the global estimates.

\begin{figure}[h]
    \centering
    \begin{subfigure}[h]{0.49\textwidth}
        \centering
        \includegraphics[width=0.95\textwidth]{figures/chap4/LastMerge.png}
        \caption{ Global Fast-BHM from merger of sub-maps}
    \end{subfigure}%
    ~
    \begin{subfigure}[h]{0.49\textwidth}
        \centering
        \includegraphics[width=0.95\textwidth]{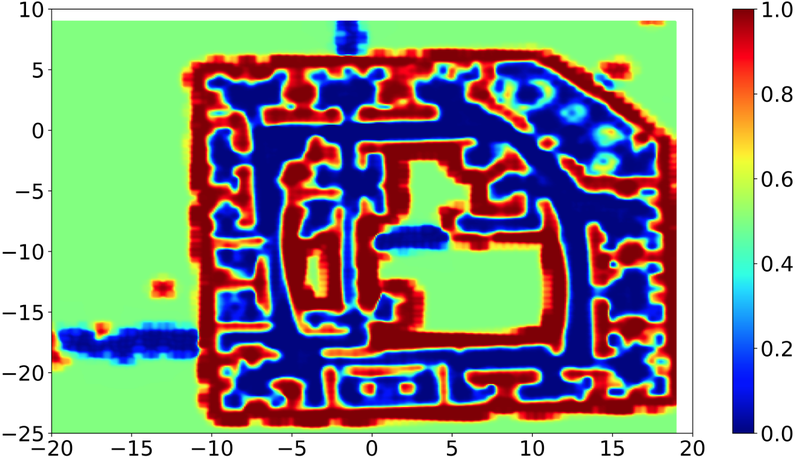}
        \caption{Single Fast-BHM trained on training set}
    \end{subfigure}
    \caption{ Merged vs entirely trained Global Fast-BHM representations (Intel Dataset)}
    \label{FigTrained}
\end{figure}

\begin{figure}[h]
    \centering
    \begin{subfigure}[h]{0.49\textwidth}
        \centering
        \includegraphics[width=0.95\textwidth]{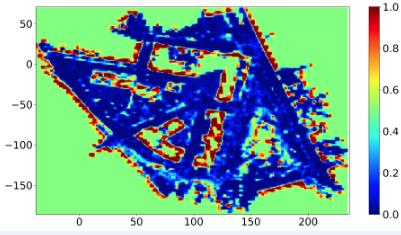}
        \caption{ Global Fast-BHM from merger of sub-maps}
    \end{subfigure}%
    ~
    \begin{subfigure}[h]{0.49\textwidth}
        \centering
        \includegraphics[width=0.95\textwidth]{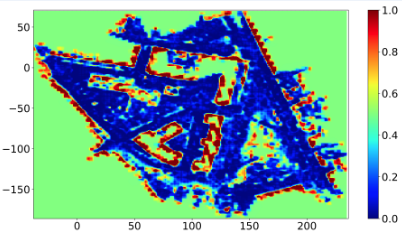}
        \caption{ Single Fast-BHM trained on training set}
    \end{subfigure}
    \caption{ Sub-maps trained on scans from the Freiburg Campus dataset}
    \label{SimSubmaps}
\end{figure}

\begin{figure}[h]
    \centering
    \begin{subfigure}[h]{0.49\textwidth}
        \centering
        \includegraphics[width=0.95\textwidth]{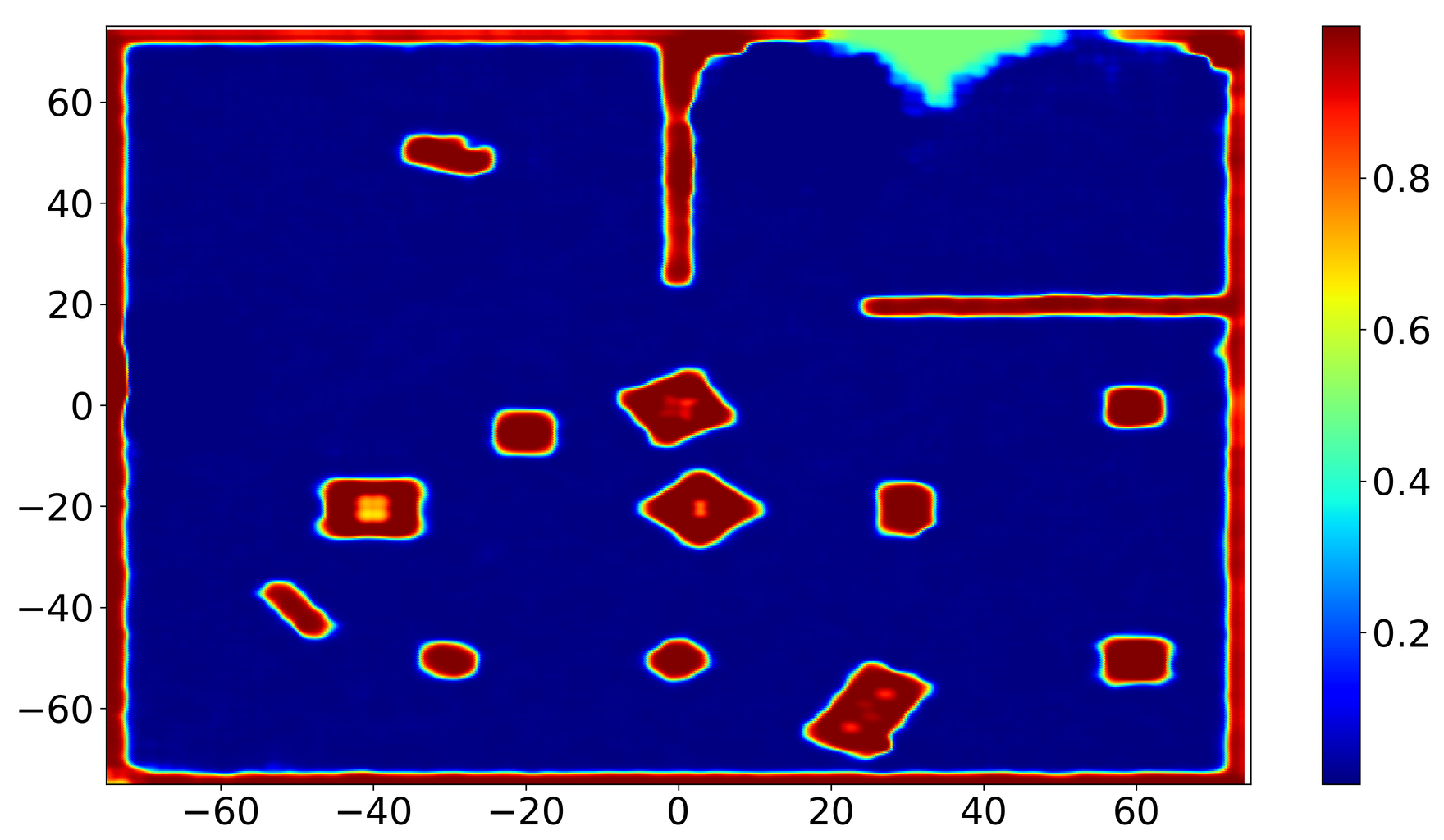}
        \caption{ Global Fast-BHM from merger of sub-maps}
    \end{subfigure}%
    ~
    \begin{subfigure}[h]{0.49\textwidth}
        \centering
        \includegraphics[width=0.95\textwidth]{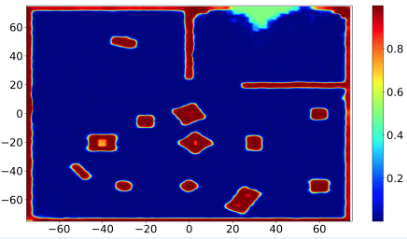}
        \caption{ Single Fast-BHM trained on training set}
    \end{subfigure}
    \caption{ Merged vs entirely trained Global Fast-BHM representations (Simulated Environment)}
    \label{FigTrainedSim}
\end{figure}

Figures \ref{FigTrained}, \ref{SimSubmaps}, and \ref{FigTrainedSim} show the resultant merges of local sub-maps and Fast-BHMs trained on data from all the agents. There are close to no visual differences in the merged and trained global Fast-BHMs. It can be seen that our method to conduct map fusion can produce a global map very similar to a map obtained by training the union of all the separate data scans. The conflation of local weight estimates has proven to be a reasonable approximation for the global weight values. 


\begin{table}[t]
\centering
\caption{ Area under ROC curve of both global maps obtained from different methods, and different datasets}
\begin{tabular}{|l|l|l|}
\hline
\textbf{Dataset}           & \textbf{Fusion Type} & \textbf{AUC} \\ \hline
\multirow{3}{*}{Intel}     & No Fusion            & 0.95$\pm$0.03   \\ 
                           & Fuse Once            & 0.94$\pm$0.04   \\ 
                           & Repeatedly Fused     & 0.94$\pm$0.03   \\ \hline

 \multirow{3}{*}{Freiburg Campus} & No Fusion            & 0.97$\pm$0.02   \\ 
                           & Fuse Once            & 0.89$\pm$0.090   \\ 
                           & Repeatedly Fused     & 0.90$\pm$0.03   \\ \hline
\multirow{3}{*}{Simulated} & No Fusion            & 0.99$\pm$5e-4 \\ 
                           & Fuse Once            & 0.99$\pm$5e-4 \\ 
                           & Repeatedly Fused     & 0.99$\pm$5e-4 \\ \hline
\end{tabular}
\label{PerformanceComp}
\end{table}

\begin{figure}[t]
        \centering
        \includegraphics[width=0.75\textwidth]{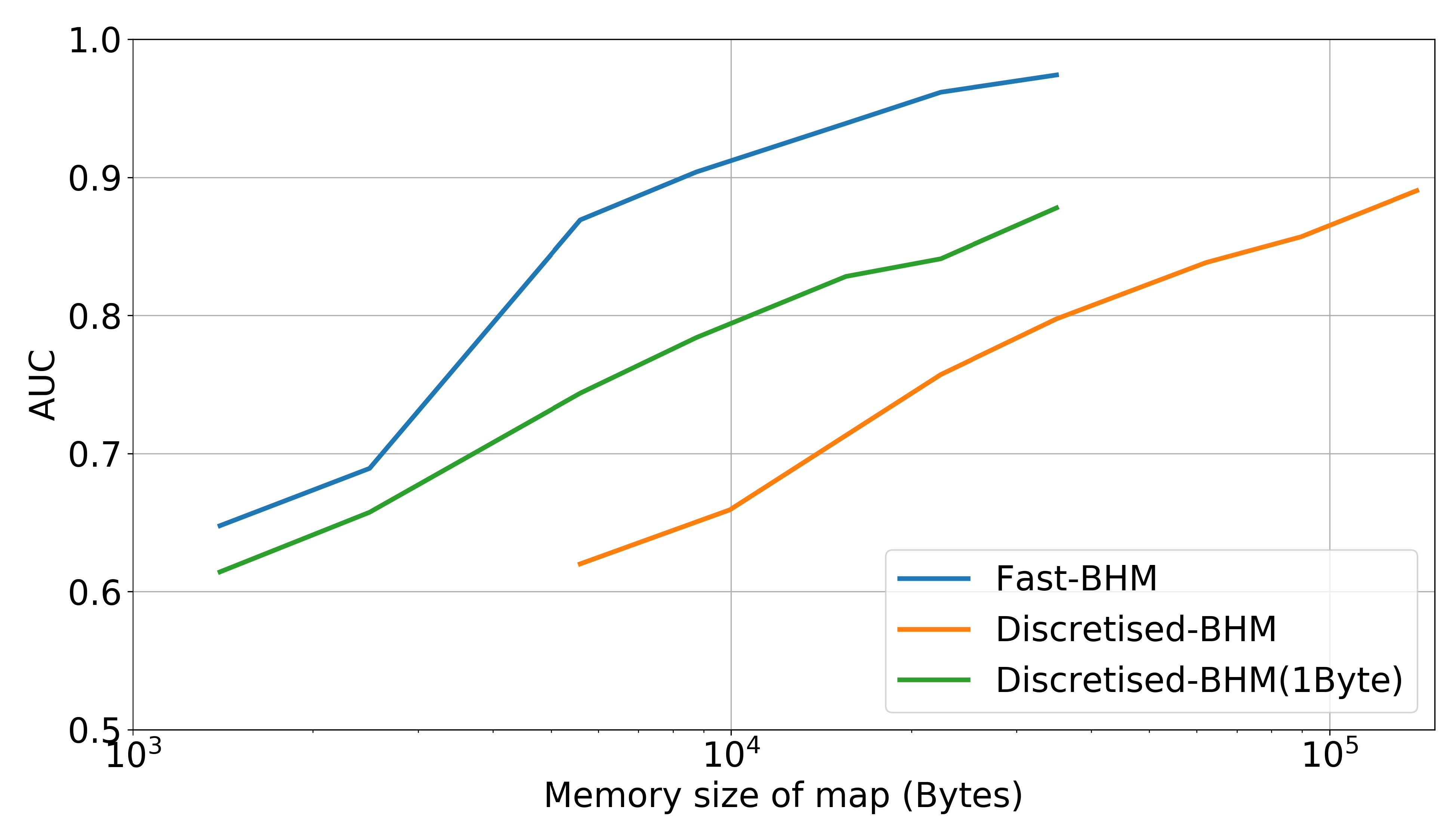}
    \caption{ AUC of the global estimate with constrained communication (Intel Dataset)}
    \label{ComparisonDiff}
    \vspace{-10pt}
\end{figure}

\begin{table}[t]
\centering
\caption{ The performance of our proposed method on merging sub-maps using the Intel \citep{Intel} dataset, using 5600 features, and the performance of method from \citep{3dMerge}, with a square cell size of 10cm}
\begin{tabular}{|l|l l l|}
\hline
                         & \textbf{AUC} & \textbf{Precision} & \textbf{Recall} \\ \hline
\textbf{Proposed Method} & 0.936$\pm$0.03   & 0.819$\pm$0.03         & 0.919$\pm$0.03      \\ 
\textbf{Method from \citep{3dMerge}}     & 0.916$\pm$0.02   & 0.864$\pm$0.03         & 0.866$\pm$0.15      \\ \hline
\end{tabular}
\label{Performance Comparison}
\end{table}

\subsection{Continuous vs Discrete Fused Map Representation}
The result of our fusion scheme is a continuous Fast-BHM model, that can be queried at any valid coordinate. Whereas, previously proposed methods \citep{Kim2014RecursiveBU} \citep{Jadidigpfusion} \citep{3dMerge} looked into merging the output predictions from separate models rather than fusing the model itself. Such methods require an {\em a priori} selection of points to query. In particular, the method proposed by \cite{3dMerge} requires a grid map to be defined and used to sample a continuous occupancy representation, such as a Hilbert Map, discretising the global map, and a fusion update equation is used to merge outputs of Hilbert Maps. We shall compare this approach to our map fusion approach.  

Continuous Fast-BHM models are more compact than grid cell representations. We want to explore the performances of continuous and discretised occupancy maps when the amount of information allowed to be communicated to each model is restricted. Using the Intel~\citep{Intel} dataset, We incrementally build a map representation with the fusion method described in \citep{3dMerge} on the outputs from Fast-BHMs, each using 10000 features, then discretising with differing cell sizes. We also explore using our proposed method to fuse sub-maps of Fast-BHMs, with different numbers of features. We can then plot the size of the map representation against the AUC measure of the fused map, as shown in \cref{ComparisonDiff}. We can see that Fast-BHMs outperform the discretised grid maps at any given memory size considered. This is true for both occupancy grid cells that used a floating point to denote the probability, and an efficient implementation of grid maps \citep{ROS}, which stores an approximation of the probability in 1 Byte. The compactness of Fast-BHMs is particularly of use when the amount of information we can communicate is limited. 

The performance of both methods on merging sub-maps using the Intel \citep{Intel} dataset is detailed in \cref{Performance Comparison}. Although the method from \citep{3dMerge} has a higher precision, our method has a higher recall and a higher overall AUC measure. Although in this case, both methods have similar performance, Fast-BHM is roughly 44.8 KB in size, while the grid map has a size of 560 KB.

\section{Summary}

This chapter introduces Fast-BHMs (Fast-BHMs) as a significantly sped-up version of Bayesian Hilbert Maps (BHMs), and presents a method to merge Fast Bayesian Hilbert Maps in a multi-robot scenario. Our merging method utilises the distributions of parameters to produce a unified global model, which is itself a Fast-BHM. We empirically demonstrate that Fast-BHMs are significantly faster than BHMs. We then demonstrate that our map fusion method is able to produce accurate global estimates, which are compact in size and can be easily transmitted when communication is limited. 

In the next chapter, we extend beyond representing the occupancy of the environment and develop learning methods to continuously represent the distribution of movement directions in an environment over time. This representation allows us to expand beyond static environments, and reason about how dynamic objects in the environment move.

\pagebreak

\chapter{Continuous Spatiotemporal Maps of Motion Directions}\label{chap4}\blfootnote{This chapter has been published in IEEE RA-L as \cite{sptemp}.}
\section{Introduction}

Robots need a representation of the environment to operate autonomously. Building static maps that capture the occupancy of an environment without considering the dynamic objects is well-studied in robotic mapping. However, simply mapping occupancy in a static environment, using for example standard occupancy grid maps \citep{OccupancyGridMaps}, does not adequately capture the dynamics in real-world environments. Robots, in particular autonomous vehicles, often need to operate in urban environments with moving objects, such as crowds of people and other moving vehicles. The understanding of movement directions can shed light on representing the trajectories of these dynamic objects in the environment. This work addresses the problem of understanding the trajectories of dynamic objects by building a directional probabilistic model that is continuous over space and time. The method provides a probability distribution over possible directions, which includes angular uncertainty, making our method a valuable tool for safe decision-making. With the proposed method, we can predict the possible directions a dynamic object would take in a given location at a given time. Then, we can use the predicted directional uncertainty to generate trajectories that dynamic objects typically take. Knowledge of such typical trajectories sheds light on long-term dynamics in an environment, improves path planning \citep{LongTermDynamicsPath}, and can plausibly be applied in autonomous driving in cities \citep{LongTerm3D}. 

Recent advances in building maps for trajectory directions \citep{DirectionalGridMaps} have been able to capture the multi-modality of uncertainty distributions associated with predictions of directions. However, grid map based methods, such as the recently presented method of \citep{DirectionalGridMaps}, require the environment to be discretised into independent grid cells specified \emph{a priori}, and interaction occurs between data points in neighbouring cells. This assumption is unrealistic and a significant impediment to the performance, as in the real world directional flows are largely continuous. In this work, we present a method to build continuous spatiotemporal maps, overcoming these limitations. This is a more realistic model of the real world, as in the natural world, movement directions are not split into uniform grids. The continuous representation provides significant improvements in performance. Taking inspiration from advances in continuous occupancy mapping \citep{HilbertMaps}, we utilise approximate kernel matrices to project data into a high-dimensional Reproducing Kernel Hilbert Space (RKHS) to build scalable continuous maps of directional uncertainty. Our proposed method does not require discretisation of the environment, and is able to be queried at arbitrary resolution after training. Most previous attempts to build maps of directions have assumed that the distribution of directions at a selected coordinate in space is stationary in time, or model how point values change over time. There have been methods to model periodic spatiotemporal changes in occupancy \citep{Krajnk}, or model temporal patterns with directions as discrete values \citep{TAROS}. Our method extends spatiotemporal representations to beyond a scalar value, to allow for multi-modal distributions as outputs.

Motivated by the limitations discussed above, we develop a method to build a map of directional uncertainty that addresses the drawbacks mentioned. We generate spatiotemporal kernel features, then pass the features into a Long Short-Term Memory (LSTM) \citep{lstm} network, and use the hidden representation to train a mixture density network \citep{Bishop94mixturedensity} of directional von Mises distributions. The proposed method builds a model with the following desirable properties: 

\begin{enumerate}
    \item models the probability distribution of movement directions over a valid  support of $[-\pi, \pi)$, and can be multi-modal; 
    \item represents the environment in a continuous manner, without assuming the discretisation of the environment into a grid of fixed resolution, with independent cells;
    \item takes into account how the probability distribution of directions changes over time, extending the map to become spatiotemporal. 
\end{enumerate}

\section{Related Work}

The method provided in our work is aimed at learning the long-term dynamics of an environment. There have been attempts \citep{TOG} to extend occupancy mapping beyond static environments by storing occupancy signals over time, and then building a representation along time in each grid. However, these methods are memory intensive and assume that time can be discretised into slices. Further extensions \citep{Conditional,IOHMM,GridBased} of learning how occupancy changes over time have also included various improvements to predicting changing occupancy maps over time.

Instead of modelling a changing occupancy map over time, methods \citep{OCallaghan2011LearningNM,embeddings} have also been developed to understand long-term occupancy by modelling trajectories. \cite{OCallaghan2011LearningNM} uses Gaussian Process learning to infer the direction an object takes in different areas of the environment. A major drawback of this method was its inability to represent multiple directions. \cite{embeddings} uses filtering with Kernel Bayes Rule, and determines the components of a mixture of Gaussian distributions. Both of these methods do not ``wrap'' around a circle, and can predict differing probability densities at $-\pi$ and $\pi$. As \cite{OCallaghan2011LearningNM} and \cite{embeddings} apply non-parametric models, the scalability of these methods is limited. Recently, a parametric directional grid maps approach \citep{DirectionalGridMaps} was proposed to model directional uncertainty in grid cells as a mixture of von Mises (VM) distributions, which capture multi-modal distributions, and have a support of $[-\pi,\pi)$. The authors showed that the directional grid maps method outperforms the Gaussian Process method presented in \cite{OCallaghan2011LearningNM}. The directional grid maps method \citep{DirectionalGridMaps} relies on \emph{a priori} discretisation of the environment into grid cells. This method allows the directional uncertainty in each grid cell to be learnt, which is assumed to take into account trajectories over different periods of time. However, the performance of this method may be hindered by the assumption of independent grid cells, and the assumption that the uncertainty is stationary over time. Both of these assumptions are relaxed in our proposed method. 

A similar method of fitting semi-wrapped Gaussian mixture models using EM is also presented in \cite{Flow}. This method also attempts to achieve a ``dense representation'' that is continuous over space, by combining predictions from several cells via imputation methods. There have also been some recent attempts to model scalar values which can be periodic in time, such as occupancy \citep{Krajnk} or a direction category \citep{TAROS}, using the Fourier Transform. Our spatiotemporal model differs from these methods in that allows for multi-modal distributions as outputs. 

\section{Methodology}
\begin{figure}
    \centering
    \includegraphics[width=.9\textwidth]{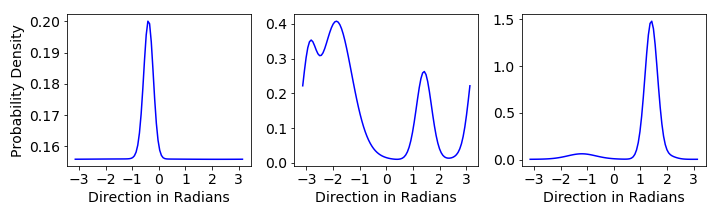}
    \caption{Example distributions of directions in different coordinates in space and time. The directions are in radians.}
    \label{distExample}
\end{figure}

\subsection{Problem Formulation}
We aim to build continuous directional maps over space and time. Let us consider a dataset, denoted as $\mathcal{D} = \{(x_m,y_m, t_m,\theta_{m})\}^{M}_{m=1}$, where $x_m$ and $y_m$ denote the longitude and latitude of the data point in space, $t_m$ denotes the time step at which the $m^{\text{th}}$ observation was made, and $\theta_m \in [-\pi,\pi)$ denotes the angular direction of a trajectory at this point. Given specific coordinates in space and time:  $(x^{*},y^{*},t^{*})$, we want the model to give us the probability distribution of the angular directions $p(\theta|x^{*}, y^{*}, t^{*})$. Essentially, our problem is to learn a mapping between a coordinate in space-time and a multi-modal distribution. Some example output distributions are shown in fig. \ref{distExample}.

\subsection{Overall Architecture of Proposed Model}
\begin{figure}[h] 
    \centering
    \includegraphics[width=.6\textwidth]{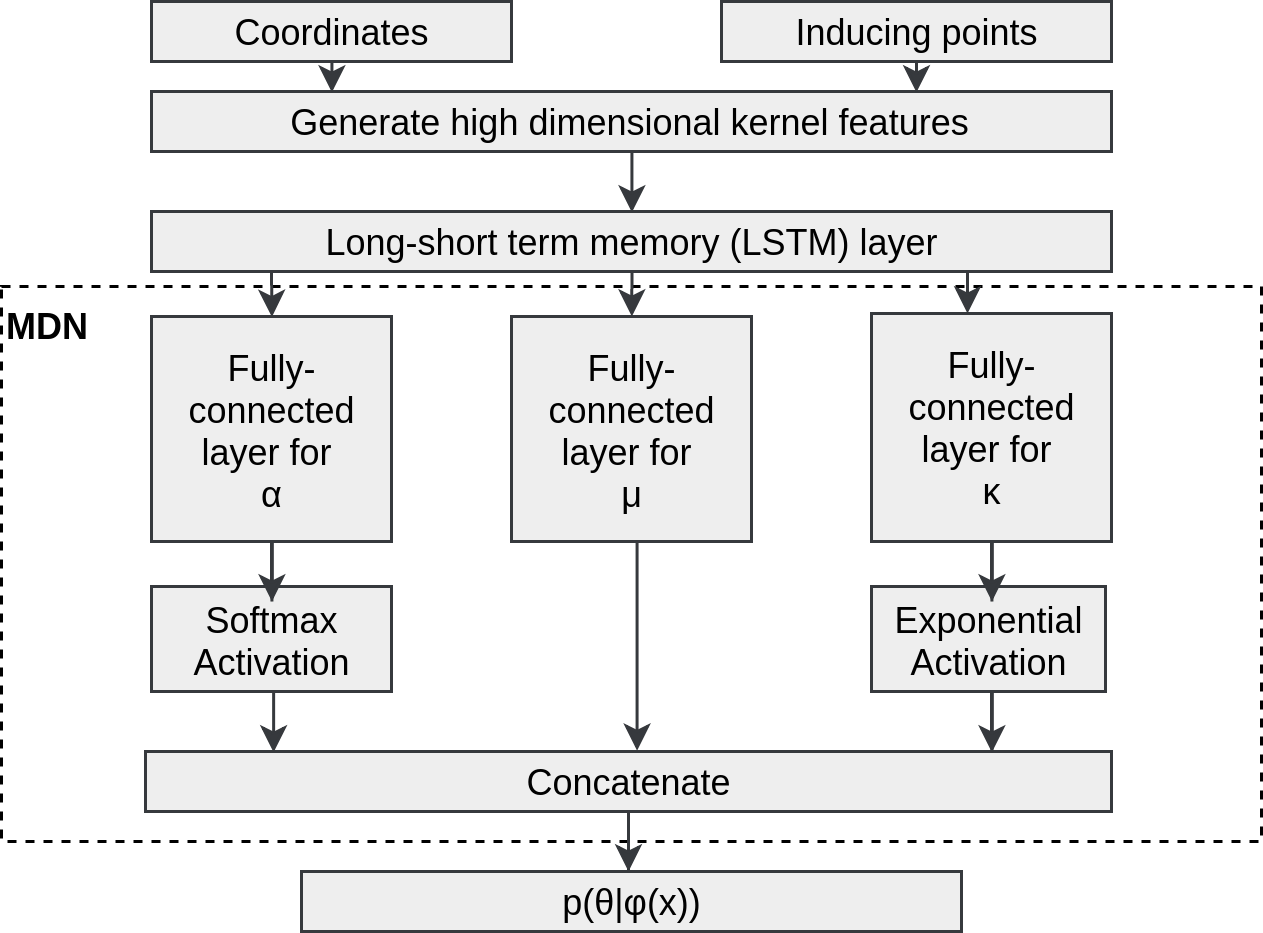}
    \caption{The architecture of our model. high-dimensional approximate kernel features, generated from an inputted coordinate with predetermined inducing points, are passed through a LSTM network, followed by a MDN.}
    \label{fig:Architecture}
\end{figure}

A brief overview of our model is given in this subsection, and important components of the model are discussed in the following subsections. Fig. \ref{fig:Architecture} summarises the main architecture of our model. Given input data and a set of coordinates, known as inducing points, we calculate high-dimensional kernel features. The features are then treated as a sequence and passed through a LSTM network. The hidden representation of the LSTM network is inputted to a mixture density network (MDN) to capture a multi-modal distribution. Fully connected layers in the MDN jointly learn the parameters $\alpha$, $\mu$, and $\kappa$, and approximate a mixture of von Mises distributions. We can then evaluate the probability distribution function of trajectory direction uncertainty given an arbitrary coordinate. Note that the coordinates inputted are a 3-dimensional vector, that provides a coordinate in space and time. Previous methods of mapping direction, such as those in \cite{DirectionalGridMaps} and \cite{OCallaghan2011LearningNM}, assume that the target distributions of directions do not change in the map over time. If we make the same assumption, and only require a continuous directional map in space independent of time, the high-dimensional features over a set of coordinates, known as inducing points, over space are inputted directly into the MDN for training. In section \ref{Experiments}, we will demonstrate the benefits of a spatiotemporal model over a purely spatial model.     

\subsection{Generating high-dimensional Features Over Space}\label{HighDim}
high-dimensional kernel features over space encode the influence each training data point holds on predictions in neighbouring points in space. This allows us to achieve maps that are continuous over space. We generate a large number of features from the inputted data that approximate the covariance matrix between $N$ data points, known as the kernel matrix, which is commonly used in kernel machines \citep{hofmann}. In this work, we approximate a kernel by projecting data to a set of inducing points, similar to the sparse Gaussian Process method \citep{SparseGP}. Inducing points can be thought of as pseudo-inputs which input coordinates are compared against. Inducing points are the centre of basis functions which are used to construct the continuous function of output parameters. We operate on the covariance between our input data points and our inducing points, rather than the covariance between each data point. 

We start by defining a set of $M_s$ inducing points over space, denoted by $\bar{\mathbf{x}}_1 \ldots \bar{\mathbf{x}}_{M_s}$. In our experiments, we place the inducing points with fixed spacing along longitude and latitude. $\mathbf{x}\in\mathbb{R}^2$ and $\bar{\mathbf{x}}\in\mathbb{R}^2$. We generate high-dimensional feature vectors from inputs by evaluating a kernel function, $k(\mathbf{x},\bar{\mathbf{x}})$, between the input and each inducing point. We denote the high-dimensional features, with spatial inducing points, as $\bm{\phi}_{s}(\mathbf{x})\in\mathbb{R}^{M_s}$. 

Similar to \cite{HilbertMaps}, the Radial Basis Function (RBF) kernel \citep{LearningWithKernels} is chosen to generate features in space. This kernel function is defined as 
\begin{equation}
    k(\mathbf{x},\bar{\mathbf{x}}) = \exp \left( - \frac{\Vert \mathbf{x}-\bar{\mathbf{x}}\Vert^{2}_{2}}{2\mathcal{\ell}^2}  \right), 
\end{equation}
where $\mathcal{\ell}$ is the length scale hyperparameter of the kernel. Intuitively, $\mathcal{\ell}$ controls how far the influence of a single training data point extends. Low $\mathcal{\ell}$ values indicates ``close''. In the next section, we will extend our high-dimensional features to include time.

\subsection{From Spatial to Spatiotemporal}
This subsection is motivated by the need to extend our continuous spatial model into a continuous spatiotemporal model. There are two steps involved in extending or model to spatiotemporal:
\begin{enumerate}
    \item Generate high-dimensional features in the space and time domains. This allows us to construct a sequence of spatial features over time;
    \item Pass the spatiotemporal features into a LSTM network to learn a more compact representation of temporal variations and use this compact representation in our training
\end{enumerate}

Spatiotemporal kernel features are needed to generate high-dimensional features that account for the influence of each training example in both space and time. A natural way to build spatiotemporal kernels, described by \cite{flaxmanThesis}, is to multiply a spatial kernel with a temporal one. Similar to our method for spatial mapping, we assume $M_t$ inducing points in time, $\bar{t}_1,\dots, \bar{t}_{M_t}$, which are evenly-spaced across a predefined range of time. The temporal kernels describe the correlation between the data and inducing points, from which we can generate a time series. The series can then be inputted into a LSTM network to learn further temporal dependencies. Our spatiotemporal kernel function is given by:
\begin{equation}
k\big((\mathbf{x},t),(\bar{\mathbf{x}},\bar{t})\big)=\underbrace{k_{s}(\mathbf{x},\bar{\mathbf{x}})}_{\text{Spatial}}\underbrace{k_{t}(t,\bar{t})}_{\text{Temporal}},
\end{equation}
where $\mathbf{x}$ and $t$ represent a coordinate in space and time, and $\bar{\mathbf{x}}$ and $\bar{t}$ are inducing points in space and time respectively.
We can calculate the temporal features over each data point as $\bm{\phi}_{t} \in \mathbb{R}^{M_t}$. We can then obtain the spatiotemporal feature of each data point as, $\bm{\phi}=\bm{\phi}_{s}\otimes\bm{\phi}_{t}$, where $\bm{\phi}\in\mathbb{R}^{M_s \times M_t}$. We also have different length scales for the spatial kernel function and the temporal kernel function, denoted by $\mathcal{\ell}_s$ and $\mathcal{\ell}_t$ respectively.

We input the spatiotemporal features into a LSTM network \citep{lstm}, then the outputted hidden representations of the LSTM network. Neural networks with LSTMs have proven to be effective for learning problems related to sequential data, in particular time series \citep{LSTMUse}. The key idea behind LSTMs is the use of memory cells which maintain their states over time and non-linear gating units that control information flowing in and out of these cells. The gating units typically include the input gate, output gate, forget gate, and memory cell. Each gate contains a non-linear transformation and a set of weights and biases. During training, the weights and biases of the gates are learnt from the data. Further details on LSTM networks can be found in \cite{lstmOdyssey}. In our method, the LSTM is able to learn a compact hidden representation of the spatiotemporal kernel features, which is a vector of the dimensionality of the specified output of the LSTM. A mixture density network is subsequently trained on the hidden representation of the LSTM.   

\subsection{Mixture Density Networks}

We expect the output of our model to be a probability distribution of movement directions, with each mode in this distribution indicating a probable trajectory direction. A probability distribution over the directions also sheds light on the uncertainty and the relative frequencies of occurrence of each associated direction.
We can achieve such outputs using mixture density networks (MDN) \citep{Bishop94mixturedensity}. A MDN allows our model to output parameters of a multi-modal mixture of distributions. Fig. \ref{distExample} shows the probability distributions of directions at three different points in space and time. Note that as the distribution is bounded between $-\pi$ and $\pi$ radians, while the integral of the probability density function is equal to 1, the probability density at a given point can have a value greater than 1.  


The probability density of the target direction is represented as a convex combination of $R$ individual probability density functions, which we will refer to as components. We denote the $r^{\text{th}}$ component as $p_{r}(\theta|\bm{\phi}(\mathbf{x},t))$, where $\bm{\phi}(\mathbf{x},t)$ denotes the high-dimensional spatiotemporal feature vector, and $\theta$ represents the target angular directions. We can then take a linear combination and write the likelihood as 
\begin{equation}
p(\theta|\bm{\phi}(\mathbf{x},t))=\sum ^{R}_{r=1}\alpha_{r}p_{r}(\theta|\bm{\phi}(\mathbf{x},t)).
\end{equation}

The parameters $\alpha_{r}$, known as mixing coefficients, are the weights of each component in the MDN. For a dataset with $N$ inputted data points, $\{(\mathbf{x}_n,t_n,\theta)\}^{N}_{n=1}$, which are assumed to be independent and identically distributed, we can write the joint likelihood as the product of the likelihood for each data point. The joint likelihood is \begin{align} p(\theta_{1},...,\theta_{N}|\{\bm{\phi}(\mathbf{x}_n,t_n)\}^{N}_{n=1})&=\prod^{N}_{n=1}p(\theta_{n}|\bm{\phi}(\mathbf{x}_n,t_n))
\\&= \prod^{N}_{n=1} \sum ^{R}_{r=1}\alpha_{r}p_{r}(\theta_n|\bm{\phi}(\mathbf{x}_n,t_n)).\end{align}

We assume that the probability density function of angular directions of trajectories at a particular coordinate can be approximated by a mixture of von Mises distributions \citep{directionStats}. For the $r^{\text{th}}$ component, the distribution has parameters $\mu_{r}\big(\bm{\phi}(\mathbf{x},t)\big)$ and $\kappa_{r}\big(\bm{\phi}(\mathbf{x},t)\big)$. These are known as the mean direction, and concentration parameters respectively. For brevity, we henceforth denote the mean direction and concentration of the $r^\text{th}$ component, with respect to the inputted high-dimensional features, as simply $\mu_r$ and $\kappa_{r}$. The probability of a single component distribution is then 
\begin{align}
p_{r}(\theta|\bm{\phi}(\mathbf{x},t))=\mathcal{V}\mathcal{M}(\theta|\mu_{r},\kappa_{r})=\frac{\exp \big( \kappa_{r} \cos\big(\theta - \mu_{r})\big)}{2\pi \mathcal{J}_{0}\kappa_{r}},
\end{align}
where $\mathcal{J}_0$ is the $0^{th}$ order modified Bessel function \citep{directionStats}.

The von Mises distribution can be viewed as an approximation of a wrapped Gaussian distribution, around the range $[-\pi,\pi)$. The mean direction of the von Mises is analogous to the mean of the Gaussian distribution, and the concentration can be viewed as the reciprocal of the variance in a Gaussian distribution.

\subsection{Learning Details}

The loss function of the MDN is the average negative log-likelihood (ANLL) over all the training examples, given by:
\begin{equation}\label{ANLL}
    \mathcal{L}=-\frac{1}{N}\sum^{N}_{n=1}\log\bigg(\sum^{R}_{r=1}  \alpha_{r}\mathcal{V}\mathcal{M}(\theta_n|\mu_{m},\kappa_{m})  \bigg).
\end{equation}

We can then use a neural network to optimise $\alpha_{r}$, $\mu_{r}$, and $\kappa_{r}$, by minimising the loss function. The mixing coefficients must satisfy $\sum^{R}_{r=1} \alpha_{r} = 1$, which can be enforced by applying a softmax activation function,
\begin{equation}
\alpha_{r}=\frac{\exp(z_{r}^{\alpha})}{\sum^{R}_{r=1}\exp(z_{r}^{\alpha})}, 
\end{equation}
where $z_{r}^{\alpha}$ denotes the network output corresponding to $\alpha_{r}$. To enforce $\kappa$ to be non-negative, an exponential activation function $\kappa_{r}=\exp(z_{r}^{\kappa})$ is applied to the network outputs.

In practice, the number of components, $R$, is determined prior to training. The larger the $R$ chosen, the more expressive the approximated distribution becomes. However, a larger $R$ also results in a larger number of parameters to learn, requiring more training data.

Due to the relatively high-dimensional outputs of MDNs, overfitting may easily occur when there are relatively few training examples. Overfitting in MDNs is discussed in detail in \cite{regMDN}. A straightforward method to combat overfitting in MDNs is to regularise the weights in the MDN, using, for instance, $L2$ regularisation. We can rewrite our regularised loss function as, 
\begin{equation}
    \mathcal{L}_{reg}=\mathcal{L}-\lambda_{\alpha}\Vert \mathbf{w}_{\alpha}\Vert_{2}^{2}-\lambda_{\mu}\Vert \mathbf{w}_{\mu}\Vert_{2}^{2}-\lambda_{\kappa}\Vert \mathbf{w}_{\kappa}\Vert_{2}^{2},
\end{equation}
where $\lambda_{\alpha}$, $\lambda_{\mu}$, $\lambda_{\kappa}\geq 0$ are regularisation coefficients, and $\mathbf{w}_{\alpha}$, $\mathbf{w}_{\mu}$ , $\mathbf{w}_{\kappa}$ are neural network weights of the mixing coefficients, mean directions, and concentrations respective.

The neural network architecture used in our experiments, with the specific number of units of each layer is shown in table~\ref{Architecture}. $M_s$ and $M_t$ denote the number of spatial and temporal inducing points respectively. This is dependent on the dataset used. $R$ denotes the number of von Mises distribution components in our output and was set to 5 for all the experiments.

\begin{table}[h]
\centering
\caption{The architecture of the spatiotemporal network used in our experiments. LSTM refers to a long short-term memory layer and FC refers to a fully connected layer. $M_s$ and $M_t$ are hyperparameters which denote the number of spatial and temporal inducing points respectively. $R$ denotes the number of von Mises distribution components in the output.} 
\begin{tabular}{|c|c|c|}
\hline
\multicolumn{3}{|c|}{Input (Size = $M_s \times M_t$)}                                         \\ \hline
\multicolumn{3}{|c|}{LSTM (Units = $40R$)}                                                    \\ \hline
FC (Units $= R$, $\alpha$) & FC(Units $= R$, $\mu$) & FC (Units $= R$, $\kappa$) \\ \hline
Softmax Activation           &                               & Exponential Activation        \\ \hline
\multicolumn{3}{|c|}{Concatenate}                                                            \\ \hline
\end{tabular}
\label{Architecture}
\end{table}


\section{Experiments and Discussions}\label{Experiments}
In this section, we investigate the behaviour of our proposed model on both simulated and real trajectory datasets. To evaluate the performance of our method, we run 10-fold cross-validation, splitting our data into training and test sets at a 90:10 ratio for each fold. We shall make use of Minimum Rotation Error and Average Likelihood metrics:

\textbf{Minimum Rotation Error (MRE)}: This metric gives the difference, in radians, between the ground truth direction and the mean of the nearest von Mises (VM) component. MRE gives a simple and relatable metric for real-world problems. However, it does not consider the mixing coefficients or the uncertainties of the VM distributions, nor does it consider components other than the nearest one to the ground truth. MRE over all of our test data is given by:
    \begin{equation}
        MRE=\frac{1}{N^\prime}\sum_{n=1}^{N^\prime}\min_{r}|\theta_{n}-\mu_{r}^{*}|,
    \end{equation}
    \noindent where $\theta_{n}$ denotes the label of test example $n$ in a test set of $N^\prime$ points in total, and $\alpha^{*}_r$, $\mu^{*}_r$, and $\kappa^{*}_r$ are predicted parameters of component $r$ in the mixture of $R$ distributions. The lower the value, the smaller the error between the ground truth and the mean of the nearest VM component.
    
\textbf{The Average Likelihood (AL)}: This metric is able to capture the uncertainty and multi-modality in angular distributions. This is equivalent to the \emph{Average Probability Density} metric in \cite{DirectionalGridMaps}. When measuring the performance of models, models with similar MRE can have significantly differing AL. Over all of our test data, AL is given by,
\begin{equation}
\text{AL}=\frac{1}{N^\prime}\sum^{N^\prime}_{n=1}\sum^{R}_{r=1}  \alpha_{r}^{*}\mathcal{V}\mathcal{M}({\theta}_n|\mu_{r}^{*},\kappa_{r}^{*})),
\end{equation}
The higher the AL value, the better the test data fit the approximated mixture. Note that the probability density at a point can be greater than 1. 

In all of our experiments, the inducing points are positioned in a grid manner in space and time. Neighbouring inducing points in each axis have a fixed distance between them, and all training and test data points lie within a regular grid of inducing points. In our experiments, the regularisation coefficients are chosen to be $\lambda_{\alpha}=0.001$, $\lambda_{\mu}=0$, $\lambda_{\kappa}=0.001$. We use a standard desktop computer with an Intel Core i7 microprocessor and 32 GB of RAM to conduct all experiments. The datasets used in our experiments are listed in Table~\ref{DatasetsTable}.

\begin{table}[t]
\vspace{10pt}
\caption{Description of datasets}
\label{DatasetsTable}
\centering
\begin{adjustbox}{width=0.95\textwidth,center} 
\begin{tabular}{l|c|l}
\toprule
Dataset                                                           & Temporal & Description of the dataset                                                                                         \\\hline
\begin{tabular}[c]{@{}l@{}}Pedestrian 1 \citep{DirectionalGridMaps} \end{tabular} & No           & Simulated unimodal.                                                                                                              \\
\begin{tabular}[c]{@{}l@{}}Pedestrian 2 \citep{DirectionalGridMaps} \end{tabular} & No           & Simulated multimodal. \\

\begin{tabular}[c]{@{}l@{}}Pedestrian 3\end{tabular} & Yes          & Simulated multimodal.                                                                                                                                                                       \\
Edinburgh \citep{Edinburgh}                                                       & Yes          & \begin{tabular}[c]{@{}l@{}}Real world pedestrian trajectories \\collected on 24 August.\end{tabular} \\
Lankershim \citep{Lankershim}                                                      & Yes          & \begin{tabular}[c]{@{}l@{}}A large real world traffic dataset \\on Lankershim Boulevard, USA.\\ \end{tabular}        \\
Peach Tree St \citep{PeachSt}                                                   & Yes          & \begin{tabular}[c]{@{}l@{}}A large real world traffic dataset \\ collected on Peach Tree Street. \end{tabular}\\ \bottomrule                
\end{tabular}
\end{adjustbox}
\end{table}

\begin{table}[h]
\centering
\caption{Results on pedestrian datasets 1 (P1) and 2 (P2), trained on the full dataset (100\%) and a random subset containing 5\% of the data. MM and UM stand for the unimodal and multimodal discrete mapping techniques in \cite{DirectionalGridMaps}.}
\begin{tabular}{l|ll|c|c|c}
\toprule
                                   Data   &   Metric & Method                  & 100\% & 5\%  & Change \\\hline
\multirow{6}{*}{P1}&
\multirow{3}{*}{AL}&Cont-           & 1.69$\pm$0.23     & 1.15$\pm$0.41   & -0.54               \\
                                    &  & MM-Dis & 1.54$\pm$0.04     & 0.60$\pm$0.14    & -0.94 \\
                                &   & UM-Dis   & 1.50$\pm$0.03     & 0.87$\pm$0.08   & -0.63              \\\cline{2-6}
             &\multirow{3}{*}{MRE}&Cont-& 0.08$\pm$0.01 &0.16$\pm$0.09& +0.08\\
             & & MM-Dis & 0.16$\pm$0.01 & 0.27$\pm$0.11&+0.11 \\
             & & UM-Dis &0.16$\pm$0.02&0.24$\pm$0.09&+0.08 \\\hline
\multirow{6}{*}{P2} & \multirow{3}{*}{AL}& Cont-           & 1.44$\pm$0.38     & 1.14$\pm$0.45   & -0.30   \\
                                    &  & MM-Dis & 1.17$\pm$0.03     & 0.57$\pm$0.19   & -0.60           \\
                                    &  & UM-Dis   & 0.86$\pm$0.02    & 0.46$\pm$0.02  & -0.40    \\ \cline{2-6}
                    &\multirow{3}{*}{MRE}& Cont-  &0.13$\pm$0.02&0.19$\pm$0.10 & +0.06\\
                    &  & MM-Dis  &0.29$\pm$0.02&0.74$\pm$0.18&+0.46  \\
                    &  & UM-Dis &0.49$\pm$0.06&0.87$\pm$0.08& +0.38 \\\hline

\end{tabular}
\label{SPcompare1}
\end{table}

For the Pedestrian 1 and 2 datasets, we conduct experiments using different numbers of inducing points. If we convert the temporal axis of all time-stamped datasets to use 0.1 hours as units, we position inducing points in a $1\times 1 \times1$ lattice when using Pedestrian 3, a $5\times 5 \times1$ when using Edinburgh, and a $20\times 20 \times 2$ for both Lankershim and Peach Tree St datasets.

\begin{figure}[h]

        \centering
        \includegraphics[width=0.95\textwidth]{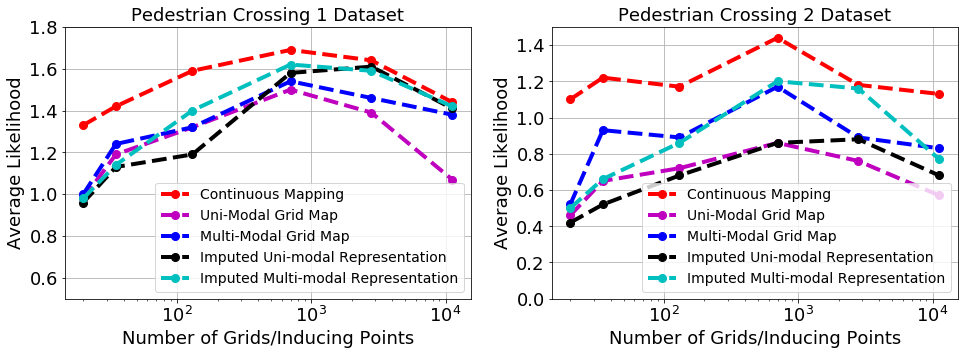}
        \includegraphics[width=0.95\textwidth]{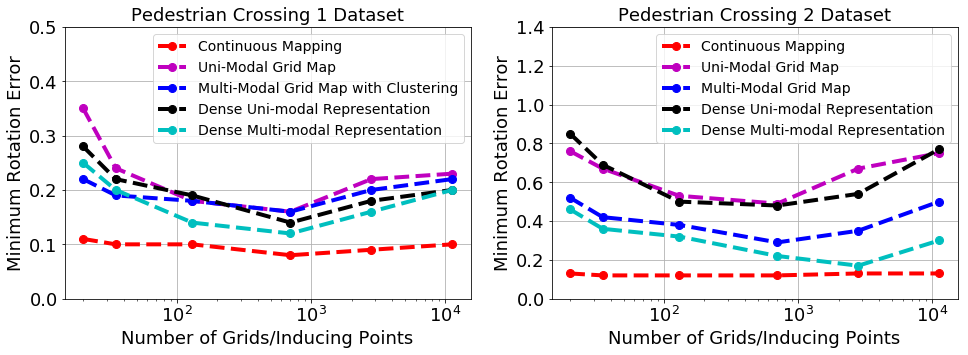}
        \caption{The performance of our continuous mapping method compared with the uni-modal grid-based method, and the multi-modal grid-based method with clustering, described in \cite{DirectionalGridMaps}, as well as dense variants similar to \cite{Flow}. We examine the effect of increasing the number of grid cells (for the discrete methods) and inducing points (for the continuous method).}
        \label{fig:Comparisons}
\end{figure}
\begin{figure}[h]
        \centering
        \includegraphics[width=0.95\textwidth]{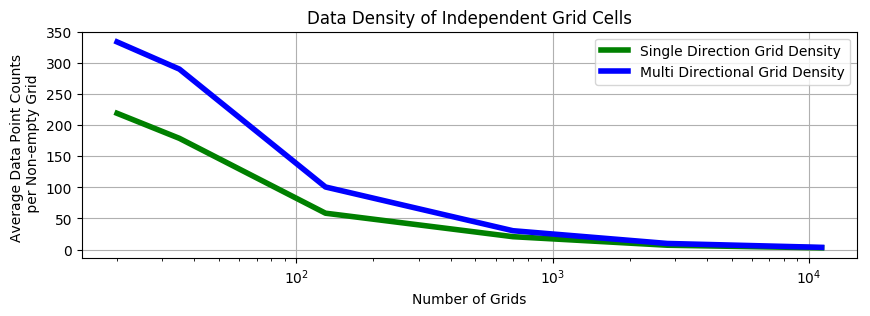}

        \caption{The average number of data points in every none empty independent grid cell, at different resolutions and P1, P2 datasets.}
        \label{fig:Density}
\end{figure}

\subsection{Continuous and Discrete Map Representations}
A major benefit of the proposed method is that the resulting maps are continuous over space and time. Removing the need to discretise the environment into independent cells is not only a more natural way to represent directions, it also facilitates building directional maps at arbitrary resolution. To investigate the benefits of continuous mapping, we conducted two experiments:
\begin{enumerate}
    \item We compare the performance of the discrete mapping methods, and dense variants of these methods, against our method at various resolutions (Fig.~\ref{fig:Comparisons}); 
    \item the performance of the discrete mapping methods against our method when we only have 5\% of the original dataset used (Table~\ref{SPcompare1}).
\end{enumerate}

We compare the performance of our continuous mapping method with the two discrete directional grid map methods presented in \cite{DirectionalGridMaps}. One of the methods presented fits a von Mises distribution to each cell individually, the other fits a mixture of von Mises distributions after finding reasonable initialisations via clustering. 

Similar to our continuous method, dense representations similar to the method described in \cite{Flow} also allow for arbitrary resolution of the map, by providing a smooth transition from one cell to the next. We make comparisons against dense variants of grid-based methods, similar to the method presented in \cite{Flow}, but using mixtures of VM distributions instead of the Wrapped Gaussian Mixture Model \citep{Flow}. Rather than segmenting the environment into independent cells, we want to lay overlapping circular cells over the environment. The same data point may lie within the radius of several cells, and be used in the training of several VM mixtures. When we query a coordinate, we impute the distribution by taking the linear combination of the VM mixtures within a fixed radius, with the weights determined by computing the RBF function applied on distances between the queried point and the centre of each relevant cell. This setup is similar to that in \cite{Flow}. In our experiments, the radius is the distance between the centre of one cell to neighbouring cells. 



The objective of the first experiment is to understand how the proposed continuous mapping method compares to discrete mapping methods. The results of the first experiment are shown in Fig. \ref{fig:Comparisons}. The number of grid cells / inducing points used is 20, 35, 130, 700, 2800, and 11200. The data density of each independent non-empty cell at each resolution is shown in Fig. \ref{fig:Density}. For the continuous method, we need to adjust the length scales, as good values for the length scale is dependent on the distance between inducing points. For the continuous method, we use a length scale, $\ell_s$, of 3.0, 2.2, 1.0, 0.6, 0.4, and 0.3 respectively for both datasets with the aforementioned resolutions. We see that our continuous representation of directional maps outperforms previously proposed grid-based methods, when there is an equal number of inducing points and grid cells. All the methods perform well at 700 grid cells/inducing points. The benefit of continuous mapping is prominent when there is a small number of grid cells or inducing points representing the map. When there are only a few cells, the cell resolution is relatively low, and there may be directional data in several directions. However, the grid-based methods give each training example within the same cell equal weighting, and do not take into account the position of the data points within the cell. For example, a training data point with coordinates at the edge of cell holds the same weight as a data point positioned at the centre of the cell. The dense representations attempts to achieve outputs which are continuous, and provide slight improvements in performance relative to discretised grid cells. It factors in the predictions from several cells, and the distances between the training data points and the cell centres. However, during training it still suffers from not weighing in these distances. The continuous representation considers the distance between each inducing point and the training coordinate, whereas both simple grid map and imputed dense models give equal weighting to data points in a certain range. The drawback of this is apparent -- a training point very close to the cell centre is considered to have the same weighting as a point on the fringes of the cell. The continuous representation takes these distances into account via the RBF distances between a data point and predefined inducing points. As the number of grid cells increase over 700, the number of data points in each grid cell decreases such that there are too few data points in each cell. Likewise with the continuous method, when the number of inducing points becomes too large, the number of parameters in the neural network increase to a level that the performance decreases.

The second experiment evaluates the performance of discrete and continuous methods when there are few data points. The results of the second experiment which shows the sensitivity of each method to the amount of data are tabulated in Table~\ref{SPcompare1}. We trained each model at the resolution of, 700 inducing points/grid cells. We see that with 95\% of the data points randomly removed and ten-fold cross-validation used to obtain the performance, the performance for all three methods deteriorate as expected. However, we see that the change of the continuous method is lower compared to the discretised methods. This is due to the continuous method considering all data points to have a degree of influence in the surrounding region. This makes more efficient use of training data, as a point that is considered in another cell in the discrete method would be considered during the training of the continuous model. This also results in the continuous method being more robust to outliers. In the discrete methods, when there are very few data points in a single grid cell, the effect of an outlier is significant, but the continuous model would be informed by data points that the discrete model treats as independent, and thus would have information from more data points.  

\begin{figure}[H]
\centering
\begin{subfigure}{0.48\textwidth}
    \centering
    \includegraphics[width=.9\textwidth]{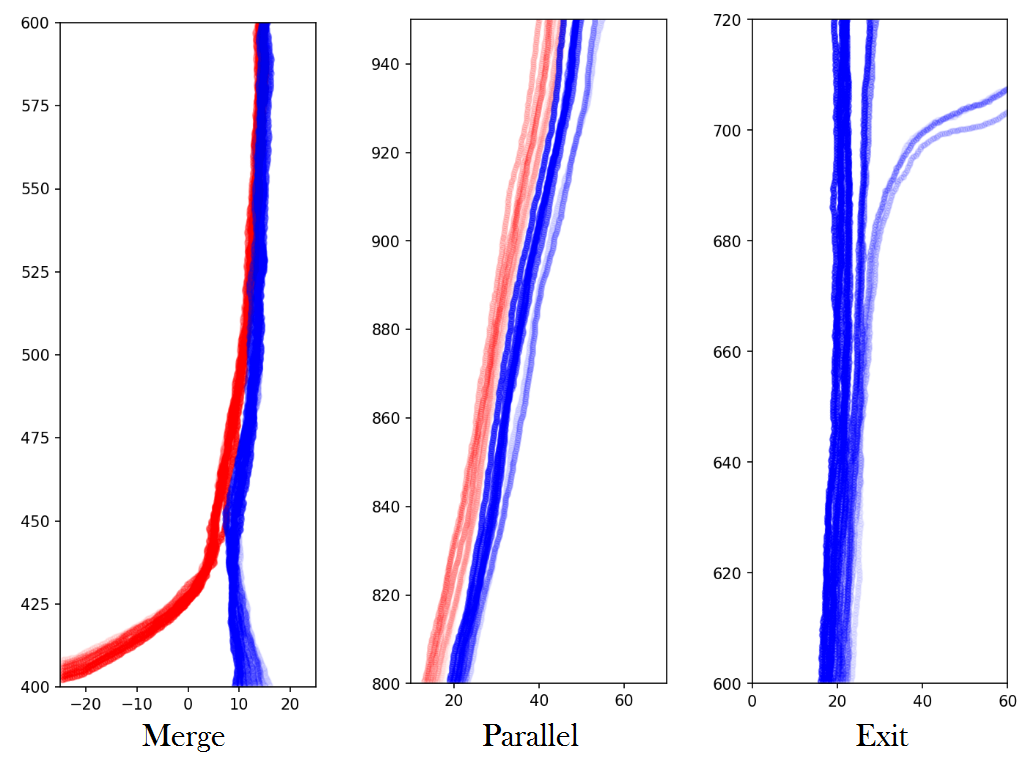}
    \caption{Generated trajectories from a model trained on traffic over sections of highway in Lankershim are shown. The vehicles move from bottom to top, with 16 such trajectories in each plot. (Left) vehicles enter the boulevard in two different directions and merge; (Centre) two separate lanes (red and blue) of the boulevard; (Right) trajectories diverge.}
    \label{TrajectoriesPlot}
\end{subfigure}%
\hspace{1em}
\begin{subfigure}{0.48\textwidth}
    \centering
    
     \includegraphics[width=0.98\textwidth]{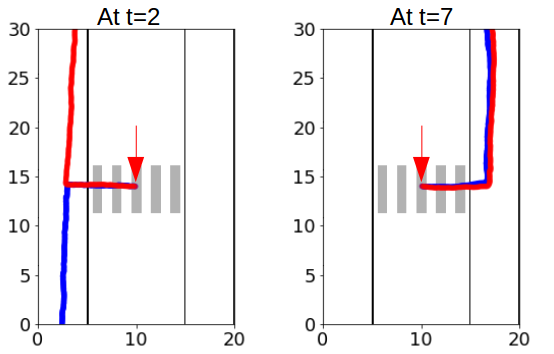}
    \caption{Predicted trajectories at times $t=2$ (left) and $t=7$ (right). At each of the two times, we generate 2 trajectories which originate at the middle of the pedestrian crossing, indicated by the red arrows. Trajectories both start at the same location at the centre of the crossing, but at different times, they take different directions.}
    \label{simfig}
\end{subfigure}
\end{figure}

\subsection{Generating Trajectories in Space}
\label{sec:gentraj}
This subsection briefly outlines a method to visualise directional predictions from our model, by generating trajectories. As the time between steps is negligible relative to the time recorded in the data, we disregard the temporal variations to generate a long-term directional map which represents the ``typical traffic patterns'' at a given position in the environment. We can generate trajectories that originate from a particular coordinate anywhere in space by recursively sampling the distribution of directions and taking small steps in the sampled direction. Visualisation of trajectories, especially on traffic datasets, allows us to qualitatively evaluate the predictions of our model.  

Given a particular starting coordinate $(x_0,y_0)$, we query our continuous map to obtain $p(\theta|x_0,y_0)$. This answers the question of in which direction should an agent take next if the current position is $(x_0,y_0)$. We sample this distribution to obtain a direction to move in, $\theta_{0}$. We can then update the coordinates by taking an increment step, $d$, in the direction of $\theta_0$. We can take an Euler integration approach, and recursively predict the current position by sampling directions and computing $x_{n+1}=x_{n}+d\cos(\theta_{n})$ and $y_{n+1}=y_{n}+{d}\sin(\theta_{n})$ for $N$ time steps. This results in a trajectory with coordinates $({x_0},{y_0}), \ldots, (x_N,y_N)$. We can generate different trajectories from $(x_0,y_0)$ by taking various samples from $p(\theta|x_n,y_n)$. Plots of such generated trajectories with a model trained on the Lankershim dataset are shown in Fig. \ref{TrajectoriesPlot}. It can be seen that the trajectories consistently follow sensible paths, and the model is able to generate realistic trajectories that enter and exit the boulevard. Such predictions can plausibly be used for risk-aware decision-making in urban environment.

\begin{figure}[h]
\centering
    \includegraphics[width=0.52\textwidth]{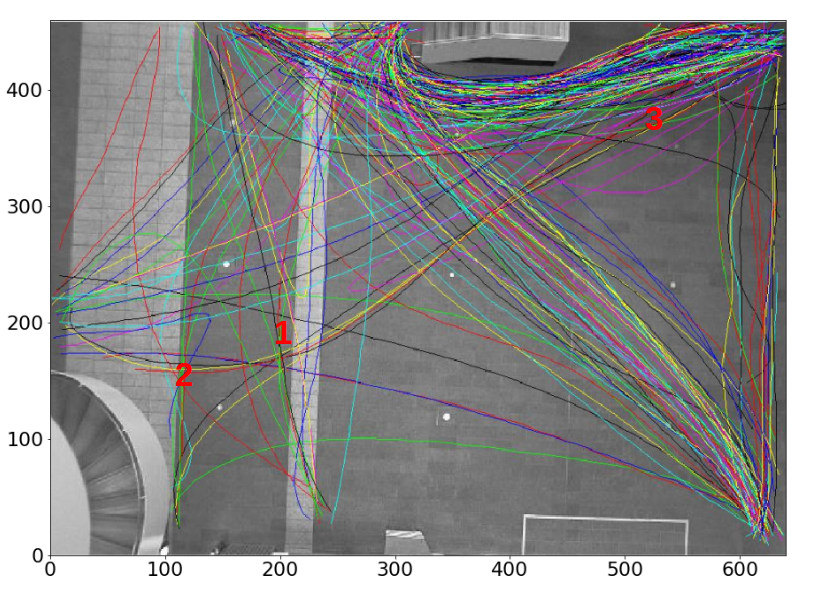}
     \includegraphics[width=0.47\textwidth]{figures/chap3/temporal_change4.png}\caption{The left figure shows trajectories people take in the atrium of the Edinburgh university \citep{Edinburgh}. We attempt to model how the directional distribution change over time in three different locations. The predicted probability distributions over four time-steps at the three locations are shown on the right.}\label{overTime}
\end{figure}

\subsection{The Spatiotemporal Representation}

Trajectories generated in Section~\ref{sec:gentraj} represent the typical paths dynamic agents such as vehicles and pedestrians take. Previous methods, such as those in \cite{OCallaghan2011LearningNM} and \cite{DirectionalGridMaps}, assume that the typical paths of objects do not change over time. However, in real-world environments, the general directions of trajectories change not only with space but also with time. An obvious example would be the different trajectory directions of people entering and exiting a building before and after an event. 

The kernel function used for temporal features affects how smooth the interpolated functions between inducing points are. After experimenting with common stationary kernels, the RBF kernel was found to have good performance for generating temporal features. Fig. \ref{simfig} shows trajectories generated at different points in time using a model trained on the pedestrian crossing 3 dataset, at the same starting coordinates $(10,15)$. We see that at different times, generated trajectories will either end at the left footpath or the right. If the temporal trends are not considered, the distribution of directions on the pedestrian crossing would be largely bi-modal, with modes representing the left-to-right and right-to-left movements simultaneously. Such a representation is incorrect. In reality, as captured by the proposed spatiotemporal model, trajectories generated starting from the centre of the pedestrian crossing could end on the left side and right side of the road, depending on the time (Fig.~\ref{simfig}). Fig. \ref{overTime} also illustrates directional probability density function change in time, predicted by a model trained on Aug 24 of the Edinburgh dataset, at three fixed points in space at four different time-steps. 

\begin{table}[h]
\vspace{10pt}
\centering
\caption{The AL values on several datasets, using both the spatial and spatiotemporal methods.}
\begin{tabular}{l|c|c}
\toprule
    Dataset   & Spatial   & Spatiotemporal \\ \hline
Pedestrian Crossing 3 & 1.09$\pm$0.06 & 2.20$\pm$0.23      \\ 
Edenburgh     & 0.38$\pm$0.02 & 0.38$\pm$0.02       \\ 
Peach Tree St & 1.34$\pm$0.04 & 1.40$\pm$0.09       \\ 
Lankershim    & 1.70$\pm$0.08 & 1.75$\pm$0.15       \\
\bottomrule
\end{tabular}
\label{SPcompare}
\end{table}

To obtain a quantitative evaluation, we compare the performance of our continuous spatiotemporal model with a solely spatial continuous model on a range of datasets, and the results are tabulated in Table \ref{SPcompare}. The spatial continuous model directly trains a MDN on the kernel features over space. We see that our spatiotemporal models give respectable improvements in performance on most of the tested datasets, and perform at least as well as the spatial models on all of the tested datasets, despite having a lower density of training data points due to the additional time axis. The greatest improvement in performance was observed on the pedestrian 3 dataset, as it contains trajectories of completely different directions at different times (illustrated in the plots in Fig. \ref{simfig}). We only use the AL metric, as MRE only considers the VM component with $\mu$ closest to the ground truth direction, even if it has a low $\alpha$. Therefore, it does not penalise inaccurate parameter estimates in VM components other than the closest component to ground truth, nor does it consider the $\alpha$ values of each component. MRE is incapable of capturing nuanced changes in the concentration and mixing coefficients, which usually occur in the time domain. For example, the mean directions of pedestrian direction distribution on a footpath would largely remain unchanged throughout the day, but the mixing coefficients would vary significantly over time. MRE would not be able to accurately reflect the performance in this scenario.

\section{Summary}
We present a novel spatiotemporal model to learn the distribution of directions of moving objects. The proposed method is continuous in both space and time, and can be queried at arbitrary resolution. We make use of high-dimensional kernel features, LSTM networks, and a MDN of von Mises distributions. Our method is capable of capturing subtle temporal changes in the multi-modal directional distribution that cannot be adequately modelled with existing techniques. Although the datasets we used are limited to urban environments, the proposed continuous spatiotemporal model is exceedingly generic, and can be applied to other domains that involve mapping coordinates in space and time to a potentially multi-modal distribution. Additional improvements on the proposed method can revolve around conditioning the probability distribution of movement directions on the history of the object for path planning in dynamic environments.

We shall next move to \cref{part2}, and explore learning for anticipatory navigation. Up next in \cref{chap5}, we develop a framework which produces probabilistic predictions of how neighbouring agents move and accounts for these predictions when controlling the ego-robot.

\pagebreak

\part{Learning for Anticipatory Navigation}\label{part2}
\chapter{Stochastic Process Anticipatory Navigation}\label{chap5}\blfootnote{This chapter has been published in ICRA as \cite{SPAN_nav}.}
\usetikzlibrary{arrows}
\usetikzlibrary{calc}
\tikzstyle{decision} = [diamond, draw, fill=blue!20, 
    text width=6em, text badly centered, node distance=3cm, inner sep=0pt]
\tikzstyle{block} = [rectangle, draw, fill=blue!20, 
    text width=6em, text centered, rounded corners, minimum height=3.2em]
\tikzstyle{line} = [draw, -latex']
\tikzstyle{cloud} = [draw, ellipse,fill=red!20, node distance=3cm,
    minimum height=2em]
\newcommand{\coorTS}{\mathbf{\hat{o}}}  

\section{Introduction}
Robots coexisting with humans often need to drive towards a goal while actively avoiding collision with pedestrians and obstacles present in the environment. To safely manoeuvre in crowded environments, humans have been found to anticipate the movement patterns in the environment and adjust accordingly \citep{anticipate}. 
Likewise, we aim to endow robots with the ability to adopt a probabilistic view of their surroundings and consider likely future trajectories of pedestrians. Static obstacles in the environment are encoded in an occupancy map, and avoided during local navigation.
We present the Stochastic Process Anticipatory Navigation (SPAN) framework, which generates local motion trajectories that takes into account predictions of the anticipated movements of dynamics objects.

\begin{figure}[t]
    \centering
    \includegraphics[width=0.8\textwidth]{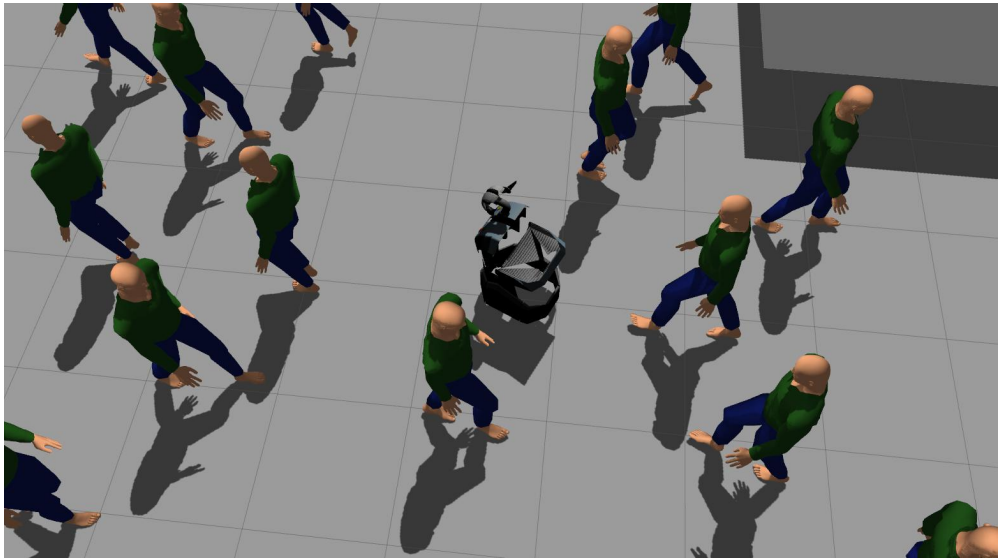}
    \caption{A non-holonomic robot drives towards a goal while interacting with pedestrians and static obstacles, and actively adapts to probabilistic predictions of surrounding movements.}\label{fig1}
\end{figure}

SPAN leverages a probabilistic predictive model to predict the movements of pedestrians. Motion prediction for dynamic objects, in particular for human pedestrians, is an open problem and has typically been studied in isolation without direct consideration of its use in a broader planning framework. Many recent methods use learning-based approaches \citep{Alahi2016SocialLH,KTM,Gupta2018SocialGS} to generalise motion patterns from collected data. To adequately capture the uncertainty of these predictions, representations of future motion patterns of pedestrians need to be probabilistic. Furthermore, collision-checking with future pedestrian positions needs to be done at a relatively high time resolution. To address these requirements, we represent uncertain future motions as continuous-time stochastic processes. Our representation captures prediction uncertainty and can be evaluated at an arbitrary time-resolution. We train a neural network to predict parameters of the stochastic process, conditional on the observed motion of pedestrians in the environment. Navigating through an environment requires online generation of collision-free and kinematically-feasible trajectories. We combine learned probabilistic anticipations of future motion and an occupancy map into a non-holonomic control problem formulation. To account for environment interactions a time-to-collision term \citep{Hayward1972NEARMISSDT} is integrated into the control problem. The formulated problem is non-smooth, and a derivative-free constrained optimiser is used to efficiently solve the problem to obtain control actions. A receding horizon approach is taken and the problem is re-optimised at a fixed frequency to continuously adapt to pedestrian prediction updates.


SPAN is novel in jointly integrating data-driven probabilistic predictions of pedestrian motions and static obstacles in a control problem formulation. The technical contributions of this chapter include: (1) A continuous stochastic process representation of pedestrian futures, that is compatible with neural networks, and allows for fast chance-constrained collision-checking, at flexible resolutions; 
(2) formulation of a control problem that utilises predicted stochastic processes as anticipated future pedestrian positions, to navigate through crowds. We demonstrate that the optimisation problem can be solved efficiently to allow the robot to navigate through crowded environments. SPAN also opens opportunities for advances in neural network-based motion prediction methods to be utilised for anticipatory navigation. The stochastic process representation of pedestrian motion is compatible with many other neural network models as the output component of the network.    


\section{Related Work}
SPAN actively predicts the motion patterns of pedestrians in the surrounding area and avoids collision while navigating towards a goal. Simple solutions to motion prediction include constant velocity or acceleration models \citep{humanSurvey}. Recent methods of motion prediction utilise machine learning to generalise motion patterns from observed data \citep{sptemp,Alahi2016SocialLH,Gupta2018SocialGS,KTM}. These methods have a particular focus on capturing the uncertainty in the predicted motion, as well as on conditioning on a variety of environmental factors, such as the position of neighbouring agents. Generally, past literature examines the motion prediction in isolation, without addressing issues arising from integrating the predictions into robot navigation, where the anticipation of pedestrian futures is valuable. 

We seek to generate collision-free motion, accounting for predicted motion patterns. Planning collision-free paths is well-studied, with methods at differing levels of locality. At one end of the spectrum, probabilistically-complete planning methods \citep{Lavalle98rapidly-exploringrandom} explore the entire search-space to find a globally optimal solution. At the other end, local methods \citep{potential,rmpflow,CHOMP} aim to perturb collision-prone motion away from obstacles, trading-off global optimality guarantees for faster run-time. Our work focuses on the local aspect, where we aim to quickly obtain a sequence of control commands to navigate through dynamic environments. Various non-linear controllers \citep{britolmpcc,NMPC,MPPI, JointM} also generate control sequences to avoid collisions. 

Local collision-avoidance methods for navigation for dynamic environments have been explored. Methods such as velocity obstacles (VO) \citep{Fiorini1998MotionPI} and optimal reciprocal collision avoidance (ORCA) \citep{Berg2009ReciprocalNC} find a set of feasible velocities for the robot, considering the current velocities of other agents. Partially observable Markov decision process (POMDP) solvers have been used in \citep{porca,hypdespot} to control linear acceleration for autonomous vehicles, with steering controls obtained from a global planner. Methods based on OCRA that indirectly solve for control sequences, such as \citep{GRCA}, are known to be overly conservative. Less conservative methods have also been developed to incorporate human intent to identify unlikely human action \citep{intent_motion_pred}. An approach was proposed in \cite{NHTTC}, where a subgradient-based method utilises a time-to-collision value \citep{Hayward1972NEARMISSDT}, under constant velocity assumptions. Similarly, SPAN also optimises against time-to-collision, and uses derivative-free constrained optimisation solvers. We extend the usage of time-to-collision to probabilistic predictions of pedestrian positions, provided by learning models, as well as integrating static obstacles captured by occupancy maps.   


\begin{figure}[t]
\centering
\begin{adjustbox}{width=0.9\textwidth,center} 
\begin{tikzpicture}[node distance = 1.2cm]

    \node [block] (Model){Learned Model};
    \node [block, right of=Model, node distance=6cm] (Pred) {Stochastic process pedestrian futures};
    \node [block, below of=Model, node distance=4cm] (Map) {Occupancy map};
    \node [block, below of=Pred, node distance=4cm] (Control) {Control problem};
    \node [block, right of=Control, node distance=6cm] (solver) {Derivative-free solver};
    \node [block, right of=solver, node distance=6cm] (controls) {Next controls};
    \node [block, above of=controls, node distance=4cm] (States) {Robot states};
    \draw[black,thick,dotted] ($(Pred.north west)+(-1.5,0.2)$) node [text width=1cm,above]{\textbf{Loop}} rectangle ($(controls.south east)+(0.5,-0.2)$);

    \path [line] (Model) -- node [text width=1cm,midway,above]{Predict}(Pred);
    \path [line] (Map) -- (Control);
    \path [line] (Pred) -- (Control);
    \path [line] (Control) -- (solver);
    \path [line] (States) -- (Control);
    \path [line] (solver) -- node [text width=1cm,midway,above]{Solve for}(controls);
    \path [line] (controls) -- node [text width=3cm,midway]{Update states}(States);

\end{tikzpicture}
\end{adjustbox}
\caption{An overview of SPAN.}\label{Overview_ch5}
\end{figure}
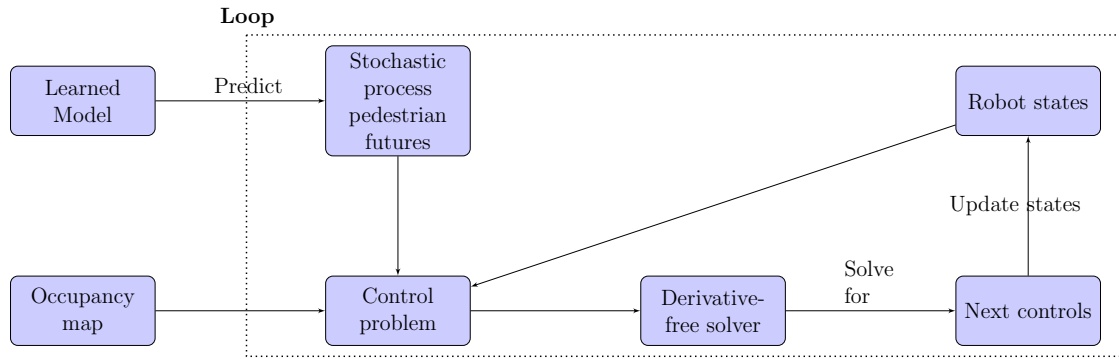

\section{Problem Overview}

A general overview of SPAN is provided in \cref{Overview_ch5}. (1) First, we train a model offline to predict probabilistic future pedestrian positions, conditioning on previous positions; (2) The predictive model and occupancy map are incorporated in an anticipatory control problem; (3) The control problem is solved online, in a receding horizon manner, with derivative-free solvers to obtain controls for the next states. This process is looped, with the learned model is continuously queried, and the problem continuously formulated and solved to obtain the next states.


\newcommand{\xCoor}{\bm{\hat{x}}}
\newcommand{\xState}{\mathbf{x}}

\subsection{Problem Formulation}
This chapter addresses the problem of controlling a robot, to navigate towards a goal $\bm{g}$, in an environment with moving pedestrians and static obstacles.
The state of the robot at a given time $t$ is given by $\xState(t)=\begin{smallmatrix}[\,x(t) & y(t) & \theta(t)\,]\end{smallmatrix}^{\top}$, denoting the robot's two-dimensional spatial position and an additional angular orientation. We focus our investigation on wheeled robots, with the robot following velocity-controlled non-holonomic unicycle dynamics, where the system dynamics are given by:  
\begin{align}
    \dot{\xState}=f(\xState,\mathbf{u})=
    [v\cos(\theta),v\sin(\theta),\omega]^{\top},&& \mathbf{u}=[v,\omega]^{\top},\label{EqnMotion}
\end{align}
where $\dot{\xState}$ denotes the state derivatives. The robot is controlled via its linear velocity $v$ and angular velocity $\omega$. We denote the controls as $\mathbf{u}$ and the non-linear dynamics as $f(\xState,\mathbf{u})$. 

%
Information about \emph{static obstacles} in the environment is provided by an occupancy map, denoted as $f_{m}(\cdot)$, which represents the probability of being occupied $p(\mathrm{Occupied}|\xCoor)\in[0,1]$ at a coordinate $\xCoor\in \mathbb{R}^{2}$. The set of positions of $N$ pedestrian obstacles, at a given time $t$ from the current time, is denoted by the set $\mathcal{O}=\{\bm{o}^{1}(t),\ldots,\bm{o}^{N}(t)\}$. 
We are assumed to have movement data, allowing for the training of a probabilistic model to predict future positions of pedestrians. 

\section{Probabilistic and Continuous Prediction of Pedestrian Futures}
In this section, we introduce a continuous \emph{stochastic process} representation of future pedestrian motion, and outline how the representation integrates into a neural network learning model. 
\subsection{Pedestrian Futures as Stochastic Processes}\label{representationSubsection}
Representations of future pedestrians positions need to capture uncertainty. Additionally, the time-resolution of the pedestrian position forecasts needs to synchronise with the frequency of collision checks. These factors motivate the use of a continuous-time stochastic process (SP) for future pedestrian positions. A SP can be thought of as a distribution over functions. Furthermore, the continuous nature allows querying of future pedestrian position at an arbitrary time, rather than at fixed resolution, without additional on-line interpolation. 

We start by considering a deterministic pedestrian motion trajectory, before extending to a \emph{distribution over motion trajectories}. A trajectory is modelled by a continuous-time function, given as the weighted sum of $m$ basis functions. We write the $n^\text{th}$ pedestrian $\bm{o}^n\in\mathcal{O}$ coordinate at $t$ as:
\begin{align}
\mathbf{o}^n(t)={\mathbf{W}^n}^{\top}\bm{\Phi}(t), && \bm{\Phi}(t)=[\bm{\phi}(t,\mathbf{t}'),\bm{\phi}(t,\mathbf{t}')],\label{GPdef_eq_ch5}
\end{align}
where ${\mathbf{W}^n}^{\top} \in \mathbb{R}^{m \times 2}$ is a matrix of weights that defines the trajectory. $\bm{\Phi}(t)$ contains two vectors of basis function evaluations, $\bm{\phi}(t,\mathbf{t}')$, one for each coordinate axis. $\bm{\phi}(t,\mathbf{t}')=[\phi(t,t'_{1}),\ldots,\phi(t,t'_{m})]^{\top}$ is a vector of basis function evaluations at $m$ evenly distanced time points, $\mathbf{t}'\in\mathbb{R}^{m}$. We use the squared exponential basis function given by $\phi(t,t')= \exp(-\gamma\|t-t'\|^{2})$, where $\gamma$ is a length-scale hyper-parameter.  

To extend our representation from a single trajectory to a distribution over future motion trajectories, we assume that the weight matrix is not deterministic, but random with a matrix normal ($MN$) distribution: 
\begin{align}
\mathbf{W}^{n}\sim MN(\mathbf{M}^{n},\mathbf{U}^{n},\mathbf{V}^{n}).
\end{align}
The matrix normal distribution is a generalisation of the normal distribution to matrix-valued random variables \citep{MLEMatN}, parameters mean matrix $\mathbf{M}^{n}\in \mathbb{R}^{m\times 2}$ and scale matrices $\mathbf{U}^{n}\in \mathbb{R}^{m\times m}$ and $\mathbf{V}\in \mathbb{R}^{2\times 2}$. $\mathbf{M}^{n}$ is analogous to the mean of a normal distribution, and $\mathbf{U}^{n}, \mathbf{V}^{n}$ capture the covariance. A log-likelihood of the matrix normal distribution is given later in \cref{NLL}. Predicting a SP requires predicting $\mathbf{M}^{n},\mathbf{U}^{n},\mathbf{V}^{n}$.

\begin{figure}[t]
    \centering
    \includegraphics[width=0.7\textwidth]{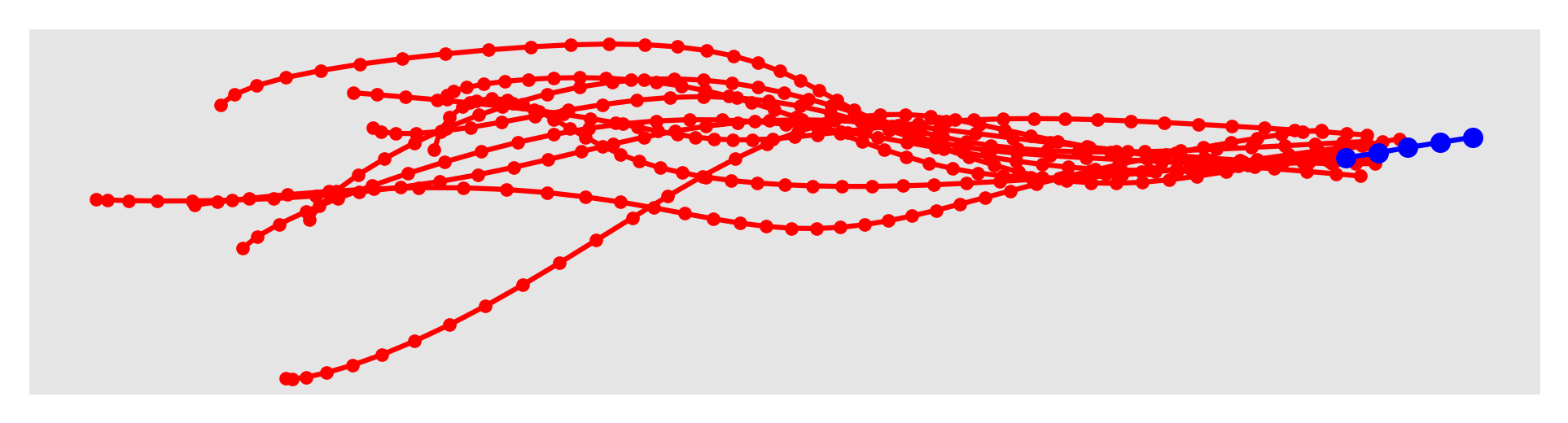}
    \caption{We can visualise the predicted stochastic process by sampling trajectories (red) from a predicted distribution, conditional on previously observed pedestrian coordinates (blue). }
    \label{pred_fig}
\end{figure}
\subsection{Learning Pedestrian Motion Stochastic Processes}\label{learn}
This subsection outlines how to train a model to predict SP representations of pedestrian motion. In the following subsection, we denote the weight matrix as $\mathbf{W}$ and parameters of the matrix normal distribution as $\mathbf{M},\mathbf{U},\mathbf{V}$, without referring to an obstacle index of a specific pedestrian. We aim to condition on a short sequence of recently observed pedestrian positions, up to the present $\{\coorTS_{1},\ldots,\coorTS_{p}\}$, and predict $\mathbf{M},\mathbf{U},\mathbf{V}$, which define a SP of the motion thereafter. A predictive model is trained offline with collected data, and then queried online. 

\subsubsection{From timestamped data to continuous functions}
Pedestrian motion data is typically in the form of a discrete sequence of $n$ timestamped coordinates $\{t_{i},\coorTS_{i}\}_{i=1}^{n}$, where $t_{i}$ gives time and $\coorTS_{i}\in\mathbb{R}^{2}$ gives coordinates. Timestamped coordinates can be converted into a continuous function by finding a deterministic $\mathbf{W}$. For a pre-defined basis function $\bm{\Phi}(\cdot)$, we solve the ridge regression problem:  
\begin{equation}
{\min_{\mathbf{W}}\Big\{\sum_{i=1}^{n}(\coorTS_{i}-\underbrace{\mathbf{W}^{\top}\bm{\Phi}(t_{i})}_{\mathbf{o}(t_i)})^2+\lambda\|\mathrm{vec}(\mathbf{W})\|^{2}\Big\}}\label{lsuqares}
\end{equation}
where $\lambda$ is a regularisation hyper-parameter, and $\mathrm{vec}(\cdot)$ is the vectorise operator. By converting timestamped data to continuous functions, we can obtain a dataset, $\mathcal{D}$, with $N_{D}$ pairs, $\mathcal{D} = \{(\{\coorTS_{1} , \ldots,\coorTS_{p}\}_{d}, \mathbf{W}_{d})\}_{d=1}^{N_{D}}$, where $\{\coorTS_{1},\ldots,\coorTS_{p}\}_{d}$ are previously observed coordinates, and $\mathbf{W}_{d}$ defines the continuous trajectory from $\coorTS_{p}$ onwards. $\mathbf{W}_{d}$ is obtained from timestamped coordinate data via \cref{lsuqares}. This dataset is used to train the predictive model.
\subsubsection{Training a Neural Network}
We wish to learn a mapping, $F$, from a sequence of observed pedestrian coordinates, $\{\coorTS_{1},\ldots,\coorTS_{p}\}$, to the parameters over distribution $\mathbf{W}$, that defines future motion beyond the sequence. We can use a neural network as,
\begin{equation}
    \overbrace{(\mathbf{M},\mathbf{U},\mathbf{V})}^{\mathclap{\text{defines SP over future motion}}}=\underbrace{F}_{\mathclap{\text{neural network function approximator}}}(\overbrace{\{\coorTS_{1},\ldots,\coorTS_{p}\}}^{\mathclap{\text{past coordinates}}})
\end{equation}
and train via the negative log-likelihood \citep{MLEMatN} of the matrix normal distribution, over $N_{D}$ training samples of dataset $\mathcal{D}$. The loss function of the network is given as:
\begin{align}
    \mathcal{L}
    =\!-\!\sum_{d=1}^{N_{D}}\!\log\frac{\exp\{-\frac{1}{2}\mathrm{tr}[{\mathbf{V}}^{-1}\!(\mathbf{W}_d-\mathbf{M})^{\top}\!\mathbf{U}(\mathbf{W}_d-\mathbf{M})]\}}{(2\pi)^{m}|\mathbf{V}||\mathbf{U}|^{\frac{m}{2}}},\label{NLL}
\end{align}
where $\mathrm{tr}(\cdot)$ is the trace operator. In our experiments, a simple 3-layer fully-connected neural network with width 100 units, and an output layer giving vectorised $\mathbf{M},\mathbf{U},\mathbf{V}$ is sufficiently flexible to learn $F$. In this work, we focus on outlining a representation of motion able to be learned and used for collision avoidance. More complex network architectures can be used to learn $F$, including recurrent network-based models \citep{Alahi2016SocialLH}.

\subsubsection{Querying the Predictive Model}
After training, the neural network outputs $\mathbf{M},\mathbf{U},\mathbf{V}$, conditional on observed sequences. $\mathbf{M},\mathbf{U},\mathbf{V}$ define random weight matrix $\mathbf{W}$, and the SP model via \cref{GPdef_eq_ch5}. Individual trajectories can be realised from the SP by sampling $\mathbf{W}$. \Cref{pred_fig} shows sampled trajectories from a predicted SP.     

\section{Anticipatory Navigation with Probabilistic Predictions}
\subsection{Time-to-Collision Cost for Control}
This subsection introduces a control formulation which makes use of probabilistic anticipations of future pedestrian positions. Human interactions have been found to be governed by an inverse power-law relation with the time-to-collision \citep{powerlaw}, and can be integrated in robot control \citep{NHTTC}. To this end, we formulate an optimal control problem, with the inverse time-to-collision in the control cost:
\begin{subequations}
\begin{align}
    \min_{\mathbf{u}}\quad&\Big\{{\|\xCoor(T)-\bm{g}\|_{2}}+\frac{\kappa}{\tau(\xState_{0},\mathbf{u},\bm{o},f_{m})}\Big\}\label{costOpt}\\
    \textrm{s.t.}\quad&\dot{\xState}(t)=f(\xState(t),\mathbf{u}), \quad t\in[0,T], \label{costOpta}\\
    & \xState(0)=\xState_{0}, \label{costOptb}\\
    & \mathbf{u}_{lower}\leq\mathbf{u}\leq \mathbf{u}_{upper}\label{costOptc},
\end{align}
\end{subequations}
where $T$ is the time horizon; 
$\xCoor$, $\bm{g}$, $\xState$, $\mathbf{u}$ are the robot coordinates, goal coordinates, states, and target controls; $\tau$ is the time-to-collision value; $\kappa$ is a weight hyper-parameter controlling the collision-averseness; $\bm{o}$ denotes the position of pedestrians, $f_{m}$ denotes an occupancy map; $\mathbf{u}_{upper}, \mathbf{u}_{lower}$ denote the control limits, and $f$ refers to the system dynamics given in \cref{EqnMotion}. Like \citep{NHTTC}, we define time-to-collision as the first time a collision occurs under constant controls over the time horizon, enabling efficient optimisation.

\subsection{Chance-constrained Collision with Probabilistic Pedestrian Movements}
This subsection outlines a method to check for collisions between the robot and SP representations of pedestrian movement, to evaluate a time-to-collision value. Collision occurs when the distance between the robot and a pedestrian is below their collision radii. The set of coordinates where collision occurs, at time $t$, with the $n^{th}$ pedestrian is given by:
\begin{equation}
    \mathcal{C}^{n}_{t}=\left\{\bm{x}\in\mathbb{R}^2\;\middle|\;\|\bm{x}-\bm{o}^{n}(t)\|_{2}<r_{\xCoor}+r_{{o}_{n}}\right\},
\end{equation}
where $r_{\xCoor}$, $r_{\bm{o}_{n}}$ denote the radii of the robot and the pedestrian, and $\bm{o}^{n}(t)$ the pedestrian position. The position of pedestrians at a given time is uncertain, and given by the stochastic process defined in \cref{GPdef_eq_ch5}. Therefore, we specify collision at $t$ in a chance constraint manner, where a collision between the robot and a pedestrian is assumed to occur if the probability of the robot position, $\xCoor(t)$, being in set $C^{n}_{t}$ is above threshold $\epsilon$, i.e. $p(\xCoor(t)\in \mathcal{C}^{n}_{t})>\epsilon$. Time-to-collision is taken as the first instance that a collision occurs, i.e., when the probability of collision with pedestrians exceeds $\epsilon$. Thus, for the chance-constraint formulation, the set of robot motions colliding at $t$ is given by, 
\begin{align}
    \mathcal{C}_{t}=\{\xCoor(t)| \max_{n}\{p(\xCoor(t)\in \mathcal{C}^{n}_{t})\}>\epsilon\}.\label{ttc}
\end{align}
As the weight parameters, $\mathbf{W}$, of the stochastic process is a matrix normal distribution, the position distribution at a specified time is a multivariate Gaussian. By considering the random weight matrix distribution of the $n^{th}$ pedestrian, $\mathbf{W}^{n}$ and \cref{GPdef_eq_ch5}, we have:
\begin{equation}
    \bm{o}^{n}(t)\sim \mathcal{N}({\mathbf{M}^{n}}^{\top}\bm{\Phi}(t),\mathbf{V}^{n}\bm{\Phi}^{\top}(t)\mathbf{U}^{n}\bm{\Phi}(t)),
\end{equation}
where $\mathbf{M}^{n}, \mathbf{U}^{n}, \mathbf{V}^{n}$ are the parameters of the distribution of weight matrix $\mathbf{W}^{n}$, and $\bm{\Phi}(t)$ contains basis function evaluations. The position of the robot $\xCoor(t)$ is deterministic, thus $\xCoor(t)-\bm{o}^{n}(t)$ is also multi-variate Gaussian, with a shifted mean. The probability of collision with the $n^{th}$ pedestrian is the integral of a Gaussian over a circle of radius $r_{\xCoor}+r_{\bm{o}_{n}}$. Let $\bm{d}^{n}=\xCoor(t)-\bm{o}^{n}(t)$, we have the expressions:
\begin{align}
    &p(\xCoor(t)\in \mathcal{C}^{n}_{t})=\int_{\|\bm{d}^{n}\|_{2}<r_{\xCoor}+r_{\bm{o}_{n}}}p(\bm{d}^{n})\mathrm{d}\bm{d}^{n}, \\
    &p(\bm{d}^{n})=\mathcal{N}(\xCoor(t)-{\mathbf{M}^{n}}^{\top}\bm{\Phi}(t),\mathbf{V}^{n}\bm{\Phi}^{\top}(t)\mathbf{U}^{n}\bm{\Phi}(t)).
\end{align}
This expression is intractable to evaluate analytically, and an upper bound to integrals of this kind is given in \citep{Zhu2019ChanceConstrainedCA} by:
\begin{align}
    p(\xCoor(t)\in \mathcal{C}^{n}_{t})\leq \frac{1}{2}\bigg\{1+erf\bigg(\frac{r_{\xCoor}+r_{\bm{o}_{n}}-\bm{a}^{\top}\bm{d}^{n}(t)}{\sqrt{2\bm{a}^{\top}\bm{\Sigma}^{n}(t)\bm{a}}}\bigg)\bigg\},\label{upperbounds}\\
    \bm{\Sigma}^{n}(t) = \mathbf{V}^{n}\bm{\Phi}^{\top}(t)\mathbf{U}^{n}\bm{\Phi}(t), && \bm{a}=\frac{\bm{d}^{n}(t)}{\|\bm{d}^{n}(t)\|_{2}}, 
\end{align}
where $\bm{\Sigma}^{n}(t) \in \mathbb{R}^{2\times 2}$ is the covariance of the Gaussian $\bm{o}^{n}(t)$, and $erf(\cdot)$ is the error function. Using the defined \cref{upperbounds}, we can efficiently evaluate the collision probability upper-bound between the robot and each pedestrian, and take the first time probability upper-bound exceeds $\epsilon$ as the time-to-collision with pedestrians. By \cref{ttc,upperbounds} a collision occurs at $t$ if:
\begin{equation}
    \max_{n=1,\ldots, N}\Bigg\{\frac{1}{2}\bigg[1+erf\bigg(\frac{r_{\xCoor}+r_{\bm{o}_{n}}-\bm{a}^{\top}\bm{d}^{n}(t)}{\sqrt{2\bm{a}^{\top}\bm{\Sigma}^{n}(t)\bm{a}}}\bigg)\bigg]\Bigg\}>\epsilon. \label{dynamicCheck}
\end{equation}
That is, collisions arise if the probability of collision with any pedestrian is above threshold $\epsilon$.
The robot-pedestrian time-to-collision, $\tau_{\bm{o}}$, is the earliest time collision with a pedestrian occurs, namely $\tau_{\bm{o}}=\min_{t\in[0,T]}(t)$, subject to \cref{dynamicCheck}.

\subsection{Chance-constrained Collision with Occupancy Maps}
Static obstacles in the environment are often captured via occupancy maps. Time-to-collision values accounting for occupancy maps can be integrated into SPAN. An occupancy map can be thought of as a function, $f_m(\xCoor)$, mapping from position, $\xCoor\in\mathbb{R}^{2}$, to the probability the position is occupied, $p(\mathrm{Occupied}|\xCoor)$. We ensure the probability of being occupied at any coordinate on the circle of radius, $r_{\xCoor}$, around our robot position is below the chance constraint threshold, $\epsilon$. By considering the occupancy map as an implicit surface, a collision occurs at time $t$ if,
\begin{align}
    \max_{\phi\in[-\pi,\pi)}\bigg\{f_m\bigg(\xCoor(t)+r_{\xCoor}\begin{bmatrix} \sin(\phi)\\ \cos(\phi) \end{bmatrix}\bigg)\bigg\}>\epsilon. \label{staticCheck}
\end{align}
Maximising the independent variable, $\phi$, sweeps a circle around the robot to check for collisions. As the domain of $\phi$ is relatively limited and querying from occupancy maps is highly efficient, taking under $10^{-6}$s each query, a brute force optimisation can be done. The time-to-collision due to static obstacles is then the first time-step where a collision occurs, namely $\tau_{m}=\min_{t\in[0,T]}(t)$, subject to \cref{staticCheck}. We take the overall time-to-collision value for the optimisation in \cref{costOpt} as the minimum of the time-to-collision for both static obstacles and pedestrians, specifically $\tau=\mathrm{minimum}(\tau_{\bm{o}},\tau_{m})$.

\subsection{Solving the Control Problem}
We take a receding horizon control approach, continuously solving the control problem defined in \cref{costOpt} - \cref{costOptc}, and updating the controls for a single time-step. Algorithm \ref{algorithm} outlines our framework, bringing together the predictive model, the formulation and solving of control problem. A predictive model of obstacle movements is trained offline with motion data. The predictive model and a static map of the environment are provided to evaluate the time-to-collision, $\tau$, in the control problem. We take a ``single-shooting'' approach and evaluate the integral in the control cost via an Euler scheme with step-size $\Delta t=0.1s$. At each time step we check for collision by \cref{dynamicCheck,staticCheck}, and $\tau$ is the first time-step a collision occurs. The control problem is non-smooth, so we use the derivative-free solver COBYLA \citep{Powell1994ADS} to solve the optimisation problem, and obtain controls $\mathbf{u}$. As the control-space is relatively small, the optimisation can be computed quickly. We can re-solve with several different initial solutions in the same iteration to refine and evade local minima. Controls are updated at $10$ Hz. 

\begin{algorithm}[tb]
    \caption{\small{Stochastic Process Anticipatory Navigation}} \label{algorithm}
\Indmm\Indmm
    \KwIn{$\xState_0,\dot{\xState}_0,\mathbf{g},\mathcal{D},f_{m},\Delta t$}
    \KwOffline{Train neural network, $F$, to predict with dataset $\mathcal{D}$ (\cref{learn})}
     \KwInit{$t \gets 0$; $\xState(0)\gets \xState_0$; $\dot{\xState}(0)\gets \dot{\xState}_0$
        }
\Indpp\Indpp
    \While{goal not reach}{
        Observe $N$ pedestrians' coordinates, within a past time window, $\{\{\coorTS_{1},\ldots,\coorTS_{p}\}\}_{i=0}^{N}$\;
        $\mathcal{O}\gets\emptyset$ \tcp*{collects predictions}
        Predict $\{\mathbf{M}^n,\mathbf{U}^n,\mathbf{V}^n\}_{n=0}^{N}=F(\{\coorTS_{1},\ldots,\coorTS_{p}\}_{i})$;
        
        \For{$n=1$ to $N$\tcp{For all peds.}}{\tcp{Construct pedestrian futures as SP}
            $\bm{o}^n(t)={\mathbf{W}^n}^{\top}\bm{\Phi}(t)$, $\mathbf{W}^n\sim MN(\mathbf{M}^n,\mathbf{U}^n,\mathbf{V}^n)$\;
            $\mathcal{O}\gets\mathcal{O}\cup \{\bm{o}^n(t)\}$\;
        }
        \tcp{optimise for control with pred.}
        $\mathbf{u} \gets$ Solve \cref{costOpt,costOpta,costOptb,costOptc}, with $\mathcal{O},f_{m},\xState(t)$ to get controls\; 
        $\xState(t+\Delta t)\gets\mathrm{NextState(\xState(t),\mathbf{u})}$ \tcp{Execute controls}
        $t \gets t+\Delta t$\;
        \If{$\|\xCoor(t)-\mathbf{g}\|_2 < \epsilon_{goal}$}{
            \texttt{Terminates}\tcp*{goal reached}
        }
    }
    
\end{algorithm}

\section{Experimental Evaluation}
We empirically evaluate the ability of SPAN to navigate in environments with pedestrians and static obstacles. Results and desirable emergent behaviour are discussed below. 
\subsection{Experimental Setup}
\begin{figure}[t]
\centering
     \begin{subfigure}{0.32\textwidth}
         \centering
         \frame{\includegraphics[width=\textwidth]{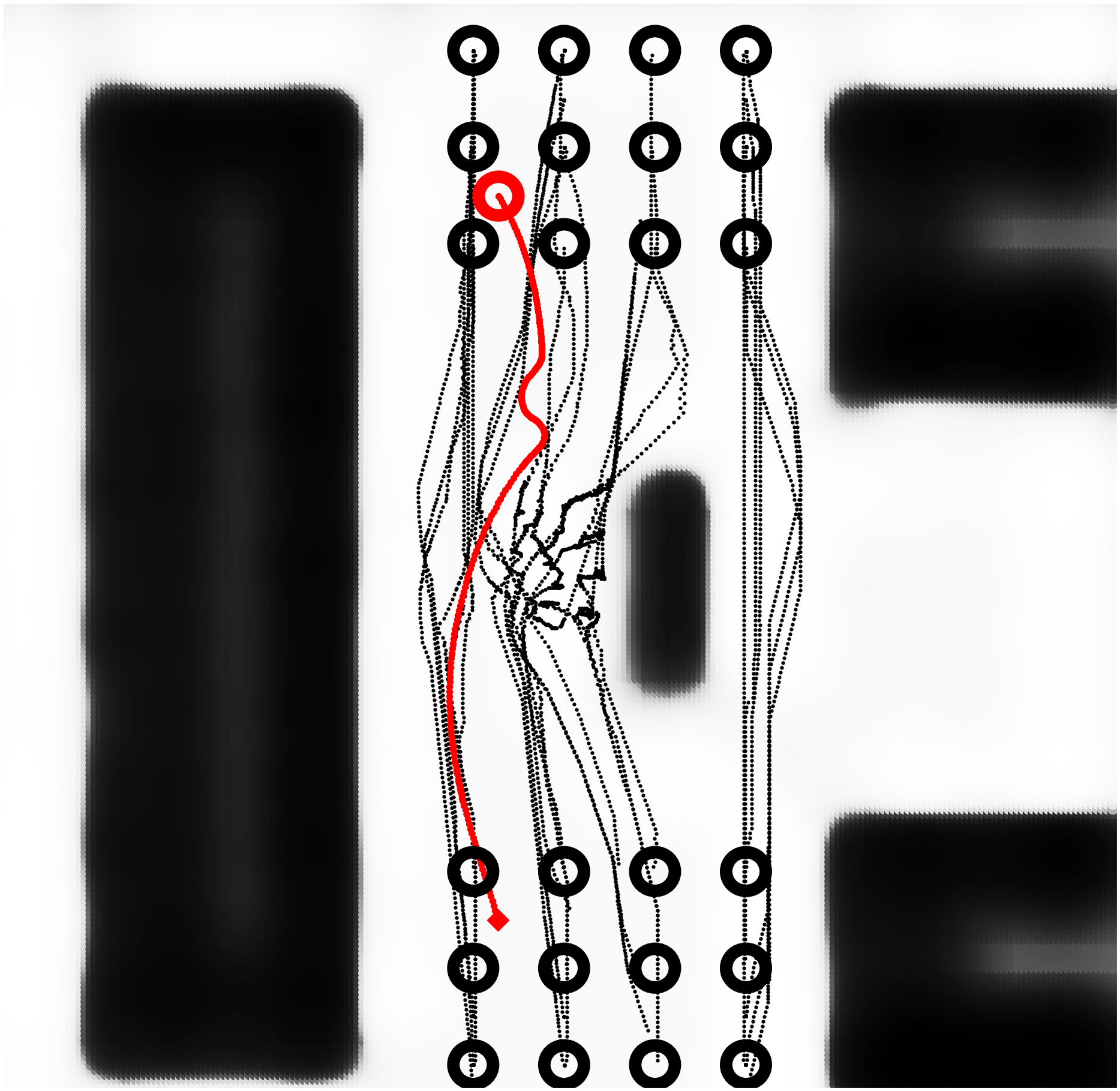}}
     \end{subfigure}
     \begin{subfigure}{0.32\textwidth}
         \centering
         \frame{\includegraphics[width=\textwidth]{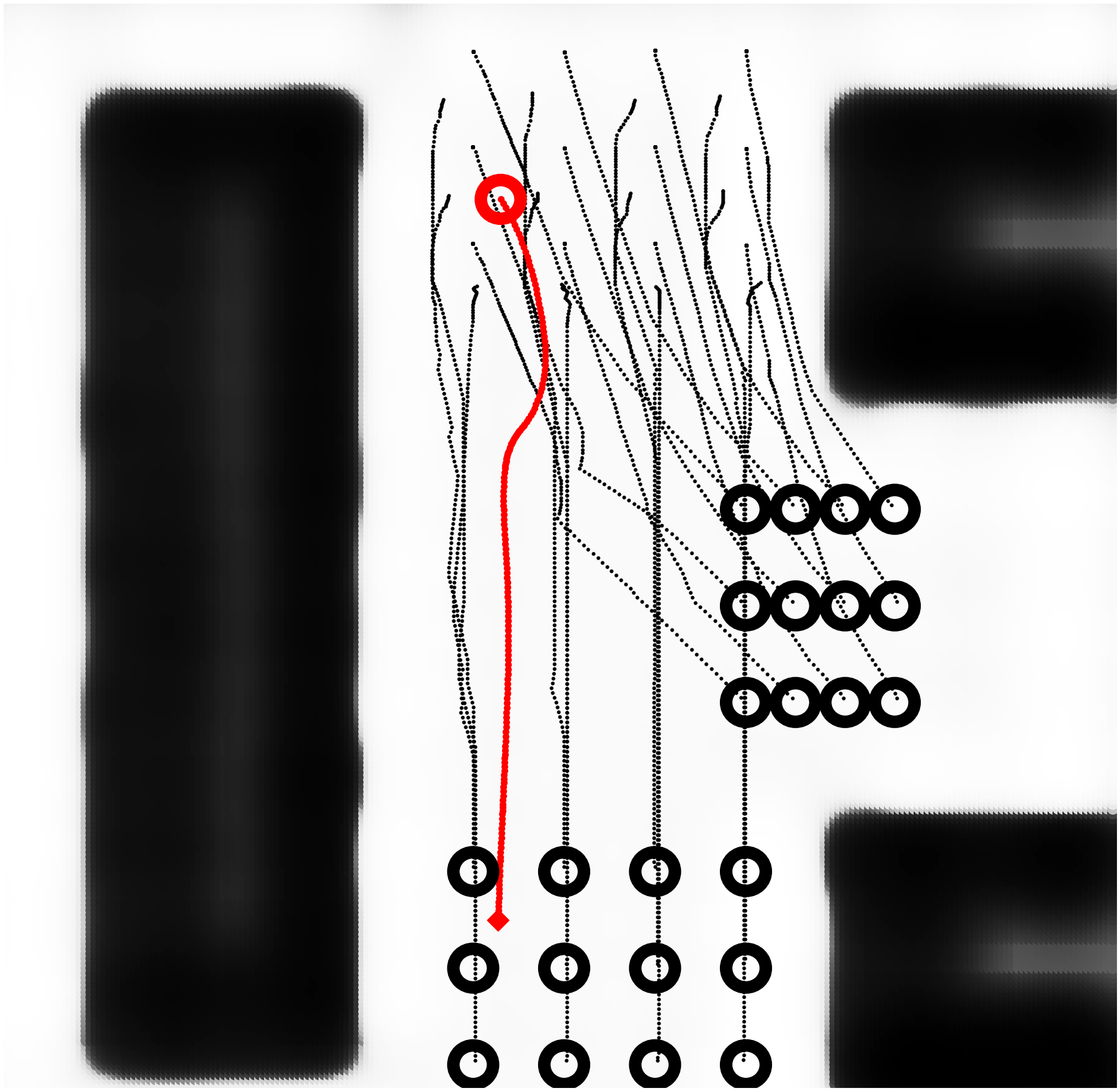}}
     \end{subfigure}
     \begin{subfigure}{0.32\textwidth}
         \centering
         \frame{\includegraphics[width=\textwidth]{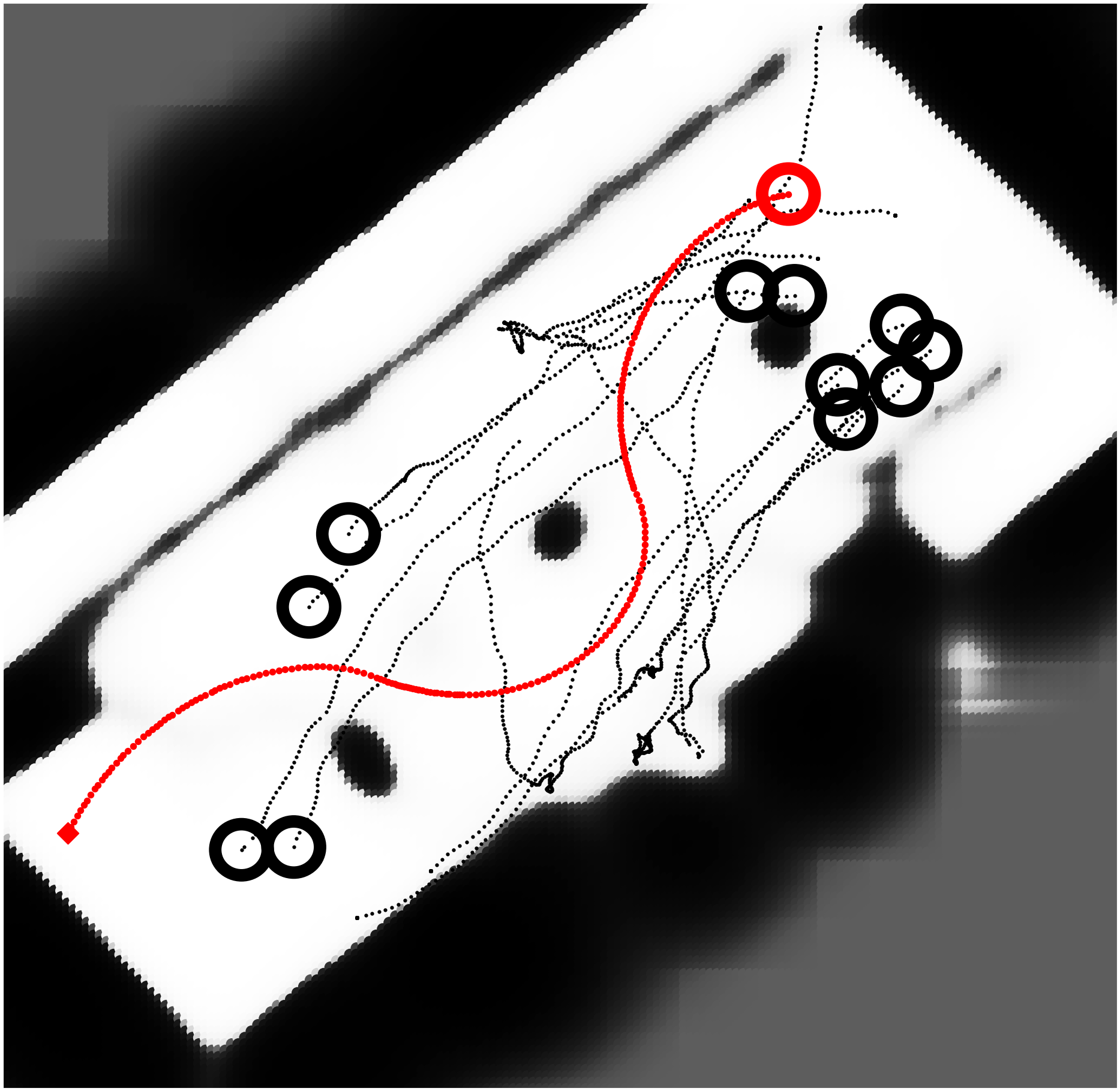}}
     \end{subfigure}
        \caption{Motion of pedestrians (black) and robot (red) overlaid on an occupancy map of the environments for Sim1 (left), Sim2 (center), and Dataset (right). Note that crossing trajectories does not necessarily indicate a collision, as moving agents could be at the same position at different times. The starting position of the pedestrians and robot are indicated by circles. The robot navigates smoothly and sensibly, remaining collision-free.}
        \label{expref0}
\end{figure}

Both simulated pedestrian behaviour and real-world pedestrian data are considered in evaluations. We use a power-law crowd simulator \citep{powerlaw,karamouzas17} with pedestrians' behaviour set as aggressive, and real-world indoor pedestrian data from \citep{thorDataset2019}. We construct two simulated environments, \emph{Sim 1} and \emph{Sim 2}, both with 24 pedestrians. \emph{Sim 1} models a crowded environment with noticeable congestion; \emph{Sim 2} models a relatively open environment with less frequent pedestrian interaction. The simulated pedestrians have a maximum speed of $1m/s$. Pedestrian trajectories collected at different times in dataset are overlaid, increasing the crowd density. The maps used in our experiments are shown in \cref{expref0}. We navigate the robot from starting point to a goal. Gazebo simulator \citep{gazebo} was used for visualisation. The following metrics are used:
\begin{enumerate}
    \item Time-to-goal (TTG): The time, in seconds, taken for the robot to reach the goal;
    \item Duration of collision (DOC): The total time, in seconds, the robot is in collision.  
\end{enumerate}
We evaluate SPAN against the following methods:
\begin{itemize}
\item Closed-loop Rapidly-exploring Random Trees$^*$ (CL-RRT$^*$). An asymptotic optimal version of \citep{kuwata2008motion} with a closed-loop prediction and a nonlinear pure-pursuit controller \citep{amidi1991integrated}. \citep{kuwata2008motion} demonstrates an ability to handle dynamic traffic, and is a DARPA challenge entry. At each iteration, a time budget of $0.1s$ is given to the planner to find a free trajectory towards the goal, with pedestrians represented as obstacles. If a feasible solution is not found, the robot will remain at its current pose. 
\item A reactive controller. The reactive controller continuously solves to find optimal collision-free controls for the next time-step, without anticipating future interaction.  
\end{itemize}
In our experiments, we set the time-to-collision weight, $\kappa$=100; the trajectory length-scale hyper-parameter, $\gamma=0.01$; the control limits on linear/ angular velocity as $-1\leq\bm{v}\leq 1$ and $-1 \leq \omega \leq 1$. Each iteration we solve for a ``look-ahead'' time horizon of $T=4.0s$. The collision threshold is set at $\epsilon = 0.25$, and the radii for the robot and the pedestrians are given as $r_{\xCoor},r_{\bm{o}} = 0.4m$. In the simulated setups, the predictive model is trained with additionally generated data from the crowd simulator; in the dataset setup, $80\%$ of the data was used to train the predictive model, with the remainder for evaluation. We condition on observed pedestrian movements in the last 0.5s, to predict the next 4.0s. Occupancy maps provided are implemented as continuous occupancy maps \citep{HM, HilbertMaps}, with the method in \citep{HilbertMaps}. Controls were solved with different initialisations 40 times per iteration. The predictive model is implemented in Tensorflow \citep{tflow}, control formulation and COBYLA \citep{Powell1994ADS} are in FORTRAN, interfacing with Python2. Videos of a controlled non-holonomic robot navigating through simulated aggressive crowds and real pedestrian data can be found in \url{https://youtu.be/bLDCnE5Lo6Q} and \url{https://youtu.be/ZJ5Uh0JgMdA}.

\begin{figure}
\centering
     \begin{subfigure}{0.49\textwidth}
         \centering
         \frame{\includegraphics[width=\textwidth]{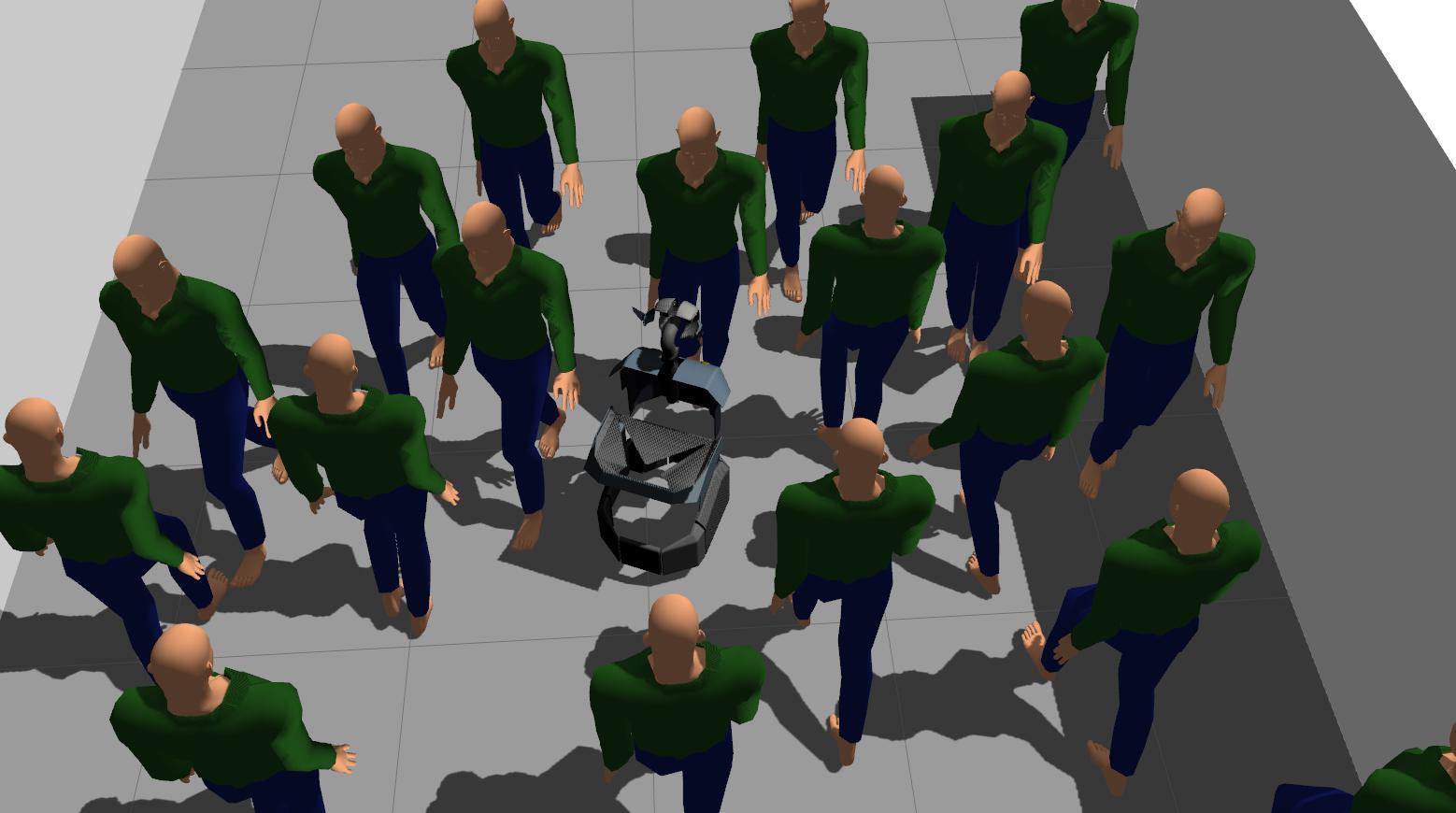}}
     \end{subfigure}
     \begin{subfigure}{0.49\textwidth}
         \centering
         \frame{\includegraphics[width=\textwidth]{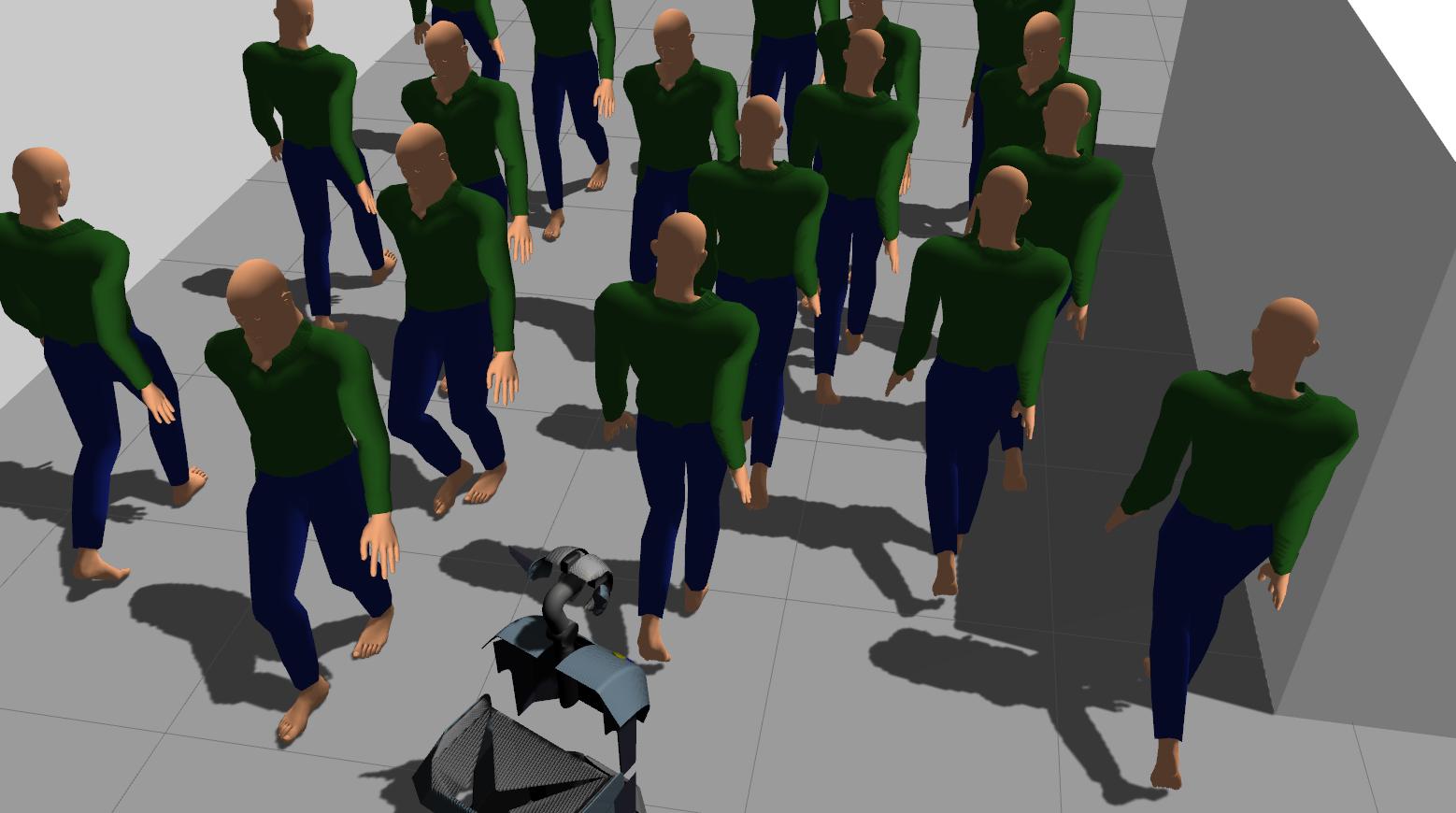}}
     \end{subfigure}
\caption{\emph{Sim 1} setup. (Left): The reactive controller often drives head-on into the crowd, leading the robot to become stuck and resulting in minor collisions. A considerable amount of time is needed to resolve the congestion. (Right): The robot controlled by SPAN evades congestion when possible, and follows behind pedestrians that are moving in the same direction.} \label{CrowdsFig}
\end{figure}

\subsection{Experimental Results}
The resulting motion trajectories by the controlled robot in the experiments are illustrated in \cref{expref0}, where the motion trajectories of the robot is marked in red, while those of pedestrians are marked in black. At a glance, we see that the controlled robot navigates smoothly and sensibly. The evaluated metrics are tabulated in \cref{tableresults}. The average computational times of an entire iteration of SPAN, including prediction and obtaining controls, are 27ms, 31ms, 33ms in  \emph{Dataset}, \emph{Sim 1}, \emph{Sim 2} setups respectively, with no iteration taking more than 50ms in any setup.

SPAN is able to control the robot to reach the goal, without collisions, in all of the experiment environments. In particular, relative to the compared methods, SPAN demonstrates markedly safer navigation through the \emph{Sim 1} and \emph{Dataset} environment setups. Both of these environments are relatively cluttered, with regions of congestion arising. We observe that both CL-RRT$^*$ and the reactive controller has a greater tendency to freeze amongst large crowds, while SPAN gives more fluid navigation amidst crowds. \Cref{CrowdsFig} shows how the reactive controller (left figure) and SPAN (right figure) navigate the robot through crowds in the dense \emph{Sim 1} environment. The reactive controller often controls the robot to drive towards incoming pedestrians head-on. This behaviour often results in driving into bottleneck positions and congestion arising, with the robot then unable to escape from a trapped position. On the other hand, the robot controlled by SPAN avoids getting stuck in crowds, as the future position of pedestrians are anticipated. Incoming pedestrians are actively evaded, while the robot follows other pedestrians through the crowd. 

\begin{figure}[t]
\centering
\begin{tikzpicture}
\node (img) {
\begin{subfigure}{0.32\textwidth}
         \centering
         \includegraphics[width=\textwidth]{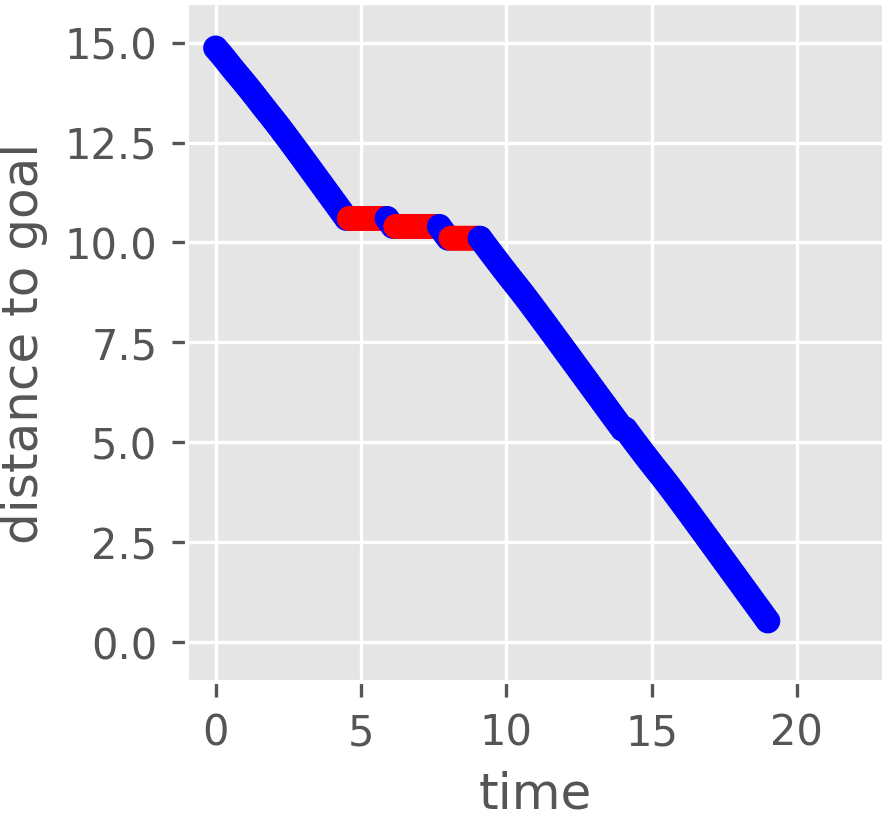}

     \end{subfigure}
     \begin{subfigure}{0.32\textwidth}
         \centering
         \includegraphics[width=\textwidth]{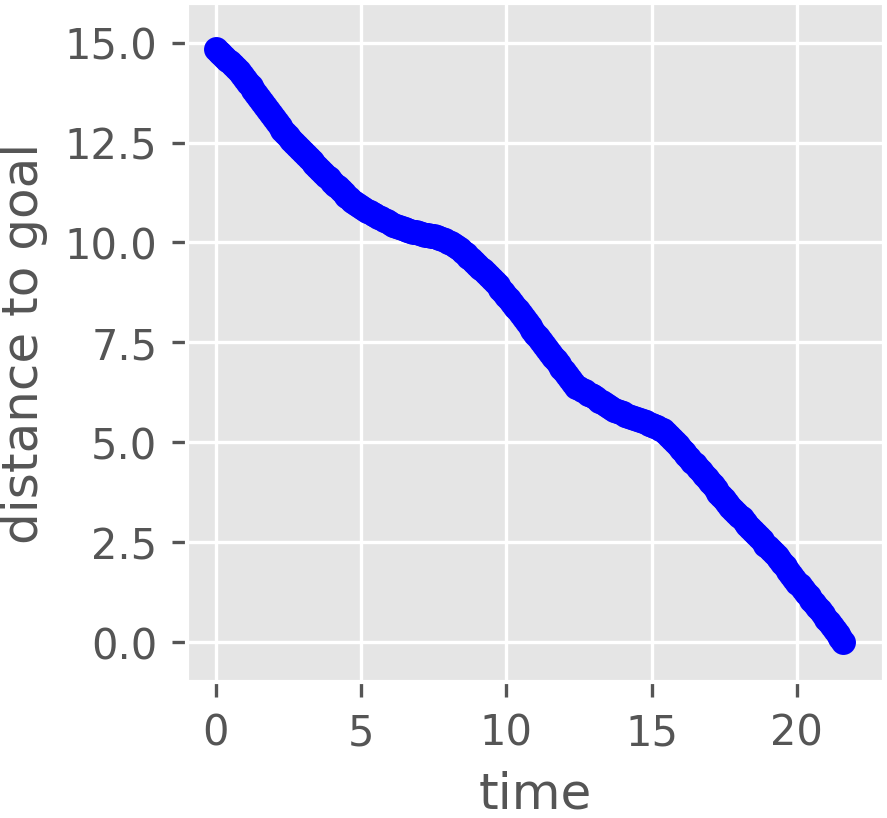}

     \end{subfigure}
     \begin{subfigure}{0.3\textwidth}
         \centering
         \includegraphics[width=\textwidth]{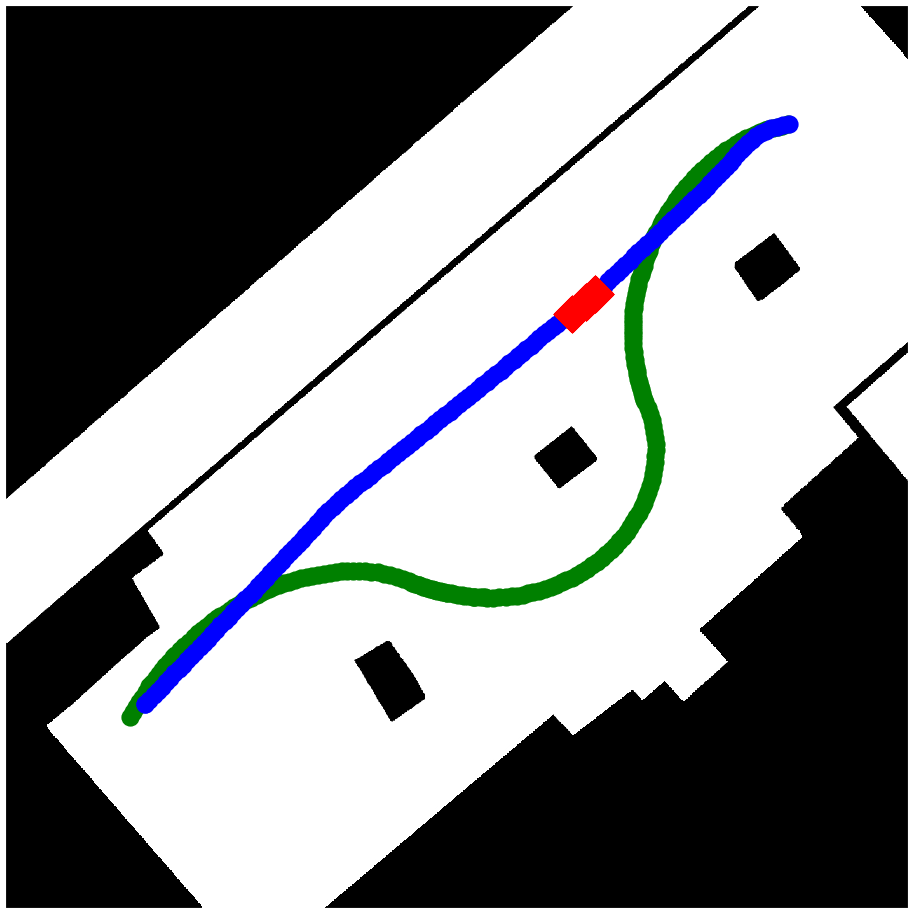}

     \end{subfigure}
     };
    \node at (-3.6,-1.15){\scriptsize \textsc{cl-rrt}$^*$};
    \node at (-0.9,-1.17){\scriptsize \textsc{span}};
\end{tikzpicture}%
     \caption{(Left, Center) shows time against the distance between the robot and goal, for the CL-RRT$^{*}$ and SPAN respectively. The times that are in collision are given in red, while the collision-free times are in blue. (Right) shows the motion of the robot as controlled by CL-RRT$^{*}$ in blue, and that as controlled by ours in green. Time-steps where the robot is in collision are in red. We see that SPAN is able to provide a safe path for the controlled robot, while CL-RRT$^{*}$ provides an aggressive path, resulting in over $3s$ of collision.}\label{aggres}
\end{figure}

In the \emph{Dataset} setup, both the compared CL-RRT$^*$ and the reactive controller produce very collision-prone navigation trajectories. This is due to the completely dominate pedestrian behaviour, which follows collected trajectory data, whereas the simulated pedestrian crowds also attempts to avoid collision with the robot. Both CL-RRT$^*$ and the reactive controller give a relatively straight trajectory towards the goal and are in collision for more than $3s$, colliding with three different pedestrians. Meanwhile SPAN produces a safe and meandering trajectory. The trajectories of the controlled robot by CL-RRT$^{*}$ and SPAN is illustrated in \cref{aggres}, along with plots of how the distance to goal changes in time. We argue, in the \emph{Dataset} setup, although the time-to-goal of SPAN is marginally longer, the navigated path is much safer with no collisions, while the compared methods have more than 3s of collision.  Additionally, all three of the evaluated methods perform similarly in the comparatively open \emph{Sim 2} environment, producing collision-free and efficient  trajectories. This is due to the ample space in the environment, with relative little congestion occurring. 

\begin{figure}[t]
\centering
     \begin{subfigure}{0.32\textwidth}
         \centering
         \frame{\includegraphics[width=\textwidth]{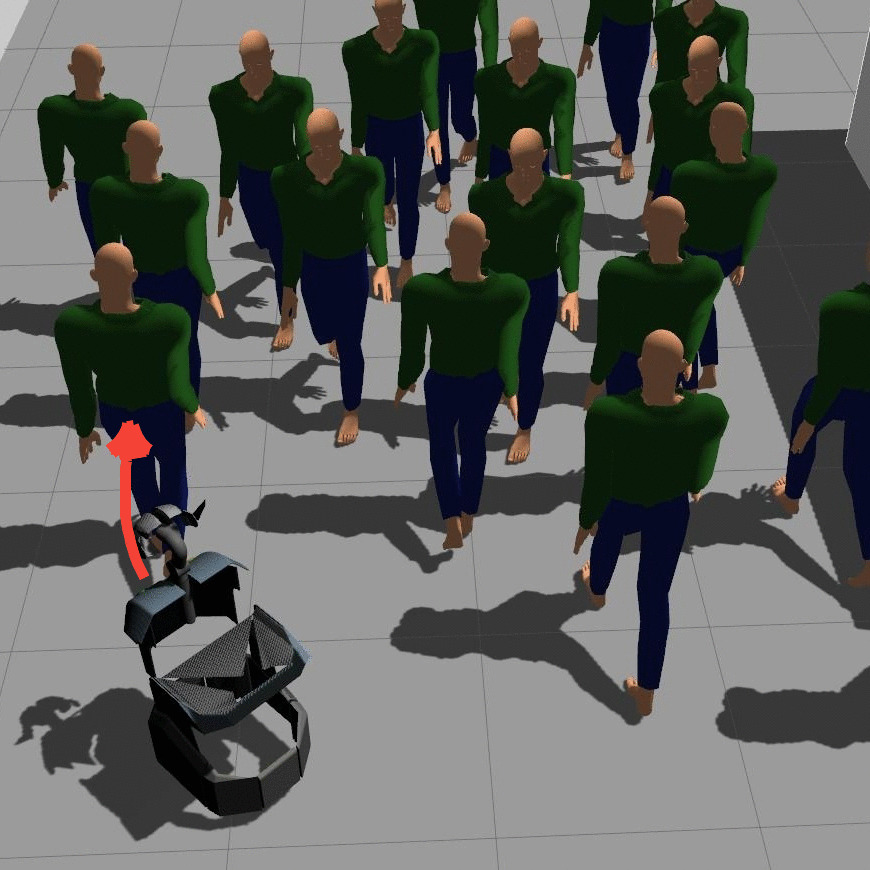}}
     \end{subfigure}
     \begin{subfigure}{0.32\textwidth}
         \centering
         \frame{\includegraphics[width=\textwidth]{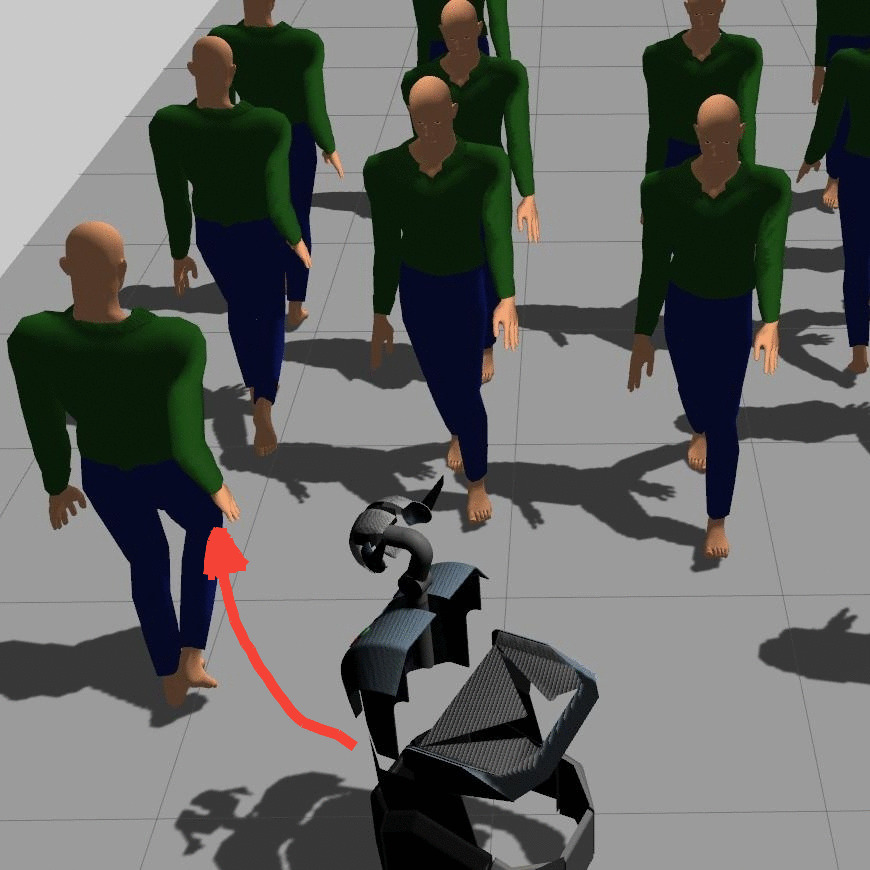}}
     \end{subfigure}
          \begin{subfigure}{0.33\textwidth}
         \centering
         \frame{\includegraphics[width=\textwidth]{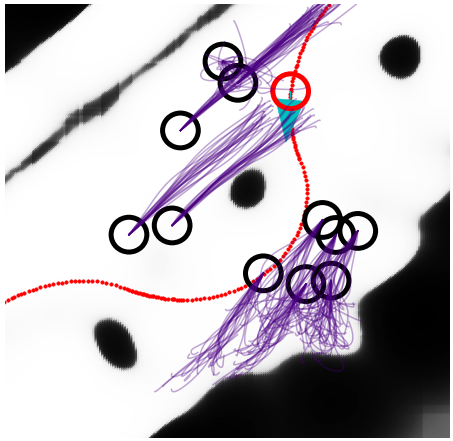}}
     \end{subfigure}
        \caption{Anticipatory behaviour of our controlled robot can be observed. (Left, center) The robot follows (red arrow) pedestrians anticipated to move ahead, while preemptively avoiding incoming pedestrians. (Right) Robot as red circle and pedestrians as black. Sampled trajectories (in purple, next 4.0s) visualise predicted SP, and the robot trajectory is in red. The robot evades an incoming group of pedestrians preemptively, and moves toward pedestrians predicted to move away.}
        \label{expref1}
\end{figure}

\begin{table}[h]
\centering
\scalebox{1.0}{
\begin{tabular}{ll|lll}
\toprule

Method & Metrics               & Dataset & Sim 1 (crowded) & Sim 2 (open) \\ \hline
SPAN (ours)   & TTG       & 21.7        & 25.9        & 17.4    \\
       & DOC & 0.0           & 0.0           & 0.0       \\
CL-RRT$^*$ & TTG        & 19.0          & 36.8        & 17.8    \\
       & DOC & 3.7         & 1.1         & 0.0      \\
Reactive & TTG        &   19.1      & 43.5       & 17.1   \\
Controller & DOC &  3.4      & 0.4        & 0.0      \\
\bottomrule
\end{tabular}}\caption{The evaluations of SPAN against compared methods. TTG gives the time (in seconds) taken by the robot to reach the goal, and DOC gives the total duration (in seconds) the robot is in collision. SPAN finds a collision-free solution in all three evaluated set-ups.}\label{tableresults}

\end{table}
\subsection{Emergent Behaviour}

The formulated control problem in \cref{costOpt,costOpta,costOptb,costOptc} does not explicitly optimise for the following of other agents. However, we observe following behaviour, behind other pedestrians, also moving in the goal direction, through crowds. \Cref{expref1} (left, center) show the controlled robot following moving pedestrians through crowds, while avoiding those moving towards the robot. This is also detailed in \cref{expref1} (right), where predicted future positions are visualised by sampling and overlaying trajectories (in purple) from the SP model. The entire navigated motion trajectory taken by the robot is outlined in red. We observe the robot taking a swerve away from the group of incoming pedestrians preemptively, and turning to follow the other group predicted to move away. At that snapshot, the pedestrians in the lower half of the map blocks the path through to the goal, but are all predicted to move away, giving sufficient space for our robot to follow and pass through. The emergence of the crowd-following behaviour facilitates smooth robot navigation through crowds.

\section{Summary}
We introduce Stochastic Process Anticipatory Navigation (SPAN), a framework for anticipatory navigation of non-holonomic robots through environments containing crowds and static obstacles. We learn to anticipate the positions of pedestrians, modelling future positions as continuous stochastic processes. Then predictions are used to formulate a time-to-collision control problem. We evaluate SPAN with simulated crowds and real-world pedestrian data. We observe smooth collision-free navigation through challenging environments. A desired emergent behaviour of following behind pedestrians, which move in the same direction, arises. The stochastic process representation of pedestrian motion is compatible as an output for many neural network models, bringing together advances in motion prediction literature with robot navigation. Particularly, conditioning on more environmental factors to produce higher quality predictions within SPAN, is a clear direction for future work.

In the upcoming \cref{chap5b}, we shall develop a framework to predict how dynamic agents in an environment should move based on the environment structure. Humans have good intuition of what movement patterns in an environment when provided a floor-plan of it. We seek to endow robots with the same capability, by learning from experience.
\pagebreak

\chapter{Trajectory Generation in New Environments from Past Experiences}\label{chap5b}\blfootnote{This chapter has been published in IROS as \cite{OTNet}.}
\usetikzlibrary{shapes.geometric, arrows, arrows.meta}
\tikzset{%
  >={Latex[width=2mm,length=2mm]},
            base/.style = {rectangle, rounded corners, draw=black,
                           minimum width=0.5cm, minimum height=0.5cm,
                           text centered},
  activityStarts/.style = {base, fill=gray!20},
       startstops/.style = {base, fill=gray!20},
    activityRuns/.style = {base, fill=gray!20},
         process/.style = {base, fill=gray!20},
         decision/.style= {diamond, fill=gray!20}
}

\newcommand\wip[1]{{\color{red}{[~#1}~]}}

\newcommand\given[1][]{\;#1\vert\;}

Understanding movement trends in dynamic environments is critical for autonomous agents, such as service robots and delivery vehicles, to achieve long-term autonomy. This need is highlighted by the increasing interest in developing mobile robots capable of coexisting and interacting safely and helpfully with humans. Anticipating likely motion trajectories allows autonomous agents to anticipate future movements of other agents and thus navigate safely as well as plan socially compliant motions by imitating observed trajectories.  

Many methods predicting motion extrapolate partially observed trajectories, without incorporating knowledge of the environment. Example  of such methods include constant acceleration\citep{SurveyDynamics}, filtering methods \citep{motionpredFilt}, and auto-regressive models \citep{Agarwal2004TrackingAM}. Advances in machine learning have lead to the development of methods which learn the motion behaviours from a dataset of trajectories. By training on motion data collected in the same environment, some learning-based methods can learn the general flow of movement \citep{sptemp, Flow, DirectionalGridMaps}, which implicitly accounts for environmental geometry. However, such methods are typically environment-specific, requiring predictions to be made in the same environment as that the training data was collected. Further methods, such as \citep{Sadeghian_2019_CVPR} attempt to encode local parts of the environment along with trajectory extrapolation. 

\begin{figure}[tb]
\centering
\includegraphics[width=0.8\linewidth]{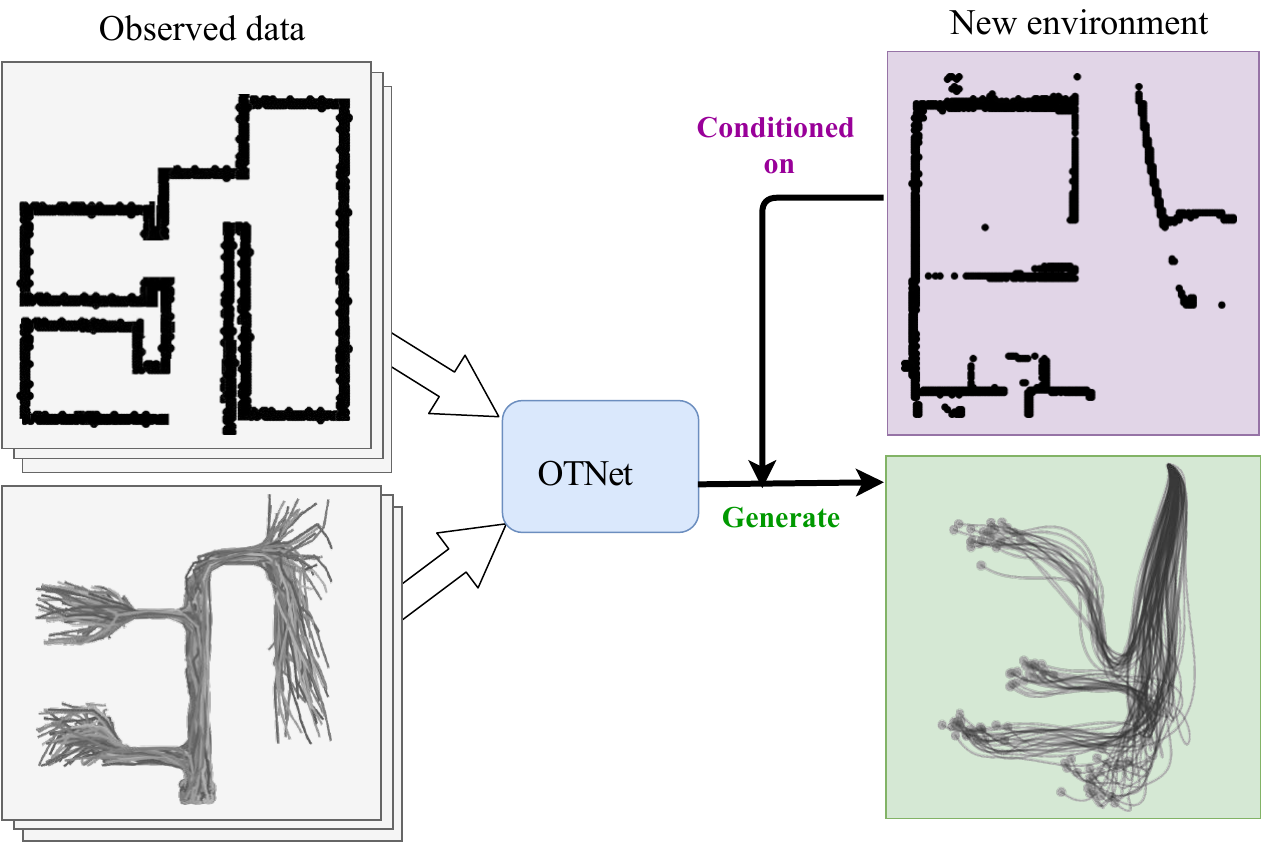}
\caption{
    We use OTNet to transfer motion patterns from maps where motion has been observed to new environments where motion has not been observed, by conditioning on the occupancy map of the new environment. 
}\label{ProblemIntro}
\end{figure}

Through experience, humans have developed the ability to anticipate movement patterns based on the layout of the environment. We hypothesise that the structure of environments contains information about how objects move within the environment. By considering 2D environment floor plans, given as occupancy grid maps, it is possible to transfer motion patterns from training environments to new environments where no motion observations have been made. We propose a probabilistic generative model, Occupancy-Conditional Trajectory Network (OTNet), capable of generating motion trajectories for new unseen environments by generalising motion trajectories from previously observed environments, as shown in \Cref{ProblemIntro}. We empirically demonstrate that our model is capable of generating motion trajectories in simulated and real-world environments by motion behaviour observed in other environments. We can additionally refine our predicted trajectory patterns, such as fixing its start point. We envision OTnet to provide priors about long-term trajectory patterns which can be integrated downstream in trajectory prediction models which reason about the immediate movement of surrounding agents, such as the predictive model in SPAN, outlined in \cref{chap5}.

\section{Related Work}\label{sec:related-work}


\subsection{Motion Trajectory Prediction}
Estimating likely motion trajectories has been studied for a long time. Early simple methods to predict motion are often dynamics-based methods which extrapolate based on the laws of physics, such as constant velocity and constant acceleration models \citep{SurveyDynamics}, with more complex Dynamics-based methods are utilised in \citep{cycle, ExDynamics2}. These models typically requires making a priori assumptions about agent motions, leaving little room to learn from observations. Other attempts at modelling motion trajectories include building dynamic occupancy grid maps based on occupancy data over time \citep{TOG,TempOg,GridBased}, inverse motion planning methods \citep{Ziebart2009PlanningbasedPF}, and hidden Markov methods \citep{Vasquez2009GrowingHM}. Recent developments in machine learning have led to an interest in data-driven models \citep{Gupta2018SocialGS}, which make fewer assumptions but rely heavily on observed data. A class of learning-based methods focus on learning a map of motion in a specific environment to implicitly capture map-aware motion patterns in the environment \citep{sptemp,DirectionalGridMaps,Flow,TAROS}.

\subsection{Trajectory Generation}
Neural network based generative models have been used in motion prediction, where they typically generate complete trajectories conditioned on an incomplete one. Generative adversarial networks (GANs) \citep{Gans} are a popular class of generative models that for trajectory generation \citep{Li2019InteractionawareMT,Gupta2018SocialGS}. Conditional Variational Autoencoders (CVAE)~\citep{CVAE} are another class of generative models \citep{CVAEPrediction,Ivanovic_2019_ICCV} that have been utilised in the same problems. Similar to this work, ideas for transferring motion patterns in specific transport scenarios are considered in \citep{Jaipuria2018ATP}. Methods using neural networks to generate trajectories based on observed trajectories have also gained attention in the motion planning community. A recent work, Motion Planning Network (MPNet)~\citep{qureshi2019motion}, aims to generate trajectories by learning from a dataset of optimal trajectories in various simulated environments.

\section{Methodology}


\subsection{Problem Formulation}
This paper addresses the problem of generating motion trajectories in new environments, by transferring trajectories data observed in other training environments. \Cref{ProblemIntro} gives a high level idea of the problem at hand: We have observed data in the form of motion trajectories and maps for training (left), and provided an occupancy map of a novel environment (upper right), we wish to generate likely motion patterns (lower right).

We assume to have a training dataset consisting of occupancy representations of diverse environments and a collection of trajectories observed within each map. We denote the dataset as $\mathcal{D}=\{\mathcal{M}_{n},\{\xi_{p}\}_{p=1}^{P_n}\}^{N}_{n=1}$, where there are $N$ maps and corresponding sets of observed trajectories, $\mathcal{M}_n$ is the $n^{th}$ occupancy map, and $\{\xi_{p}\}_{p=1}^{P_n}$ is the set of $P_n$ trajectories collected in the corresponding environment. Note that the number of trajectories in each environment may differ.  

We now have a new map, $\mathcal{\mathcal{M}^{*}}$, where no trajectory observations have been made, and seek to generate new likely motion trajectories on $\mathcal{\mathcal{M}^{*}}$, based on patterns we see in our training dataset. Key requirements of a solution to the problem include: (1) the ability to generate trajectories are diverse enough to capture multiple trajectory patterns; (2) the ability to generate trajectories that start from a specified coordinate.   

\subsection{Overview of OTNet}\label{sec:OTNet}
On a high level, our method encodes maps as similarities between themselves and a pre-defined collection of maps, and represents trajectories as parameters of a function. A mixture density network \citep{Bishop94mixturedensity} is then used to predict distributions over function parameters, conditional on an encoded map. 


The training process is illustrated in \Cref{TrainingModel}, and can be summarised as:
\begin{enumerate}[leftmargin=.6cm]
    \item Construct feature vectors of similarities, $\bm{\phi}$, from each training map. Intuitively, we similarly shaped floor-plans to have similar motion patterns. Our map encoding explicitly introduces the notion of map similarity to the learning process. Details in \Cref{EncodingOcc}.
    \item Following the continuous trajectory representation outlined in \cref{chap5}, We represent trajectory data as vectors of weight parameters, $\mathbf{w}$, of a function. This allows us to operate on on trajectory data sequences which contain differing numbers of waypoints. Each trajectory can typically be represented with much fewer parameters than waypoint coordinates.

    \item We use a mixture density network (MDN) \citep{Bishop94mixturedensity} to learn the distribution over weight parameters conditioned on the map feature vectors, $p(\mathbf{w}|\bm{\phi})$. Details in \Cref{MDN}.
\end{enumerate}
A brief overview of the generative process is illustrated in \Cref{QueryModel}. 

After the MDN has been trained, we construct a feature vector $\bm{\phi}^{*}$ of a new map $\mathcal{M}^{*}$, and query the MDN to obtain $p(\mathbf{w}|\bm{\phi}^{*})$. Vectors of $\mathbf{w}$ can be sampled from $p(\mathbf{w}|\bm{\phi}^{*})$, and each sample $\mathbf{w}$ can be used to generate a new trajectory. As there are no explicit constraints in the MDN to prevent trajectories from overlapping with occupied regions, and we can efficient sample check for collisions, we accept collision-free trajectories to output. If we are only interested in trajectories that start at a certain point, we can generate trajectories conditional on a specified start-point.
\begin{figure}[t]
\centering
\begin{minipage}[t]{.4\textwidth}
\centering
\begin{tikzpicture}[node distance=1.6cm,
    scale=0.6, every node/.style={scale=0.6},
    align=center]
  \node (MapIn)[activityStarts] {Occupancy Maps\\ $\{\mathcal{M}_1,\ldots,\mathcal{M}_N\}$};
  \node (GenerateFeatures)     [process, below of=MapIn, yshift=-1.6cm]{Encode each map\\ as feature vector $\bm{\phi}$};
 \node (TrajIn)[activityStarts, right of=MapIn, xshift=3cm] {Trajectories\\ $\{\{\xi_{p}\}_{p=1}^{P_1},\ldots, \{\xi_{p}\}_{p=1}^{P_{N}}\}$};
\node (GenerateEmbeddings)     [process, below of=TrajIn, yshift=-1.6cm]          {Abstract each\\ trajectory as\\ vector $\mathbf{w}$ };
\node (TrainModel)      [activityRuns, below of=GenerateFeatures, xshift=3cm, yshift=-0.7cm] {Train MDN to model $p(\mathbf{w}|\bm{\phi})$};
                                                    
  \draw[->]             (MapIn) -- (GenerateFeatures);
  \draw[->]             (TrajIn) -- (GenerateEmbeddings);
  \draw[->]             (GenerateFeatures) -- (TrainModel);
  \draw[->]             (GenerateEmbeddings) -- (TrainModel);
  \end{tikzpicture} \caption{Process of learning model to generate $p(\mathbf{w}|\bm{\phi})$}\label{TrainingModel}
\end{minipage}
\begin{minipage}[t]{.4\textwidth}
\centering
\begin{tikzpicture}[node distance=1.5cm,
    scale=0.6, every node/.style={scale=0.6},
    align=center]
    \tikzstyle{decision} = [diamond, inner sep=-.3ex, text centered, draw=black, fill=green!30]
    \node (FeatureIn)[activityStarts] {Queried map, $\mathcal{M}^{*}$};
    \node (Generatefeatures) [process, below of=FeatureIn]{Generate feature $\bm{\phi}^{*}$};
    \node (ObtainProbability)     [process, below of=Generatefeatures]{Predict $p(\mathbf{w}|\bm{\phi}^{*})$};
    \node (SampleW)[process, below of=ObtainProbability] {Sample $\mathbf{w}$\\ to construct trajectories};
     \node (retain)     [process, below of=SampleW]          {Collision check};
    \node (Finish)     [activityRuns, below of=retain]          {Output generated trajectory};   
  \draw[->]             (FeatureIn) -- (Generatefeatures);
  \draw[->]             (Generatefeatures) -- (ObtainProbability);
  \draw[->]             (ObtainProbability) -- (SampleW);
  \draw[->]             (SampleW) -- (retain);
 \draw[->]             (retain) -- (Finish);
  \end{tikzpicture} \caption{Process of generating trajectories}\label{QueryModel}
 \end{minipage}
\end{figure}

\subsection{Encoding of Environmental Occupancy}\label{EncodingOcc}
We represent occupancy maps as a vector of similarities between the given environment and a representative database of other maps. Intuitively, we can think of this as operating in the \emph{space of maps}, where we pin-point the map of interest by its relation with other maps, where maps similar to one another are closer together in the space of map. We expect similar maps to have similar motion trajectory patterns. Related ideas have been explored in the context of pseudo-inputs for sparse Gaussian process \citep{SparseGP}.

 The Hausdorff distance is a widely used distance measure to shapes and images \citep{ImageH}, and can be efficiently computed in linear time. The Hausdorff distance measures the distance between two finite sets of points, and allows us to make comparisons between our maps. We place the \emph{Hausdorff distance} into a \emph{distance substitute kernel} \citep{DistanceKernel}, to obtain our similarity function. 

Given two sets of points $A=\{a_1, a_2, \ldots, a_n\}$ and $B=\{b_1, b_2,\ldots, b_m\}$, and in general $n$ and $m$ are not required to be equal, the one-sided Hausdorff distance between the two sets is defined as:
\begin{equation}
    \hat{\delta}_{H}(A,B)=\max_{a\in A}\min_{b \in B}||a-b||.
\end{equation}
The one-sided Hausdorff distance is not symmetric, we enforce symmetry by taking the average of $\hat{\delta}_{H}(A,B)$ and $\hat{\delta}_{H}(B,A)$, i.e.:
\begin{equation}
    \delta_{H}(A,B)=\frac{1}{2}(\hat{\delta}_{H}(A,B)+\hat{\delta}_{H}(B,A)).
\end{equation}
We can then define a similarity function between two sets $A$ and $B$, analogous to a distance substitute kernel described in \citep{DistanceKernel}, as:
\begin{equation}
    S_{H}(A,B)=\exp\Big\{-\frac{\delta_{H}(A,B)^2}{2\ell_{H}}\Big\},
\end{equation}
where $\ell_{H}$ is a length scale hyper-parameter. 

We extract occupied edge points of binary grid maps and evaluate the similarity function between each map in the representative database. We can select a set of maps within our training data to be in the representative database, or when the number of maps is low, we can consider all the training maps to be in the database. A simple method of selecting the set of examples to be in a representative database are described in \citep{sptemp}. If we have $N$ training maps and $M$ representative maps in the database, feature vector for the $n^{th}$ map, $\bm{\phi}_n$, is:

\begin{equation}
    \begin{bmatrix}
    \bm{\phi_1}\\
    \vdots\\
    \bm{\phi_N}
    \end{bmatrix}
    =
    \begin{bmatrix} 
    S_{H}(\mathcal{M}_1,\mathcal{M}_1) & \ldots & S_{H}(\mathcal{M}_1,\mathcal{M}_{M}) \\
    \vdots & \ddots & \vdots\\
    S_{H}(\mathcal{M}_N,\mathcal{M}_1), & \ldots, & S_{H}(\mathcal{M}_N,\mathcal{M}_{M}).
    \end{bmatrix}. 
\end{equation}
For every map $\mathcal{M}$ in our dataset, there is a corresponding vector of similarities $\bm{\phi}\in\mathbb{R}^{M}$. The $m^{th}$ element in the feature vector $\bm{\phi}_n$ denotes the similarity between the $n^{th}$ occupancy map in the training dataset, and the $m^{th}$ occupancy map in the database.

\subsection{Learning a Mixture of Stochastic Processes}\label{MDN}
Like \cref{learn}, we begin by modelling motion as continuous trajectories. We differ slightly from \cref{learn}, by considering the $x$ and $y$ dimensions independently. This produces fewer parameters we shall need to eventually learn. We define a normalised timestep parameter $\tau \in [0,1]$. A continuous trajectory can be modelled function, $\bm{\Xi}(\tau)=[x(\tau),y(\tau)]$, which maps $\tau$ to the x and y coordinates of the trajectory. We model $x(\tau)$ and $y(\tau)$ as weighted sums of $M$ fixed radial basis functions centred on evenly spaced $\tau$ values, with weight vectors $\mathbf{w}_x$ and $\mathbf{w}_y$. These weight parameters can be found from discrete trajectory data $\xi$ via solving regression equations akin to \label{GPdef_eq}. We can then query the trajectory coordinates at $\tau^{*}$ by evaluating $x(\tau^*)=\mathbf{w}_x^{T}\mathbf{k}(\tau^*)$ and $y(\tau^*)=\mathbf{w}_y^{T}\mathbf{k}(\tau^*)$. We denote the concatenation of $\mathbf{w}_x$ and $\mathbf{w}_y$ as $\mathbf{w}=[\mathbf{w}_x,\mathbf{w}_y]^{T}\in\mathbb{R}^{2M}$. 


Recall that from \Cref{EncodingOcc}, each occupancy representation is encoded as a vector of similarities $\bm{\phi}$. Our goal is now to estimate $p(\mathbf{w}|\bm{\phi})$, the conditional distribution over trajectory parameters. There may exist many distinct groupings of trajectories in each environment. The distribution of trajectories is expected to be multi-modal, and we need to use a model capable of predicting multi-modal conditional distributions over the weights, $p(\mathbf{w}|\bm{\phi})$. Mixture density networks (MDN) \citep{Bishop94mixturedensity} are a class of neural networks capable of representing conditional distributions. We slightly modify the classical MDN described in \citep{Bishop94mixturedensity} to learn a mixture of vectors of conditional distributions, corresponding to the conditional distribution for each element in $\mathbf{w}$. We model the conditional distribution $p(\mathbf{w}|\bm{\phi})$ as a mixture of $Q$ vectors of distributions, which we call mixture components. To reduce the number of parameters to predict, we make the mean-field Gaussian assumption \citep{Bishop:2006} on the weights giving,
\begin{equation}
    p(\mathbf{w}|\bm{\phi})=\sum_{q=1}^{Q}\alpha_{q}p_{q}(\mathbf{w}|\bm{\phi})=\sum_{q=1}^{Q}\alpha_{q}\mathcal{N}(\mathbf{w}|\bm{\mu}_q,\bm{\Sigma}_q),
\end{equation}
where $p_{q}(\mathbf{w}|\bm{\phi})$ denotes the $q^{th}$ component, and $\alpha_q$ is the associated component weight. From the mean-field Gaussian assumption, each mixture component has mean vector, $\bm{\mu}_q=[\mu_{q,1},\mu_{q,2},\ldots,\mu_{q,2M}]^{T}$, and diagonal covariance, $\mathrm{diag}(\bm{\Sigma}_q)=\bm{\bm{\Sigma}}_{q}^{2}=[\bm{\Sigma}_{q,1}^{2},\bm{\Sigma}_{q,2}^{2},\ldots,\bm{\Sigma}_{q,2M}^{2}]^{T}$. We can write each component of the conditional distribution as:
    \begin{equation}\label{NormalAss}
        p_{q}(\mathbf{w}|\bm{\phi})=\prod_{m=1}^{2M}\frac{1}{\sqrt{2\pi\bm{\Sigma}_{q,m}^2}}\exp\Big\{-\frac{(w_{m}-\mu_{q,m})^{2}}{2\bm{\Sigma}_{q,m}^{2}}\Big\},
    \end{equation}
giving us the negative log-likelihood loss function over $N$ maps, and $P_n$ trajectories observed in the environment corresponding to the $n^{th}$ map in the dataset as:
\begin{equation}\label{lossF}
    \mathcal{L}(\bm{\theta})=-\log\Big[\prod^{N}_{n=1}\prod^{P_n}_{p=1}\sum^{Q}_{q=1}\alpha_{q}p_{q}(\mathbf{w}|\bm{\phi})\Big],
\end{equation}
where we denote the set of parameters to optimise as $\bm{\theta}=\{\alpha_q,\bm{\mu}_q,\bm{\Sigma}_q\}_{q=1}^{Q}.$
Using a neural network to minimise the loss function defined in \Cref{lossF}, we can learn a model that maps from the feature vector of similarities $\bm{\phi}$ to the parameters required to construct $p(\mathbf{w}|\bm{\phi})$. The neural network is relatively simple with a sequence of fully-connected layers.

The standard MDN constraints are applied using the activation functions highlighted in \citep{Bishop94mixturedensity}. This includes:
\begin{enumerate}[leftmargin=.6cm]
\item $\sum^{Q}_{q=1}\alpha_q=1$, such that component weights sum up to one, by applying the softmax activation function on associated network outputs;
\item $\bm{\Sigma}_{q,m}\geq0$, by applying an exponential activation function on associated network outputs.
\end{enumerate}
As a distribution is estimated for each of the elements in weight vector, $\mathbf{w}$, the predicted $p(\mathbf{w}|\bm{\phi})$ results in a mixture of discrete processes, where each realisation is a vector of $\mathbf{w}$. $\mathbf{k}$ denotes the basis functions.


\subsection{Trajectory Generation and Conditioning}
After we complete the training of our MDN model, we can generate trajectories in environments with no observed trajectories. We generate the feature vector of similarities, $\bm{\phi}^*$, from the map of interest, $\mathcal{M}^{*}$, and into the MDN. We obtain parameters, that define conditional distributions over $\mathbf{w}$. Realisations of $\mathbf{w}$ can be sampled randomly from the predicted $p(\mathbf{w}|\bm{\phi}^{*})$, and a possible continuous trajectory, $\bm{\Xi}$, can be found by evaluating
\begin{equation}
\bm{\Xi}(\tau)=[x(\tau),y(\tau)]^{\top}=[\mathbf{w}_x\mathbf{k}(\tau), \mathbf{w}_y\mathbf{k}(\tau)]^{\top}, 
\end{equation}
where $\mathbf{k}(\tau)$ gives evaluations of the basis functions for the continuous trajectories. As there are no explicit constraints in the MDN to prevent the generation of trajectories which overlap with occupied regions, we apply collision checking. The trajectories can be generated and checked very efficiently, as it involves randomly sampling a mixture of Gaussian distributions and checking a map. 

In order to predict how agents observed at a known position move, we are often interested in generating likely trajectories which begin at a certain start-point. To achieve this, let us consider the distribution of trajectories $x,y$-coordinates. Intuitively, rather than the distribution on weight parameters $\mathbf{w}$, we want to consider the joint distribution between trajectory coordinates. We evaluate the continuous trajectories at a set of $L$ times of interest, $\bm{\tau}=[\tau_1,\tau_2\ldots\tau_{L}]$, 
\begin{align}
p(\bm{\Xi}(\!\bm{\tau})\!)
&\!=\!
\!p\Big(\begin{bmatrix}
\mathbf{w}^{T}_{x}\mathbf{K}(\bm{\tau})\\
\mathbf{w}^{T}_{y}\mathbf{K}(\bm{\tau})
\end{bmatrix}\Big)
\!=\!
\sum_{q=1}^{Q}\!\alpha_{q}\!\begin{bmatrix}\mathcal{N}(\hat{\bm{\mu}}^{x}_{q},\hat{\bm{\Sigma}}^{x}_{q})\\\mathcal{N}(\hat{\bm{\mu}}^{y}_{q},\hat{\bm{\Sigma}}^{y}_{q})\end{bmatrix},
\end{align}
where $\mathbf{K}(\bm{\tau})=[\mathbf{k}(\tau_1),\mathbf{k}(\tau_2),\dots,\mathbf{k}(\tau_L)]^{T}$ is a matrix containing basis function evaluations at $L$ time points of interest $\bm{\tau}$. We can now find mean and covariance parameters ($\hat{\bm{\mu}}^{x}_{q}, \hat{\bm{\mu}}^{y}_{q},\hat{\bm{\Sigma}}^{x}_{q},\hat{\bm{\Sigma}}^{y}_{q}$) for the trajectory coordinates at times of interest as,
\begin{align}
\hat{\bm{\mu}}^{x}_{q}={\bm{\mu}_{q}^{x}}^{T}\mathbf{K}(\bm{\tau}), && \hat{\bm{\Sigma}}^{x}_{q}=\mathbf{K}(\bm{\tau}){\bm{\Sigma}_{q}^{x}}\mathbf{K}(\bm{\tau})^{T},\\ \hat{\bm{\mu}}^{y}_{q}={\bm{\mu}_{q}^{y}}^{T}\mathbf{K}(\bm{\tau}), && \hat{\bm{\Sigma}}^{y}_{q}=\mathbf{K}(\bm{\tau}){\bm{\Sigma}_{q}^{y}}\mathbf{K}(\bm{\tau})^{T},
\end{align}
The covariances of the $q^{th}$ component between the elements in weight parameter vectors $\mathbf{w}_x$ and $\mathbf{w}_y$ are denoted as $\bm{\Sigma}^{x}_{q}$ and $\bm{\Sigma}^{y}_{q}$ respectively. The component covariance $\bm{\Sigma}_{q}$ between all the weights $\mathbf{w}=[\mathbf{w}_{x},\mathbf{w}_{y}]$ and $\bm{\Sigma}_{q}^{x}$, $\bm{\Sigma}_{q}^{y}$ are $\mathrm{diag}(\bm{\Sigma}_{q})=[\mathrm{diag}(\bm{\Sigma}_{q}^{x}),\mathrm{diag}(\bm{\Sigma}_{q}^{y})]$. Note that although $\bm{\Sigma}_q$ is diagonal, the covariance of Gaussian components between times of interest $\hat{\bm{\Sigma}_{q}^{x}},\hat{\bm{\Sigma}_{q}^{y}}$ are typically fully-connect. 

For any component, we can then enforce locations that the trajectories must pass through $(x^{*},y^{*})$ at $\tau_{i}$, by finding the distributions of coordinates the times of interest conditioned on the fixed point. Consider when we wish to condition on the start point, i.e. $\tau_{1}=(x^{*},y^{*})$, a single component in the mixture can be written as:
\begin{align}
    p_{q}(\bm{\Xi}(\bm{\tau}_{1:L})|\bm{\Xi}(\bm{\tau}_{1})=[x^{*},y^{*}])=\alpha_{q}\begin{bmatrix}\mathcal{N}(\bar{\bm{\mu}}^{x},\bar{\bm{\Sigma}}^{x})\\\mathcal{N}(\bar{\bm{\mu}}^{y},\bar{\bm{\Sigma}}^{y})\end{bmatrix}.
\end{align}
If we consider elements in mean vectors $\hat{\bm{\mu}}^{x}$, $\hat{\bm{\mu}}^{y}$ and covariance matrices $\hat{\bm{\Sigma}^{x}}$, $\hat{\bm{\Sigma}^{y}}$ of the unconditioned distribution as:
\begin{align}
    \hat{\bm{\mu}}^{z}\!=\!\begin{bmatrix}\hat{\bm{\mu}}_{1}^{z} \\ \hat{\bm{\mu}}_{2:L}^{z}\end{bmatrix},&&
    \!\hat{\bm{\Sigma}}^{z}\!=\!\begin{bmatrix}
    \hat{\bm{\Sigma}}^{z}_{1,1} & {\hat{\bm{\Sigma}}^{z}_{1,2:L}} {}^{T}\\
    {\hat{\bm{\Sigma}}^{z}_{1,2:L}} & {\hat{\bm{\Sigma}}^{z}_{2:L,2:L}}
    \end{bmatrix}, &&\! \text{for } z\!=\!\{\!x,y\!\},
\end{align}
then for $z=\{x,y\}$, we can express the parameters of the conditional distribution as \citep{Bishop:2006}:
\begin{align}
    \bm{\bar{\mu}^{z}}&=\hat{\bm{\mu}}_{2:L}^{z}+\hat{\bm{\Sigma}}_{1,2:L}^{z}{\hat{\bm{\Sigma}}_{1,1}^{z}} {}^{-1}(z^{*}-\bm{\hat{\mu}}_{1})\\
    \bar{\bm{\Sigma}}^{z}&=\hat{\bm{\Sigma}}_{2:L,2:L}^{z}-\hat{\bm{\Sigma}}_{1,2:L}^{z}{\hat{\bm{\Sigma}}_{1,1}^{z}} {}^{-1}{\hat{\bm{\Sigma}}_{1,2:L}^{z}} {}^{T}.
\end{align}

We can condition each mixture component to start at a certain point, or employ strategies to only condition selected mixture components, such as the nearest component to the conditioned coordinate, and only generate trajectories belonging to the conditioned mixture component.
\section{Experimental Results and Discussion}


\subsection{Dataset and Metrics}
Training an OTNet requires a dataset containing occupancy maps of multiple environments along with observed trajectories in each environment. To the best of our knowledge, there exists no real-world dataset of sufficient size with different occupancy maps and trajectories observed in each of the different environments. Therefore, we conduct our experiments with our simulated dataset, \textit{Occ-Traj120} \citep{occtraj}. This dataset contains 120 binary occupancy grid maps of indoor environments with rooms and corridors, as well as simulated motion trajectories. We aim to utilise our method to learn transfer the motion to new environments. The dataset is split with a 80-20 ratio for training and testing respectively. Our aim is to transfer the trajectory behaviour, from a training subset to unseen test environments. 
\begin{figure}[H]
\centering
\begin{subfigure}{.3\textwidth}
  \centering
  \includegraphics[width=\linewidth]{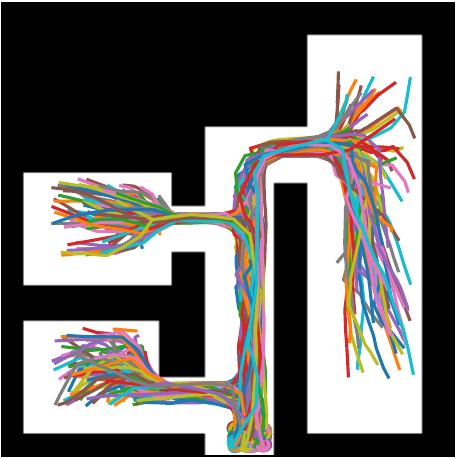}  
\end{subfigure}
\begin{subfigure}{.3\textwidth}
  \centering
  \includegraphics[width=\linewidth]{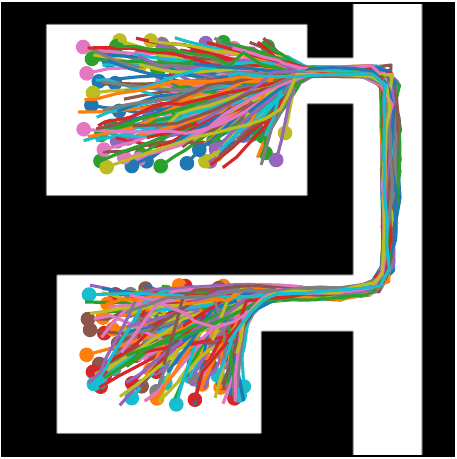}  
\end{subfigure}
\begin{subfigure}{.3\textwidth}
  \centering
  \includegraphics[width=\linewidth]{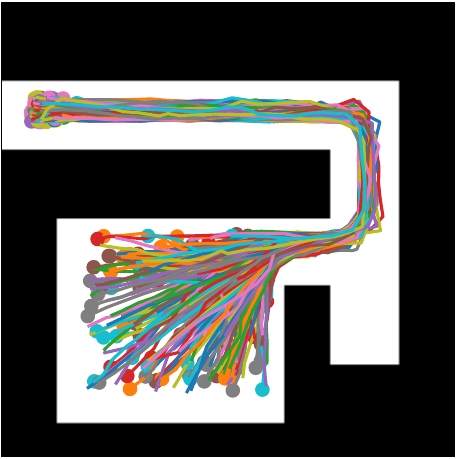}  
\end{subfigure}
\caption{Occupancy maps in the Occ-Traj 120 dataset, along with associated trajectories. We aim to generalise trajectories to unseen maps.}\label{datasetfig}
\end{figure}

We evaluate the generated trajectories against a test set with hidden ground truth trajectories. Continuous trajectories outputted are discretised for evaluation by querying at uniform intervals. Due to the probabilistic and multi-modal nature of our output, the metric used is \emph{minimum trajectory distance} (MTD), and is defined by:
${MTD = \min_{i=1,\ldots, P}{\mathcal{D}(\xi_{gen},\xi_{i})}}$,
where $P$ denotes the number of trajectories observed in the environment, with $i$ indexing each trajectory, and $\mathcal{D}(\xi_{gen},\xi_{i})$ is a distance measure of trajectory distance between the generated $\xi_{gen}$ and a ground truth trajectory $\xi_i$. In our evaluations the Euclidean \emph{Hausdoff distance} and \emph{discrete Frech\'et distance} are considered. These trajectory distances are commonly used in distance-based trajectory clustering to quantify the dissimilarity between trajectories, and a review of these distances can be found in \citep{traj_dist}. We also wish to evaluate the quality of the uncertainty captured. To this end, we calculate the average negative log-likelihood (ANLL) between the predicted distribution at 100 uniform time-steps and each ground truth trajectory. When only samples of generated trajectories are available, we fit Gaussian distributions over world-space coordinates at uniform time-coordinates. ANLL can take into account of probabilistic multi-modal distributions, and a relative lower ANLL indicates better performance. 


\subsection{Experimental Setup}
\begin{figure}[t]
\centering
\begin{subfigure}{.3\textwidth}
  \centering
  \includegraphics[width=\linewidth]{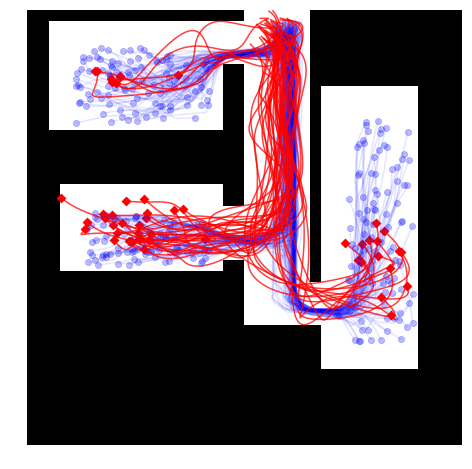}  
\end{subfigure}
\begin{subfigure}{.3\textwidth}
  \centering
  \includegraphics[width=\linewidth]{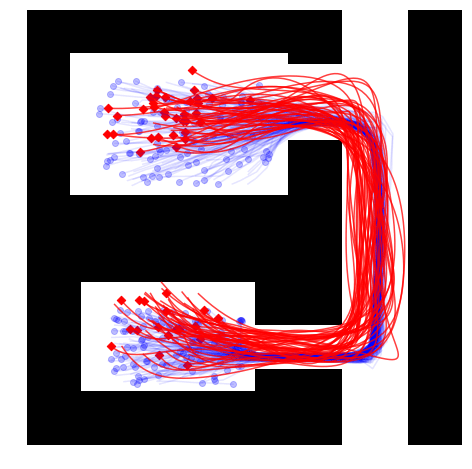}  
\end{subfigure}
\begin{subfigure}{.3\textwidth}
  \centering
  \includegraphics[width=\linewidth]{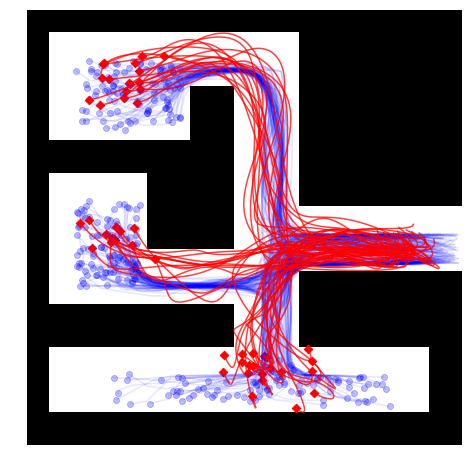}  
\end{subfigure}
\caption{Generated likely motion trajectories (in red) in new environments. End points are indicated by scatter points. OTNet captures the probabilistic, multi-modal nature of motion trajectories. The ground truth trajectories (in blue) are hidden during training.}
\label{fig:Pred}
\end{figure}

We compare our proposed method, OTNet, with neural-network based generative models, a learning-based motion reconstruction method, and a k-nearest neighbour (k-NN) based method. The details of these models are as follows:

\subsubsection{OTNet} We train OTNet with the length-scale hyper-parameters $\ell_{H}=50$ and $\ell_{b}=5$ for 20 epochs. The number of bases are set to be $M=15$, and the number of mixture components, $Q=4$. These hyper-parameters are hand-picked and fairly not sensitive, with values in the same ball-park giving similar results. Cross-validation could be applied for further improvements. As the number of map examples in the training set is relatively small at 96, we select all our training maps to be in our database of maps, when encoding maps of interest as feature vectors. In each of our experiments, trajectories can be generated efficient during test time. On a standard desktop machine, predicting the distribution of trajectories can be done in under one second, negligible time is required to randomly sample the distribution to generate trajectories. 
\subsubsection{Generative network models}
Our proposed method is generative, we evaluate two popular generative models: GANs \citep{Gans} CVAEs \citep{CVAE}, trained for 300 epochs to generate trajectory parameters $\mathbf{w}$. The discriminator uses convolutional layers, and the generator samples a 100-dimensional latent vector, concatenated with the map for conditioning. Five fully-connect layers are used to output $\mathbf{w}$. The hyperparameters of the CVAE model, are identical to the GAN model.


\subsubsection{Learning-based motion reconstruction}
We evaluate the performance of the learning-based MPNet~\citep{qureshi2019motion}, which conditions on the environment and imitates training trajectories. We pre-train on MPNet's public 2D dataset with further training with the \emph{Occ-Traj120 dataset} for 300 epochs. MPNet requires trajectory start and end coordinates, during evaluation we provide ground truth mean start and end points of groups of trajectories. To guarantee valid trajectories, we use the hybrid MPNet-RRT variant suggested by the authors. We emphasise that MPNet can only generate a single prediction, and cannot generate distributions of trajectories.


\subsubsection{Nearest-neighbour}
We also consider a baseline k-NN based approach. Using the symmetric Hausdoff distance, we find the nearest map training map, and transfer the trajectories directly to the queried map. As we cannot average trajectories, we use $k=1$. Note storage of not only all maps, but also trajectories are required.


\subsection{Comparisons to Other Generative Models}

\begin{table}[h]
\centering
\begin{tabular}{|l|lll|}
\hline
                     & Hausdoff & Frechet & ANLL    \\ \hline
OTNet  & 1.98     & 2.13    &   29.83\\
GANs        & 11.79   & 16.66   & NA \\ 
CVAE        & 9.48 & 14.67 & NA \\
MPNet-RRT   & 2.99 & 5.14  &   NA\\
k-NN        & 2.03 & 2.14  &  31.46\\\hline
\end{tabular}
\caption{Performance of OTNet and comparisons models. Lower values are better. OTNet outperforms the comparisons, although the performance of k-NNs are strong, k-NNs lack the ability to generate new trajectories that do not exist in the provided dataset.}\label{NormalvLaplace}
\end{table}

The performance results of our experiments are tabulated in \Cref{NormalvLaplace}, we see that OTNet outperform the other generative models compared. The ANLL for CVAE and GANs suffer from a loss of precision due to an extremely low likelihood, resulting in taking log of a near 0 value, the ANLL will be much higher than OTNet. The ANLL for MPNet-RRT is also not applicable, as it generates single trajectories. The map representations are of too high dimensions to directly train neural networks with a limited dataset. In particular, the encoding of each occupancy map as a feature vector of similarities, $\bm{\phi}$, allows for flexible representations even when the number of maps in the dataset is small. Comparatively, other neural network-based models which attempt to directly extract map features using convolutional neural networks are unable to successfully condition on the occupancy maps. The nearest neighbour model, which transfers the trajectories of the nearest map, performs surprising well, comparable to the OTNet on the simulated on the Haussdoff and Frech\'{e}t measures. This is due to OTNet often containing several relatively similar-looking maps. However, OTNet is able to take into consideration several maps, and weigh each by their similarity with the map of interest, while the nearest neighbour only considers the closest map. We also note that the single nearest neighbour approach often results in relative over-confidence of biased predictions, hence the higher ANLL relative to OTNet. We note that we can directly obtain closed form equations of trajectory distributions from OTNet, allowing us to repeatedly generate new trajectories. Whereas, MPNet-RRT produces single predictions. Although using k-NN can give trajectories with competitive performance, it cannot generate new trajectories, whereas OTNet is capable of generating trajectories that have not been observed in the original dataset. Examples of trajectories generated by OTNet are shown in red in \Cref{fig:Pred}, along with ground truth trajectories (blue).   

\begin{figure*}[!t]
\centering
\begin{subfigure}{.15\linewidth}
    \centering
    \includegraphics[width=\linewidth,trim=6em 6em 6em 6em, clip]{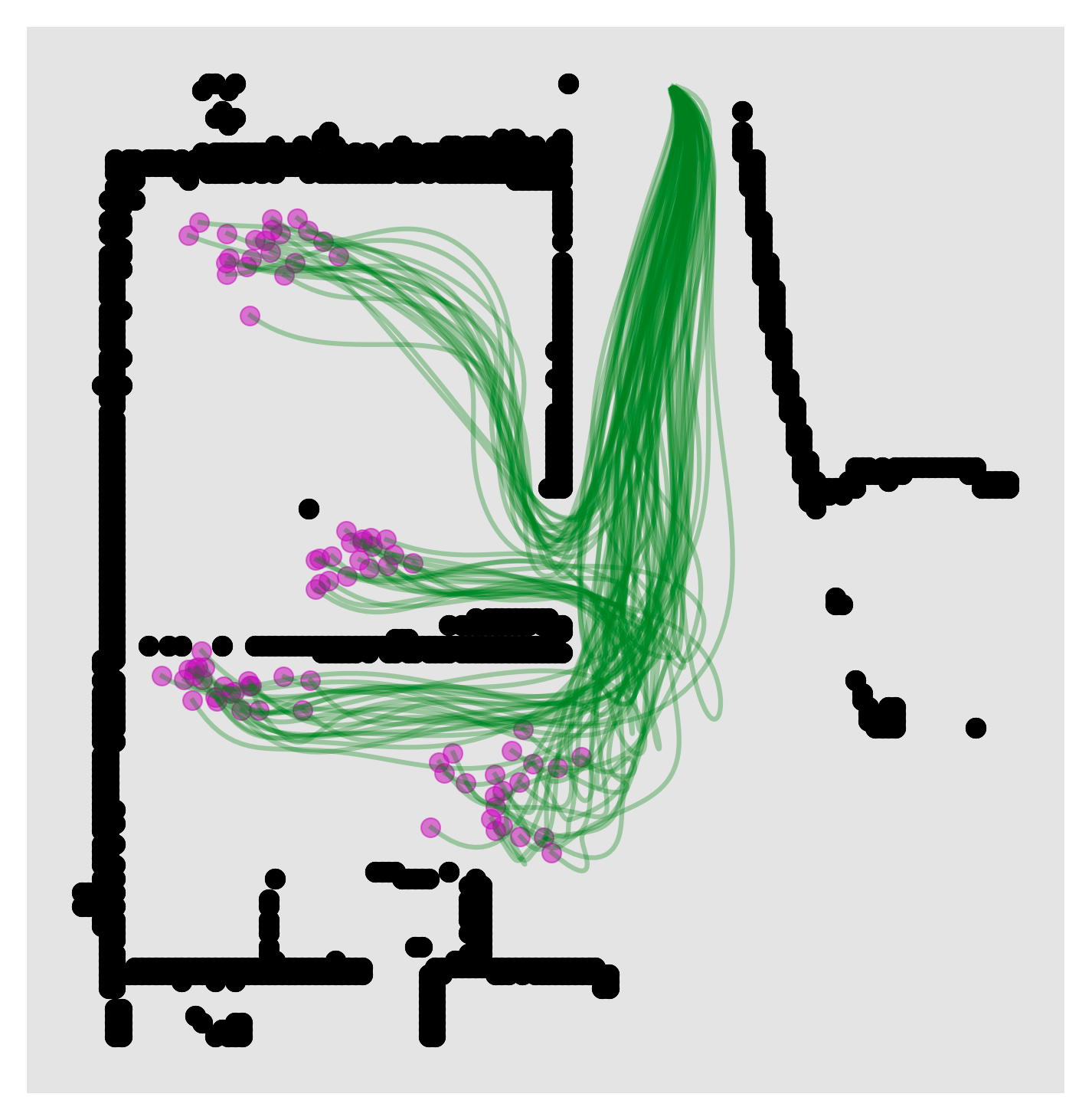}
    \includegraphics[width=\linewidth,trim=6em 6em 6em 6em, clip]{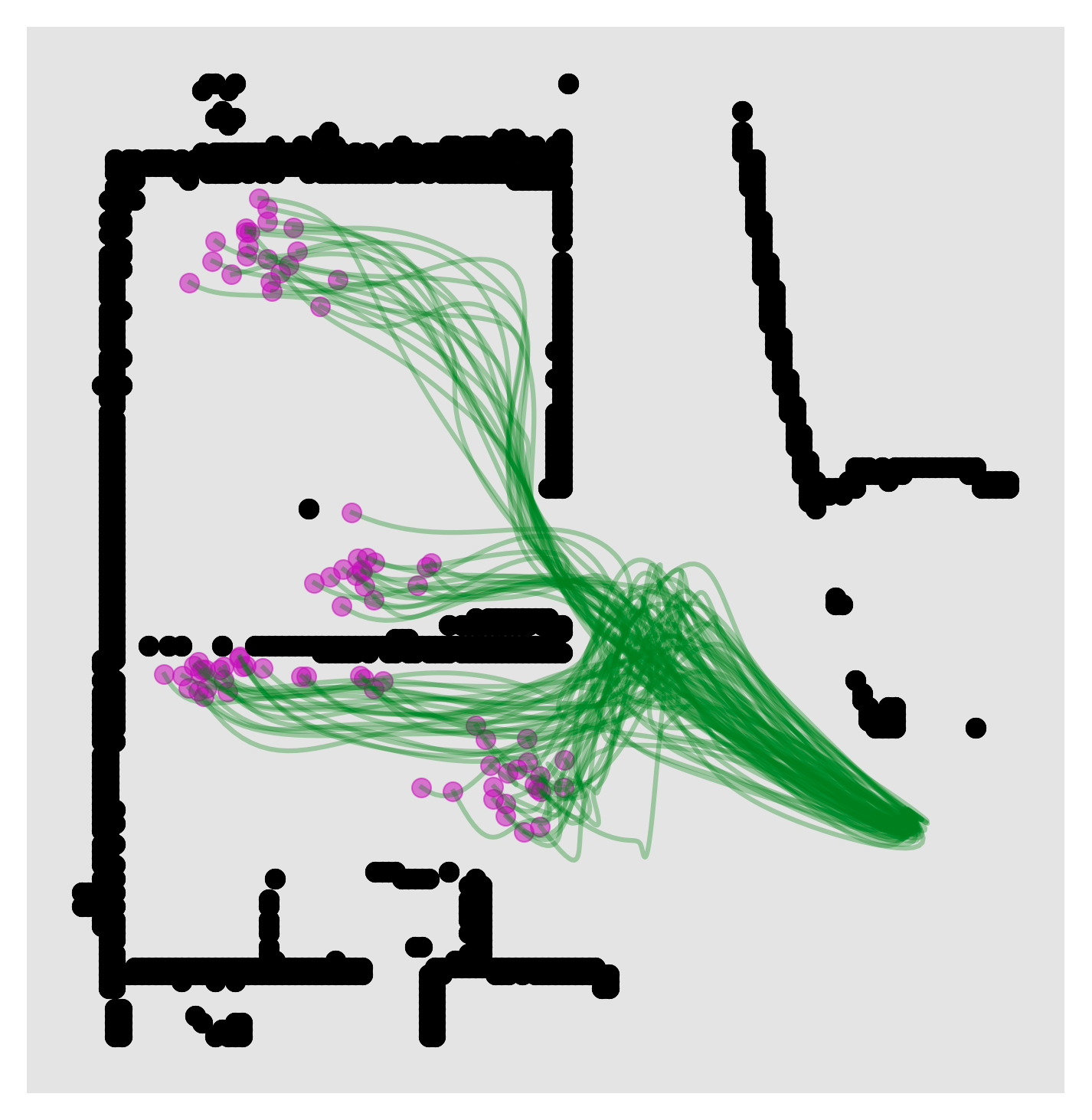}
    \includegraphics[width=\linewidth,trim=6em 6em 6em 6em, clip]{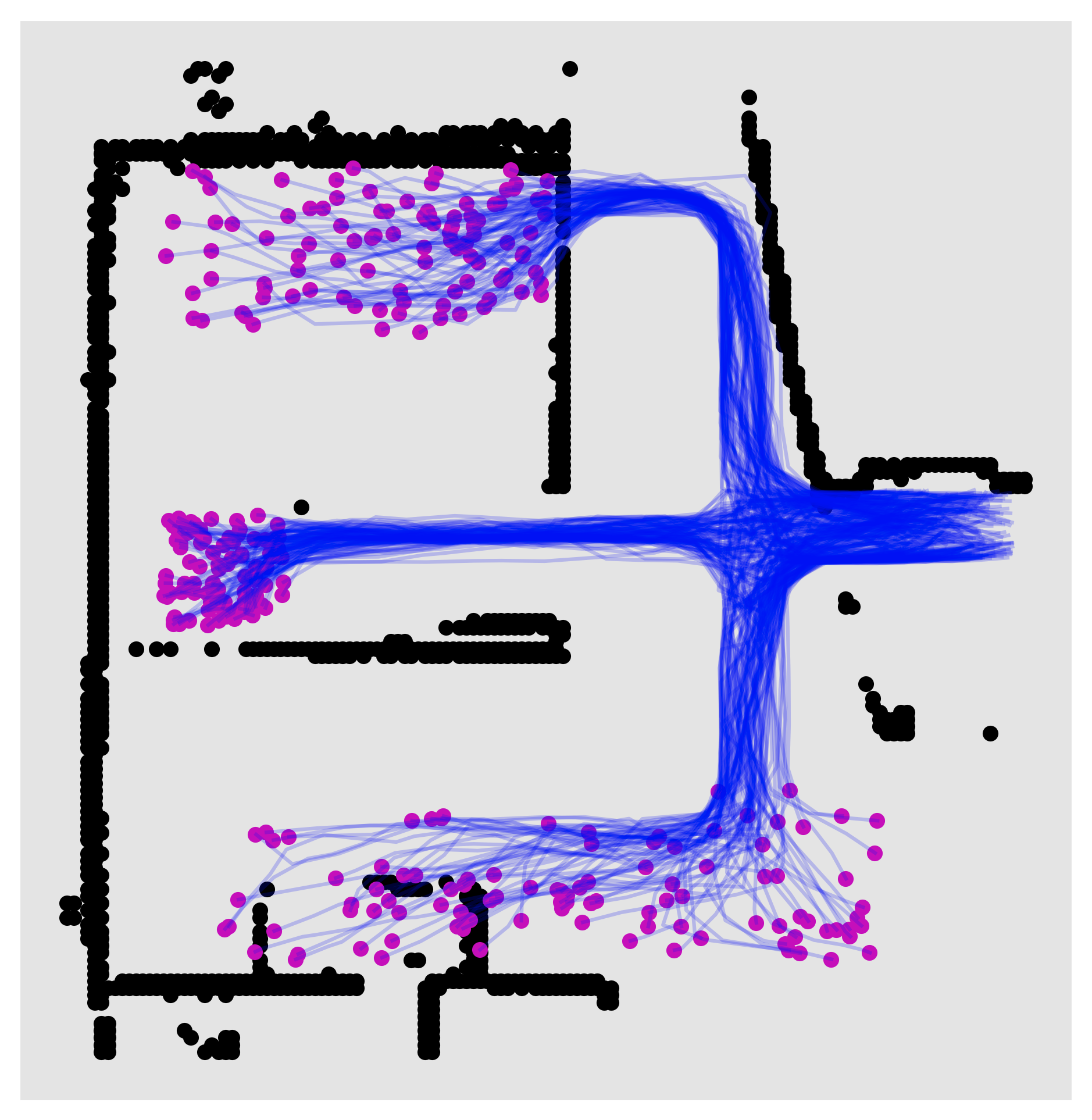}
\end{subfigure}
\begin{subfigure}{.15\linewidth}
\centering
    \includegraphics[width=\linewidth,trim=6em 6em 6em 6em, clip]{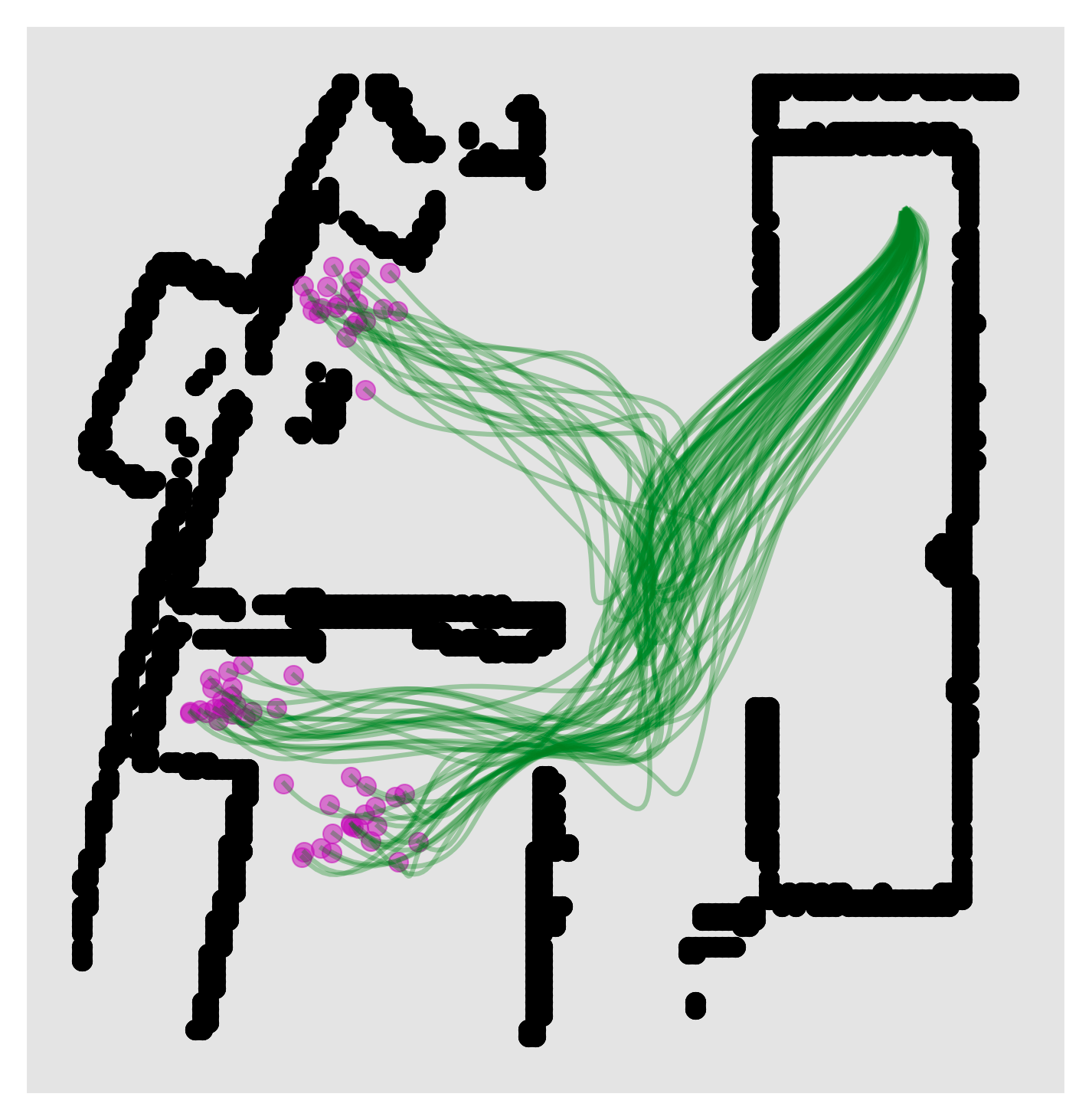}
    \includegraphics[width=\linewidth,trim=6em 6em 6em 6em, clip]{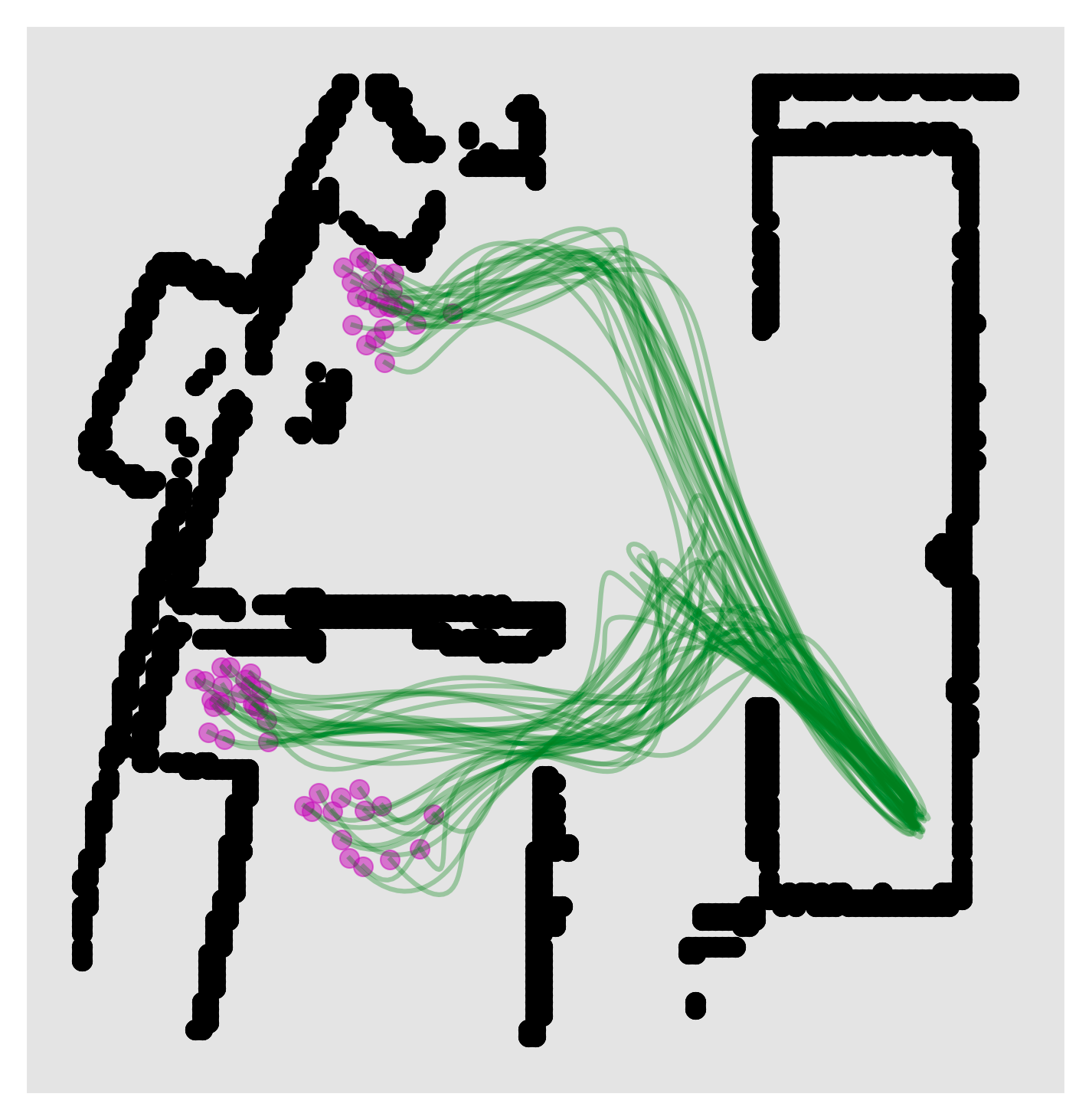}
    \includegraphics[width=\linewidth,trim=6em 6em 6em 6em, clip]{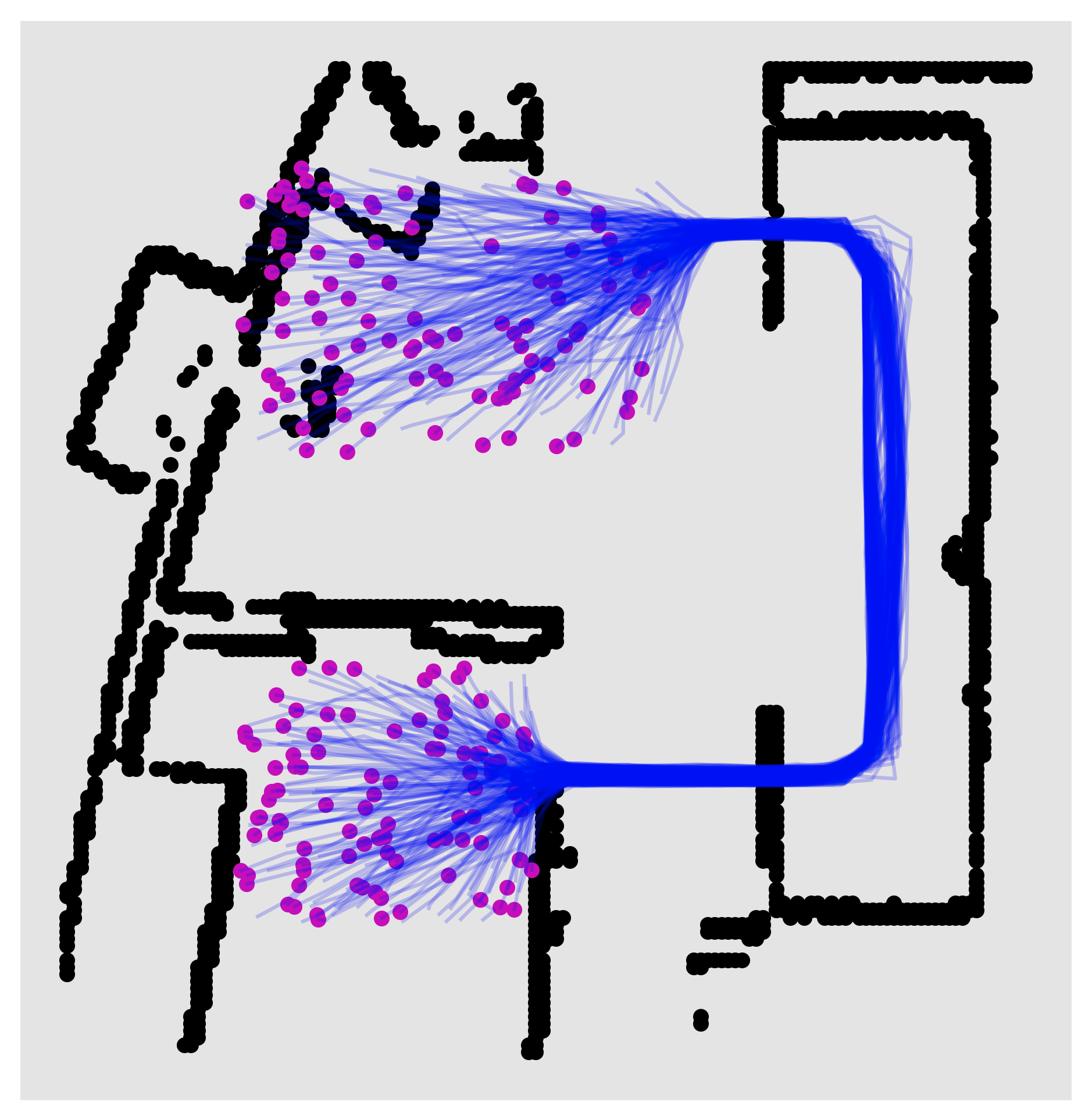}
\end{subfigure}
\begin{subfigure}{.15\linewidth}
\centering
    \includegraphics[width=\linewidth,trim=6em 6em 6em 6em, clip]{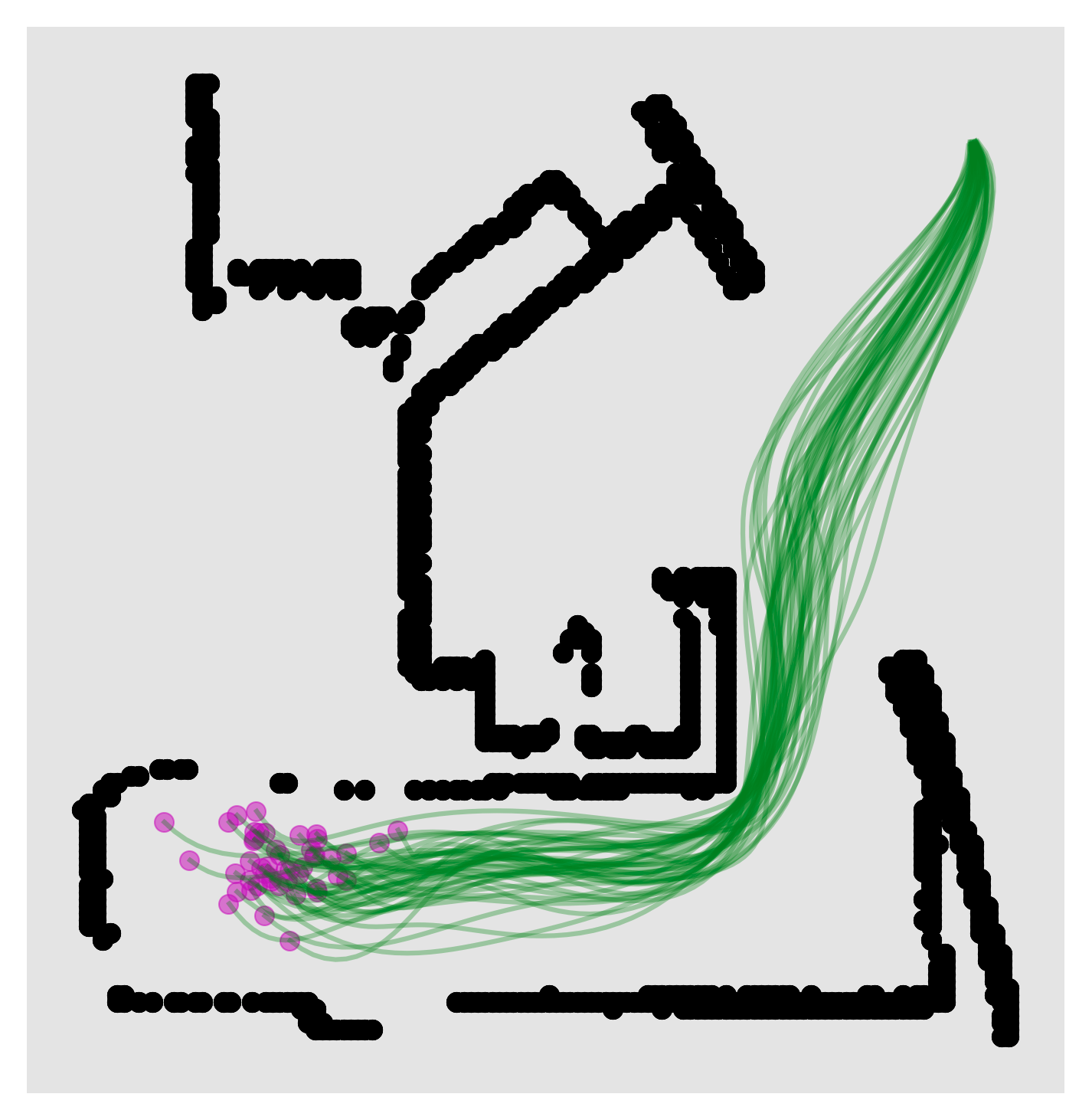}
    \includegraphics[width=\linewidth,trim=6em 6em 6em 6em, clip]{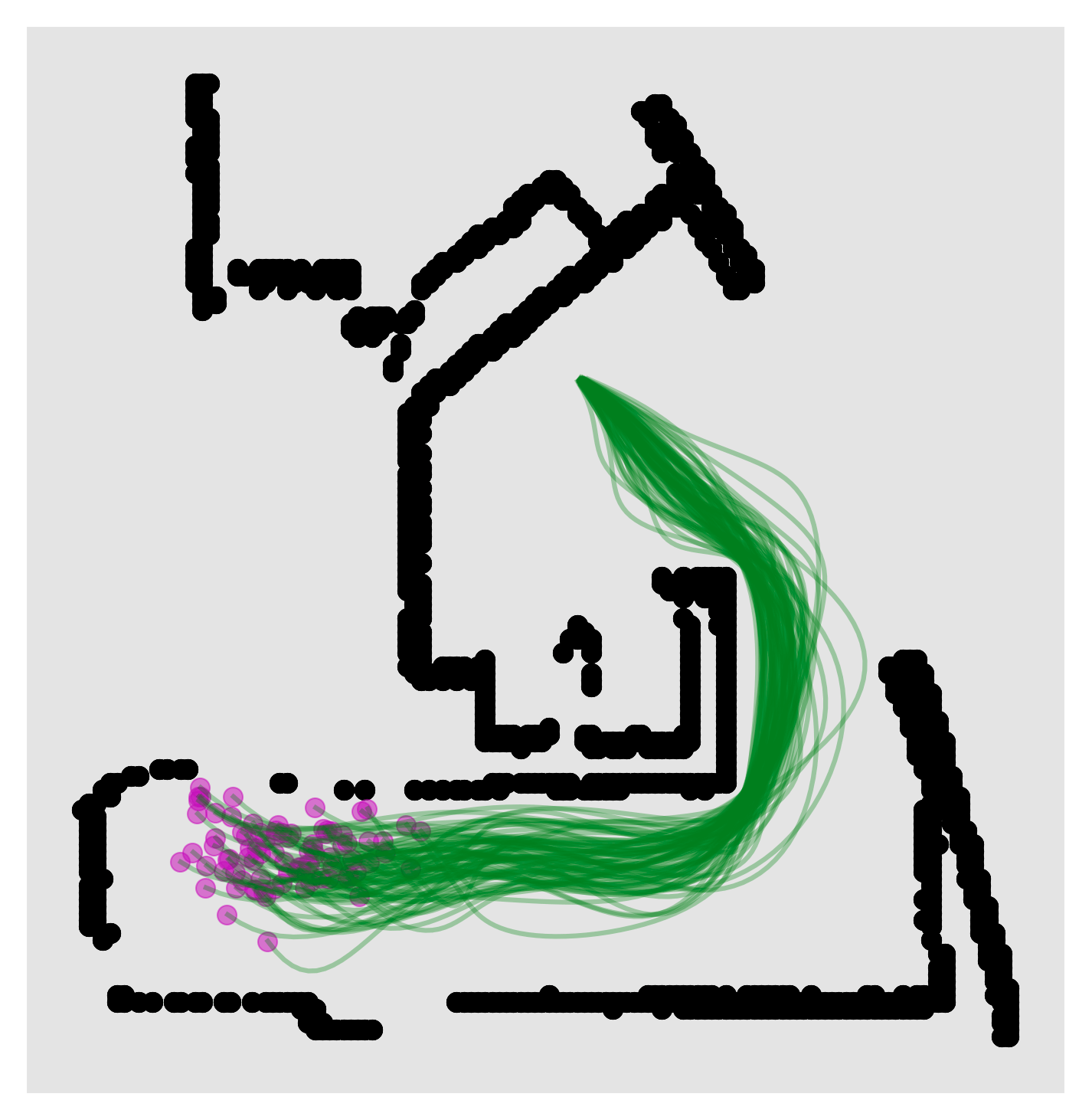}
    \includegraphics[width=\linewidth,trim=6em 6em 6em 6em, clip]{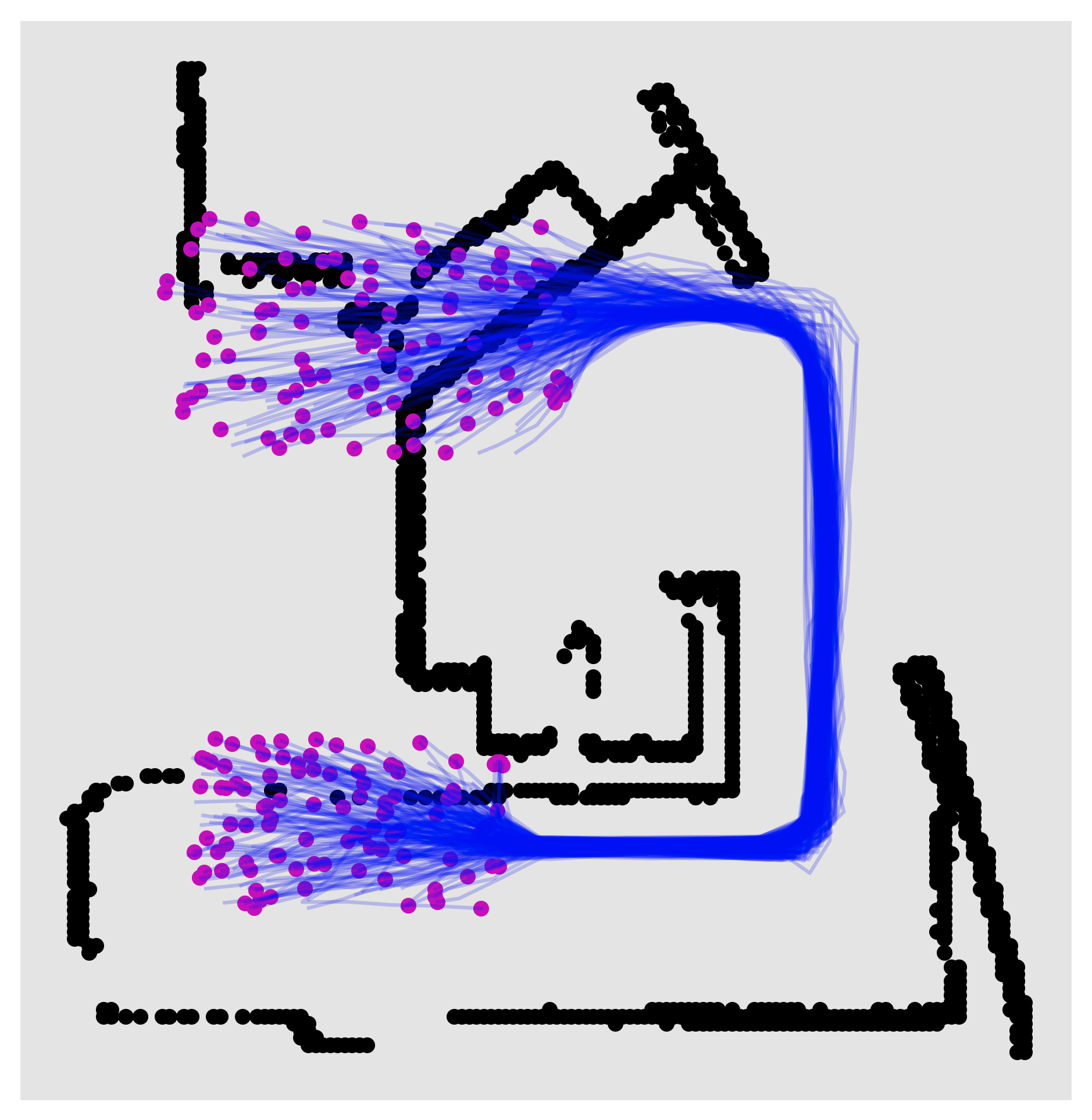}
\end{subfigure}
\begin{subfigure}{.48\linewidth}
\includegraphics[width=\linewidth]{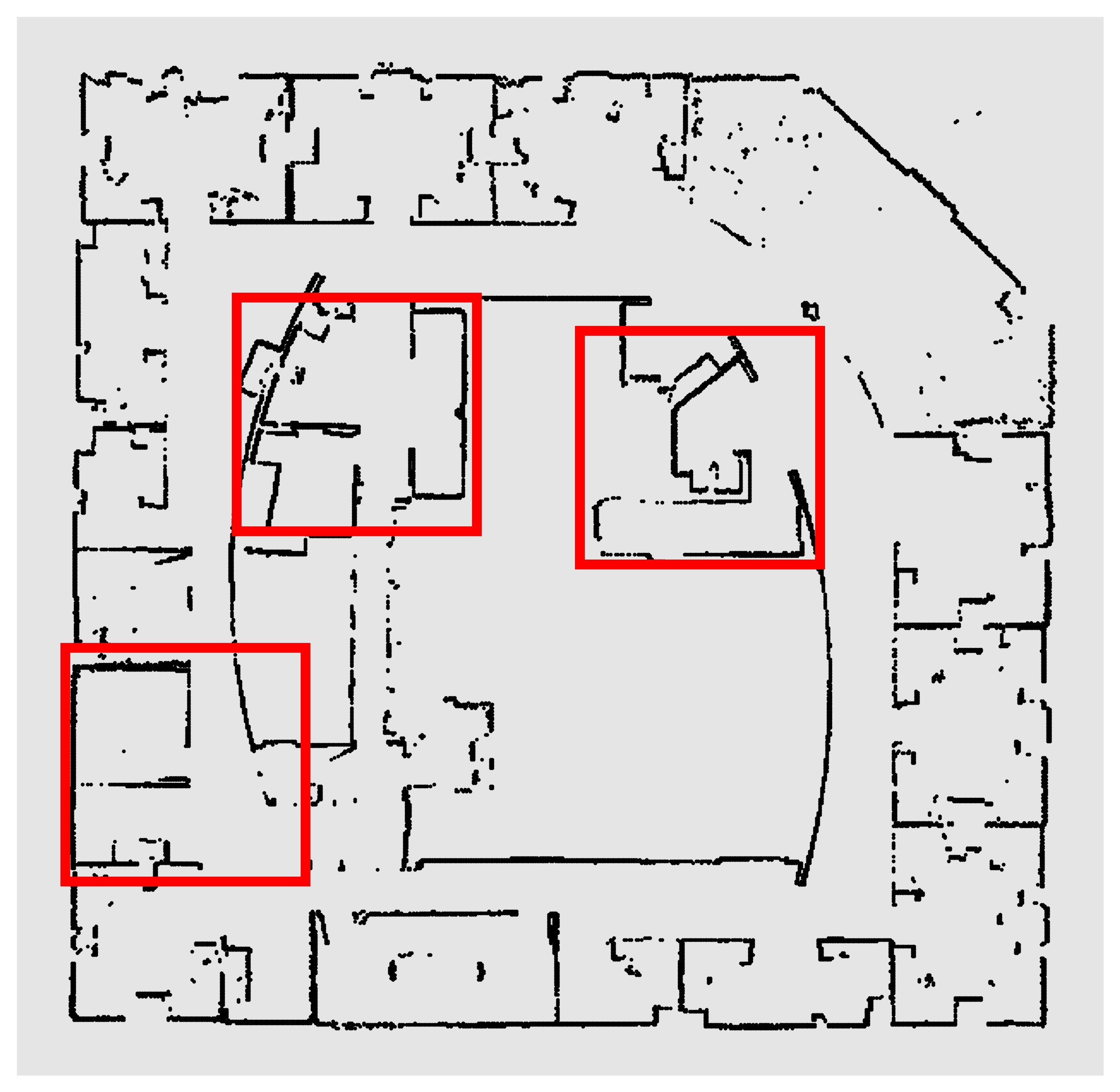}
\end{subfigure}
\caption{After training on a simulated dataset, we condition on sections of real world occupancy data, from the Intel-Labs dataset \citep{Intel}, to generate motion trajectories (top two rows). The start point of the trajectories are fixed, with the end-points illustrated with magenta markers. We can compare this to trajectories from a baseline nearest neighbour model (bottom row), which transfers trajectories from the most similar training map, and is unable to adapt to the new environments. }\label{TransferEx}
\end{figure*}

%
\begin{figure}[h]
    \centering
    \begin{subfigure}{.45\linewidth}
        \centering
        \vspace{1em}
        \includegraphics[width=\linewidth]{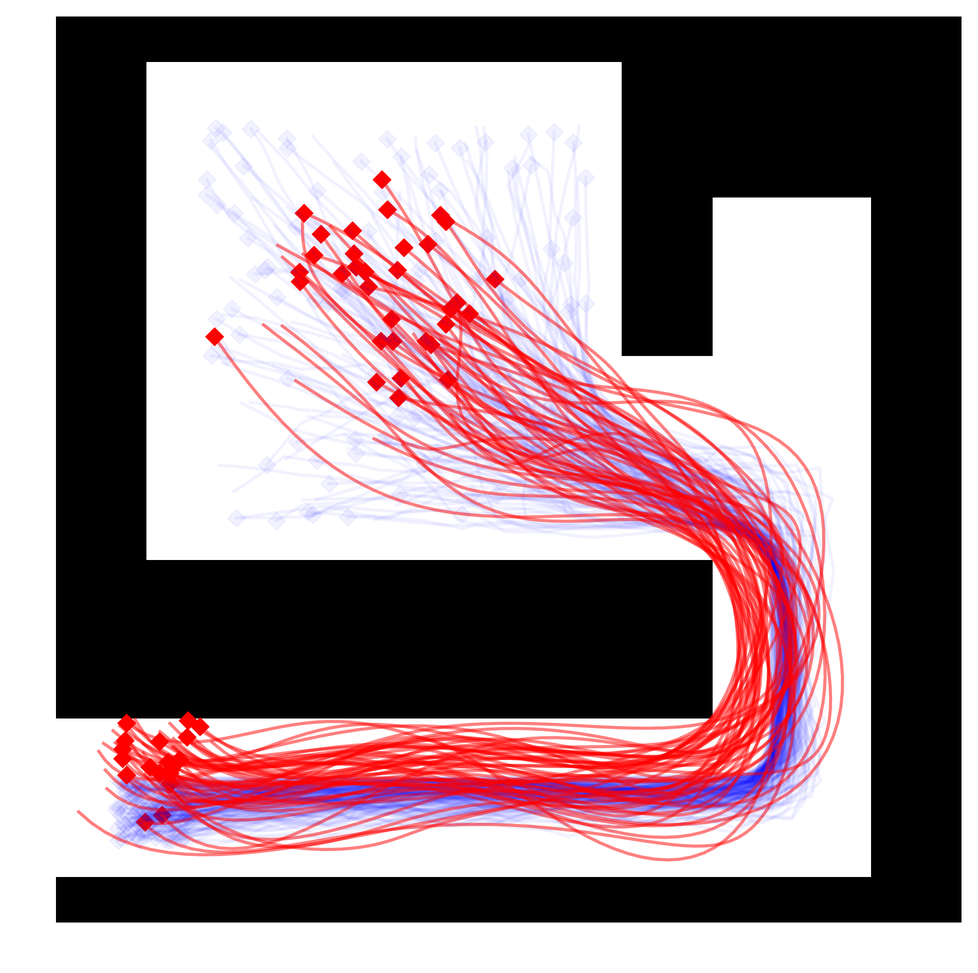}  
    \end{subfigure}
    \hspace{0.5em}
    \begin{subfigure}[t]{.45\linewidth}
        \centering
        \includegraphics[width=\linewidth]{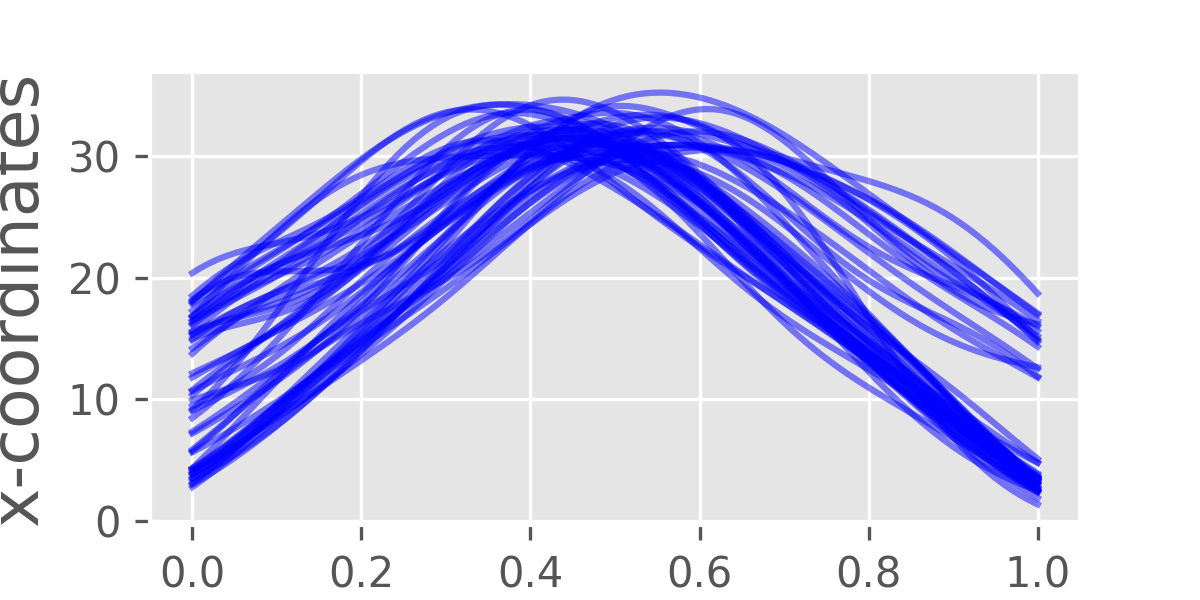} 
        \includegraphics[width=\linewidth]{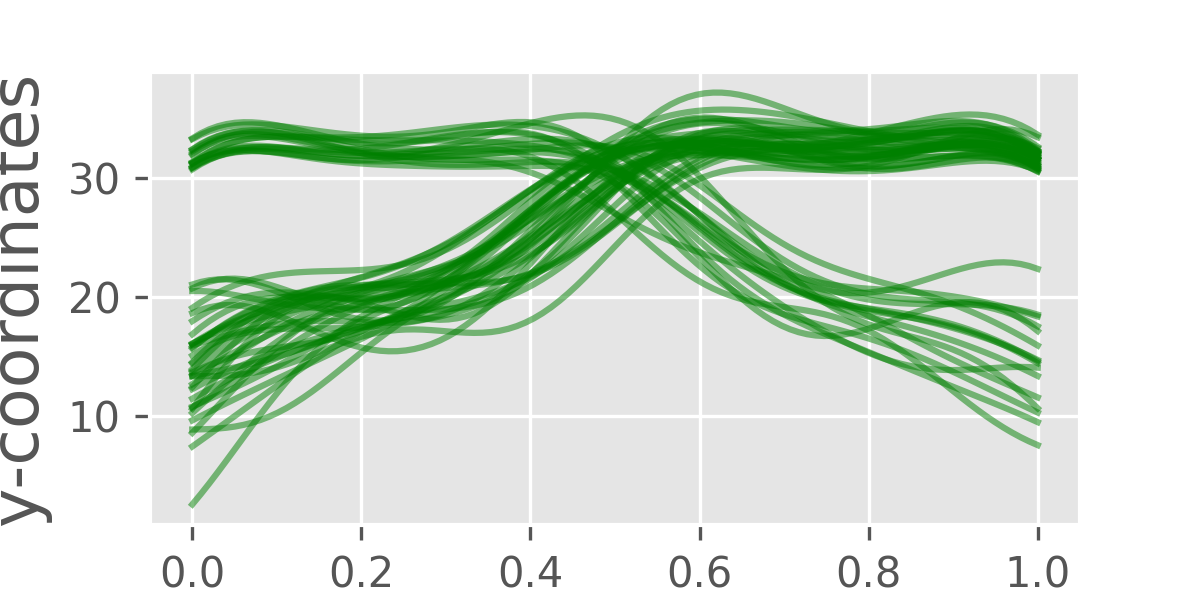}  
    \end{subfigure}
    \caption{(Left) Generated trajectories (red) overlaid on the corridor and room test map, with markers indicating the endpoints. The hidden ground truth trajectories are under-laid in blue. We see two different groups of trajectories: one starts from inside the room and end in the corridor, the other moves in the opposite direction. (Right) Plots of coordinates wrt $\tau$, $x(\tau)$ and $y(\tau)$  (top, bottom respectively), corresponding to the trajectories generated on the left plot. The ability to generate distinct groups of trajectories is clear.} \label{functionswithT}
\end{figure}

\begin{figure}[h]
    
    \begin{subfigure}[t]{.32\linewidth}
        \centering
        \includegraphics[width=0.95\linewidth]{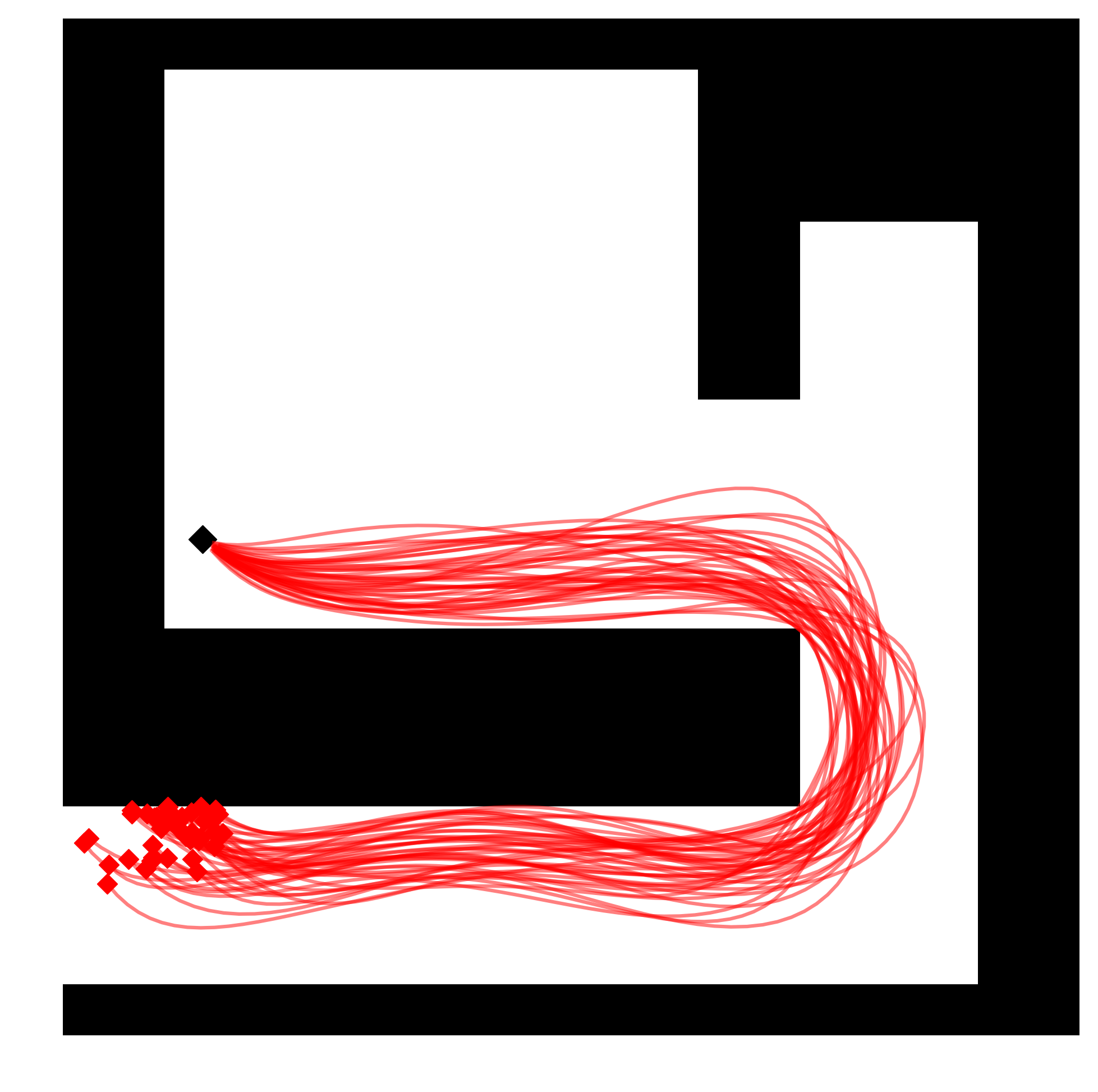} 
        \includegraphics[width=\linewidth]{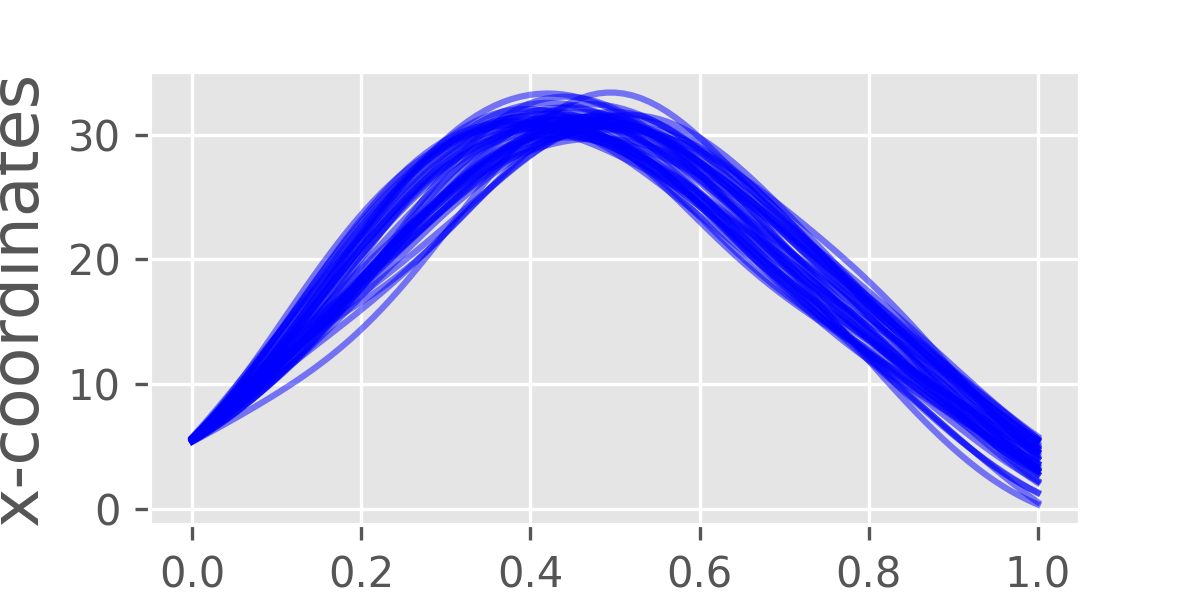}  
        \includegraphics[width=\linewidth]{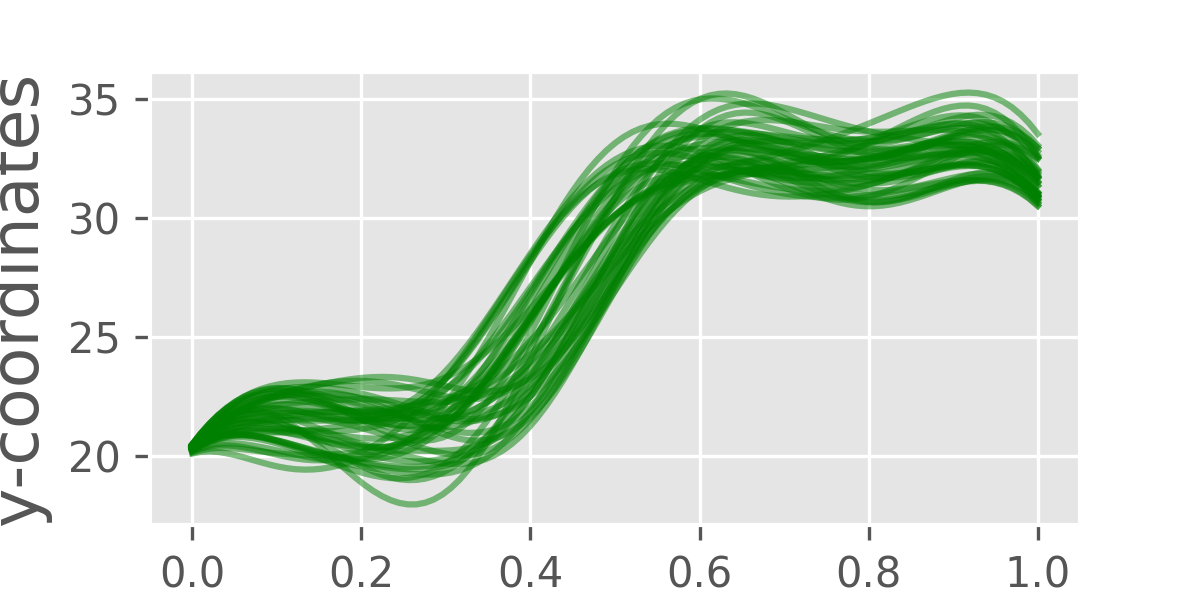}  
    \end{subfigure}
    \begin{subfigure}[t]{.32\linewidth}
        \centering
        \includegraphics[width=0.95\linewidth]{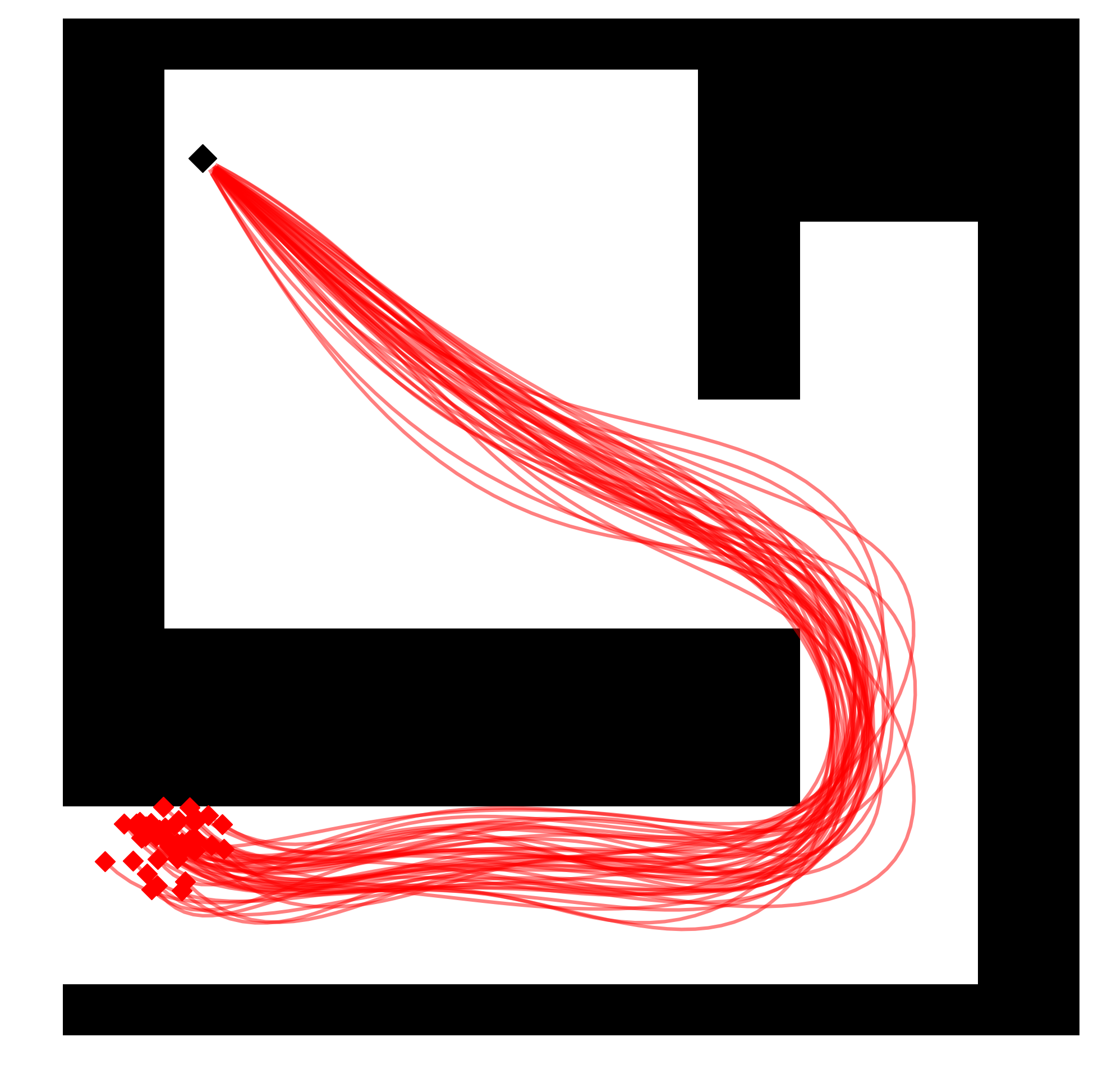} 
        \includegraphics[width=\linewidth]{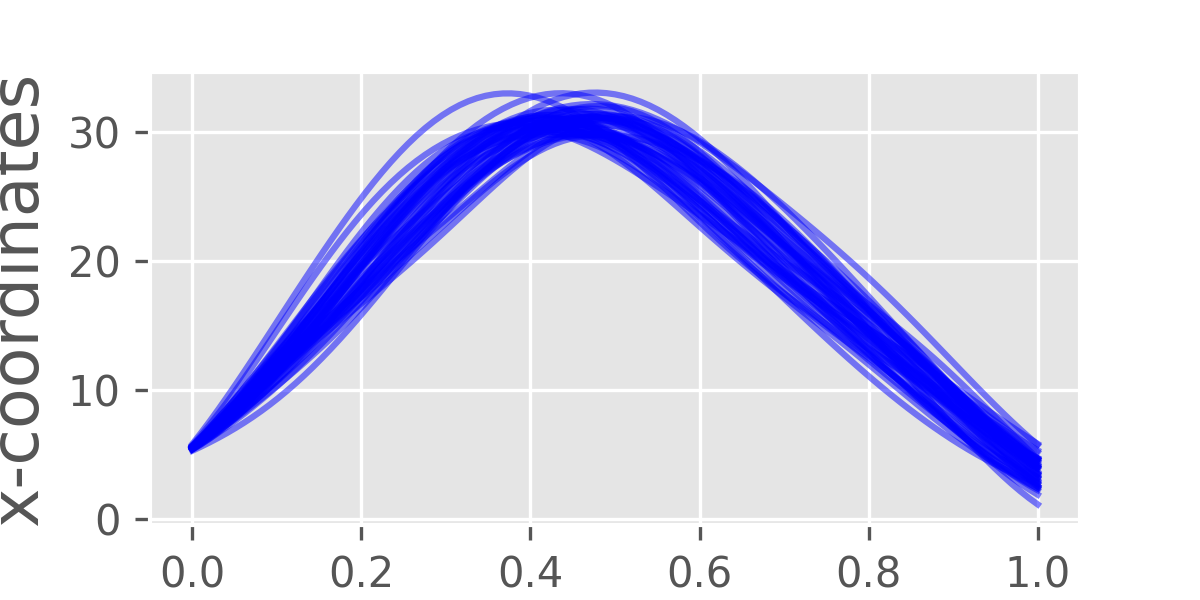}  
        \includegraphics[width=\linewidth]{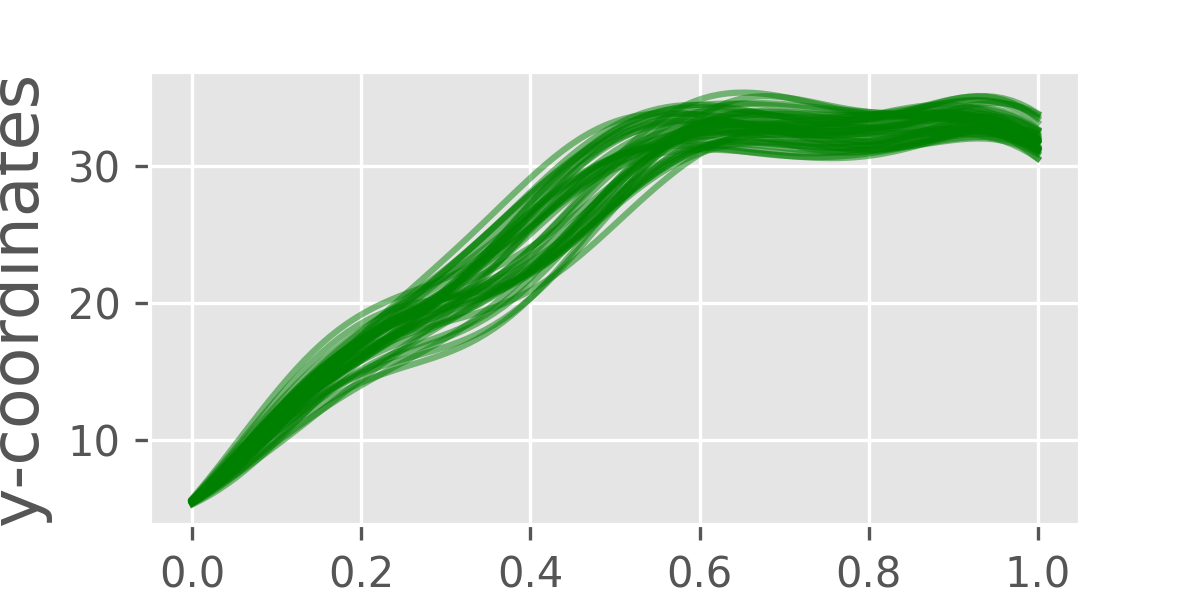}
    \end{subfigure}
    \begin{subfigure}[t]{.32\linewidth}
        \centering
        \includegraphics[width=0.95\linewidth]{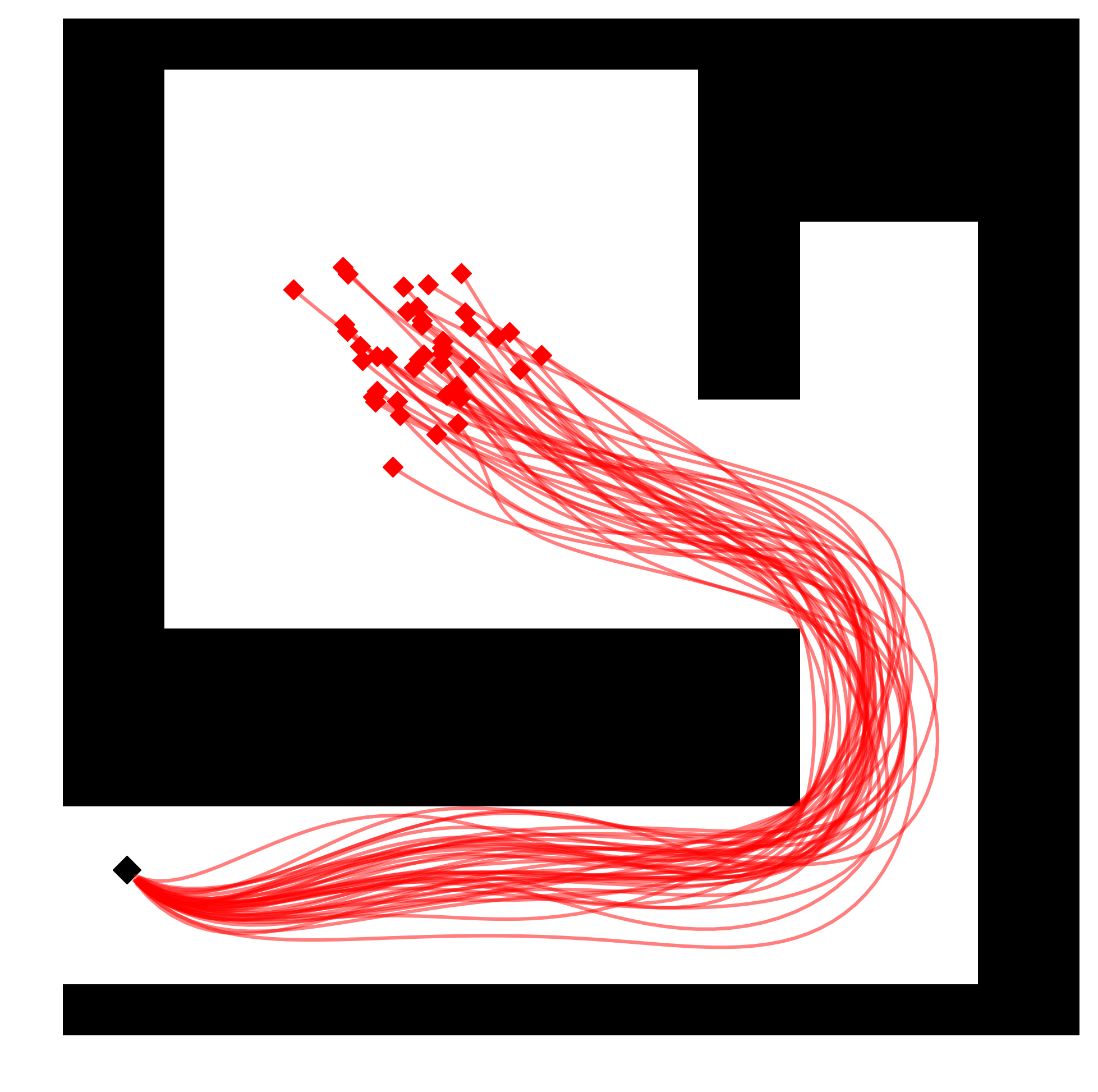} 
        \includegraphics[width=\linewidth]{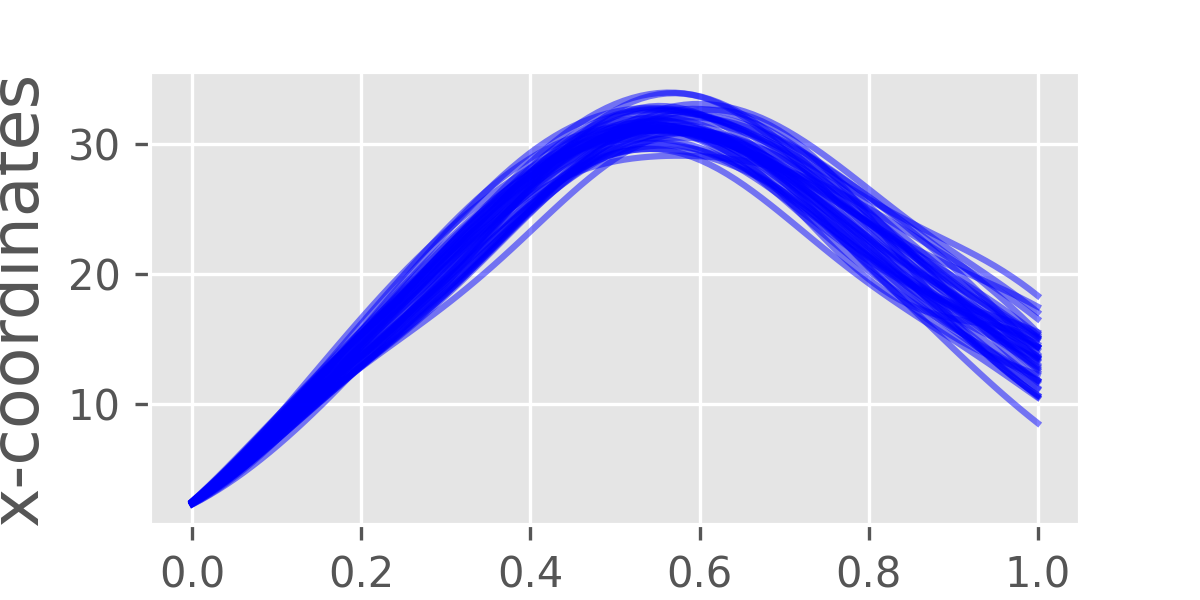}  
        \includegraphics[width=\linewidth]{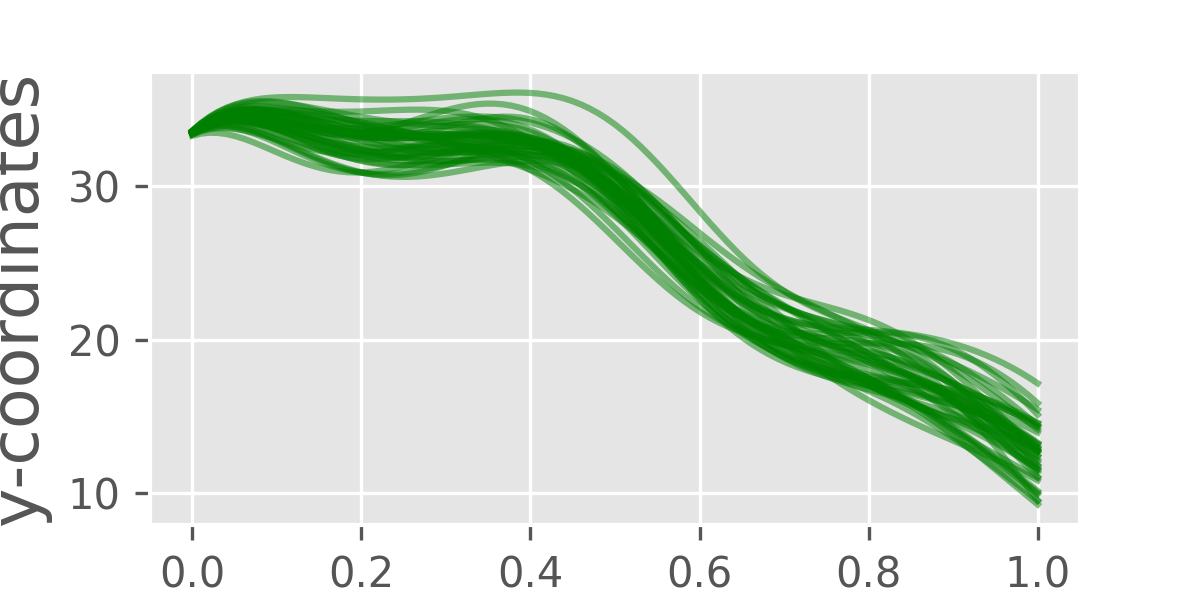}
    \end{subfigure}
    \caption{We generate trajectories (in red, with red markers as endpoints) conditional on the start points (black marker). We condition on two start points inside the room (left, centre), and one point in the corridor (right). The corresponding plots of $x$, $y$-coordinates wrt to $\tau$, $x(\tau)$ and $y(\tau)$  (top, bottom respectively). We see that the trajectories can be conditioned on a variety of start points.}\label{ConditionedTraj}
\end{figure}
\subsection{Diverse Trajectories and Conditioning Trajectory Start Points}
We also want to further investigate whether generated trajectories capture different groups of motion, and whether we can specify start-points for generated trajectories. \Cref{functionswithT} shows generated trajectories in an unseen test environment, along with plots of the trajectory coordinates wrt the normalised time parameter, $\tau$. The test environment is an indoor environment of a corridor with a connected room. Predicted trajectories are coloured in red, and hidden ground truth trajectories are under-laid in blue. The end coordinates of each trajectory have an attached marker. We clearly observe that the trajectories generated can belong to different groupings -- one group starts from inside the room and exit into the corridor, while the other starts in the corridor and end in the room. The multi-modality of the distributions of trajectories learned is even more clear when observing the plots of coordinates against $\tau$. If we observe the y-coordinates, $y(\tau)$, we clearly see two groupings of functions, one of which is at around $x(0)=15$ and around $y(1)=32$, the other grouping has values around $x(0)=32$ and $y(1)=15$.

In the same environment, we attempt to condition the generated trajectories to different start-points. The generated trajectories (top), as well as trajectory coordinates $x(\tau)$ (middle) and $y(\tau)$ (bottom) of the trajectories, are shown in \Cref{ConditionedTraj}. The generated trajectories are in red, with red marker end-points, and black start-points. In the left and centre plots, we condition the trajectories' start point on two coordinates in the room, ${(x(0)=5, y(0)=20)}$ (left) and ${(x(0)=5,y(0)=5)}$ (centre), as well conditioning on a start-point in the corridor, ${(x(0)=2, y(0)=35)}$. We see that in all three cases, OTNet was able to generate reasonable trajectories, which follow the environment structure, with each starting at their designated start coordinates.

\subsection{Transferring Trajectories in Simulated Environments to Real-world Data}

We also investigate the transfer of trajectories from simulated maps to real-world environments. We select three different regions in the Intel-lab dataset \citep{Intel}, and transfer trajectory patterns trained on the simulated Occ-Traj dataset. \Cref{TransferEx} shows trajectories (top two rows, green with magenta end-points) generated on three different sub-maps, and conditioned on two different starting locations. We see that OTNet is able to generate multi-modal distributions of trajectories, with distinct groupings, and largely conforms to the environmental geometry, even though the real-world occupancy differs from the simulated training environments significantly. We visually compare trajectories generated by OTNet to those by the 1-nearest neighbour approach (bottom row, blue with magenta end-points), which transfers the trajectories from the most similar map to the queried map. Qualitatively, we evaluate the trajectories generated by OTNet to conform better to the real-world environment compared to the nearest-neighbour trajectories. The 1-nearest neighbour approach can only draw from the most similar map. We note that the 1-nearest neighbour approach performs well when transferring to the simulated dataset, as it could often find a relatively similar map that has been observed. Whereas, the real-world occupancy map is not very similar to any single map we have experienced, but OTNet can generalise trajectories from a combination of maps which resembles the test environment closer. Hence, OTNet is more capable of generalising to unseen environments. The nearest neighbour approach is also unable to immediately generate similar new trajectories beyond those included in the data, nor is it possible to condition predicted trajectories on a start-point. Both generating similar new trajectories, and conditioning on points can be achieved with OTNet.

\section{Summary}
We present a novel generative model, OTNet, capable of producing likely motion trajectories in new environments where no motion has been observed. We generalise observed motion trajectories in alternative training environments. The OTNet encodes maps as a feature vector of similarities, and embeds observed trajectories as function parameters. A neural network is used to learn conditional distributions over the parameters. Realisations of the vectors can then be sampled from the conditional distribution, and used to reconstruct generated trajectories. We empirically show the strong performance of OTNet against popular generative methods. Further improvements on OTNet could include incorporating temporal changes in trajectory patterns into the framework.

In the following \cref{chap6}, we take an alternative combined learning and optimisation approach to accounting for environment structure when understanding motion patterns. This allows for hard-constraints to be applied on the distribution of trajectories predicted. 

\pagebreak

\chapter{Structurally Constrained Motion Prediction}\label{chap6}\blfootnote{This chapter has been published in IROS as \cite{Prob_struct_const}.}
\renewcommand\vec[1]{\mathbf{#1}}
\section{Introduction}
Autonomous robots, such as service robots and self-driving vehicles, are often required to coexist in environments with other dynamic objects. For robots to safely navigate and plan in an anticipatory manner in dynamic environments, accurate motion predictions for other nearby objects are needed. Many recent developments \cite{Alahi2016SocialLH,Gupta2018SocialGS,KTM} in motion prediction have relied on learning based approaches, whereby future positions, or probabilities over future positions, are learned from motion trajectories of previously observed dynamic objects. Learning-based methods to motion prediction typically extract patterns from the training data. These patterns are then used to extrapolate future positions based on new observations without explicitly considering the feasibility of the prediction. 

However, there are often rules for motion trajectories that are known {\em a priori}, but are not explicitly enforced during extrapolation by learning-based models. In particular, we address the problem of incorporating constraints that arise from the environment's structure, such as obstacles, into motion predictions. In most environments, motion trajectories are not highly unstructured, and depend closely on the environment occupancy structure. 

To address the shortcomings of purely learning-based methods, we view motion prediction as a combined probabilistic trajectory learning and constrained trajectory optimisation problem. Constrained optimisation methods allow for constraints to be imposed directly onto the distribution of trajectories without incorporating the collected data. We propose a novel framework that incorporates structural constraints into probabilistic motion prediction problems. Our framework leverages the ability of learning based models to extract patterns from data, as well as the ability of constrained trajectory optimisation approaches to explicitly specify constraints, giving better generalisability and higher quality extrapolations. Specifically, our contributions consist of: (1) a method to learn a distribution over trajectories from observations that is amenable to being optimised; (2) an approach to enforce constraints on distributions of trajectories, particularly collision chance constraints when provided an occupancy map.

\begin{figure}[t]
    \centering
    \includegraphics[width=0.5\textwidth]{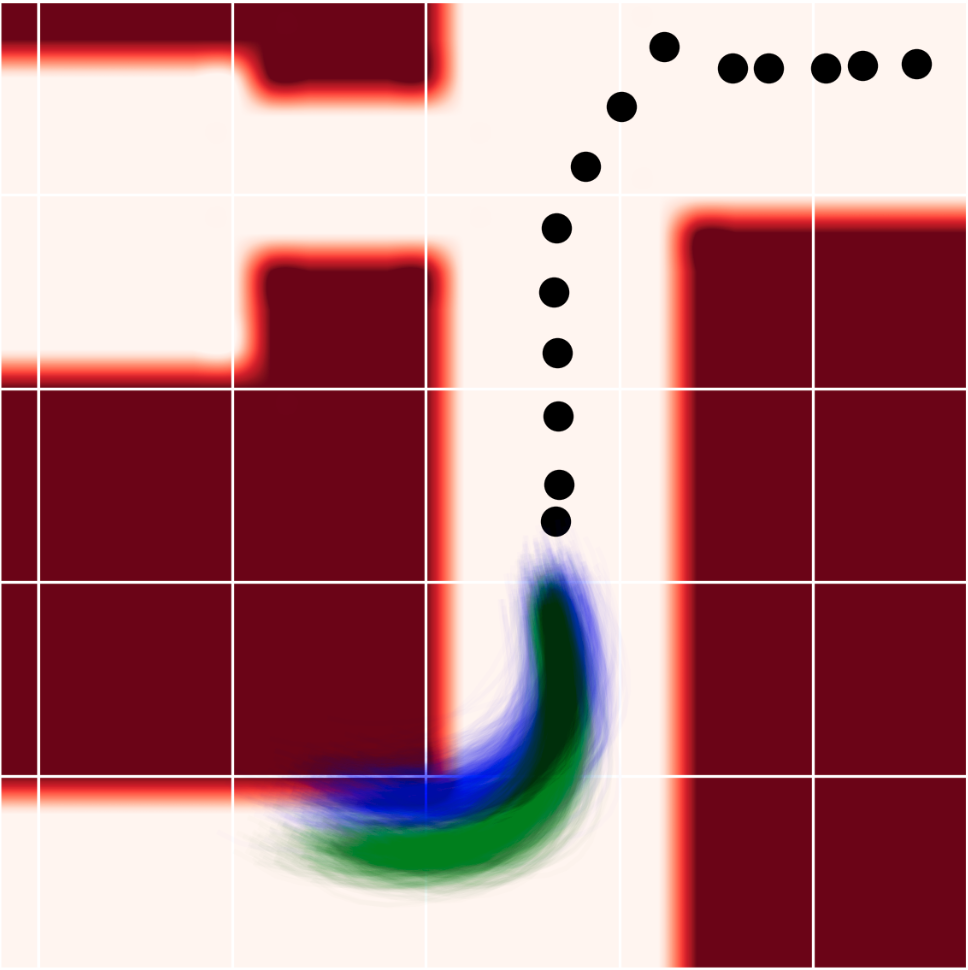}
    \caption{A simple example of enforcing predictions to be consistent with environment: Provided an observed trajectory segment (black), a purely learning method may predict collision-prone distributions of trajectories (blue). Our framework formulates an optimisation problem with chance constraints arising from environmental structure, as given by an occupancy map. We solve to obtain predictions (green) which are compliant with the environment, while remaining similar to the learned future.}
    \label{Fig1}
\end{figure}

\section{Related Work}\label{RelatedWork}
\subsubsection{Motion Trajectory Prediction}
Motion prediction is a problem central to robot autonomy. The most simple methods of motion prediction are physics-based models, such as constant velocity/ acceleration models \citep{Schubert2008ComparisonAE}. More complex physics-based models often fuse multiple dynamic models, or select a best fit model from a set \citep{Kooij2018ContextBasedPP}. Recent developments in machine learning have led to learning-based methods that learn motion patterns \citep{Alahi2016SocialLH,sptemp} from a dataset of observed trajectories. In particular, long-short term memory (LSTM) and generative adversarial network (GAN)-based models \citep{Alahi2016SocialLH,Zhang2019SRLSTMSR,Amirian2019SocialWL,Gupta2018SocialGS} have gained popularity. Methods of learning entire trajectories, similar to our learning component, but with relaxed dependencies, is outlined in \cite{KTM,OTNet}. Further attempts to incorporate static obstacles in learning attempt to do so as a part of the learning pipeline \citep{Sadeghian_2019_CVPR} and attempts to extrapolate behaviour. Approaches of this kind cannot guarantee that valid solutions satisfy specified chance constraints, and there have not been approaches that directly incorporate occupancy maps, which are typically built by robots from depth sensor data.\\
\subsubsection{Trajectory Optimisation}
Trajectory optimisation has been successfully applied in motion planning, by directly expressing obstacles as constraints to solve for collision-free paths between start and goal points. Popular methods include \emph{Trajopt} \citep{Schulman2013FindingLO}, \emph{CHOMP} \citep{Zucker2013CHOMPCH}, \emph{STOMP} \citep{Kalakrishnan2011STOMPST}, and \emph{GPMP2} \citep{Dong2016MotionPA}. Trajopt utilises sequential quadratic programming to find locally optimal trajectories compliant with constraints. CHOMP exploits obstacle gradients for efficient optimisation. Further work on gradient-based trajectory optimisation for planning can be found in \cite{Marinho2016FunctionalGM,Francis2019FastSF}. Although trajectory optimisation is a popular choice to obtain collision-free trajectories for motion-planning, there has yet to be extensive work in utilising trajectory optimisation for motion prediction. The optimisation required in this work is also different in that we optimise to obtain distributions of trajectories, rather than a single trajectory as required in motion planning.
\subsubsection{Combined Learning and Optimisation} 
There has been previous explorations of combining learning and trajectory optimisation. \cite{Toussaint1} introduces a framework which combines a known analytical cost function with black box functions in reinforcement learning for manipulation. \cite{clamp} introduces CLAMP, a framework that combines learning from demonstration with optimisation based motion planning. Theoretical aspects of using learning in conjunction with optimisation, in a Bayesian framework known as posterior regularisation has also been investigated in \cite{PReg}. All of these methods aim to add more structure to be encoded in learning problems. Similarly our proposed framework allows for encoding of structure, but specifically for the trajectory prediction problem, where we are required to do learning and optimisation on distributions of trajectories with respect to environment occupancy.

\section{Continuous-time Motion Trajectories}\label{continuousTsec}
Our framework operates over continuous motion trajectories, capable of being queried at arbitrary resolution, without forward simulating. Here, we briefly revisit the continuous-time motion trajectory representation used in this work, and introduce the relevant notation. We restrict ourselves to the 2-D case. 

Similar to methods described in \cite{Francis2019FastSF,Marinho2016FunctionalGM,KTM}, we represent motion trajectories as functions $\xi: [0,T]\rightarrow \mathbb{R}^{2}$. $T$ represents the prediction time horizon. In this chapter, we limit our discussion to two-dimensional continuous trajectories. We construct our trajectories as dot products between weight vectors and \emph{reproducing kernels}. We limit our discussion to squared exponential radial basis function (RBF) kernels, though methods in this study generalise to various kernels with minimal modifications. Using RBFs result in smooth trajectories, with good empirical convergence properties for functional trajectory optimisation \citep{Marinho2016FunctionalGM}. A trajectory, $\xi(t)$, can be expressed as:
\begin{align}
    \vec \xi(t) = 
    \begin{bmatrix}
        \vec {w^{x}}^{\top} \bm \phi(t) \\
        \vec {w^{y}}^{\top} \bm \phi(t)
    \end{bmatrix},
    &&
    \bm{\phi}(t) = \left[
        \phi(t, t'_{1}), \dots, \phi(t, t'_M)
    \right]^\top,
    \label{trajDef}
\end{align}
where the basis functions are given as
\begin{equation}
\bm{\phi}(t, t') =\exp(-\gamma ||t - t'||^2_{2}).
\end{equation}
The coordinates of a dynamic object at time, $t\in[0,T]$, is given as $x(t)$ and $y(t)$, $\vec{w}^{x}$, $\vec{w}^{y}\in \mathbb{R}^M$ are weight parameters, $\bm{\phi}(t)\in \mathbb{R}^M$, and $\bm{t'}=[t'_1,t'_2,\ldots,t'_{M}]^{\top}$ are $M$ pre-define time points at which RBF kernels are centered. $\gamma$ is a length-scale hyper-parameter which determines the impact a time point neighbouring times on the trajectory. 

Motion data typically comes in the form of discrete sequences of timestamped coordinates $\{t_i, x_i,y_i\}, i = 0, \dots, N$. To construct a best-fit continuous trajectory of discrete sequences from $N$ timestamped coordinates, we solve the ridge regression problem:
 \begin{align}
     \min\limits_{\vec{w}^{x},\vec{w}^y}\sum\limits_{n=1}^{N}\big\{(x_n-\vec{w}^{{x}^{\top}}\bm{\phi}(t_n))^{2}&+(y_n-\vec{w}^{{y}^{\top}}\bm{\phi}(t_n))^{2}+\lambda(||\vec{w}^x||^{2}+||_{2}\vec{w}^y||^{2}_{2})\big\},\label{ridgeeqn}
 \end{align}
where $\lambda$ is a regularisation hyper-parameter, with the set of $M$ uniform times $\bm{t}'$ selected \emph{a priori}, as centers where the kernels are fixed. Evaluating the closed form ridge regression solution results in weight vectors $\vec{w}^{x}$ and $\vec{w}^{x}$ which forms the trajectory $\xi(t)=[\vec{w}^{{x}^{\top}}\bm{\phi}(t),\vec{w}^{{y}^{\top}}\bm{\phi}(t)]$. We denote the stacked vector of $\vec{w}^{x}$ and $\vec{w}^{y}$ as $\vec{w}$. As future motion trajectories are inherently uncertain, the motion prediction problem often requires predicting a distribution over future trajectories a dynamic object may follow. We denote distributions of trajectories as $\bm{\Xi}$, with the vector of probabilistic weight parameters, $\bm{\omega}$, and the parameters of the distribution over $\bm{\omega}$ is denoted as $\Psi$. $\bm{\Xi}(t)$ gives the distribution of coordinate position at time $t$. Every sample function of $\bm{\Xi}$ is a trajectory $\xi$, and samples of vector of random variables $\bm{\omega}$ give $\vec{w}$. Denoting the parameter distributions associated with $x$ and $y$ as $\bm{\omega}^{x}$ and $\bm{\omega}^{y}$, distributions of trajectories can be constructed as $\bm{\Xi}=[{\bm{\omega}^x}^{\top}\bm{\phi}(t),{\bm{\omega}^y}^{\top}\bm{\phi}(t)]$. In practice, to obtain $\bm\xi$ we can also sample $\vec{w}$ from the distribution $\bm{\omega}$, and take dot products with $\bm{\phi}$.

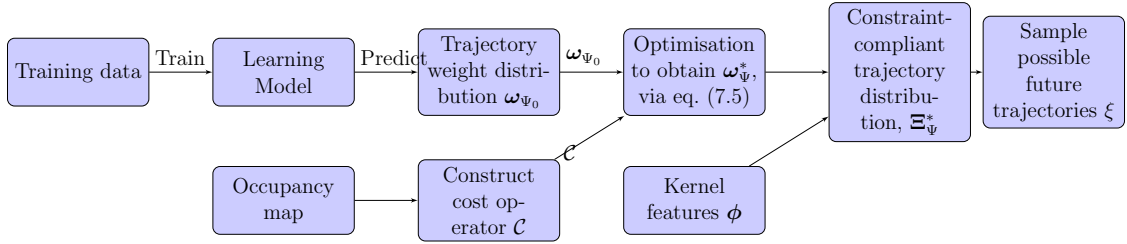
\begin{figure}[t]
\begin{adjustbox}{width=0.9\textwidth,center} 
\begin{tikzpicture}[node distance = 1.6cm]
    \node [block] (Data){Training data};
    \node [block, right of=Data, node distance=4cm] (Pred) { Learning Model};
    \node [block, right of=Pred, node distance=4cm] (Params) { Trajectory weight distribution $\bm{\omega}_{\Psi_{0}}$};
    \node [block, below of=Pred, node distance=2.5cm] (Occ){ Occupancy map};
    \node [block, below of=Params, node distance=2.5cm] (Obs_cost) { Construct cost operator $\mathcal{C}$};
    \node [block, right of=Params, node distance=4cm] (Optimise){ Optimisation to obtain $\bm{\omega}_{\Psi}^{*}$, via \cref{formulationWhole}};
    \node [block, right of=Optimise, node distance=4cm] (Generate){ Constraint-compliant trajectory distribution, $\bm{\Xi}_{\Psi}^{*}$};
    \node [block, right of=Generate, node distance=3cm] (Sample){ Sample possible future trajectories $\xi$};
    \node [block, below of=Optimise, node distance=2.5cm] (featuremaps){Kernel features $\bm{\phi}$};
    
    
    \path [line] (Data) -- node [text width=1cm,midway,above]{Train}(Pred);
    \path [line] (Occ) -- (Obs_cost);
    \path [line] (Pred) -- node [text width=1cm,midway,above]{Predict}(Params);
    \path [line] (Obs_cost) -- node [text width=1cm,midway,below]{$\mathcal{C}$}(Optimise);
    \path [line] (Params) -- node [text width=1cm,midway,above]{$\bm{\omega}_{\Psi_{0}}$}(Optimise);
    \path [line] (Optimise) -- (Generate);
    \path [line] (featuremaps) -- (Generate);
    \path [line] (Generate) -- (Sample);

\end{tikzpicture}
\end{adjustbox}
\caption{Overview of our proposed framework. We combine learning and optimisation to generalise data and enforce constraints.}\label{Overview}
\end{figure}

\section{Motion Prediction with Environmental Constraints}
\subsection{Problem Overview}
\begin{figure}[t]
  \begin{center}
    \includegraphics[width=0.45\textwidth]{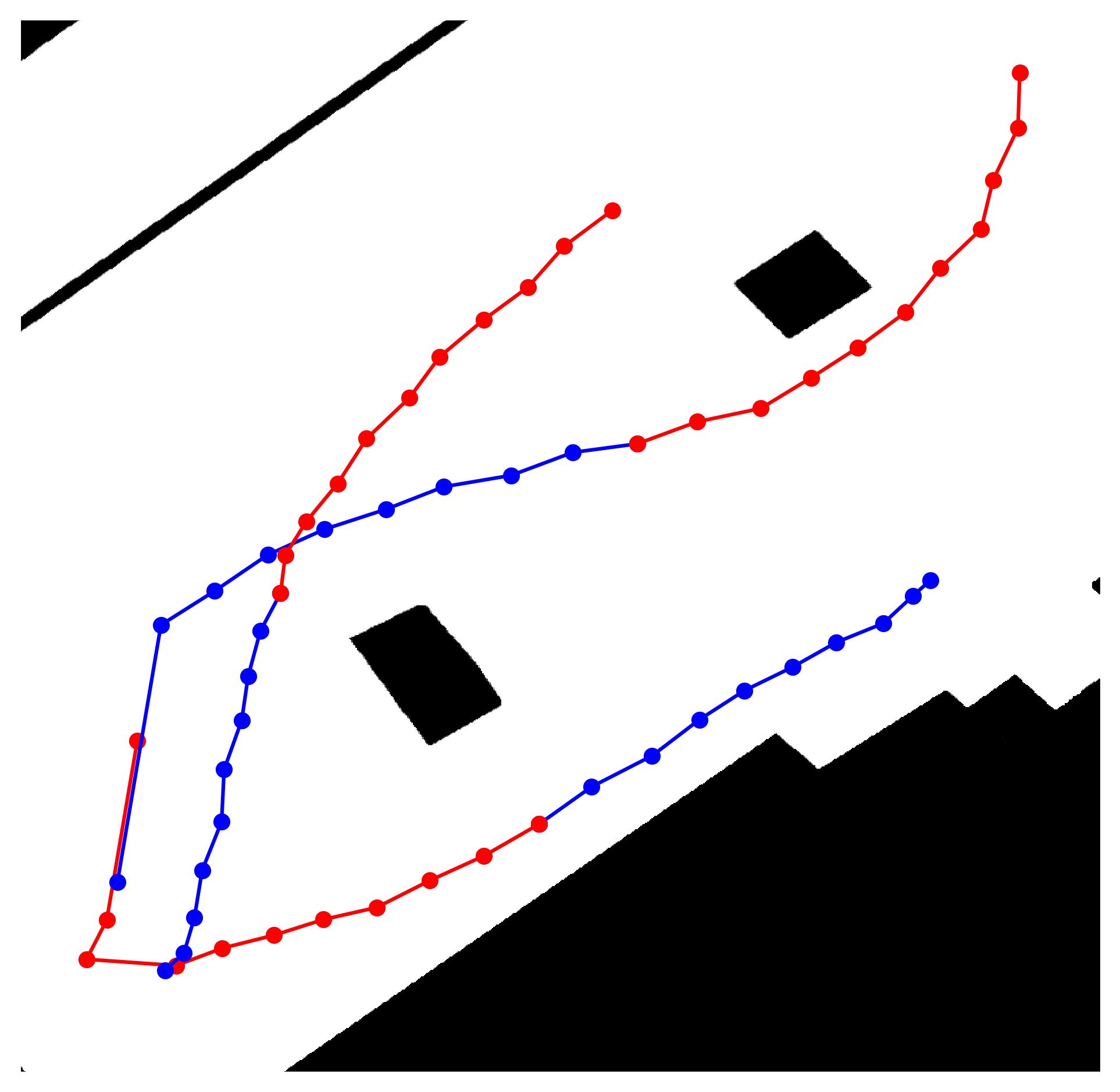}
    \includegraphics[width=0.45\textwidth]{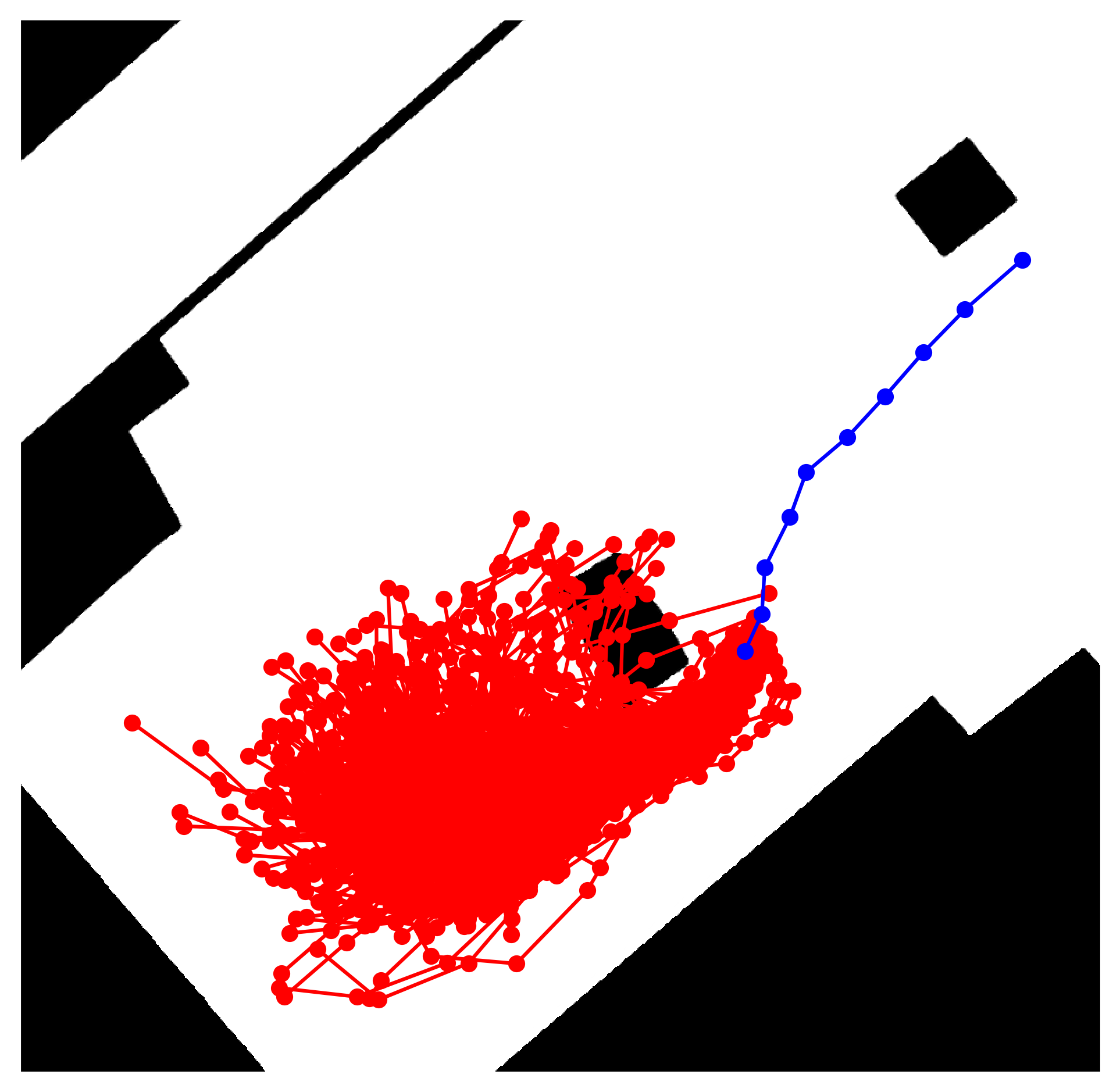}
  \end{center}
  \caption{(Left) Examples from real-world dataset \citep{thorDataset2019}. Historical trajectories in blue, and future in red. (Right) Provided a new unseen trajectory (blue), extrapolate futures (red). Learned future trajectories may collide with obstacles. Our framework tackles this by directly enforcing constraints on trajectory distributions, such that the trajectories are collision free. }\label{datasetProb}
\end{figure}
Our goal is to reason about the possible future trajectories of dynamic objects, based on the immediate past trajectories of the objects, along with occupancy information from the environment. In this work, we limit our discussion to trajectories in 2-D, which usually arise from the tracking of humans and vehicles in video data. We assume to have a dataset containing coordinates of an object up to a time, along with its coordinates thereafter, and an occupancy map of the environment. Examples from a real-world dataset are shown in \cref{datasetProb}, as well as a prediction given by a learning-based model. Given the observed historical coordinates, we aim to extrapolate the distribution of trajectories representing likely future motion thereafter. We require the distribution of trajectories compliant with the environment structure indicated in the occupancy map, such that the probability of hypothesising infeasible trajectories is constrained to a small value.

\subsection{Framework Overview}

An overview of our framework is shown in \cref{Overview}. We learn from our training data to predict a vector of random variables of weights $\bm{\omega}_{\Psi_{0}}$ with distribution parameters $\Psi_{0}$, and produce a prior distribution of trajectories, $\bm{\Xi}_{0}$, as described in \cref{trajDef}. Our aim is to find an alternative random vector $\bm{\omega}_{\Psi}^{*}$, defined by optimised parameters $\Psi^{*}$, which satisfies enforced constraints, while remaining close to the learned distribution $\bm{\omega}_{\Psi_{0}}$. The optimisation objective uses $\bm{\omega}_{\Psi_{0}}$, while constraints are provided through the \textit{obstacle cost operator}, $\mathcal{C}$, which requires an occupancy map of the environment. The desired random vector $\bm{\omega}_{\Psi}^{*}$ defines a constraint-compliant distribution of trajectories, $\bm{\Xi}_{\Psi}^{*}$, provided pre-defined RBF features $\bm{\phi}$, by \cref{trajDef}. Individual constraint-compliant future trajectories, $\xi$, can then be realised from $\bm{\Xi}_{\Psi}^{*}$. 

To find the constrain-compliant parameters $\Psi^{*}$, we formulate the optimisation problem in a similar manner to Bayesian posterior regularisation methods \citep{PReg}. The desired parameters $\Psi^{*}$ is the solution to the constrained optimisation problem:
\begin{align}
    \min_{\Psi}\quad\mathrm{D}_{KL}(\bm{\omega}_{\Psi}\lVert\bm{\omega}_{\Psi_{0}}) \\ \textrm{s.t.}\quad  \mathcal{C}(\bm{\Xi}_{\Psi})-\epsilon\leq 0 \label{formulationWhole}
\end{align}
where $\mathrm{D}_{KL}$ is the Kullback-Leibler (KL) divergence \citep{Bishop:2006}, and $\mathrm{D}_{KL}(\bm{\omega}_{\Psi}\lVert\bm{\omega}_{\Psi_{0}})$ gives us a definition of ``close-ness'' between our predicted $\bm{\omega}_{\Psi_{0}}$ and optimised $\bm{\omega}_{\Psi}$ distributions. $\mathcal{C}(\bm{\Xi}_{\Psi})$ is the obstacle cost operator, which maps distributions of trajectories $\bm{\Xi}_{\Psi}$ to a scalar cost value. The cost is interpreted as the time-averaged probability of generating a trajectory that collides with obstacles. We designate $\epsilon$ to be the allowed limit of the obstacle cost. Intuitively, our optimisation problem returns the closest distribution of trajectories, as defined by the KL divergence term, to that given by our predictive model, subject to the obstacle collision constraint. By separating the learning and optimisation, we disentangle the complexities of the learning model and the structural constraints. Our learned $\bm{\omega}_{\Psi_{0}}$ also provides us with a ``warm-start'' initial solution to our optimisation. In contrast to learning-based approaches, the constraints in our optimisation problem are enforced in valid solutions. The main challenges to solving this trajectory optimisation problem lie in how to obtain the vector of random variables $\bm{\omega}_{\Psi_{0}}$, which defines our prior trajectory distribution $\bm{\Xi}_{0}$, as well as how to derive the obstacle cost operator $\mathcal{C}$ required for optimisation. These are addressed in the following subsections.

\subsection{Learning Multi-modal Distributions of Trajectories}\label{Learndists}
In this subsection, we outline a method of learning a mapping from observed historical trajectories to prior distributions of future trajectories, parameterised by $\bm{\omega}_{\Psi_{0}}$. The resulting prior distributions of trajectories are to be used directly in the optimisation problem \cref{formulationWhole}. We model the distribution of trajectories, $\bm{\Xi}$, as mixtures of multi-output stochastic processes. To capture the correlation between x and y coordinates, the distribution over weight parameters are assumed to be mixtures of matrix normal distributions \citep{Dawid1981SomeMD}. A matrix normal distribution is the generalisation of the normal distribution to matrix-valued random variables, where each sample from a matrix normal distribution gives a matrix. We work with the matrix of random variables defined by vertically stacking the parameters of coordinate dimensions, denoted as $\mathbf{W}=[\bm{\omega}^{x},\bm{\omega}^{y}]\in\mathbb{R}^{M\times 2}$:
\begin{align}
    \mathbf{W}=
    \begin{bmatrix}
    {w}^{x}_{1} & {w}^{y}_{1}\\
    \smash{\vdots} & \smash{\vdots}\\
    {w}^{x}_{M} & {w}^{y}_{M}
    \end{bmatrix}, &&
     p(\mathbf{W})&=\sum_{r=1}^{R}\alpha_{r}\mathrm{MN}\Big(
    \mathbf{M}_{r}, \mathbf{U}_{r}, \mathbf{V}_{r}
    \Big).
\end{align}
We assume that there are $R$ mixture components, similar to a Gaussian mixture model \citep{Bishop:2006}, with each component being a matrix normal distribution, and associated weight components $\alpha_r$. Matrix normal distributions have location matrices $\mathbf{M}_{r}=[\bm{\mu}_{r}^{x},\bm{\mu}_{r}^{y}]\in\mathbb{R}^{M \times 2}$, and positive-definite scale matrices $\mathbf{U}_{r}\in\mathbb{R}^{M \times M}$, $\mathbf{V}_{r}\in\mathbb{R}^{2 \times 2}$ as parameters \citep{MLEMatN}. 

We utilise a neural network to learn a mapping from a fixed-length vectorised coordinate sequences, $\bm{\varphi}$, that encodes observed partial trajectories, to component weights $\bm{\alpha}$, along with location matrices $\mathbf{M}_{r}$ the lower triangular $\mathbf{V}_{r}^{\frac{1}{2}}$ and diagonal $\mathbf{U}_{r}^{\frac{1}{2}}$ matrices. $\mathbf{V}$ is positive-definite and will be constructed by taking the product of two lower triangular matrices. The dependence between different times on the trajectory are largely captured the RBF features in the time domain $\bm{\phi}(t)$, as described in \cref{continuousTsec}. Hence, it is sufficient for $\mathbf{U}$ to be a positive definite diagonal matrix. We obtain the input vector $\bm{\varphi}$ by taking a 10 time-step window of the latest observed coordinates, and vectorising to obtain a vector of 20 dimensions. The neural network used is relatively light-weight, consisting of 4 Relu-activated dense layers with number of neurons: $(15\times M \times R)\rightarrow (5\times M \times R)\rightarrow (5\times M \times R)\rightarrow (R\times [3M+3])$. In this work, our main goal is to outline a representation of future trajectories outputted by the network, such that the result can be directly optimised. Our method does not preclude alternative encoding schemes of conditioned history trajectories to be inputted into the network, such as features from LSTM networks \citep{lstm} or trajectory features outlined in \citep{KTM}. 

We are provided with a dataset of history and future trajectories, given by $N$ pairs of timestamped coordinate sequences. After flattening the history trajectory sequence to obtain $\bm{\varphi}$, and applying \cref{ridgeeqn} on trajectory futures to obtain weight matrices, we have pairs denoted as $\{\bm{\varphi}_{n}, \mathbf{W}_n\}_{n=1}^{N}$. We learn a mapping from $\bm{\varphi}_n$ to parameters $\{\alpha_{r},\mathbf{M}_{r},\mathbf{V}_{r},\mathbf{U}_{r}\}_{r=1}^{R}$ with the fully-connected neural network, with the negative log-likelihood of the weight matrices distributions as the loss function:
\begin{align}
    \mathcal{L}
    &=-\sum_{n=1}^{N}\log \sum_{r=1}^{R}\alpha_{r}\frac{\exp\{-\frac{1}{2}\mathrm{tr}[{\mathbf{V}_{r}}^{-1}(\mathbf{W}_n-\mathbf{M}_{r})^{\top}\mathbf{U}_{r}(\mathbf{W}_n-\mathbf{M}_{r})]\}}{(2\pi)^{M}|\mathbf{V}_{r}||\mathbf{U}_{r}|^{\frac{M}{2}}}.\label{fullLL}
\end{align}
and by applying the activation functions:
\begin{align}
    \mathrm{diag}[(\mathbf{U}_{r})^{\frac{1}{2}}]&=\exp(z^{\mathrm{diag}(\mathbf{U}_{r})}), \\ \mathrm{LowerTrig}[(\mathbf{V}_{r})^{\frac{1}{2}}]&=z^{\mathrm{LowerTrig}(\mathbf{V}_{r})}, \\
    \mathrm{diag}[(\mathbf{V}_{r})^{\frac{1}{2}}]&=\exp(z^{\mathrm{diag}(\mathbf{V}_{r})}), \\
    \alpha_{r}&=\mathrm{softmax}(z^{\alpha_r}),
\end{align}
where $z^{\alpha_r},z^{\mathrm{LowerTrig}(\mathbf{V}_{r})},z^{\mathrm{diag}(\mathbf{V}_{r})},z^{\mathrm{diag}(\mathbf{U}_{r})}$ are neural network outputs, and $\mathrm{diag}(\cdot)$, $\mathrm{LowerTrig}(\cdot)$ indicate the matrix diagonal, and strictly lower triangular matrix elements respectively. The $\exp$ activation function guarantees the validity of the scale matrices, by enforcing $z^{\mathrm{diag}(\mathbf{V}_{r})}>0, z^{\mathrm{diag}(\mathbf{U}_{r})}> 0$. We also enforce $\sum_{r=1}^{R}\alpha_{r}=1$ using the $\mathrm{softmax}$ function. After we train the neural network, provided a vectorised historical trajectory, $\bm{\varphi}$, the neural network gives the corresponding random vector of weights of our prior trajectory distribution, $\bm{\omega}_{\Psi_{0}}=\mathrm{vec}(\mathbf{W})$, where $\mathrm{vec}(\cdot)$ is the vectorise operator. The prior distributions of trajectories is then the dot product of the trajectory parameters and the RBF features, $\bm{\phi}(t)$, as described in \cref{trajDef}.

\subsection{Constraints on Distributions of Trajectories}\label{CostoperatorSubs}
After obtaining the weights, $\bm{\omega}_{\Psi_{0}}$, of a prior distribution of trajectories via learning, our main challenge is to design a cost operator, $\mathcal{C}(\bm{\Xi})$, to encode our desired constraints. Specifically we wish to design $\mathcal{C}(\bm{\Xi})$ to account for collisions of distributions of trajectories, provided an occupancy map. $\mathcal{C}(\bm\Xi)$ maps a distribution of smooth trajectories, $\bm{\Xi}$, to a real-valued scalar cost, which quantifies how collision-prone a given $\bm\Xi$ is. $\mathcal{C}$ differs from the cost operator in trajectory optimisation motion planning problems, such as those in \cite{Zucker2013CHOMPCH,Marinho2016FunctionalGM,Francis2019FastSF}, in that the input is a distribution over trajectories rather than a single trajectory. Our optimisation can not only optimise over trajectory mean, but also over trajectory variances.

\begin{figure}[h]
{\caption{The abscissae (black) for estimating the distribution in world space with the Gauss-Hermite quadrature scheme}\label{fig:absc}}
\centering
{\includegraphics[width=0.65\textwidth]{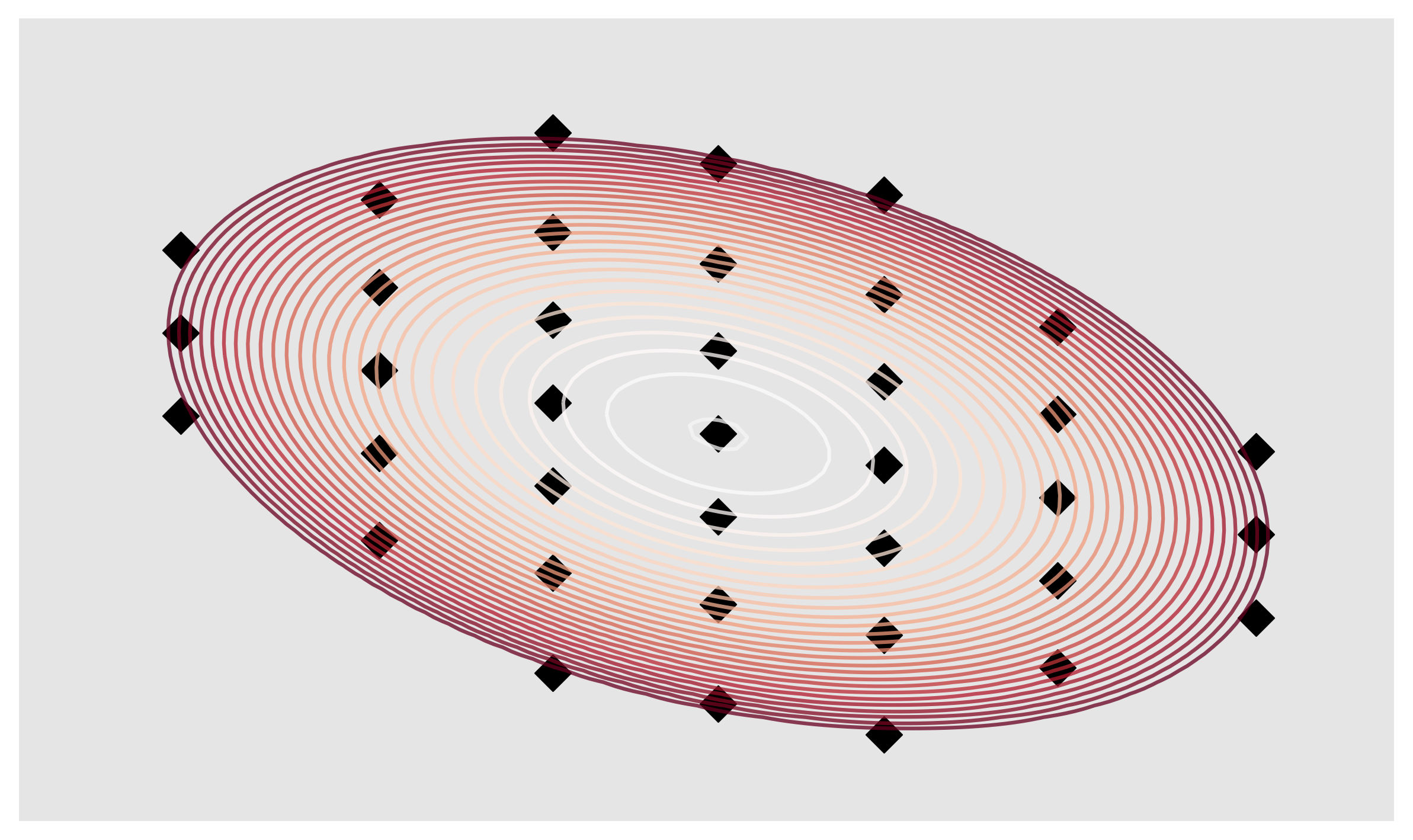}}
\end{figure}

Like trajectory optimisation methods \citep{Zucker2013CHOMPCH}, we assume the cost operator takes the form of average costs over prediction horizon $T$, where the cost at a given point in time, $\mathcal{C}(\bm{\Xi}_{\Psi}(t))$, is the probability of a collision at a given time. $\bm{\Xi}_{\Psi}$ contains weight variables with a distribution parameterised by $\Psi$. We can then interpret $\mathcal{C}(\bm{\Xi}_{\Psi})$ as the average probability of collision over a time horizon, defined as: 
\begin{align}
    \mathcal{C}(\bm\Xi_{\Psi})=&\frac{1}{T}\int\displaylimits_{0}^{T}\mathcal{C}(\bm{\Xi}_{\Psi}(t))\mathrm{d}t\\=&\frac{1}{T}\int\displaylimits_{0}^{T}\int\displaylimits_{\mathbb{R}^2}p(m=1|\vec{x})p(\vec{x}|\Psi,t)\mathrm{d}\vec{x}\mathrm{d}t.\label{Costeqn}
\end{align}
The probability of collision at some time and location is computed by taking the product of the probability of the trajectory reaching the coordinate at the given time and the probability that the coordinate is occupied. We are provided with an occupancy map and the probability a coordinate of interest, $\vec{x}\in \mathbb{R}^{2}$, being occupied is denoted as $p(m=1|\vec{x})$. To evaluate the cost, we could marginalise over the all trajectory parameters, but this may be difficult due to the high dimensionality of $\bm{\omega}$. We instead work in the 2D world space, where the environmental constraints are defined. As $\bm\Xi_{\Psi}$ is constructed by taking the dot product of weight vectors and feature vectors, trajectory coordinate distribution in world space at time $t$ is:
\begin{align}
p(\bm{\Xi}_{\Psi}(t))&=\sum_{r=1}^{R}\alpha_{r}\overbrace{{\mathrm{MN}(\mathbf{M}_{r}^{\top}\bm{\phi}(t),\mathbf{V}_{r},\bm{\phi}(t)^{\top}\mathbf{U}_{r}\bm{\phi}(t))}}^{\text{matrix normal distribution}}\\&=\sum_{r=1}^{R}\alpha_{r}\overbrace{\mathcal{N}(\underbrace{\mathbf{M}_{r}^{\top}\bm{\phi}(t)}_{=:\bm{\mu}_{r}},\underbrace{\bm{\phi}(t)^{\top}\mathbf{U}_{r}\bm{\phi}(t)\mathbf{V}_{r}}_{=:\Sigma_{r}}).}^{\text{multivariate normal}}\label{worldspacedist}
\end{align}
By the equivalences between matrix normal and multivariate normal distributions \citep{Dawid1981SomeMD}, we have a mixture of two-dimensional multivariate normal distributions in world-space, $\bm{\Xi}_{\Psi}(t)\sim \sum_{r=1}^{R}\alpha_r\mathcal{N}(\bm{\mu}_r,\Sigma_r)$. This is the probability distribution over trajectory coordinates at time $t$, $p(\vec{x}|\Psi,t)$. The mixture means and covariances of each component are defined as 
\begin{align}
{\bm{\mu}_r:=\mathbf{M}_{r}^{\top}\bm{\phi}(t)}, && \Sigma_{r}:={\bm{\phi}(t)^{\top}\mathbf{U}_{r}\bm{\phi}(t)\mathbf{V}_{r}}.
\end{align}
Substituting \cref{worldspacedist} into the inner integral of \cref{Costeqn}, gives the cost operator equation at given time:
\begin{align}
&\mathcal{C}(\bm{\Xi}_{\Psi}(t))=\sum_{r=1}^{R}\alpha_{r}\int\displaylimits_{\mathbb{R}^{2}}\Big\{\frac{\exp[-\frac{1}{2}(\vec{x}-\bm{\bm{\mu}_r})^{\top}\Sigma_{r}^{-1}(\vec{x}-\bm{\mu}_r)]}{(2\pi)|\Sigma_{r}|^{\frac{1}{2}}}\Big\}\underbrace{p(m=1|\vec{x})}_{\text{From map}}\mathrm{d}\vec{x}\label{costEqnTime}
\end{align}
Note that $\bm{\mu}_r$ and $\Sigma_{r}$ are dependent on $\mathbf{M}_r,\mathbf{U}_r,\mathbf{V}_r$. During the optimisation, we assume that each mode can be optimised independently and that the importance of each mode, represented by $\alpha$, is accurately captured by the learning model and does not change during trajectory optimisation. We then define the set of parameters to optimise as $\Psi=\big\{\mathbf{M}_{r},\mathbf{U}_{r},\mathbf{V}_{r}\}_{r=1}^{R}$.
The integral defined in \cref{costEqnTime} may be intractable to evaluate analytically, so we aim to find an approximation to it. The coordinate distribution at a given time in world space, $\bm{\Xi}_{\Psi}(t)$, is a mixture of bi-variate Gaussian distributions. We use Gauss-Hermite quadrature, suitable for numerically integrating exponential-class functions, to approximate \cref{costEqnTime}. Quadrature methods approximate an integral by taking the weighted sum of integrand values sampled at computed points, called abscissae. We introduce the change of variables, 
\begin{equation}
\bm{z_{r}}=\frac{1}{\sqrt{2}}L_{r}^{-1}(\vec{x}-\bm{\mu}_r),
\end{equation}
where the Cholesky factorisation gives $\Sigma_{r}=L_{r}L_{r}^{T}$. Using Gauss-Hermite quadrature schemes, we have:
\begin{align}
    \mathcal{C}(\bm\Xi_{\Psi}(t))
    &\approx\sum_{r=1}^{R}\alpha_r\pi^{-1}\sum_{i=1}^{I}\sum_{j=1}^{J}\beta^{i}\beta^{j}p(m=1|\sqrt{2}L\bm{z}+\bm{\mu}_r),\label{eqnQuad}
\end{align}
where we find the values of $\bm{z}^{i,j}=[z^{i},z^{j}]^{\top}$ as abscissa obtained from roots of Hermite polynomials, and $\beta^{i}$, $\beta^{j}$ are the associated quadrature weights. Details of abscissae and weight calculations, as well as a review on multi-dimensional Gauss-Hermite quadrature, can be found in \citep{GaussHermite}. An example of abscissae points when estimating a multi-variate Gaussian is shown in \cref{fig:absc}. Next, we evaluate the outer integral in \cref{Costeqn} using rectangular numerical integration for 
\begin{equation}
{\mathcal{C}(\bm\Xi_{\Psi})=\int_{0}^{T}\mathcal{C}(\bm{\Xi}_{\Psi}(t))}\mathrm{d}t. 
\end{equation}

With $\mathcal{C}(\bm\Xi_{\Psi})$ defined, we are in a position to solve the combined learning and constrained optimisation problem, defined earlier in \cref{formulationWhole}, using a non-linear optimisation solver, such as the sequential least squares optimiser SLSQP \citep{slsqp}. Analytical gradients with respect to occupancy is provided when using a continuous differential occupancy map such as those introduced in \cite{HilbertMaps, HM}. By utilising the equivalence between matrix normal distributions and multivariate Gaussians \citep{MLEMatN}, the KL divergence term between $\bm{\omega}_{\Psi}$ and $\bm{\omega}_{\Psi_{0}}$ components can be computed \citep{Bishop:2006}. After obtaining the solution of the optimisation we have the optimised distribution of $\bm{\omega}_{\Psi}^{*}$. By taking dot products between $\bm{\omega}_{\Psi}^{*}$ and $\bm{\phi}$, we can construct $\bm{\Xi}_{\Psi}^{*}$ the constraint-compliant trajectory distribution of futures. Individual realised trajectories $\xi$ can then be generated from $\bm{\Xi}_{\Psi}$. 

Although we have focused on collision constraints with respect to environment structure, the optimisation allows for constraints on velocity or acceleration. The continuous nature of our trajectories allows time derivatives of trajectories to be obtained. Specifically, the \emph{derivative reproducing property}, ensures the derivatives of smooth kernels also correspond to kernels \citep{DerivativeRKHS}. We can reason about the $n^{th}$-order derivatives of displacement by replacing $\bm{\phi}(t)$ with $\frac{\partial^{n}\bm{\phi}(t)}{\partial t^{n}}$. 

\section{Experimental Evaluation}
We empirically evaluate (1) the ability of our learning component to learn probabilistic representations of future trajectories to provide good prior trajectories distributions; (2) the ability of the proposed framework to enforce collision constraints, and its effect on trajectory quality.

\subsection{Datasets and Metrics}
We run our experiments on one simulated dataset, and two real-world datasets, including:
\begin{itemize}
    \item Simulated dataset \textbf{(Simulated)}: This contains simulated trajectories on a floor-plan;
    \item TH{\"O}R dataset \citep{thorDataset2019} \textbf{(Indoor)}: This dataset contains real human trajectories walking in a room with three large obstacles in the center of the room. 
    \item Lankershim dataset \textbf{(Traffic)}: This dataset contains real vehicle data along a long boulevard. We use a particular busy intersection between $x=[-80, 80]$ and $y=[250, 500]$. 
\end{itemize}
Throughout each of our experiments, we ensure that the observed trajectory conditioned on contains 10 timesteps. The prediction horizons, $T$, for the simulated, indoor and traffic datasets are 15, 10, and 20 timesteps respectively. We split long trajectories, such that we have 200 pairs of partial trajectories in the simulated dataset, while the indoor dataset contained 871 pairs of partial trajectories, and the traffic dataset contained 6175 pairs. Trajectories that were shorter than the length of the prediction horizon were filtered out. A train-to-test ratio of 80:20 was used for each dataset. For our learning model, we set the number of mixture components $R=2$, hyper-parameters $M=8, 10, 11$, and $\gamma=0.05, 0.1, 0.2$ for the simulated, indoors, and traffic datasets respectively. Maps for the simulated and indoor datasets are available, and we construct a map for the traffic dataset by considering the road positions. Our occupancy map is represented as a continuous Hilbert Map \citep{HilbertMaps}, giving smooth occupancy probabilities over coordinates, and occupancy gradients. 

Metrics used in the quantitative evaluation include: 
\begin{itemize}
\item{Average displacement error (ADE)}: The average euclidean distance error between the expectation of the nearest distribution component of trajectories with ground truth trajectory; 
\item{Final displacement error (FDE)}: The euclidean distance error at the end of the time horizon between the expectation of the nearest distribution component of trajectories with ground truth trajectory. 
\item{Average Likelihood (AL)}: To take into account the uncertain nature of motion prediction, for probabilistic models, we record the likelihood of drawing the trajectory averaged over the timesteps. 
\end{itemize}
Note that likelihoods are difficult to interpret and not comparable over different datasets. Lower ADE and FDE, and higher AL indicate better performance. For the evaluation of collisions, we also record the percentage of constraint-violating trajectory distributions in the test set.

\begin{table}[t]
\centering
\begin{adjustbox}{width=0.6\textwidth} 
\begin{tabular}{llccc}
\toprule
                           &          & \small ADE  & \small FDE   & \small AL \\
                           \midrule
\multirow{4}{*}{\small Simulated} &\small Ours      & 1.29 & 2.27  & 0.12       \\
                           &\small KTM \citep{KTM}     & 1.44 & 2.41  & 0.11       \\
                           &\small CV       & 4.27 & 11.66 & -          \\
                           &\small NN-naive & 2.11 & 6.07  & -          \\
                           \midrule
\multirow{4}{*}{\small Indoors}   &\small Ours      & 0.94 & 1.32  & 3.32       \\
                           &\small KTM      & 0.65 & 1.26  & 2.88       \\
                           &\small CV       & 2.08 & 4.66  & -          \\
                           &\small NN-naive & 1.84 & 3.35  & -          \\
                           \midrule
\multirow{4}{*}{\small Traffic}   &\small Ours      & 4.99 & 7.24  & 0.04       \\
                           &\small KTM      & 4.98 & 7.03  & 0.043      \\
                           &\small CV       & 5.14 & 10.38 & -          \\
                           &\small NN-naive & 21.1 & 39.5  & -      \\ 
                           \bottomrule
\end{tabular}
\end{adjustbox}
\caption{We evaluate the quality of our learned trajectory prior with benchmark models. Lower ADE and FDE, and higher AL indicate better performance.}\label{tableRes1} 
\end{table}

\subsection{Learning Continuous-Time Trajectory Distributions as Priors}
The trajectory distribution learning component of our framework provides a prior distribution for optimisation. We examine whether the learning component of our framework is able to learn distributions of future trajectories from data, comparing the performance with several benchmark models. The benchmark models are (1) Kernel trajectory maps (KTMs) \citep{KTM}, (2) Constant velocity (CV) model, and (3) Naive neural-network (NN) model, where we learn a mapping between past trajectories and future trajectories by minimising the mean squared error loss. The neural network comprised of Relu-activated fully-connect layers with the number of neurons: $(560)\rightarrow(180)\rightarrow(180)\rightarrow(N_t)$, where $N_t$ gives the trajectory time horizon. 

\begin{figure}[t]
\centering
        \includegraphics[width=0.3\textwidth]{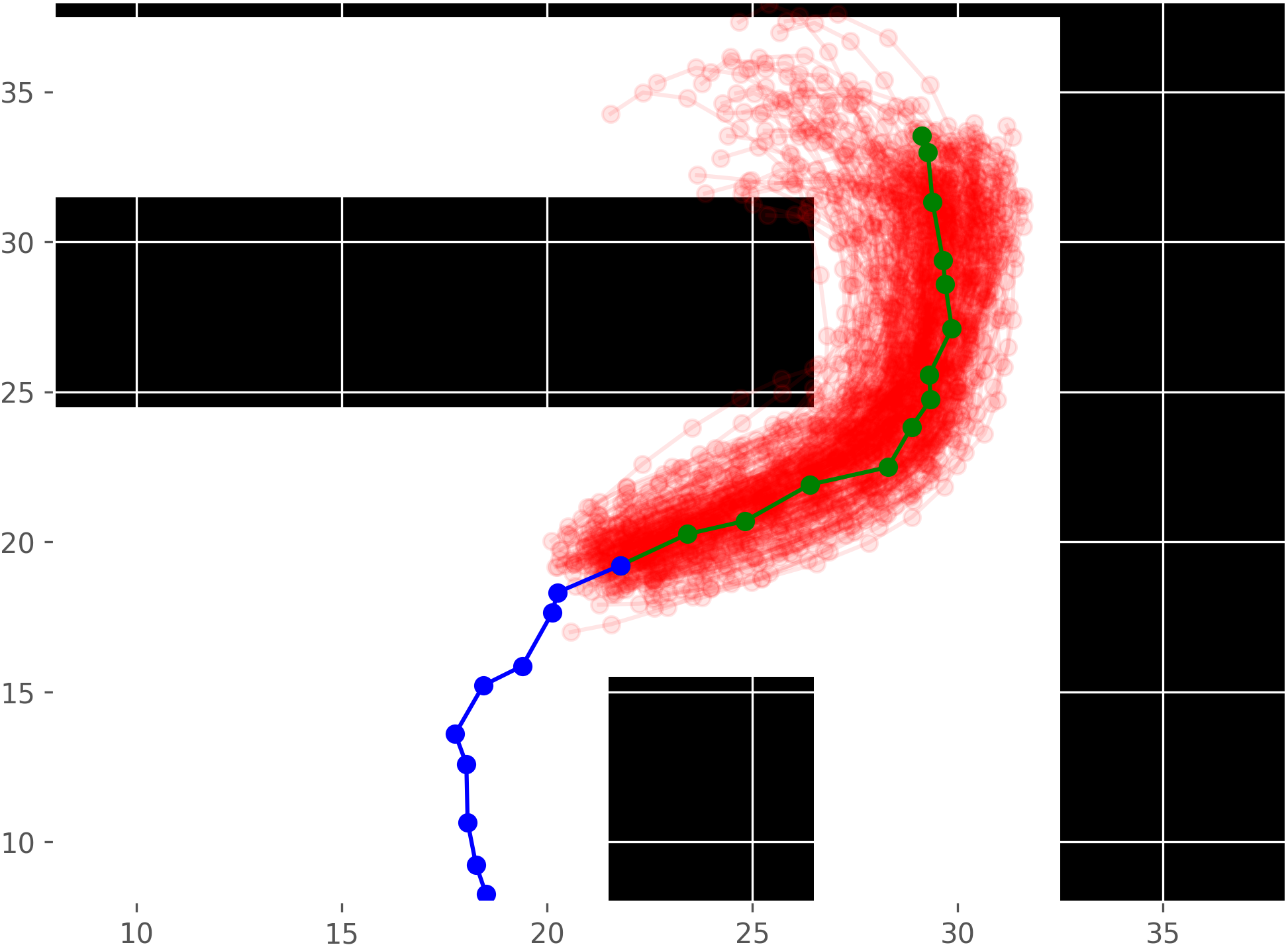}
        \includegraphics[width=0.3\textwidth]{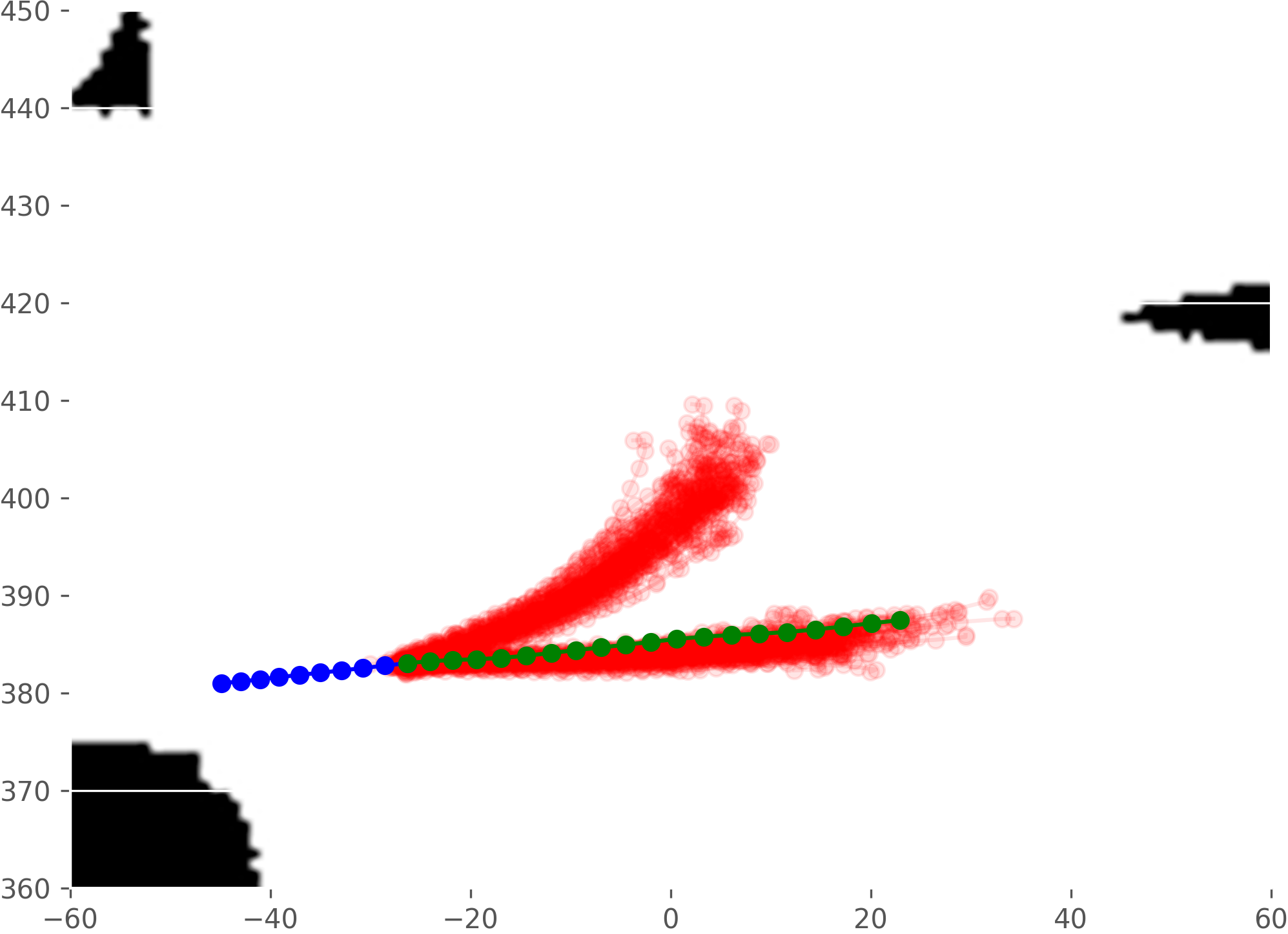}
        \includegraphics[width=0.3\textwidth]{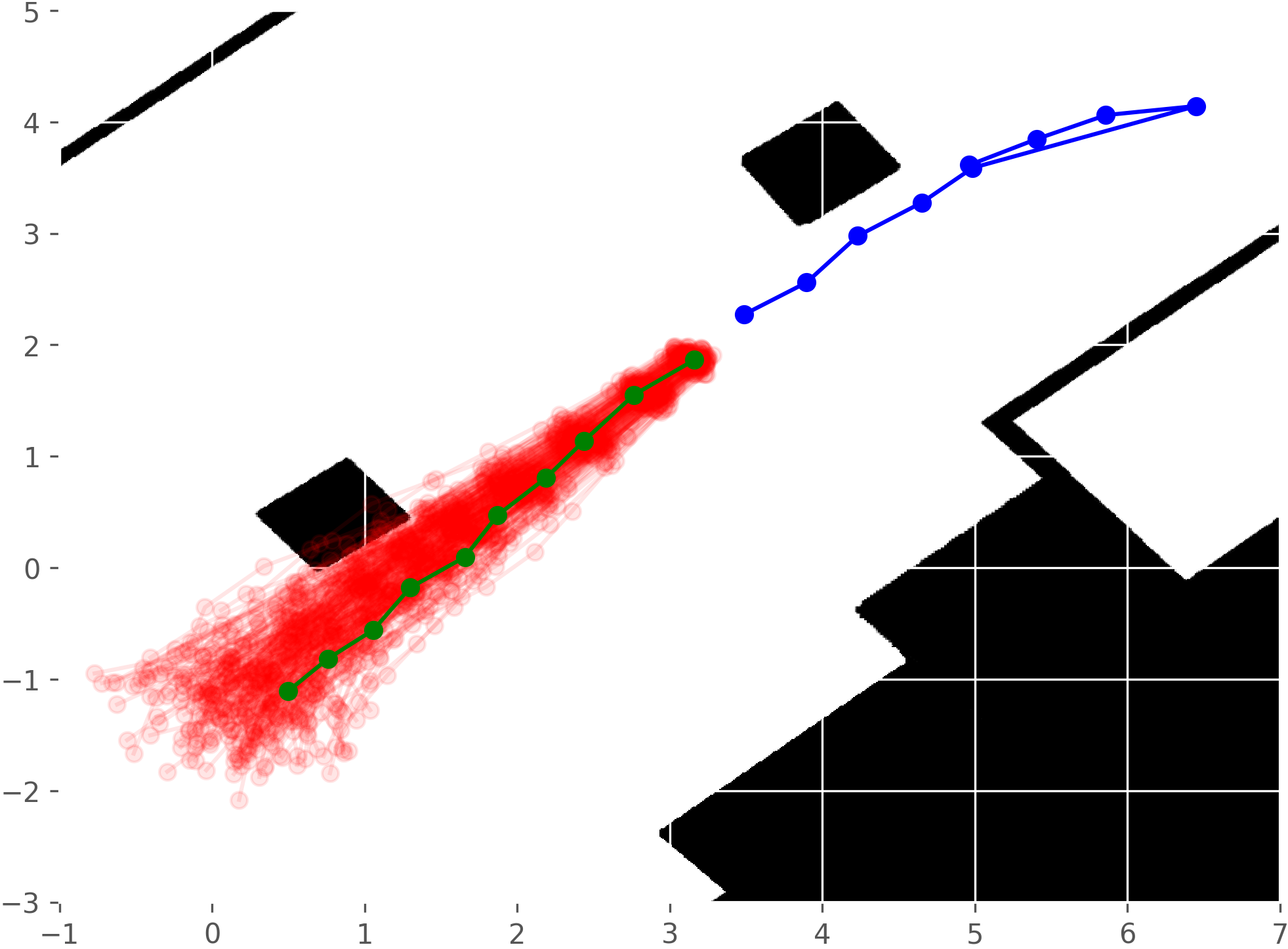}
        
        \includegraphics[width=0.3\textwidth]{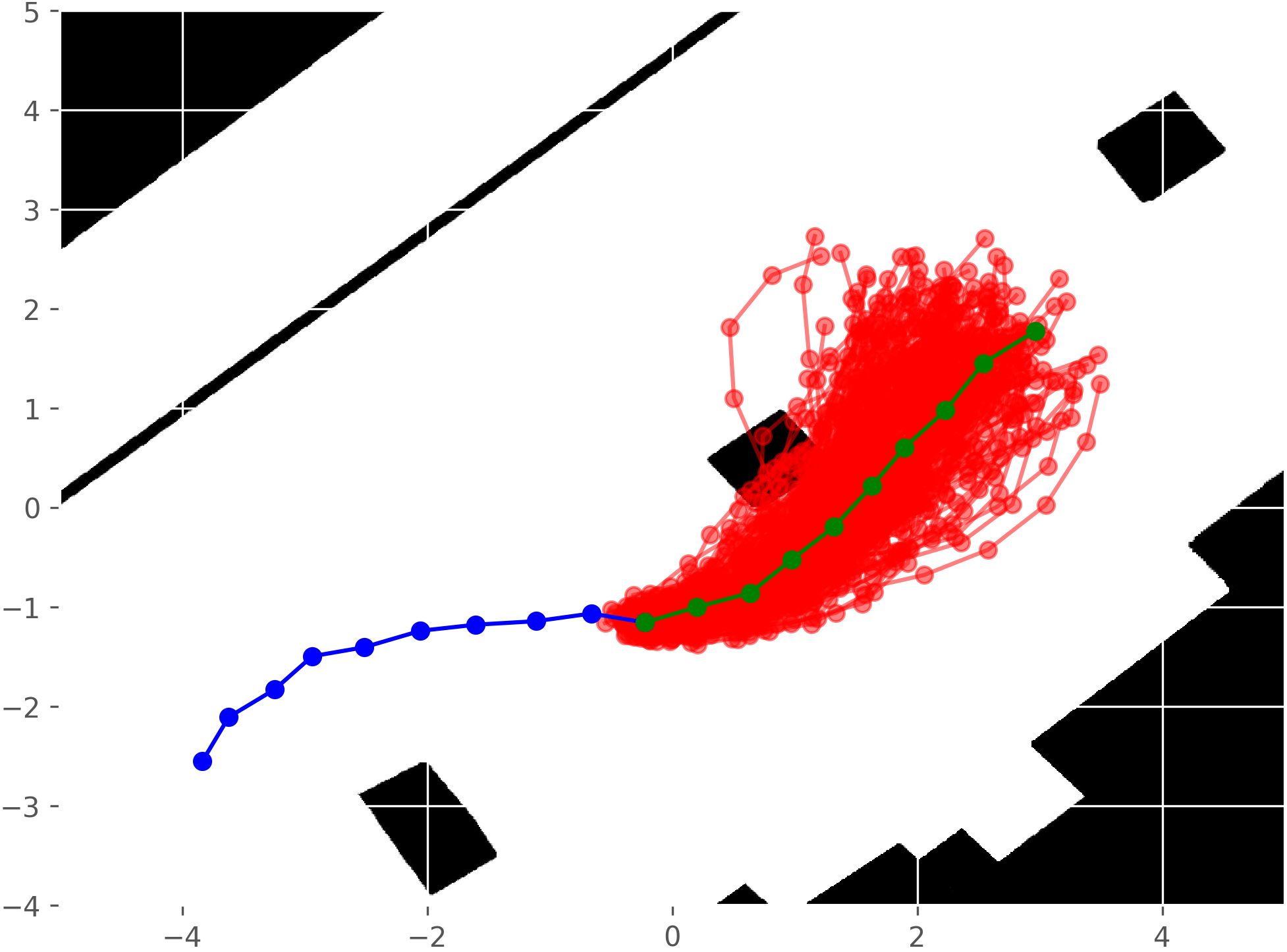}
        \includegraphics[width=0.3\textwidth]{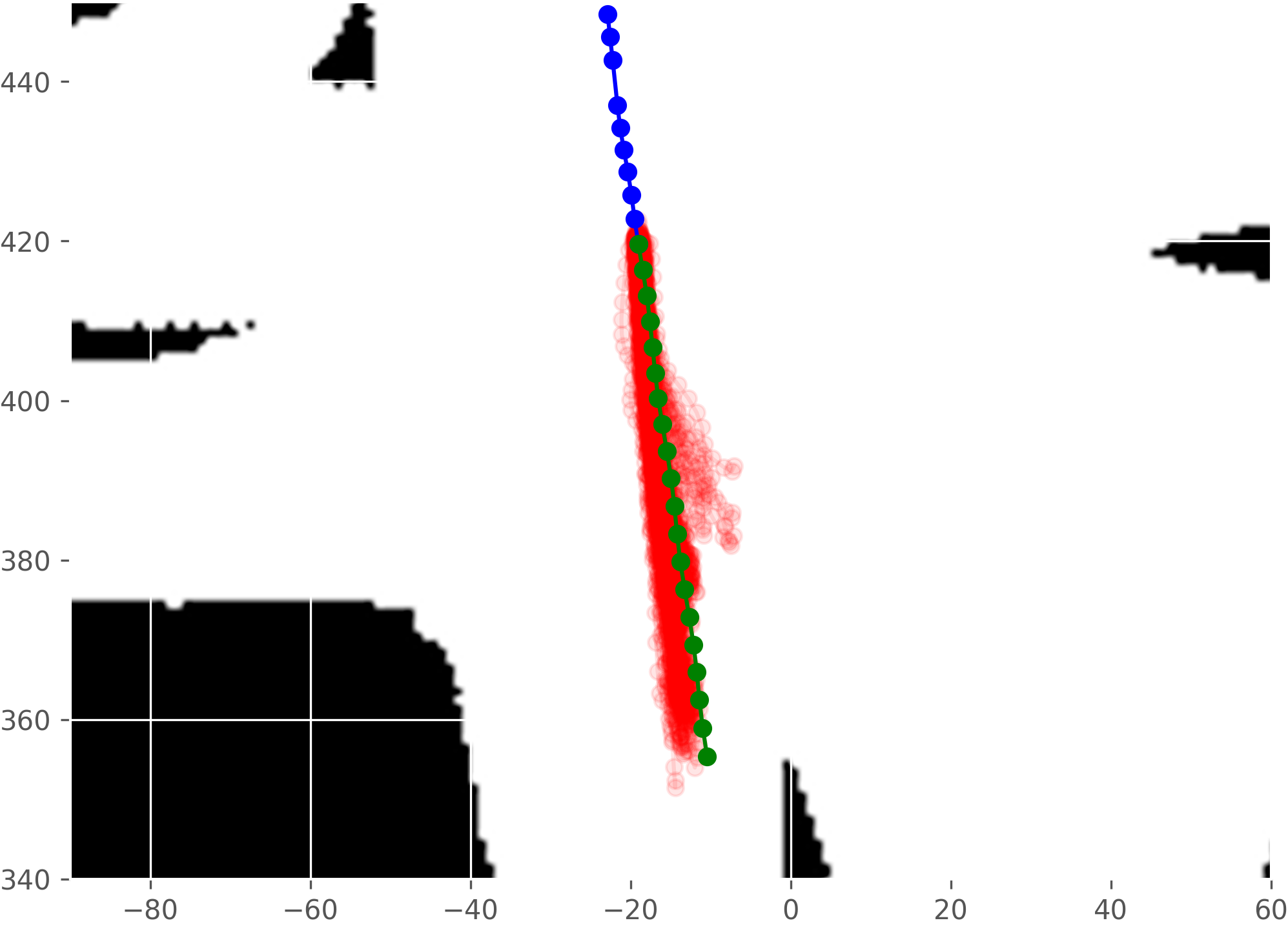}
        \includegraphics[width=0.3\textwidth]{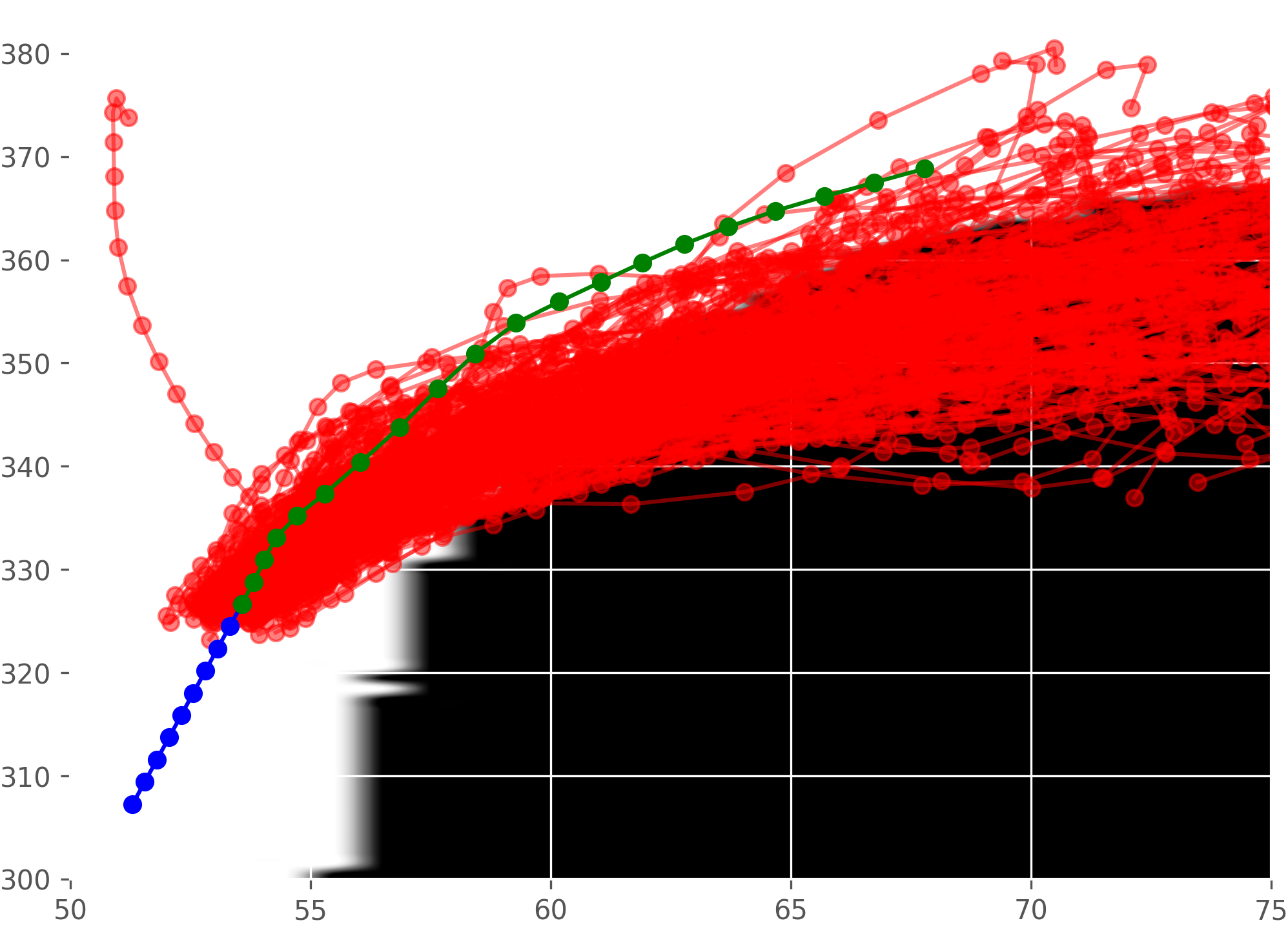}
        
\captionof{figure}{Quantitative evaluations of learned distributions of trajectories (red), Conditioned observations (blue), ground truth (green). We see that our learning model gives relatively good priors, but many sampled trajectories collide with obstacles. The prior distributions are further enforced to be collision-free (See \cref{xyTraj})}\label{ExamplePred}
\end{figure}

We see that the learning component of our framework is able to learn high-quality distributions of trajectories. \Cref{ExamplePred} shows examples of extrapolated trajectory distributions in red, and the observed trajectory in blue. We see that the learned distribution is relatively close to the ground truth green trajectory, and is able to capture the uncertain nature of motion prediction. Although a non-negligible number of sampled trajectories collide with the environment. \Cref{tableRes1} summarises the performance results of our method against our benchmarks. We see that our learning model is comparable to the method proposed in \cite{KTM}, while outperforming a constant velocity and naive neural network model. While the KTM method performs slightly better on the traffic dataset, it is purely learning-based and lacks mechanisms to enforce constraints. We note that our method encodes the observed trajectory by simply flattening the input coordinates, as described in \cref{Learndists}. More sophisticated methods of encoding the observed trajectory (such as with LSTM networks \citep{lstm}), independent of our outlined output of trajectory distributions, could improve the performance of the learning model.


\subsection{Enforcing Structure Compliance}

After obtaining a prior from our learned model, we apply constraints on the prediction via trajectory optimisation. We will evaluate whether collision constraints can be enforced, and their effect on trajectory quality. We define the constraint such that the time-averaged probability of collision is less than or equal to 0.05, $\mathcal{C}-0.05\leq 0$. The SLSQP optimiser was able to solve the optimisation quickly, under half a second, with a python implementation. We obtain trajectory distribution priors from our learned model, and select all of the observations where the learned prior trajectory is constraint-violating. We evaluate the performance of the trajectory priors, and compare them with the trajectory distribution predictions after enforcing constraints. The quantitative results are listed in \cref{tableRes2}. We see after solving with structural constraints, all predictions are feasible. Additionally, the trajectory distribution quality improves after enforcing constraints for all the datasets considered.

\begin{table}[h]
\centering
\begin{tabular}{llcccc}
\toprule
          &             & \small ADE  & \small FDE   & \small AL    & \small C.V.P\\
          \midrule
\small Simulated & \small Unoptimised & 1.48 & 2.87  & 0.11  & 40\%                 \\
          & \small Optimised   & 1.28 & 2.33  & 0.17  & 0\%                  \\
\small Indoors   & \small Unoptimised & 1.62 & 2.41  & 3.42  & 15.4\%               \\
          & \small Optimised   & 1.44 & 2.38  & 3.3   & 0\%                  \\
\small Traffic   & \small Unoptimised & 6.56 & 11.81 & 0.034 & 5.7\%                \\
          & \small Optimised   & 5.14 & 8.95  & 0.043 & 0\%\\
          \bottomrule
\end{tabular}
\caption{We evaluate the quality of our optimised trajectory distribution with collision constraints, relative to the prior. We see that providing constraints improves prediction quality across all dataset. The percentage of trajectory distributions violating constraint, $\mathcal{C}-0.05\leq 0$, relative to the entire test set is also given, as Constraint Violation Percentage (C.V.P). After optimisation all trajectory distributions are constraint-compliant.}\label{tableRes2}
\end{table}

\begin{figure}[t]
        \begin{subfigure}[]{0.195\textwidth}
        \centering
        \includegraphics[width=\textwidth]{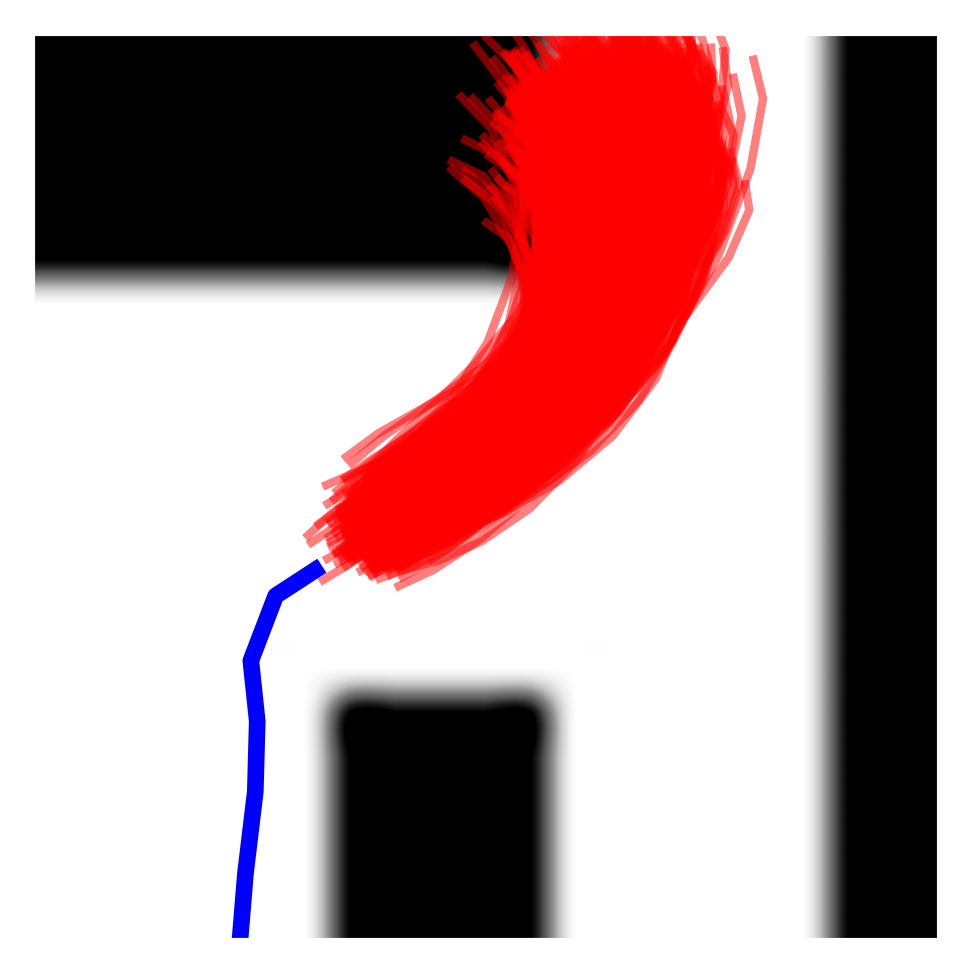}
    \end{subfigure}%
    \begin{subfigure}[]{0.195\textwidth}
        \centering
        \includegraphics[width=\textwidth]{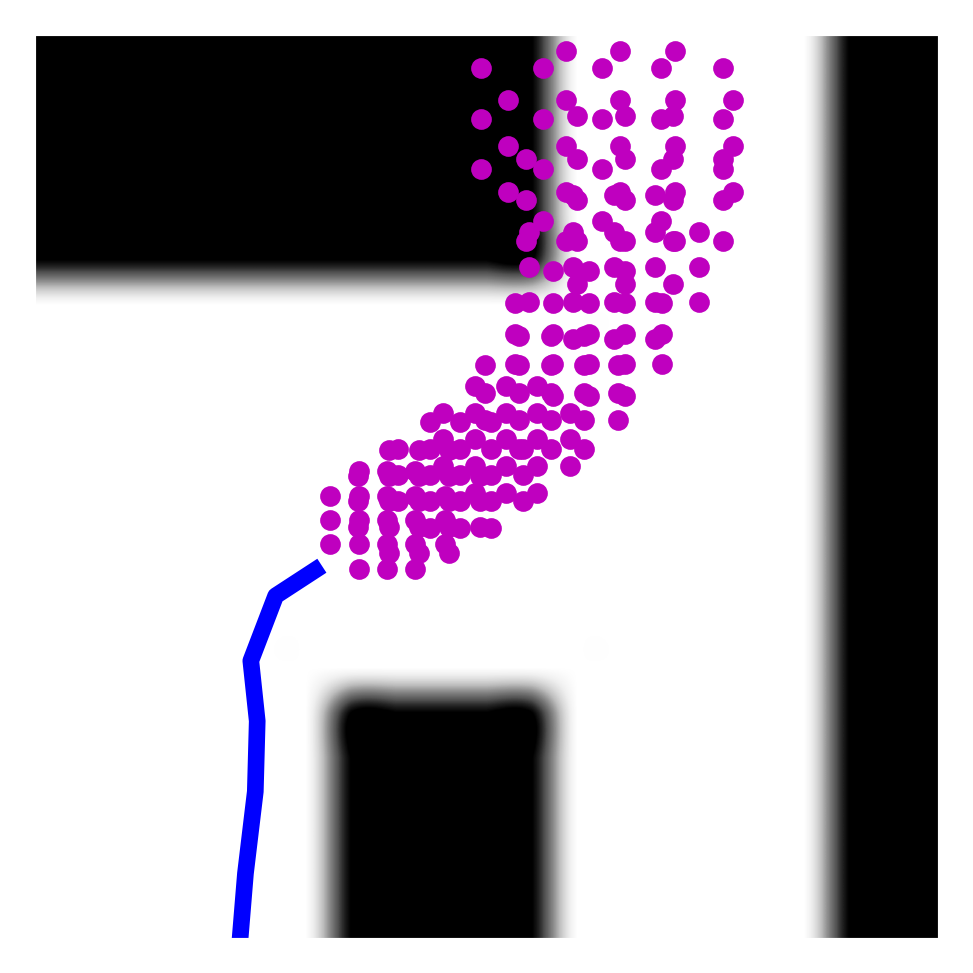}

    \end{subfigure}%
    \begin{subfigure}[]{0.195\textwidth}
        \centering
        \includegraphics[width=\textwidth]{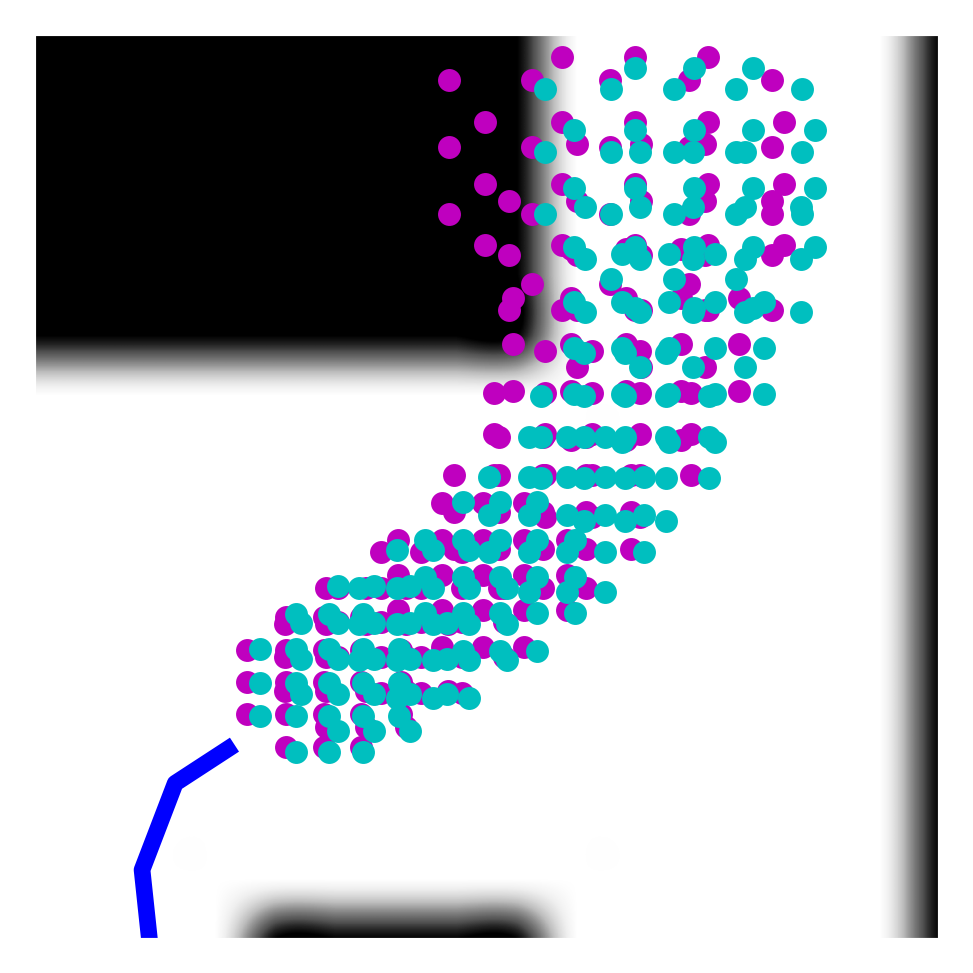}

    \end{subfigure}%
    \begin{subfigure}[]{0.195\textwidth}
        \centering
        \includegraphics[width=\textwidth]{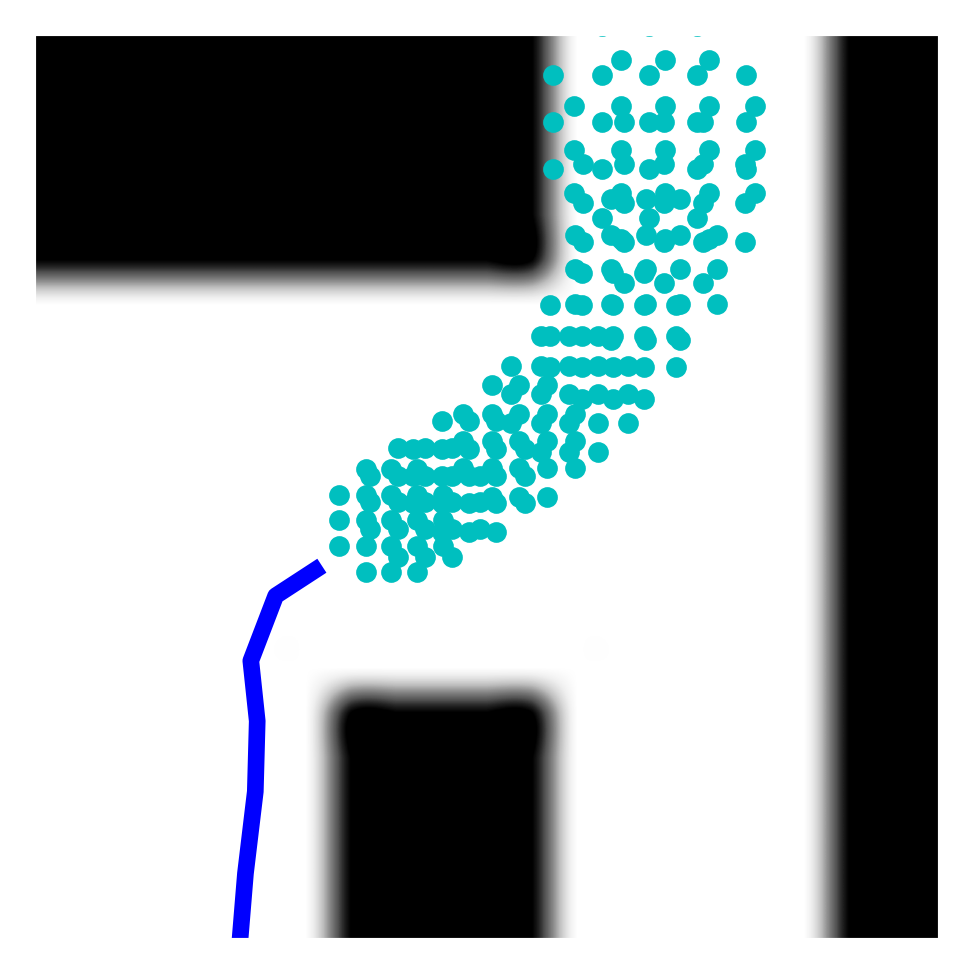}

    \end{subfigure}%
    \begin{subfigure}[]{0.195\textwidth}
        \centering
        \includegraphics[width=\textwidth]{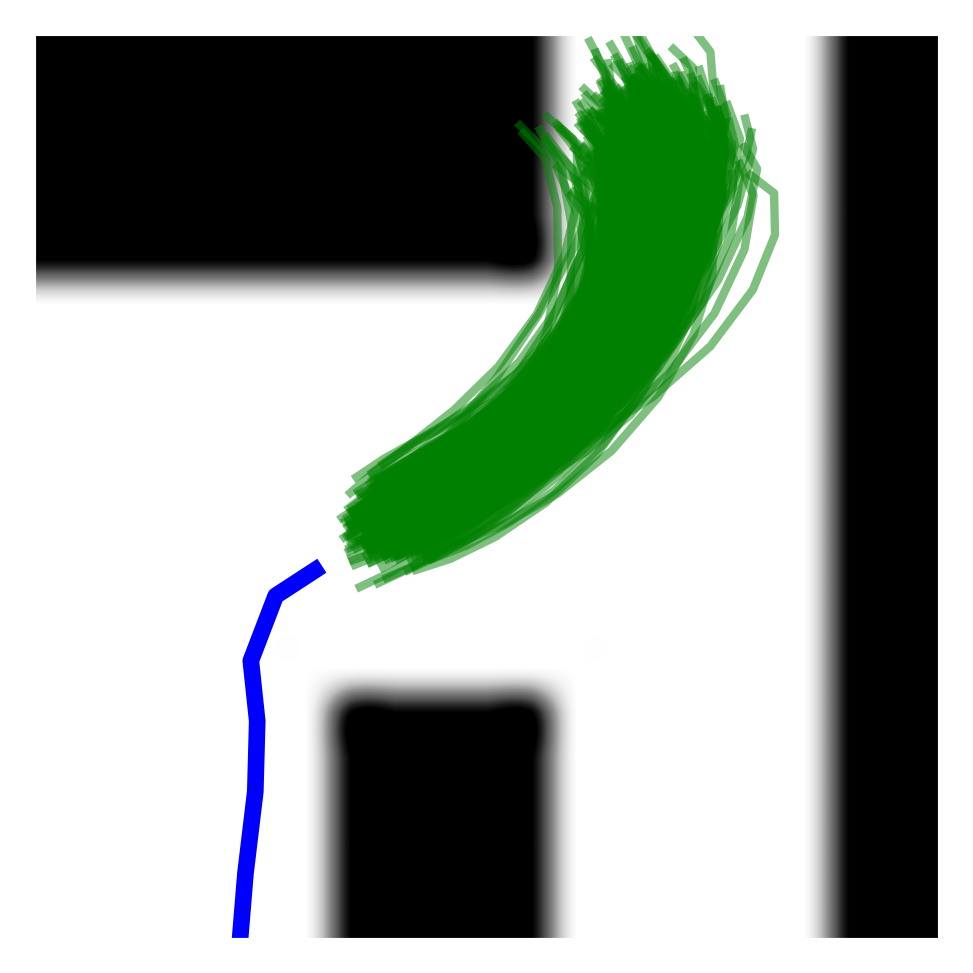}

    \end{subfigure}%

    \caption{An example of an originally non-compliant prior distribution of trajectories becomes more constraint compliant after the constraint enforcement. Left to right: (1) Non-compliant distribution of trajectories; (2) Abscissae of constraint violating distribution; (3) Abscissae of before and after constraint enforcement (zoomed in); (4) Abscissae of constraint compliant distribution; (5) Compliant distribution of trajectories.}\label{example_opt}
\end{figure}

\begin{figure}[h]
\centering
        \begin{subfigure}[]{0.45\textwidth}
        \centering
        \includegraphics[width=\textwidth]{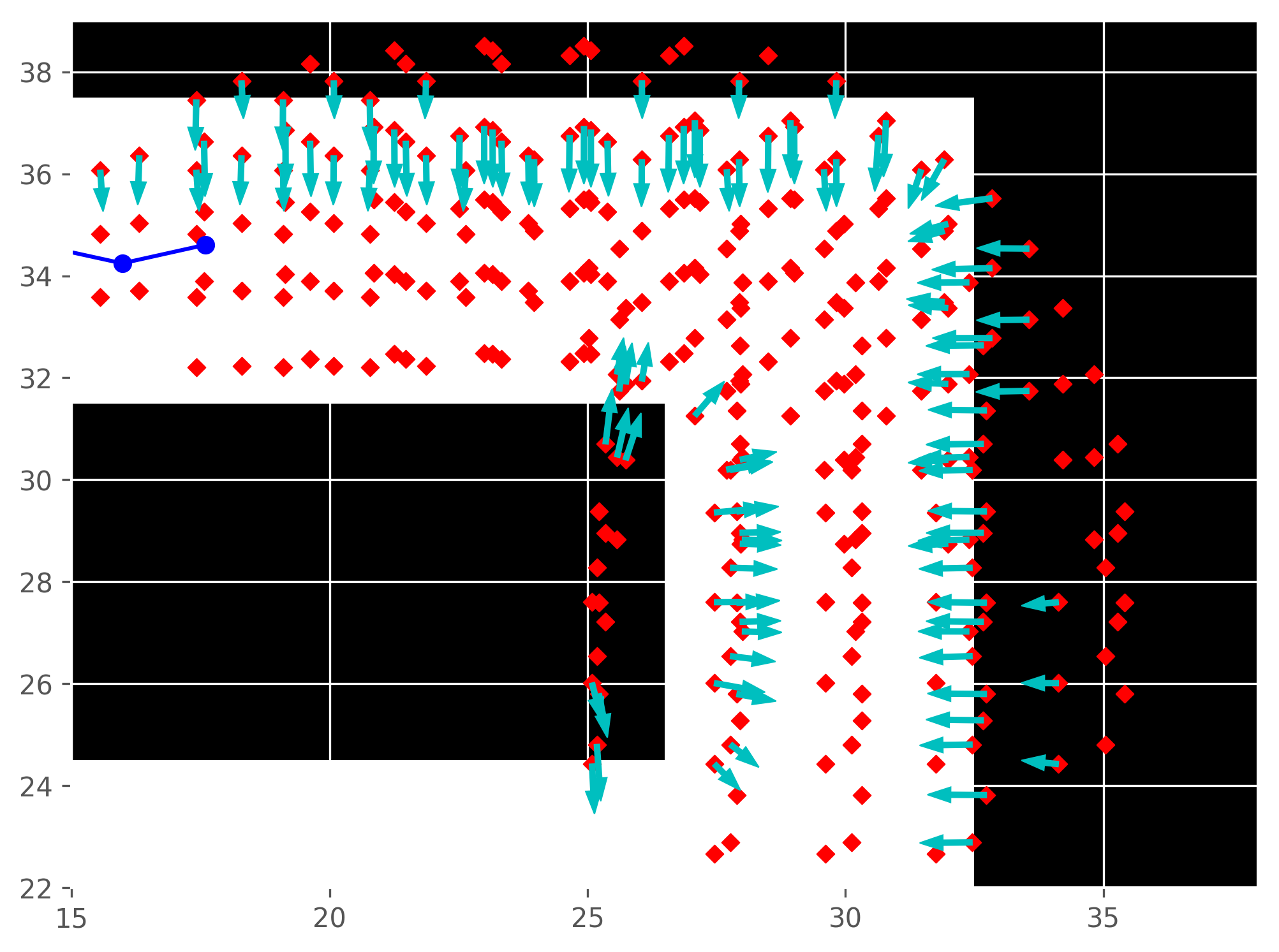}
    \end{subfigure}%
    \hspace{2mm}
    \begin{subfigure}[]{0.45\textwidth}
        \centering
        \includegraphics[width=\textwidth]{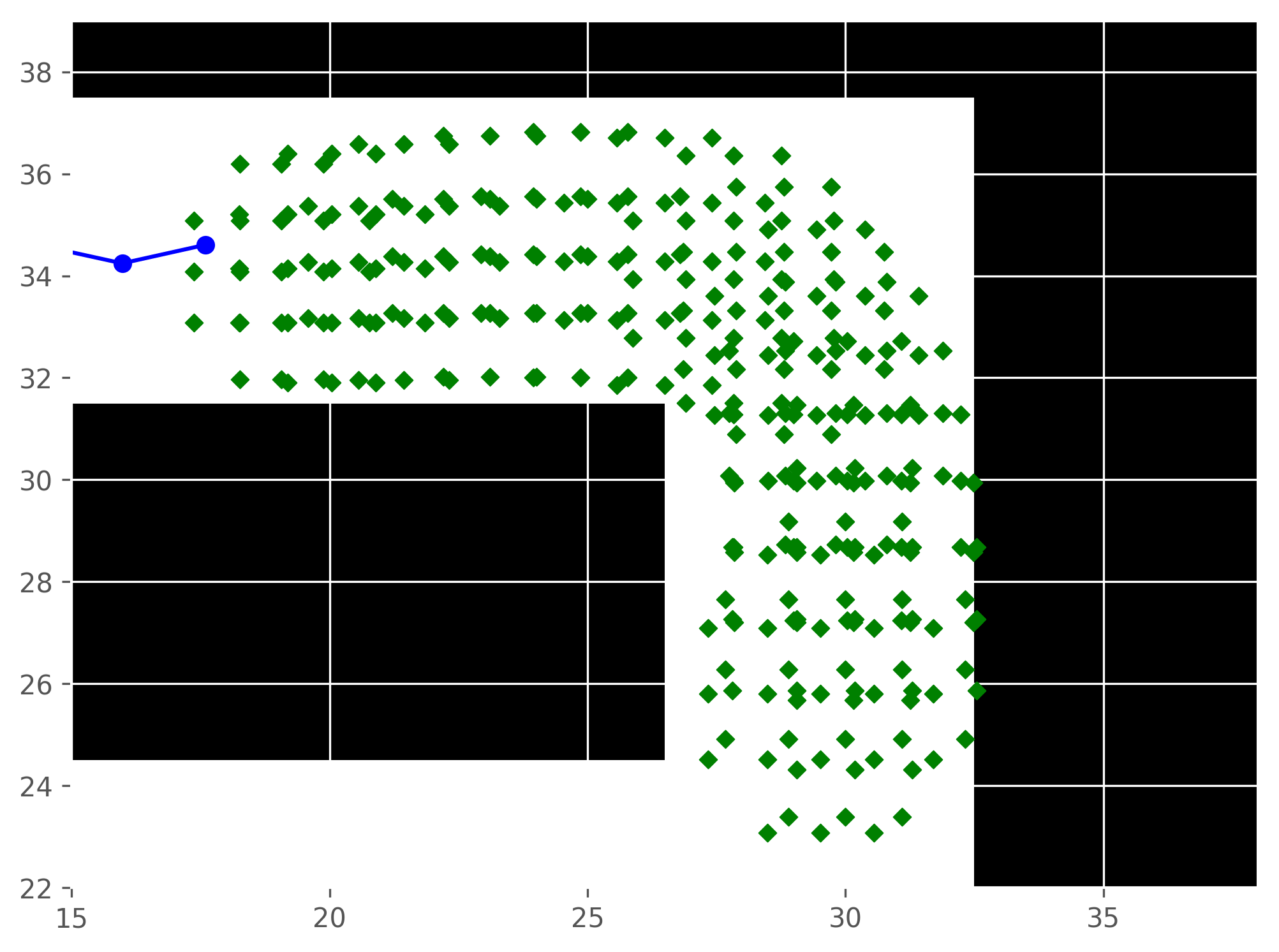}
    \end{subfigure}%
    
    \begin{subfigure}[]{0.45\textwidth}
        \centering
        \includegraphics[width=\textwidth]{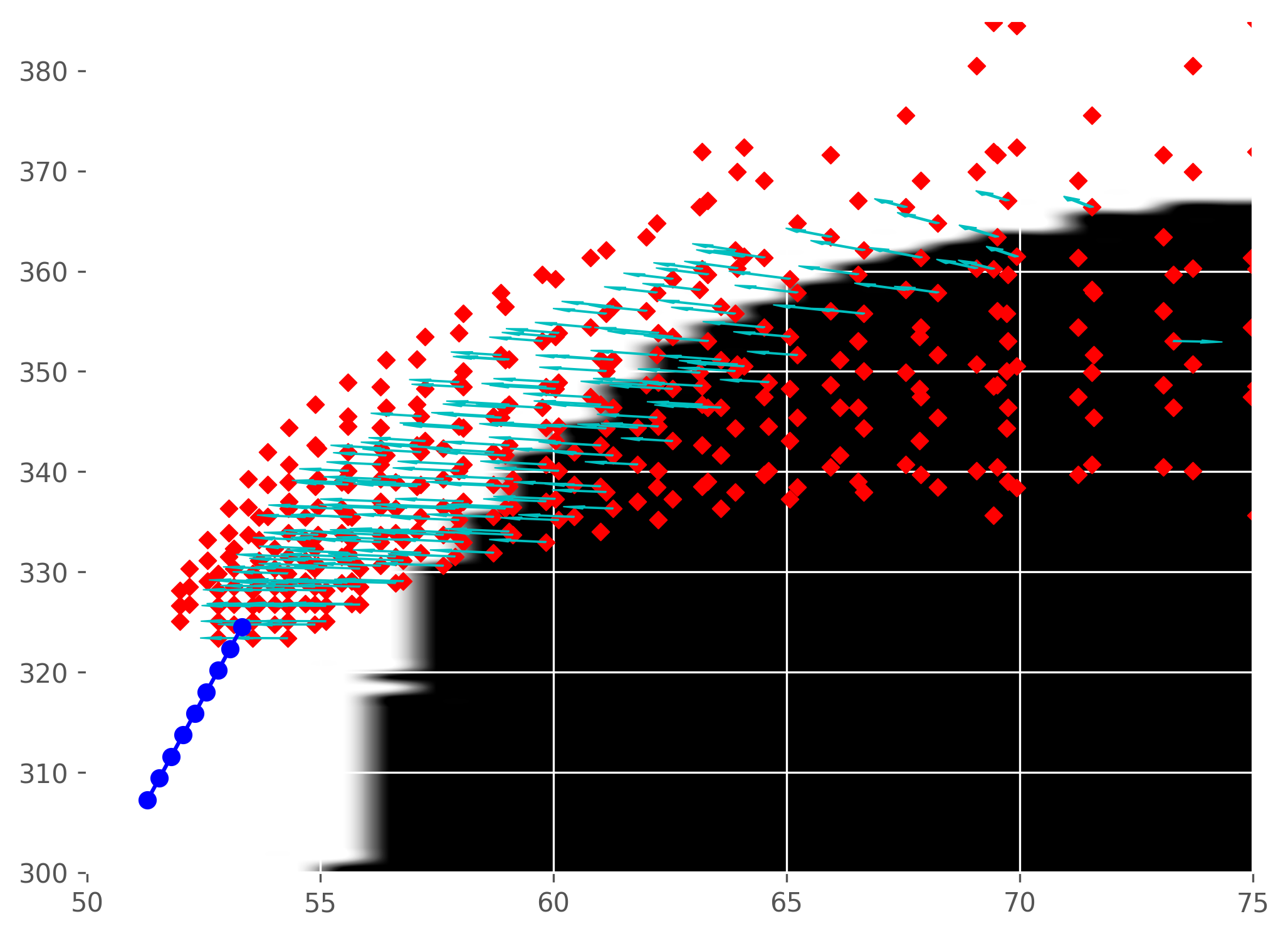}
    \end{subfigure}%
    \hspace{2mm}
     \begin{subfigure}[]{0.45\textwidth}
        \centering
        \includegraphics[width=\textwidth]{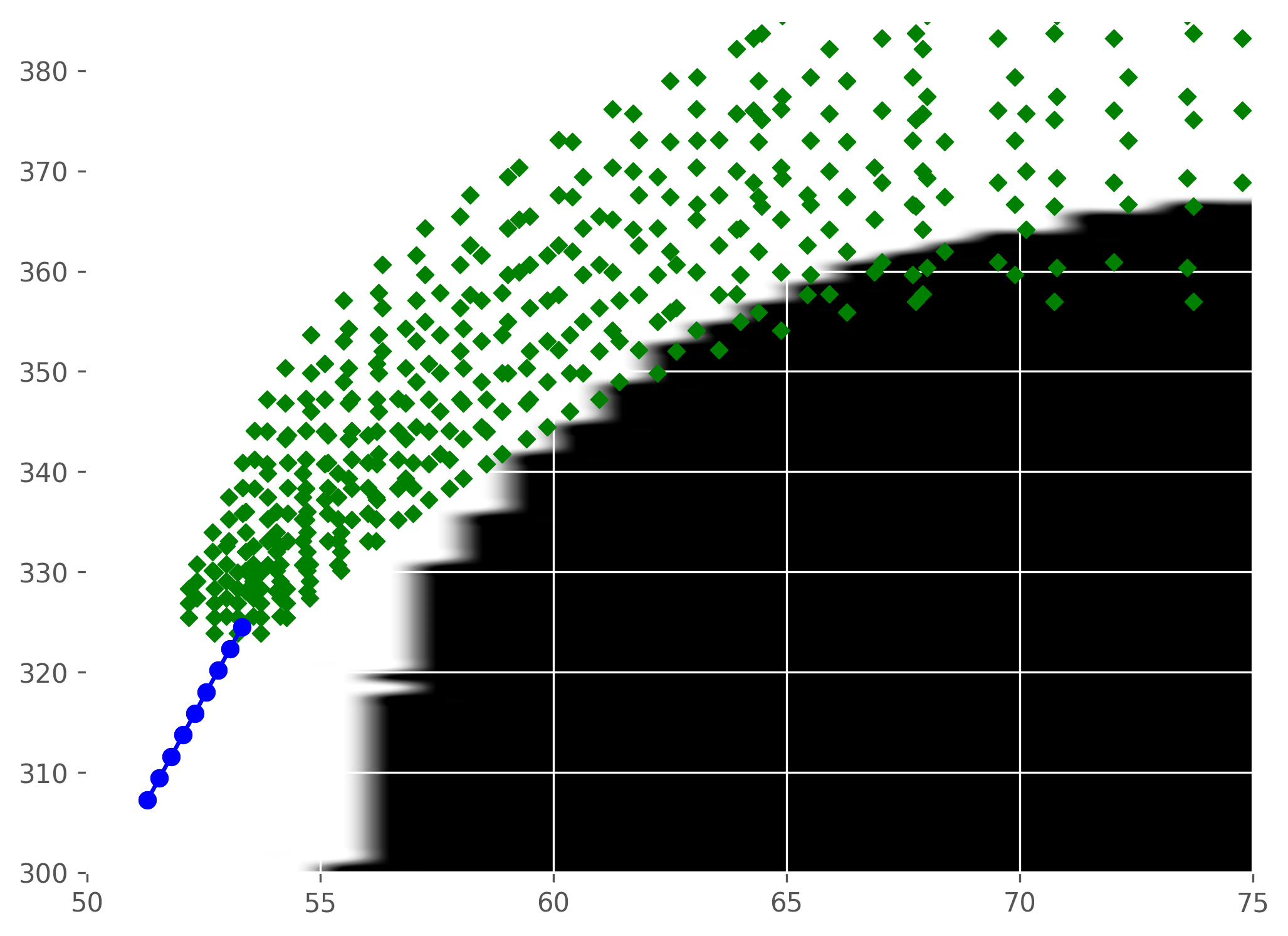}
    \end{subfigure}%
    
     \begin{subfigure}[]{0.45\textwidth}
        \centering
        \includegraphics[width=\textwidth]{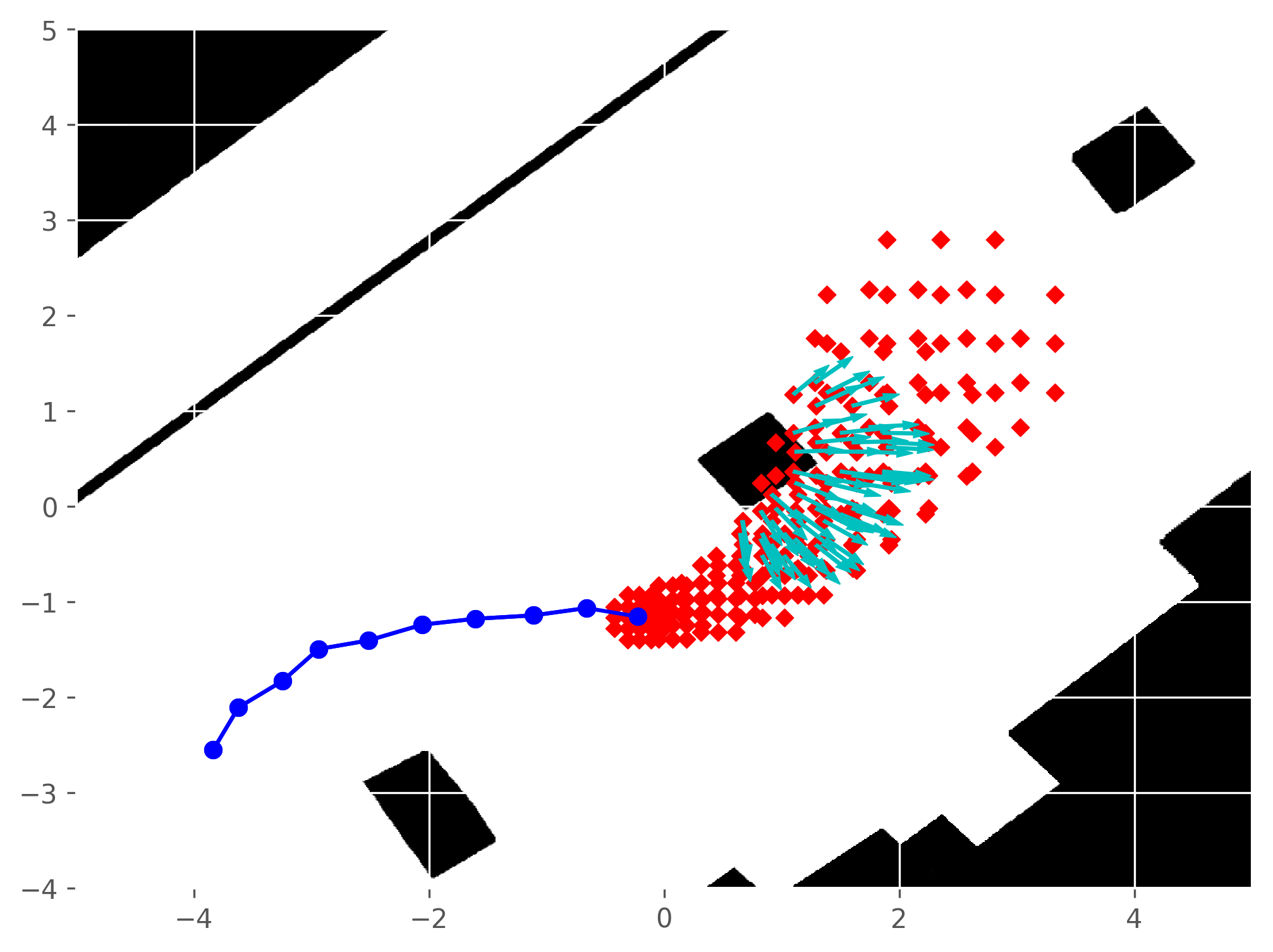}
    \end{subfigure}%
    \hspace{2mm}
     \begin{subfigure}[]{0.45\textwidth}
        \centering
        \includegraphics[width=\textwidth]{figures/chap6/c_obs_arr_solved4.png}
    \end{subfigure}%
    \caption{Examples from each dataset with the optimisation to apply constraints. The left column shows abscissae points of the main distribution component (component with the largest weight), with notable gradients with respect to occupancy shown as arrows (cyan) at the abscissae. The right column shows the same abscissae points after optimisation. We see that the constrained distribution subtly conforms better to structures in the environment, while still closely resembling the learned trajectory distribution. The left, middle and right sub-figures are examples from the simulated, indoors, and traffic datasets respectively.}\label{xyTraj}
\end{figure}

Qualitative results are shown in \cref{xyTraj} and \cref{example_opt}. We see the abscissae points of the largest trajectory distribution component before (top) and after (bottom) the trajectory optimisation of examples in each dataset. Our framework is able to ensure that the distributions comply with environmental structure. Notably, we see that the trajectory optimisation not only adjusts the mean of the trajectory distribution, but also optimises the variances, giving more robust trajectory distributions. This effect is most prominent in the left sub-figure, where the predicted distribution of trajectories in a simulated hallway recovers a much tighter variance which follows the environment structure. We also observe instances of collision-avoidance with objects in open space (demonstrated in the middle sub-figure), as well as increased adherence to road structure (right sub-figure). 

\FloatBarrier


  
\section{Summary}
We introduced a novel framework to learn motion predictions from observations while conforming to constraints, such as obstacles. Our framework consists of a learning component which learns a distribution over future trajectories. This distribution is used as a prior in a trajectory optimisation step which enforces chance constraints on the trajectory distribution. We empirically demonstrate that our framework can learn complex trajectory distributions and enforce compliance with environment structures via optimisation. This results in reduced variance and the avoidance of obstacles by the predicted trajectories, leading to improved prediction quality.

We shall now move to \cref{part3}, and explore the usage of learning in the generation of motion trajectories for manipulator arm robots. Commercially-available robot manipulators have much higher degrees-of-freedom relative to wheeled ground robots and present a different set of challenges to researchers. In the upcoming \cref{chap7}, we shall contribute a framework which allows robots to learn from expert demonstrations, and generalise the knowledge when the surroundings change.
\pagebreak

\part{Learning for Robot Manipulator Motion Generation}\label{part3}
\chapter{Diffeomorphic Transforms for Generalised Imitation Learning}\label{chap7}\blfootnote{This chapter has been published in L4DC as \cite{Diff_templates}.}
\tikzstyle{decision} = [diamond, draw, fill=blue!20, 
    text width=6em, text badly centered, node distance=3cm, inner sep=0pt]
\tikzstyle{block} = [rectangle, draw, 
    text width=8.5em, text centered, rounded corners, minimum height=3.2em]
\tikzstyle{block2} = [rectangle, 
    text width=10em, text centered,  minimum height=3.2em]
\tikzstyle{line} = [draw, -latex']
\tikzstyle{cloud} = [draw, ellipse,fill=red!20, node distance=3cm,
    minimum height=2em]
\tikzset{square arrow/.style={to path={-- ++(0,-0.25) -| (\tikztotarget)}}}  

\section{Introduction}
Imitation learning, or learning by demonstration, is an approach where robots are taught to execute novel motions from a few human expert demonstrations. Such an approach is natural and intuitive for generating complex motions, as it circumvents hand coding movements or specifically designing control costs \citep{ilbook}. Crucial to imitation learning is a generative model mimicking the provided demonstrations, as well as methods to reproduce motions in novel situations. Generative models for imitation learning can typically be described as having a time-dependent or state-dependent representation. In this chapter, we focus on the state-dependent category, where motions are generated by integrating a time-invariant non-linear dynamical system.

A major challenge of representing robot motion with a state-dependent dynamical system is the need to ensure the system's stability. Various methods \citep{SEDS,Saveriano2020AnEA,Euclideanising,urain2020imitationflows} have been developed to learn stable dynamics. In particular, recent methods such as \cite{Euclideanising,urain2020imitationflows} attempt to construct inherently stable dynamical systems by learning a smooth bijective mapping (i.e. a diffeomorphism) linking a known stable system and a desired target system, using invertible neural networks. Although these methods are capable of learning stable systems to accurately reproduce demonstrated examples, it is difficult to gauge how the neural network deforms the known stable system. The opaqueness of the network offers little insight into how to inject additional knowledge to alter the system upon encountering a change in the environment. 

\begin{figure}[t]
 \setlength{\fboxsep}{0pt}%
\setlength{\fboxrule}{1pt}%

\centering
        \fbox{\includegraphics[width=.99\linewidth]{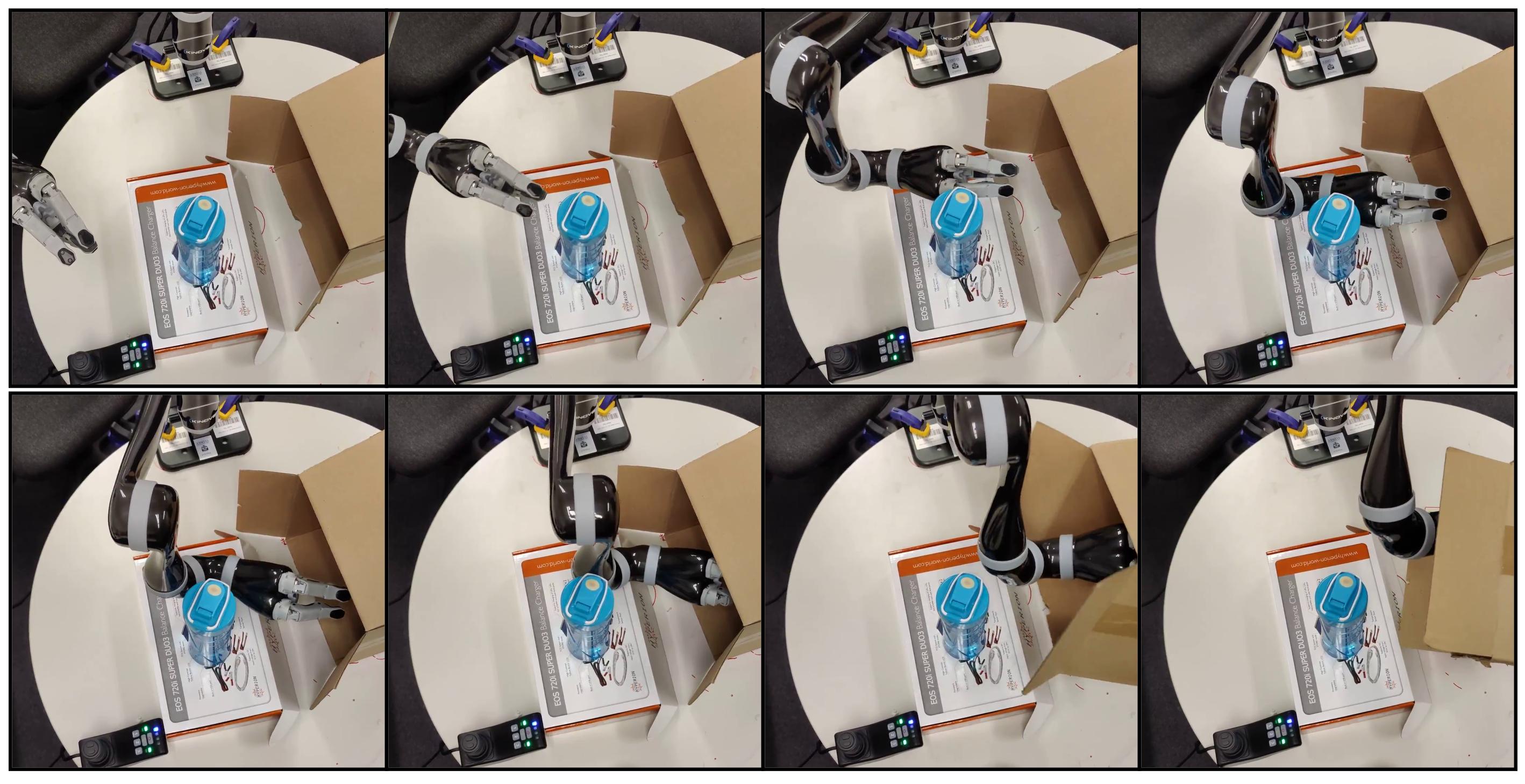}}
\caption{A motivating application for our work is generalising stable systems learned from expert demonstrations, when surroundings change. Here, a 6 degrees-of-freedom JACO arm is shown 8 recorded human demonstrations. Then, an obstacle bottle is placed right in the path of the recorded demonstrations. The JACO arm is able to warp smoothly around obstacle and successfully reach in and lift the box. }
    \label{fig:Jaco}

\end{figure}

Furthermore, prior work often focuses solely on learning a stable representation to imitate demonstrations, and struggles to operate in different environments where no new demonstrations are provided. A specific example is described in \cref{fig:Jaco}, where additional obstacles cause the previous demonstrations to be in collision. We contribute the diffeomorphic transforms framework for generalised imitation learning. We place emphasis on the ability of the system to generalise when surroundings change, by factoring in other sources of information, such as occupancy data or additional user specification. We represent each behaviour, such as reproducing demonstrations and obstacle avoidance, as individual diffeomorphic transforms. Each transform can be built in various ways, such as learning from demonstrated examples, learning from occupancy data, or be hand-crafted. By combining diffeomorphic transforms, generated motions can not only be learned from demonstrations, but also be adjusted based on changes in the environment, such as avoiding new obstacles or warping towards a designated point, all while conserving the stability of the resulting system. 

\section{Related Work}
Imitation learning is an approach that replicates motion based on expert demonstrations, allowing human operators to transfer skills to robot systems. One class of methods in imitation learning represent robot motions as time-dependent functions, these include methods proposed in \cite{promp,kmp,Kulak2020FourierMP}. Another class of methods represents robot motion by a stable state-dependent dynamical system. Such a representation is more robust to perturbations in the starting position, particular in regions with no data, as the system is guaranteed to stabilise at some point. A pioneer example of a stable state-based representation is SEDS \citep{SEDS}, which selects a candidate from a family of quadratic Lyapunov functions and is used to constrain the learned system via sequential quadratic programming. However, the expressiveness of SEDs is restricted by being limited to candidates from the quadratic Lyapunov functions family. Further work in \cite{SEDS2} expands the family of Lyapunov functions for candidates. 

Recent approaches on learning stable systems often take a different strategy. Instead of explicitly parameterising and learning the class of stable systems, these approaches learn smooth bijective mappings, diffeomorphisms, which map between a stable base system and some target. The resulting dynamical system can be thought of as a ``morphing'' of the base system, and preserves the stability properties of the original system. This approach could be thought of as analogous to normalising flows \citep{normalisingflows} for density estimation, which utilises invertible mappings to map a known probability distribution to the target distribution. The methods outlined in \cite{Neumann2015LearningRM,urain2020imitationflows,Euclideanising} take this approach. Particularly, \cite{Euclideanising} interprets the target system as a stable natural gradient system \citep{natural} on a manifold with an associated metric provided by the diffeomorphism. Our work takes a similar approach, by using diffeomorphisms to transform a stable base system, we can ensure the resulting system is stable in a principled manner.

Despite the advancements made in imitation learning, there have been relatively few works addressing adapting imitations to new environments. \cite{TPGMM} propose to adjust parameters of Gaussian mixture models on based tasks, while the authors of \cite{clamp} make attempts to alter generated trajectories, by post-processing with motion planning. However, these method are restricted to time-dependent trajectories, rather than state-dependent systems. In particular, the method in \cite{clamp} is restricted to discrete-time trajectories. Our approach is able to learn stable dynamical systems, and generalise the learned stable systems based on environment occupancy and user specifications. 

\section{Preliminaries}
\subsection{Motion Representation}


Robot motion is represented as a mapping between the robot state $\bm{x}\in\mathbb{R}^{d}$ and its velocity $\dot{\bm{x}}\in\mathbb{R}^{d}$. We model the motion as a time-invariant non-linear first-order dynamical system:
\begin{align}
    \dot{\bm{x}}(t)=f(\bm{x}(t)),&& \bm{x}(0)=\bm{x}_{0}, \label{syst}
\end{align}
where $f:\mathbb{R}^{d}\rightarrow\mathbb{R}^{d}$ is a non-linear mapping, and $\bm{x}_{0}$ is the initial state. A trajectory of the system is given by the solution of \cref{syst}, i.e. $\xi(t, \bm{x}_{0})=\bm{x}_{0}+\int_{0}^{t}\dot{\bm{x}}(s)\mathrm{d}s$. Additionally, like many other works in this area \citep{Neumann2015LearningRM, urain2020imitationflows, Euclideanising}, we require the motion to be \emph{globally asymptotically stable}, as outlined in \cref{bk_asymtotically_sec}. Stable systems behave in an expected manner during robot execution, and have vastly improved generalisation capabilities, as demonstrated in \cite{stable_vect_field_example}.



\subsection{Imitating demonstrations}\label{imitateSub} 
Provided a set of $N$ demonstrations, $\Xi$, where the $i^{th}$ demonstration is given by a sequence of states, of length $l_{i}$, i.e. $\Xi=\{\{\bm{x}_{i,1},\bm{x}_{i,2},\ldots,\bm{x}_{i,l_{i}}\}\}_{i=1}^{N}$. The velocities, $\dot{\bm{x}}_{i,j}$, are assumed to be available. To imitate the demonstrations, we aim to learn an $f(\bm{x})$ such that trajectories generated using \cref{syst} match the empirical demonstrations $\Xi$. Assuming that $\bm{\theta}$ parameterises $f$, we can optimise $\bm{\theta}$ in a least squares manner:
\begin{align}
    \min_{\bm{\theta}}\sum_{i=1}^{N}\sum_{j=1}^{l_{i}}\lvert\lvert\dot{\bm{x}}_{i,j}-f_{\bm{\theta}}(\bm{x}_{i,j})\lvert\lvert_{2}^{2}, &&
    \text{s.t. $\dot{\bm{x}}=f_{\bm{\theta}}(\bm{x})$ is stable.} \label{min_opt}
\end{align}

\subsection{Generalising the Learned System} 
After optimising for parameters $\bm{\theta}$, we can generate trajectories from $f_{\bm{\theta}}$ which imitate the provided demonstrations. Beyond simply mimicking the provided demonstrations, we are also interested in modifying $f_{\bm{\theta}}$ to satisfy further requirements, particularly when the situation changes and no further demonstrations are provided. A motivating example is when the position of objects in the environment changes, resulting in the original demonstrations no longer being collision-free. With no access to new demonstrations but only the environment's occupancy, many open questions exist: How do we make use of occupancy data to update the system such that generated trajectories avoid collision and imitate demonstrations while enforcing the stability of the system? Furthermore, how should we modify the system such that generated trajectories incorporate more general additional user specifications, such as a bias towards a certain direction?  


\subsection{Preserving Stability with Diffeomorphisms} \label{diffintro}
An elegant method, as presented in \cite{Euclideanising}, of ensuring the stability of $\dot{\bm{x}}=f(\bm{x})$, is by viewing the target system as a \emph{natural gradient system} defined on some $d$-dimensional manifold $\boldsymbol{M}$, where integral curves of the target system on $\boldsymbol{M}$ correspond to integral curves of another known simple system in Euclidean space. A bijective mapping is a \emph{diffeomorphism} if $\psi$ and $\psi^{-1}$ are continuously differentiable. A diffeomorphism $\psi:\boldsymbol{M}\rightarrow \mathbb{R}^{d}$ maps points on the manifold to Euclidean space.

Given a differentiable potential function, $\Phi:\mathbb{R}^{d}\rightarrow\mathbb{R}$, and diffeomorphism $\psi$, and following \cite{Euclideanising}, we can define the \emph{natural gradient descent} system \citep{natural} on $\boldsymbol{M}$, corresponding to the system $\dot{\bm{z}}=\nabla_{\bm{z}}(\bm{z})$ defined in Euclidean space as:
\begin{equation}
    \dot{\bm{x}}:=f(\bm{x})=-\bm{G}(\bm{x})^{-1}\nabla_{\bm{x}}\Phi(\psi(\bm{x})),\label{naturalDesc}
\end{equation}
where $\bm{G}:\boldsymbol{M}\rightarrow\mathbb{R}^{d\times d}_{++}$ is a real positive definite matrix \emph{Riemannian metric} which varies smoothly on $\boldsymbol{M}$. We take the metric as 

\begin{equation}
\mathbf{G}(\bm{x})=\bm{J}_{\psi}(\bm{x})^{\top}\bm{J}_{\psi}(\bm{x}), 
\end{equation}

where $\bm{J}_{\psi}(\bm{x})$ is the Jacobian of $\psi$. 
\begin{theorem}[\citep{Euclideanising}]
Let $\Phi:\mathbb{R}^{d}\rightarrow\mathbb{R}$ be a potential function, and $\psi:\boldsymbol{M}\rightarrow \mathbb{R}^{d}$ be a diffeomorphism mapping between some manifold $\boldsymbol{M}$ to $\mathbb{R}^{d}$ Euclidean space. If the system defined by $\dot{\bm{z}}=-\nabla_{\bm{z}}\Phi(\bm{z}) \in \mathbb{R}^{d}$ is globally asymptotically stable, then the system defined by the natural gradient descent in \cref{naturalDesc} is globally asymptotically stable.
\end{theorem}

Building inherently stable systems can be done in a principled fashion by specifying a stable potential function $\bm{\Phi}$ in Euclidean space, and constructing diffeomorphisms, $\psi$, such that the natural gradient system of $\bm{\Phi}$, as defined in \cref{naturalDesc}, matches a desired \emph{target system}. The problem at hand becomes how to construct diffeomorphisms such that the natural gradient system of a known stable system exhibits the desired behaviours.
 
\section{Diffeomorphic Transforms}

In this section, we introduce diffeomorphic transforms (DTs) as a tool to generalise imitation learning. Diffeomorphic Transforms are diffeomorphisms which encode desired behaviour. By composing multiple DTs sequentially, we can build a stable system which is capable of incorporating various behaviours such as imitating demonstrations and generalising to novel environments. For example, by composing two DTs, one encoding the imitation of demonstrations and the other encoding obstacle avoidance, we can obtain a stable system that attempts to follow the given demonstrations while avoiding obstacles in the environment.

This section is structured as follows: We start by discussing how multiple DTs can sequentially be composed together (\cref{composition}). Next, we outline the general theory on which DTs are built and how to imbue DTs with desirable properties (\cref{infinitesimal-generatpr}). Thereafter, we derive DTs with the following behaviours: (i) imitating demonstration data (\cref{learnsection}), (ii) avoiding collision (\cref{collsion}), and (iii) biasing towards specified coordinates (\cref{bias}).      

\subsection{Composition of Diffeomorphic Transforms}\label{composition}
Diffeomorphisms provide smooth bijective mappings between manifolds. If we have two manifolds which are linked by a diffeomorphism and a system on one manifold, we can find a natural gradient system on the other manifold via \cref{naturalDesc}. In this chapter, diffeomorphisms map between manifolds of the same dimension and can be thought of as a change of coordinate systems. 

We denote diffeomorphisms by $\psi$ and manifolds by $\boldsymbol{M}$ with superscripts for specific diffeomorphisms or manifolds. Previous work \cite{Euclideanising,Ratliff_learninggeometric}, define forward diffeomorphisms $\psi$ as a mapping from some warped manifold to Euclidean space. We adopt this convention and define the diffeomorphisms $\psi$ of a DT as a mapping from some manifold which encodes desirable behaviour to one that does not. Correspondingly, the inverse $\psi^{-1}$ maps to a manifold with encoded behaviour. Compositions of DTs can be done in a sequential manner. Consider a simple system in Euclidean space linked by a DT to its natural gradient on some manifold, which in turn is linked to another natural gradient system on another manifold by another DT, and so forth. Each of the defined dynamical systems shares the same stability attributes as they are linked via a chain of diffeomorphic transforms.

A specific example of such a composition is given in \Cref{composition plot}, where we obtain a system which imitates demonstrations while also avoiding collisions. We start with a stable attractor in Euclidean space, with potential $\Phi(\bm{z})=\lvert\lvert \bm{z}-\bm{z}_{e}\lvert\lvert_{2}$. The DT, $\psi^{\text{i}}$, is learned from demonstration data, and maps from the $d$-dimensional manifold, $\boldsymbol{M}^{\text{i}}$ to Euclidean space. The natural gradient system 

\begin{align}
\dot{\bm{x}}=f_{i}(\bm{x})=-\bm{G}^{\text{i}}(\bm{x})^{-1}\nabla_{\bm{x}}\Phi(\psi^{\text{i}}(\bm{x})), && \bm{x}\in\boldsymbol{M}^{\text{i}} 
\end{align}

generates trajectories that mimic expert demonstrations. The DT $\psi^o$ is constructed from occupancy data, and maps from another $d$-dimensional manifold $\boldsymbol{M}^{\text{o}}$ to $\boldsymbol{M}^{\text{i}}$. The resulting natural gradient system $\dot{\bm{y}}=f_{\text{o}}(\bm{y})=-\bm{G}^{\text{o}}(\bm{y})^{-1}f_{i}(\psi^{\text{o}}(\bm{y}))$ thus encodes both the imitation and collision-avoidance properties.

Throughout the chapter, we denote system states in Euclidean space by $\bm{z}$, those on a manifold encoding demonstrated behaviour by $\bm{x}$, and those on a manifold which encodes generalisation behaviour, such as collision avoidance, by $\bm{y}$. The system in Euclidean space is always given as an attractor with potential $\Phi(\bm{z})=\lvert\lvert \bm{z} - \bm{z}_{e}\lvert\lvert_{2}$ in Euclidean space, with equilibrium point $\bm{z}_{e}\in\mathbb{R}^{d}$. Subscripts or superscripts of i, o, b, in our notation relate to the \emph{imitation}, \emph{obstacle-avoiding}, \emph{bias-incorporation} capabilities we wish to endow DTs with.

\begin{figure}[bt]
\begin{adjustbox}{width=0.95\textwidth,center} 
\fbox{\begin{tikzpicture}[node distance = 2cm]

    \node [block] (Simple){$d$-dimensional Euclidean space};
    \node [block, right of=Simple, node distance=8cm] (Imitate) {Manifold, $\boldsymbol{M}^{\text{i}}$, encoding imitating from examples};
    \node [block, right of=Imitate, node distance=8cm] (Generalise) {Manifold, $\boldsymbol{M}^{\text{o}}$, generalising to avoid obstacles};
    \node [block, above left of=Imitate, node distance=3.8cm and 1.5cm] (Examples) {Expert demonstrations};
    \node [block, above left of=Generalise, node distance=3.8cm and 1.5cm] (Occupancy) {Occupancy data};
    
    \node [block2, below of=Simple, node distance=2.cm] (Sys1) {$\dot{\bm{z}}=\nabla_{\bm{z}}\bm{\Phi}(\bm{z})$};
    \node [block2, below of=Imitate, node distance=2.cm] (Sys2) {$\dot{\bm{x}}=f_{i}(\bm{x})=-\bm{G}^{\text{i}}(\bm{x})^{-1}\nabla_{\bm{x}}\Phi(\psi^{\text{i}}(\bm{x}))$};
    \node [block2, below of=Generalise, node distance=2.25cm] (Sys2) {$\dot{\bm{y}}=f_{\text{o}}(\bm{y})=-\bm{G}^{\text{o}}(\bm{y})^{-1}f_{i}(\psi^{\text{o}}(\bm{y}))$};
    
    \path [line] (Imitate) -- node (psi_imitate) [text width=0.5cm,midway,above]{$\psi^{\text{i}}$}(Simple);
    \path [line] (Generalise) -- node  (psi_collision) [text width=0.5cm,midway,above]{$\psi^{\text{o}}$}(Imitate);
    
    \path [line] (Examples) -- (psi_imitate);
    \path [line] (Occupancy) -- (psi_collision);

\end{tikzpicture}}
\end{adjustbox}
\caption{To obtain a stable system capable of both imitating demonstrations and avoiding collisions, we compose two DTs: $\psi^{\text{i}}$ mapping from manifold $\boldsymbol{M}^{\text{i}}$ to Euclidean space, and $\psi^{\text{o}}$ from $\boldsymbol{M}^{\text{o}}$ to manifold $\boldsymbol{M}^{\text{i}}$. The natural gradient system of $\bm{\Phi}$ on $\boldsymbol{M}^{\text{i}}$ reproduces the expert demonstrations, while that on $\boldsymbol{M}^{\text{o}}$ avoids obstacles and imitates the demonstrations.}\label{composition plot}
\end{figure}

\subsection{Infinitesimal Generators of Flows}
\label{infinitesimal-generatpr}

In this section, we introduce the building block of diffeomorphic transforms: Smooth flows (integral curves) generated by infinitesimal generators (smooth vector fields):

\begin{definition}[Flows]
A (global) flow on $\mathbb{R}^{d}$ is a continuous mapping $\gamma:\mathbb{R}\times\mathbb{R}^{d}\rightarrow\mathbb{R}^{d}$, such that $\forall\bm{s} \in \mathbb{R}^{d}$ and $\forall t_{1},t_{2}\in\mathbb{R}$, there is
$\gamma(0,\bm{s})=\bm{s}$ and $\gamma(t_{2}, \gamma(t_{1},\bm{s}))=\gamma(t_{1}+t_{2},\bm{s})$.
\end{definition}

\begin{definition}[Infinitesimal Generator]
The infinitesimal generator, $V:\mathbb{R}^{d}\rightarrow\mathbb{R}^{d}$, of a smooth flow $\gamma$ is a smooth vector field mapping from each $\bm{s}\in\mathbb{R}^{d}$ to the tangent vector $\gamma'(0,\bm{s})$. 
\end{definition}


We shall use the following property of flows to build DTs: For every smooth infinitesimal generator $V$ where flows exist for all points in $\mathbb{R}^{d}$, the flow for time $t\in\mathbb{R}$, $\gamma(t,\cdot)$, is a diffeomorphism. Specifically, for any initial point $\bm{s}_{0}\in\mathbb{R}^{d}$ and time $t$, if $\bm{s}_{t}=\gamma(t,\bm{s}_{0})\in\mathbb{R}^{d}$, then $\bm{s}_{0}=\gamma(-t,\bm{s}_{t})$. That is, the inverse of $\gamma(t,\cdot)$ is unique and given by $\gamma(-t,\cdot)$. This is provided, with proof, in Theorem 9.12 of \cite{intro_geo}. 

Intuitively, this means that following a smooth vector field $V$ (infinitesimal generator) for time $t$ from $\bm s_0$ leads to a new location $\bm s_t$. To return to $\bm s_0$ from $\bm s_t$ requires the reversed vector field $-V$, which for an infinitesimal generator is equivalent to following the vector field $V$ for the negative amount of time, i.e. $-t$. This property provides us with a principled mechanism to construct a diffeomorphism: The smooth infinitesimal generator $V$ and its associated flow $\gamma$ form our diffeomorphism and its inverse, i.e.
\begin{align}
    \psi(\cdot) & = \gamma(t, \cdot) &
    \psi^{-1}(\cdot) & = \gamma(-t, \cdot)\;.
\end{align}

Therefore the following sections describe how to construct infinitesimal generators with different properties. Once constructed these infinitesimal generators form a flow which we use as diffeomorphic transforms which we combine to form complex yet stable motion generators.



\subsection{Learning Infinitesimal Generators from Demonstrations}\label{learnsection}

We outline a method to learn smooth infinitesimal generators which encode demonstrated motions.
We exploit the smoothness of Gaussian radial basis functions to parametrise the infinitesimal generator $V^{\text{i}}: \mathbb R^d \rightarrow \mathbb R^d$ with a weighted combination of basis functions,
\begin{equation}
    V^{\text{i}}(\bm{s})=\bm{W}^{\top}K(\bm{s},\hat{\bm{s}}),\label{vecfieldeqn}
\end{equation}
where $K(\bm{s},\hat{\bm{s}})=[k(\bm{s},\hat{\bm{s}}_{1}),k(\bm{s},\hat{\bm{s}}_{2}),\ldots k(\bm{s},\hat{\bm{s}}_{m})]^{\top}\in\mathbb{R}^{d\times m}$ is the projection of point $\bm{s}\in\mathbb{R}^{d}$ to $m$ pre-determined \emph{inducing states}, $\hat{\bm{s}}_{i}\in\mathbb{R}^{d}$, for $i=1,2,\dots,m$. The radial basis functions $k(\bm{s},\hat{\bm{s}}_{i})=\exp(-\ell\lvert\lvert\bm{s}-\hat{\bm{s}}_{i}\lvert\lvert_{2})$ are centered on each inducing state. The vector field $V^{\text{i}}$ is parameterised by weight matrix $\bm{W}\in\mathbb{R}^{d\times m}$. Like \cite{Euclideanising,urain2020imitationflows}, the policies are defined for world space position, and are typically coordinates in $\mathbb R^3$. The hyper-parameter $\ell$ of the radial basis function controls its ``width''. 


The flow on the infinitesimal generator is obtained by taking integral curves. As outlined in \cref{diffintro}, assume there exists a $d$-dimensional manifold $\boldsymbol{M}^{\text{i}}$, where trajectories follow demonstrations. Points on $\bm{x}\in \boldsymbol{M}^{\text{i}}$ are mapped to the corresponding point in Euclidean space, $\bm{z}$, by following the flow $\gamma^{\text{i}}$ on $V^{\text{i}}$ for time $t$. We define the diffeomorphism $\psi_{\bm{W}}^{\text{i}}:\boldsymbol{M}^{\text{i}}\rightarrow\mathbb{R}^{d}$ and its inverse as:
\begin{align}
    \psi_{\bm{W}}^{\text{i}}(\bm{x})&:=\gamma^{\text{i}}(t,\bm{x})=\bm{x}+\int_{0}^{t}V^{\text{i}}(\gamma^{\text{i}}(u,\bm{x}))\mathrm{d}u = \bm{z},\label{diffeo}\\
    \psi_{\bm{W}}^{\text{i}}(\bm{z})^{-1}&:=\gamma^{\text{i}}(-t,\bm{z})=\bm{z}+\int_{-t}^{0}V^{\text{i}}(\gamma^{\text{i}}(u,\bm{z}))\mathrm{d}u = \bm{x}.\nonumber
\end{align}
As analytical integrals are often difficult to obtain we use numerical integration techniques, Euler's method in our case due to its simplicity.

\begin{figure}[t]
\centering
\begin{subfigure}{.3\textwidth}
  \centering
  \includegraphics[width=\linewidth]{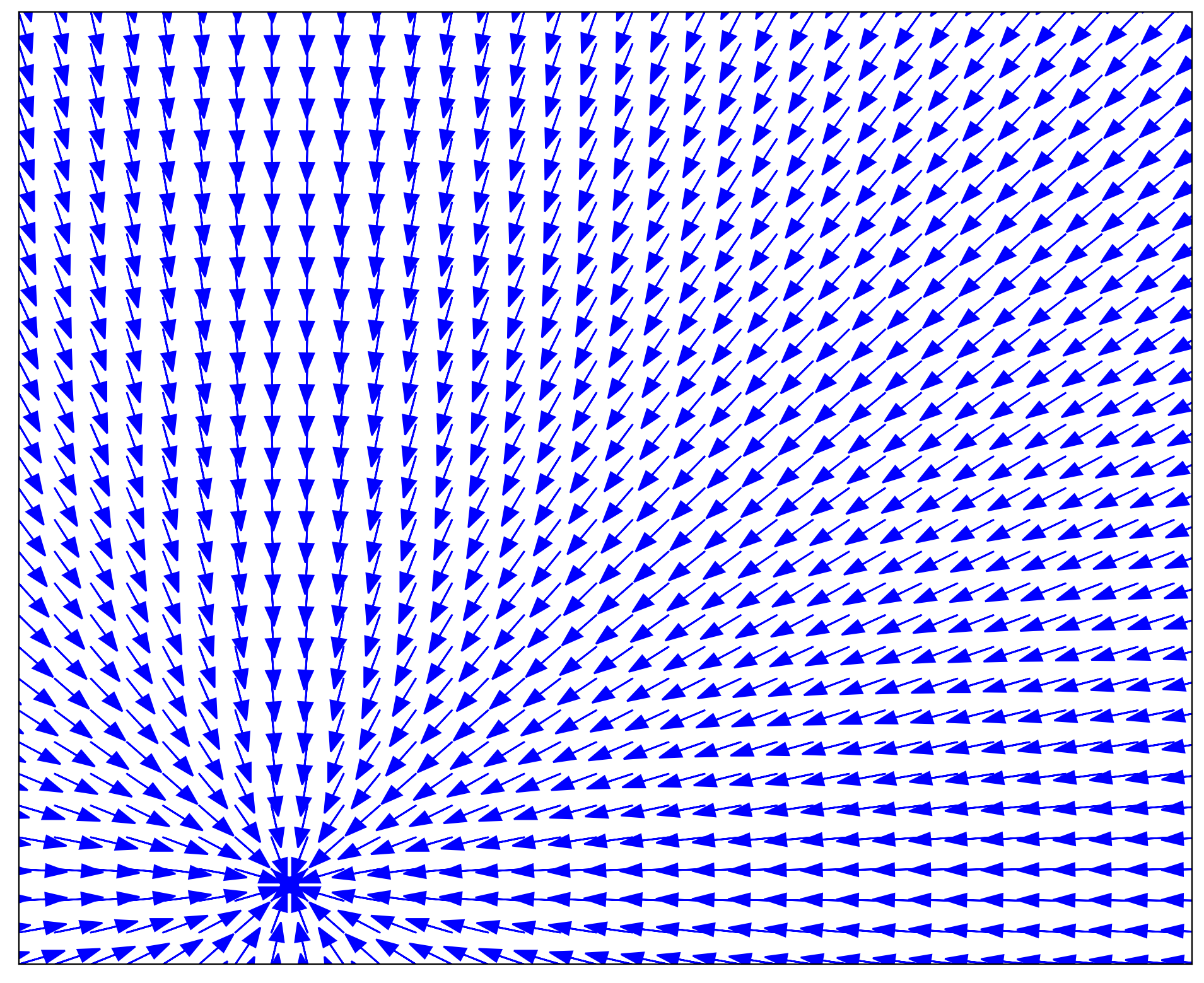}  
\end{subfigure}%
\hspace{1em}
\begin{subfigure}{.3\textwidth}
  \centering
  \includegraphics[width=\linewidth]{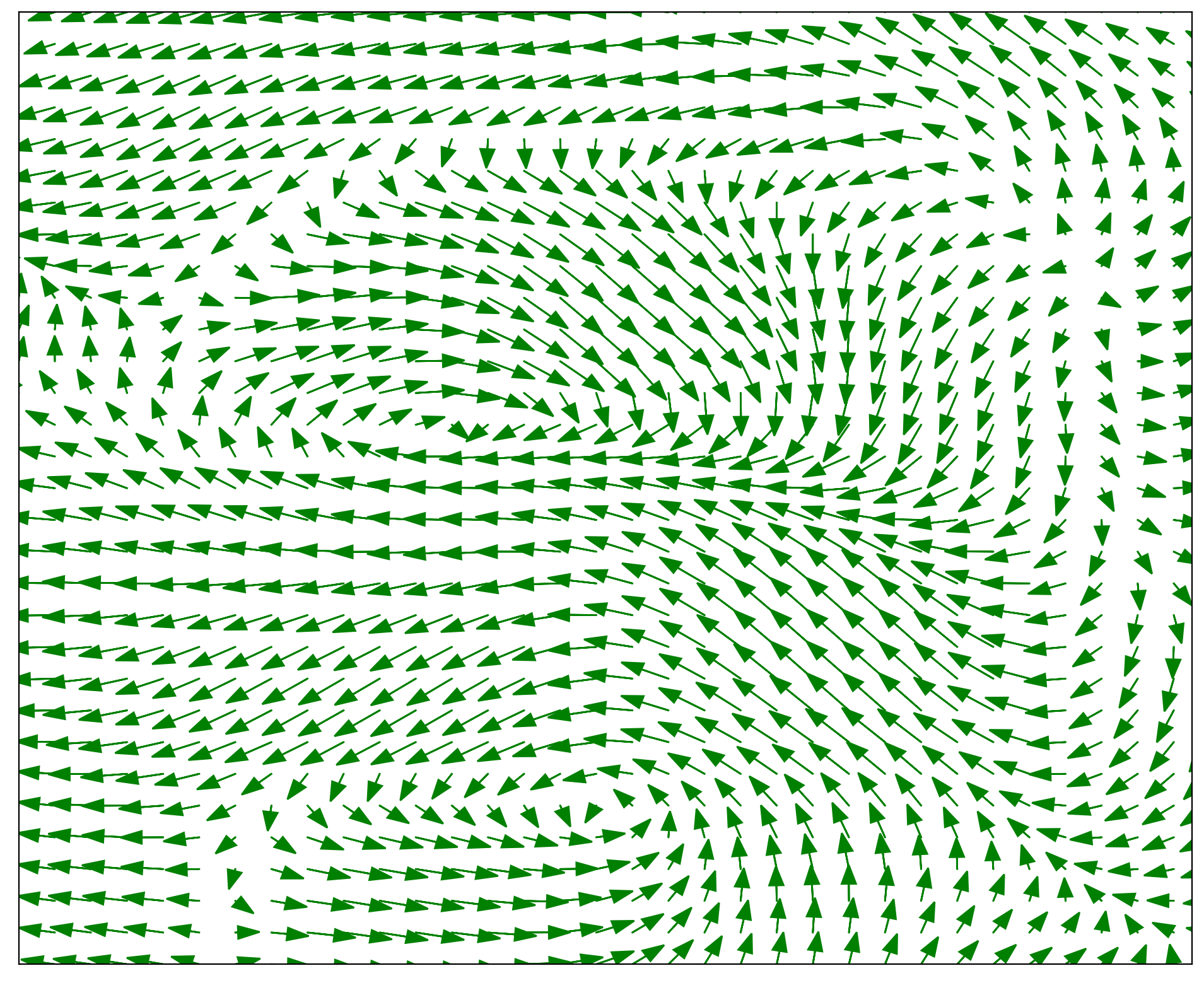}  
\end{subfigure}%
\hspace{1em}
\begin{subfigure}{.3\textwidth}
  \centering
  \includegraphics[width=\linewidth]{figures/chap7/S_vec_b.png}
\end{subfigure}%
\caption{Visualisation of vector fields, learning to generate an ''S'' motion. Left: a stable base system, a simple attractor; Center: the infinitesimal generator $V_{\bm{W}}^{\text{i}}$ learned on demonstrations of ``S''; Right: the stable system $\dot{\bm{x}}=f_{i}(\bm{x})$ imitating demonstrations, with black trajectory rollouts.}
\label{imitate}
\end{figure}

In the context of the imitating demonstrations, as outlined in \cref{imitateSub}, we are assumed to have $N$ demonstrations, each a sequence of states. That is, $\hat{\xi}_{i}=\{\bm{x}_{i,1},\bm{x}_{i,2},\ldots,\bm{x}_{i,l_{i}}\}$, where $l_{i}$ is the length of the $i^{th}$ demonstration. The state derivatives are also assumed to be available or obtained via finite differences. By \cref{diffeo,naturalDesc,min_opt}, we can formulate the optimisation problem as:
\begin{align}
    \min_{\bm{W}}\Big\{\sum_{i=1}^{N}\sum_{j=1}^{l_{i}}\lvert\lvert\dot{\bm{x}}_{i,j}+\bm{G}^{\text{i}}(\bm{x}_{i,j})^{-1}\nabla_{\bm{x}_{i,j}}\Phi(\psi_{\bm{W}}(\bm{x}_{i,j}))\lvert\lvert_{2}^{2}\Big\} \label{learn_cost}\\
    \bm{G}^{\text{i}}(\bm{x}_{i,j})=\bm{J}_{\psi_{\bm{W}}}(\bm{x}_{i,j})^{\top}\bm{J}_{\psi_{\bm{W}}}(\bm{x}_{i,j}).
\end{align}
We can optimise \cref{learn_cost} to obtain $\bm{W}$. By \cref{naturalDesc}, the natural gradient system which generates imitated trajectories, with $\bm{x}\in\boldsymbol{M}^{\text{i}}$, is then, 
\begin{equation}
\dot{\bm{x}}=f_{i}(\bm{x})=-\bm{G}^{\text{i}}(\bm{x})^{-1}\nabla_{\bm{x}}\Phi(\psi_{\bm{W}}^{\text{i}}(\bm{x})). \label{ds1}
\end{equation}


\subsection{Building Infinitesimal Generators for Collision Avoidance}\label{collsion}

Beyond learning to reproduce motions from demonstration data, we wish to adapt the generated motions so that they remain collision-free when the occupied space in the environment changes. In this section, we introduce diffeomorphic transforms which encode collision-avoiding behaviours. Again, we aim to generate a diffeomorphism by constructing an appropriate smooth infinitesimal generator.

We use smoothly varying obstacle gradients to build the collision-avoiding diffeomorphic transforms. These are obtained via a continuous occupancy representation, such as the representation described in \cite{HilbertMaps}. We exploit its continuous nature to find analytical gradients of the probability of being occupied. Using occupancy information about the new environment in the form of a training dataset of $n$ pairs, $\{(\bm{s}_{k},o_{k})\}_{k=1}^{n}$, where each sample $\bm{s}_{k}\in\mathbb{R}^{d}$ is a point in the state-space, and $o_{k}\in\{0,1\}$ is a binary variable indicating whether $\bm{s}_{k}$ is occupied or free. \cite{HilbertMaps} use a kernelised logistic regressor \cite{Bishop:2006} wrapped around Gaussian radial basis functions to learn a mapping between coordinates to the probability of them being occupied (occupancy). Likewise, we use the continuous occupancy representation introduced in \cite{HilbertMaps}. The continuous representation is given as $f^{\text{map}}(\bm{s}):=p(o=1|\bm{s})$, and is smooth as it consists of sigmoid and Gaussian functions. Unlike signed distance representations \citep{SDF}, the produced gradients of occupancy, w.r.t. coordinates, are smooth over the entire domain. We exploit the smooth obstacle gradients to construct an infinitesimal generator: 
\begin{equation}
    V^{o}(\bm{s})=\nabla_{\bm{s}}p(o=1\lvert \bm{s})=\nabla_{\bm{s}}f^{\text{map}}. \label{map}
\end{equation}

\begin{figure}[t]
\centering
\begin{subfigure}{.3\textwidth}
  \centering
  \includegraphics[width=\linewidth]{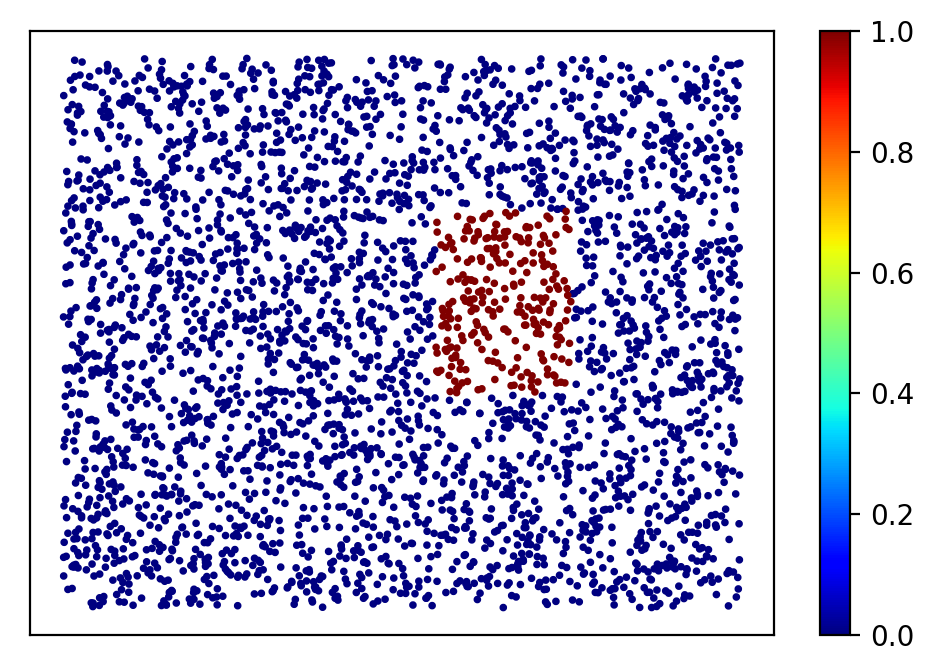}  
\end{subfigure}%
\begin{subfigure}{.3\textwidth}
  \centering
  \includegraphics[width=\linewidth]{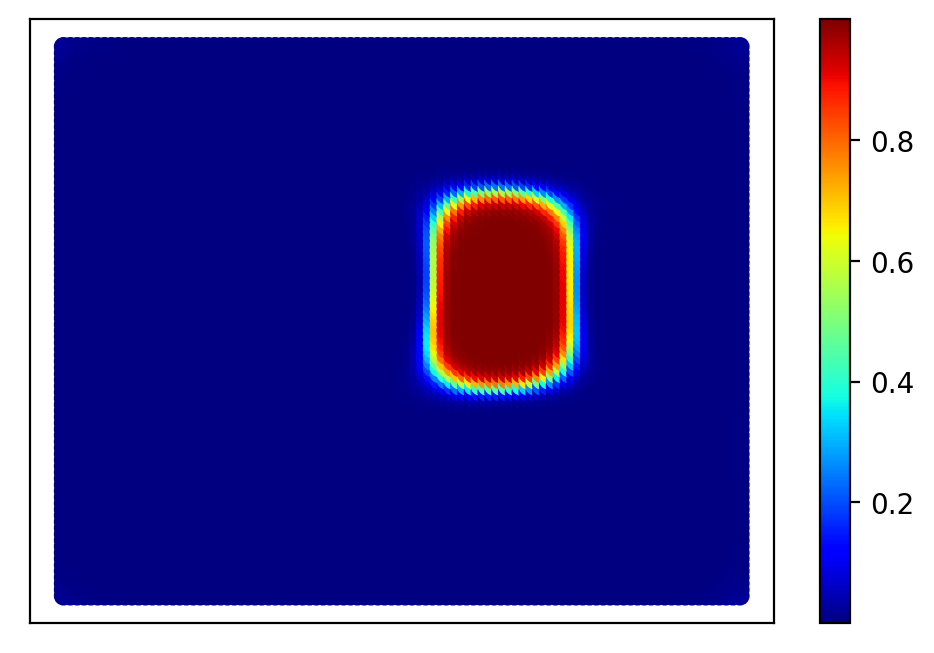}  
\end{subfigure}%
\begin{subfigure}{.3\textwidth}
  \centering
  \includegraphics[width=\linewidth]{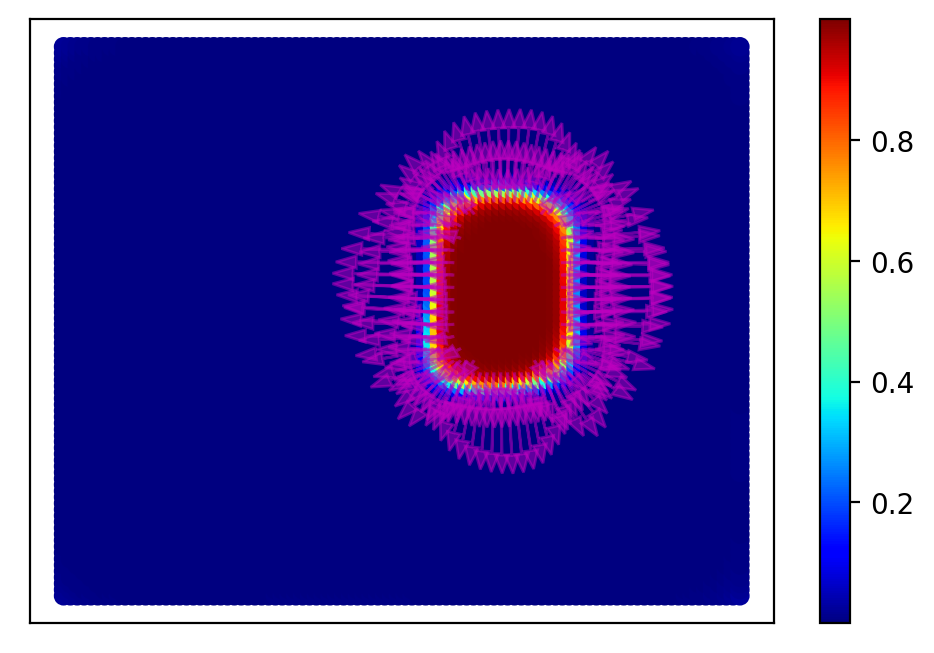}  
\end{subfigure}%
\caption{A 2d example of building a continuous occupancy representation. (Left) Occupancy data; (Centre) A continuous function mapping from coordinates to the probability of being occupied; (Right) Gradients of the probability of being occupied, wrt coordinates, are available and displayed in magenta.}
\label{simple2d}
\end{figure}

A simple example of a continuous occupancy representation from occupancy data is given in \cref{simple2d} (left and center). Gradient vectors with large negative values in $-V$ are shown in \cref{simple2d} (right). Following the flow on $V^{o}$ in reverse time reduces the observed occupancy, i.e. lower collision likelihood. This captures the goal of the inverse diffeomorphism ${\psi^{o}}^{-1}$, which is designed to map to a manifold encoding collision-avoiding behaviour. Accordingly, we define a $d$-dimensional manifold $\boldsymbol{M}^{\text{o}}$ where trajectories exhibit obstacle avoidance behaviour. 

Provided some manifold $\boldsymbol{M}$, the infinitesimal generator $V^{o}$ generates flow $\gamma^{o}$. For some $t\in\mathbb{R}$, let $\bm{y}\in \boldsymbol{M}^{\text{o}}$ and $\bm{x}\in \boldsymbol{M}$, we can define the diffeomorphism $\psi^{o}: \boldsymbol{M}^{\text{o}}\rightarrow\boldsymbol{M}$ as:
\begin{equation}
\psi^{o}(\bm{y}):=\gamma^{o}(t,\bm{y})=\bm{y}+\int_{0}^{t}V^{o}(\gamma^{o}(u,\bm{y}))\mathrm{d}u  = \bm{x}, 
\end{equation}
and its inverse as 
\begin{equation}
{\psi^{o}}(\bm{x})^{-1}:=\gamma^{o}(-t,\bm{x})=\bm{x}+\int_{-t}^{0}V^{o}(\gamma^{o}(u,\bm{x}))\mathrm{d}u = \bm{y}, 
\end{equation}
where a numerical integrator is used to evaluate the integral. The natural gradient system on $\boldsymbol{M}^{\text{o}}$ can then be defined as:
\begin{align}
    \dot{\bm{y}}&=-\bm{G}^{\text{o}}(\bm{y})^{-1}f(\psi^{o}(\bm{y})),
\end{align}
where $f$ corresponds to a dynamical system on $\boldsymbol{M}$. This could be for example, the original simple attractor or $f_{\text{i}}$ that encodes demonstrations. 

\begin{figure}[h]
\centering
\begin{subfigure}{.4\textwidth}
  \centering
  \includegraphics[width=0.95\linewidth]{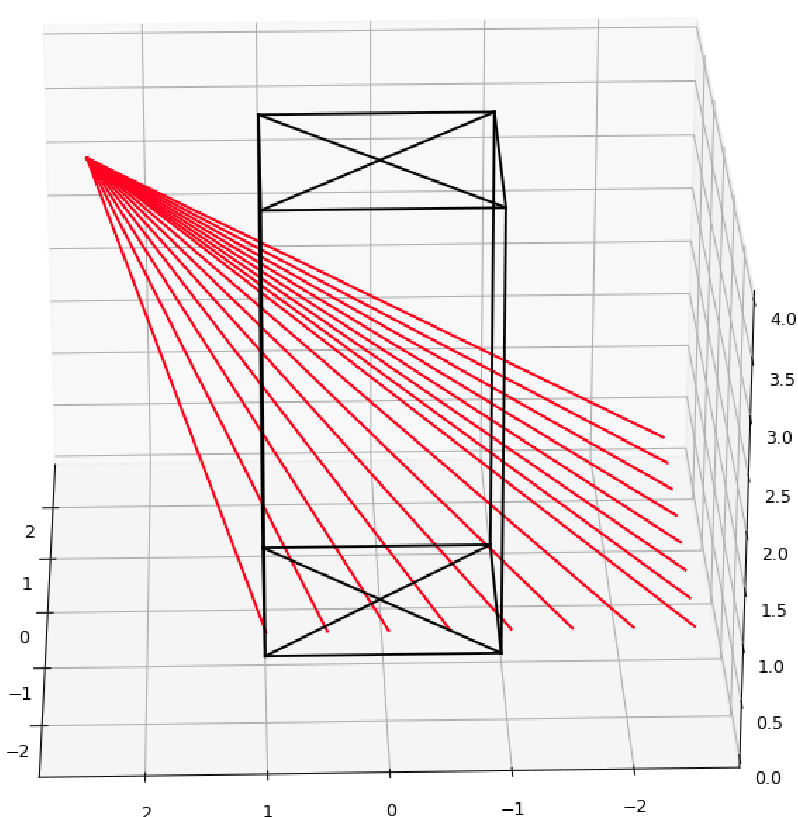}  

\end{subfigure}%
\hspace{3em}
\begin{subfigure}{.4\textwidth}
  \centering
  \includegraphics[width=0.95\linewidth]{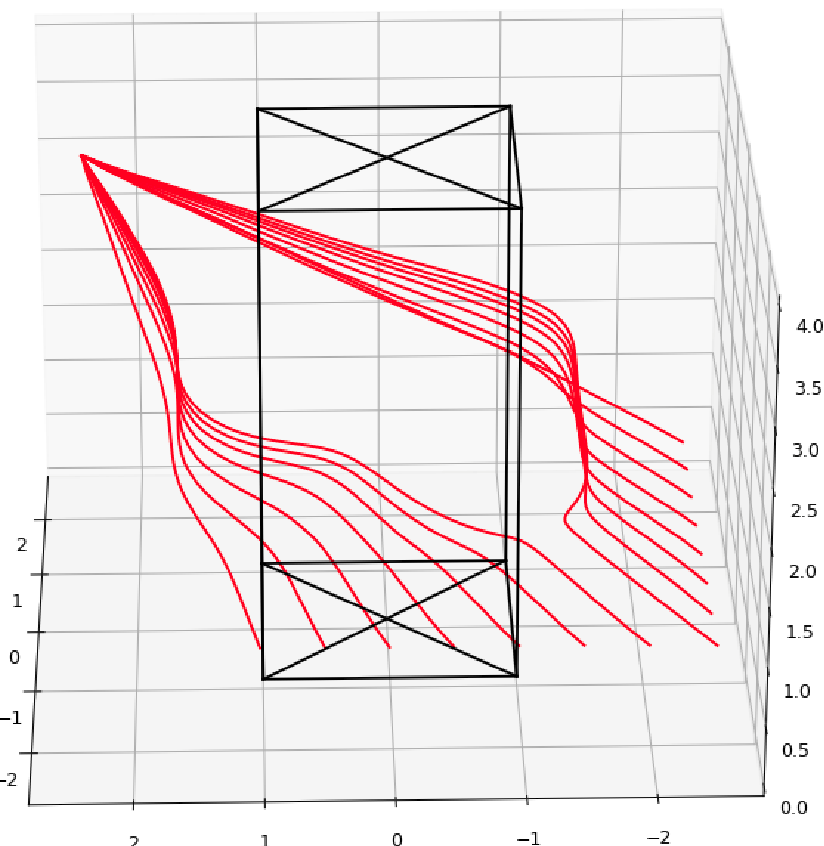}  
\end{subfigure}%
\caption{A 3d example of avoiding a cube obstacle (black). (Left) As a simple example, an attractor $\dot{\bm{x}}=f(\bm{x})$ is on $\boldsymbol{M}$; (Right) The corresponding natural gradient system on $\boldsymbol{M}^{\text{o}}$, given by $\dot{\bm{y}}=f_{\text{o}}(\bm{y})$ generates trajectories (red) which avoid the obstacle by smoothly warping around.}
\label{simple3d}
\end{figure}

When rolling out trajectories with the collision-avoiding system, the initial conditions given by some $\bm{y_{0}}\in\boldsymbol{M}^{\text{o}}$ are assumed to be collision-free. As $\bm{y}$ approaches the boundary of the obstacle, the natural gradient system smoothly warps around the obstacle. \Cref{simple3d} (right) gives a simple 3d example of trajectories of the natural gradient system on $\boldsymbol{M}^{\text{o}}$ which smoothly avoids the cube obstacle. The corresponding system on $\boldsymbol{M}$, illustrated in \cref{simple3d} (left), is a simple attractor. Although the example given is a simple attractor, the system on $\boldsymbol{M}$ can encode imitation. 

\subsection{Building Infinitesimal Generators for Collision Avoidance in C-space}\label{CspaceCol}

In imitation learning setups with a manipulator, it is often necessary to learn a stable dynamical system that governs the Cartesian \emph{world-space} coordinates of the end-effector. At the same time, collision avoidance in such a scenario is typically not limited to the end-effector but required for the entire manipulator based on its \emph{configuration-space} parametrisation. We can define diffeomorphic transforms which encode avoidance behaviours for multiple body points on the manipulator operating in the configuration space (C-space) \citep{lavalle:2006} of the robot. We focus our discussion on the common case where the manipulator configuration is given by joint angles.

A $d_{\bm{q}}$-dimensional configuration of the manipulator is defined as $\bm{q}\in\mathcal{Q}\subset\mathbb{R}^{d_{\bm{q}}}$, where the C-space is denoted as $\mathcal{Q}$. Robot configurations can be mapped into the world-space coordinates of $c$ body points on the robot by a set of $c$ kinematic functions, $\{\mathbf{x}^{\text{f}}_{i}\}_{i=1}^{c}$. Each kinematic function, $\mathbf{x}^{\text{f}}_{i}:\mathcal{Q} \rightarrow \mathbb{R}^{d}$, represents the Cartesian world space coordinates of the $i^{th}$ body point from a given configuration. For multiple-link manipulators, these kinematic functions are often modelled by of simple operations on basic trigonometry functions \citep{denavit195554} and are smooth with respect to the configurations. We set the manipulator end-effector as the first body point, thus $\mathbf{x}^{\text{f}}_{1}(\cdot)$ is the forward kinematics of the robot. Denoting the end-effector position as $\mathbf{x}^{e}$, the velocity of the end-effector and the configurations are described by:
\begin{align}
    \dot{\mathbf{x}^{e}}=\bm{J}_{\mathbf{x}^{\text{f}}_{1}}(\bm{q})\dot{\bm{q}}, && \dot{\bm{q}}=\bm{J}_{\mathbf{x}^{\text{f}}_{1}}^{\dagger}(\mathbf{x}^{e})\dot{\mathbf{x}^{e}}, && \bm{J}_{\mathbf{x}^{\text{f}}_{1}}(\cdot)=\frac{\partial\mathbf{x}^{\text{f}}_{1}}{\partial \bm{q}}(\cdot).
\end{align}
 The dimensions of $\bm{q}$ are typically greater than that of the end-effector coordinates $\mathbf{x}^{e}$, and many configurations can map to the same end-effector coordinates. To obtain a unique $\dot{\bm{q}}$ for some $\dot{\bm{x}}$, we take $\bm{J}_{\mathbf{x}^{\text{f}}_{1}}^{\dagger}=\bm{J}_{\mathbf{x}^{\text{f}}_{1}}^{\top}(\bm{J}_{\mathbf{x}^{\text{f}}_{1}}\bm{J}_{\mathbf{x}^{\text{f}}_{1}}^{\top})^{-1}$ as the the Moore-Penrose pseudoinverse. For some $\dot{\bm{x}}$, its product with $\bm{J}_{\mathbf{x}^{\text{f}}_{1}}^{\dagger}$ provides the unique $\dot{\bm{q}}$ where $\lvert\lvert\dot{\bm{q}}\lvert\lvert$ is also minimised. The Moore-Penrose pseudoinverse preserves stability of the resulting system \citep{INVJac_proof}. Provided a dynamical system in the world-space coordinates of the end-effector, we can use $\bm{J}_{\mathbf{x}^{\text{f}}_{1}}^{\dagger}$ to \emph{pull} the end-effector velocity to C-space and obtain $\dot{\bm{q}}$. If the end-effector follows a dynamical system given by $\dot{\bm{x}}=f(\bm{x})$, we can define the system in C-space:
\begin{equation}
    \dot{\bm{q}}=\bm{J}_{\mathbf{x}^{\text{f}}_{1}}^{\dagger}f(\mathbf{x}^{\text{f}}_{1}(\bm{q})).\label{Pull}
\end{equation}
Using inverse kinematics, we can find a starting configuration $\bm{q}_{0}$ which matches up to the initial world-space coordinate, i.e. $\mathbf{x}^{\text{f}}_{1}(\bm{q}_{0})=\bm{x}_{0}$, then we can numerically integrate to roll out a sequence of configurations. 

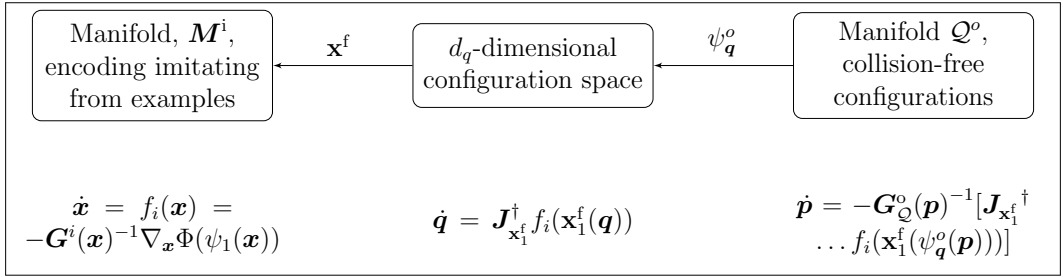
\begin{figure}[bt]
\begin{adjustbox}{width=0.85\textwidth,center} 
\fbox{\begin{tikzpicture}[node distance = 2cm]

    \node [block] (Imitate) {Manifold, $\boldsymbol{M}^{\text{i}}$, encoding imitating from examples};
    \node [block, right of=Imitate, node distance=6cm] (Cspace) {$d_{q}$-dimensional configuration space};
    \node [block, right of=Cspace, node distance=6cm] (warppedC) {Manifold $\mathcal{Q}^{o}$, collision-free configurations};
    
    \node [block2, below of=Imitate, node distance=2.5cm] (Sys2) {$\dot{\bm{x}}=f_{i}(\bm{x})=-\bm{G}^{i}(\bm{x})^{-1}\nabla_{\bm{x}}\Phi(\psi_{\text{1}}(\bm{x}))$};
    \node [block2, below of=Cspace, node distance=2.5cm] (Sys3) {$\dot{\bm{q}}=\bm{J}_{\mathbf{x}^{\text{f}}_{1}}^{\dagger}f_{i}(\mathbf{x}^{\text{f}}_{1}(\bm{q}))$};
    \node [block2, below of=warppedC, node distance=2.5cm] (Sys4) {$\dot{\bm{p}}=-{\bm{G}^{\text{o}}_{\mathcal{Q}}(\bm{p})}^{-1}[{\bm{J}_{\mathbf{x}^{\text{f}}_{1}}}^{\dagger}$\\$\ldots f_{i}(\mathbf{x}^{\text{f}}_{1}(\psi^{o}_{\bm{q}}(\bm{p})))]$};
    
    \path [line] (Cspace) -- node  [text width=0.5cm,midway,above]{$\mathbf{x}^{\text{f}}$} (Imitate);
    \path [line] (warppedC) -- node [text width=0.5cm,midway,above]{$\psi^{o}_{\bm{q}}$}(Cspace);
    
\end{tikzpicture}}
\end{adjustbox}
\caption{An example of a system on $\mathcal{Q}^{\text{o}}$, for imitating demonstrations and avoiding collision along body points. The mappings between manifolds in world-space and C-space, along with the system constructed on each manifold is shown.}\label{flowchart2}
\end{figure}

In order for body points on the robot to be collision-free, we shall extend our approach in \cref{collsion} to construct an infinitesimal generator to encode collision-avoidance in C-space. Instead of using the derivative of occupancy with respect to world-space coordinates, we make use of the derivative of occupancy over all body points with respect to configurations. Provided a continuous occupancy representation, as given in \cref{map}, the derivative of the combined occupancy of body points, wrt to configurations, gives smooth vector field $V^{\text{o}}_{q}$,
\begin{align}
    V^{o}_{\bm{q}}(\bm{q}):&=\nabla_{\bm{q}}\sum_{j=1}^{c}f^{map}(\mathbf{x}^{\text{f}}_{j}(\bm{q})),
\end{align}
where $f^{map}$ is the continuous occupancy representation as outlined in \cref{collsion}. We can then use $V^{o}_{\bm{q}}$ as the infinitesimal generator of a diffeomorphism in a manner similar to \cref{diffeo}. This gives us a $d_{\bm{q}}$-dimensional manifold, which we shall denote as $\mathcal{Q}^{o}$, where natural gradient systems corresponding to systems in $\mathcal{Q}$ are collision-free. The diffeomorphism, $\phi^{o}_{\bm{q}}$, mapping from $\mathcal{Q}^{o}$ to $\mathcal{Q}$ is given by following the flow, $\gamma_{\bm{q}}^{o}$, on $V^{o}_{\bm{q}}$. For a configuration $\bm{q}\in\mathcal{Q}$, and its corresponding collision-avoiding configuration $\bm{p}\in\mathcal{Q}^{o}$, we define the diffeomorphism:  
\begin{equation}
        \psi^{o}_{\bm{q}}(\bm{p}):=\gamma^{o}(t,\bm{y})=\bm{p}+\int_{0}^{t}V^{o}_{\bm{p}}(\gamma^{o}(u,\bm{p}))\mathrm{d}u  = \bm{q}
        \end{equation}
and its inverse:
\begin{equation}
    {\psi^{o}_{\bm{q}}}(\bm{q})^{-1}:=\gamma^{o}(-t,\bm{q})=\bm{q}+\int_{-t}^{0}V^{o}_{\bm{q}}(\gamma^{o}(u,\bm{q}))\mathrm{d}u = \bm{p}.
\end{equation}    
Following \cref{naturalDesc}, we can use the diffeomorphism to define our system on $\mathcal{Q}^{o}$ as:
\begin{align}
    \dot{\bm{p}}&=-{\bm{G}^{\text{o}}_{\mathcal{Q}}(\bm{p})}^{-1}[{\bm{J}_{\mathbf{x}^{\text{f}}_{1}}}^{\dagger}f_{i}(\mathbf{x}^{\text{f}}_{1}(\psi^{o}_{\bm{q}}(\bm{p})))], \label{CspaceNG}\\ \bm{G}^{\text{o}}_{\mathcal{Q}}(\bm{p})&={\bm{J}_{\psi^{o}_{\bm{q}}}(\bm{p})}^{\top}\bm{J}_{\psi^{o}_{\bm{q}}}(\bm{p}).
\end{align}
\Cref{flowchart2} shows a specific example of using diffeomorphic transforms to construct a system that avoids collision along chosen body points and also imitates demonstrations. Provided a stable dynamical system on $\mathbf{M}^{i}$, $\dot{\bm{x}}=f_{i}(\bm{x})$, learned from demonstration data in world-space (as outlined in \cref{learnsection}), we can pull back the system to C-space $\mathcal{Q}$ via \cref{Pull}. The natural gradient system on the collision-avoiding manifold in C-space is then given in \cref{CspaceNG}, we can roll out $\bm{p}$ to obtain collision-free configurations.

\subsection{Infinitesimal Generators for User-Specified Bias}\label{bias}

Sometimes it is desirable to deform learned motions based on user input without recording new expert demonstrations. In this section, we outline a method to construct a diffeomorphism which can apply deformation biases to specific coordinates along a trajectory.

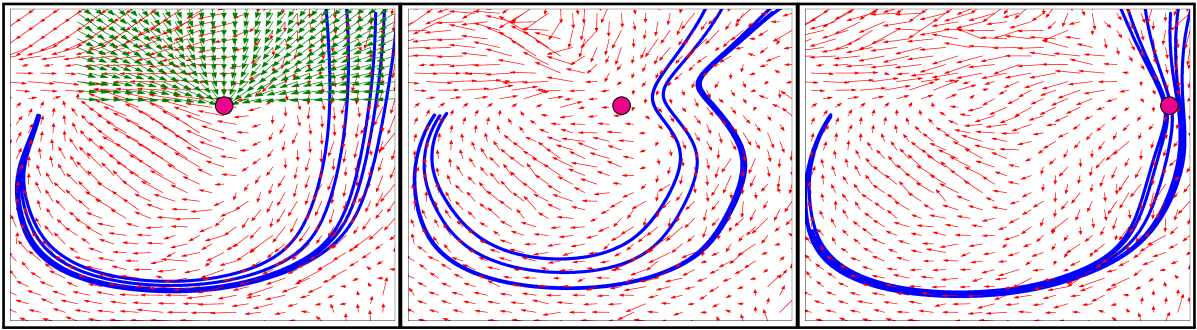
\begin{figure}[bt]
\centering
 \setlength{\fboxsep}{0pt}%
\setlength{\fboxrule}{1pt}%
\begin{subfigure}{.32\textwidth}
    \begin{tikzpicture}
        \node (img) {\fbox{\includegraphics[clip,width=0.99\linewidth]{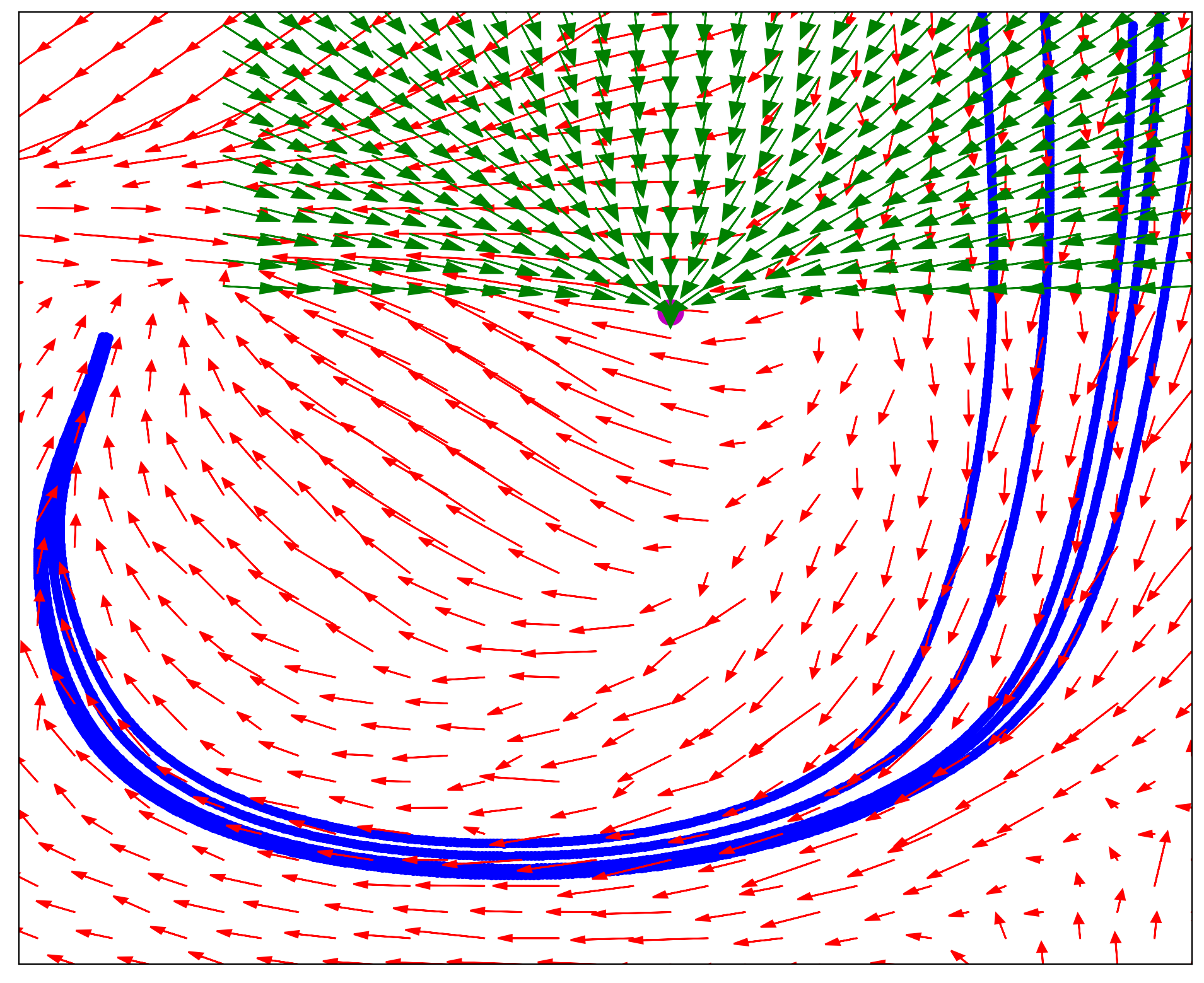}}};
        \node[style={draw,shape=circle,fill=magenta,scale=0.6}] at (.30,0.8){};
    \end{tikzpicture}%
  \centering
\end{subfigure}%
\begin{subfigure}{.32\textwidth}
  \centering
    \begin{tikzpicture}
        \node (img) {\fbox{\includegraphics[clip,width=0.99\linewidth]{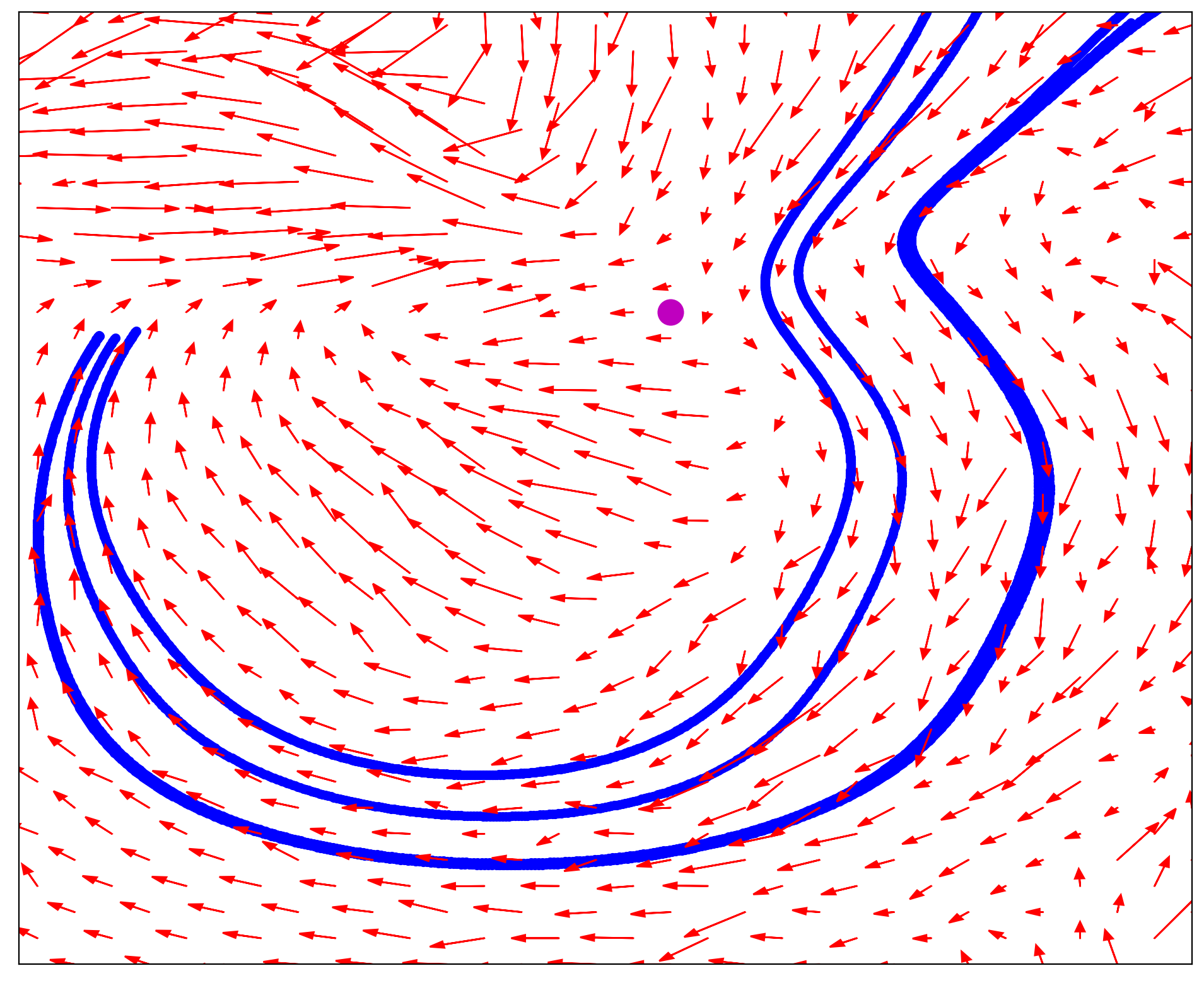}}};
        \node[style={draw,shape=circle,fill=magenta,scale=0.6}] at (.30,0.8){};
    \end{tikzpicture}%
\end{subfigure}%
\begin{subfigure}{.32\textwidth}
  \centering
    \begin{tikzpicture}
        \node (0,0) {\fbox{\includegraphics[clip,width=0.99\linewidth]{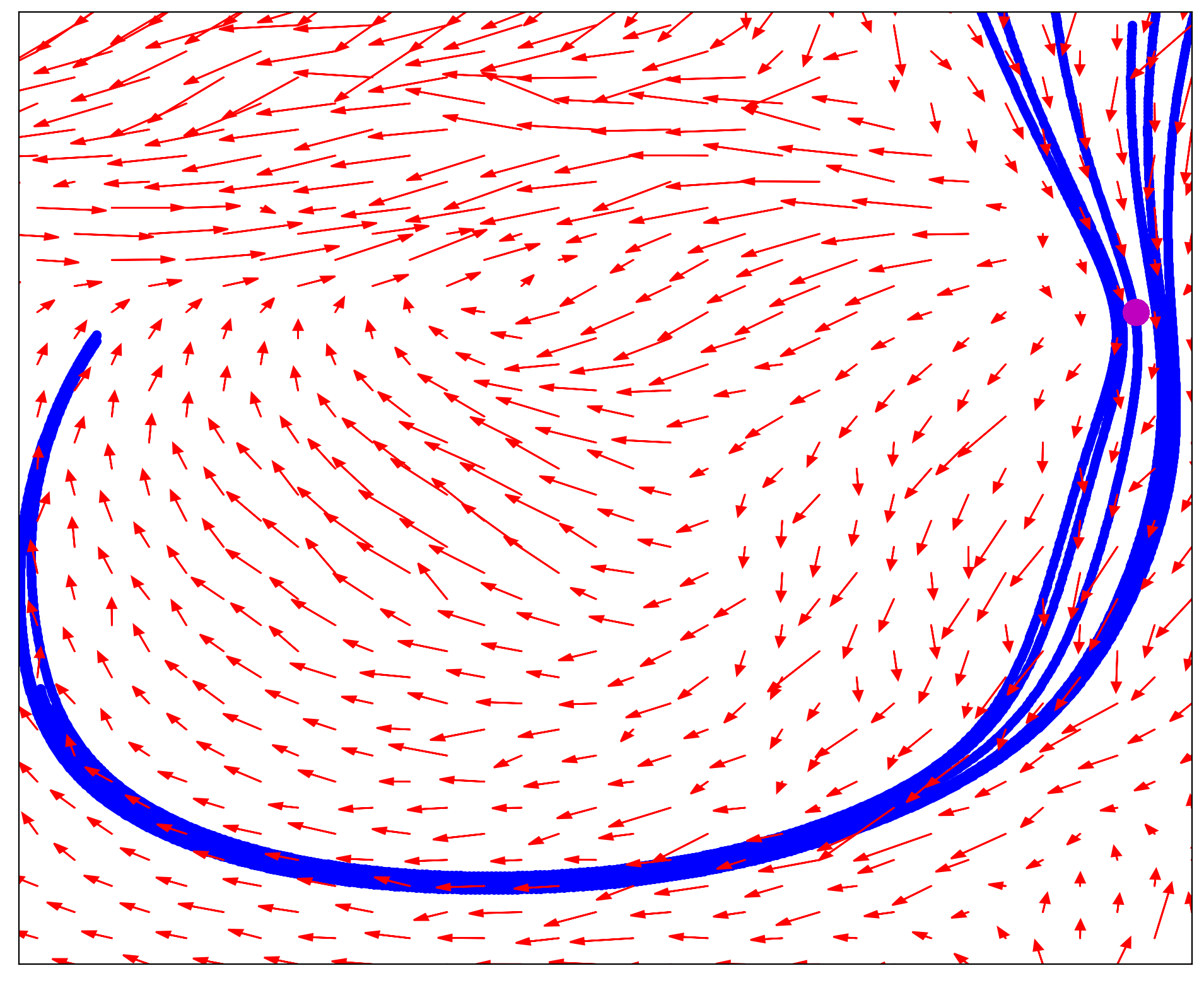}}};
        \node[style={draw,shape=circle,fill=magenta,scale=0.6}] at (2.3,0.8){};
    \end{tikzpicture}%
\end{subfigure}%
\caption{(Left) a stable dynamical system (visualised as red vector field) on $\boldsymbol{M}^{\text{i}}$ learned to imitate drawing a 'J' (example trajectories in blue). We wish to bias the system to the magenta point, the green vector field illustrates, $-V^{\text{b}}(\bm{s})$; (Center) the resulting system by biasing towards the magenta point is shown; (Right) biasing the stable system towards another coordinate (magenta point).}\label{biasingplots}
\end{figure}

Similar to previous sections, we focus on building diffeomorphic transforms by directly crafting their infinitesimal generator. We design the manifold $\boldsymbol{M}^{\text{b}}$ by constructing the diffeomorphism $\psi^{b}$ such that the inverse ${\psi^{b}}^{-1}$ maps straight trajectories in a specific region to trajectories with the specified bias. As we define ${\psi^{b}}^{-1}$ from reverse-time integral curves of the generator, we need to build the generator $V^{\text{b}}$ such that the biased direction is given by $-V^{\text{b}}$. This is achieved by regressing specified vectors onto the infinitesimal generator representation introduced in \cref{vecfieldeqn}. We collect sample points within a fixed region in the state-space and compute a unit vector pointing away from the point of interest, giving us a dataset of $n_{\bm{s}}$ pairs $\mathcal{D}=\{(\bm{s}_{i}^{\text{a}},\nabla_{\bm{s}_{i}^{\text{a}}}\lvert\lvert \bm{s}_{i}^{\text{a}}-\bm{s}^{tar}\lvert\lvert_{2})\}_{i=1}^{n_{\bm{s}}}$, where $\bm{s}^{\text{a}}\in\mathcal{S}$ are sample points in some region $\mathcal{S}$ where the bias takes effect, and $\bm{s}^{tar}$ is the coordinate of interest. To produce a valid diffeomorphism, we require an infinitesimal generator to be smooth. This is achieved by fitting a smooth function approximator introduced in \cref{vecfieldeqn} on $\mathcal{D}$, by minimising:
\begin{align}
    \bm{W}^{*}&:=\arg\min_{\bm{W}}\sum_{i=1}^{n_{\bm{s}}}\lvert\lvert \nabla_{\bm{s}_{i}^{\text{a}}}\lvert\lvert \bm{s}_{i}^{\text{a}}-\bm{s}^{tar}\lvert\lvert_{2})-\bm{W}^{\top}K(\bm{s}_{i},\hat{\bm{s}})\lvert\lvert_{2}^{2},
\end{align}
where we now have a smooth $V^{\text{b}}:=\bm{W}^{{*}^{\top}}K(\bm{s},\hat{\bm{s}})$. Like \cref{diffeo}, we can integrate $V^{\text{b}}$ to obtain diffeomorphism $\psi^{b}$. With \cref{naturalDesc} we can take natural gradients on the resulting manifold $\boldsymbol{M}^{\text{b}}$.

An example of biasing a learned stable system towards two different coordinates of interest is illustrated in \cref{biasingplots}. Constructing a DT for complicated morphing on the original system may be difficult, as it is challenging to intuit what the infinitesimal generator needs to look like. Furthermore, unexpected behaviour may arise in complex setups, as stretching in one space contracts another \citep{Ratliff_learninggeometric}. However, for relatively simple cases such as bias towards a point or direction, a simple generator such as $V^{b}$, which points away from the point of interest, results in the desired behaviour.

\section{Experimental Evaluation}


We empirically evaluate the performance of diffeomorphic transforms to imitate and generalise robot motions, both in simulation and on a real-world 6-DOF JACO manipulator. In \cref{demo}, we evaluate the ability of our method to accurately reproduce human demonstrations. In \cref{colReproduction}, we evaluate the generalisation performance of our approach when occupancy in the environment changes and compare our results against multiple baselines. Additionally, in \cref{Userinfluenced}, we demonstrate the ability of diffeomorphic transforms to generalise based on user-specified requirements. 

\subsection{Imitating Expert Demonstrations}\label{demo}
\begin{figure}[t]                                                  \setlength{\fboxsep}{0pt}%
\setlength{\fboxrule}{1pt}%
    \centering                                                       
    \begin{subfigure}{0.2\linewidth}                   
        \fbox{\includegraphics[width=.99\linewidth]{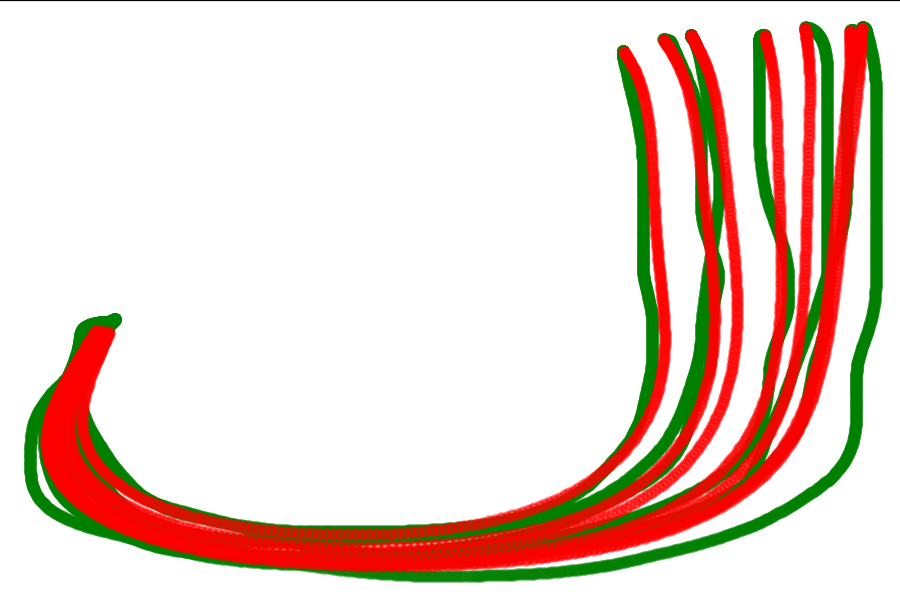}}
        \fbox{\includegraphics[width=.99\linewidth]{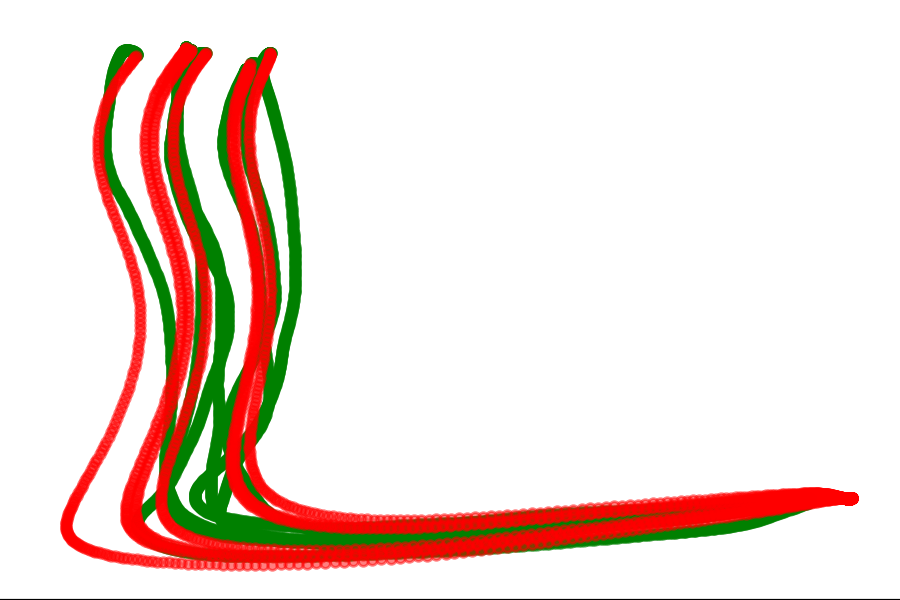}}

    \end{subfigure}%
    \begin{subfigure}{0.2\linewidth}                  
        \fbox{\includegraphics[width=.99\linewidth]{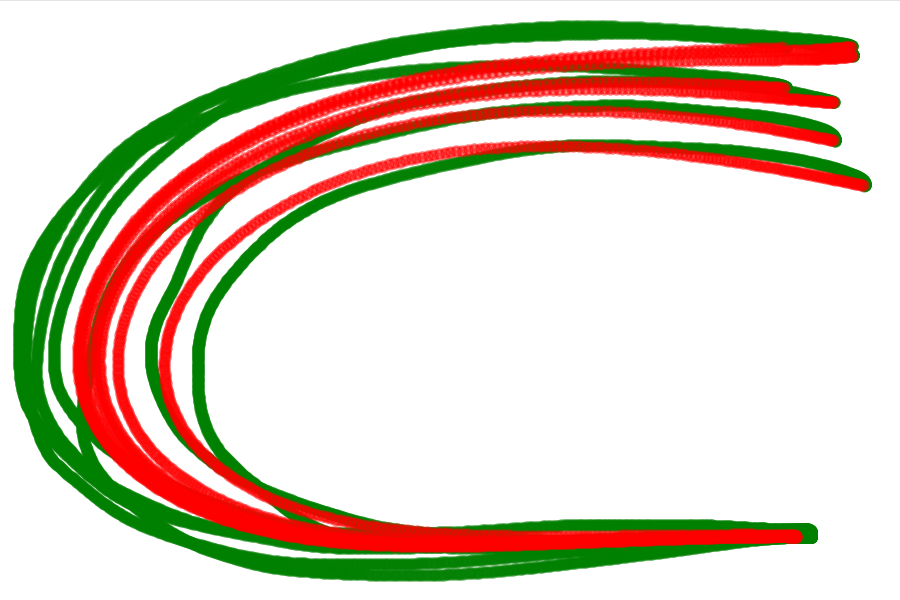}}
        \fbox{\includegraphics[width=.99\linewidth]{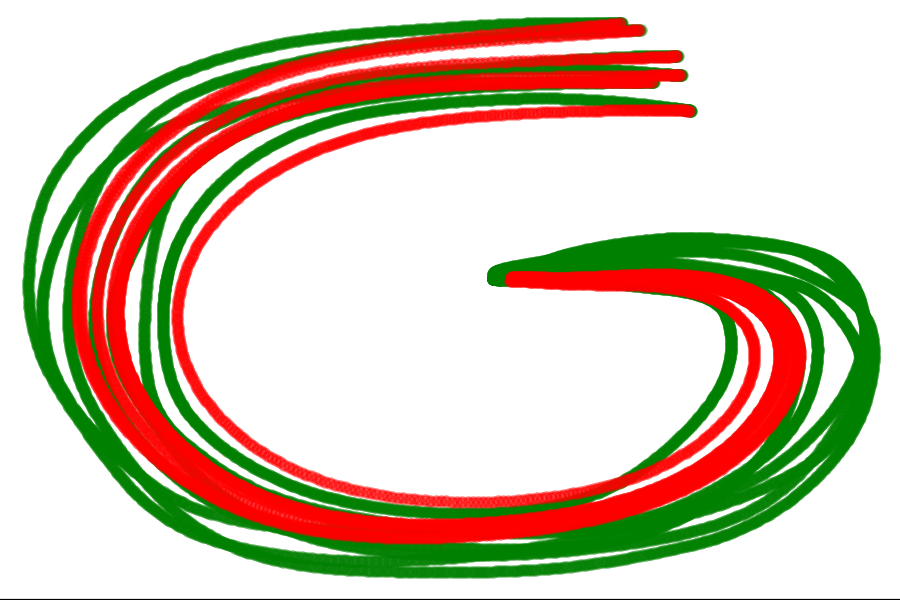}}

    \end{subfigure}%
    \begin{subfigure}{0.2\linewidth}                      
        \fbox{\includegraphics[width=.99\linewidth]{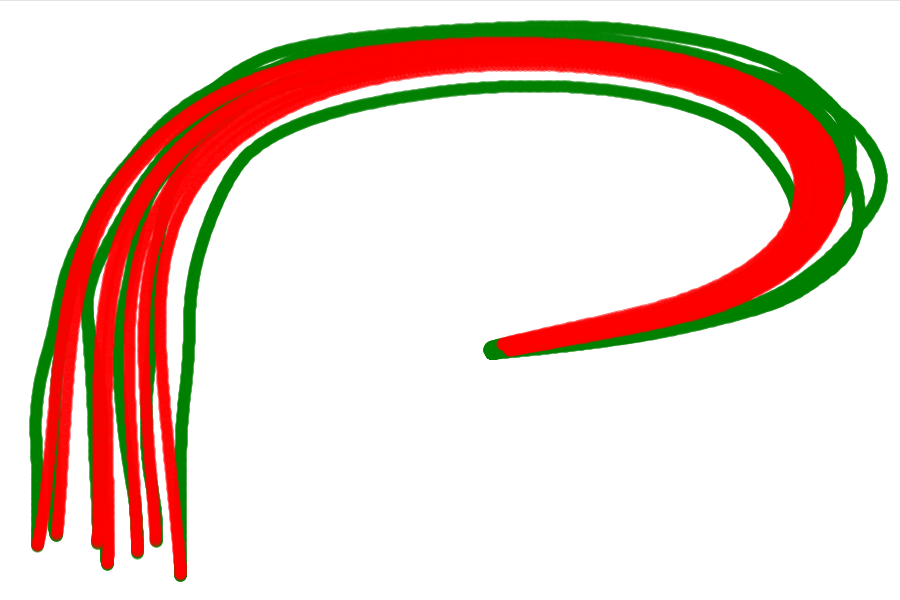}}
        \fbox{\includegraphics[width=.99\linewidth]{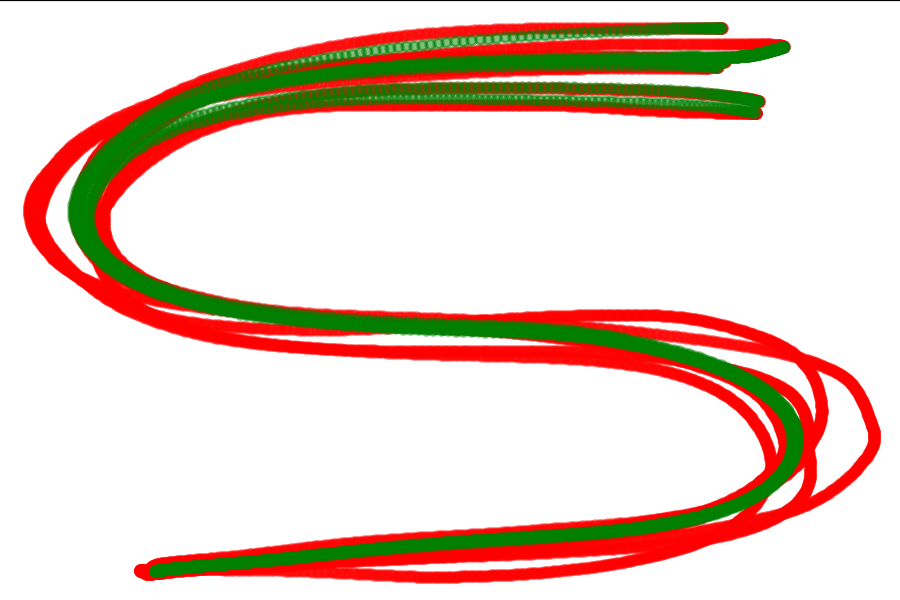}}

    \end{subfigure}%
    \begin{subfigure}{0.2\linewidth}                      
        \fbox{\includegraphics[width=.99\linewidth]{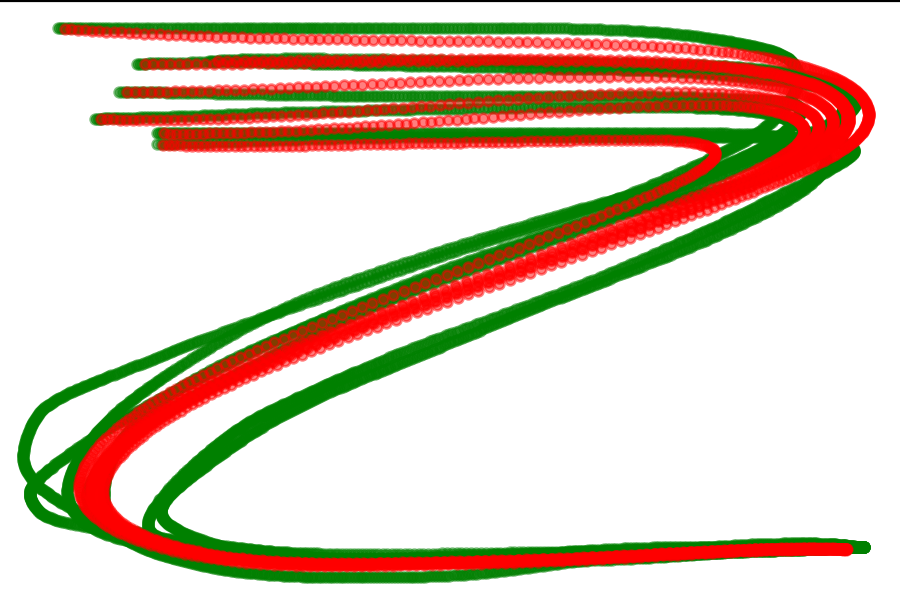}}
        \fbox{\includegraphics[width=.99\linewidth]{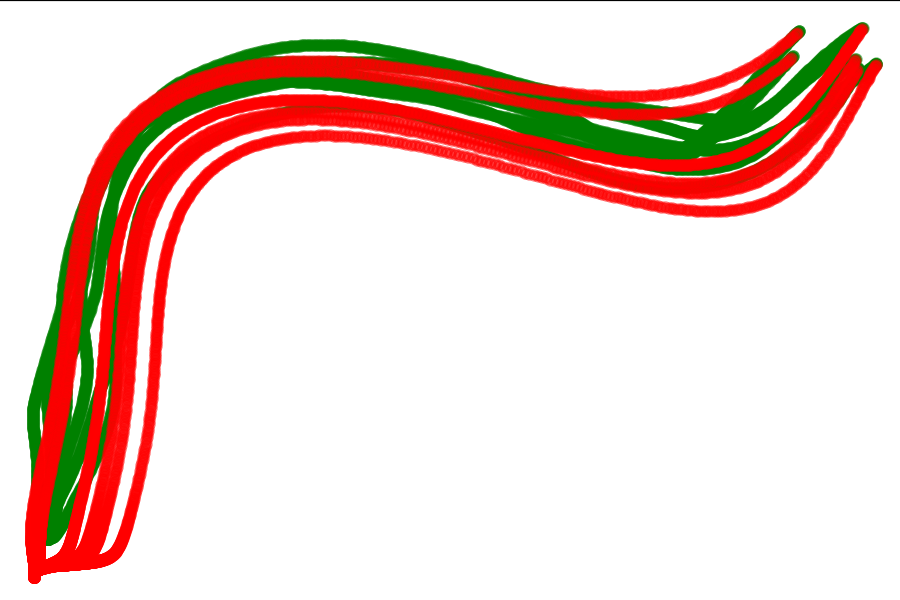}}
    \end{subfigure}
    \caption{      
Imitating expert examples from the LASA dataset \citep{SEDS}. Reproduced motion in red and demonstrations in green.
    }\label{CharRes}                                                                
\end{figure}

The first analysis we carry out is on the ability of diffeomorphic transforms to imitate provided demonstrations. We train diffeomorphic transforms from 8 alphabet characters from the LASA handwriting dataset \citep{SEDS}. There are 7 demonstrations for each character. We use $m=720$ inducing states for the Gaussian basis function, each with hyperparameter $\ell=0.005$, arranged in an equally-spaced grid covering the range of the demonstrations. To build the diffeomorphism. Qualitative results are presented in \cref{CharRes}, where generated trajectories are given in red with demonstrations in green. We observe that the generated trajectories are smooth and consistent with the demonstrated motions. We note that first-order dynamical systems cannot model intersections which exist in the data \citep{aug_ode}, and explains some of the differences between generated and ground truth motion. These results demonstrate that our method is sufficiently flexible to learn diffeomorphic transforms to morph the simple base attractor from complex demonstrations. 

\begin{figure}[h]
    \centering
    \includegraphics[width=0.95\textwidth]{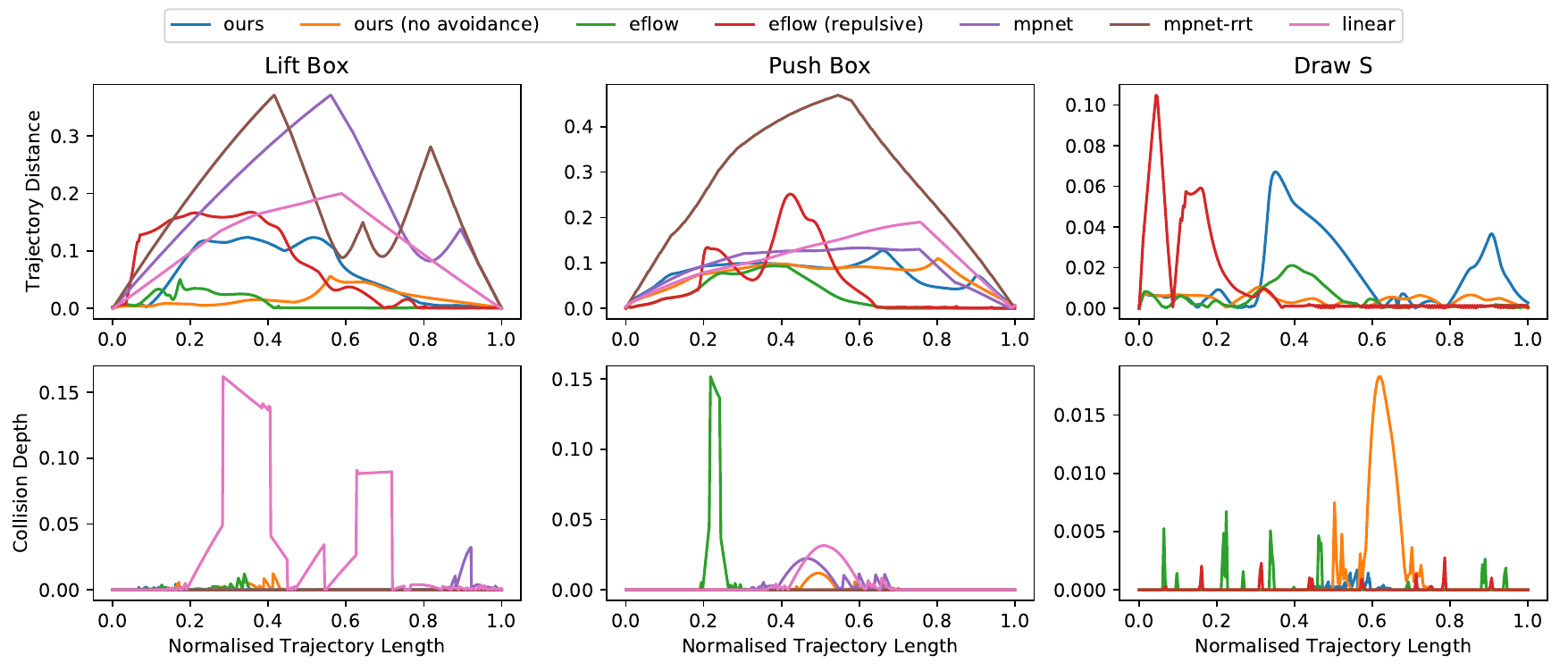}
    \caption{Qualitative results of diffeomorphic transforms along with comparisons. We normalise the generated trajectories by arc-length, and report the trajectory distance and M.C.D. at fixed intervals. For the draw ``S'' task, results for MPNet, MPNet-RRT, and linear are discarded. These methods were unable to imitate, and departed greatly from the ground truth.}
    \label{resultsFig}
\end{figure}

\subsection{Generalised Collision-Avoiding Motion Reproduction}\label{colReproduction}
In this section, we evaluate the efficacy of diffeomorphic transforms to adapt the motion according to the occupancy of the surroundings and avoid collisions. We work with demonstrations from the following tasks:
(i) pushing a box (7 demonstrations collected with JACO arm);
(ii) lifting a box, by reaching inside and abruptly moving upwards (8 demonstrations collected with JACO arm);
(iii) drawing an ``S" character (7 demonstrations from LASA dataset).
We train diffeomorphic transforms to imitate expert demonstrations to complete these tasks. Then, we place an obstacle of diameter 15cm such that the original trajectories are in collision. We then provide occupancy data of the new obstacle, and construct diffeomorphic transforms in C-space to provide collision avoidance for the entire manipulator.

\subsubsection{Baseline Methods}
We evaluate the performance of the following methods and variants: 
\begin{enumerate}
\item Our method with DTs to imitate demonstrations and avoid collisions. 
\item Our method with only imitation transformation. 
\item Euclideanizing Flows (Eflow) \citep{Euclideanising}. 
\item Eflow with an additional repulsor overlaid for collision avoidance. 
\item Motion-planning networks (MPNet) \citep{MPNet}: Although MPNet is typically used for planning, at its core, it uses a neural network to imitate demonstrations from a motion planner. 
\item MPNet with RRT replanning (MPNet-RRT) \citep{MPNet}: a variant where results are touched up by a RRT planner \citep{Lavalle98rapidly-exploringrandom}. 
\item Linear motion in the C-space provided start and end configurations.
\end{enumerate}

It is tempting to generalise imitation learning by simply learning on more demonstrations gathered in more diverse scenarios. Hence, MPNet provides a learning-based comparison, as it attempts to learn avoidance directly from examples of a planner. Training data of non-colliding trajectories in a variety of environment occupancy scenarios are required to train MPNet and MPNet-RRT. For each task, we simulate 200 environments each with a simulated collision-free demonstration obtained with planners along with demonstration waypoints. 

\begin{table}[t]
\centering
\caption{Performance, as measured by Frechet Distance (F.D.) and Maximum Collision Distance (M.C.D.) (3 d.p.) \label{table:traj-distance}}
\begin{tabular}{@{}llccc@{}}
\toprule
 & Distance & Push Box & Lift Box & Draw S \\ \midrule
\multirow{2}{*}{Ours}                & F.D   & 0.124 & 0.127 & 0.067 \\
                                     & M.C.D & 0     & 0     & 0.002 \\
\multirow{2}{*}{Ours (no avoidance)} & F.D   & 0.070 & 0.112 & 0.011 \\
                                     & M.C.D & 0.012 & 0.012 & 0.018 \\
\multirow{2}{*}{EFlow}               & F.D   & 0.068 & 0.093 & 0.021 \\
                                     & M.C.D & 0.012 & 0.152 & 0.007 \\
\multirow{2}{*}{EFlow (repulsive)}   & F.D   & 0.168 & 0.252 & 0.195 \\
                                     & M.C.D & 0     & 0.001 & 0.003 \\
\multirow{2}{*}{MPNet}               & F.D   & 0.371 & 0.133 & N/A  \\
                                     & M.C.D & 0.038 & 0.022 & N/A  \\
 \multirow{2}{*}{MPNet-RRT}          & F.D   & 0.371 & 0.478 & N/A  \\
                                     & M.C.D & 0     & 0     & N/A  \\
\multirow{2}{*}{Linear}              & F.D   & 0.242 & 0.205 & N/A  \\
                                     & M.C.D & 0.147 & 0.031 & N/A  \\ \bottomrule 
\end{tabular}
\end{table}

\begin{figure}[t] \setlength{\fboxsep}{0pt}%
\setlength{\fboxrule}{1pt}%
    \centering                                                       
    \begin{subfigure}{0.32\linewidth}                   
        \fbox{\includegraphics[clip,width=.99\linewidth]{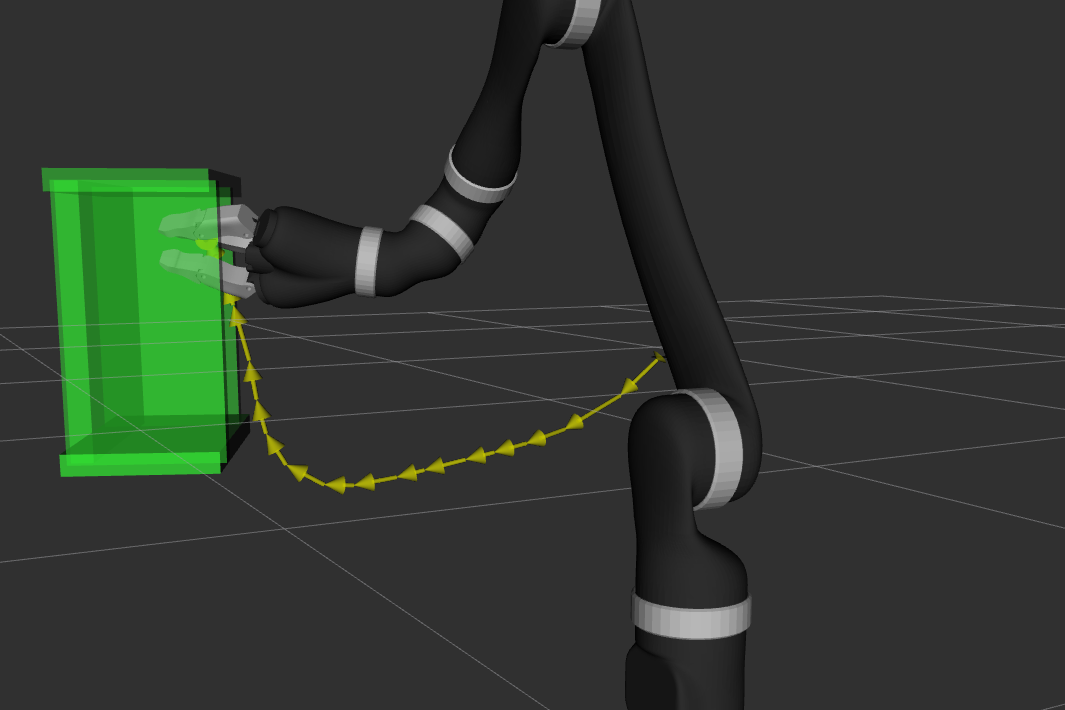}}
        \fbox{\includegraphics[clip,width=.99\linewidth]{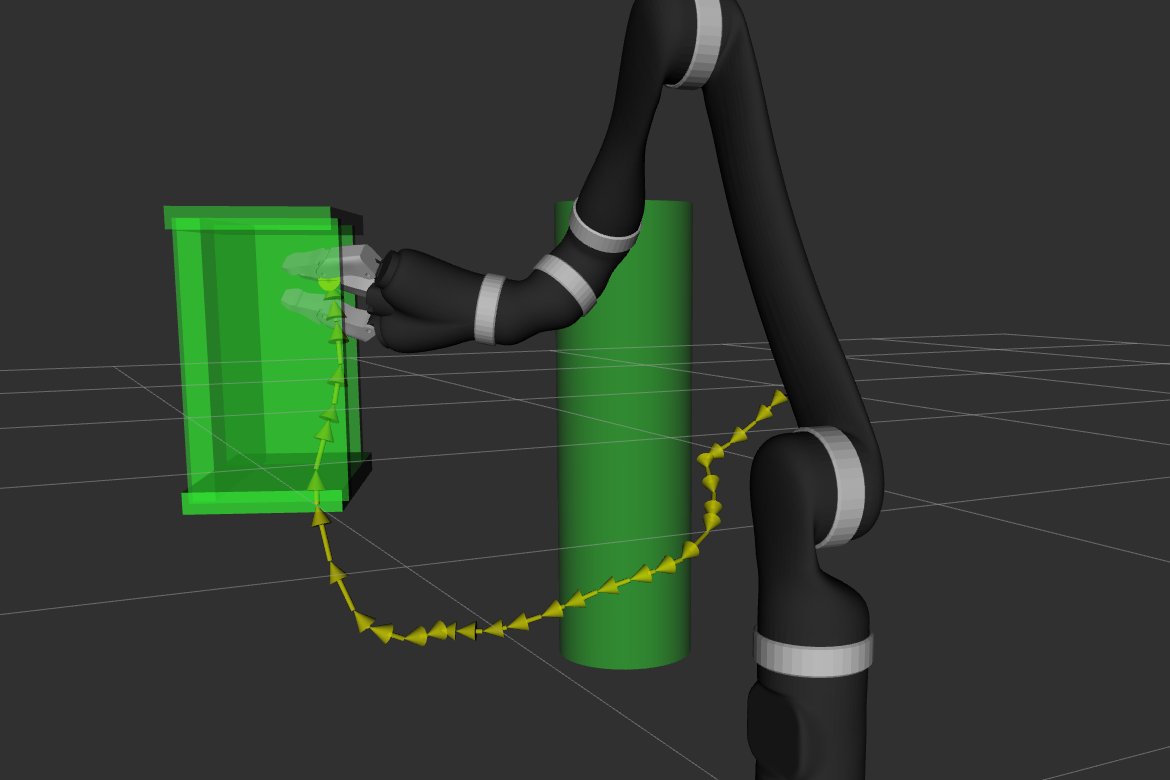}}
        \caption{
            Lift Box\label{fig:liftbox-sim}
        }
    \end{subfigure}%
    \begin{subfigure}{0.32\linewidth}                  
        \fbox{\includegraphics[clip,width=.99\linewidth]{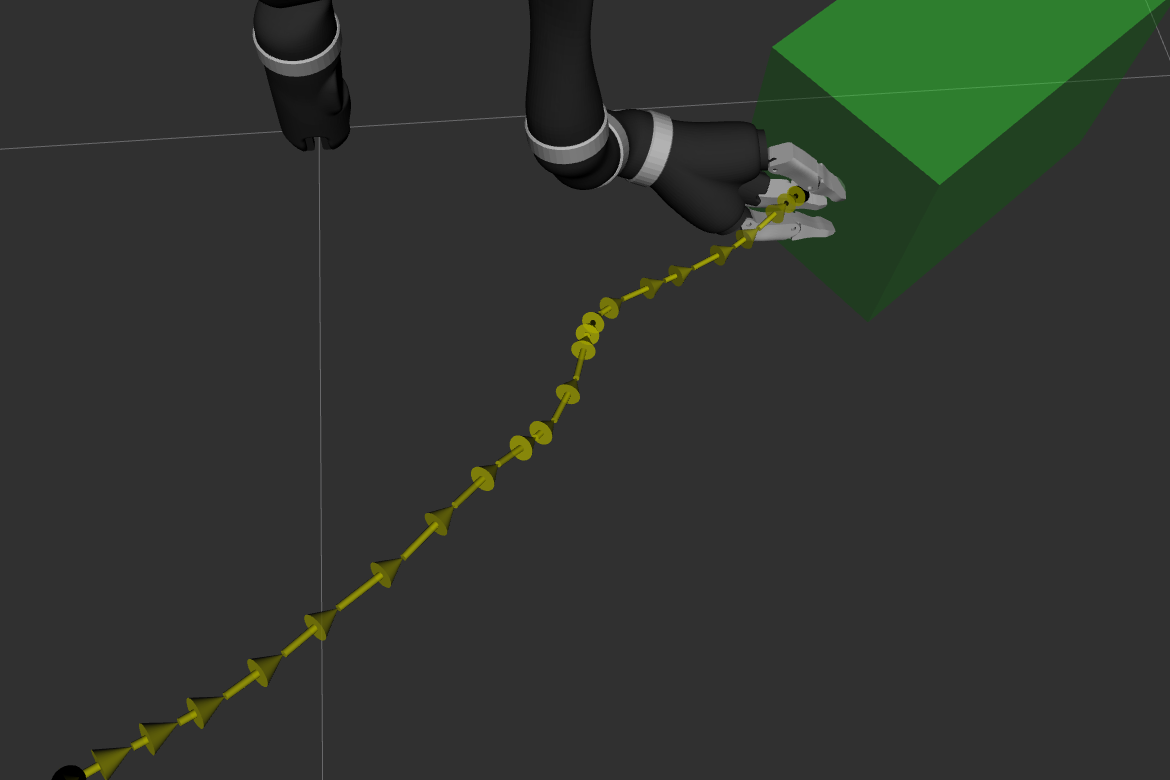}}
        \fbox{\includegraphics[clip,width=.99\linewidth]{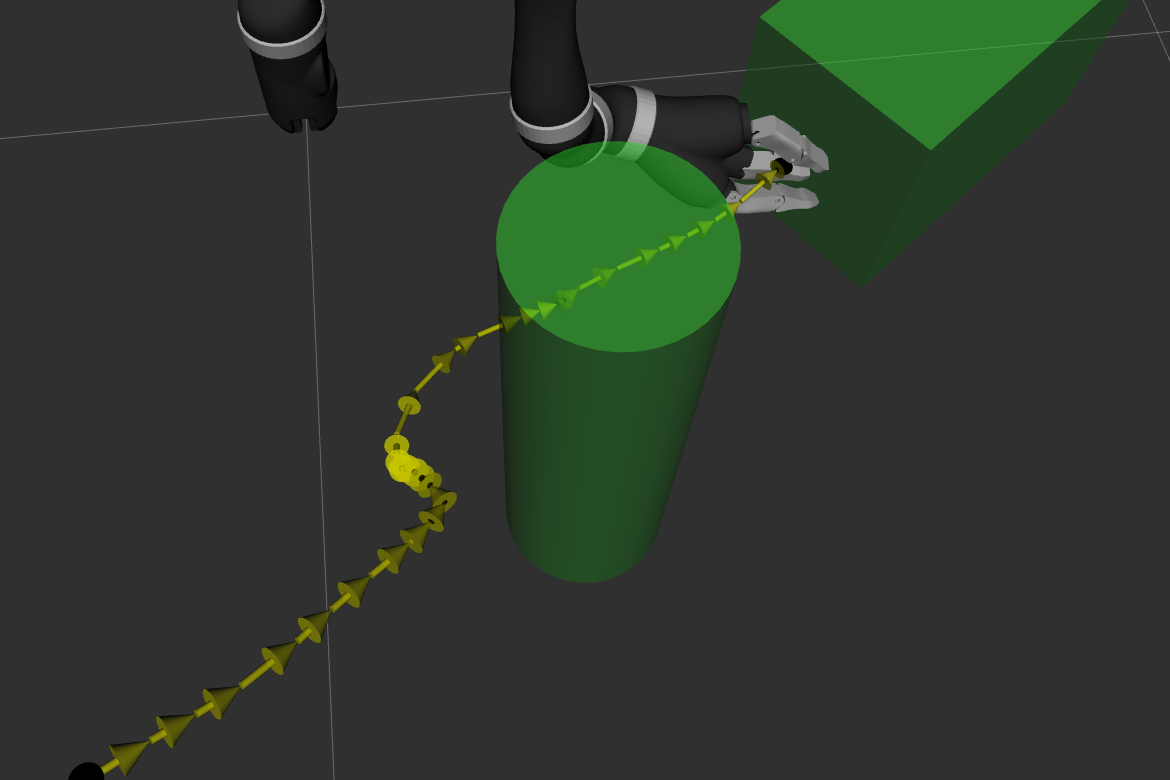}}
        \caption{
            Push Box\label{fig:pushbox-sim}
        }
    \end{subfigure}%
    \begin{subfigure}{0.32\linewidth}                      
        \fbox{\includegraphics[clip,width=.99\linewidth]{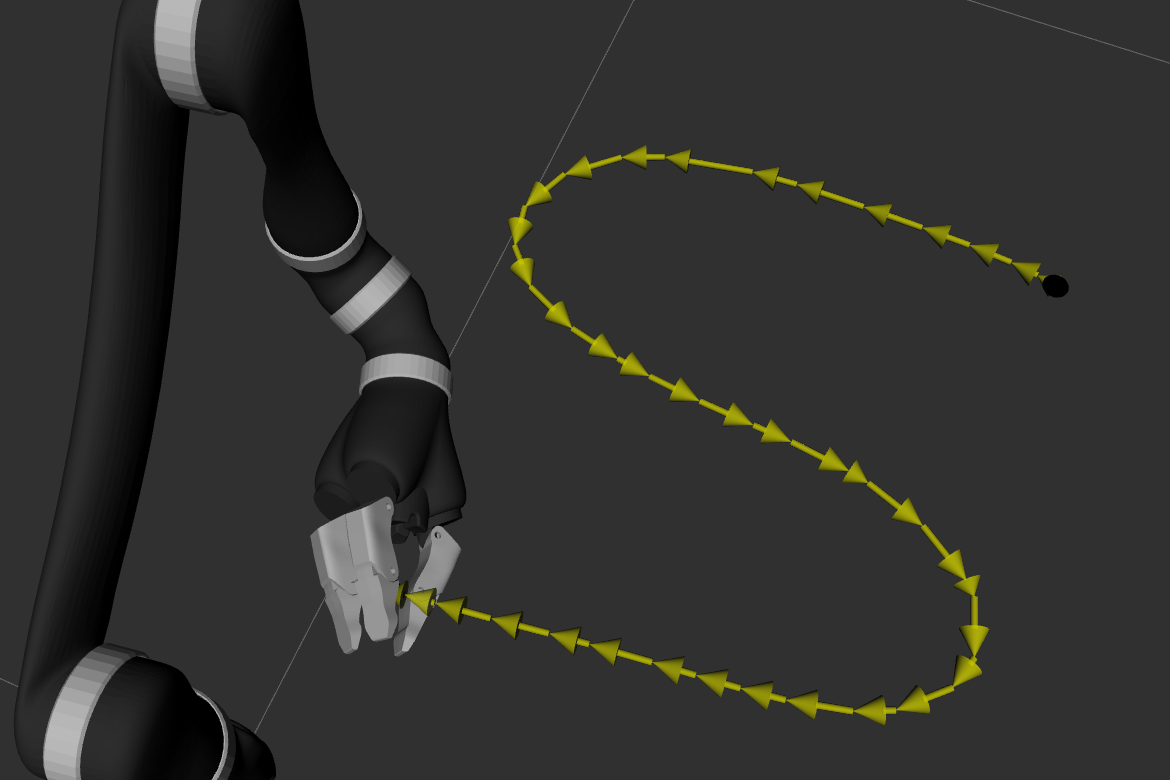}}
        \fbox{\includegraphics[clip,width=.99\linewidth]{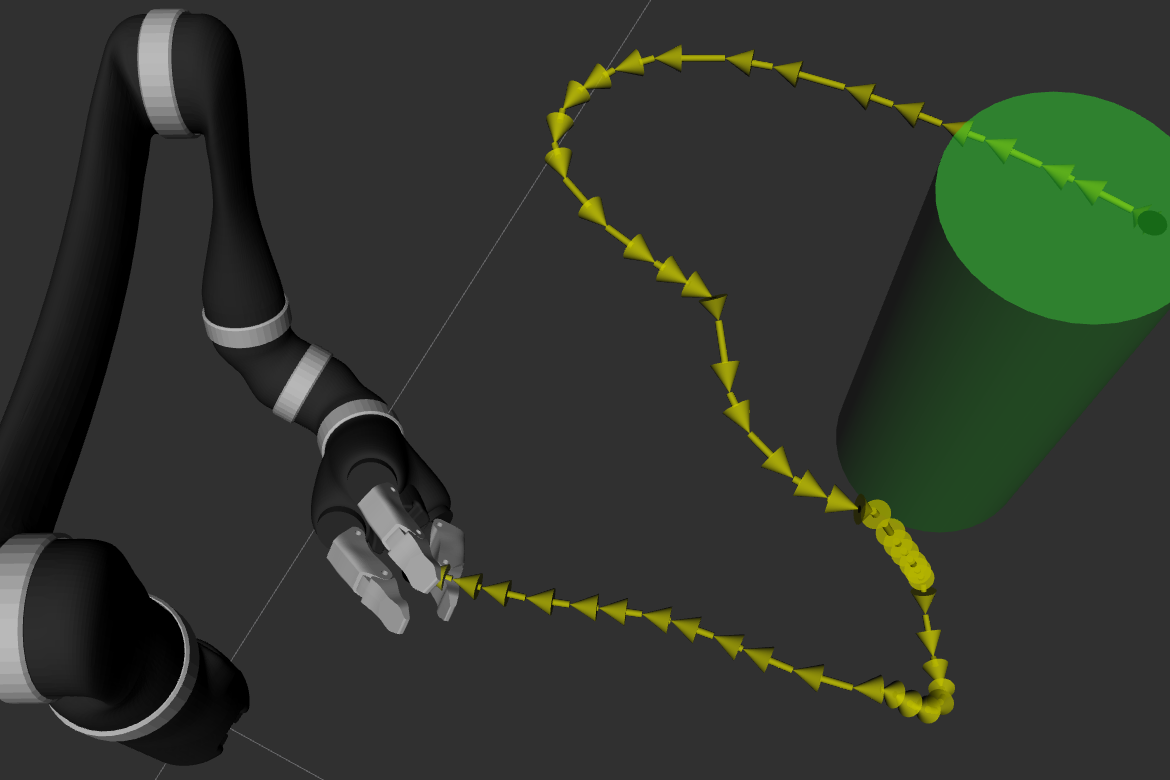}}
        \caption{
            Draw S\label{fig:drawS-sim}
        }
    \end{subfigure}
    \caption{      
        The simulated environments under the MoveIt framework \citep{moveitRef}, where the top row illustrates the trajectories integrated from the infinitesimal generator and the bottom row contains the corresponding trajectories after applying diffeomorphic transformation for obstacle avoidance.    }\label{Simres}                                                                
\end{figure} 

\subsubsection{Evaluation Results}
We quantitatively evaluate the quality of the generalised imitations produced under environment occupancy changes. Results are shown in \cref{resultsFig}, and tabulated in \cref{table:traj-distance}. In \cref{resultsFig}, the metrics used are: (i) trajectory distance (metres), the distance between a point on a generated trajectory and the nearest coordinate of a demonstration; (ii) Maximum Collision Distance (M.C.D.), the depth of the manipulator is in collision with the obstacle. We evaluate both metrics at fixed intervals along arc-length normalised generated trajectories. In \cref{table:traj-distance}, we report: (i) the Fr\'echet distance \citep{frechet} between the generated and ground truth trajectories, which gives a single value over the entire trajectories; (ii) M.C.D..

We observe from tabulated figures, for pure imitation, our method generates trajectories with a relatively low Frechet Distance to the ground truth, comparable to Eflow. After factoring in the obstacle for both our method and Eflow, our method is able to remain relatively close to the original demonstration, without large changes. In contrast, Eflow (repulsive) typically has a higher distance, particularly when an obstacle is reached, the repulsion results in sharper and larger changes relative to the original demonstration, while our method manages to diverge from the original demonstration in a smoother and more gradual manner. This is highlighted in the sharper peaks of trajectory distance in \cref{resultsFig} by Eflow (repulsive). The M.C.D. results in \cref{table:traj-distance} show that both our method and Eflow (repulsive) are able to significantly reduce collision, with ours performing slightly better. In addition, unlike our approach, Eflow with added repulsor does not have theoretical guarantees on stability. Our method produces non-colliding trajectories for the first two tasks, and is in contact with the obstacle briefly in the ``S''-drawing task, with a M.C.D. of 2mm. MPNet and MPNet-RRT produced trajectories that are less similar to given examples. Notably, the collision-avoidance capability of MPNet (without RRT-replanning) is significantly less than our approach, which indicates the difficulty to learn directly from collision-avoiding motions. This is particularly the case in an imitation learning setup, when the number of demonstrations is few. 

In \cref{Simres}, we illustrate generated motion in both the absence and in the presence of the added obstacles. We see in all cases, with the addition of the obstacle, the manipulator warps smoothly around the cylinder obstacle, while maintaining a similar shape required the original task upon passing the obstacles. This is particularly pronounced in both the lift-box and draw ``S'' tasks, where the upwards movement to lift box and the general shape of the ``S'' character is largely intact.


\subsubsection{Evaluation on Real Robot}
To evaluate the robustness of our method, we repeat the three tasks on a real-world JACO manipulator, placing obstacles in the path of demonstrations. For all three tasks, the JACO arm was able to warp around the obstacle with no visible collision, and successfully complete the task. A video of the real-world experiments can be found at \url{https://youtu.be/LoLQ0bzfw9I}. 

\subsection{Generalised User-Influenced Motion Reproduction}\label{Userinfluenced}
We also demonstrate the ability diffeomorphic transforms possess to be built based on user specifications. A simulated case studied of warping a dynamic system trained on imitations of drawing ``J'' characters, based user specified points, is illustrated in \cref{flowchart2}. We observe by crafting diffeomorphic, the original system is morphed to bias the designated positions. 

\begin{figure}[t]                                                    
\setlength{\fboxsep}{0pt}%
\setlength{\fboxrule}{1pt}%
    \centering                                                       
    \begin{subfigure}{0.245\linewidth}                   
        \fbox{\includegraphics[clip,width=.99\linewidth]{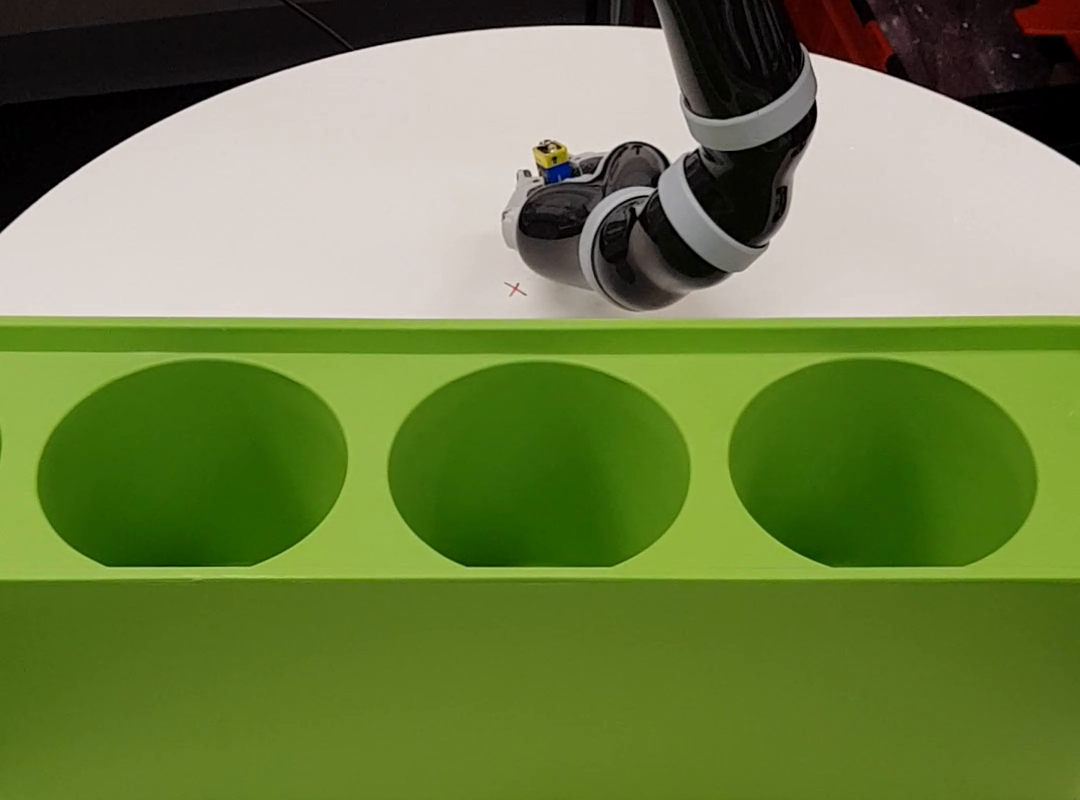}}
        \caption{Start position}
    \end{subfigure}%
    \begin{subfigure}{0.245\linewidth}                  
        \fbox{\includegraphics[clip,width=.99\linewidth]{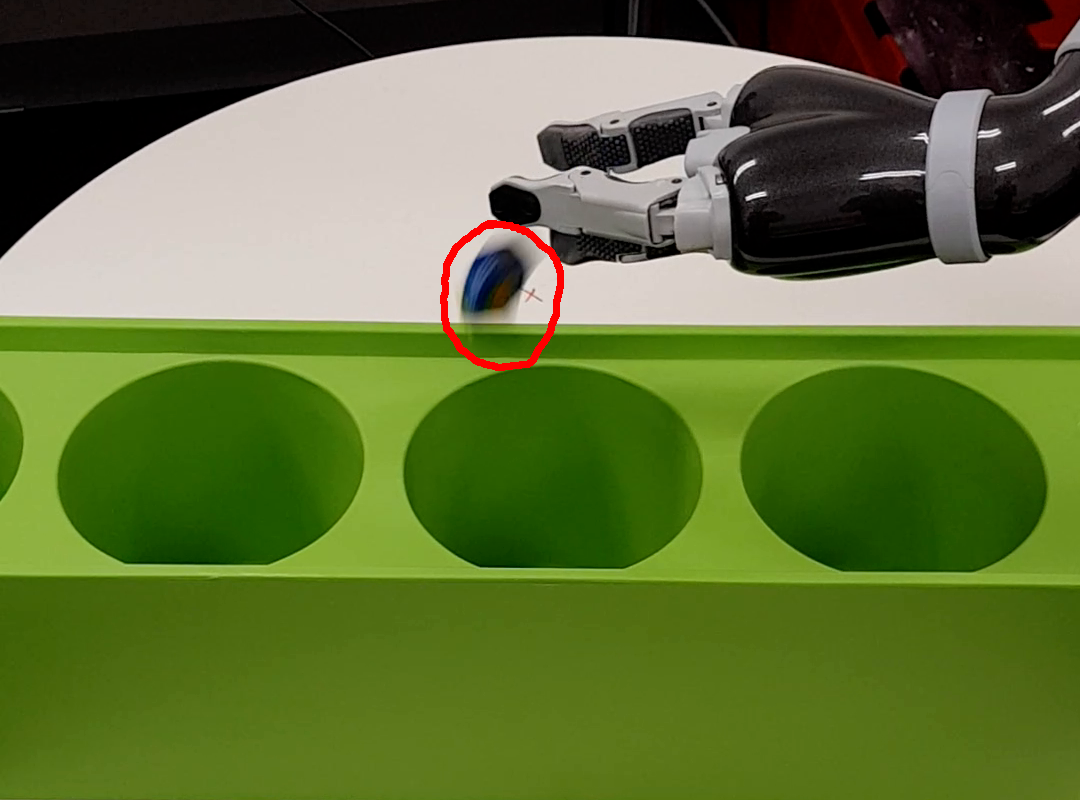}}
    \caption{center }
    \end{subfigure}%
    \begin{subfigure}{0.245\linewidth}                  
        \fbox{\includegraphics[clip,width=.99\linewidth]{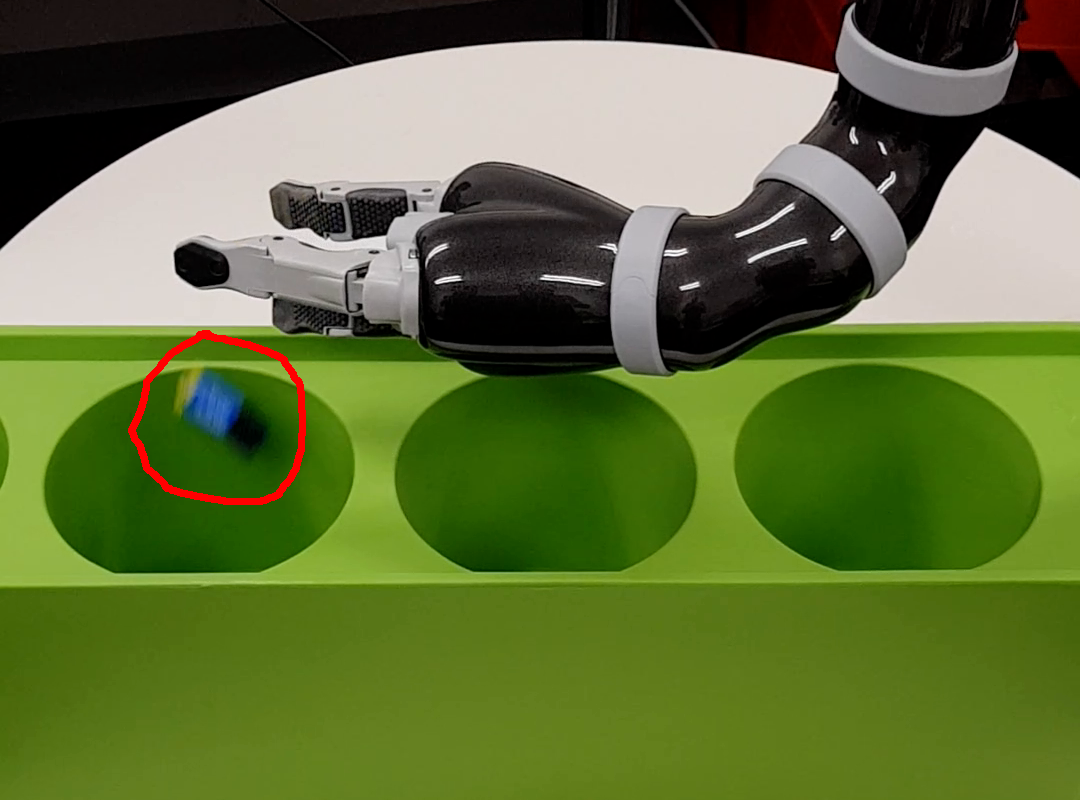}}
    \caption{Left pot}
    \end{subfigure}%
        \begin{subfigure}{0.245\linewidth}                  
        \fbox{\includegraphics[clip,width=.99\linewidth]{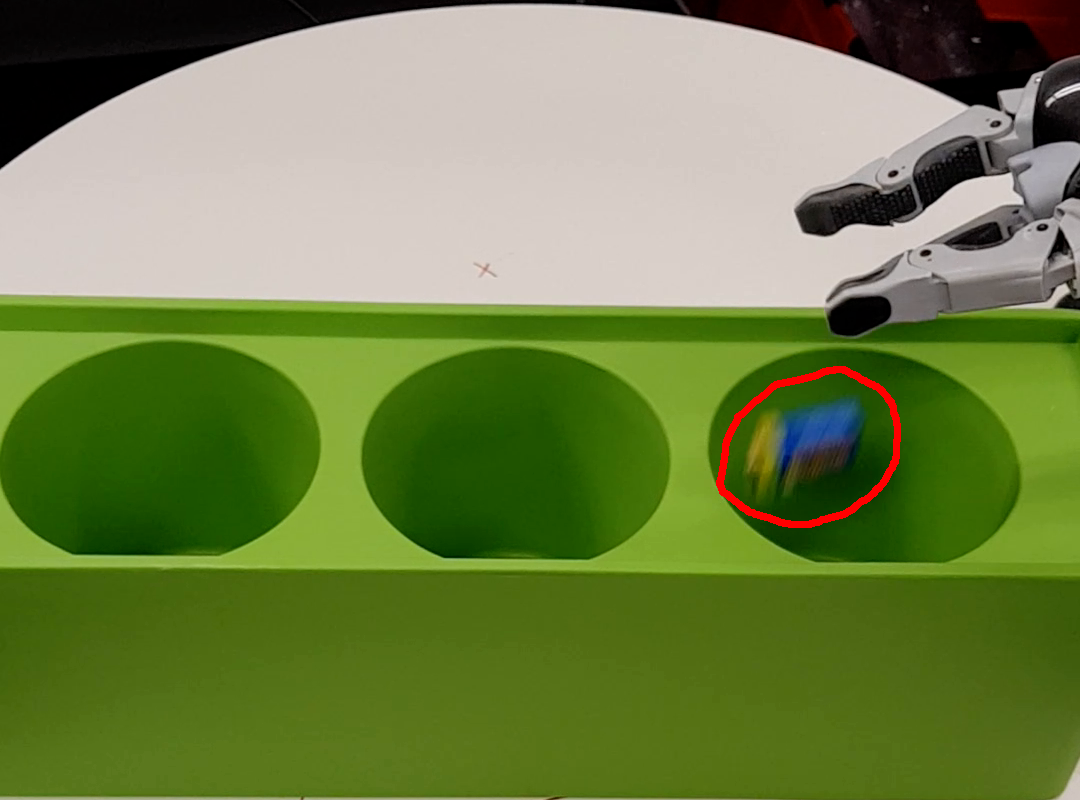}}
        \caption{Right pot}

    \end{subfigure}%
    \caption{      
We provide 8 demonstrations of dropping moving over the center pot to drop an object, we can create user-specific diffeomorphisms to warp the reproduced movement to drop the object in the left or right pots. Left-most image shows starting position. The dropped objects are circled in red.
    }\label{GeneralisedDrop}                                                            
\end{figure}

We extend our experiments to real-world robots and evaluate the ability of the dynamical system to be biased in user-specified positions with the JACO: We provide demonstrations of moving on top of a pot to drop an object. Then, we craft diffeomorphic transforms to morph the motion to either side such that the reproduced motion drops the object in neighbouring pots. This is performed by following the outline in \cref{bias} and setting the position above the neighbouring pots as biased coordinates. Qualitative evaluations are provided in \cref{GeneralisedDrop}. We see that without additional demonstrations, by applying diffeomorphic transforms, we can shift the robot motion to neighbouring pots. The objects were then successfully dropped in the corresponding pots. 

\section{Summary}
Diffeomorphisms provide a principled method to transform dynamical systems such that stability properties are not altered. We utilise this property and propose a framework which generalises imitation learning in scenarios where changes in the environment renders collected demonstrations infeasible due to. We modularise robot motion using diffeomorphic transforms, where each diffeomorphic transform encodes specific desirable behaviours, including imitating demonstrations, avoiding obstacles, and incorporating simple specified biases. When these individual transforms are sequentially applied to a known stable system the resulting system inherits the stability properties of the known system. We demonstrate the capabilities of this novel approach on a range of imitation learning tasks in both simulation and on a real robot.

In the next \cref{chap8}, we shall contribute a method for generating reactive and also global motion in a motion planning and control problem setup. We additionally develop a self-supervised learning framework to warm-start and speed-up the motion generation process.

\pagebreak

\chapter{Motion Generation with Geometric Fabric Command Sequences}\label{chap8} \blfootnote{This chapter has been accepted and will appear in ICRA as \cite{GeoFab_gloabL_opt}.}
\section{Introduction}
Motion generation is a central problem in robotics. It is concerned with finding collision-free and executable trajectories from start to goal. Generating robot motions in dynamic environments is particularly challenging: the motion needs to be reactive to changes, and also be intuitive to any spectating humans. Recent reactive approaches to motion generation, such as Riemannian Motion Policies \citep{RMPs} and Geometric Fabrics \citep{geoFabs}, produce local policies to efficiently generate smooth and natural motions. These approaches produce a policy that find the instantaneous optimal controls, and allow for perturbations and deviations in the environment. However, these locally optimal solutions are often globally infeasible, resulting in robots getting stuck in local minima. Sampling-based planners, such as RRTs \citep{rrts}, can find global solutions, but often produce non-intuitive paths, and typically require re-planning to handle deviations and local changes in the environment.  

In this work, we introduce \emph{Geometric Fabric Command Sequences} (GFCS), a new approach to motion generation that produces globally feasible solutions while retaining local reactivity. In broad strokes, our method solves the global motion generating problem by optimising over a sequence of attractor states, which we call \emph{commands}, for a \emph{Geometric Fabric} policy \citep{geoFabs}. Geometric Fabrics produce local motion by combining decoupled dynamical systems, which each represent an individual behaviour, such as goal reaching, obstacle avoidance, and joint limit handling. We propose to directly apply global optimisation over the commands. Furthermore, we hypothesise the solutions of the optimisation are transferable over different problem setups. We take a \emph{self-supervised} approach and learn, over a dataset of motion generation problems and solutions, to warm-start the optimisation procedure. Specifically, we use an implicit generative model to progressively learn to recommend candidate solutions, as more and more problems are solved, with solutions of past optimisation runs used as training labels. This allows for faster and higher quality solutions as more problems are solved. As time progresses, the implicit model itself can accurately generate optimal solutions, without further optimisation. 

Concretely, we contribute: (1) \emph{Geometric Fabric Command Sequences} (GFCS), a formulation of global and reactive motion generation as optimisation over sequentially connected commands, inputted to \emph{Geometric Fabrics} \citep{geoFabs}; (2) \emph{Self-supervised GFCS}, a learning approach to warm-start the optimisation, using self-labelled data. Thus, improving the speed and quality of motion generation.

\begin{figure}[H]
\centering
\begin{subfigure}{0.2\textwidth}
    \frame{\includegraphics[width=\textwidth]{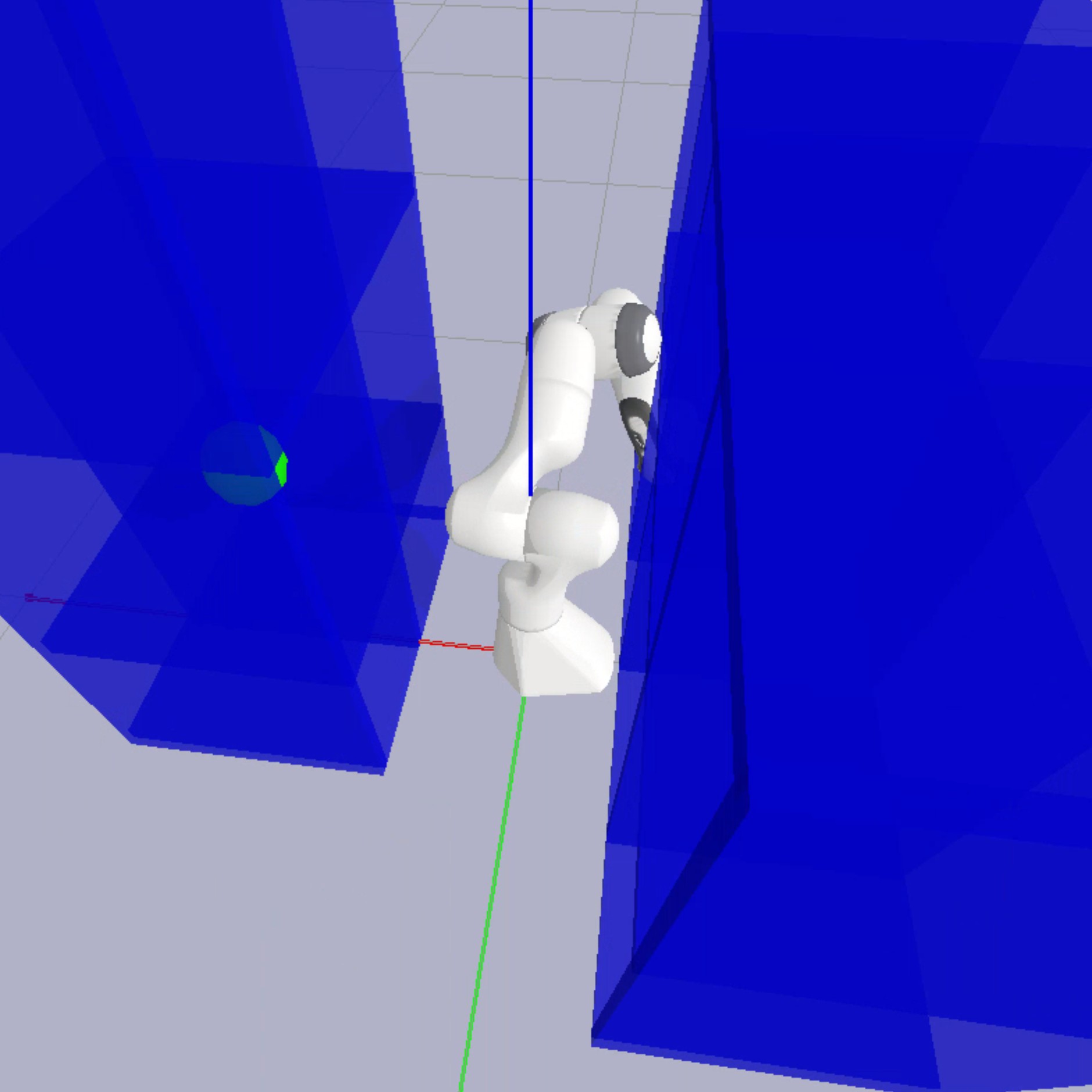}}
\end{subfigure}%
\begin{subfigure}{0.2\textwidth}
    \frame{\includegraphics[width=\textwidth]{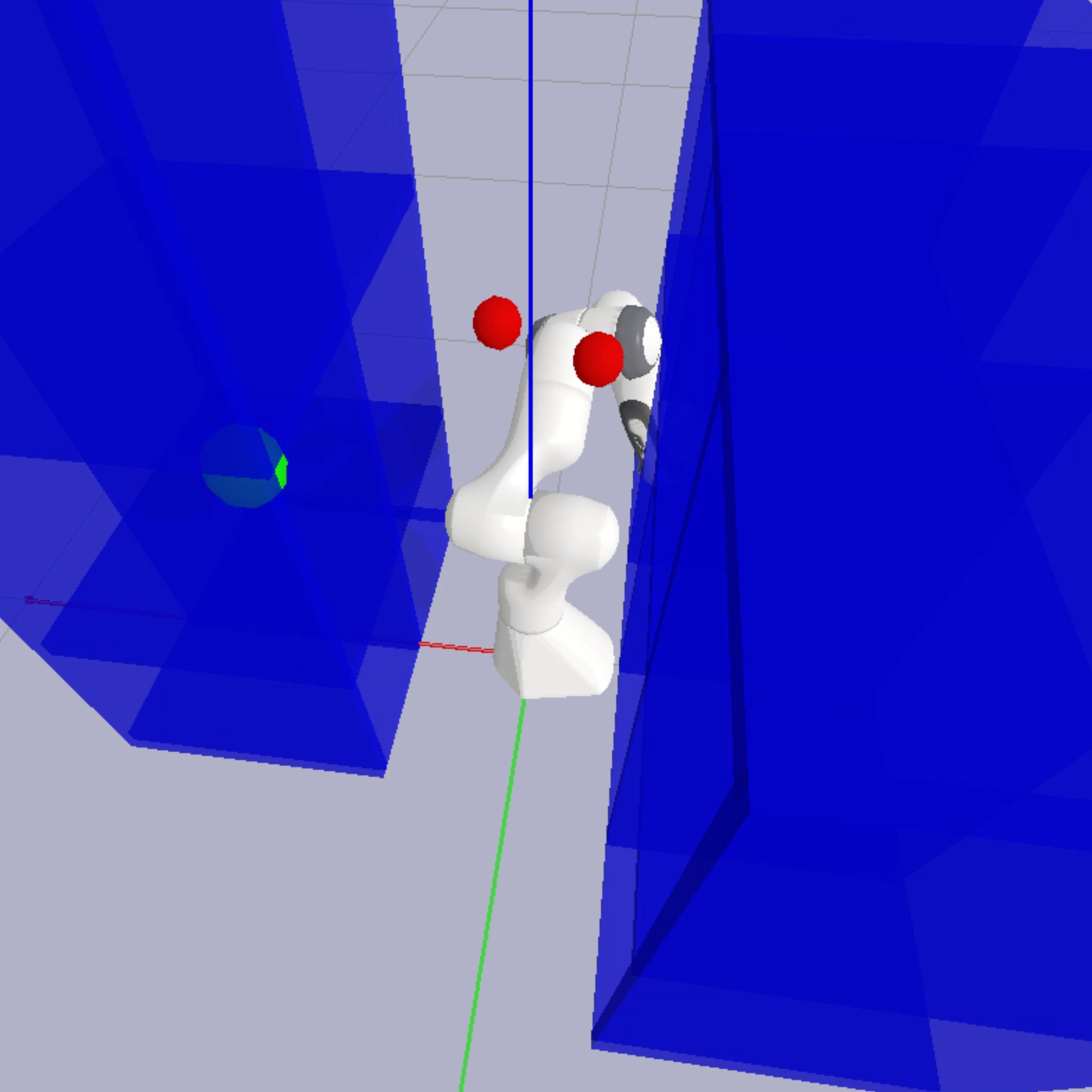}}
\end{subfigure}%
\begin{subfigure}{0.2\textwidth}
    \frame{\includegraphics[width=\textwidth]{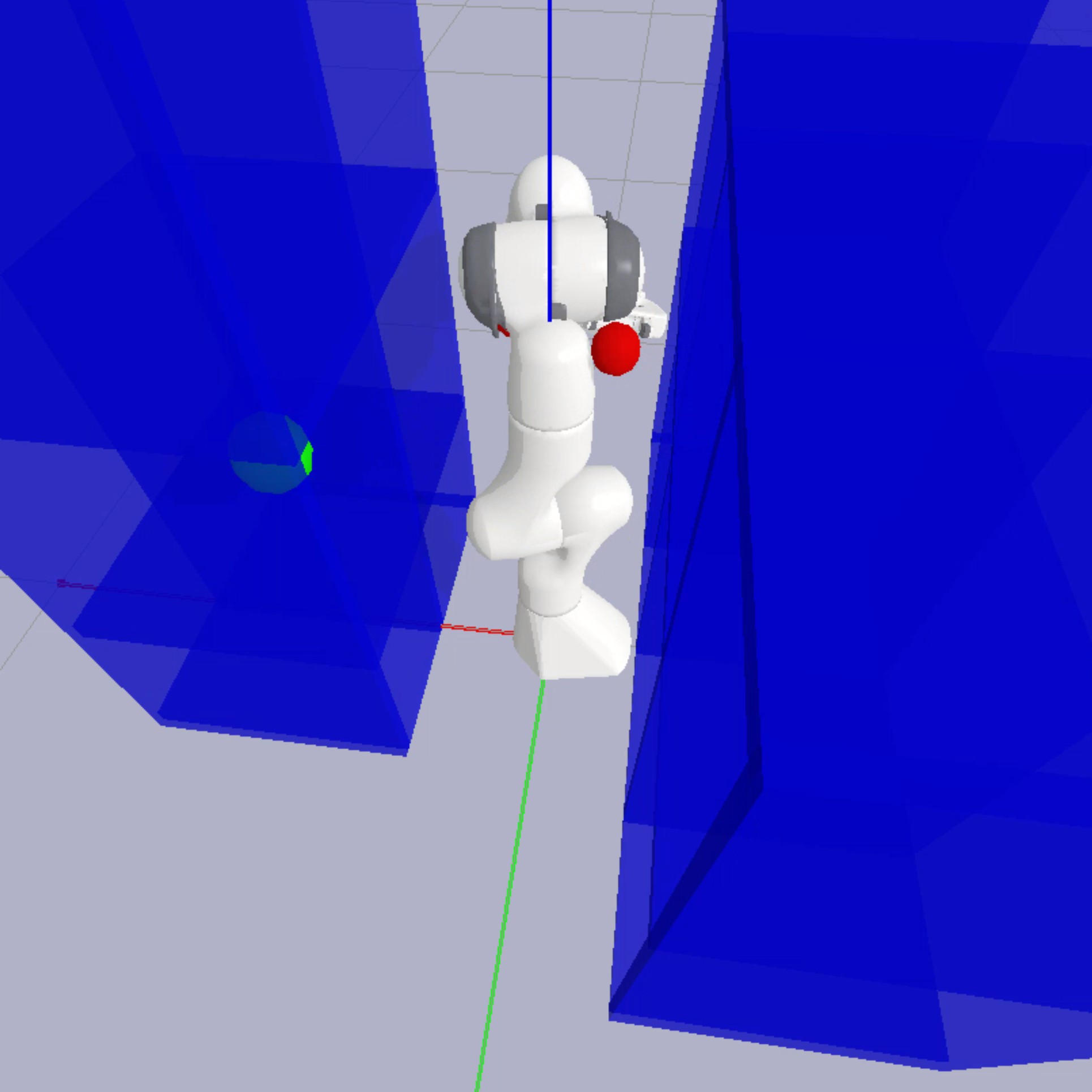}}
\end{subfigure}%
\begin{subfigure}{0.2\textwidth}
    \frame{\includegraphics[width=\textwidth]{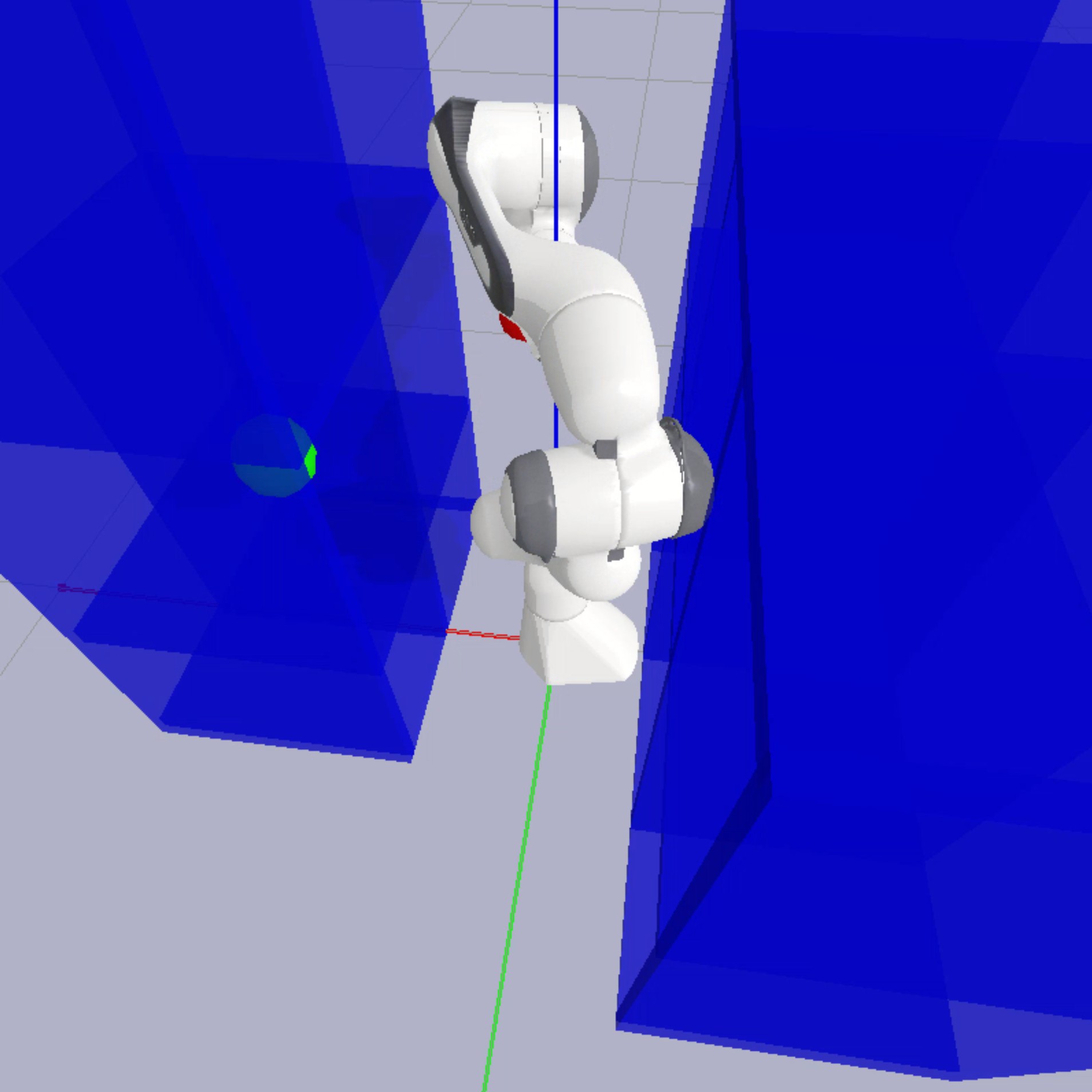}}
\end{subfigure}%
\begin{subfigure}{0.2\textwidth}
    \frame{\includegraphics[width=\textwidth]{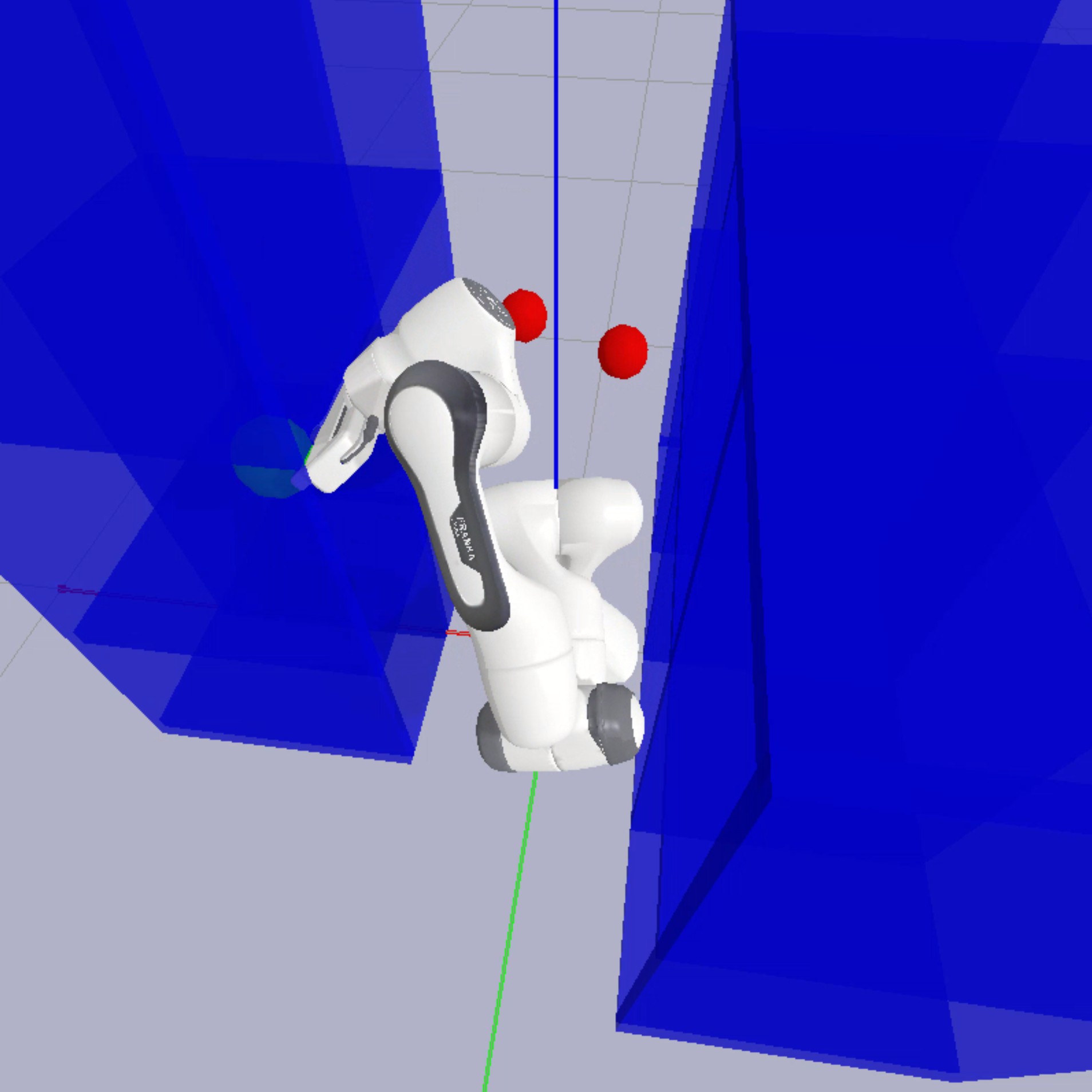}}
\end{subfigure}%

\begin{subfigure}{0.2\textwidth}
    \frame{\includegraphics[width=\textwidth]{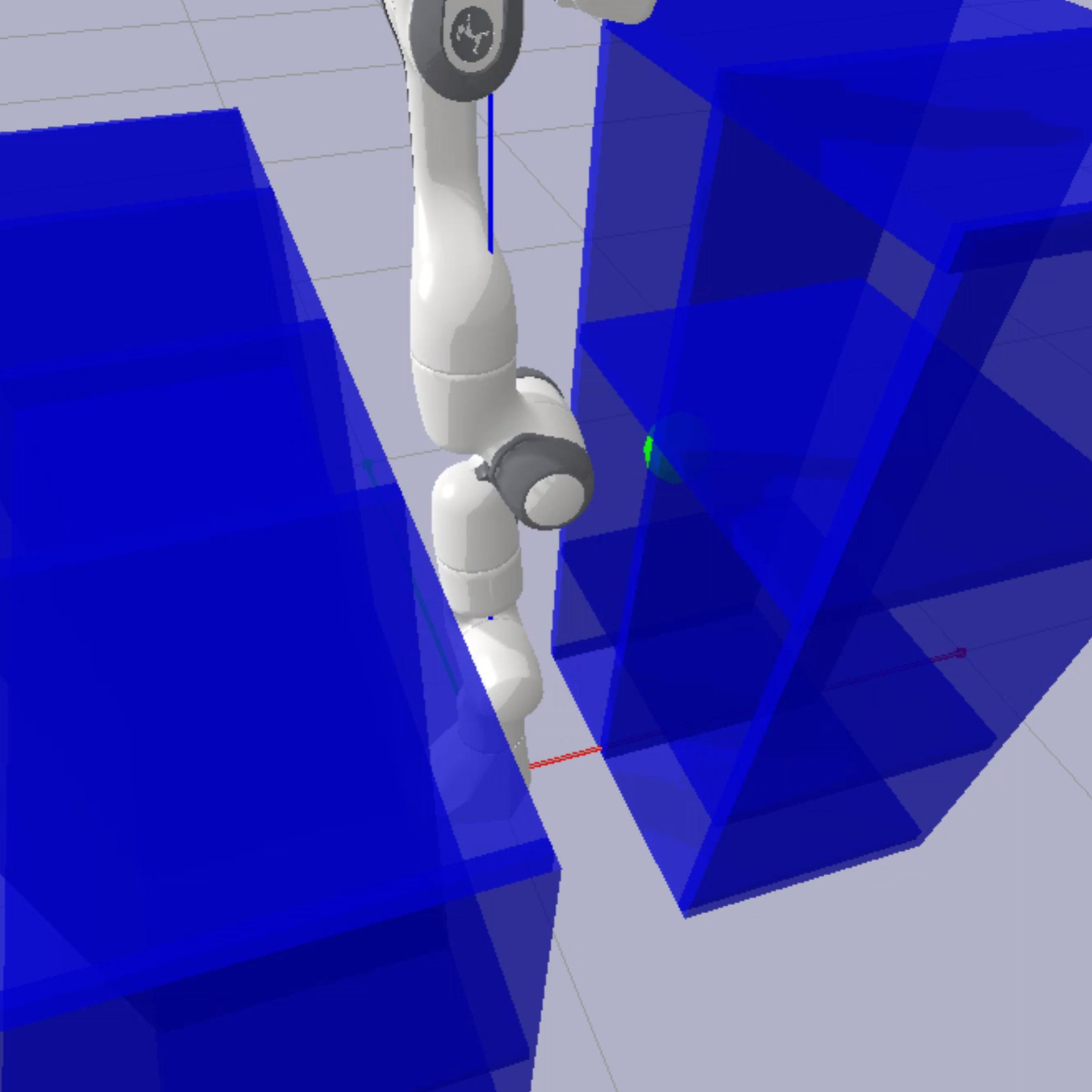}}
\end{subfigure}%
\begin{subfigure}{0.2\textwidth}
    \frame{\includegraphics[width=\textwidth]{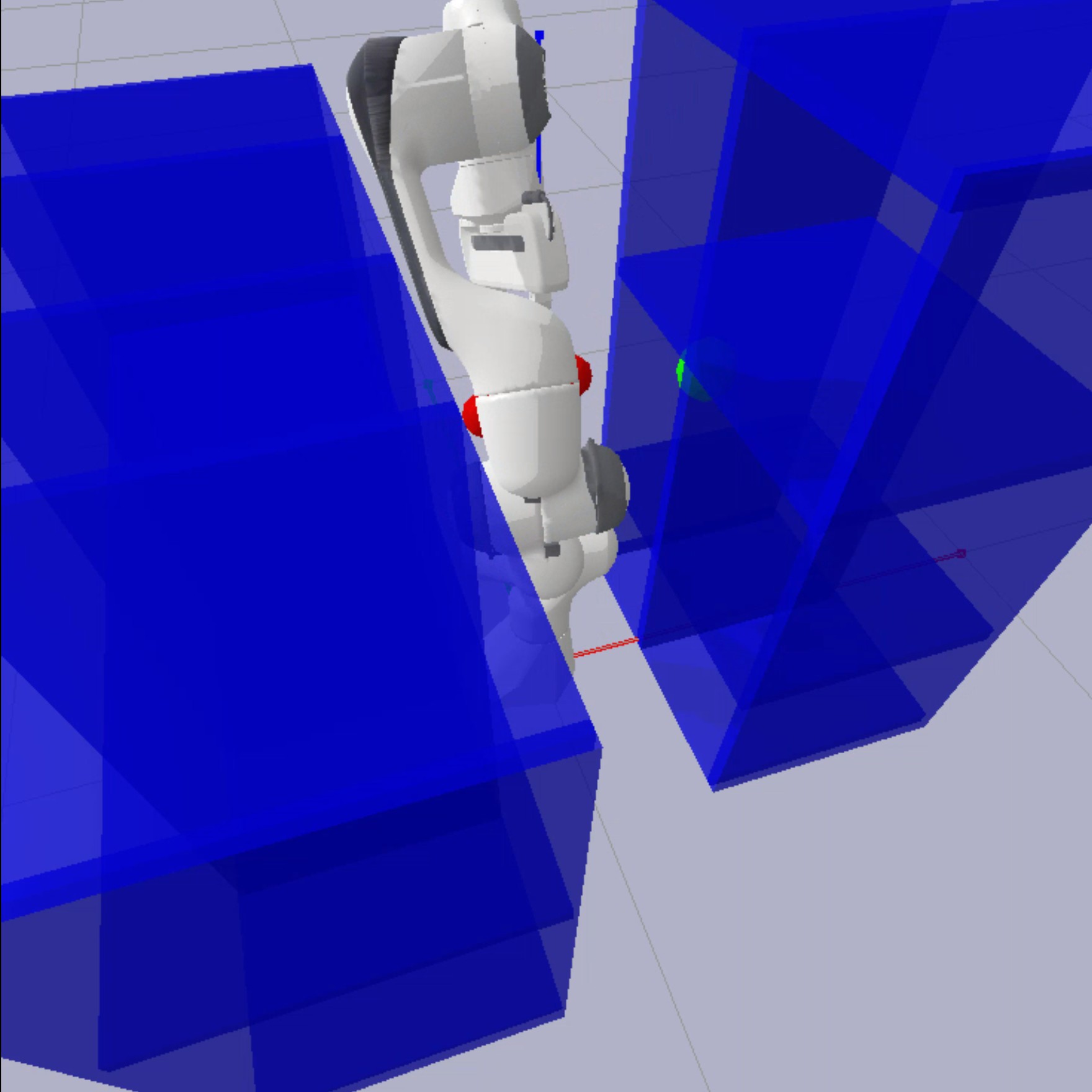}}
\end{subfigure}%
\begin{subfigure}{0.2\textwidth}
    \frame{\includegraphics[width=\textwidth]{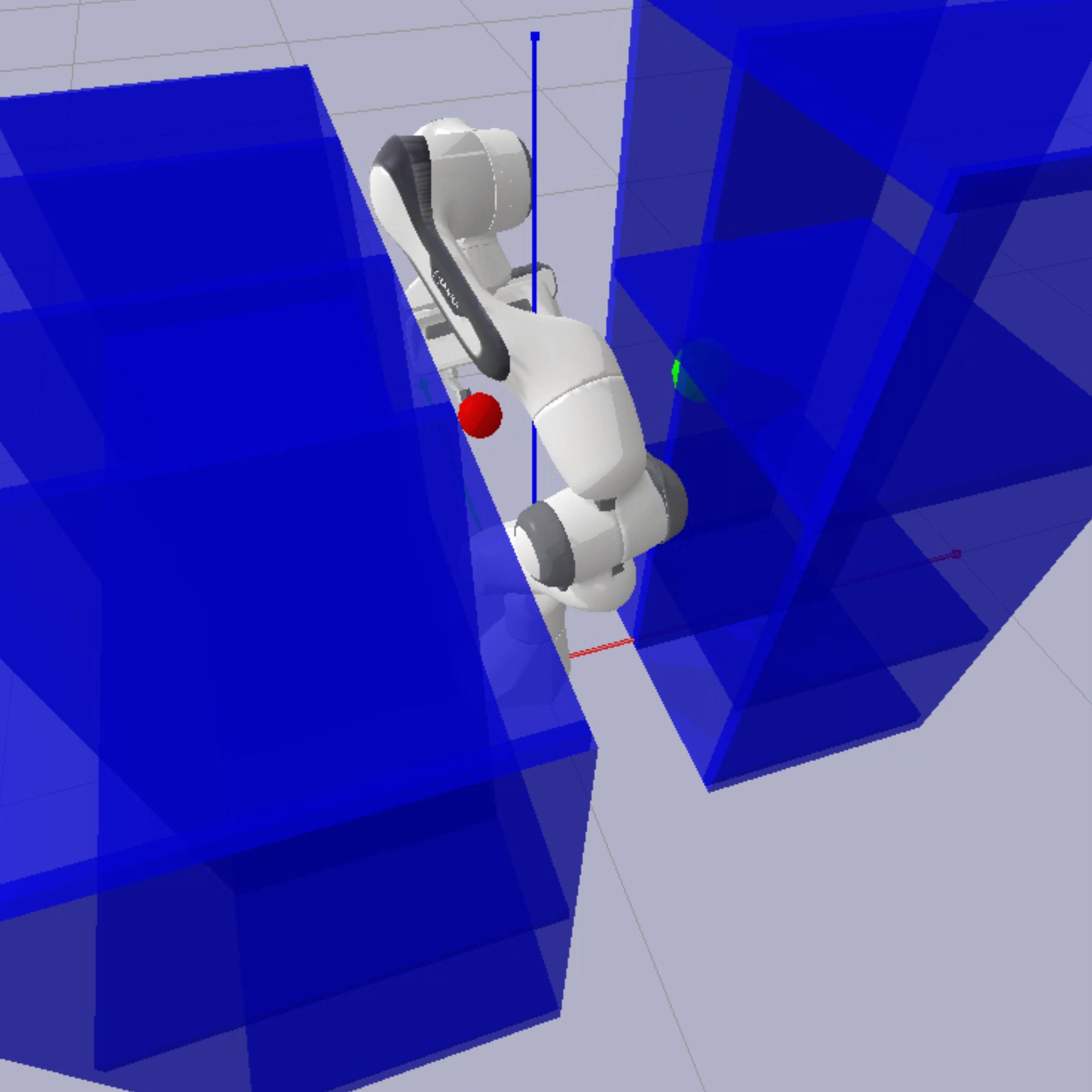}}
\end{subfigure}%
\begin{subfigure}{0.2\textwidth}
    \frame{\includegraphics[width=\textwidth]{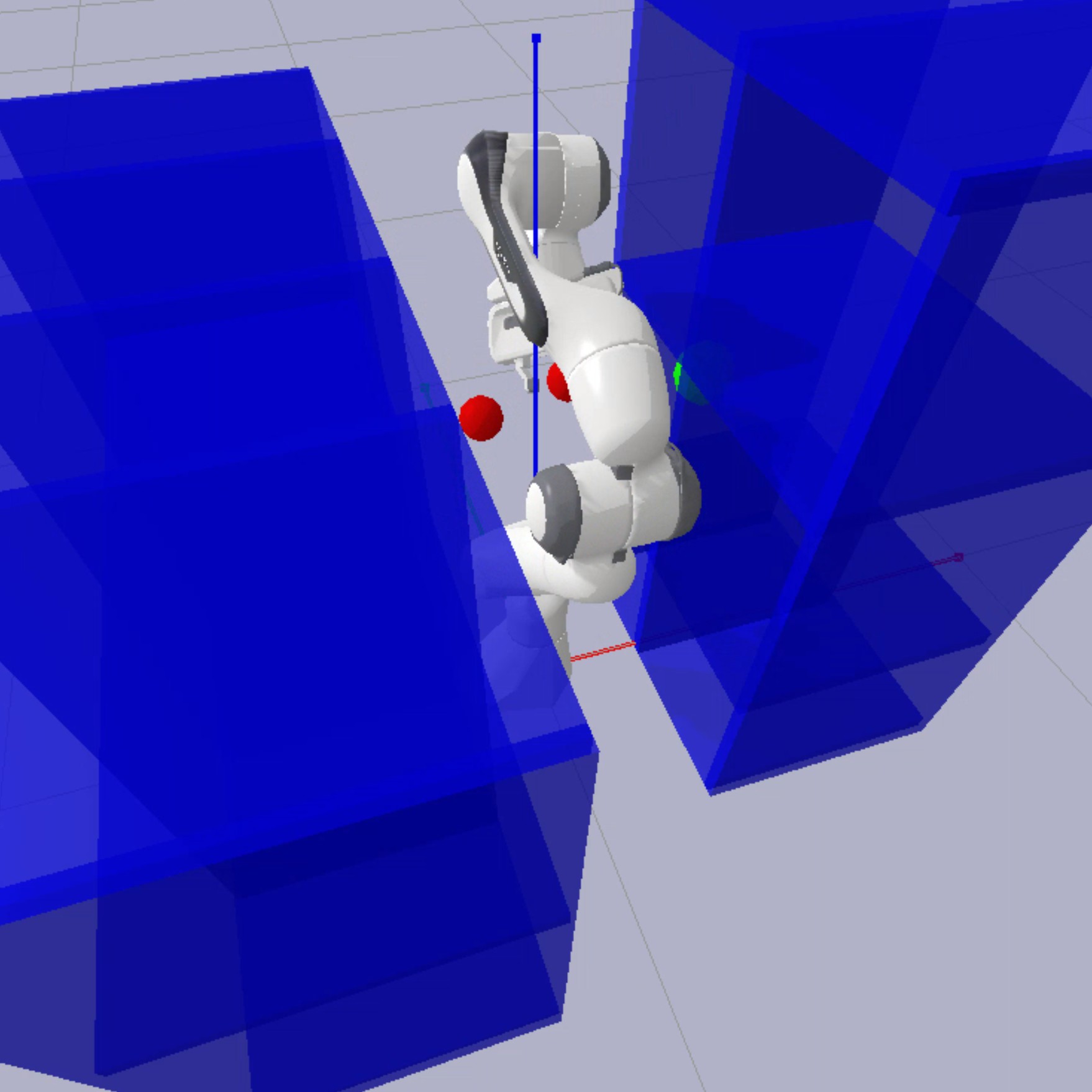}}
\end{subfigure}%
\begin{subfigure}{0.2\textwidth}
    \frame{\includegraphics[width=\textwidth]{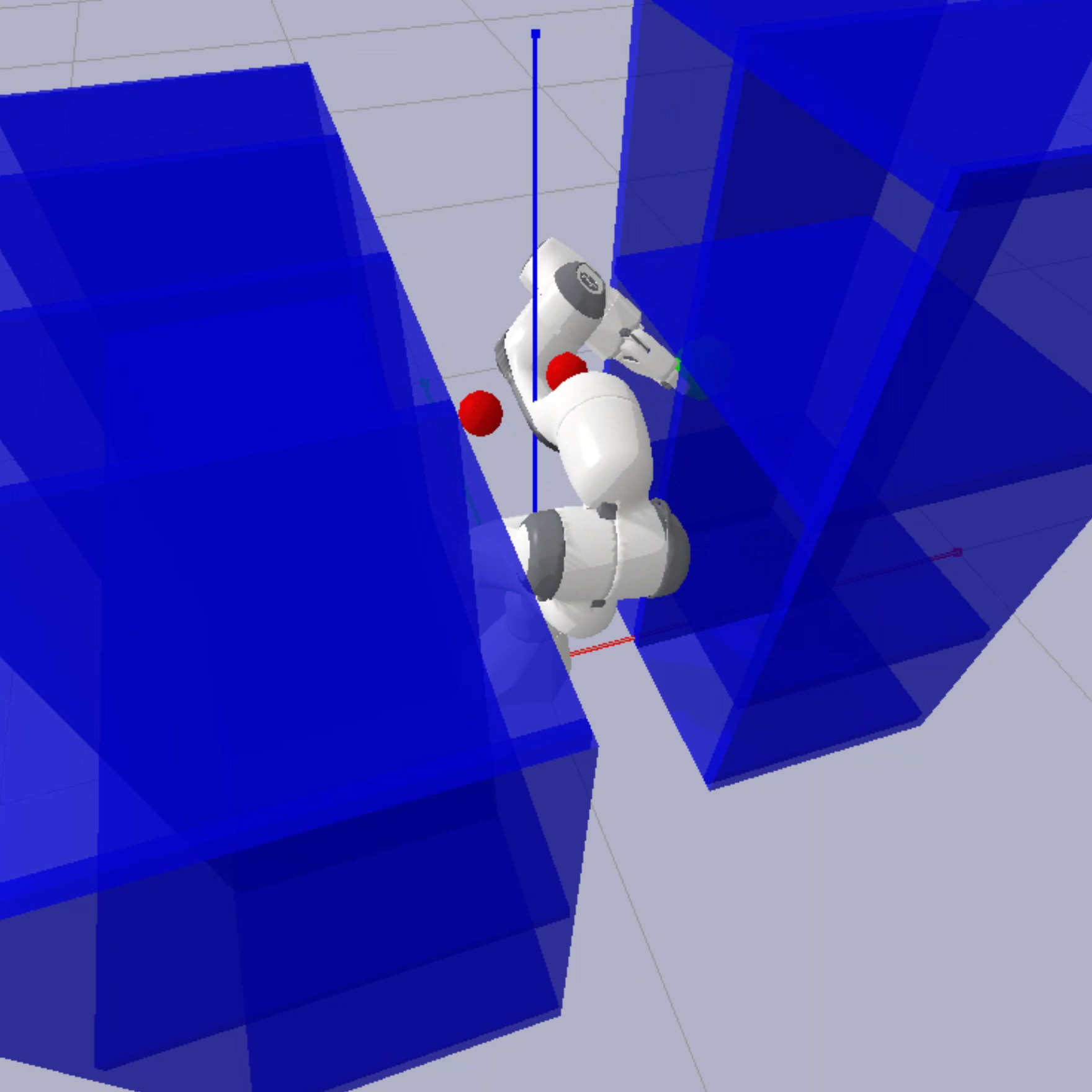}}
\end{subfigure}%

\begin{subfigure}{0.2\textwidth}
    \frame{\includegraphics[width=\textwidth]{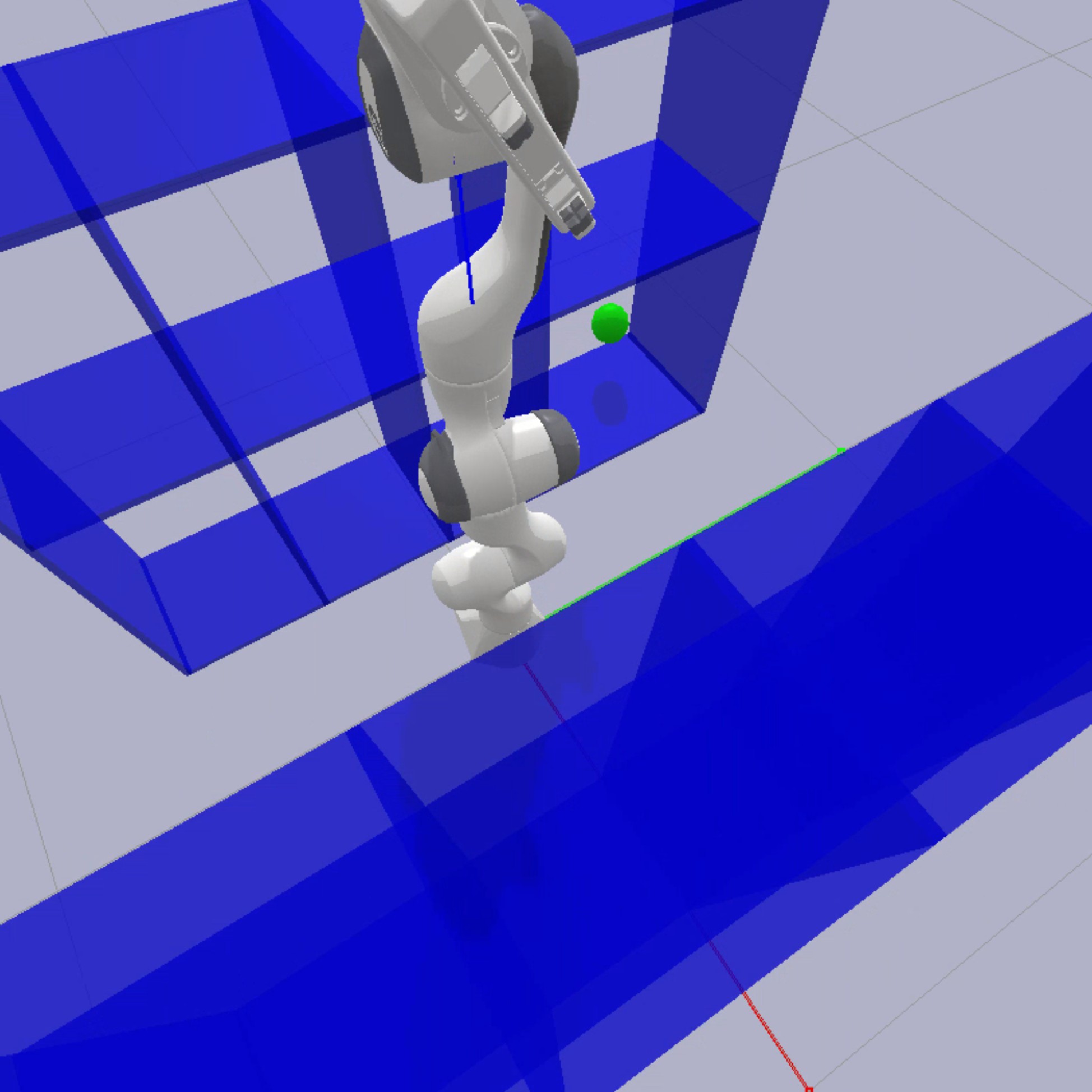}}
\end{subfigure}%
\begin{subfigure}{0.2\textwidth}
    \frame{\includegraphics[width=\textwidth]{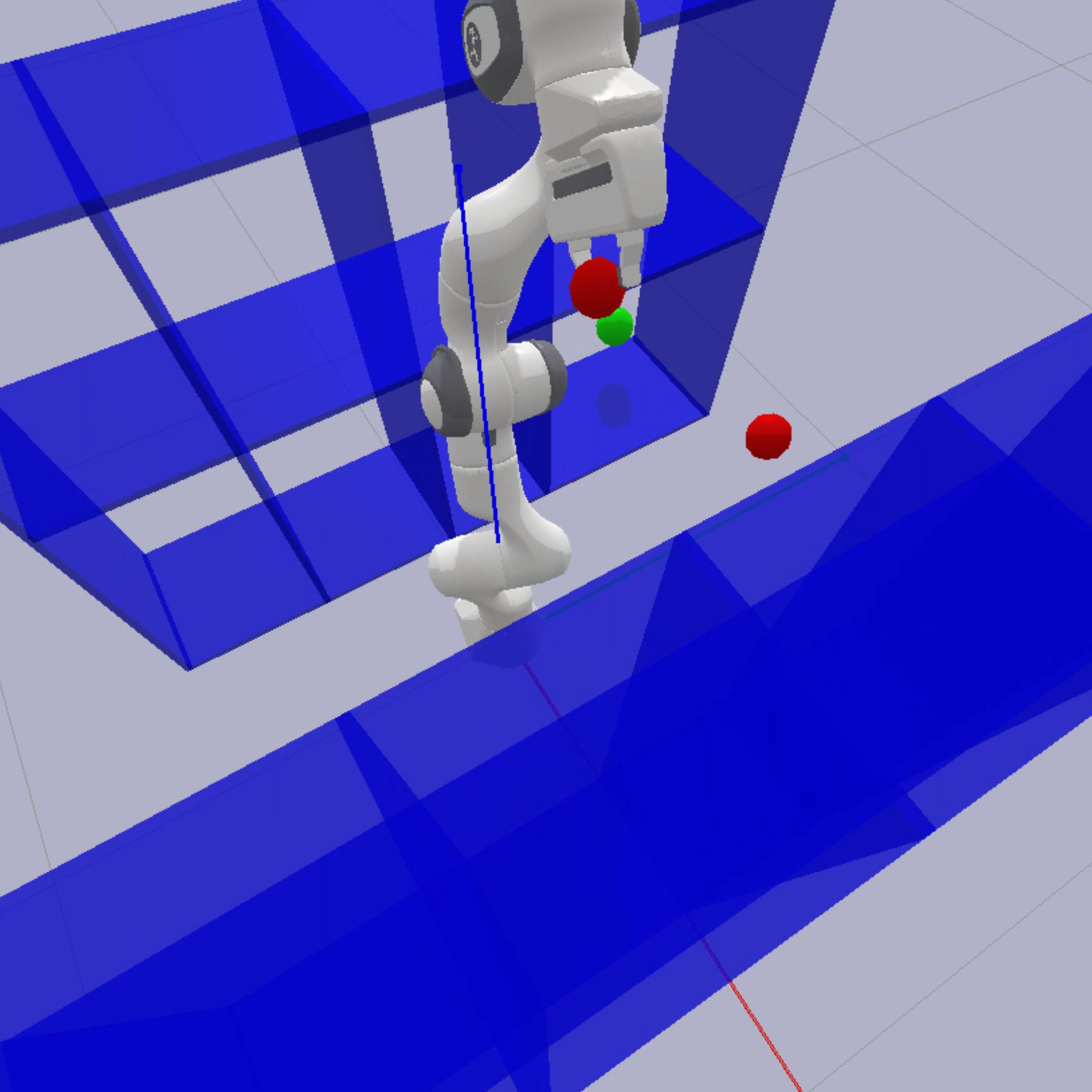}}
\end{subfigure}%
\begin{subfigure}{0.2\textwidth}
    \frame{\includegraphics[width=\textwidth]{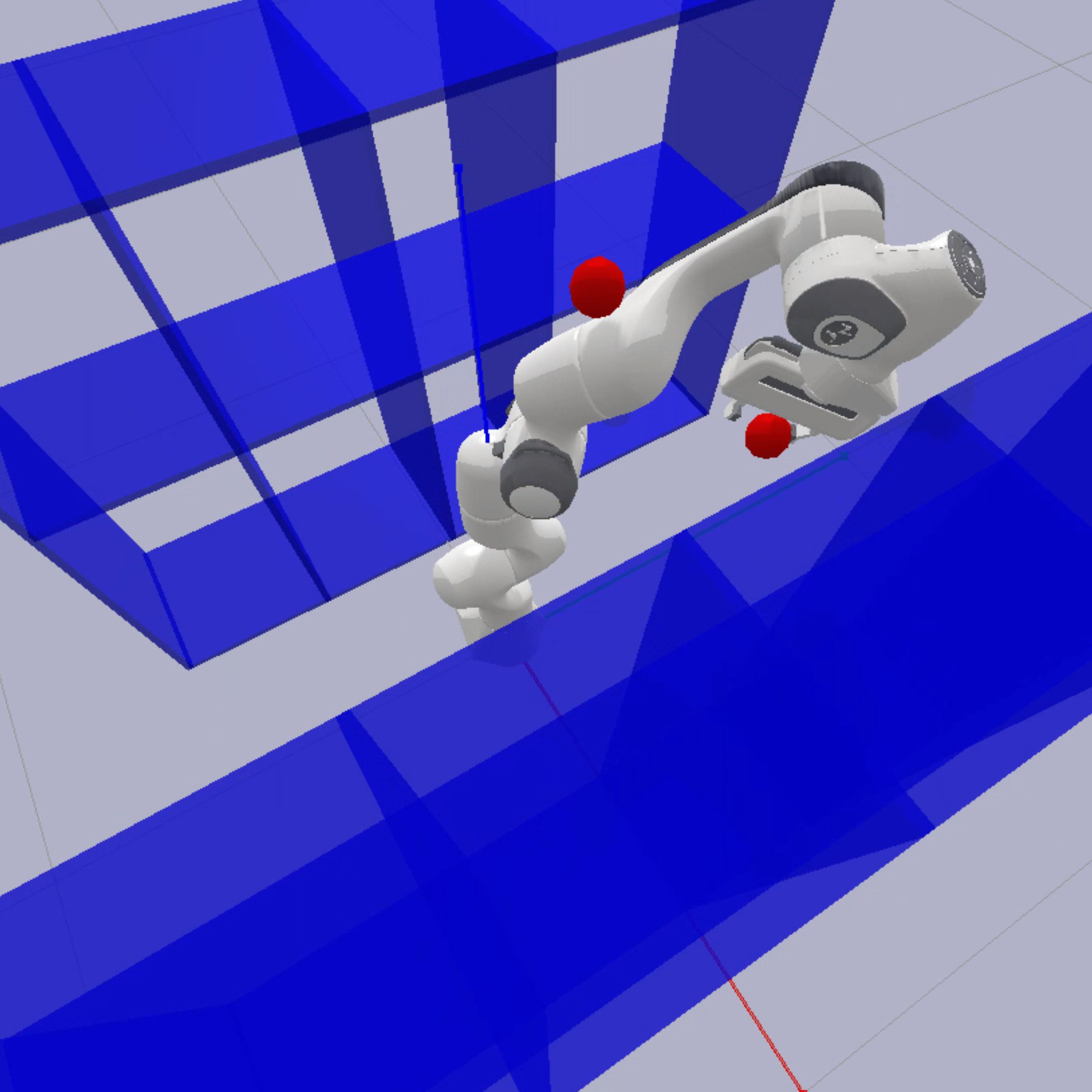}}
\end{subfigure}%
\begin{subfigure}{0.2\textwidth}
    \frame{\includegraphics[width=\textwidth]{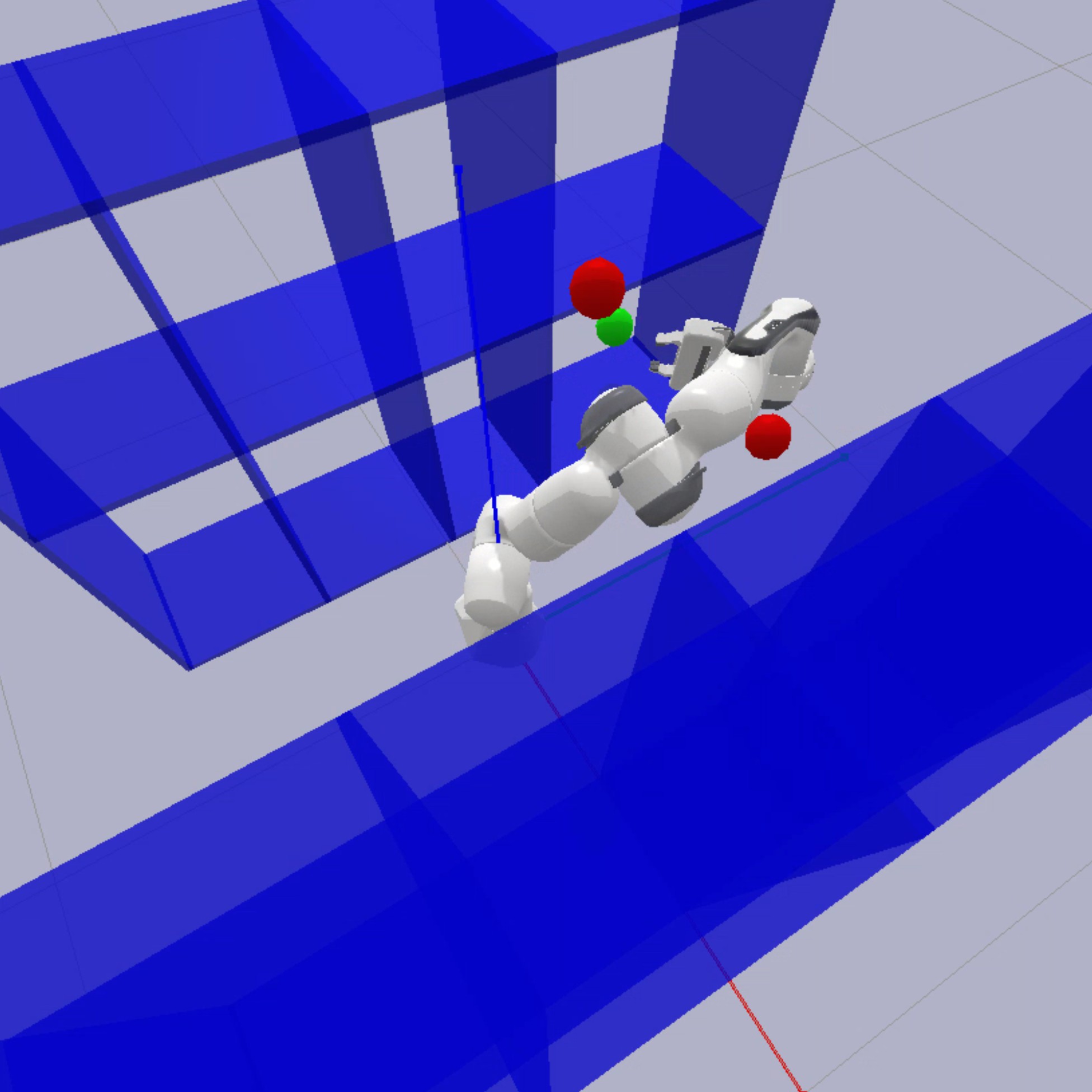}}
\end{subfigure}%
\begin{subfigure}{0.2\textwidth}
    \frame{\includegraphics[width=\textwidth]{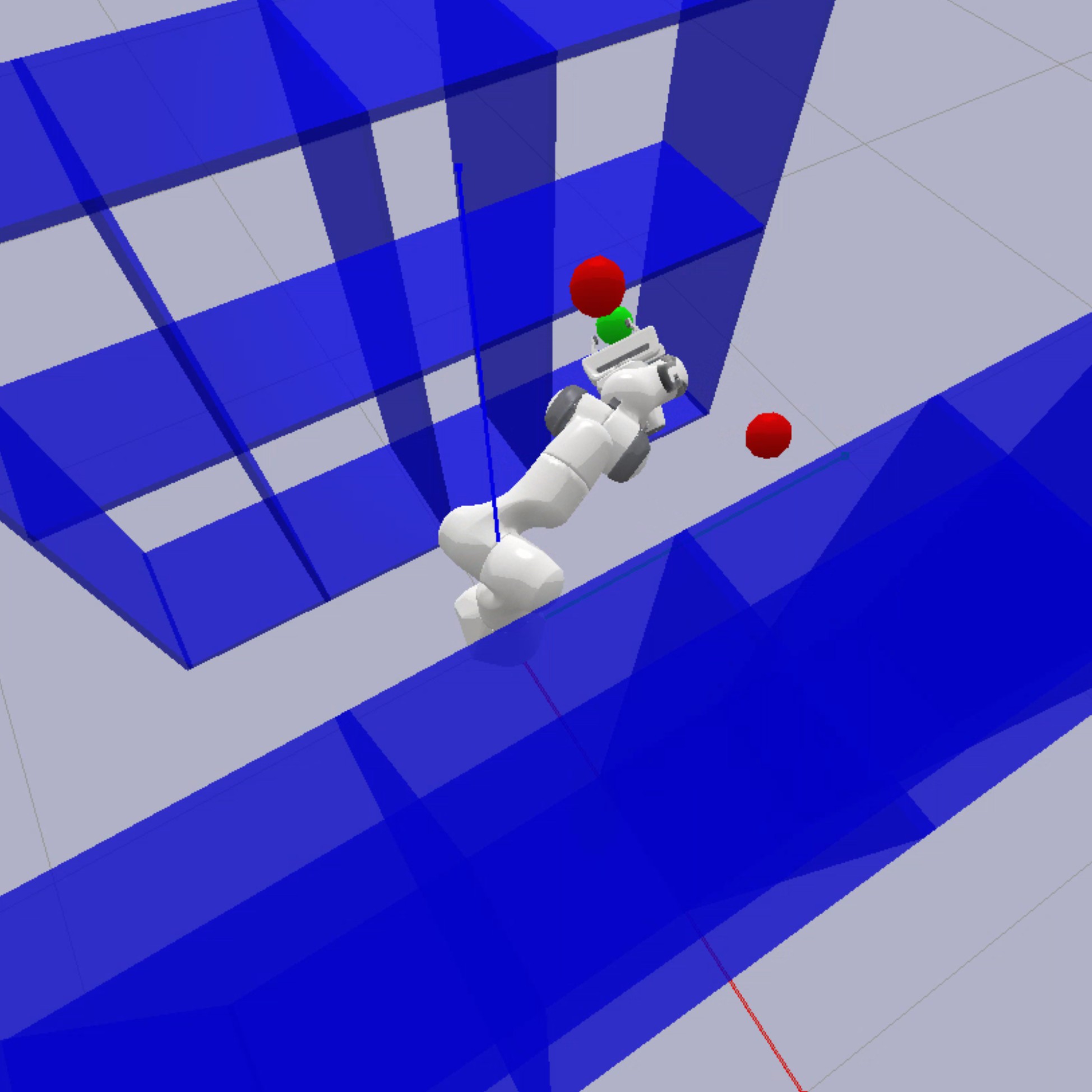}}
\end{subfigure}%

\begin{subfigure}{0.2\textwidth}
    \frame{\includegraphics[width=\textwidth]{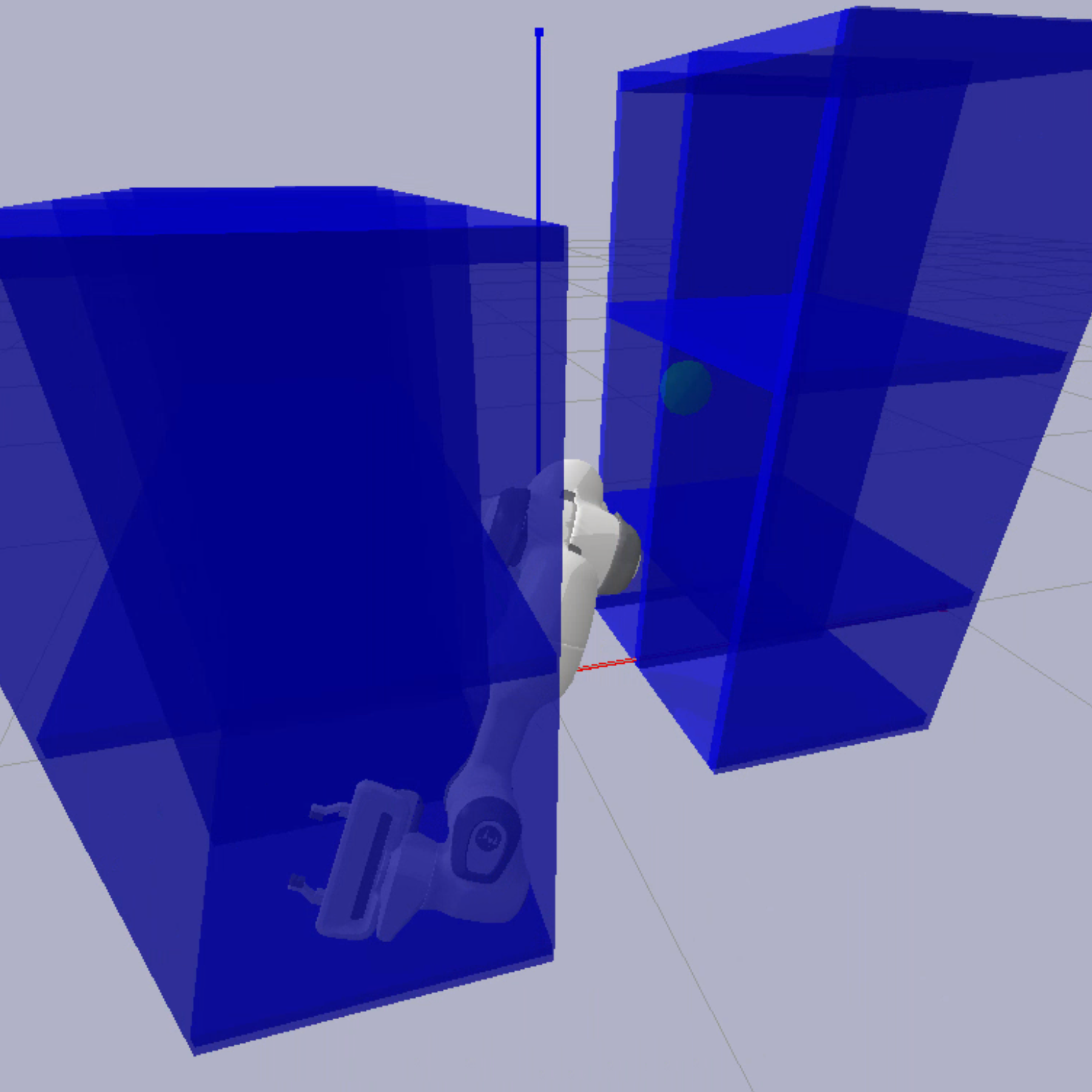}}
\end{subfigure}%
\begin{subfigure}{0.2\textwidth}
    \frame{\includegraphics[width=\textwidth]{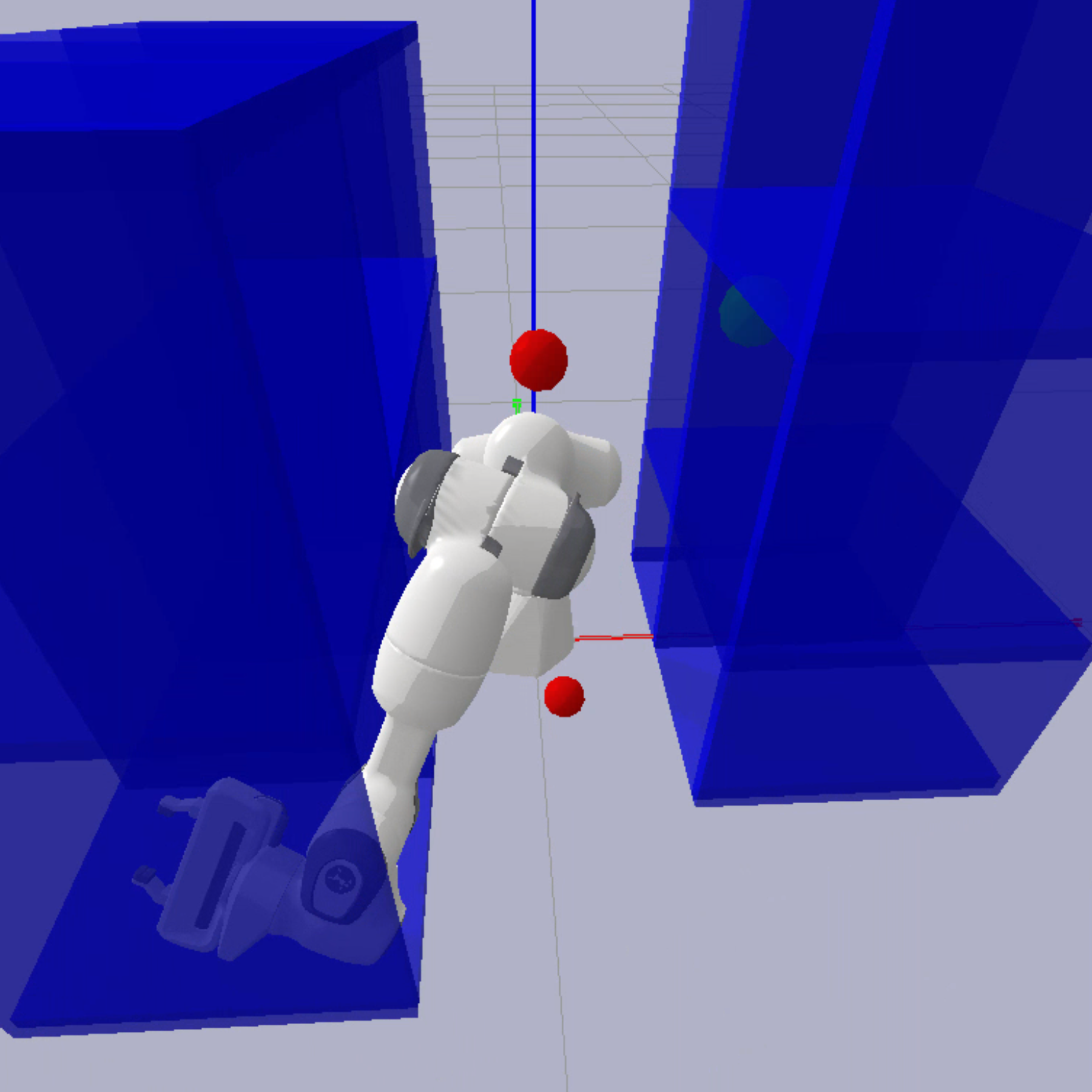}}
\end{subfigure}%
\begin{subfigure}{0.2\textwidth}
    \frame{\includegraphics[width=\textwidth]{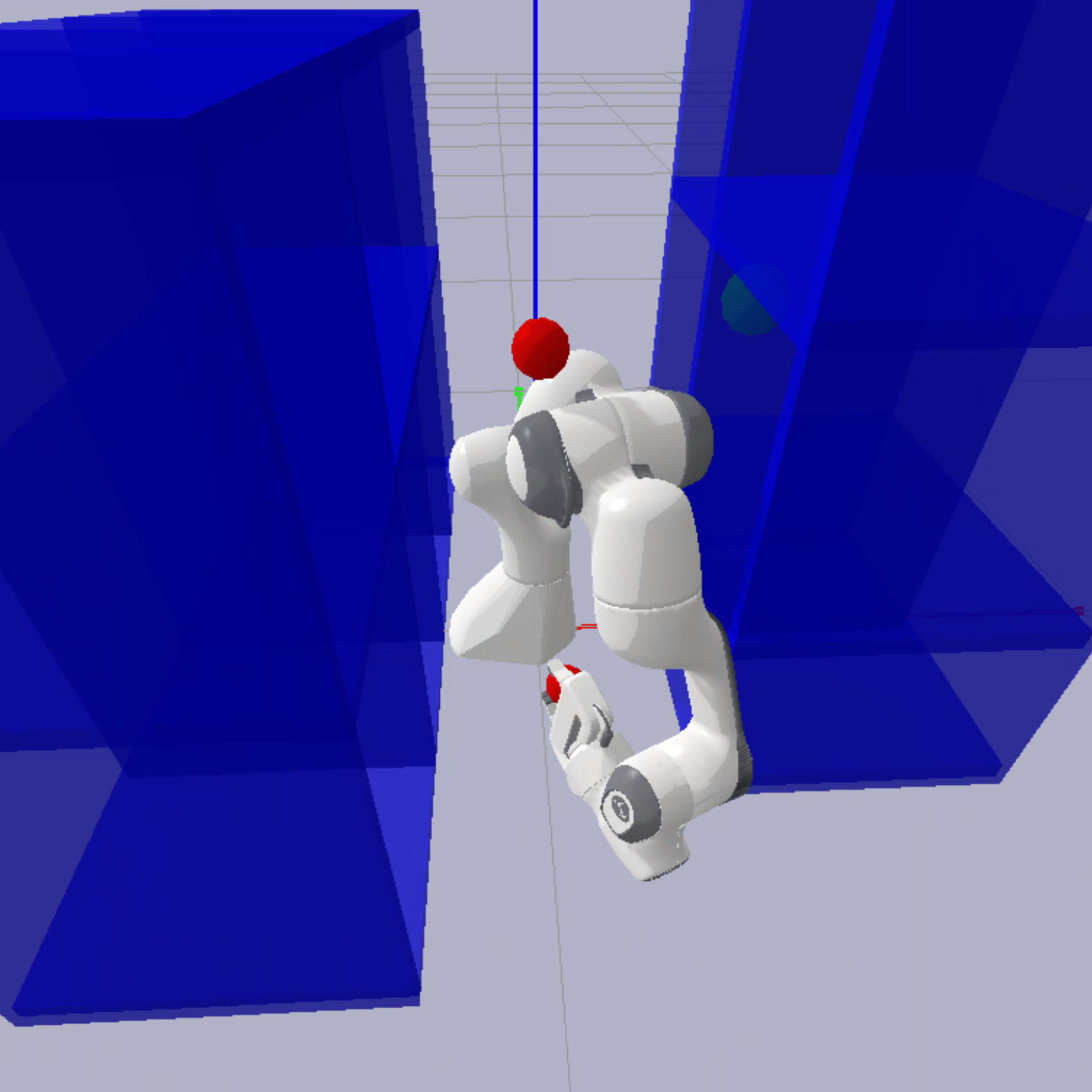}}
\end{subfigure}%
\begin{subfigure}{0.2\textwidth}
    \frame{\includegraphics[width=\textwidth]{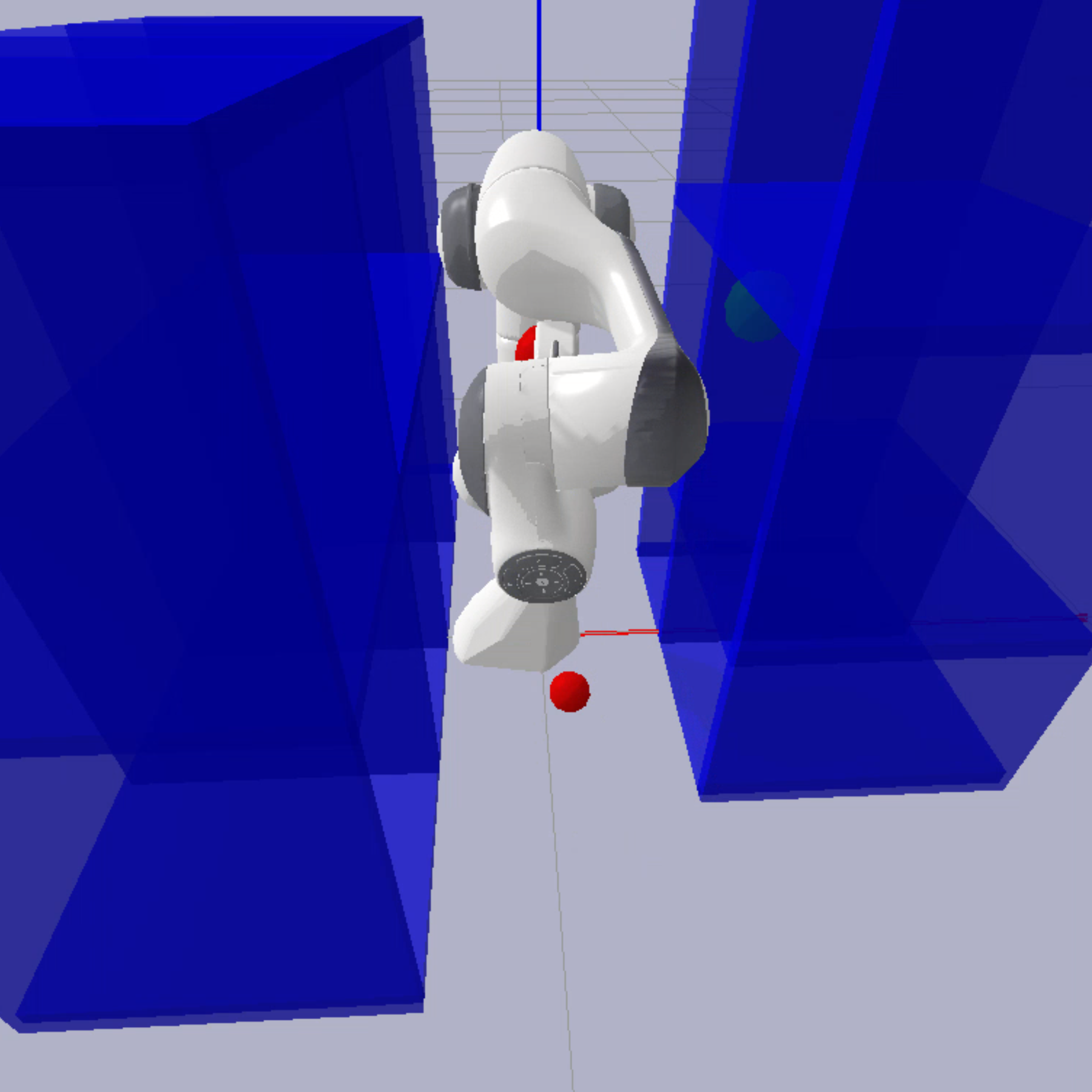}}
\end{subfigure}%
\begin{subfigure}{0.2\textwidth}
    \frame{\includegraphics[width=\textwidth]{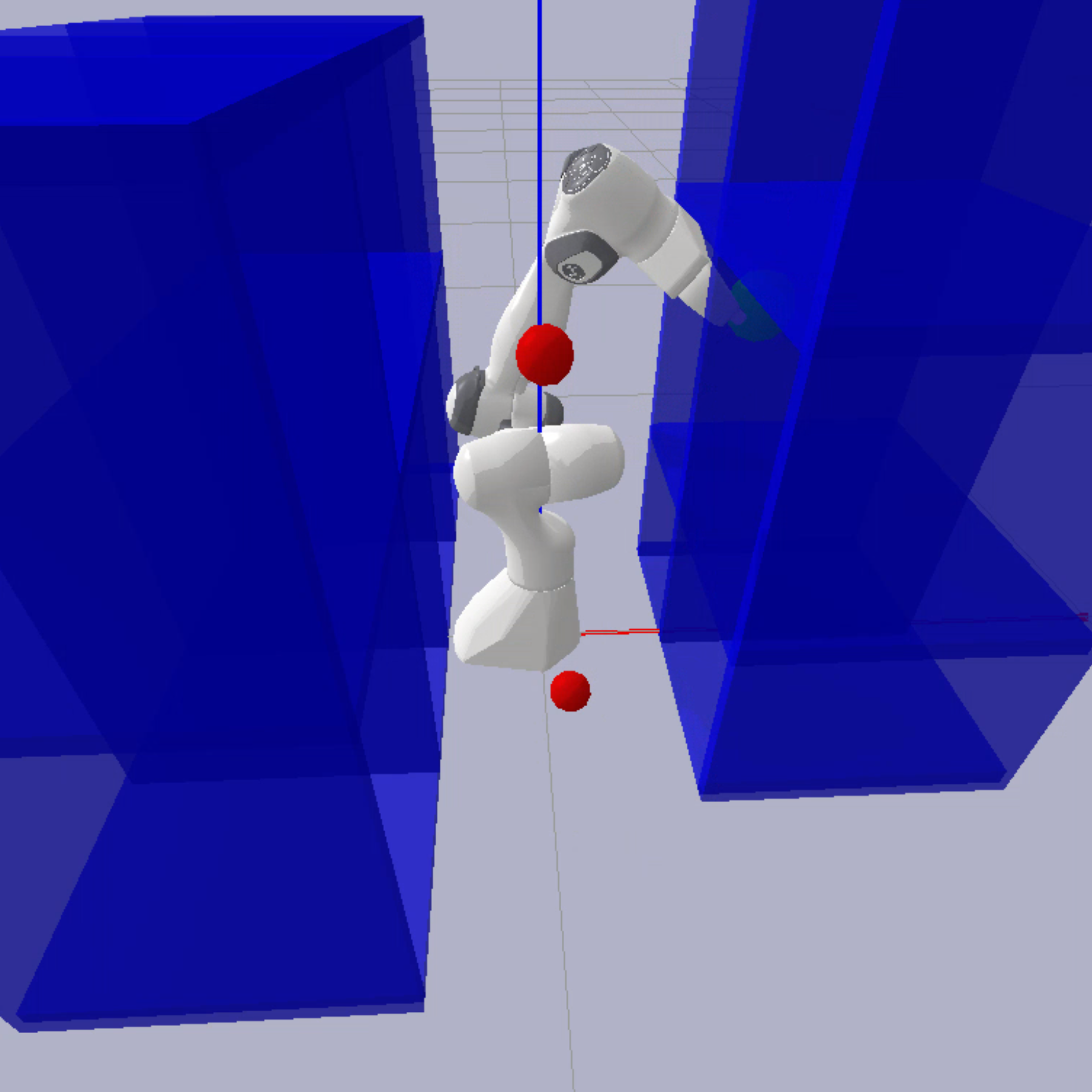}}
\end{subfigure}%
\caption{{A manipulator reaching from a cupboard to the goal (green sphere) in another. The environment is tightly-packed, as the cupboards sandwich the manipulator. Previous local reactive methods fall into local minima, while our approach finds a sequence of \emph{commands} to produce reactive motion to the goal. The end-effector coordinates of the command states are shown in red spheres.}}\label{fig:img_seq}
\end{figure}

\section{Related Work}

\subsubsection{Reactive Motion Generation} 
Our method is built atop reactive methods for motion generation. Reactive methods take into account the local environment and the current robot state to compute the optimal immediate next control. Methods in this category include \cite{potential,Ijspeert2013DynamicalMP,RMPs,geoFabs}. These approaches are able to react to changes in the environment, perturbations, and sensor uncertainty, as the instantaneous best control is found at every timestep. However, these methods can often get stuck in local minima when the workspace geometry is non-convex.

\subsubsection{Probabilistic Complete Motion Planning:} Probabilistic Complete Motion Planning lies on the other end of the spectrum, where planners aim to find a path (represented as waypoints) from start to goal configuration, by iteratively sampling and maintaining a graph structure. As the number of sample points approaches infinity, the probability of finding a path, if it exists, tends to one. Representative approaches in this category include PRMs \citep{PRMs}, RRTs \citep{rrts}, as well as subsequent variants such as RRT$^{*}$ \citep{rrtstar}, Bi-RRT \citep{birrt}. Like our proposed method, these motion planners aim to find a globally optimal solution rather than immediate optimal controls. However, methods in this category cannot reactively adapt to changes, and often produce non-smooth, non-intuitive motion with drastic swings. 

\subsubsection{Learning to Plan} 
Learning approaches to aid planning have recently gained popularity. Machine learning models can be used to memorise solutions of planning, by training over a large dataset containing problem setups, and with corresponding labelled solutions. During inference, the learning model can speed up or improve planning by suggesting promising candidates to use in the downstream planning algorithm. Motion planning networks \citep{MPNet} condition on the problem to iteratively predict deterministic waypoints. Further work in trajectory or sample generation take a probabilistic approach and draw samples from mixture \citep{OTNet}, normalising flow \citep{PDMP}, or variational autoencoder models \citep{Ichter2018LearningSD}. Similarly, our method also applies learning to aid motion generation. However, we tackle this in a self-supervised learning manner, where learning and motion generation is combined into the same loop. An implicit generative model capable of flexibly modelling unnormalised multi-modal distributions is applied.  

\section{Preliminaries: Geometric Fabrics}
\emph{Geometric Fabrics} (GFs) are the central building block of our method. Our work does not alter the machinery of GFs, but uses GFs as building blocks. For completeness, here we give an intuitive introduction to GFs, with further details and theoretical analysis available in \cite{geoFabs}. 

\subsubsection{Fabric terms} 
Geometric fabrics (GFs) modularise robot motion by representing each individual local behaviour as position and velocity-dependent second order dynamical systems, referred to as \emph{fabric terms}. Examples of these local behaviour include: accelerating towards a specified goal; reactively avoiding an obstacle by getting ``repelled'' away; avoiding the robot joint limits. Each of these behaviours are defined independently, and may be defined in different task spaces. For example, obstacle avoidance is defined with respect to body points on the robot, while joint limits are defined in the configuration space. These fabric terms can then be fused into a single system. We denote the position in some task space as $\bb{x}$, its velocity and acceleration as $\dot{\bb{x}}$, $\ddot{\bb{x}}$. We denote a configuration belonging to a $d$-dimensional C-space $\mathcal{Q}\subseteq\mathbb{R}^{d}$ as $\bb{q}\in\mathcal{Q}$, its time derivatives as $\dot{\bb{q}}\in\mathbb{R}^{d}$, $\ddot{\bb{q}}\in\mathbb{R}^{d}$. Formally, each fabric term is defined by a pair $(\mathcal{L}_{e},\pi)$, where $\mathcal{L}_{e}(\bb{x},\dot{\bb{x}})\rightarrow \mathbb{R}$ is a \emph{Finsler energy} \citep{Ratliff2021GeneralizedNA}, and $\ddot{\bb{x}}=\pi(\bb{x},\dot{\bb{x}})$ is a policy. $\mathcal{L}_{e}$ satisfies: (1) $\mathcal{L}_{e}(\bb{x},\dot{\bb{x}})\geq 0$ with equality only when $\dot{\bb{x}}=0$;  (2) $\mathcal{L}_{e}(\bb{x},\lambda\dot{\bb{x}})=\lambda^{2}\mathcal{L}_{e}(\bb{x},\dot{\bb{x}})$, for $\lambda\geq 0$; (3) $\bb{M}=\partial^{2}_{\dot{\bb{x}}\dot{\bb{x}}}$ is positive definite, and $\pi$ satisfies $\pi(\bb{x},\lambda\dot{\bb{x}})= \lambda^{2} \pi(\bb{x},\dot{\bb{x}})$ for $\lambda\geq 0$. Here, the policy $\pi$ defines time-independent paths, an operation $\mathtt{Energize}_{\mathcal{L}_{e}}(\pi)$ is then used to select an execution velocity, and $\bb{M}$ is a priority metric.

\subsubsection{Fusion} 
Fabric terms defined in different task spaces can be fused, by first finding the configuration space coordinates of each term, in a \emph{transform tree} \citep{rmpflow} manner. Suppose, from each fabric term, we have a set of $N_{ft}$ triples, $\{(\phi_{i},\bb{M}_{i},\ddot{\bb{x}}_{i})\}_{i=1}^{N_{ft}}$, where $\phi_{i}$ maps a $\bb{q}$ to task space coordinates $\bb{x}_{i}$. The combined acceleration, $\ddot{\bb{q}}$, is given by:
\begin{equation}
\ddot{\bb{q}}=\Big(\sum_{i=1}^{N_{ft}}\bb{J}_{i}^{\top}\bb{M}_{i}\bb{J}_{i}\Big)^{-1}\Big(\sum_{i=1}^{N_{ft}}\bb{J}_{i}^{\top}\bb{M}_{i}(\ddot{\bb{x}}_{i}-\dot{\bb{J}}_{i}\dot{\bb{q}})\Big),
\end{equation}
where $\bb{J}_{i}$ is the Jacobian of $\phi_{i}$, $\dot{\bb{J}}_{i}$ its time derivative.

\section{Self-supervised Learning of Geometric Fabric Command Sequences}
We shall first formulate motion generation as a global optimisation over a sequence of \emph{commands} (\cref{opt_section}). Then, we develop a generative model to \emph{learn} to optimise (\cref{learn_section}), and a self-supervised learning framework that encompasses both optimisation and learning (\cref{ssl:subsection}).

\subsection{Geometric Fabrics Command Optimisation} \label{opt_section}
Let us consider how Geometric Fabrics can be used to generate locally optimal motion trajectories. A combined Geometric Fabric, $\ddot{\bb{q}}=f(\bb{q},\dot{\bb{q}},|\bb{q}_{0},\dot{\bb{q}}_{0},\bb{q}_{g},E,C)$, can be viewed as a policy that maps from $\bb{q}$ and $\dot{\bb{q}}$ to $\ddot{\bb{q}}$. It is also dependent on problem-specific parameters: the start configuration $\bb{q}_{0}\in\mathcal{Q}$ and velocity $\dot{\bb{q}}_{0}\in\mathbb{R}^{d}$, goal $\bb{q}_{g}\in \mathcal{Q}$, environment $E$, and a set of robot-specific configuration parameters $C$. Trajectories can be obtained by a double integrator:
\begin{align}
    \begin{bmatrix}
           \dot{\bb{q}}_{t} \\
          \bb{q}_{t}
         \end{bmatrix} 
         = \int_{0}^{t}\begin{bmatrix} f(\bb{q}_{u},\dot{\bb{q}}_{u},|\bb{q}_{0},\dot{\bb{q}}_{0},\bb{q}_{g},E,C)\\
         \dot{\bb{q}}_{u}
         \end{bmatrix}\mathrm{d}u+
         \begin{bmatrix}
           \dot{\bb{q}}_{0} \\
          \bb{q}_{0}
         \end{bmatrix}. \label{eq:integrate}
\end{align}
We define $\mathtt{rollout_t}$ as an operation that finds the time steps used, $t^{*}\in \mathbb{R}^{+}$, to roll-out near the goal,
\begin{align}
    t^{*}=\mathtt{rollout_t}(\bb{q}_{0},\dot{\bb{q}}_{0},\bb{q}_{g},E,C):&=\min\{\min(t),t_{max}\}, && \text{ s.t. }\lvert\lvert \bb{q}_{t}-\bb{q}_{g}\lvert\lvert_{2}<\epsilon,
\end{align}
where $\epsilon$ denotes some tolerance. In practice $\bb{q}_{t}$ is obtained via numerical integration, such as Euler's method, of \cref{eq:integrate}, with a time step budget of $t_{max}\in \mathbb{R}^{+}$. Instead of defining a single goal, we inject global behaviour into the policy by stringing together a sequence of attraction states, or commands $\bb{q}_{g}^{i}$, while sharing local collision-avoidance and joint-limit handling fabric terms. We define the Geometric Fabrics Command Optimisation Problem (GFCOP), where we have a sequence of $n+1$ commands with different goal configurations, and optimise over the goals of the first $n$ commands. The final command is specified as the final global goal. Specifically, we define: 
\begin{align}
    [\text{GFCOP}]:&\min_{\bb{q}^{1}_{g}\ldots\bb{q}^{n}_{g}}\sum_{i=1}^{n+1} t^{*}_{i} +M\mathbbm{1} \label{Cost}\\
    \text{s.t. }t^{*}_{i}&=\mathtt{rollout_t}(\bb{q}^{i}_{0},\dot{\bb{q}}^{i}_{0},\bb{q}^{i}_{g},E,C),  \text{ for } i=1,\ldots,n+1 \label{Rollout}\\
    \bb{q}_{0}^{1}&=\bb{q}_{0},\;  \dot{\bb{q}}_{0}^{1}=\dot{\bb{q}}_{0}, \; \bb{q}^{n}_{g}=\bb{q}_{g}, && \text{(Boundary conditions)}\label{boundary}\\
    \bb{q}_{0}^{i}&=\bb{q}_{t^{*}_{i-1}}^{i-1}, \; \dot{\bb{q}}_{0}^{i} =\dot{\bb{q}}_{t^{*}_{i-1}}^{i-1}, \; \text{ for } i=2,\ldots,n+1, && \text{(Intermediate continuity)}\label{intermediate}\\
    \mathbbm{1}&=\{
               0, \text{ if } t^{*}_{n+1}<t_{max};
               1, \text{ otherwise.}\}&& \text{(Penalty Indicator)}\label{indicator}
\end{align}
where the cost (\cref{Cost}) is defined as the sum of the integration times exhausted until we reach our goal. Note that superscripts indicate the command index, while the subscript specifies the integration steps, i.e. $\bb{q}^{i}_{0}$ indicates the initial configuration at the $i^{th}$ command. If the roll-out towards the final command in the sequence cannot reach our goal, then a large penalty of $M \in \mathbb{R}$ is applied. This is controlled by the indicator $\mathbbm{1}\in\{0, 1\}$, as defined in \cref{indicator}. \Cref{boundary} designates that the start configuration and its velocity, match given initial conditions $\bb{q}_{0}$ and $\dot{\bb{q}}_{0}$, and the final command is also the specified global goal, $\bb{q}_{g}$. \Cref{intermediate} enforces continuity, specifying that the initial conditions when integrating towards a command, shall be the terminating configuration and velocity of that of the previous. Constraints \cref{Rollout,boundary,intermediate} are satisfied by running the roll-outs in a sequential manner, using the terminal conditions of the former integration as the initial conditions of the next. 

The underlying local Geometric Fabric, given by $f$ in \cref{eq:integrate}, performs local obstacle avoidance, self-collision avoidance, joint coordinate and velocity limit handling. This allows the GFCOP to be concisely defined. Additionally, as local avoidance fabric terms are repulsion-based, motion slows down as the robot approaches obstacles, resulting in larger $\mathtt{rollout_t}$ values. Therefore, solutions of GFCOP tend to smoothly move around obstacles, and not scrape past them.

The defined GFCOP is discontinuous and non-convex. We can apply the black-box optimiser \emph{Covariance Matrix Adaptation Evolution Strategy} (CMA-ES) \citep{CMA_tut}, which iteratively updates a multivariate Gaussian (MVG) sampling distribution over our command decision variables, $\bb{q}_{g}^{1}\ldots\bb{q}_{g}^{n}$, after ranking the computed cost of previously evaluated candidates. The probability of CMA-ES drawing samples over the entire domain is non-zero, and has been shown to efficiently find globally optimal solutions \citep{benchmark-cma,benchmark-es}. We denote each solution, $\bb{y}_{i} = [{\bb{q}_{g}^{1}}^{\top},\ldots, {\bb{q}_{g}^{n}}^{\top}]_{i}^{\top}$, as a sequentially concatenated vector of the command goals. In each iteration, we sample a batch of candidates from the MVG, compute the cost of the candidates in parallel, and then update the sampling distribution. Details of CMA-ES can be found in the tutorial \cite{CMA_tut}. Although CMA-ES finds globally optimal solutions, under a fixed time budget, the quality and speed of the optimisation is impacted by the initial batch of samples. Vanilla CMA-ES simply initialises a MVG based on random guesses of the mean and covariance, or simply sample uniformly. This motivates us to imbue prior knowledge into the optimisation and warm-start \citep{warmstart} the process by learning to sample.

\begin{algorithm}[H]
\caption{{$\mathtt{SolveGFCOP}$ with warm-started CMA-ES}}\label{alg:solve}

\SetKwInOut{Input}{input}
\SetKwInOut{Output}{output}
\SetKwInOut{Intialise}{intialise}
\newcommand{\lIfElse}[3]{\lIf{#1}{#2 \textbf{else}~#3}}
\SetKwFor{DoParallel}{Do in Parallel for each (}{) $\lbrace$}{$\rbrace$}
\Input{$\bb{q}_{0}$,$\dot{\bb{q}}_{0}$, $\bb{q}_{g}$, $E$, $f_{sampler}$, $C$, number of commands in sequence $n+1$, iterations $N_{iters}$, samples batch-size $N_{samp}$, $M$ large penalty for failure.}
$\bb{y}_{1},\ldots, \bb{y}_{N_{samp}}\sim f_{sampler}$ \tcp*{Draw candidate solutions from $f_{sampler}$ to warm-start}
$AllCandidatesWithCosts\leftarrow\{\}$ \tcp*{Empty set for all solutions with costs}
\For{$j=1$; $j<=N_{iters}$; $j=j+1$}{
$BatchCandidatesWithCosts \leftarrow \{\}$ \tcp*{Empty set for solutions with costs}
\textbf{$\rhd$ Compute costs for each candidate in parallel}\;
\DoParallel{$\bb{y}$ $\mathrm{in}$ $[\bb{y}_{1},\ldots, \bb{y}_{N_{samp}}]$}{
$\bb{q}_{g}^{1}, \ldots,\dot{\bb{q}}_{g}^{n}\leftarrow \bb{y}$ \tcp*{Unpack $\bb{y}$ to get candidate solution goals}
$\bb{q}_{g}^{n+1}\leftarrow \bb{q}_{g}$ \tcp*{Specify the command goal as the global goal}
$\bb{q}_{0}^{1}\leftarrow \bb{q}_{0}$, $\dot{\bb{q}}_{0}^{1}\leftarrow \dot{\bb{q}}_{0}$ \tcp*{Set Initial configuration and velocity}

$Cost \leftarrow 0$\tcp*{Initialise the cost of a candidate solution}
\For{$i=1$; $i<=n+1$; $i=i+1$}{
$t_{i}^{*}\leftarrow \mathtt{rollout_t}(\bb{q}^{i}_{0},\dot{\bb{q}}^{i}_{0},\bb{q}_{g}^{i}, E, C)$ \tcp*{Roll-out the $i^{th}$ command in sequence}
$Cost\leftarrow Cost+t_{i}^{*}$
}
\lIfElse {$t^{*}<t_{max}$}{$\mathbbm{1}\leftarrow 0$}{$\mathbbm{1} \leftarrow 1$} 

$Cost\leftarrow Cost+M\mathbbm{1}$ \tcp*{If we don't reach before time-out, apply penalty}
$BatchCandidatesWithCosts.\mathtt{insert(\{\bb{y}, Cost\})}$\;
$AllCandidatesWithCosts.\mathtt{insert(\{\bb{y}, Cost\})}$\;
}
\textbf{$\rhd$ Update CMA-ES and sample the next batch of candidates}\;
CMA-ES.$\mathtt{update}(BatchCandidatesWithCosts)$ \tcp*{Update CME-ES with new costs}
$\bb{y}_{1},\ldots, \bb{y}_{N_{samp}}\sim$ CMA-ES.sampler()\tcp*{Sample next batch with CMA-ES sampler}
}
$top$-$m\leftarrow\mathtt{GetTopMByCost}(AllCandidatesWithCosts)$ \tcp*{Sort by cost and get lowest $m$ candidates}
\Output{$top$-$m$}
\end{algorithm}

\subsection{Learning to Sample for Optimisation}\label{learn_section}
We postulate that the optimal motions across problem setups with similar start, $\bb{q}_{0}$ $\dot{\bb{q}}_{0}$, goal, $\bb{q}_{g}$, and environments, $E$, are similar, and that we can \emph{learn} to generate promising candidates from the optimisation solutions in alternative problem setups. We can use the generated candidates to warm-start CMA-ES, or directly use the candidate solution with the lowest cost as an approximate solution. Next, we define operations needed for the training, and outline the training, sampling and encoding used to construct the generative model.

\subsubsection{Solutions of GFCOP} 
Let us define the operation $\mathtt{SolveGFCOP}(\bb{q}_{0},\dot{\bb{q}}_{0},\bb{q}_{g},E \lvert f_{sampler})$ to solve a GFCOP problem for a specific initial configuration, velocity, goal, environment, and return the top-$m$ candidate solutions. A detailed algorithm is shown in algorithm \ref{alg:solve}. A sampler $f_{sampler}$ is passed on to provide the initial samples, and warm-start CMA-ES. We keep track of candidate solutions and their cost. For each call of $\mathtt{SolveGFCOP}$, we obtain a set of $\{\bb{y}_{i}\}_{i=1}^{m}$ of $m$ solutions, where $\bb{y}_{i} = [{\bb{q}_{g}^{1}}^{\top},\ldots, {\bb{q}_{g}^{n}}^{\top}]_{i}^{\top}$, is the sequentially concatenated vector of the command goals. Note that the GFCOP is also dependent on other inputs, such as the length of the sequence of commands, $n+1$, as well as the robot-specific parameters, $C$. However, we shall drop the explicit dependence on these, as they are held constant during learning, i.e. learning is conducted over command sequences of the same length and on the same robot. 

\subsubsection{Training the Generative Model}
Implicit energy-based models (EBMs) \citep{implicitBC} excel at learning complex \emph{unnormalised} distributions, where only sampling and not density estimation is performed using the model. Here, we take an EBM approach and learn an energy function $E_{\theta}(\bb{x}^{c},\bb{y})\rightarrow \mathbb{R}$, mapping from a (context, target) pair to an energy value, by contrasting positive and negative samples, where positive samples have high energy and negative samples have low energy values. In our setup, the context is an encoding of the problem, while the target is the optimised command sequence. Suppose we have a dataset of positive examples $D^{pos}=\{\bb{x}^{c}_{j},\bb{y}^{*}_{j}\}_{j=1}^{N_{pos}}$. Each $\bb{x}^{c}_{j}$ encodes the inputs to $\mathtt{SolveGFCOP}$, and each $\bb{y}^{*}_{j}$ is outputted from $\mathtt{SolveGFCOP}$. For each positive example $\bb{y}^{*}_{j}$, we actively generate $N_{neg}$ negative examples $\{\widehat{\bb{y}}^{i}_{j}\}_{i=1}^{N_{neg}}$. Then, we can construct a InfoNCE-like loss \citep{implicitBC,infoNCE}:
\begin{align}
    \mathcal{L}_{\mathrm{InfoNCE}}=\sum^{N_{pos}}_{j=1} -\log\Big\{\big(e^{-E_{\theta}(\bb{x}^{c}_{j},\bb{y}^{*}_{j})}\big)\big(e^{-E_{\theta}(\bb{x}^{c}_{j},\bb{y}^{*}_{j})}+\sum_{k=1}^{N_{neg}}e^{-E_{\theta}(\bb{x}^{c}_{j},\bb{\widehat{y}}_{j}^{k})}\big)^{-1}\Big\}. \label{eq:lossimpli}
\end{align}
The energy function $E_{\theta}$ is modelled by a neural network with parameters $\theta$, which can be trained via stochastic gradient descent and its variants. 

\subsubsection{Generating Samples with Stochastic Gradient Langevin Dynamics} 
Stochastic gradient Langevin dynamics (SGLD) \citep{SGLD} is a technique to obtain a full posterior distribution from some energy function $E_{\theta}$, rather than a single maximum {\em a posteriori} mode. This allows us to draw samples from the energy function to: (1) provide challenging negative examples to contrast against \citep{ImplicitLange}; (2) warm-start CMA-ES. Given a trained energy function $E_{\theta}$, and an context encoding $\bb{x}^{c}$, we randomly initialise $n_{p}$ particles $\bb{y}^{0}_{1},\ldots,\bb{y}^{0}_{n_{p}}\sim\mathcal{U}(\bb{q}^{upper},\bb{q}^{lower})$ from a uniform distribution, where $\bb{q}^{upper},\bb{q}^{lower}$ are the upper and lower joint limits. The particles are updated according to the rule:
\begin{align}
     \Delta \bb{y}^{k+1}_{i}=\bb{y}_{i}^{k}+\frac{\sigma^{2}_{k}}{2}(\nabla_{\bb{y}_{i}}E_{\theta}(\bb{x}^{c},\bb{y}^{k}_{i}))+\eta^{k}, && \text{where }\eta\sim \mathcal{N}(0,\sigma^{2}_{k}), && \text{for }i=0,\ldots,n_{p}\label{update_rule}.
\end{align}
Additionally, we decrease $\sigma^{2}$ according to the scheduler $\sigma^{2}_{k}=(1+k)^{-0.5}$. After a maximum number of gradient ascent iterations has been exhausted, we take the updated particles and either select them as negative examples during training, or use them as solutions during inference. 

\subsubsection{Encoding a GFCOP}

The EBM requires a fixed length context vector, $\bb{x}^{c}$, to condition on. The initial configuration, its velocity, and the goal are all fixed in size. We aim to find a fixed-sized encoding for the environment {\em E}. Here, we assume that each environment is represented by a point set $\mathcal{S}=\{\bb{s}_{i}\}^{N_{E}}_{i=1}$ of size $N_{E}$, where $\bb{s}_{i}\in\mathbb{R}^{3}$ are coordinate points in Euclidean space. This represents surfaces of objects, and can be obtained from an RGBD camera. The size of the point set $N_{E}$ varies with each environment. We take a reference point-set approach \citep{Yang_2019_ICCV} and compute the minimum Euclidean distance between the point set, and a set of fixed $N_{R}$ reference coordinates $\widehat{\mathcal{S}}=\{\widehat{\bb{s}}_{i}\}_{i=1}^{N_{R}}$

\begin{equation}
\bb{d}_{E}(\mathcal{S},\widehat{\mathcal{S}}):=\big[\min_{\bb{s}\in\mathcal{S}}\lvert\lvert\bb{s}-\widehat{\bb{s}}_{1}\lvert\lvert_{2},\;\min_{\bb{s}\in\mathcal{S}}\lvert\lvert\bb{s}-\widehat{\bb{s}}_{2}\lvert\lvert_{2},\;\ldots,\;\min_{\bb{s}\in\mathcal{S}}\lvert\lvert\bb{s}-\widehat{\bb{s}}_{N_{R}}\lvert\lvert_{2}\big].
\end{equation}

This provides us a fixed length (of length $N_{R}$) vector representation for environments with point sets of different sizes. The reference points are laid out in a lattice, with equal distance. If the size of the fixed length vector is excessively large, its dimensions are further reduced using an autoencoder, to obtain encoding $\widehat{\bb{d}}_{E}$. We concatenate $\bb{q}_{0}$, $\dot{\bb{q}}_{0}$, $\bb{q}_{g}$, and $\widehat{\bb{d}}_{E}$ to produce the input context $\bb{x}^{c}$.

\subsection{Self-supervision: Optimisation and Learning in the Loop}\label{ssl:subsection}

A key insight is that the optimisation outlined in \cref{opt_section} and the learning outlined in \cref{learn_section} are complimentary. Specifically, as more GFCOPs are solved, the more self-labelled training data is available for our implicit EBM. The EBM is trained by \cref{eq:lossimpli}, with negative samples generated from SGLD. On the other hand, as our EBM is progressively more well-trained, the quality of its generated candidate solutions also improve. This section brings both the optimisation and learning into a unified loop, and outlines a self-supervised learning framework. Following the definition in \cite{ssl_robotics}, in this context, self-supervision refers to the absence of external supervision signal (i.e. no human labelling) \citep{ssl_robotics}. Algorithm \ref{alg:self_super} outlines the self-supervised GFCS learning framework. We are assumed to have a set of $N_{prob}$ motion generation problems, $\{\bb{q}_{0}^{n},\dot{\bb{q}}_{0}^{n},\bb{q}_{g}^{n},E^{n}\}_{n=1}^{N_{prob}}$. To self-generate data, we iteratively add solutions of $\mathtt{SolveGFCOP}$ to a buffer $D^{buffer}$ of some maximum length; we train our EBM, $E_{\theta}$, on the data in the buffer, after every batch, of size $n_{train}$, of problems have been solved; $E_{\theta}$ is then used as the sampler $f_{sampler}$ for future calls of $\mathtt{SolveGFCOP}$.   

\begin{algorithm}[H]
\caption{{Self-supervised Geometric Fabric Command Sequence Learning Framework}}\label{alg:self_super}

\SetKwInOut{Input}{input}
\SetKwInOut{Output}{output}
\Input{A set of $N_{prob}$ motion generation problems $\{\bb{q}_{0}^{n},\dot{\bb{q}}_{0}^{n},\bb{q}_{g}^{n},E^{n}\}_{n=1}^{N_{prob}}$; an untrained implicit model $E_{\theta}(\bb{x},\bb{y})\rightarrow\mathbb{R}$, $N_{neg}$, $\bb{q}^{upper},\bb{q}^{lower}$, $\mathcal{D}^{buffer}$.}
$f_{sampler}\leftarrow \mathcal{U}(\bb{q}^{upper},\bb{q}^{lower})$ \tcp*{Initialise the sampler to be uniform}
\For{$n=1$; $n<=N_{prob}$; $n=n+1$}{
\textbf{$\rhd$ Self-generate positive examples via optimisation}

    $\mathcal{D}^{pos}\leftarrow\mathtt{SolveGFCOP}(\bb{q}_{0}^{n},\dot{\bb{q}}_{0}^{n},\bb{q}_{g}^{n},E^{n}\lvert f_{sampler})$ \tcp*{Solve for top-$m$ solutions}
    $\mathcal{D}^{buffer}\mathtt{.insert}(\mathcal{D}^{pos})$ \tcp*{Add positive data into buffer}
    $\mathcal{D}^{buffer}\leftarrow \mathtt{MaintainBuffer}(\mathcal{D}^{buffer})$ \tcp*{Maintain the length of buffer}
    $\mathcal{D}^{neg}\leftarrow\{\}$ \tcp*{Initialise empty set for negatives}
    \For{$j=1$; $j<=\mathtt{length}(\mathcal{D}^{buffer})$; $j=j+1$}{
    $\bb{y}^{0}_{1},\ldots,\bb{y}^{0}_{N_{neg}}\sim\mathcal{U}(\bb{q}^{upper},\bb{q}^{lower})$\;
    $\{\widehat{\bb{y}}^{i}_{j}\}_{i=1}^{N_{neg}}\leftarrow \mathtt{SGLD}(\bb{y}^{0}_{1},\ldots,\bb{y}^{0}_{N_{neg}}\lvert E_{\theta})$ \tcp*{Run SGLD using \cref{update_rule} on $E_{\theta}$}
    $\mathcal{D}^{neg}.\mathtt{insert}(\{\widehat{\bb{y}}^{i}_{j}\}_{i=1}^{N_{neg}})$ \tcp*{Insert into set of negatives}
    }
    $\bb{x}\leftarrow \mathtt{encode}(\bb{q}_{0}^{n},\dot{\bb{q}}_{0}^{n},\bb{q}_{g}^{n},E^{n})$ \tcp*{Encode problem.}
\textbf{$\rhd$ Train on self-generated data, and update the optimisation sampler}

\If {$n\% n_{train}==0$}
    {$E_{\theta}.\mathtt{train}(\mathcal{D}^{buffer},\mathcal{D}^{neg})$ \tcp*{train $E_{\theta}$ with \cref{eq:lossimpli}, after $n_{train} solves$}}
    
    $f_{sampler}\leftarrow \mathtt{SGLD}(\bb{y}^{0}_{1},\bb{y}^{0}_{2}\ldots \lvert E_{\theta})$, \text{ where } $\bb{y}^{0}_{1},\bb{y}^{0}_{2}\ldots \sim \mathcal{U}(\bb{q}^{upper},\bb{q}^{lower})$ \tcp*{Update the sampler as SGLD to the trained $E_{\theta}$}
}
\Output{Trained $E_{\theta}$}
\end{algorithm}

\section{Experimental Evaluation}
\begin{figure}[h]
\centering
\begin{subfigure}{0.32\textwidth}
    \frame{\includegraphics[width=\textwidth]{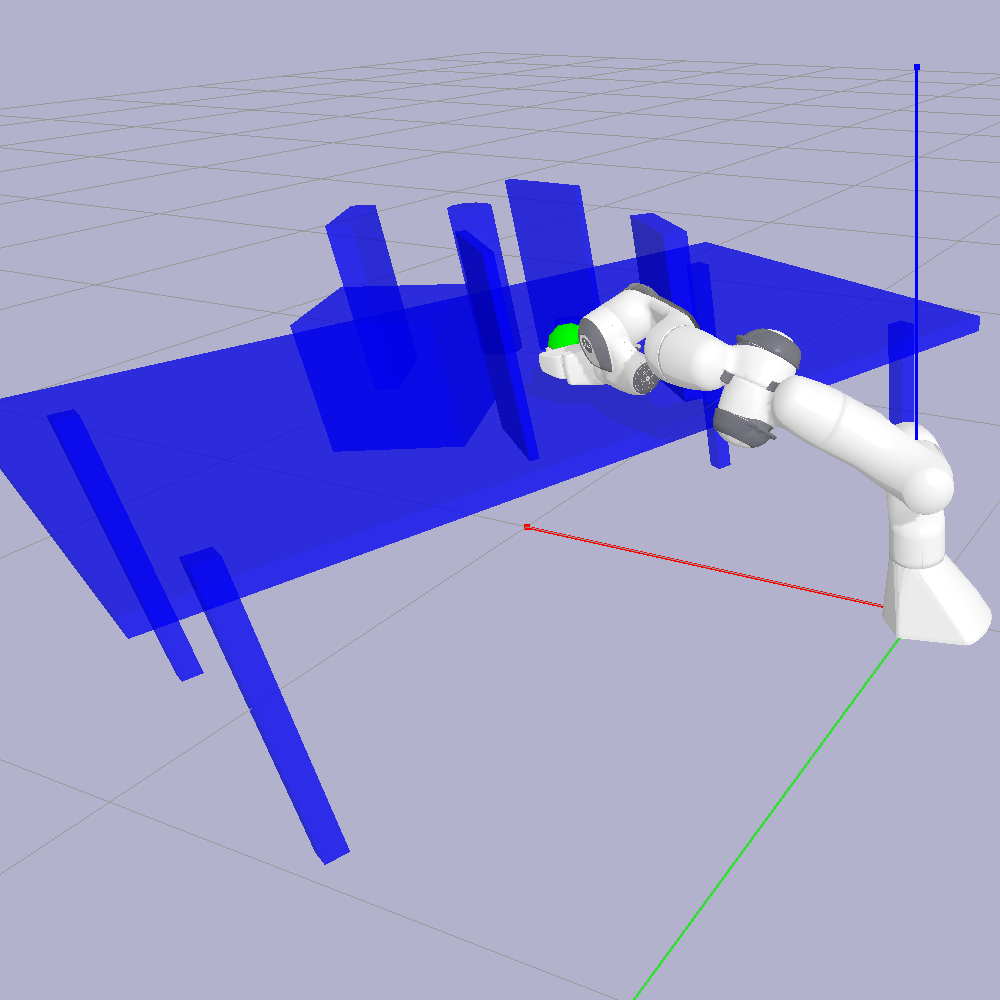}}
\end{subfigure}%
\begin{subfigure}{0.32\textwidth}
    \frame{\includegraphics[width=\textwidth]{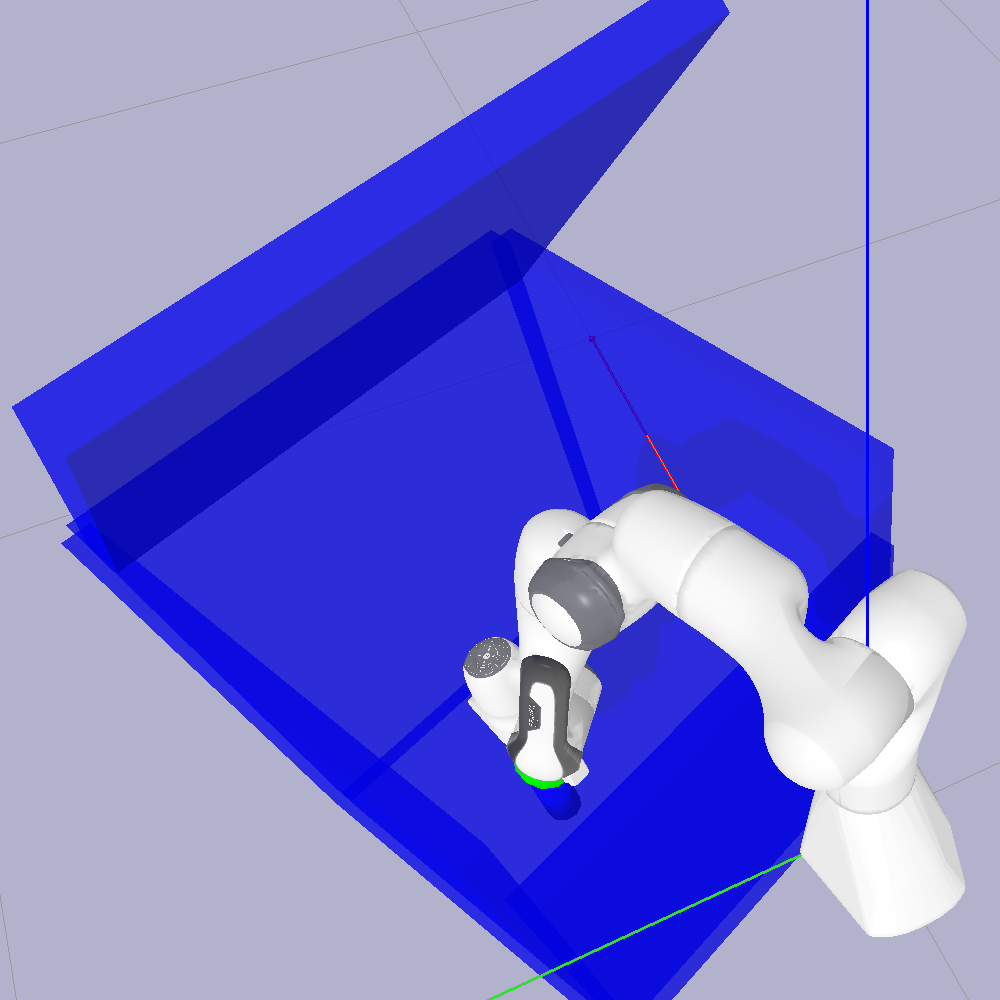}}
\end{subfigure}%
\begin{subfigure}{0.32\textwidth}
    \frame{\includegraphics[width=\textwidth]{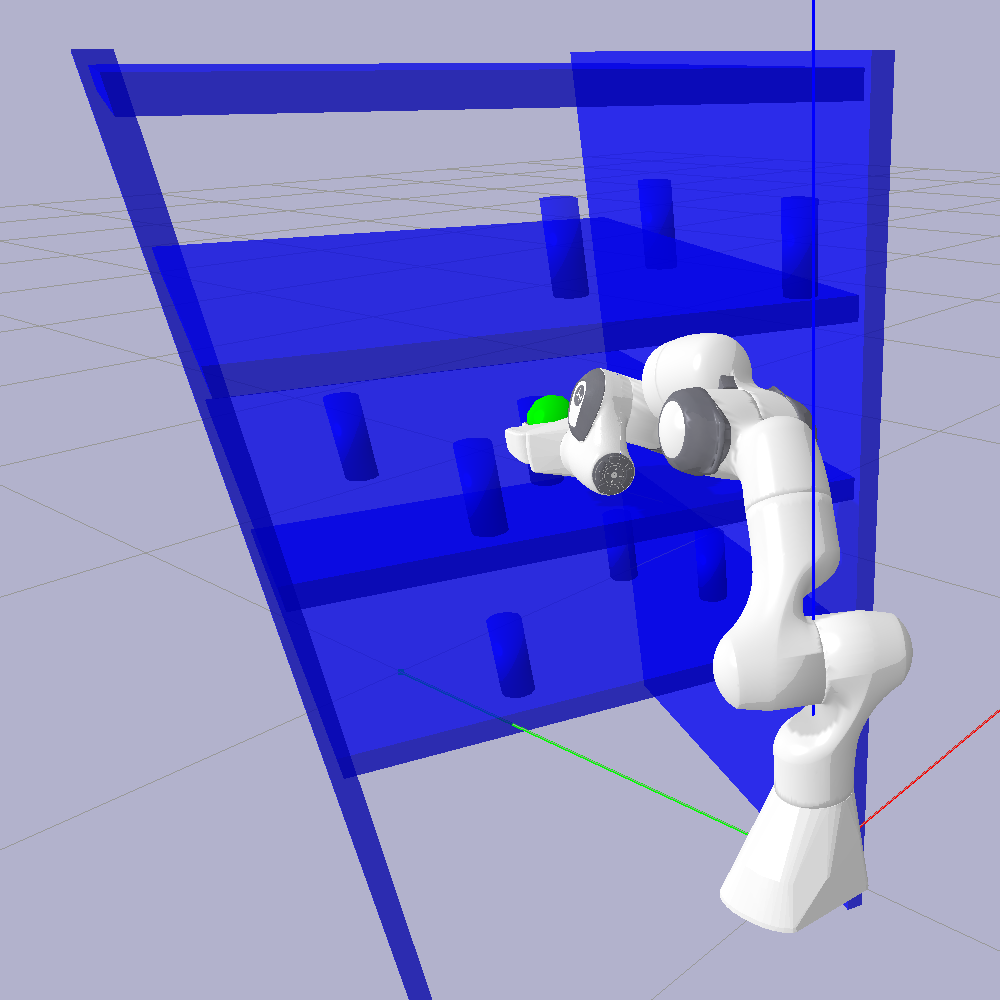}}
\end{subfigure}%

\begin{subfigure}{0.32\textwidth}
    \frame{\includegraphics[width=\textwidth]{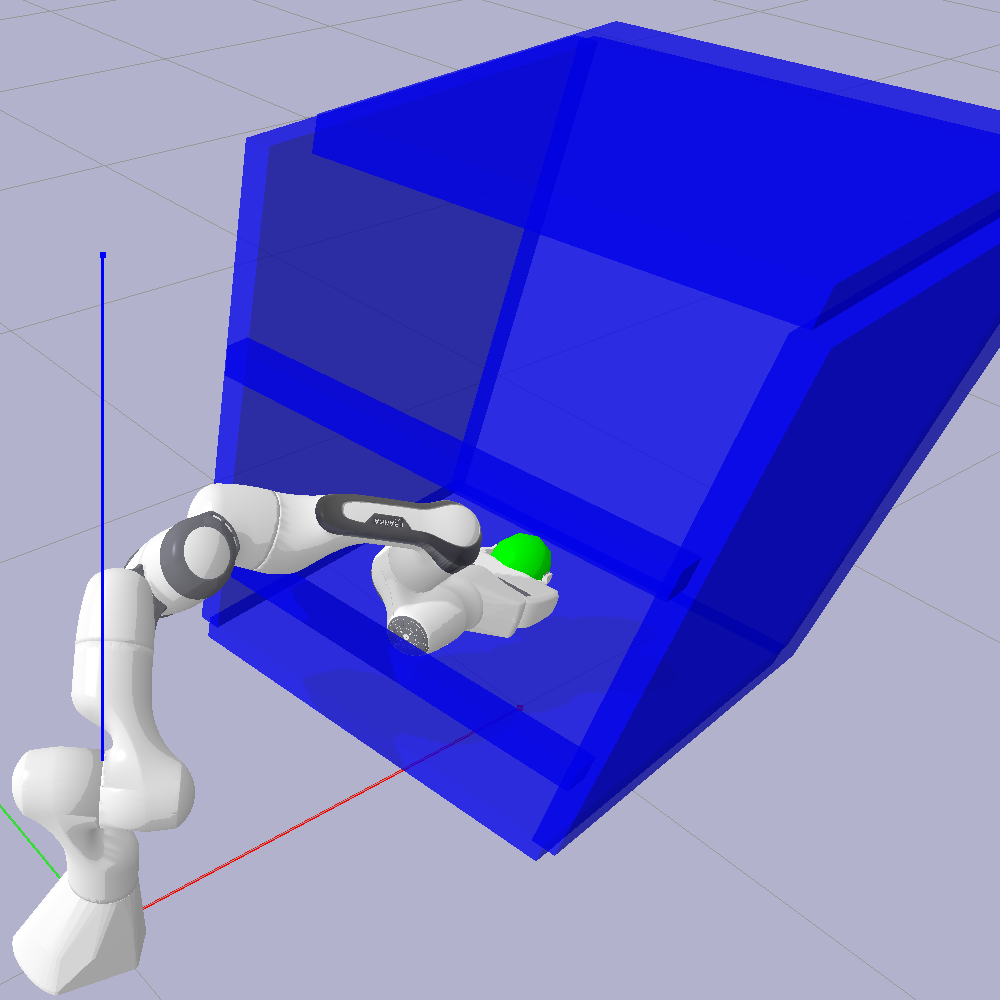}}
\end{subfigure}%
\begin{subfigure}{0.32\textwidth}
    \frame{\includegraphics[width=\textwidth]{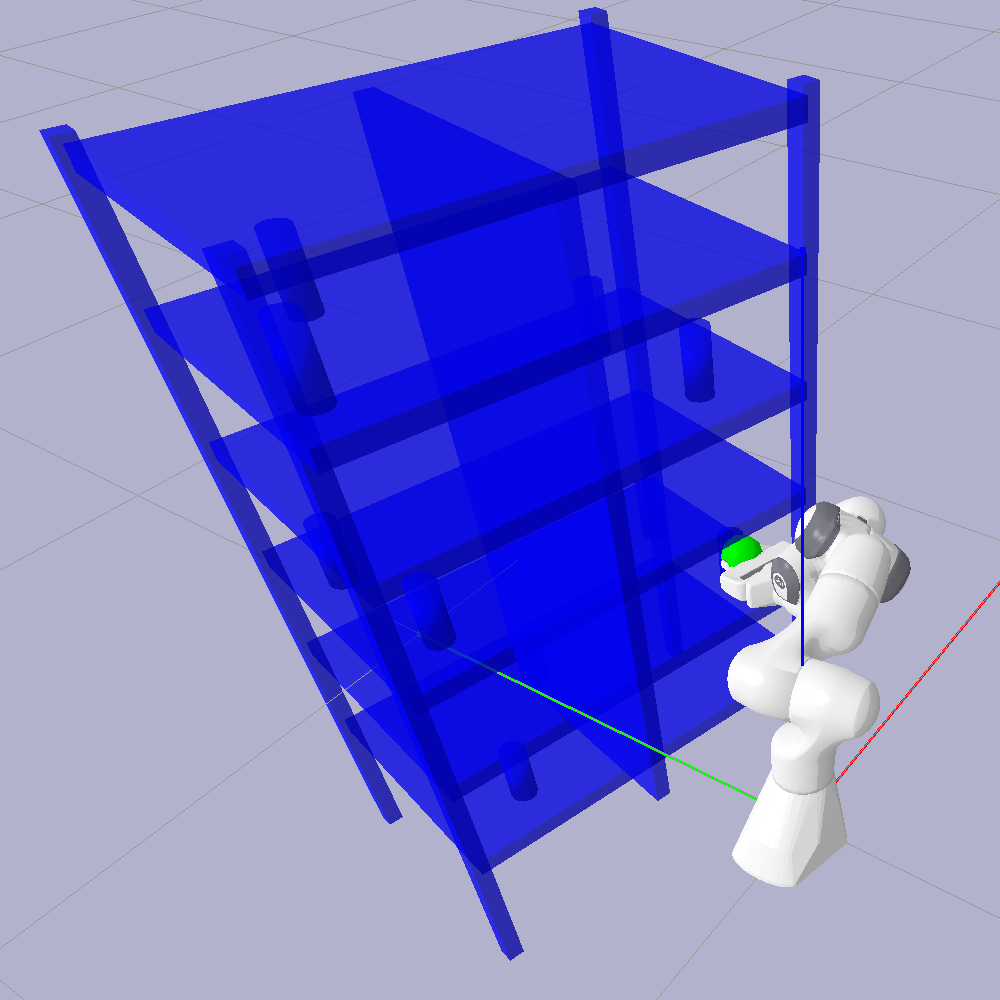}}
\end{subfigure}%
\begin{subfigure}{0.32\textwidth}
    \frame{\includegraphics[width=\textwidth]{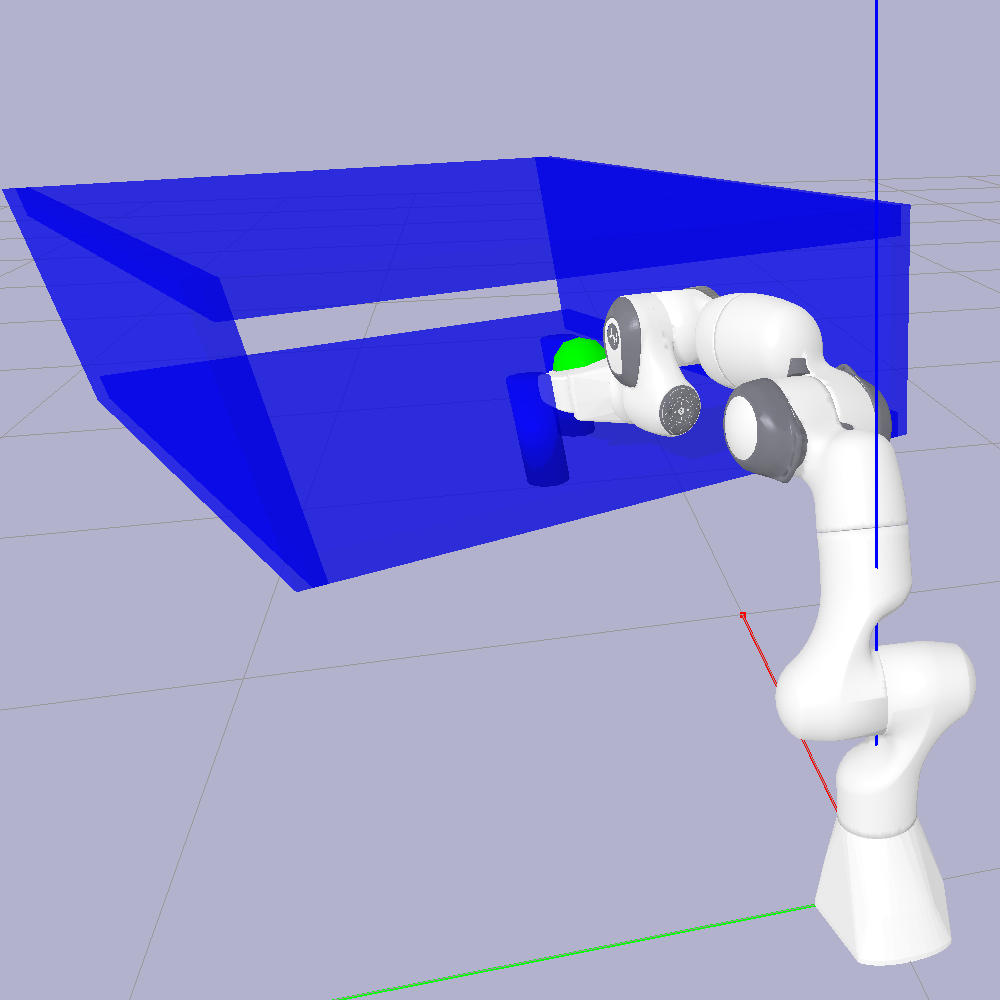}}
\end{subfigure}
\caption{{Examples from the different problem classes (Table under, Box, Tall shelf, Cage, Thin shelf, Single shelf) from \cite{Chamzas2022MotionBenchMakerAT}, goals shown in green.}}\label{fig:problems}
\end{figure}

We empirically investigate the performance of self-supervised GFCS to reliably generate global and reactive motion in various complex environments, both on a simulation FRANKA EMIKA Panda and on a real-world JACO manipulator. We evaluate 6 different classes of benchmark problems from \cite{Chamzas2022MotionBenchMakerAT}, each class of problems contains 100 environments with different start/goals. Example problems are illustrated in \cref{fig:problems}. For each problem class, we use 50 problems for self-supervised training, and report our evaluation on the remaining 50 test problems. We report on the following metrics: 
\begin{itemize}
\item \textit{\% Success}: the percentage of problems a solution is found; \textit{Times}: the time taken to generate the trajectories; 
\item \textit{EE length}: the length of the end-effector trajectory; \textit{C-space length}: the length of the C-space trajectory; 
\item \textit{EE-LDJ}: The log-dimensionless jerk (LDJ) \citep{LogdimJerk} of the end-effector trajectory; \item \textit{C-space LDJ}: The LDJ \citep{LogdimJerk} of the C-space trajectory.
\end{itemize}
EE length reflects trajectory legibility, with a low length indicating that little superfluous motion is present. LDJ reflects the trajectory smoothness, where closer to zero indicates a smoother trajectory.

\subsection{Experimental Setup Details}

In our experimental evaluations of trajectory quality, we use we evaluate 6 problem classes of benchmark problems. For each class, 50 problems are used within the self-supervised training loop, and we evaluate and report our results on the remaining 50. In total, 300 problems are used for self-supervised learning, and 300 used in evaluated results. 

When rolling out trajectories for Geometric Fabric Command Sequences, we set the time budget per command, $t_{max}$, to 7s, and integrate with step-size $0.01$. The Geometric Fabric terms for attractors, joint limit avoidance, and collision-avoidance follow the original implementation from the authors of Geometric Fabrics \citep{geoFabs}. Over the 6 benchmark problems, we use two commands for every problem. During self supervised training, for each problem, we sample and optimise for 30 batches with 30 candidate samples per batch, while during test time, we draw 20 candidate samples per batch. We iterate through all the training problems twice, and add the top-4 solutions into the buffer for training, and the buffer is maintained to have 200 solutions. Training on solutions occurs after every 25 problems have been solved, that is $n_{train} = 25$. 

During training, to encode the environment, we lay out fixed reference point in intervals of 0.15. After obtaining an encoding by projecting the obstacle points to the reference points, we further lower the dimensions of the encoding to size 300, using an auto-encoder. We then concatenate the environment encoding with the start and goal configurations along with the input commands to produce an input vector $\bb{x}^{c}$. This is passed into a neural network of with layers:

\noindent $Dense(input\_size, 800) \rightarrow ReLU() \rightarrow Dense(800,800) \rightarrow ReLU() \rightarrow Dense(800,800) \rightarrow ReLU() \rightarrow Dense(800,1)$. 

We generate negative samples by running SLGD on 200 particles, for 50 iterations with $\epsilon^{2}=0.2$. For each time we train the generative model, training is run for 800 iterations with an ADAM optimiser with a step-size of $0.0001$. When we generate new obstacles to evaluate the reactive-ness of our method, obstacles positions are sampled randomly from the solved trajectory coordinates which we solve without the additional obstacles. Additionally, we require that the addition of the obstacles do not cause the initial robot configuration to be in collision, and is not in the trajectory portion corresponding to the last second of the original solution.

\subsection{Global Solutions and Motion Trajectory Quality}
We hypothesis that our method, like local motion generation approaches, is able to generate legible motion trajectories, without superfluous end-effector motion, while capable of handling non-convexities in the task space to generate globally feasible motion. We observe that a GFCS that is very short is sufficient to generate complex non-local behaviour. We compare GFCS, of length two, against: 
\begin{enumerate}
\item \textit{Geometric Fabrics (GFs)}, a reactive method capable of producing high quality motion trajectories; 
\item Bi-directional Expansive Space Trees (Bi-EST) \citep{biEST}, a bi-direction sampling-based motion planner; 
\item Batch Informed Trees (BBT$^{*}$) \citep{BITstar}, an asymptotically optimal sampling-based planner. 
\end{enumerate}
We use the Bi-EST and BIT$^{*}$ implementations in the Open Motion Planning Library (OMPL) \citep{OMPL}. As our method can continuously optimise and refine the solution, we evaluate our method after the first and fifth batches of samples have been evaluated. Likewise, BIT$^{*}$ is asymptotically optimal and will continuously refine the solution found. We evaluate BIT$^{*}$ with a budget of 1.2 seconds and 6 seconds, which is slightly longer than the times our method takes to evaluate the first and fifth batch of samples.

\begin{figure}[t]
\centering
\includegraphics[width=0.45\textwidth]{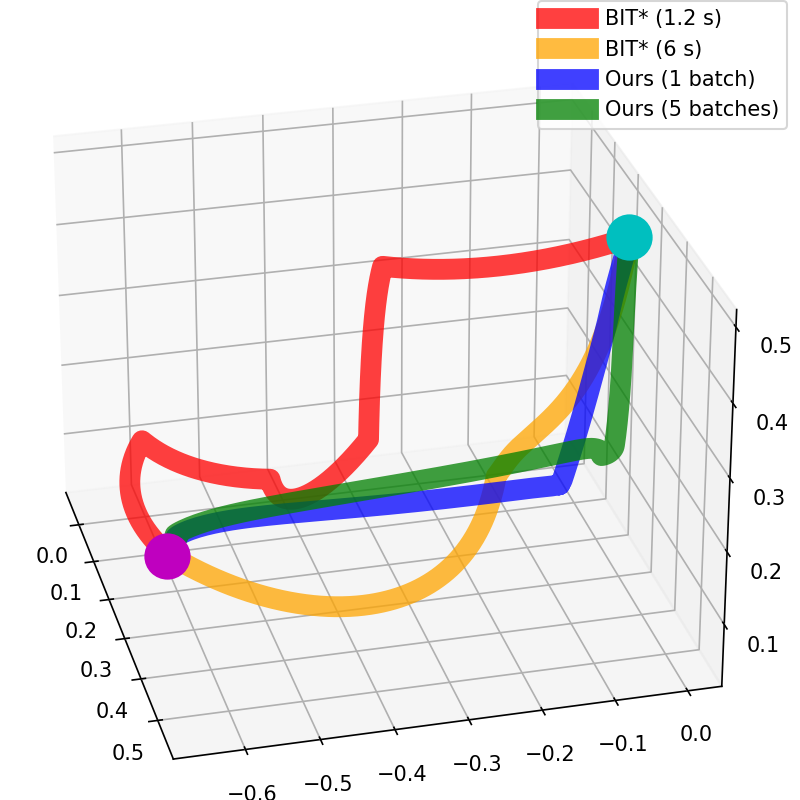}
\hspace{1em}
\includegraphics[width=0.45\textwidth]{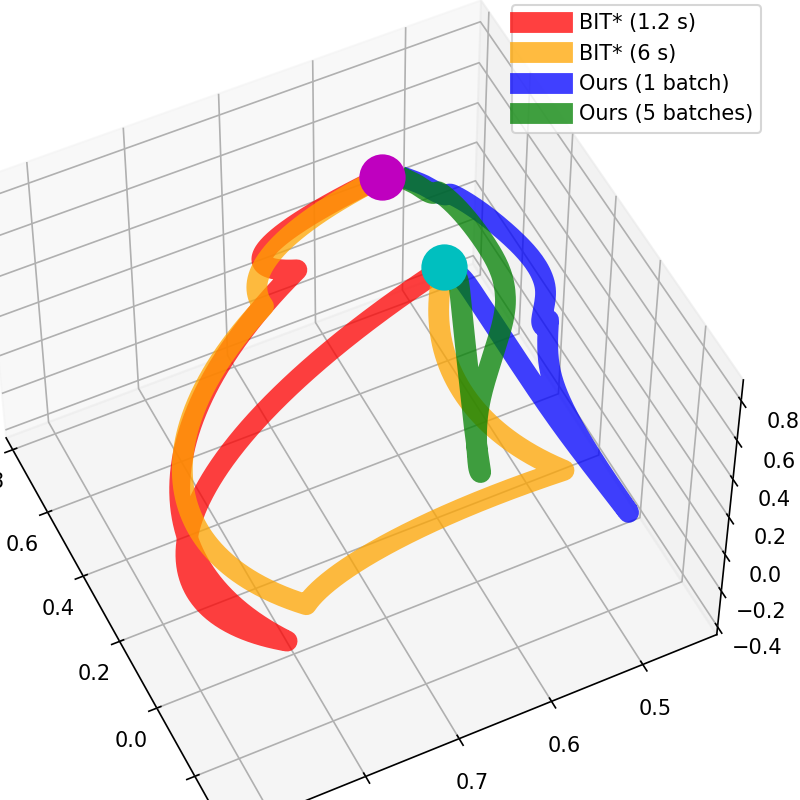}
\caption{{Examples of end-effector trajectories, from start (cyan) to goal (magenta), in ``Tall shelf'' (Left) and ``Table under'' (Right). GFCS consistently produces shorter and smoother motions. BIT$^{*}$(1.2s) given in red, BIT$^{*}$(6s) in yellow, ours (1 batch) in blue, and ours (5 batch) in green.}}\label{eef_traj}
\end{figure}

We tabulate the results over the six classes of benchmark environments in \cref{results_table}. We observe that the local geometric fabrics approach fails to solve many of the problems, particularly in the ``table under'' class of problems, where the manipulator is tasked with starting from under a table and reaching a pose on a table. Completing such a task requires the manipulator to exhibit global behaviour and escape from a local minimum. On the other hand, GFCS is able to generate motions which escape from local minima. Bi-EST can typically find feasible solutions quickly, but produce undesirable trajectories, as measured by our metrics of trajectory quality. We also observe that the ``cage'' problems require the manipulator to traverse a narrow passage, through the bars of a cage, to reach inside. Sampling-based motion planning methods struggle at this: Bi-ESTs become inefficient and are slower than local approaches to find a feasible trajectory, while BIT$^{*}$ simply cannot find a feasible solution for most of the ``cage'' problems. Our method is able to generate high-quality trajectories. GFs and GFCS consistently produce trajectories with shorter and smoother end-effector motions than sampling-based planners. End-effector trajectories in two example problems are shown in \cref{eef_traj}.

\begin{table}[h]
    \caption{Evaluated metrics (mean $\pm$ std) for motion generated in multiple benchmark environments} \label{results_table}
    \centering
    \begin{subtable}{.78\linewidth}
      \centering
        \caption{ Evaluation in the ``Table Under'' problems}
        \begin{adjustbox}{width=0.98\textwidth,center} 
\begin{tabular}{lllllll}
\toprule

                 & \% Success & EE length                   & C-space length               & Times (s)                     & EE-LDJ                        & C-space-LDJ                   \\
                 \midrule
GF               & 0          & NA                          & NA                           & NA                            & NA                            & NA                            \\
Bi-EST           & 100        & $8.24 \pm 3.2$ & $23.69 \pm 9.2$ & $0.028 \pm 0.04$ & $-11.09 \pm 1.7$ & $-10.58 \pm 1.3$ \\
BIT* (1.2s)      & 100        & $3.93 \pm 0.9$ & $7.71  \pm 2.8$  & NA                            & $-7.86 \pm 1.3$  & $-7.34 \pm 0.9$  \\
BIT* (6s)        & 100        & $3.71 \pm 1.0$ & $7.12  \pm 2.8$  & NA                            & $-7.95 \pm 1.0$  & $-7.13 \pm 1.1$  \\
Ours (1 batch)   & 96         & $1.43 \pm 0.5$ & $9.35  \pm 2.4$  & $0.96 \pm 0.05$  & $-7.45 \pm 1.8$  & $-7.51 \pm 2.0$  \\
Ours (5 batches) & 98         & $1.32 \pm 0.3$ & $8.65  \pm 2.0$  & $5.29 \pm 0.4$  & $-7.05 \pm 1.6$  & $-7.09 \pm 1.8$ \\
\bottomrule
\end{tabular}
\end{adjustbox}
\end{subtable}%

\centering
\begin{subtable}{.78\linewidth}
\centering
\caption{ Evaluation in the ``Box'' problems}
\begin{adjustbox}{width=0.98\textwidth,center} 
\begin{tabular}{lllllll}
\toprule

                 & \% Success & EE length     & C-space length & Times (s)      & EE-LDJ         & C-space-LDJ    \\
                 \midrule
GF               & 100       & $0.92 \pm 0.07$ & $5.25 \pm 0.6$   & $0.066 \pm 0.01$ & $-5.32 \pm 0.9$  & $-5.57 \pm 1.5$  \\
Bi-EST           & 100       & $5.52 \pm 3.3$  & $19.68 \pm 9.3$  & $0.030 \pm 0.03$ &$-10.55 \pm 1.8$ & $-10.36 \pm 1.3$ \\
BIT* (1.2s)      & 92        & $1.97 \pm 0.7$  & $8.21 \pm 2.2$   & NA             & $-8.99 \pm 1.1$  & $-8.30 \pm 0.8$  \\
BIT* (6s)        & 100       & $1.84 \pm 0.6$  & $7.39 \pm 2.3$   & NA             & $-8.69 \pm 1.0$  & $-7.99 \pm 1.0$  \\
Ours (1 batch)   & 100       & $1.00 \pm 0.1$  & $5.43 \pm 0.8$   & $0.74 \pm 0.03$  & $-5.84 \pm 1.1$  & $-5.55 \pm 1.5$  \\
Ours (5 batches) & 100       & $0.97 \pm 0.1$  & $5.23 \pm 0.6$   & $4.01 \pm 0.2$   & $-5.89 \pm 1.6$ & $-5.40 \pm 1.6$ \\
\bottomrule
\end{tabular}
\end{adjustbox}
\end{subtable}%

\centering
\begin{subtable}{.78\linewidth}
\centering
\caption{ Evaluation in the ``Tall shelf'' problems}
\begin{adjustbox}{width=0.98\textwidth,center} 
\begin{tabular}{lllllll}
\toprule

                 & \% Success & EE length     & C-space length & Times (s)      & EE-LDJ        & C-space-LDJ   \\
                 \midrule
GF               & 88        & $0.700 \pm 0.2$ & $6.51 \pm 1.1$   & $0.1 \pm 0.03$   & $-4.97 \pm 1.3$ & $-6.03 \pm 1.9$ \\
Bi-EST           & 100       & $4.57 \pm 2.6$  & $14.01 \pm 5.8$  & $0.036 \pm 0.08$ & $-9.42 \pm 2.2$ & $-9.37 \pm 1.5$ \\
BIT* (1.2s)      & 90        & $1.73 \pm 0.7$  & $5.78 \pm 1.2$   & NA             & $-6.80 \pm 2.4$ & $-5.63 \pm 2.4$ \\
BIT* (6s)        & 94        & $1.60 \pm 0.6$  & $5.72 \pm 1.3$   & NA             & $-6.94 \pm 2.4$ & $-5.75 \pm 2.5$ \\
Ours (1 batch)   & 100       & $1.02 \pm 0.3$  & $5.89 \pm 1.0$   & $0.75 \pm 0.04$  & $-6.39 \pm 1.3$ & $-6.02 \pm 1.2$ \\
Ours (5 batches) & 100       & $1.00 \pm 0.2$  & $5.74 \pm 0.9$   & $3.69 \pm 0.25$  & $-6.19 \pm 1.4$ & $-5.76 \pm 1.3$\\
\bottomrule
\end{tabular}
\end{adjustbox}
\end{subtable}%

\centering
\begin{subtable}{.78\linewidth}
\centering
\caption{ Evaluation in the ``Cage'' problems}
\begin{adjustbox}{width=0.98\textwidth,center} 
\begin{tabular}{lllllll}
\toprule

                 & \% Success & EE length     & C-space length & Times (s)     & EE-LDJ         & C-space-LDJ    \\
                 \midrule
GF               & 88        & $0.66 \pm 0.1$  & $7.87 \pm 0.8$  & $0.16 \pm 0.02$ & $-5.3 \pm 1.4$   & $-7.02 \pm 0.8$  \\
Bi-EST           & 100       & $10.67 \pm 5.2$ & $37.3 \pm 15.4$  & $1.99 \pm 2.1$  & $-12.57 \pm 1.4$ & $-12.02 \pm 1.4$ \\
BIT* (1.2s)      & 2         & NA            & NA             & NA            & NA             & NA             \\
BIT* (6s)        & 2         & NA            & NA             & NA            & NA             & NA             \\
Ours (1 batch)   & 98        & $0.75 \pm 0.1$  & $6.91 \pm 0.9$   & $0.82 \pm 0.04$ & $-7.27 \pm 2.2$  & $-7.97 \pm 1.7$  \\
Ours (5 batches) & 100       & $0.73 \pm 0.1$  & $6.31 \pm 0.9$   & $4.06 \pm 0.2$  & $-7.46 \pm 1.7$  & $-7.94 \pm 1.4$  \\
\bottomrule
\end{tabular}
\end{adjustbox}
\end{subtable}%

\centering    
\begin{subtable}{.78\linewidth}
\centering
\caption{ Evaluation in the ``Thin shelf'' problems}
\begin{adjustbox}{width=0.98\textwidth,center} 
\begin{tabular}{lllllll}
\toprule

                 & \% Success & EE length    & C-space length & Times (s)      & EE-LDJ        & C-space-LDJ   \\
                 \midrule
GF               & 80        & $0.68 \pm 0.2$ & $6.49 \pm 1.4$   & $0.11 \pm 0.02$  & $-6.06 \pm 2.3$ & $-6.56\pm 1.2$  \\
Bi-EST           & 100       & $5.11 \pm 2.4$ & $16.50 \pm 7.1$  & $0.033 \pm 0.04$ & $-9.89 \pm 1.7$ & $-9.86 \pm 1.5$ \\
BIT* (1.2s)      & 86        & $1.89 \pm 0.7$ & $6.16 \pm 1.2$   & NA             & $-8.04 \pm 1.4$ & $-6.79 \pm 1.3$ \\
BIT* (6s)        & 92        & $1.66 \pm 0.6$ & $6.01 \pm 1.1$   & NA             & $-8.51 \pm 1.2$ & $-6.70 \pm 1.1$ \\
Ours (1 batch)   & 100       & $0.93 \pm 0.2$ & $5.64 \pm 0.8$   & $0.82 \pm 0.04$  & $-7.32 \pm 1.7$ & $-7.05 \pm 1.4$ \\
Ours (5 batches) & 100       & $0.92 \pm 0.2$ & $5.38 \pm 0.8$   & $4.33 \pm 0.4$   & $-7.38 \pm 1.4$ & $-6.68 \pm 1.4$ \\
\bottomrule
\end{tabular}
\end{adjustbox}
\end{subtable}%

\centering
\begin{subtable}{.78\linewidth}
\centering
\caption{ Evaluation in the ``Single shelf'' problems}
\begin{adjustbox}{width=0.98\textwidth,center} 
\begin{tabular}{lllllll}
\toprule

                 & \% Success & EE length    & C-space length & Times (s)      & EE-LDJ         & C-space-LDJ   \\
                 \midrule
GF               & 94        & $0.69 \pm 0.2$ & $6.51 \pm 1.46$  & $0.084 \pm 0.02$ & $-5.11 \pm 1.8$  & $-5.78 \pm 1.4$ \\
Bi-EST           & 100       & $5.64 \pm 2.8$ & $16.26 \pm 7.2$  & $0.066 \pm 0.13$ & $-10.04 \pm 1.8$ & $-9.89 \pm 1.6$ \\
BIT* (1.2s)      & 86        & $1.51 \pm 0.6$ & $5.79 \pm 1.7$   & NA             & $-6.75 \pm 2.8$  & $-5.38 \pm 2.7$ \\
BIT* (6s)        & 94        & $1.61 \pm 0.8$ & $5.48 \pm 1.5$   & NA             & $-6.88 \pm 2.9$  & $-5.26 \pm 2.7$ \\
Ours (1 batch)   & 100       & $0.90 \pm 0.2$ & $5.79 \pm 1.2$   & $0.73 \pm 0.03$  & $-6.97 \pm 1.4$  & $-6.65 \pm 1.3$ \\
Ours (5 batches) & 100       & $0.91 \pm 0.2$ & $5.68 \pm 1.1$   & $3.62 \pm 0.3$   & $-7.03 \pm 1.5$  & $-6.64 \pm 1.2$ \\
\bottomrule
\end{tabular}
\end{adjustbox}
\end{subtable} 
\end{table}

\subsection{The Effect of Learning}

\begin{table}
\caption{First batch motion by uniformly sampling, and by a learned model for ``Table under''.} \label{after_training}
\begin{adjustbox}{width=0.8\textwidth,center} 
\begin{tabular}{llllll}
\toprule
                                & \% Success & EE length    & C-space length & EE-LDJ        & C-space-LDJ   \\
                                \midrule
No learning & 90        & $1.85 \pm 0.5$ & $13.53 \pm 2.9$  & $-8.61 \pm 1.6$ & $-8.76 \pm 1.4$ \\
Learning                  & \textbf{96}        & $\bb{1.43 \pm 0.5}$ & $\bb{9.35 \pm 2.4}$   & $\bb{-7.45 \pm 1.8}$ & $\bb{-7.51 \pm 2.0}$\\
\bottomrule
\end{tabular}
\end{adjustbox}
\end{table}

Our implicit generative model learns to produce good samples for GFCS to optimise over. Here, we investigate the performance improvements provided by our learnt model. We observe that the learned generative model is able to suggest high quality candidate to the global optimiser. We investigate the performance of our learnt generative model compared against drawing from a uniform distribution, on ``table under'' problems, which particularly requires global solutions. The results in \Cref{after_training} show that using the learned sampler gives rises to more successful solutions, and higher quality motion trajectories. 

\subsection{Global and Reactive Motion Generation}
\begin{figure}[t]
\centering
\frame{\includegraphics[width=0.32\textwidth]{figures/chap8/blocking.png}}
\frame{\includegraphics[width=0.32\textwidth]{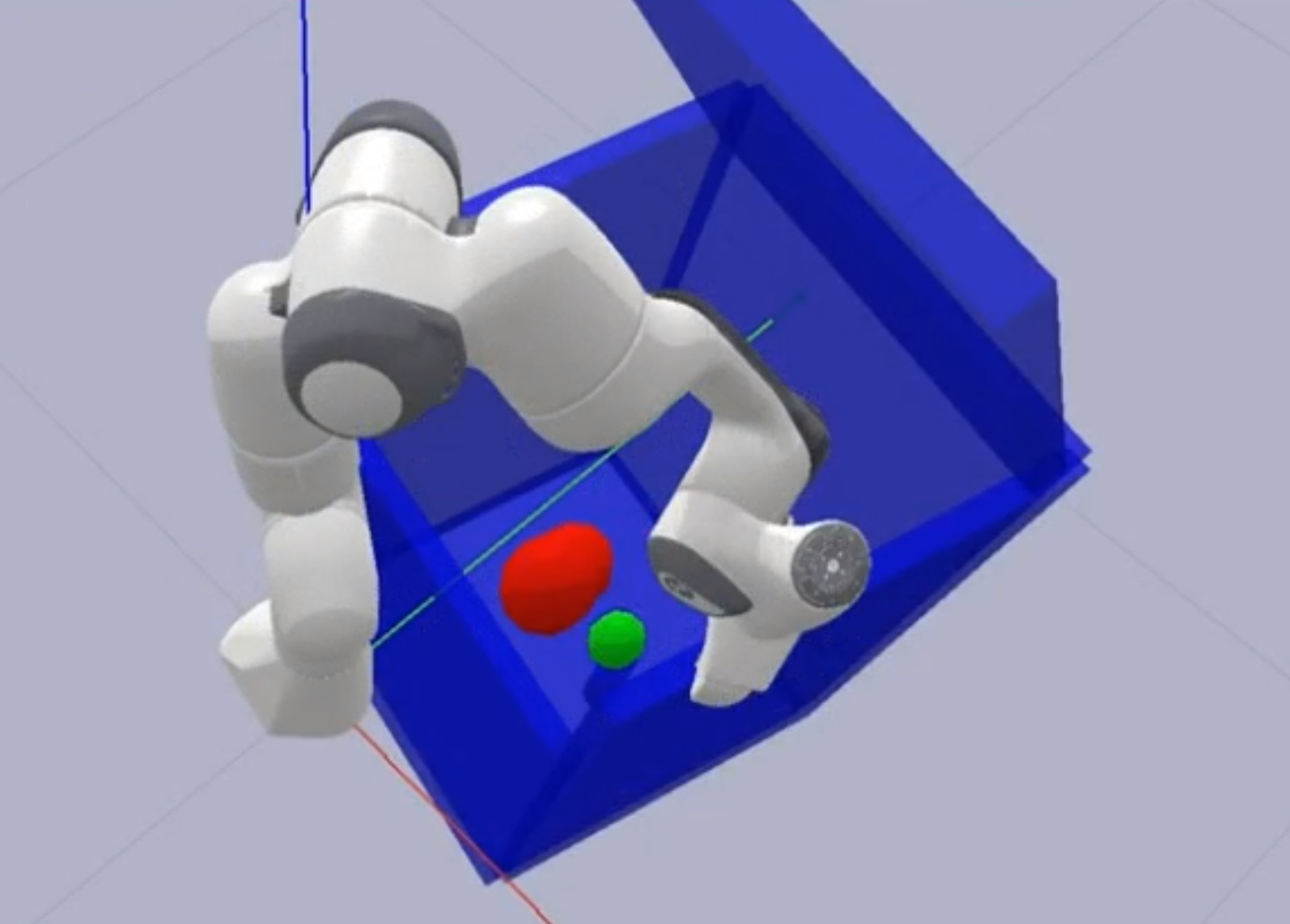}}
\frame{\includegraphics[width=0.32\textwidth]{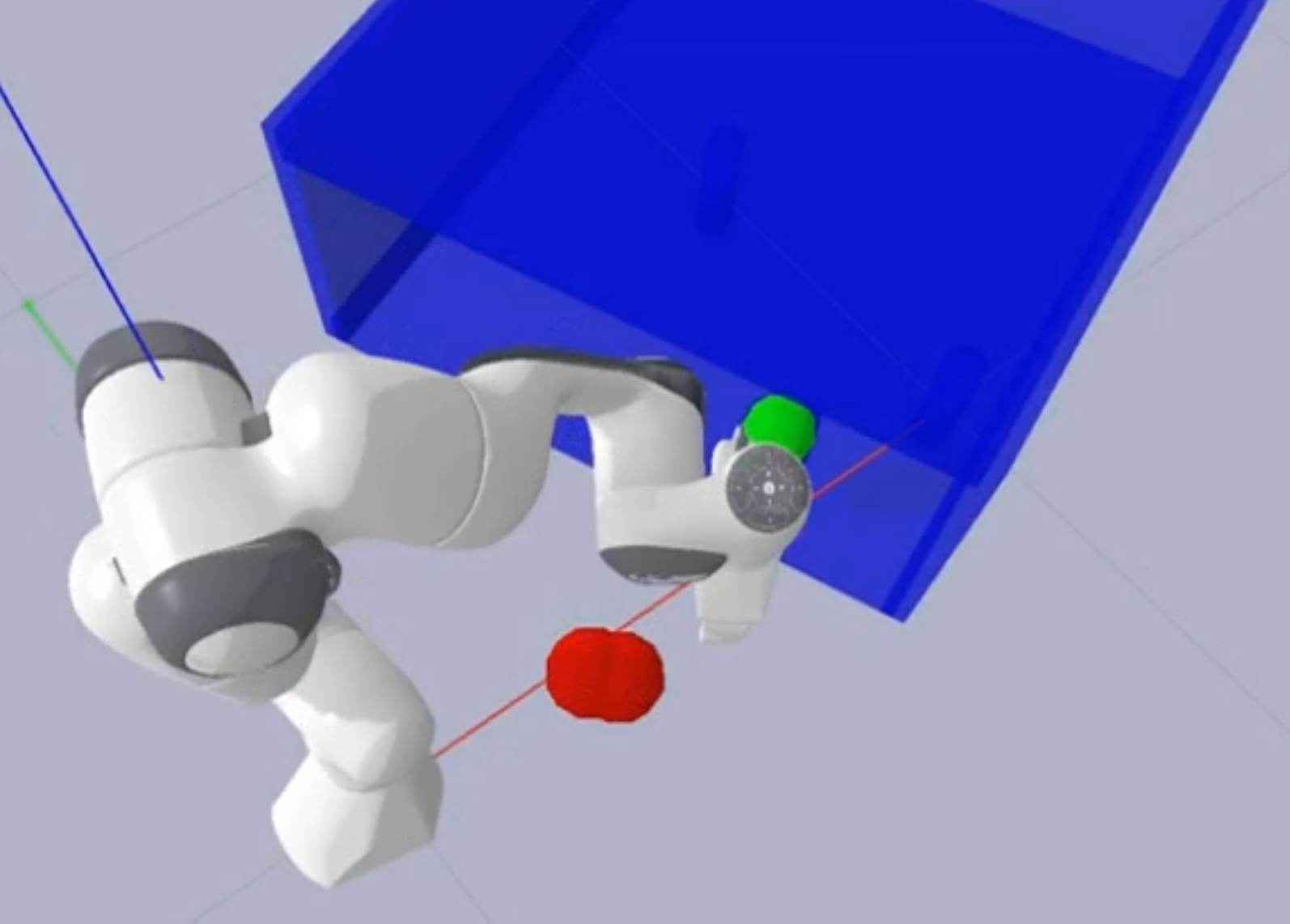}}
\caption{ After obtaining a solution to the goal (in green), we add new obstacles (in red), GFCS reactively avoid the new obstacles without re-optimising.}\label{fig:block}
\end{figure}

Geometric Fabric Sequences inherit reactive behaviour from Geometric Fabrics while finding global solutions. Specifically, after commands in the GFCS have been optimised, upon encountering previously unseen obstacles, local collision avoidance fabric terms will execute local avoidance. We randomly generate new obstacles, blocking the previously solved motion trajectory, and check whether the motion can instantaneously adapt to the new obstacles. Examples are illustrated in \cref{fig:block}, where the red obstacles, after optimisation, are added in the way of the manipulator. We observe that GFCS can consistently generate collision-free trajectories which reactively avoid new obstacles, while escaping local minima.

\begin{figure}[b]
\caption{ Geometric Fabrics (4 figs on the top) perform local avoidance and fail to reach the blue goal. GFCS (4 figs on the bottom) performs global optimisation and finds a non-local solution.}
\label{fig:jaco_arm}
\centering
\begin{subfigure}{0.24\textwidth}
    \frame{\includegraphics[width=\textwidth]{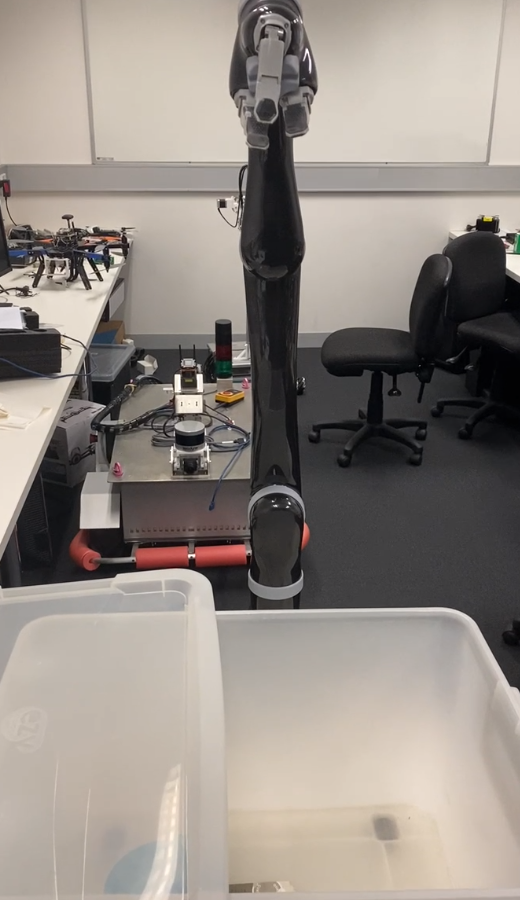}}
\end{subfigure}%
\begin{subfigure}{0.24\textwidth}
    \frame{\includegraphics[width=\textwidth]{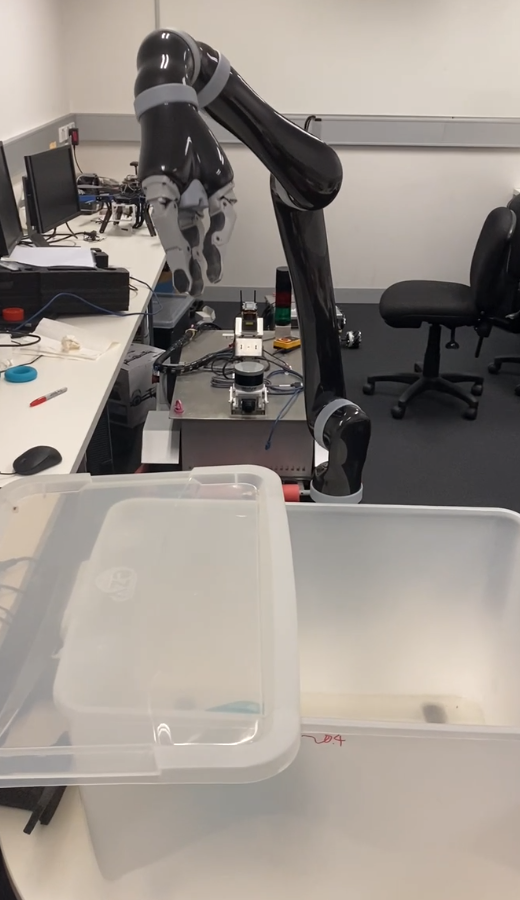}}
\end{subfigure}%
\begin{subfigure}{0.24\textwidth}
    \frame{\includegraphics[width=\textwidth]{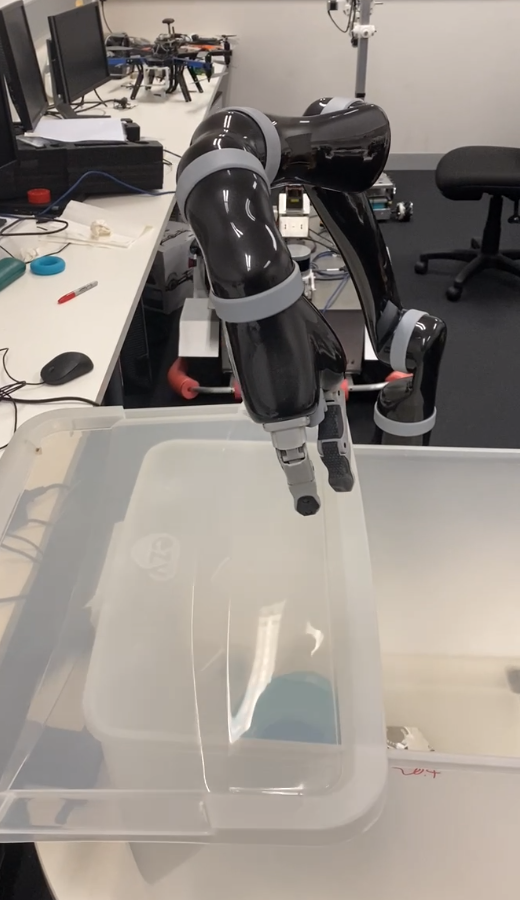}}
\end{subfigure}%
\begin{subfigure}{0.24\textwidth}
    \frame{\includegraphics[width=\textwidth]{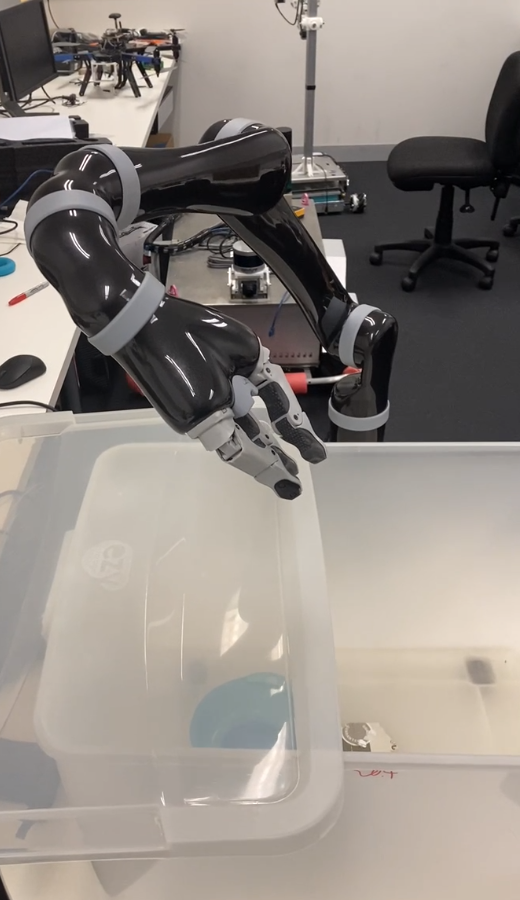}}
\end{subfigure}%

\begin{subfigure}{0.24\textwidth}
    \frame{\includegraphics[width=\textwidth]{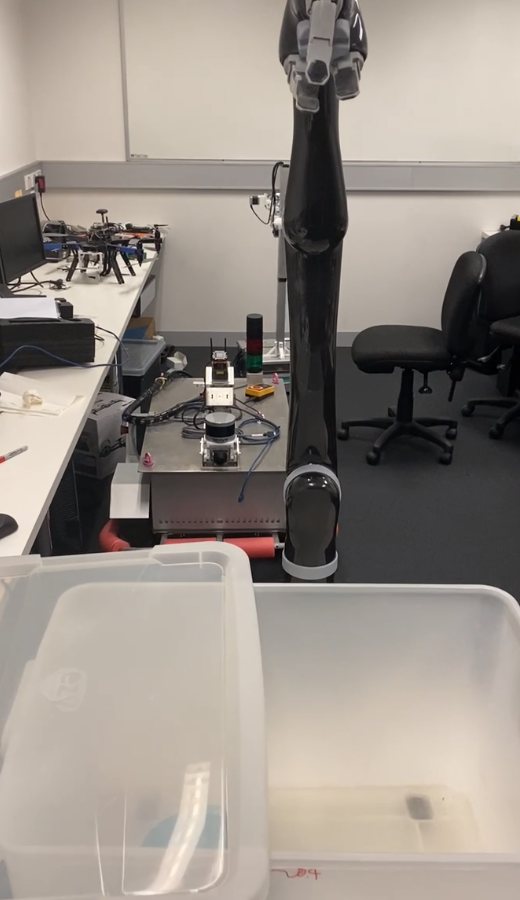}}
\end{subfigure}%
\begin{subfigure}{0.24\textwidth}
    \frame{\includegraphics[width=\textwidth]{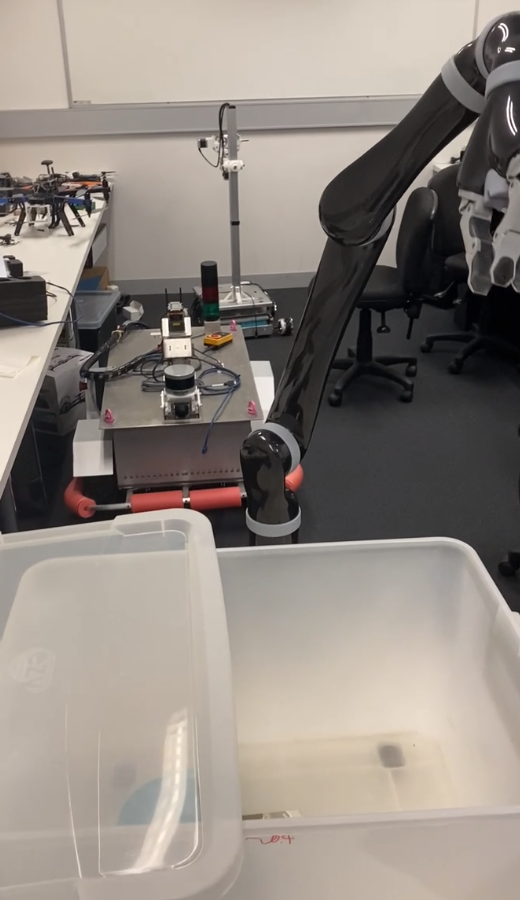}}
\end{subfigure}%
\begin{subfigure}{0.24\textwidth}
    \frame{\includegraphics[width=\textwidth]{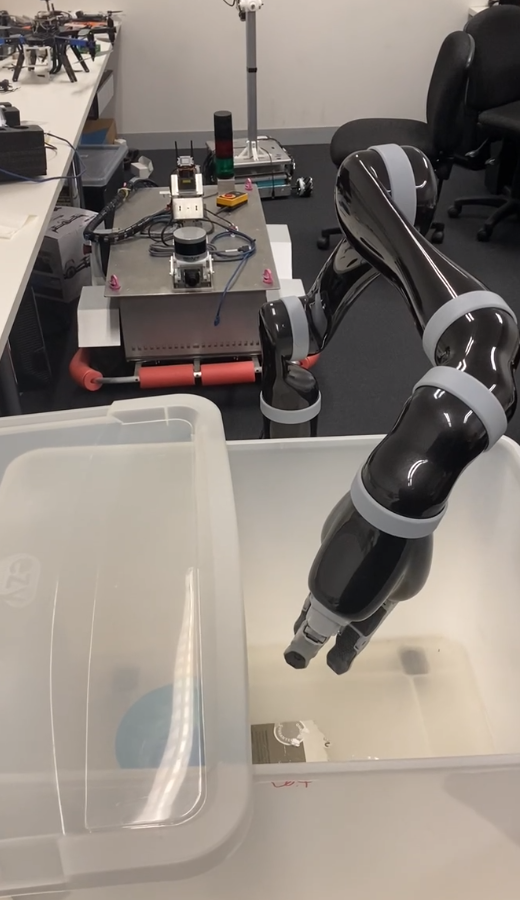}}
\end{subfigure}%
\begin{subfigure}{0.24\textwidth}
    \frame{\includegraphics[width=\textwidth]{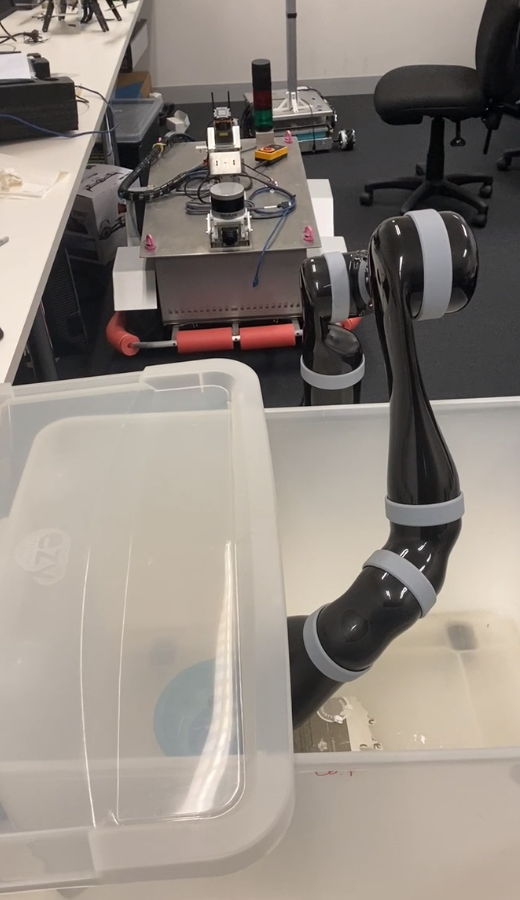}}
\end{subfigure}
\end{figure}
\FloatBarrier

\subsection{Execution on a Real Robot}
To evaluate the robustness of global motion found from GFCS, we generate motion on a real-world JACO manipulator to reach a goal in a half-covered box. Qualitative results from Geometric Fabrics and GFCS are shown in \cref{fig:jaco_arm}, where local avoidance by Geometric Fabrics is insufficient to reach the goal, and the manipulator stabilises on the cover. GFCS avoids the local minimum and reaches into the box towards the goal. A video illustrating GFCS controlling a manipulator to reactively avoid new obstacles in simulation, and execution on the real-world robot is available at \url{https://youtu.be/8LFFhriblLs}.

\section{Summary}
We present Geometric Fabric Command Sequences, a novel approach to generate global and reactive motion. This is achieved by running global optimisation over selected parameters of several sequentially joined command parameters for \emph{Geometric Fabrics}. The speed and performance of the optimisation can then be improved by transferring the knowledge of solving similar problems. We introduce a self-supervised framework, where an implicit generative model iterative learns to provide informed candidate solutions to the global optimisation procedure. We validate our approach on a range of problem classes, both in simulation and on a real-world JACO manipulator.

In the following \cref{chap9}, we shall summarise the contributions of this thesis, explore their connecting themes, and discuss potential future directions.
\pagebreak

\bookmarksetup{startatroot}
\chapter{Conclusions and Future Work}\label{chap9}
\section{Summary of Contributions}
This thesis presents methods for robots to learn from experience to understand, extrapolate and make decisions in unstructured environments. Contemporary robots often struggle to resiliently handle the endless possible permutations that can arise in the real-world, as doing so would require a prohibitively large set of hand-crafted rules to be specified. Learning approaches enable robots to move away from hand-designed rules and to generalise from experience. Learning can be flexibly integrated to a variety different components of the robotics autonomy stack. Under the overarching motivation of enabling greater autonomy in unstructured and dynamic environments with robot learning, we divide the thesis into three main themes:
\begin{itemize}
    \item Learning for environment representation;
    \item Learning for anticipatory navigation;
    \item Learning for robot manipulator motion generation.
\end{itemize}
Within each of these themes, we contribute novel robot learning methods and evaluate them thoroughly, both in simulation and using world-real data or on real robots. While each of these themes deals with a different problem setting, there are underlying commonalities behind the techniques proposed, and each connects with discussions that arise from the previous. In this section, we shall summarise these contributions and discuss possible avenues for future research.

\subsection{Learning Representations of the Environment}
In \cref{part1}, we focus our efforts on developing methods that learn continuous representations of the environment. Traditionally, grid-based methods have been used to represent the environment -- these approaches divide the domain into fixed-resolution cells and compute attributes for each cell independently. Learning-based approaches instead learn a continuous mapping from the domain to the attributes of interest, and the representation is abstracted away as a concise set of parameters. These representations are a more natural way to represent the continuous real-world and provide benefits such as greater memory-efficiency and improved performance when data is sparse. Here, we describe the main contributions of each chapter under this theme.

\subsubsection{Fusion of Continuous Occupancy Maps}
In \cref{chap3}, we study the problem of continuously representing the occupancy, i.e. the probability of a location being occupied. Specifically, we explore the application of continuous representation in multi-agent problem setups.

We introduce Fast Bayesian Hilbert Maps (Fast-BHM), a reformulation of Bayesian Hilbert Maps (BHM) \citep{Senanayke:2017}. We observe that BHM are prohibitively slow to build. To speed-up map building, we make the mean-field assumption when estimating the covariance matrix and re-derive equations for updating the map model. We demonstrate that imposition of the mean-field covariance assumption provides significant efficiency gains, with little degradation of performance.  

More importantly, we contribute a method to iteratively combine multiple Fast-BHM models. We demonstrate that Fast-BHMs are more memory-efficient than grid-based approaches, owing to the number of parameters of Fast-BHMs being much smaller than number of grid-cells required to sufficiently represent the environment. This property makes Fast-BHMs particularly attractive for multi-agent systems where the bandwidth for map transmission is restricted.

\subsubsection{Continuous Spatiotemporal Maps of Motion Directions}
In \cref{chap4}, we study the representation of continuous environments beyond the static setting, into the dynamic setting. In particular, we aim to build models capable of capturing the distribution of movement directions in the environment over time. These models endow robots with the ability to reason about how dynamic objects in the environment generally move, and may inform downstream planning tasks.

We contribute continuous spatiotemporal maps of motion directions, a model that learns a mapping from the space-time domain to a distribution of motion directions. Our model is capable of (1) handling the directional distributions which are multi-model, and (2) enforcing the support of the distribution to be between $[-\pi, \pi]$. We empirically demonstrate that the continuous nature of our model delivers performance benefits over discretised representations, particularly when the data is sparse. This is due to discretised representations assuming that each cell is independent from one another, and completely disregard information from neighbouring cells.

\subsection{Anticipatory Navigation with Learning}
Following the discussions presented in \cref{chap4}, where we explored continuous representations for dynamic environments, in \cref{part2}, we now look at imbuing a ground robot with the ability to navigate smoothly through dynamic environments. This is a setting where we require the robot to learn to continuously predict the dynamics of moving agents in the surroundings. To smoothly negotiate a path through the crowd, we develop methods to probabilistically \emph{anticipate} the movement of others in the vicinity, and then incorporate the predictions of motions into the decision-making process. 

\subsubsection{Stochastic Process Anticipatory Navigation}
In \cref{chap5}, we contribute the Stochastic Process Anticipatory Navigation (SPAN) framework. SPAN learns the motion patterns of other moving agents in the surroundings, and models the agents' movements as stochastic processes -- distributions over continuous functions mapping time from position. The outputs of the probabilistic predictive models are then integrated into the chance-constraints of a time-to-collision (TTC) control problem, which is solved in a receding horizon manner. 

We empirically evaluate the performance of SPAN to navigate through simulated crowds and real-world pedestrian data. We demonstrate that for our problem setup of navigating through a dynamic crowd with a non-holonomic ground robot, reactive methods would often lead the robot into unrecoverable locations in the crowd and become stuck. On the other hand, SPAN is able to smoothly maneuver through the crowd. Additionally, we observe the emergent behaviour of the robot ``hitch-hiking'' behind pedestrians moving in the same direction.  

\subsubsection{Trajectory Generation in New Environments from Past Experiences}
In \cref{chap5b}, we propose a framework, called Occupancy-Conditional Trajectory Network (OTNet), to condition on the structure of an environment to generate motion trajectories which are likely to appear in the environment. Humans have a great intuition to how agents in an environment would move, by observing a floor plan of the environment. We seek to empower robots to do the same, and have the ability to deduce motion patterns from the environment structure. OTNet encodes maps of environments by considering their Hausdorff distances from a set of representative maps to obtain fixed length encoding vectors. These are then inputted into a neural network to predict parameters of mixtures of trajectory distributions, which represent the motion trajectories we are expected to observe in the environment. We demonstrate that we can additionally enforce the start position of the distribution of trajectories. We evaluate on a simulated dataset containing environments and trajectories, and additionally qualitatively evaluate the quality of generating trajectories on a real world environment dataset, by generalising experience on the simulated dataset.

\subsubsection{Structurally Constrained Motion Prediction}
In \cref{chap6}, we improve the quality of motion prediction and develop methods capable of modelling multi-modal distributions of trajectories, and can incorporate prior knowledge of the structure of the surrounding environment as constraints. We know \emph{a priori} that agents in the environment are unlikely to move into occupied regions. Therefore, provided an occupancy representation of surroundings, our method allows us to impose chance constraints such that the probability of a moving agent entering occupied space is bounded. We leverage the availability of gradients from continuous representations of the environments, similar to the models discussed in \cref{part1}, and apply gradient-based constrained optimisers. We demonstrate, on both simulated and real-world data, that our method encourages probabilistic predictions to be more compliant with the environment structure.

\subsection{Learning to Generate Robot Manipulator Motion}
In \cref{part3}, we study how to generate desired motion trajectories for manipulators in unstructured environments. As discussed in \cref{part2}, anticipation are particularly helpful in enabling ground robots to avoid ``freezing'' in a crowd, while in robot manipulation, we rarely encounter a ``crowd'' of dynamic objects in the manipulator's vicinity. Additionally, manipulators typically possess higher degree-of-freedoms, and are not subject to non-holonomic constraints. We instead opt for handling dynamic obstacles in the environment with reactive approaches which compute instantaneous control solutions by considering the current states only. Here, we use learning to enable robot manipulators to imitate and generalise human demonstrations (\cref{chap7}), and to find more globally optimal solutions (\cref{chap8}).


\subsubsection{Diffeomorphic Templates for Generalised Imitation Learning}
In \cref{chap7}, we tackle the \emph{generalised imitation learning} problem. It is often challenging to accurately design or specify the robot's behaviour when faced with new tasks. Instead, imitation learning, also known as learning from demonstration, can be applied for the robot to learn the behaviour from a fairly small set of human expert demonstrations. Beyond simply reproducing the demonstrated motions, we additionally wish to enable robots to generalise the learned skills, under changes in the problem setup or environment. These changes can include additional obstacles in the environment, or newly specified biases provided by the user. 

We introduce the \emph{Diffeomorphic Templates} (DT) framework for generalised imitation learning. Under this framework, robot motions are represented as integrals of asymptotically stable dynamical systems. DTs are defined as differentiable and invertible functions (diffeomorphisms) which encode a specific behaviour. We can learn DTs to imitate demonstrations, or craft DTs to produce generalisation behaviour. These DTs can encode various behaviours, such as avoiding new obstacle, or to accounting for additional instructions. We devise an approach to compose multiple DTs in the configuration space of the robot, producing a combined system while remaining asymptotically stable. We demonstrate the ability of the Diffeomorphic Templates framework to learn and generalise novel skills in both simulation and on a real-world manipulator.

\subsubsection{Global and Reactive Motion Generation with Geometric Fabric Command Sequences}
In \cref{chap8}, we introduce \emph{Geometric Fabric Command Sequences} (GFCS) -- a global and reactive method to produce safe motion between start and goal configurations. GFCS builds upon Geometric Fabrics \citep{geoFabs}, a reactive motion generation approach capable of producing smooth and legible trajectories. Purely reactive approaches, such as Geometric Fabrics, produce actions which only take into account the immediate state, often leading to undesirable solutions that are infeasible and trapped in a local minimum. To combat this drawback, GFCS casts motion generation as a global optimisation problem over a concise set of parameters of a sequence of Geometric Fabrics, which we call commands. We then solve the optimisation problem via a black-box optimiser.

Additionally, we contribute a self-supervised learning framework to learn to warm-start the optimiser. Global optimisers can be greatly sped up by providing informative initial batches of solutions. We hypothesise that the experience of finding solutions in one environment may be transferable to alternative environments. To this end, we train an implicit generative model, conditional on the environment and problem setup, to generate informative initial solutions. A key insight is that the training of the generative model and the solving of the optimisation problem are complementary. That is, as we continuously solve to obtain more solution, we have more training data to refine the generative model, which in turn boost the efficiency and quality of the optimisation. Therefore, we incorporate both the learning and optimisation into a single loop, and incrementally solve-and-train in a self-supervised manner.

\section{Future Directions}
In this section, we shall outline specific future lines of enquiry based on the methods and results presented in this thesis.
\subsection{Time-to-Collision for Collaborative Manipulation}
In \cref{chap5}, we have demonstrated that Time-to-Collision (TTC) is effective as a component of a control problem where the robot is closely interacting with moving obstacles in the environment. We posit that TTC can also be effective to generate interactive behaviour for robot manipulators. Methods for approximating TTC can be computed very efficiently, for example in \cite{NHTTC}, by assuming robot controls are held and agents move at constant velocities. This potentially allows for the approximate TTC to be included in settings where the manipulator needs to remain fast and reactive, such as those in \cref{chap8}. Additionally, instead of simply trying to maximise the expected TTC to reduce the risk of being in-collision, we envision TTC to be constrained within a range to facilitate collaborative behaviour. For example, to enable the robot to stay close to humans in the environment, while also ensuring that collisions do not occur.

\subsection{Differentiable SPAN for Learning to Navigate}
The focus of \cref{part2} has been to develop probabilistic predictive models to model the motion of other agents in the environment and to use these predictions in a downstream control problem. Here observe that how the robot behaves, conditioned on the predictions, is specified via a cost and not learned. However, it is improbable that a constructed control cost is able to capture all the nuances of navigating through crowds. A future line of work can focus on injecting learning directly into the decision-making process, while still making use of the strong inductive biases provided by the SPAN model. This can potentially be achieved by developing a differentiable Stochastic Process Anticipatory Navigation (SPAN) model, allowing the control problem to be differentiated end-to-end. This shall allow imitation learning or reinforcement learning methods to be applied to SPAN. 

\subsection{Environment Generation for Active Self-supervised Learning}
The self-supervised learning framework to learn to warm-start optimisation, presented in \cref{chap8}, brings optimisation for motion generation and learning a generative into a unified loop. A great opportunity for future work is to also incorporate an environment generator into the loop. The environment generator will then actively generate environments such that they provide the greatest improvement to the training of the generative model, while being feasible to solve. The environment generator may take the form of a generative model over continuous occupancy maps models, such as Fast-BHMs (\cref{chap3}). Continuous occupancy representation is concise, requiring a relatively small number of parameters, and thus reduces the dimension of the data distribution for the generative model.

\pagebreak

\addcontentsline{toc}{chapter}{Bibliography}
\bibliography{bib.bib}
\bibliographystyle{abbrvnat}

\pagebreak

\end{document}